\documentclass{book}
\usepackage{fullpage}

\newcommand{\textdir}{../..}

\newcommand{\treplay}{\text{replay}}
\newcommand{\tnew}{\text{new}}
\newcommand{\topt}{\text{opt}}
\newcommand{\told}{\text{old}}
\newcommand{\wnew}{\vw_{\tnew}}

\newcommand{\myeat}[1]{}

\usepackage[usenames,dvipsnames]{xcolor}
\usepackage{longtable}

\usepackage[toc,page]{appendix}

\usepackage{bm}
\usepackage{frcursive} 
\usepackage{xfrac}
\usepackage{mathrsfs}

\usepackage{fancyvrb} 
\usepackage{xspace}
\usepackage{amsmath}
\usepackage{amssymb}
\usepackage{amsthm}

\usepackage[mathscr]{euscript}

\mathchardef\mhyphen="2D 
\mathchardef\mdash="2D 

\usepackage{accents}

\usepackage[font={small,it}]{caption}
\usepackage{subcaption} 

\usepackage{url}
\usepackage{cancel} 
\usepackage{multirow}
\usepackage{multicol}

\usepackage{verbatim}

\usepackage[T1]{fontenc}    

\usepackage[utf8]{inputenc}
\DeclareUnicodeCharacter{00B4}{'}
\DeclareUnicodeCharacter{03BB}{\lambda}
\usepackage[T1]{fontenc}
\usepackage{mathalpha}

\usepackage{url}            
\usepackage{booktabs}       
\usepackage{amsfonts}       
\usepackage{nicefrac}       
\usepackage{microtype}      

\usepackage[
  style=alphabetic,
  citestyle=alphabetic,
  natbib=true,
  backend=bibtex,
  maxcitenames=3,  mincitenames=1,
  maxbibnames=10,  minbibnames=1,
  firstinits=true,
  backref=false, 
  hyperref=true,
  doi=false,  isbn=false,  url=true,
  arxiv=abs
]{biblatex}

\AtBeginBibliography{%
  \iffieldundef{url}{}{%
    \printfield{url}%
  }%
}

\DefineBibliographyStrings{english}{%
  backrefpage = {page},
  backrefpages = {pages},
}

\usepackage{xpatch}

\usepackage{booktabs}
\usepackage{comment}

\usepackage{paralist} 
\usepackage{tabularx}

\usepackage{tikz}
\usetikzlibrary{fit, arrows, arrows.meta, calc, positioning, shadows.blur}
\usetikzlibrary{matrix, positioning, arrows.meta, shapes, trees, shapes.geometric}

\usepackage{xr}

\usepackage{mathtools}

\usepackage{enumitem}

\usepackage[ruled,vlined,linesnumbered]{algorithm2e}

\newcommand{\dontprintsemicolon}{}

\usepackage{mdframed}
\usepackage{multicol}
\usepackage{rotating}
\usepackage{setspace}

\usepackage{blkarray}

\usepackage{imakeidx}
\usepackage[columns=3]{idxlayout}

\usepackage[bookmarksnumbered]{hyperref}

\usepackage[capitalize]{cleveref}
\Crefname{equation}{Equation}{Equations}
\crefname{equation}{Equation}{Equations}
\Crefname{section}{Section}{Sections}
\crefname{section}{Section}{Sections}
\Crefname{chapter}{Chapter}{Chapters}
\crefname{chapter}{Chapter}{Chapters}
\Crefname{figure}{Figure}{Figures}
\crefname{figure}{Figure}{Figures}
\Crefname{appsec}{Appendix}{Appendices}
\crefname{appsec}{Appendix}{Appendices}

\definecolor{mycyan}{cmyk}{1,0,0,0.4}
\definecolor{mymagenta}{cmyk}{0,1,0,0.15}

\hypersetup{
    colorlinks,
    citecolor=mymagenta,
    linkcolor=mymagenta,
    urlcolor=mycyan,
}

\usepackage{eqnalign}
\usepackage{version}

\usepackage{etoolbox}

\usepackage{listings}
\newcommand{\hier}{\text{hier}}
\newcommand{\RTG}{\text{RTG}}
\newcommand{\BR}{\text{BR}}
\newcommand{\Regret}{\text{Regret}}

\newcommand{\fmix}{f_{\text{mix}}}
\newcommand{\mcts}{\text{MCTS}}
\newcommand{\tmin}{\text{min}}
\newcommand{\tmax}{\text{max}}

\makeatletter
\newcommand*\rel@kern[1]{\kern#1\dimexpr\macc@kerna}
\newcommand*\widebar[1]{%
	\begingroup
	\def\mathaccent##1##2{%
		\rel@kern{0.8}%
		\overline{\rel@kern{-0.8}\macc@nucleus\rel@kern{0.2}}%
		\rel@kern{-0.2}%
	}%
	\macc@depth\@ne
	\let\math@bgroup\@empty \let\math@egroup\macc@set@skewchar
	\mathsurround\z@ \frozen@everymath{\mathgroup\macc@group\relax}%
	\macc@set@skewchar\relax
	\let\mathaccentV\macc@nested@a
	\macc@nested@a\relax111{#1}%
	\endgroup
}
\makeatother

\newcommand{\ob}{o}
\newcommand{\ac}{a}
\newcommand{\ach}{\hat{\ac}}
\newcommand{\Ac}{\calA}
\newcommand{\rew}{r}
\newcommand{\rewh}{\hat{\rew}}
\newcommand{\Rew}{\mathcal{R}}
\newcommand{\ci}{c}
\newcommand{\his}{h}
\newcommand{\batchsize}{B}
\newcommand{\dsc}{\gamma}
\newcommand{\pol}{\pi}

\newcommand{\Pol}{\Pi}
\newcommand{\st}{s}
\newcommand{\sth}{\hat{\st}}
\newcommand{\Stf}{\mathcal{T}}

\newcommand{\ul}[1]{\underline{#1}}

\newcommand{\Adv}{\text{Adv}}

\newcommand{\ppi}{p_{\pi}}
\newcommand{\tprior}{\text{prior}}
\newcommand{\optimality}{{\mathcal O}}

\newcommand{\opt}{\optimality}
\newcommand{\penv}{p_{\text{env}}}
\newcommand{\done}{\text{done}}
\newcommand{\stopgrad}{\text{sg}}

\def\1{\bm{1}}

\def\eps{{\epsilon}}

\newcommand*{\fullref}[1]{\hyperref[{#1}]{\cref{#1} (\nameref*{#1})}}

\newcommand{\eat}[1]{}
\newcommand{\new}{\text{new}}

\mathchardef\mhyphen="2D 
\mathchardef\mdash="2D 

\newcommand{\myvec}[1]{\mathbf{#1}}
\newcommand{\myvecsym}[1]{\boldsymbol{#1}}

\makeatletter
\newcommand{\oset}[3][-0.3ex]{%
  \mathrel{\mathop{#3}\limits^{
    \vbox to#1{\kern-4\ex@
    \hbox{$\scriptstyle#2$}\vss}}}}
\makeatother

\newcommand{\vzero}{\myvecsym{0}}

\newcommand{\vdelta}{\myvecsym{\delta}}

\newcommand{\vepsilon}{\myvecsym{\epsilon}}

\newcommand{\veta}{\myvecsym{\eta}}
\newcommand{\vgamma}{\myvecsym{\gamma}}

\newcommand{\vmu}{\myvecsym{\mu}}

\newcommand{\vphi}{\myvecsym{\phi}}

\newcommand{\vPhi}{\myvecsym{\Phi}}
\newcommand{\vpi}{\myvecsym{\pi}}

\newcommand{\vpsi}{\myvecsym{\psi}}
\newcommand{\vPsi}{\myvecsym{\Psi}}

\newcommand{\vtheta}{\myvecsym{\theta}}

\newcommand{\vsigma}{\myvecsym{\sigma}}
\newcommand{\vSigma}{\myvecsym{\Sigma}}

\newcommand{\vtau}{\myvecsym{\tau}}

\newcommand{\va}{\myvec{a}}
\newcommand{\vb}{\myvec{b}}

\newcommand{\vc}{\myvec{c}}

\newcommand{\ve}{\myvec{e}}

\newcommand{\vg}{\myvec{g}}
\newcommand{\vh}{\myvec{h}}

\newcommand{\vm}{\myvec{m}}

\newcommand{\vo}{\myvec{o}}
\newcommand{\vp}{\myvec{p}}
\newcommand{\vq}{\myvec{q}}
\newcommand{\vr}{\myvec{r}}
\newcommand{\vs}{\myvec{s}}

\newcommand{\vu}{\myvec{u}}
\newcommand{\vv}{\myvec{v}}
\newcommand{\vw}{\myvec{w}}

\newcommand{\vx}{\myvec{x}}

\newcommand{\vy}{\myvec{y}}

\newcommand{\vz}{\myvec{z}}

\newcommand{\vA}{\myvec{A}}
\newcommand{\vB}{\myvec{B}}

\newcommand{\vD}{\myvec{D}}

\newcommand{\vF}{\myvec{F}}

\newcommand{\vH}{\myvec{H}}
\newcommand{\vI}{\myvec{I}}

\newcommand{\vL}{\myvec{L}}
\newcommand{\vM}{\myvec{M}}

\newcommand{\vT}{\myvec{T}}

\newcommand{\vW}{\myvec{W}}
\newcommand{\vX}{\myvec{X}}

\newcommand{\vZ}{\myvec{Z}}

\DeclareMathAlphabet{\mathsfit}{\encodingdefault}{\sfdefault}{m}{sl}
\SetMathAlphabet{\mathsfit}{bold}{\encodingdefault}{\sfdefault}{bx}{n}

\def\sR{{\mathbb{R}}}

\newcommand{\mymathcal}[1]{\mathcal{#1}}

\newcommand{\calA}{\mymathcal{A}}
\newcommand{\calB}{\mymathcal{B}}
\newcommand{\calC}{\mymathcal{C}}
\newcommand{\calD}{{\mymathcal{D}}}

\newcommand{\calE}{\mymathcal{E}}

\newcommand{\calF}{\mymathcal{F}}
\newcommand{\calG}{\mymathcal{G}}
\newcommand{\calH}{\mymathcal{H}}

\newcommand{\calI}{\mymathcal{I}}
\newcommand{\calJ}{\mymathcal{J}}

\newcommand{\calL}{\mymathcal{L}}
\newcommand{\calM}{\mymathcal{M}}

\newcommand{\calO}{\mymathcal{O}}

\newcommand{\calR}{\mymathcal{R}}
\newcommand{\calS}{\mymathcal{S}}

\newcommand{\calT}{\mymathcal{T}}
\newcommand{\calU}{\mymathcal{U}}

\newcommand{\calX}{\mymathcal{X}}

\newcommand{\betadist}{\mathrm{Beta}}

\newcommand{\Ber}{\mathrm{Ber}}

\newcommand{\gauss}{\mathcal{N}}

\newcommand{\softmax}{\mathrm{softmax}}

\newcommand{\lr}{\eta}

\newcommand{\argmax}{\operatornamewithlimits{argmax}}
\newcommand{\argmin}{\operatornamewithlimits{argmin}}

\newcommand{\half}{\frac{1}{2}}

\newcommand{\defeq}{\triangleq}

\newcommand{\real}{\sR}

\newcommand{\trans}{{\mkern-1.5mu\mathsf{T}}}

\newcommand{\ra}{\rightarrow}

\newcommand{\myst}{\ \  \mathrm{s.t.} \ \ }

\newcommand{\old}{\mathrm{old}}

\newcommand{\params}{\vtheta}

\newcommand{\xdata}{\vx}

\newcommand{\pemp}{p_{\data}}

\DeclareMathOperator{\KL}{D_\mathbb{KL}}
\newcommand{\KLpq}[2]{D_{\mathbb{KL}}\left({#1} \mrel{\|} {#2}\right)}

\DeclareMathOperator{\entropy}{\mathbb{H}}

\newcommand{\crossentropy}{\entropy_{ce}}

\DeclareMathOperator{\fdiv}{D_f}
\newcommand{\fdivpq}[2]{\fdiv\left({#1}\middle\Vert{#2}\right)}

\newcommand{\sigmoid}{\sigma}

\newcommand{\relu}{\ensuremath{\mathrm{ReLU}}\xspace}

\newcommand{\ind}[1]{\mathbb{I}\left({#1}\right)}

\newcommand{\vFisher}{\vF}

\newcommand{\E}{\mathbb{E}} 
\newcommand{\expect}[1]{\mathbb{E}\left[{#1}\right]} 
\newcommand{\expectQ}[2]{\mathbb{E}_{{#2}}\left[ {#1} \right]} 
\newcommand{\Var}{\mathbb{V}}
\newcommand{\var}[1]{\mathbb{V}\left[ {#1}\right]}

\newcommand{\union}{\cup}
\newcommand{\intersect}{\cap}

\newcommand{\assign}{\leftarrow}

\newcommand{\ptran}{p_S}    
\newcommand{\preward}{p_R}    
\newcommand{\policy}{\pi}
\newcommand{\policydet}{\mu_{\vtheta}}
\newcommand{\behavior}{\pi_b}
\newcommand{\policyexp}{\policy_{\mathrm{exp}}}
\newcommand{\traj}{\vtau}
\newcommand{\return}{G}
\newcommand{\initdist}{p_0}

\newcommand{\tinit}{\text{init}}
\newcommand{\tpg}{\text{pol}}
\newcommand{\tpol}{\text{pol}}

\newcommand{\tenv}{\text{env}}
\newcommand{\timag}{\text{imag}}
\newcommand{\tmodel}{\text{model}}
\newcommand{\ttot}{\text{total}}
\newcommand{\Mtrue}{M_{\text{env}}}
\newcommand{\Mest}{\hat{M}}

\newcommand{\ttrace}{\text{trace}}
\newcommand{\tref}{\text{ref}}
\newcommand{\policyref}{\pi_{\tref}}

\newcommand{\statmeasure}{\rho}
\newcommand{\statmeasuredet}{\rho_{\policydet}}

\newcommand{\statdist}{p^{\gamma}}
\newcommand{\statdistpol}{\statdist_{\policy}}
\newcommand{\statdistpolk}{\statdist_{\policy_k}}
\newcommand{\statdistpolki}{\statdist_{\policy_{k-i}}}
\newcommand{\statdistpolref}{\statdist_{\policyref}}
\newcommand{\statdistpolapprox}{\statdist_{\polapprox}}
\newcommand{\statdistbehave}{\statdist_{\behavior}}

\newcommand{\statdistpolexp}{\statdist_{\policyexp}}

\newcommand{\Vapprox}{V_{\vw}}
\newcommand{\Qapprox}{Q_{\vw}}
\newcommand{\Vpol}{V_{\policy}}

\newcommand{\Vpolopt}{V_{\polopt}}
\newcommand{\Vopt}{V^{*}}
\newcommand{\Vdet}{V_{\policydet}}
\newcommand{\Qpol}{Q_{\policy}}
\newcommand{\Qpolapprox}{Q_{\polapprox}}
\newcommand{\Qopt}{Q^{*}}
\newcommand{\Qdet}{Q_{\policydet}}
\newcommand{\advantage}{\text{Adv}}
\newcommand{\Apol}{\advantage_{\policy}}

\newcommand{\Aapprox}{A_{\vw}}
\newcommand{\polopt}{\policy^{*}}
\newcommand{\polapprox}{\policy_{\vtheta}}

\newcommand{\regret}{l}    
\newcommand{\totalRegret}{L}    
\newcommand{\estR}{\hat{R}}    
\newcommand{\optR}{\tilde{R}}    
\newcommand{\bstate}{\vb}    
\newcommand{\polval}{J}    
\newcommand{\polvallin}{L}    
\newcommand{\polvalest}{\hat{\polval}}    
\newcommand{\polvalis}{\polvalest_{\mathrm{IS}}}    
\newcommand{\polvalpdis}{\polvalest_{\mathrm{PDIS}}}    
\newcommand{\polvalbeh}{\polval_b}    
\newcommand{\policytgt}{\policy}    
\newcommand{\policybeh}{\policy_b}    
\newcommand{\psiRatio}{\rho}    

\newcommand{\targetV}{y}
\newcommand{\TargetV}{\targetV}

\newcommand{\loss}{\calL}

\newcommand{\elbo}{\text{\L}}

\newcommand{\ielbo}{\tilde{\end{description}\elbo}}

\newcommand{\lossfn}{\ell} 

\newcommand{\history}{\vh}
\newcommand{\hist}{\history}
\newcommand{\data}{\calD}

\newcommand{\dataExp}{\calD_{\mathrm{exp}}}

\newcommand{\vfeatures}{\vphi}

\newcommand{\latent}{\hiddenNode}

\newcommand{\be}{\begin{equation}}
\newcommand{\ee}{\end{equation}}

\newcommand{\bea}{\begin{eqnarray}}
\newcommand{\eea}{\end{eqnarray}}
\newcommand{\beaa}{\begin{eqnarray*}}
\newcommand{\eeaa}{\end{eqnarray*}}
\newcommand{\ba}{\begin{align*}}
\newcommand{\ea}{\end{align*}}

\newcommand{\figgen}[1]{Generated by #1} 

\newcommand{\figbased}[1]{Adapted from #1}
\newcommand{\figtaken}[1]{From  #1}
\newcommand{\figthanks}[1]{Used with kind permission of #1}

\newcommand{\figfix}[1]{}

\newcommand{\addfig}[4] 
{
    \begin{figure}
    \centering
    \includegraphics[#1]{\figdir/#4}
    \caption{#2}
    \label{fig:#3}
    \end{figure}
}

\newcommand{\addtwofigs}[5]    
{
    \begin{figure}
    \centering
    \begin{subfigure}{0.45\textwidth}
      \centering
      \includegraphics[#1]{\figdir/#4}
      \caption{ }
     \end{subfigure}
~
    \begin{subfigure}{0.45\textwidth}
            \centering
            \includegraphics[#1]{\figdir/#5}
                  \caption{ }
    \end{subfigure}
    \caption{#2}
    \label{fig:#3}
    \end{figure}
}

\newcommand{\addthreefigs}[6]    
{
    \begin{figure}
    \centering
    \begin{subfigure}{0.3\textwidth}
      \centering
      \includegraphics[#1]{\figdir/#4}
      \caption{ }
     \end{subfigure}
~
    \begin{subfigure}{0.3\textwidth}
            \centering
            \includegraphics[#1]{\figdir/#5}
                  \caption{ }
  \end{subfigure}
~
    \begin{subfigure}{0.3\textwidth}
            \centering
            \includegraphics[#1]{\figdir/#6}
                  \caption{ }
    \end{subfigure}
    \caption{#2}
    \label{fig:#3}
    \end{figure}
}

\newcommand{\addfourfigs}[7]    
{
    \begin{figure}
    \centering
    \begin{subfigure}{0.45\textwidth}
      \centering
      \includegraphics[#1]{\figdir/#4}
      \caption{ }
     \end{subfigure}
~
    \begin{subfigure}{0.45\textwidth}
      \centering
      \includegraphics[#1]{\figdir/#5}
      \caption{ }
     \end{subfigure}
\\
    \begin{subfigure}{0.45\textwidth}
      \centering
      \includegraphics[#1]{\figdir/#6}
      \caption{ }
     \end{subfigure}
~
    \begin{subfigure}{0.45\textwidth}
      \centering
      \includegraphics[#1]{\figdir/#7}
      \caption{ }
     \end{subfigure}
    \caption{#2}
    \label{fig:#3}
    \end{figure}
}

\newcommand{\notebook}[1]{\href{https://probml.github.io/notebooks\##1.ipynb}{#1.ipynb}}

\newcommand{\twodigits}[1]{\ifnum #1 < 10 0\fi #1}

\eat{
\makeatletter
\newcommand{\chapterauthor}[1]{%
  {\parindent0pt\vspace*{-25pt}%
  \linespread{1.1}\large\scshape#1%
  \par\nobreak\vspace*{35pt}}
  \@afterheading%
}
\makeatother
}

\newcommand{\keywordIndex}[1]{#1\index{#1}}
\newcommand{\keyword}[1]{\keywordIndex{#1}}
\newcommand{\keywordSpecial}[2]{{\bf #1}\index{#2}} 
\newcommand{\keywordDef}[1]{{\bf #1}\index{#1|textbf}}

\mdfdefinestyle{fearns}{%
    linecolor=red,
    outerlinewidth=50pt,
    roundcorner=20pt,
    innertopmargin=20pt,
    innerbottommargin=20pt,
    innerrightmargin=20pt,
    innerleftmargin=20pt,
    backgroundcolor=yellow!50!white}

\mdfdefinestyle{coptcomment}{%
    linecolor=blue,
    innertopmargin=10pt,
    innerbottommargin=10pt,
    innerrightmargin=10pt,
    innerleftmargin=10pt,
    backgroundcolor=green!25!white}

\mdfdefinestyle{alemi}{%
    linecolor=blue,
    outerlinewidth=50pt,
    roundcorner=20pt,
    innertopmargin=20pt,
    innerbottommargin=20pt,
    innerrightmargin=20pt,
    innerleftmargin=20pt,
    backgroundcolor=blue!30!white}

\newcommand{\points}[1]{}

\newcommand{\mrel}[1]{\mathrel{#1}}

\def\vtheta{{\bm{\theta}}}
\def\va{{\bm{a}}}
\def\vb{{\bm{b}}}
\def\vc{{\bm{c}}}

\def\ve{{\bm{e}}}

\def\vg{{\bm{g}}}
\def\vh{{\bm{h}}}

\def\vm{{\bm{m}}}

\def\vo{{\bm{o}}}
\def\vp{{\bm{p}}}
\def\vq{{\bm{q}}}
\def\vr{{\bm{r}}}
\def\vs{{\bm{s}}}

\def\vu{{\bm{u}}}
\def\vv{{\bm{v}}}
\def\vw{{\bm{w}}}
\def\vx{{\bm{x}}}
\def\vy{{\bm{y}}}
\def\vz{{\bm{z}}}

\usepackage{shorttoc}

\newif\ifarxiv
 \arxivtrue 

\ifarxiv
\usepackage[frozencache,cachedir=minted-cache]{minted}
\setminted{fontsize=\footnotesize,baselinestretch=1}
\else
\usepackage[cachedir=minted-cache]{minted}
\setminted{fontsize=\footnotesize,baselinestretch=1}
\fi

\ifarxiv
  \usepackage{listings}
  \newenvironment{codeblock}
                 {\begin{lstlisting}}
                 {\end{lstlisting}}
\else
  \usepackage{minted}
  
  \fi

\begin{document}

\title{Reinforcement Learning: An Overview}
\author{Kevin P. Murphy}
\date{\today}
\maketitle

\shorttableofcontents{Brief Table of Contents}{1}
\newpage

\tableofcontents
\newpage








\chapter{Introduction}



\section{Sequential decision making}

\newcommand{\envstate}{w}
\newcommand{\envmodel}{M}
\newcommand{\embedding}{e}
\newcommand{\agentstate}{z}

\begin{figure}
\centering
\includegraphics[height=2in]{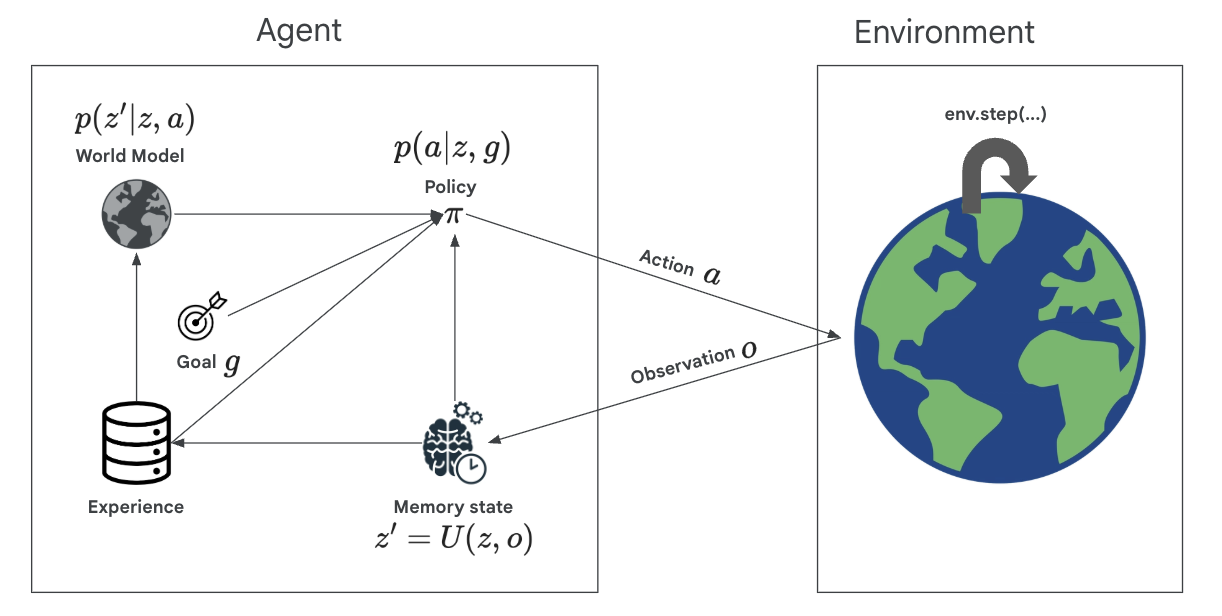}
\caption{
  A small agent interacting with a big external world.
  The observation $o_t$ (which, for notational simplicity,
  includes the previous action $a_t$)
  is used to update the
  internal agent state $z_t$, which is passed to the policy $\pi$
  which picks the next action $a_{t+1}$ based on the agent's goal $g_t$.
    Rewards are computed internally by the agent,
  by comparing $z_t$  with its internal goal $g_t$.
  The observations, actions and rewards are stored in
  a replay buffer, which can be used to learn the policy,
  a value function (not shown), and optionally an internal world model
  (for use in model-based RL, see \cref{sec:MBRL}).
  }
\label{fig:agentEnv}
\end{figure}

\keywordDef{Reinforcement learning}
or \keywordDef{RL} is a class of methods for solving
various kinds of sequential decision making tasks.
In such tasks, we want to design an  \keywordDef{agent}
that interacts with an external \keywordDef{environment}.
The agent maintains an internal state $\agentstate_t$,
which it passes to its \keywordDef{policy}
  $\pi$ to choose an action $a_{t}=\pi(\agentstate_t)$.
  The environment responds by sending back an observation $o_{t+1}$,
  which the agent uses to update its internal state using the state-update
  function $\agentstate_{t+1}=SU(\agentstate_t,a_{t},o_{t+1})$.
See \cref{fig:agentEnv} for an illustration.

To simplify things,
we often assume that the environment is also a Markovian
process, which has internal world state $\envstate_t$, from which the observations
$o_t$ are derived. (This is called a POMDP --- see \cref{sec:POMDP}).
We often simplify things even more by assuming that 
  the observation $o_t$
  reveals the hidden environment state;
  in this case, we denote the internal agent state and external environment
  state by the same letter, namely $s_t=o_t=\envstate_t=\agentstate_t$.
  (This is called an MDP --- see \cref{sec:MDP}).
 We discuss these assumptions in more detail in \cref{sec:universal}.

    RL is  more complicated
than supervised learning (e.g., training a classifier)
or self-supervised learning (e.g., training a language model),
because this framework is very general:
there are many assumptions we can make about the environment
and its observations $o_t$,
  and many choices we can make about the form the
  agent's internal state $\agentstate_t$
  and policy $\pi$, as well  the ways to update these objects
  as we see more data.
  We will study many different combinations in the rest of this document.
  The right choice ultimately depends on which real-world application you are
  interested in solving.\footnote{
For a list of real-world applications of RL, see e.g.,
\url{https://bit.ly/42V7dIJ}
from Csaba szepesvari (2024),
\url{https://bit.ly/3EMMYCW}
from Vitaly Kurin (2022),
and
\url{https://github.com/montrealrobotics/DeepRLInTheWorld},
which seems to be kept up to date.
}.

\subsection{Maximum expected utility principle}

The goal of the agent is to choose a policy $\pi$ so as to
maximize the sum of expected rewards:
\be
V_{\pi}(s_0) = \expectQ{\sum_{t=0}^T R(s_t,a_t) |s_0}{p(a_0,s_1,a_1,\ldots,a_T,s_T|s_0,\pi)}
\label{eqn:valueFn}
\ee
where $s_0$ is the agent's initial state,
$R(s_t,a_t)$ is the \keywordDef{reward function} that the agent
uses to measure the value of performing an action in a given state,
$V_{\pi}(s_0)$ is the \keywordDef{value function} for  policy $\pi$ evaluated at $s_0$,
and the expectation is wrt
\begin{align}
p(a_0,s_1,a_1,\ldots,a_T,s_T|s_0,\pi)
&= \pi(a_0|s_0) \penv(o_1|a_0) \delta(s_1=U(s_0,a_0,o_1))  \\
& \times \pi(a_1|s_1) \penv(o_2|a_1,o_1) \delta(s_2=U(s_1,a_1,o_2)) \\
& \times \pi(a_2|s_2) \penv(o_3|a_{1:2},o_{1:2}) \delta(s_3=U(s_2,a_2,o_3)) 
\ldots
\label{eqn:distrib}
\end{align}
where $\penv$ is the environment's distribution over observations 
(which  is usually  unknown).

We define the optimal policy as
\be
\pi^* = \arg\max_{\pi} \expectQ{V_{\pi}(s_0)}{p_0(s_0)}
\ee
Note that picking a policy to maximize
the sum of expected rewards is an instance
of the  \keywordDef{maximum expected utility} principle.
(In \cref{sec:regret}, we discuss the closely related concept
of choosing a policy which minimizes
the \keywordDef{regret},
which can be thought of as the difference between
the expected reward of the agent's policy compared
to a reference policy.)
There are  various ways to design or learn such an optimal policy,
depending on the assumptions we make about the environment,
and the form of the agent. We will discuss some of these options below.

\subsection{Episodic vs continual tasks}

If the agent can potentially interact with the environment forever,
we call it  a \keywordDef{continual task} \citep{Naik2021}.
In this case, we replace the sum of rewards (when defining the value
function) with the \keywordDef{average reward} \citep{Wan2021}.

Alternatively, we say the agent is in an
\keywordDef{episodic task}
if its interaction terminates once the system enters a
\keywordDef{terminal state} or \keywordDef{absorbing state},
which is a state which transitions to itself with 0 reward.
After entering a terminal state,
we may start a new \keywordDef{episode}
from a new initial world state $z_0 \sim \initdist$.
(The agent will typically also reinitialize its own internal state $s_0$.)
The episode length is in general random.
(For example, the length of an interaction with a chatbot
may be quite variable, depending on the decisions taken by the chatbot agent
and the randomness in the environment
(i.e., the responses from the user)).
Finally, if the trajectory length $T$
in an episodic task is fixed and known,
it is called a \keywordDef{finite horizon problem}.

We define the \keywordDef{return}  for a state at time $t$
to be the sum of expected rewards obtained
going forwards, where each reward is multiplied
by a \keywordDef{discount factor} $\gamma \in [0,1]$:
\begin{align}
\return_t &\defeq r_t + \gamma r_{t+1} + \gamma^2 r_{t+2} + \cdots
+ \gamma^{T-t-1} r_{T-1}
\label{eqn:rl-return} \\
&= \sum_{k=0}^{T-t-1} \gamma^k r_{t+k}
= \sum_{j=t}^{T-1} \gamma^{j-t} r_j
\end{align}
where $r_t=R(s_t,a_t)$ is the reward,
and
$\return_t$ is  the \keywordDef{reward-to-go}.
For episodic tasks that terminate at time $T$,
we define $\return_t = 0$ for $t \ge T$.
Clearly, the return satisfies the following
recursive relationship:
\begin{align}
\return_t 
= r_{t} + \gamma (r_{t+1} + \gamma r_{t+2} + \cdots) = r_{t} + \gamma \return_{t+1}
\end{align}
Furthermore, we define the value function to be the expected
reward-to-go:
\be
V_{\pi}(s_t) = \expect{\return_t | \pi}
\ee

The discount factor $\gamma$ plays two roles.
First, it ensures the return is finite
even if  $T=\infty$ (i.e., infinite horizon),
provided we use $\gamma<1$ and the rewards $r_t$ are bounded.
Second, it puts more weight on short-term rewards,
which generally has the effect of encouraging the agent
to achieve its goals more quickly.
(For example, if $\gamma=0.99$, then an agent that reaches a terminal reward
of 1.0 in 15 steps will receive
an expected discounted reward of
$0.99^{15}=0.86$, whereas if it takes 17 steps it will only
get $0.99^{17}=0.84$.)
However, if $\gamma$ is too small,
the agent will become too greedy.
In the extreme case where $\gamma=0$,
the agent is completely \keywordDef{myopic},
and only tries to maximize its immediate reward.
In general,  the discount factor reflects
the assumption that  there is a probability of $1-\gamma$ that the
interaction will end at the next step.
(If $\gamma=1-\frac{1}{T}$, the agent expects to live on the order
of $T$ steps; for example, if each step is 0.1 seconds,
then $\gamma=0.95$ corresponds to 2 seconds.)
For finite horizon problems, where $T$ is known,
we can set $\gamma=1$,
since we know the life time of the agent a priori.

\subsection{Universal model}
\label{sec:universal}

\begin{figure}
\centering
\includegraphics[height=3.5in]{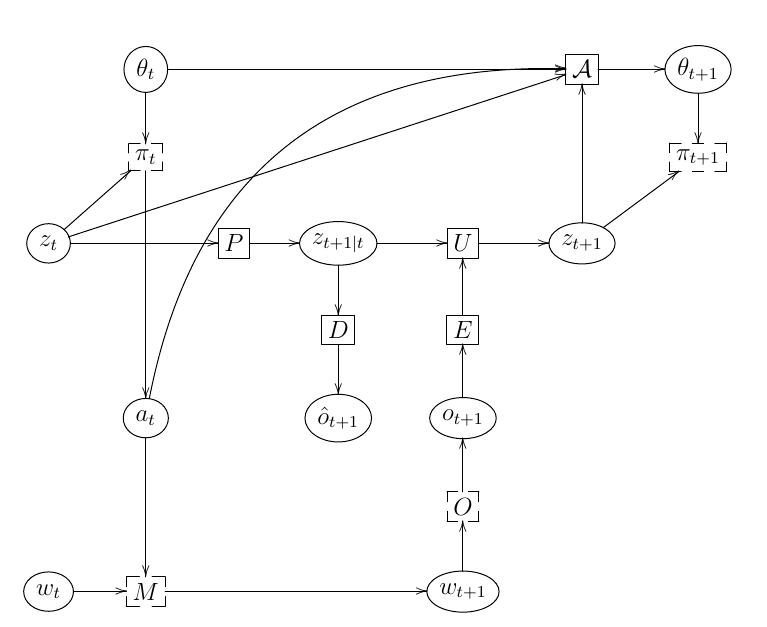}
\caption{
  Detailed illustration of the interaction of an agent in an environment.
  The agent has internal state $\agentstate_t$, and chooses action $a_t$
  based on its policy $\pi_t$ using $a_t \sim \pi_t(\agentstate_t|\theta_t)$.
  It then predicts its next internal states, $\agentstate_{t+1|t}$, via
  the predict function $P$,
  and optionally predicts the resulting observation,
  $\hat{o}_{t+1}$, via the observation decoder $D$.
  The environment has (hidden) internal state $\envstate_t$, which gets updated
  by the environment model $M$ to give the new state
  $\envstate_{t+1} \sim \envmodel(\envstate_t,a_t)$
  in response to the agent's action.
  The environment also emits an observation $o_{t+1}$ via
  its observation model,    $o_{t+1} \sim O(\envstate_{t+1})$.
  This gets encoded to $\embedding_{t+1}=E(o_{t+1})$
  by the agent's observation encoder $E$,
  which the agent uses to update its internal state using
  $\agentstate_{t+1}=U(\agentstate_t,a_t,\embedding_{t+1})$.
  The policy is parameterized by $\theta_t$,
  and these parameters may be updated (at a slower time scale) by an RL
  algorithm denoted by $\calA$.
  Square nodes are functions, circles are variables (either random or deterministic),
  dashed square nodes are stochastic functions that take an extra source of randomness
  (not shown).
  }
\label{fig:POMDP}
\end{figure}

A generic representation for sequential decision making problems
is shown in \cref{fig:POMDP}.
This  is an extended version of the ``universal modeling framework'' proposed in
\citep{Powell2019,Powell2022},
and is related to the ``common model of the intelligent decision maker''
discussed in \citep{Sutton2022}.
This common model  assumes the environment can be modeled by a
\keywordDef{controlled Markov process}\footnote{
The Markovian assumption is without loss of generality,
since we can always condition on the entire past sequence of states
by suitably expanding the Markovian state space.
} %
with hidden state $\envstate_t$, which gets updated at each step in response
to the agent's action $a_t$.
To allow for non-deterministic dynamics, we write this as 
 $\envstate_{t+1}=\envmodel(\envstate_t,a_t,\epsilon^\envstate_t)$,
where
$\envmodel$ is the environment's state transition function (which is usually not
known to the agent)
and
$\epsilon^\envstate_t$ is random system noise.\footnote{
Representing a stochastic function as a deterministic function with some noisy
inputs is known as a functional causal model,
or structural equation model.
This is standard practice in the control theory and causality communities.
},
The agent does not see the world state $\envstate_t$, but instead
sees a potentially noisy and/or partial observation
$o_{t+1}=O(\envstate_{t+1},\epsilon^o_{t+1})$ at each step,
where $\epsilon^o_{t+1}$ is random observation noise.
For example, when navigating a maze, the agent may only see what is in front of it,
rather than seeing everything in the world all at once;
furthermore, even the current view may be corrupted by sensor noise.
Any given image, such as one containing a door, could correspond to many
different locations in the world (this is called \keywordDef{perceptual aliasing}),
each of which may require a different action.

Thus the agent needs use  these observations to
main an internal \keywordDef{belief state}
about the world, denoted by $\agentstate$.
This gets updated
using the state update function
\be
\agentstate_{t+1}=SU(\agentstate_t,a_t,o_{t+1})
\ee
In the simplest setting, the internal $\agentstate_t$
can just store all the past observations,
$\vh_t=(\vo_{1:t},\va_{1:t-1})$, but such non-parametric models
can take a lot of time and space to work with, so we will usually
consider parametric approximations.
The agent can then pass its state to its policy to pick actions,
using $a_{t+1}=\pi_t(\agentstate_{t+1})$.

We can further elaborate the behavior of the agent by breaking the state-update
function into two parts.
First the agent predicts its own next state, $\agentstate_{t+1|t}=P(\agentstate_t,a_t)$,
using a \keywordDef{prediction function} $P$, and then it updates
this prediction given the observation using \keywordDef{update function} $U$,
to give $\agentstate_{t+1}=U(\agentstate_{t+1|t},o_{t+1})$.
Thus the $SU$ function is defined as the composition of the predict and update
functions
\be
\agentstate_{t+1} = U(P(\agentstate_t,a_t), o_{t+1})
\ee
If the observations are high dimensional (e.g., images),
the agent may choose to encode its observations into a low-dimensional
embedding $e_{t+1}$ using an encoder, $e_{t+1}=E(o_{t+1})$;
this can encourage the agent to focus on the relevant
parts of the sensory signal.
In this case, the state update becomes
\be
\agentstate_{t+1} = U(P(\agentstate_t,a_t), E(o_{t+1}))
\ee
Optionally the agent can also learn to invert this encoder by training
a decoder to predict the next observation using $\hat{o}_{t+1}=D(\agentstate_{t+1|t})$;
this can be a useful training signal, as we will discuss in \cref{sec:MBRL}.
Finally, the agent needs to learn the action policy 
$\pi_t(\agentstate_t)=\pi(\agentstate_t;\vtheta_t)$.
We can update the policy parameters using
a learning algorithm, denoted
\be
\vtheta_t = \calA(o_{1:t},a_{1:t},r_{1:t})
=\calA(\vtheta_{t-1},  a_t, \agentstate_t, r_t)
\ee
See \cref{fig:POMDP} for an illustration.

We see that, in general, there are three interacting stochastic processes
we need to deal with:
the environment's states $\envstate_t$ (which are usually affected
by the agents actions);
the  agent's internal states $\agentstate_t$
(which reflect its beliefs about the environment based on the observed data);
and the agent's policy parameters $\vtheta_t$
(which are updated based on the information stored in the belief state
and the external observations).

\eat{
The reason
there are so many RL algorithms is that this framework is very general.
In the rest of this manuscript we will study special cases,
where we make different assumptions
about the environment's state $\envstate_t$ and dynamics $\envmodel$,
the agent's state $\agentstate_t$ and dynamics $U$,
the form of the action policy $a_t=\pi(\agentstate_t,\vtheta_t)$,
and  the form of the policy update method $\calA$.
}

\subsection{Further reading}

In later chapters, we will describe methods for learning the best
policy to maximize $V_{\pi}(s_0)=\expect{G_0|s_0,\pi})$.
More details on RL can be found
in textbooks such as 
\citep{Suttonv2,KochenderferDecisions,Plaat2022,Li2023RL,Szepesvari10},
and reviews such as \citep{Arulkumaran2017,FrancoisLavet2018,Li2018RL,
  weng2018bandit,Ivanov2019,Jaeger2024}.
For a more theoretical treatment, see e.g., \citep{AgarwalBook,MannorBook,Foster2023}.
For details on how RL relates to \keywordDef{control theory},
see e.g., \citep{Sontag98,Recht2019,BertsekasRL,Meyn2022};
 for connections to operations research,
 see \citep{Powell2022};
 for connections to finance, see \citep{RaoRLBook}.


\section{Canonical models}

In this section, we describe different forms of model for the environment and the agent
that have been  studied in the literature.

\subsection{Partially observed MDPs}
\label{sec:POMDP}

The model shown in \cref{fig:POMDP} 
 is called
a \keywordDef{partially observable Markov decision process}
or \keywordDef{POMDP} (pronounced ``pom-dee-pee'')
\citep{Kaelbling98,Lauri2022,Subramanian2022}.
Typically the environment's dynamics model is represented by a stochastic transition function,
rather than a deterministic function with  noise as an input.
We can derive this transition function as follows:
\be
p(\envstate_{t+1}|\envstate_t,a_t) =
\expectQ{\ind{\envstate_{t+1}=W(\envstate_t, a_t, \epsilon^\envstate_t)}}{\epsilon^\envstate_t}
\ee
Similarly the stochastic observation function is given by
\be
p(o_{t+1}|\envstate_{t+1}) = \expectQ{\ind{o_{t+1}=O(\envstate_{t+1}, \epsilon^o_{t+1})}}{\epsilon^o_{t+1}}
\ee
Note that we can combine these two distributions to derive
the joint world model
$p_{WO}(\envstate_{t+1},o_{t+1}|\envstate_t,a_t)$.
Also, we can use these distributions to derive the environment's non-Markovian
observation distribution, $\penv(o_{t+1}|o_{1:t},a_{1:t})$,
used in \cref{eqn:distrib},
as follows:
\begin{align}
  \penv(o_{t+1}|o_{1:t},a_{1:t})
  &= \sum_{\envstate_{t+1}} p(o_{t+1}|\envstate_{t+1}) p(\envstate_{t+1}|a_{1:t}) \\
  p(\envstate_{t+1}|a_{1:t}) &= \sum_{\envstate_1} \cdots \sum_{\envstate_t}
  p(\envstate_1|a_1) p(\envstate_2|\envstate_1,a_1) \ldots p(\envstate_{t+1}|\envstate_t,a_t)
  \end{align}

If the world model (both $p(o|\envstate)$ and $p(\envstate'|\envstate,a)$) is known, then we can
--- in principle --- solve
for the optimal policy.
The method requires that the agent's internal state correspond to
the \keywordDef{belief state}
$s_t = \vb_t = p(\envstate_t | \vh_t)$,
where $\vh_t=(o_{1:t},a_{1:t-1})$ is the observation history.
The belief state  can be updated recursively using Bayes rule.
See \cref{sec:beliefStateMDP} for details.
The belief state 
 forms a  sufficient statistic for the optimal policy.
 Unfortunately, computing the belief state and
 the resulting optimal policy is wildly intractable
 \citep{Papadimitriou87,Kaelbling98}.
We discuss some approximate methods in \cref{sec:partialObs}.

\subsection{Markov  decision process (MDPs)}
\label{sec:MDP}

\begin{figure}
\centering
    \includegraphics[height=1.8in]{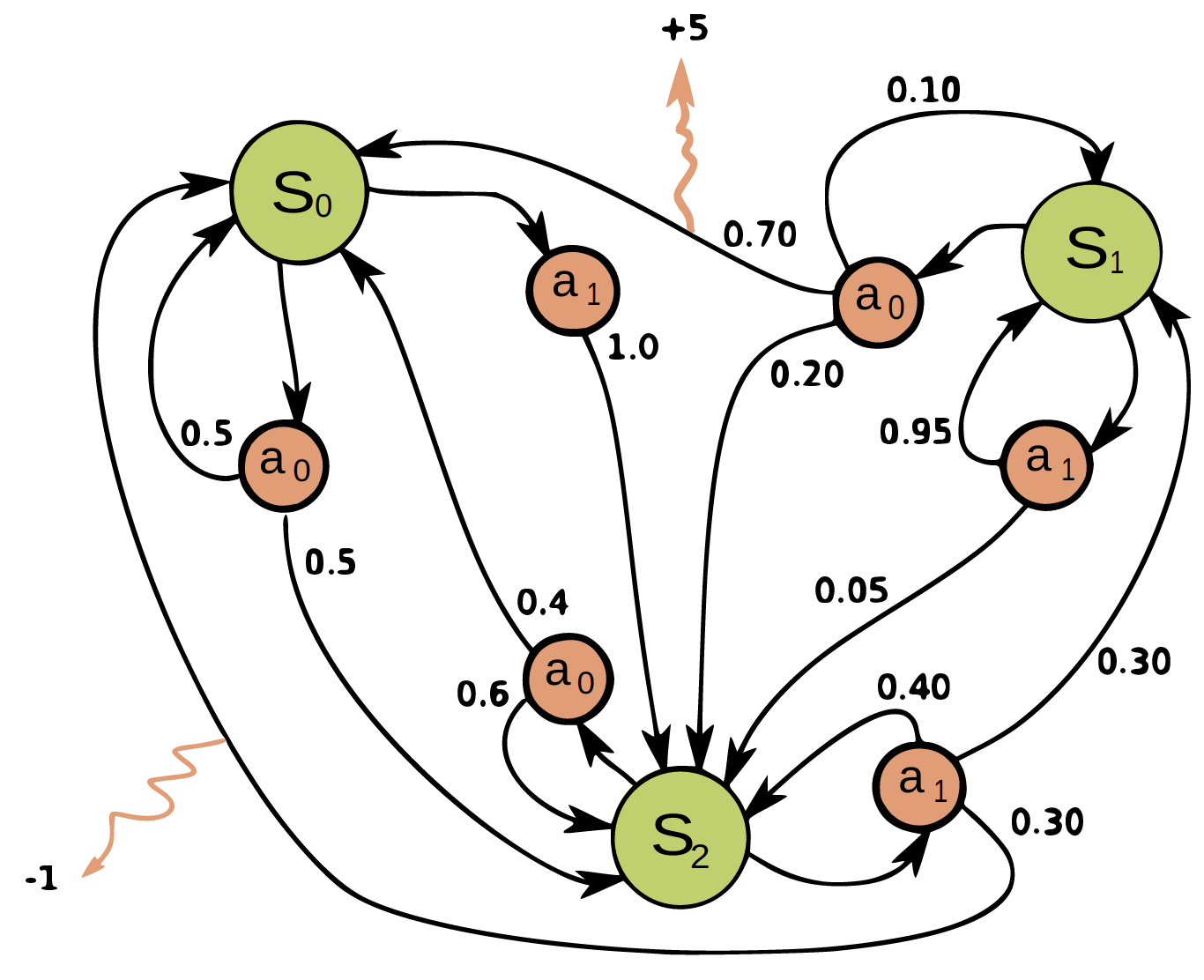}
\caption{
Illustration of an MDP as a finite state machine (FSM).
The MDP has three discrete states (green cirlces),
two discrete
actions (orange circles), and two non-zero rewards (orange arrows).
The numbers on the black
edges represent state transition
probabilities, e.g.,  $p(s'=s_0|a=a_0,s'=s_1)=0.7$;
most state transitions are impossible
(probability 0), so the graph is sparse.
The numbers on the yellow wiggly edges represent
expected rewards, e.g., $R(s=s_1, a=a_0, s'=s_0) = +5$;
state transitions   with zero reward are not annotated.
\figtaken{\url{https://en.wikipedia.org/wiki/Markov_decision_process}}.
\figthanks{Wikipedia author waldoalvarez}.
}
\label{fig:MDP}
\end{figure}

A \keywordDef{Markov decision process}~\citep{Puterman94}
is a special case of a POMDP in which the environment states are observed,
so $\envstate_t=o_t=s_t$.
We usually define an MDP in terms of the state transition matrix
induced by the world model:
\be
\ptran(s_{t+1}|s_t,a_t) = \expectQ{\ind{s_{t+1}=W(s_t,a_t,\epsilon^s_t)}}{\epsilon^s_t}
\ee
In lieu of an observation model, we assume the environment
(as opposed to the agent) sends out a reward signal,
sampled from $p_R(r_t|s_t,a_t,s_{t+1})$.
The expected reward is then given by
\begin{align}
  R(s_t,a_t,s_{t+1}) &= \sum_r r  \;\; \preward(r|s_t,a_t,s_{t+1}) \\
  R(s_t,a_t) &= \sum_{s_{t+1}} \ptran(s_{t+1}|s_t,a_t) R(s_t,a_t,s_{t+1})
\end{align}
Note that
the field of \keyword{control theory}
uses slightly different terminology and notation when describing
the same setup:
the environment is called the \keywordDef{plant},
the agent is called the \keywordDef{controller},
States are denoted by $\vx_t \in \calX \subseteq \real^D$,
actions are denoted by $\vu_t \in \calU \subseteq \real^K$,
and rewards are replaced by costs $c_t \in \real$.

Given a stochastic policy $\pi(a_t|s_t)$,
the agent can interact with the environment over many steps.
Each step is called a \keywordDef{transition},
and consists of the tuple $(s_t,a_t,r_t,s_{t+1})$,
where $a_t\sim \policy(\cdot|s_t)$,
$s_{t+1} \sim \ptran(s_t,a_t)$,
and $r_t \sim \preward(s_t,a_t,s_{t+1})$.
Hence, under policy $\policy$,
the probability of generating a \keywordDef{trajectory}  length $T$,
$\traj=(s_0,a_0,r_0,s_1,a_1,r_1,s_2,\ldots, s_T)$,
can be written explicitly as
\begin{align}
\label{eqn:rl-traj-prob}
p(\traj) = \initdist(s_0) \prod_{t=0}^{T-1} \policy(a_t|s_t) \ptran(s_{t+1}|s_t,a_t)
\preward(r_t|s_t,a_t,s_{t+1})
\end{align}

In general, the state and action sets of an MDP can be discrete or continuous.
When both sets are finite,
we can represent these functions as lookup tables;
this is known as a \keywordDef{tabular representation}.
In this case, we can represent the MDP as a
\keywordDef{finite state machine},
which is a graph where nodes correspond to states,
and edges correspond to actions and the resulting rewards and next states.
\cref{fig:MDP} gives a simple example
of an MDP with 3 states and 2 actions.

If we know the world model $\ptran$ and $\preward$,
and if the state and action space is tabular,
then we can solve for the optimal policy
using dynamic programming techniques,
as we discuss in  \cref{sec:planning}.
However, typically the world model is unknown,
and the states and actions may need 
complex nonlinear models to represent their transitions.
In such cases, we will have to use RL methods to learn a good policy.

\subsection{Goal-conditioned MDPs}
\label{sec:GCRL}

A \keywordDef{goal-conditioned MDP} is one
in which the reward is defined as $R(s,a|g)=1$
iff the goal state is achieved,
i.e., $R(s,a|s) = \ind{s=g}$.
We can also define a dense reward signal using
some state abstraction function $\phi$,
by definining $R(s,a|g)=\text{sim}(s, g)$,
where sim is some kind of similarity metric.
For example, if $s$ is an image and $g$
is a sentence, we may use cosine similarity
\be
\text{sim}(s, g) =
\frac{\phi(s)^\trans \psi(g)}{||\phi(s)|| \;  ||\psi(g)||}
\ee
where $\phi(s)$ is an embedding of the image (state).
and $\psi(g)$ is an embedding of the text (goal).
Such embeddings can be computed by
using a VLM or vision-language model
(see \cref{sec:LLMforReward}).

A goal-conditioned policy of the form
$\pi(a|s,g)$ is sometimes called a \keywordDef{universal policy}
\citep{Schaul2015}.
We can learn such policies
using \keywordDef{goal-conditioned RL} methods
(see e.g., \citep{Liu2022GCRL} and \cref{sec:HER}).

Note that multi-goal RL  is different to multi-task RL.
The latter refers to the ability to solve different ``tasks'',
which correspond to entire MDPs (with different dynamics
as well as different rewards).

\subsection{Contextual MDPs}

A \keywordDef{Contextual MDP} \citep{Hallak2015}
is an MDP where the 
dynamics and rewards of the environment depend on a
hidden static parameter referred to as the context.
(This is different to a contextual bandit, discussed in \cref{sec:bandits},
where the context is observed at each step.)
A simple example of a contextual MDP is a
video game, where each level of the game
is \keywordDef{procedurally generated},
that is, it is randomly generated each time
the agent starts a new episode.
Thus the agent must solve a sequence of
related MDPs, which are drawn from a common distribution.
This requires the agent to \keywordDef{generalize} across multiple MDPs,
rather than overfitting to a specific environment
\citep{Cobbe2019,Kirk2021,Tomilin2022}.
(This form of generalization is different
from generalization within an MDP,
which requires generalizing across states,
rather than across environments; both are important.)

A contextual MDP is a special kind of POMDP
where the hidden variable corresponds to the unknown
parameters of the model.
In \citep{Ghosh2021RL}, they call this an \keywordDef{epistemic POMDP},
which is closely related to the concept
of belief state MDP which we discuss in \cref{sec:beliefStateMDP}.

\subsection{Contextual bandits}
\label{sec:contextualBandit}
\label{sec:bandits}

A \keywordDef{contextual bandit} is a special case of a POMDP
where the world state transition function is independent of the
action of the agent and the previous state,
i.e., $p(\envstate_t|\envstate_{t-1},a_t) = p(\envstate_t)$.
In this case, we call the world states ``contexts'';
these are observable by the agent, i.e., $o_t=\envstate_t$.
Since the world state distribution is independent of the agents actions,
the agent has no effect on the external
environment.
However, its actions do affect  the rewards that it
receives. Thus the agent's internal belief state --- about the underlying
reward function $R(o,a)$ --- does change over time,
as the agent learns a model of the world
(see \cref{sec:beliefStateMDP}).

A special case of a contextual bandit is a regular bandit,
in which there is no context, or equivalently, $s_t$ is
some fixed constant that never changes.
When there are a finite number of possible actions,
$\calA=\{a_1,\ldots,a_K\}$,
this is called a \keywordDef{multi-armed bandit}.\footnote{
The terminology arises by analogy to a slot machine
(sometimes called a ``bandit'', because it steals your money)
in a casino.
If there are $K$ slot machines, each with different rewards
(payout rates), then the agent (player) must explore
the different machines (by pulling the arms)
until they have discovered
which one is best, and can then stick to exploiting it.
}
In this case the reward model  has the form $R(a) = f(\vw_a)$,
where $\vw_a$ are the parameters for arm $a$.

Contextual bandits have many applications.
For example, consider an \keywordDef{online advertising system}.
In this case, the state $s_t$ represents features
of the web page
that the user is currently looking at,
and the action $a_t$ represents the identity
of the  ad which the system
chooses to show.
Since the relevance of the ad depends on the page,
the reward function has the form $R(s_t,a_t)$,
and hence  the problem is contextual.
The goal is to maximize the expected reward,
which is equivalent to the expected number of times people click on
ads; this is known as
 the \keywordDef{click through rate} or \keywordDef{CTR}.
(See e.g.,  \citep{Graepel10,Li10linucb,McMahan13,Agarwal2014laser,Du2021ads,Yang2022CTR}
for more information about this application.)
Another application of contextual bandits
arises in \keywordDef{clinical trials}
\citep{Villar2015}.
In this case, the state $s_t$ are features of the current patient we are treating,
and the action $a_t$ is the treatment the doctor chooses to give them
(e.g., a new drug or a \keywordDef{placebo}).

For more details on bandits,
see e.g., \citep{Lattimore2019,Slivkins2019}.

\subsection{Belief state MDPs}
\label{sec:beliefStateMDP}

In this section, we describe a kind of MDP where the state
represents a probability distribution,
known as a \keywordDef{belief state} or \keywordDef{information state},
which is updated by the agent (``in its head'') as it receives
information from the environment.\footnote{
Technically speaking,  this is a POMDP, where we assume the states are observed,
and the parameters are the unknown hidden random variables.
This is in contrast to \cref{sec:POMDP},
where the states were not observed, and the parameters were assumed to be known.
} %
More precisely, consider a contextual bandit problem,
where the agent approximates the unknown reward by a function
$R(o,a)=f(o,a;\vw)$.
Let us denote the posterior over the unknown parameters
by $\bstate_t=p(\vw|\history_t)$,
where $\history_t = \{ o_{1:t}, a_{1:t}, r_{1:t} \}$
is the history of past observations, actions and rewards.
This belief state can be updated deterministically using Bayes' rule;
we denote this operation by
$\bstate_{t+1}=\mathrm{BayesRule}(\bstate_{t},o_{t+1},a_{t+1},r_{t+1})$.
(This corresponds to the state update $SU$ defined earlier.)
Using this, we can define the following
\keywordDef{belief state MDP},
with deterministic dynamics given by
\begin{align}
  p(\bstate_{t+1}|\bstate_{t},o_{t+1},a_{t+1},r_{t+1})
  &= \ind{\bstate_{t+1}=\mathrm{BayesRule}(\bstate_{t},o_{t+1},a_{t+1},r_{t+1})} 
\end{align}
and reward function given by
\begin{align}
p(r_t|o_t,a_t,\bstate_t) &= \int \preward(r_t|o_t,a_t;\vw) p(\vw|\bstate_t) d\vw
\end{align}
If we can solve this (PO)MDP, we have the optimal solution to the exploration-exploitation
problem
(see  \cref{sec:explorationExploitation}).

\begin{figure}
\centering
\includegraphics[height=2in]{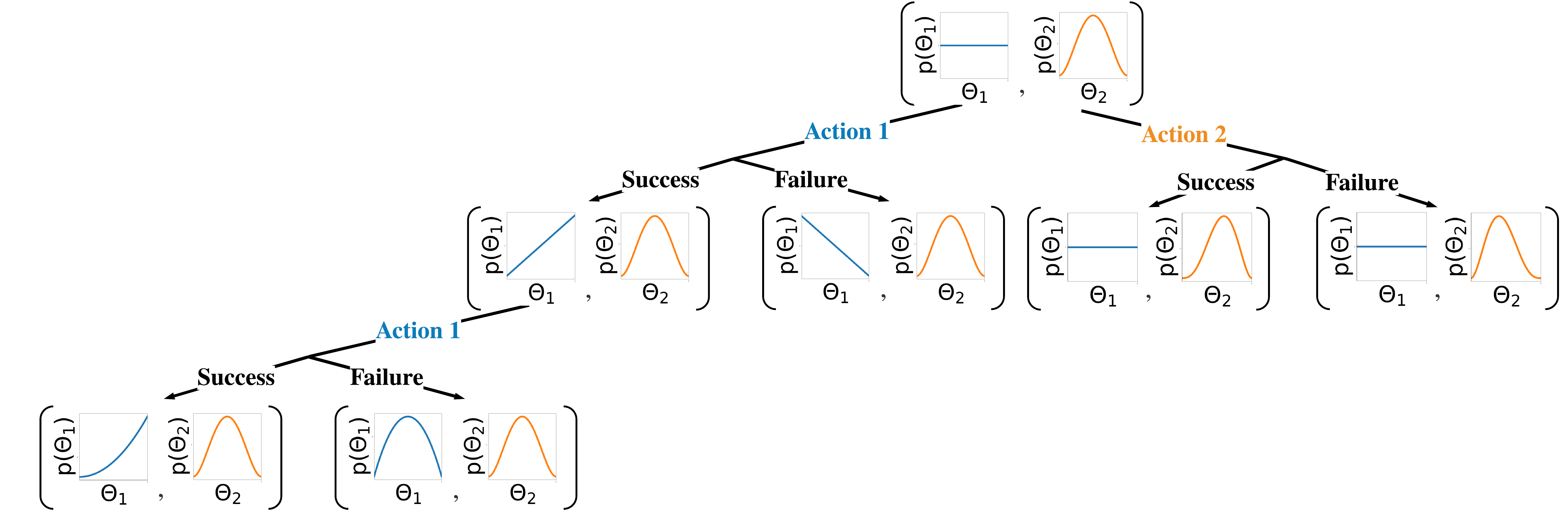}
\caption{
Illustration of sequential belief updating
for a two-armed beta-Bernoulli bandit.
The prior for the reward for action 1
is the (blue) uniform distribution $\betadist(1,1)$;
the prior for the reward for action 2
is the (orange) unimodal distribution $\betadist(2,2)$.
We update the parameters of the belief state based on the chosen action,
and based on whether the observed reward is success (1)
or failure (0).
}
\label{fig:bernoulliBanditBAMDP}
\end{figure}

As a simple example,
consider 
a context-free \keywordDef{Bernoulli bandit},
where $\preward(r|a) = \Ber(r|\mu_a)$,
and $\mu_a=\preward(r=1|a)=R(a)$ is the expected reward for taking
action $a$.
The only unknown parameters are $\vw=\mu_{1:A}$.
Suppose  we use  a factored beta prior
\be
p_0(\vw)=\prod_a \betadist(\mu_a|\alpha_0^a,\beta_0^a)
\ee
where $\vw=(\mu_1,\ldots,\mu_K)$.
We can compute the posterior in closed form
to get
\be
p(\vw|\data_t) =
\prod_a \betadist(\mu_a|\underbrace{\alpha_0^a + N_{t}^0(a)}_{\alpha_t^a},
\underbrace{\beta_0^a + N_{t}^1(a)}_{\beta_t^a})
\label{eqn:postBetaBandit}
\ee
where
\be
N_t^r(a) = \sum_{i=1}^{t-1} \ind{a_i=a, r_i=r}
\ee
This is illustrated in \cref{fig:bernoulliBanditBAMDP}
for a two-armed Bernoulli bandit.
We can use a similar method for a \keywordDef{Gaussian bandit},
where $\preward(r|a)=\gauss(r|\mu_a,\sigma_a^2)$.

In the case of contextual bandits, the problem is conceptually
the same, but becomes more
complicated computationally.
If we assume a \keywordDef{linear regression bandit},
$\preward(r|s,a;\vw) = \gauss(r|\vfeatures(s,a)^\trans \vw,\sigma^2)$,
we can use Bayesian linear regression to compute
$p(\vw|\data_t)$ exactly in closed form.
If we assume a \keywordDef{logistic regression bandit},
$\preward(r|s,a;\vw) = \Ber(r|\sigmoid(\vfeatures(s,a)^\trans \vw))$,
we have to use approximate methods for approximate Bayesian logistic regression to compute
$p(\vw|\data_t)$.
If we have a \keywordDef{neural bandit}
of the form
$\preward(r|s,a;\vw) = \gauss(r|f(s,a; \vw))$
for some nonlinear function $f$,
then  posterior inference is even more challenging
(this is equivalent to the problem of inference
in Bayesian neural networks, see e.g., \citep{Arbel2023} for a review paper
for the offline case, and \citep{Duran-Martin2022,bong} for some recent online methods).

We can generalize the above methods to compute the belief state
for  the parameters of an MDP in the obvious way,
but modeling both the reward function and state transition function.

Once we have computed the belief state,
we can derive a policy with optimal  regret using the
methods  like UCB (\cref{sec:UCB})
or Thompson sampling (\cref{sec:thompson}).

\eat{
We can compute an optimal policy
for a contextual bandit by defining
the agent belief state  to be $s_t=\vb_t = p(\vw|\vh_t)$,
where $\vw$ are the weights of the reward model,
$R(o,a)=f(o,\vw)_a$ for some function $f$.
We can update this state using Bayes rule to give
$s_{t+1}=SU(s_t,a_t,r_{t+1})$.
The optimal policy $\pi(s_t)$ can be computed using the methods
described in
\cref{sec:beliefStateMDP},
although in practice is more common to use simpler heuristics
like UCB (\cref{sec:UCB})
or Thompson sampling (\cref{sec:thompson}).
}

\subsection{Optimization problems as decision problems}
\label{sec:opt}

The  bandit problem is an example of a problem where the agent
must interact with the world in order to collect information,
but it does not otherwise affect the environment.
Thus the agents internal belief state changes over time,
but the environment state does not.\footnote{
In the contextual bandit problem, the environment state (context) does change,
but not in response to the agent's actions.
Thus $p(o_t)$ is usually assumed  to be a static distribution.
}
Such problems commomly arise when we are trying to optimize a fixed but unknown
function $R$. We can ``query'' the function by evaluating it at different
points (parameter values), and in some cases, the resulting observation
may also include gradient information.
The agent's goal is to find the optimum of the function in as few steps
as possible.\footnote{
If we only care about the final
performance of the agent,
we can try to minimize the \keywordDef{simple regret},
which is just the regret
at the last step, namely $\regret_T$.
This is the difference between the function value we chose
and the true optimum.
Minimizing simple regret results in a problem
known as  \keywordDef{pure exploration}
\citep{Bubeck2011},
where the agent needs to interact with the environment
to learn the underlying MDP;
at the end, it can then solve for the resulting policy
using planning methods
(see \cref{sec:planning}).
However, in general RL problems, it is more common to focus
on 
the \keywordDef{cumulative regret},
also called the
\keywordDef{total regret} or just the \keywordDef{regret},
which is defined as
$\totalRegret_T \defeq \expect{\sum_{t=1}^T \regret_t}$.
}
We give some examples of this problem setting below.

\subsubsection{Best-arm identification}

In the standard multi-armed bandit problem
our goal is to maximize the sum of expected
rewards.
However, in some cases, the goal is to determine 
the best arm given a fixed budget of $T$ trials;
this variant is
known as \keywordDef{best-arm identification}~\citep{Audibert10Best}.
Formally, this corresponds to optimizing the \keywordDef{final reward}
criterion:
\be
V_{\pi,\pi_T} = \expectQ{R(\hat{a})}{p(a_{1:T}, r_{1:T}|s_0,\pi)}
\ee
where $\hat{a} = \pi_T(a_{1:T}, r_{1:T})$ is the estimated optimal arm
as computed by the \keywordDef{terminal policy} $\pi_T$
applied to the sequence of observations obtained by the exploration policy $\pi$.
This can be solved by a simple adaptation of the methods used for standard bandits.

\subsubsection{Bayesian optimization}
\label{sec:BO}

Bayesian optimization is a gradient-free approach to optimizing expensive blackbox functions.
That is, we want to find
\be
\vw^* = \argmax_{\vw} R(\vw)
\ee
for some unknown function $R$,
where $\vw \in \real^N$,
using as few actions (function evaluations of $R$) as possible.
This is essentially  an ``infinite arm'' version of the best-arm identification problem
\citep{Toussaint2014},
where we replace the discrete choice of arms $a \in \{1,\ldots,K\}$
with the parameter vector $\vw \in \real^N$.
In this case, the optimal policy can be computed
if the agent's state $s_t$ is a belief state over 
the unknown function, i.e., $s_t=p(R|\vh_t)$.
A common way to represent this distribution is to use Gaussian processes.
We can then use heuristics like expected improvement, knowledge gradient
or Thompson sampling
to implement the corresponding policy, $\vw_t=\pi(s_t)$.
For details, see e.g., \citep{Garnett2023}.

\subsubsection{Active learning}

Active learning is similar to BayesOpt, but instead of trying to find the point
at which the function is largest (i.e., $\vw^*$), we are trying to learn
the whole function $R$, again by querying it at different points $\vw_t$.
Once again, the optimal strategy again requires maintaining a belief state over the unknown function,
but now the best policy takes a different form, such as choosing query points
to reduce the entropy
of the belief state.
See e.g., \citep{Smith2023}.

\subsubsection{Stochastic Gradient Descent (SGD)}

Finally we discuss how to interpret SGD as a sequential decision making process,
following \citep{Powell2022}.
The action space  consists of querying the unknown function $R$ at
locations $\va_t=\vw_t$, and observing the function value $r_t=R(\vw_t)$;
however, unlike BayesOpt, now we also observe
the corresponding gradient $\vg_t = \nabla_{\vw} R(\vw)|_{\vw_t}$,
which gives non-local information about the function.
The environment state contains the true function $R$
which is used to generate the observations given the agent's actions.
The agent state contains the current parameter estimate
$\vw_t$, and may contain  other information such as first and second moments
$\vm_t$ and $\vv_t$,
needed by methods such as Adam.
The update rule (for vanilla SGD) takes the form
$\vw_{t+1}=\vw_t + \alpha_t \vg_t$,
where the stepsize $\alpha_t$ is chosen by the policy,
$\alpha_t=\pi(s_t)$.
The terminal policy has the form $\pi(s_T)=\vw_T$.

Although in principle it is possible to learn the learning rate (stepsize) policy
using RL (see e.g., \citep{Xu2017RL}), the policy is usually
chosen by hand, either using a \keywordDef{learning rate schedule}
or some kind of manually designed \keywordDef{adaptive learning rate} policy
(e.g., based on second order curvature information).

\section{Reinforcement Learning: a high-level summary}
\label{sec:RLoverview}

In this section, we give a
brief overview of how to compute
optimal policies when the model of the environment is unknown;
this is the core problem tackled by RL.
We mostly  focus on the MDP case, but discuss the POMDP case
in \cref{sec:partialObs}.

We can categorize RL methods along mutiple 
dimensions, such as the following:
\begin{itemize}
 \item What does    the agent learn?
   Options include
the value function,
 the policy,
 the model,
 or some combination of the above.

\item How does the agent represent its unknown functions?
  The two main choices are to use non-parametric or
  \keywordDef{tabular representations}, or to use parametric representations
based on function approximation.
If these functions are based on neural networks,
this approach is called ``\keywordDef{deep RL}'',
where the term ``deep'' refers to the use of neural networks
with many layers.

\item How are the  actions are selected?
  Options include
  \keywordDef{on-policy} methods,
  where actions must be selected by the agent's
current policy),
and \keywordDef{off-policy} methods,
where actions can be select by any kind of policy,
including human demonstrations.
\end{itemize}

\cref{tab:RL} lists a few common examples of RL methods,
classified along these lines.
More details are given in the subsequent sections.

\begin{table}
  \centering
\begin{tabular}{lllll}
Approach &  Method & Functions learned & On/Off & Section
  \\ \hline
Value-based &  SARSA & $Q(s,a)$ & On & \cref{sec:SARSA} \\          
Value-based &   $Q$-learning & $Q(s,a)$ & Off & \cref{sec:Qlearning} 
  \\ \hline
Policy-based &  REINFORCE & $\policy(a|s)$ & On & \cref{sec:REINFORCE} \\
Policy-based &   A2C & $\policy(a|s)$, $V(s)$ & On &  \cref{sec:A2C} \\      
Policy-based &   TRPO/PPO & $\policy(a|s)$, $\advantage(s,a)$ & On & \cref{sec:PPO} \\
Policy-based &   DDPG & $a=\policy(s)$, $Q(s,a)$ & Off & \cref{sec:DDPG} \\
Policy-based &   Soft actor-critic & $\policy(a|s)$, $Q(s,a)$ & Off & \cref{sec:SAC} 
  \\ \hline
Model-based &  MBRL & $p(s'|s,a)$ & Off & \cref{sec:MBRL}
\end{tabular}
\caption{
  Summary of some popular methods for RL.
  On/off refers to on-policy vs off-policy methods.
}
\label{tab:RL}
\end{table}

\subsection{Value-based RL}
\label{sec:rl-intro-valuebased}

In this section, we give a brief introduction to \keywordDef{value-based RL},
also called \keywordDef{Approximate Dynamic Programming} or \keywordDef{ADP};
see \cref{sec:valueRL} for more details.

We introduced the value function $\Vpol(s)$ in \cref{eqn:valueFn},
which we repeat here for convenience:
\begin{align}
\Vpol(s)
\defeq \expectQ{\return_0 | s_0=s}{\policy}
=\expectQ{\sum_{t=0}^{\infty} \gamma^t r_{t} | s_0=s}{\policy}
\end{align}
The value function for the optimal policy $\pi^*$
is known to satisfy the following recursive condition,
known as 
\keywordDef{Bellman's equation}:
\begin{align}
  \Vopt(s) &=
   \max_a R(s,a) + \gamma \expectQ{\Vopt(s')}{\ptran(s'|s,a)} 
\end{align}
This follows from the principle of \keywordDef{dynamic programming},
which computes the optimal solution to a problem (here the value of state $s$)
by combining
the optimal solution of various subproblems (here the values
of the next states $s'$).
This can be used to derive the following learning rule:
\be
V(s) \leftarrow V(s) + \lr[r + \gamma V(s') - V(s)]
\ee
where $s' \sim \ptran(\cdot|s,a)$
is the next state sampled from the environment,
and $r=R(s,a)$ is the observed reward.
This is called  \keywordDef{Temporal Difference} or \keywordDef{TD} learning
(see \cref{sec:TD} for details).
Unfortunately,  it is not clear how to derive a policy if all we know is the
value function.
We now describe a solution to this problem.

We first generalize the notion of value function
to assigning a value to a state and action pair, by defining the
\keywordDef{Q function} as follows:
\begin{align}
\Qpol(s,a)
\defeq \expectQ{\return_0 | s_0=s,a_0=a}{\policy}
=\expectQ{\sum_{t=0}^{\infty} \gamma^t r_{t} | s_0=s,a_0=a}{\policy}
\end{align}
This quantity represents the expected return obtained
if we start by taking action $a$ in state $s$,
and then follow $\policy$ to choose actions thereafter.
The $Q$ function for the optimal policy
satisfies a modified Bellman equation
\begin{align}
    \Qopt(s,a) &= R(s,a) + \gamma  \expectQ{\max_{a'} \Qopt(s',a')}{\ptran(s'|s,a)}
\end{align}
This gives rise to the following TD update rule:
\begin{align}
  Q(s,a) \leftarrow
r + \gamma \max_{a'} Q(s',a')- Q(s,a)
\end{align}
where we sample $s' \sim \ptran(\cdot|s,a)$ from the environment.
The action is chosen at each step from the implicit policy
\be
a = \argmax_{a'} Q(s,a')
\ee
This is called \keywordDef{Q learning}
(see \cref{sec:Qlearning} for details),

\eat{
we try to learn the optimal $Q$-function from experience,
and then derive a policy from it using \cref{eqn:optPolFromQ}.
Typically, a function approximator
(e.g., a neural network), $\Qapprox$,
is used to represent the $Q$-function,
which is trained iteratively.
Given a transition $(s,a,r,s')$, we define
the \keywordDef{temporal difference}
(also called the \keywordDef{TD error}) as
\begin{align}
r + \gamma \max_{a'} \Qapprox(s',a')- \Qapprox(s,a)
\end{align}
Clearly, the expected TD error is the Bellman error evaluated at $(s,a)$.
Therefore, if $\Qapprox=\Qopt$, the TD error is $0$ on average by Bellman's optimality equation.
Otherwise, the error provides a signal
for the agent to change $\vw$ to make
$\Qapprox(s,a)$ closer to $R(s,a) + \gamma \max_{a'} \Qapprox(s',a')$.
The update on $\Qapprox$ is based on a target
that is computed using $\Qapprox$.
This is called \keywordDef{Q-learning}.
This kind of update is known as \keywordDef{bootstrapping} in RL,
and should not be confused with the
statistical bootstrap.
Value based methods are discussed in more detail in \cref{sec:RLvalue}.
}

\subsection{Policy-based RL}

In this section we give a brief introductin
to \keywordDef{Policy-based RL}; for details see  \cref{sec:policySearch}.

In policy-based methods,
we try to directly maximize $J(\pi_{\vtheta})=\expectQ{V_{\pi}(s_0)}{p(s_0)}$
wrt the parameter's $\vtheta$;
this is called \keywordDef{policy search}.
If $\polval(\polapprox)$ is differentiable wrt $\vtheta$,
we can use stochastic gradient ascent to optimize
$\vtheta$, which is known as
\keywordDef{policy gradient} (see \cref{sec:PG}).

Policy gradient methods have the advantage that
they provably converge to a local optimum
for many common policy classes,
whereas $Q$-learning may diverge
when approximation is used
(\cref{sec:offpolicyrl-deadlytriad}).
In addition, policy gradient methods can easily be applied to
continuous action spaces, 
since they do not need to compute $\argmax_a Q(s,a)$.
Unfortunately,
the score function estimator
for $\nabla_{\vtheta} \polval(\polapprox)$ can have a very high 
variance, so the resulting method can converge slowly.

One way to reduce the variance is to learn
an approximate value function, $\Vapprox(s)$,
and to use it as a baseline
in the score function estimator.
We can learn $\Vapprox(s)$ using 
using TD learning.
Alternatively, we can learn an advantage function,
$\Aapprox(s,a)$, and use it as a baseline.
These policy gradient variants are called
\keywordDef{actor critic} methods,
where the actor refers to the policy $\polapprox$
and the critic refers to $\Vapprox$ or $\Aapprox$.
See \cref{sec:AC} for details.

\subsection{Model-based RL}

In this section, we give a brief introduction to \keywordDef{model-based RL};
for more details, see
\cref{sec:MBRL}.

Value-based methods, such as Q-learning,
and policy search methods, such as policy gradient,
can be very \keywordDef{sample inefficient},
which means they may need to interact with the environment many times
before finding a good policy,
which can be problematic when real-world interactions are expensive.
In \keyword{model-based RL},
we first learn the MDP,
including  the $\ptran(s'|s,a)$ and $R(s,a)$ functions,
and then compute the policy, either using approximate dynamic programming
on the learned model, or doing lookahead search.
In practice, we often interleave the model learning
and planning phases,
so we can use the partially learned policy to decide
what data to collect, to help learn a better model.

\subsection{State uncertainty (partial observability)}
\label{sec:partialObs}

In an MDP, we assume that the state of the environment $s_t$
is the same as the observation $o_t$ obtained by the agent.
But in many problems, the observation only gives partial information
about the underlying state of the world
(e.g., a rodent or robot navigating in a maze).
This is called \keywordDef{partial observability}.
In this case, using a policy of the form
$a_t = \pi(o_t)$ is suboptimal, since $o_t$
does not give us complete state information.
Instead we need to use a policy of the form
$a_t = \pi(\vh_t)$,
where $\vh_t=(a_1, o_1,  \ldots, a_{t-1}, o_t)$
is the entire past history of observations and actions,
plus the current observation.
Since depending on the entire past is not tractable
for a long-lived agent, various approximate solution methods
have been developed, as we summarize below.

\subsubsection{Optimal solution}

If we know the true latent structure of the world
(i.e., both $p(o|z)$ and $p(z'|z,a)$, to use the notation
of \cref{sec:universal}),
then we can use solution methods designed
for  \keyword{POMDPs}, discussed in \cref{sec:POMDP}.
This requires using Bayesian inference to  compute a belief state,
$\vb_t = p(\envstate_t|\vh_t)$ (see \cref{sec:beliefStateMDP}),
and then using this belief state
to guide our decisions.
\eat{
In particular, we can derive the corresponding
\keywordDef{belief state MDP},
in which the states are belief states,
and the state dynamics are deterministic (conditional on the next
observation), as given by Bayes' rule:
\be
p(\vb_{t+1}|\vb_t,\vo_{t+1},\va_t)
\propto p(\vb_{t+1}|\vb_t, \va_t) p(\vo_{t+1}|\vb_{t_1})
\ee
}
However, learning the parameters of a POMDP (i.e., 
the  generative latent world model) is very difficult,
as is recursively computing and updating the belief state,
as is computing the policy given the belief state.
Indeed, optimally solving POMDPs is known to be
computationally very difficult for any method
\citep{Papadimitriou87,Kaelbling98}.
So in practice simpler approximations are used.
We discuss some of these below.
(For more details, see \citep{Murphy00pomdp}.)

Note that it is possible to 
marginalize out the POMDP
latent state $\envstate_t$,
to derive a prediction over the next observable state,
$p(\vo_{t+1}|\vh_t,\va_t)$.
This can then become a learning  target for a model, that is trained
to directly
predict  future observations, without explicitly invoking the concept
of latent state.
This is called a \keywordDef{predictive state representation}
or \keywordDef{PSR} \citep{PSR}.
This is related to the idea of
\keywordDef{observable operator models}
\citep{Jaeger2000},
and to the concept of
successor representations
which we discuss in  \cref{sec:SR}.

\subsubsection{Finite observation history}

The simplest solution to the partial
observability problem is to  define
the state to be a finite history of the last $k$ observations,
$\vs_t = \vh_{t-k:t}$;
when the observations $\vo_t$ are images,
this is often called \keywordDef{frame stacking}.
We can then use standard MDP methods.
Unfortunately, this cannot capture long-range dependencies in the data.

\subsubsection{Stateful (recurrent) policies}
  
A more powerful approach is to use a stateful policy,
that can remember the entire past, and not just respond
to the current input or last $k$ frames.
For example, we can represent the policy by an RNN (recurrent neural
network),
 as proposed in  the \keywordDef{R2D2} paper \citep{R2D2},
 and used in many other papers.
Now the hidden state $\envstate_t$ of the RNN
will implicitly summarize the past observations, $\vh_t$,
and can be used in lieu of the state $s_t$ in any standard RL algorithm.

RNNs policies are widely used, and this method is often effective in solving partially
observed problems. However, they typically will not plan to perform
information-gathering actions, since there is no explicit notion of
belief state or uncertainty. However, such behavior can arise
via meta-learning \citep{Mikulik2020}.

\eat{
Alternatively, \citep{Moss2024} present an approach
called \keywordDef{BetaZero}, that amortizes look ahead search
in belief space into a neural network,
analogous to MuZero (\cref{sec:muzero}).
Since this plans in the space of distributions,
it can choose to perform information gathering actions.
}

\subsection{Model uncertainty (exploration-exploitation tradeoff)}
\label{sec:exploreExploit}
\label{sec:explorationExploitation}
\label{sec:explore-exploit}

In RL problems, we typically assume the underlying
transition and reward models are not known.
We can either try to explicitly learn these models (as in model-based RL),
and then solve for the policy,
or just learn the policy directly (as in model-free RL).
But in either case, we need to explore the environment
in order to collect enough data to figure out what to do.
This may involve  choosing between
actions that the agent knows will yield high reward,
vs choosing actions which might not been known to yield high reward
but which will be informative about potential future gains.
This is called the 
\keywordDef{exploration-exploitation tradeoff}.
In this section, we discuss some simple heuristic solutions to this problem.
See \cref{sec:exploreExploit2} for more sophisticated methods.

\label{sec:epsGreedy}

If we just want to exploit our current knowledge
(without trying to learn new things),
we can use  the 
\keywordDef{greedy policy}:
\be
a_t=\argmax_a Q(s,a)
\ee
We can add exploration to this by 
sometimes picking some other, non-greedy action,
as we discuss below.

One approach
is to use an
\keywordSpecial{$\epsilon$-greedy}{epsilon-greedy} policy
$\policy_\epsilon$, parameterized by $\epsilon\in[0,1]$. 
In this case,  we pick
the greedy action wrt the current model,
$a_t = \argmax_a \estR_t(s_t,a)$ with probability
$1-\epsilon$, and a random action with probability $\epsilon$.
This rule ensures the agent's continual exploration
of all state-action combinations.
Unfortunately, this heuristic can be shown to be suboptimal,
since it explores every action with at least a constant
probability $\epsilon/|\calA|$,
although this can be solved by annealing $\epsilon$ to 0 over time.

Another problem with $\epsilon$-greedy is that it can result in ``dithering'',
in which the agent continually changes its mind about what to do.
In \citep{Dabney2021} they propose a simple solution to this
problem, known as
\keywordSpecial{$\epsilon z$-greedy}{epsilon-z-greedy},
that often works well.
The idea  is that with probability $1-\epsilon$ the agent
exploits, but with with probability $\epsilon$ the agent
explores by repeating the sampled action for $n \sim z()$ steps
in a row, where $z(n)$ is a distribution over the repeat duration.
This can help the agent escape from local minima.
(See also \citep{Treven2023}, who learn a policy
to not only pick an action, but also how long to use that action for,
by solving an augmented MDP where the action space is augmented
by duration.)

\label{sec:boltzmannExploration}
Another simple approach to exploration is to use 
\keywordDef{Boltzmann exploration},
which assigns higher probabilities to explore more promising actions,
taking into account the reward function.
That is, we use a policy of the form
\be
\policy_{\tau}(a|s) = \frac{\exp(\estR_t(s_t,a)/\tau)}{\sum_{a'} \exp(\estR_t(s_t,a')/\tau)}
\ee
where $\tau>0$ is a temperature parameter
that controls how entropic the distribution is.
As $\tau$ gets close to $0$, $\policy_{\tau}$ becomes close to a greedy policy.  
On the other hand, higher values of $\tau$ will make
$\policy(a|s)$ more uniform, and encourage more exploration.
Its action selection probabilities can be much ``smoother'' with
respect to changes in the reward estimates 
than $\epsilon$-greedy, as illustrated in
\cref{tab:epsGreedy}.

\begin{table}
  \centering
\begin{tabular}{llllll}
  $\estR(s,a_1)$ & $\estR(s,a_2)$
  & $\policy_{\epsilon}(a|s_1)$     & $\policy_{\epsilon}(a|s_2)$  
    & $\policy_{\tau}(a|s_1)$     & $\policy_{\tau}(a|s_2)$     \\
  \hline
  1.00 & 9.00 & 0.05 & 0.95 & 0.00 & 1.00 \\
  4.00 & 6.00 & 0.05 & 0.95 & 0.12 & 0.88 \\
  4.90 & 5.10 & 0.05 & 0.95 & 0.45 & 0.55 \\
  5.05 & 4.95 & 0.95 & 0.05 & 0.53 & 0.48 \\
  7.00 & 3.00 & 0.95 & 0.05 & 0.98 & 0.02 \\
  8.00 & 2.00 & 0.95 & 0.05 & 1.00 & 0.00
\end{tabular}
\caption{Comparison of $\epsilon$-greedy policy (with $\epsilon=0.1$)
  and Boltzmann policy (with $\tau=1$) for a simple MDP with 6
  states and 2 actions.
  \figbased{Table 4.1 of \citep{Graesser2019}}.
}
  \label{tab:epsGreedy}
\end{table}

The Boltzmann policy explores equally widely in all states.
An alternative approach is to try to explore (state,action)
combinations where the consequences of the outcome
might be uncertain.
This can be achived using  an \keywordDef{exploration bonus}
$R^b_t(s,a)$, which is large if the number
of times we have tried actioon $a$ in state $s$ is small.
We can then add $R^b$ to the regular reward,
to bias the behavior in a way that will hopefully
cause the agent to learn useful information
about the  world.
This is called an \keywordDef{intrinsic reward} function
(\cref{sec:intrinsicReward}).

\subsection{Reward functions}
\label{sec:reward}

Sequential decision making relies on the
 user to define the reward function in order to
 encourage the agent to exhibit some desired behavior.
In this section, we discuss this crucial aspect of the problem.

\subsubsection{The reward hypothesis}

The ``\keywordDef{reward hypothesis}'' states that
``all of
what we mean by goals and purposes can be well thought
of as maximization of the expected value of the cumulative
sum of a received scalar signal (reward)''
\citep{Sutton2004}.
(See also the closely related ``reward is enough'' hypothesis
\citep{Silver2021}.)
Whether this hypothesis is true or not depends on what
one means by ``goals and purposes''.
This can be formalized in terms of preference relations
over (state,action) trajectories,
as discussed in \citep{Bowling2023}.
(See also \citep{booth23perils,Knox2024} for some related work on reward
function design.)

\subsubsection{Non-Markovian rewards}

Most of the literature assumes the reward can be defined
in terms of the current state and action, $R(s,a)$,
or in terms of the most recent state transition, $R(s,a,s')$.
In \citep{Bowling2023},  they discuss when a utility function over trajectories
can be converted into a Markovian reward of the form
$R(s,a,s')$.

\begin{figure}
\centering
\includegraphics[height=2in]{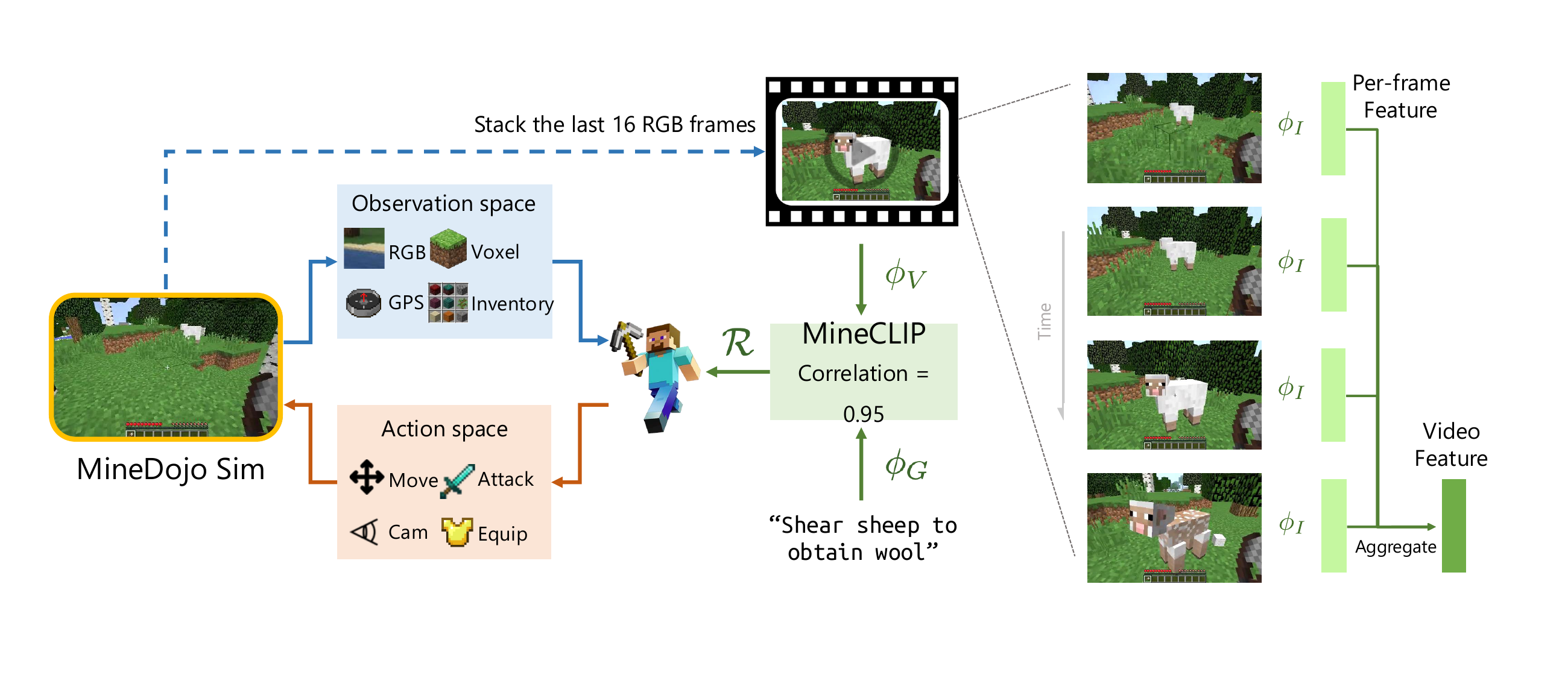}
\caption{
  Illustration of how the MineClip reward function can be used
  to help train an agent to play Minecraft in the MineDojo simulator.
  \figtaken{Figure 4 of \citep{MineDojo}.}
  \figthanks{Jim Fan}.
  }
\label{fig:mineClip}
\end{figure}

In general, the reward function will need to be non-Markovian.
For example, consider training an agent to solve various goals,
specified in natural language, inside the Minecraft video game.
(For a general discussion of goal-conditioned RL, see \cref{sec:GCRL}.)
In this case, we do not have access to the underlying world state,
and even if we did, it can be hard to determine from a single state,
or single state transition pair,
whether a generic goal (such as ``shear the sheep to obtain wool'')
has been satisifed.
In the \keywordDef{MineDojo} paper
\citep{MineDojo}, they tackled this problem by pre-training
a reward model of the form $R(o(t-K:t),g)$,
where $o(t-K:t)$ are the last $K$ frames,
and $g$ is the goal.
This model, known as \keywordDef{MineCLIP},
was trained using contrastive learning
applied to a large corpus of video-text pairs.\footnote{
To make this reward function fast to compute,
they computed it using a simple comparison
between the embedding of the goal, $\phi_G(g)$,
and the aggregated embeddings of each image, $1/K \sum_{k=0}^{K-1} \phi_I(o_{t-k})$.
By caching the embeddings of previously seen frames,
and using a frozen image encoder which is shared between the reward
and the agent, computation could be significantly sped up.
}

\subsubsection{Reward hacking}
\label{sec:rewardHacking}
\label{sec:hacking}

In some cases, the reward function may be misspecified, so even though the agent may
maximize the reward, this might turn out not to be what the user desired.
For example, suppose the user rewards the agent for making as many
paper clips as possible.
An optimal agent may convert the whole world into a paper clip factory,
because the user forgot to specify various constraints, such as not
killing people (which might otherwise be necessary
in order to use as many resources as possible for paperclips).
In the \keywordDef{AI alignment} community,
this example is known as  the \keywordDef{paperclip maximizer problem},
and is due to Nick Bostrom \citep{Bostrom2016}.
(See e.g., \url{https://openai.com/index/faulty-reward-functions/} for some examples
that have occurred in practice.)
This is an example of a more general problem
known as
\keywordDef{reward hacking}
\citep{Skalse2022}.
For a potential solution, based on the assistance game paradigm,
see \cref{sec:assistance}.

\eat{
For example, a rat which can electrically stimulate its dopamine
center by pressing a lever, will press the lever indefinitely,
until it starves to death \citep{Bennett2023}; this is known as
\keywordDef{wire heading}, and illustrates
the dangers of changing the reward function.
}

\subsubsection{Sparse reward}

Even if the reward function is correct, optimizing it is not always easy.
In particular, many problems suffer from \keywordDef{sparse reward},
in which $R(s,a)=0$ for almost all states and actions, so the agent
only every gets feedback (either positive or negative) on the rare occasions
when it achieves some unknown goal.
This requires \keywordDef{deep exploration}
\citep{Osband2019}
to find the rewarding states.
One approach to this is use to use PSRL (\cref{sec:PSRL}).
However, various other heuristics have been developed,
some of which we discuss below.

\subsubsection{Reward shaping}
\label{sec:rewardShaping}

In  \keywordDef{reward shaping},
we add prior knowledge about what we believe good states should look like,
as a way to combat the difficulties of learning from sparse reward.
That is, we define a new reward function $r' = r + F$,
where $F$ is called the shaping function.
In general, this can affect the optimal policy.
For example, if a soccer playing agent is ``artificially''
rewarded for making contact with the ball,
it might learn to repeatedly touch and untouch the ball
(toggling between $s$ and $s'$), rather than
trying to win the original game.
But in \citep{Ng99},
the prove that if the shaping function has the form
\be
F(s,a,s') = \gamma \Phi(s') - \Phi(s)
\ee
where $\Phi: \calS \ra \real$
is a \keywordDef{potential function},
then we can guarantee that the sum of shaped rewards
will match the sum of original rewards plus a constant.
This is called \keywordDef{Potential-Based Reward Shaping}.

In \citep{Wiewiora2003}, they prove that (in the tabular case)
this approach
is equivalent to initializing the value function
to $V(s)=\Phi(s)$.
In \citep{Tessler2019},
they propose
an extension called potential-based advice,
where they show that a potential of the form
$F(s,a,s',a') = \gamma \Phi(s',a') - \Phi(s,a)$
is also valid (and more expressive).
In \citep{Hu2020}, they introduce a reward shaping function $z$
which can be used to down-weight or up-weight the
shaping function:
\be
r'(s,a) = r(s,a) + z_{\phi}(s,a) F(s,a)
\ee
They use bilevel optimization to optimize $\phi$
wrt the original task performance.

%

\subsubsection{Intrinsic reward}

In \cref{sec:intrinsicReward}, we discuss \keywordDef{intrinsic reward},
which is a set of methods for encouraging agent behavior without the need
for any external reward signal.
For example, we might want agents to explore their environment just so
they can ``figure things out'', without any other specific goals in mind.
This can be useful even if there is an external reward, but it happens to be sparse.

\subsection{Best practices for experimental work in RL}

Implementing RL algorithms is much trickier than
methods for supervised learning,
or generative  methods such as language modeling
and diffusion,
all of which have stable (easy-to-optimize) loss functions.
Therefore it is often wise to build on existing
software rather than starting from scratch.
We list  some useful 
libraries  in Table~\ref{tab:software}.

Even with good code,
RL experiments can be very high variance,
making it hard to draw valid conclusions from an experiment.
See \citep{Agarwal2021RL,Patterson2024,Jordan2024} for some recommended
experimental practices.
For example, when reporting performance across different environments,
with different intrinsic difficulties (e.g., different kinds
of Atari games),
\citep{Agarwal2021RL} recommend
reporting the \keywordDef{interquartile mean} (IQM) of the performance metric,
which is the mean of the samples between the 0.25 and 0.75 percentiles,
(this is a special case of a trimmed mean).
Let this estimate be denoted by $\hat{\mu}(\data_i)$,
where $\data$ is the empirical data (e.g., reward vs time)
from the $i$'th run.
We can estimate the uncertainty in this estimate using
a nonparametric method, such as bootstrap resampling,
or a parametric approximation, such as
a Gaussian approximation. (This requires computing
the standard error of the mean, $\frac{\hat{\sigma}}{\sqrt{n}}$,
where $n$ is the number of trials,
and $\hat{\sigma}$ is the estimated  standard deviation
of the (trimmed) data.)

\begin{table}
  \centering
  \footnotesize
  \begin{tabular}{lll}
    URL & Language & Comments \\ \hline
    \href{https://github.com/EdanToledo/Stoix}{Stoix}
    & Jax & Mini-library with many methods (including MBRL)
    \\
    \href{https://github.com/luchris429/purejaxrl/tree/main}{PureJaxRL}
    & Jax & Single files with DQN; PPO, DPO
    \\
    \href{https://github.com/ikostrikov/jaxrl}{JaxRL}
    & Jax & Single files  with AWAC, DDPG, SAC, SAC+REDQ
        \\
    \href{https://stable-baselines3.readthedocs.io/en/master/guide/sbx.html}
         {Stable Baselines Jax}
         & Jax & Library with  DQN, CrossQ, TQC;  PPO, DDPG, TD3, SAC
     \\
     \href{https://github.com/tinker495/jax-baseline}{Jax Baselines}
     &  Jax & Library with many methods
     \\
     \href{https://github.com/keraJLi/rejax}{Rejax}
     & Jax & Library with DDQN, PPO, (discrete) SAC, DDPG
     \\
         \href{https://github.com/google/dopamine}{Dopamine}
         & Jax/TF & Library with many methods
\\
     \href{https://github.com/google-deepmind/rlax/tree/master}{Rlax}
     & Jax & Library of RL utility functions (used by Acme)
     \\
     \href{https://github.com/google-deepmind/acme/tree/master}{Acme}
     & Jax/TF & Library with many methods (uses rlax)
     \\
         \href{https://github.com/vwxyzjn/cleanrl}{CleanRL}
    & PyTorch & Single files with many methods
    \\
    \href{https://stable-baselines3.readthedocs.io/en/master/}{Stable
      Baselines 3}
    & PyTorch & Library with DQN; A2C, PPO, DDPG, TD3, SAC, HER
    \\
    \href{https://github.com/thu-ml/tianshou/}{TianShou}
    & PyTorch & Library with many methods (including offline RL)
    \end{tabular}
  \caption{Some open source RL software.}
    \label{tab:software}
\end{table}



\eat{

  Google code
  
  We may be about to open pandora's box of online RL methods. There are several external JAX implementations such as PureJaxRL  (environment and agent both in JAX) and Gymnax (only agent is in JAX, IIUC). 
For the record, I am also  copying a list from Pablo about internal codebases:
-Xebulba is a very good starting point (and initial experiments have already been conducted there), as it supports distributed training and already includes implementations of many algorithms.
-Reverb is a general-purpose storage system, typically used as a replay buffer for distributed RL training. It’s distributed design is different from that of Xebulba, so using both may provide us with a diverse set of viewpoints/insights into this problem.
-Dopamine is another option, in particular for synchronous training. We have recently added a new replay buffer which will allow us to integrate with Reverb, thus enabling distributed training.
-RLX is used for fine-tuning LLMs with RLHF, and supports both
synchronous and distributed actors and learners
}

\chapter{Value-based RL}
\label{sec:RLvalue}
\label{sec:valueBased}
\label{sec:valueRL}

\section{Basic concepts}

In this section we introduce some definitions and basic concepts.

\subsection{Value functions}
\label{sec:valueFn}

Let $\policy$ be a given policy.  We define the \keywordDef{state-value function},
or \keywordDef{value function} for short,
as follows (with $\expectQ{\cdot}{\policy}$ indicating that actions are selected by $\policy$):
\begin{align}
\Vpol(s)
\defeq \expectQ{\return_0 | s_0=s}{\policy}
=\expectQ{\sum_{t=0}^{\infty} \gamma^t r_{t} | s_0=s}{\policy}
\label{eqn:Vfn}
\end{align}
This is the expected return obtained if we start in state $s$ and follow $\policy$ to choose actions
in a continuing task (i.e., $T=\infty$).

Similarly, we define 
the \keywordDef{state-action value function},
also known as the \keywordDef{$Q$-function},
as follows:
\begin{align}
\Qpol(s,a)
\defeq \expectQ{\return_0 | s_0=s,a_0=a}{\policy}
=\expectQ{\sum_{t=0}^{\infty} \gamma^t r_{t} | s_0=s,a_0=a}{\policy}
\label{eqn:Qfn}
\end{align}
This quantity represents the expected return obtained
if we start by taking action $a$ in state $s$,
and then follow $\policy$ to choose actions thereafter.

Finally, we define the \keywordDef{advantage function} as follows:
\be
\Apol(s,a) \defeq \Qpol(s,a)  - \Vpol(s)
\ee
This tells us the benefit of picking action $a$
in state $s$ then switching to policy $\policy$, relative to the baseline return of always following $\policy$.
Note that $\Apol(s,a)$ can be both positive and negative, and $\expectQ{\Apol(s,a)}{\policy(a|s)}=0$
due to a useful equality:
$\Vpol(s) = \expectQ{\Qpol(s,a)}{\policy(a|s)}$.

\subsection{Bellman's equations}
\label{sec:optPolicy}
\label{sec:bellman}
\label{sec:Bellman}

Suppose $\polopt$ is a policy such that $\Vpolopt \ge \Vpol$ for all $s \in \calS$ and all policy $\policy$, then it is an \keywordDef{optimal policy}.
There can be multiple optimal policies for the same MDP, but by definition their value functions must be the same, and are denoted by $\Vopt$ and $\Qopt$, respectively.
We call $\Vopt$ the \keywordDef{optimal state-value function}, and $\Qopt$ the \keywordDef{optimal action-value function}.
Furthermore, any finite MDP must have at least one deterministic optimal policy~\citep{Puterman94}.

A fundamental result about the optimal value function is
\keywordDef{Bellman's optimality equations}:
\begin{align}
  \Vopt(s) &=
   \max_a R(s,a) + \gamma \expectQ{\Vopt(s')}{\ptran(s'|s,a)} 
\label{eqn:bellmanOptV} \\
    \Qopt(s,a) &= R(s,a) + \gamma  \expectQ{\max_{a'} \Qopt(s',a')}{\ptran(s'|s,a)}
    \label{eqn:bellmanOptQ}
\end{align}
Conversely, the optimal value functions are the only solutions that satisfy the equations.
In other words, although the value function is defined as the
expectation of a sum of infinitely many rewards, it can be
characterized by a recursive equation
that involves only one-step transition and reward models of the MDP.
Such a recursion play a central role in many RL algorithms we will see later.

Given a value function ($V$ or $Q$),
the discrepancy between 
the right- and left-hand sides of
\cref{eqn:bellmanOptV,eqn:bellmanOptQ}
are called \keywordDef{Bellman error}
or \keywordDef{Bellman residual}.
We can define the \keywordDef{Bellman operator}
$\calB$ given an MDP $M=(R,T)$ and policy $\pi$
as a function that takes a value function $V$
and derives a few value function $V'$ that satisfies
\be
V'(s) = \calB^{\pi}_{M} V(s)
\defeq \expectQ{R(s,a) + \gamma \expectQ{V(s')}{T(s'|s,a)}}{\pi(a|s)}
\ee
This reduces the Bellman error.
Applying the Bellman operator to a state is called a
\keywordDef{Bellman backup}.
If we iterate this process, we will converge to the optimal
value function $V_*$, as we discuss in \cref{sec:VI}.

Given the optimal value function, we can derive an optimal policy using
\begin{align}
\polopt(s) &= \argmax_a \Qopt(s,a)
\label{eqn:optPolFromQ} \\
  &= \argmax_a \left[ R(s,a) + \gamma \expectQ{\Vopt(s')}{\ptran(s'|s,a)} \right]
\label{eqn:optPolFromV}
\end{align}
Following such an optimal policy ensures
the agent achieves maximum expected return
starting from any state. 

The problem of solving for $\Vopt$, $\Qopt$ or $\polopt$
is called \keywordDef{policy optimization}.
In contrast, solving for $\Vpol$ or $\Qpol$ for a
given policy $\policy$ is called
\keywordDef{policy evaluation},
which constitutes an important subclass of RL
problems as will be discussed in later sections.
For policy evaluation, we have similar Bellman equations,
which simply replace $\max_a\{\cdot\}$
in \cref{eqn:bellmanOptV,eqn:bellmanOptQ} with 
$\expectQ{\cdot}{\policy(a|s)}$.

\eat{
We have assumed a discrete state space in
\cref{eqn:optPolFromV}, so that $\ptran$ is
a conditional probability mass function.
For continuous state spaces, $\ptran$ is
a probability density function, in which case
we can simply replace the summation with integration.
In this chapter, both forms will be used interchangeably, if there is no ambiguity.
}

In \cref{eqn:optPolFromQ,eqn:optPolFromV}, as in the Bellman optimality equations, we must take a maximum over all actions in $\calA$, and the maximizing action is called the \keywordDef{greedy action} with respect to the value functions, $\Qopt$ or $\Vopt$.
Finding greedy actions is computationally easy if $\calA$ is a small finite set.
For high dimensional continuous spaces,  see \cref{sec:QTopt}.

\subsection{Example: 1d grid world}
\label{sec:Q1d}

\begin{figure}
\centering
\includegraphics[height=2.5in]{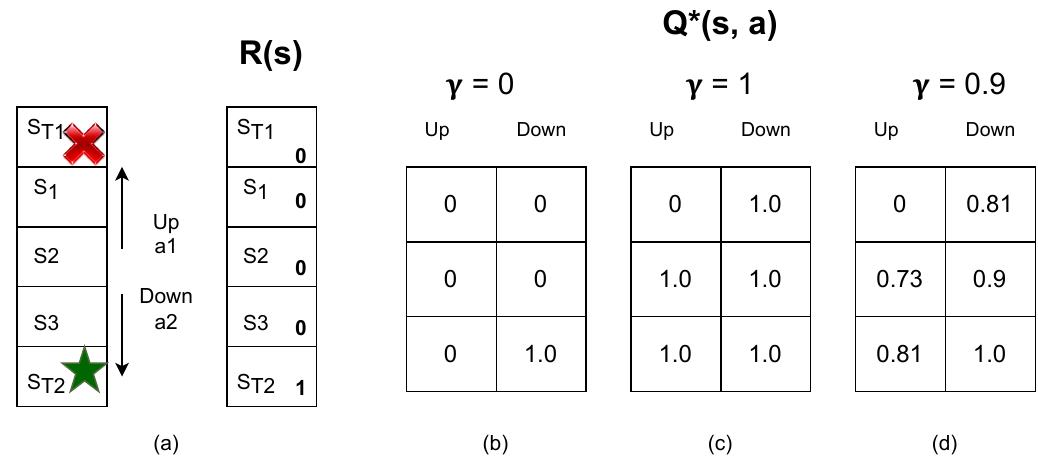}
\caption{
  Left: illustration of a simple MDP
  corresponding to a 1d grid world
  of 3 non-absorbing states and 2 actions.
  Right: optimal $Q$-functions for different values of $\gamma$.
\figbased{Figures 3.1, 3.2, 3.4 of \citep{Graesser2019}}.
}
\label{fig:Q1d}
\end{figure}

In this section, we show a simple example,
to make some of the above concepts more concrete.
Consider the 1d \keywordDef{grid world}
shown in \cref{fig:Q1d}(a). There are 5 possible states,
among them $S_{T1}$ and $S_{T2}$ are absorbing states,
since the interaction ends once the agent enters them.
There are 2 actions, $\uparrow$ and $\downarrow$.
The reward function is zero everywhere except
at the goal state, $S_{T2}$,
which gives a reward of $1$ upon entering.
Thus the optimal action in every state is to move down.

\cref{fig:Q1d}(b) shows the $\Qopt$ function for $\gamma=0$.
Note that we only show the function for non-absorbing states,
as the optimal $Q$-values are $0$ in absorbing states
by definition.
We see that $\Qopt(s_3,\downarrow)=1.0$, since the agent will get a reward of $1.0$
on the next step if it moves down from $s_3$;
however, $\Qopt(s,a)=0$ for all other state-action pairs,
since they do not provide nonzero immediate reward.
This optimal $Q$-function reflects the fact that
using $\gamma=0$ is completely \keyword{myopic},
and ignores the future.

\cref{fig:Q1d}(c) shows $\Qopt$ when $\gamma=1$.
In this case, we care about all future rewards equally.
Thus $\Qopt(s,a)=1$ for all state-action pairs,
since the agent can always reach the goal eventually. 
This is infinitely far-sighted.
However, it does not give the agent
any short-term guidance on how to behave.
For example, in $s_2$, it is not clear if it is should go up or down,
since both actions will eventually reach the goal
with identical $\Qopt$-values.

\cref{fig:Q1d}(d) shows $\Qopt$ when $\gamma=0.9$.
This reflects a preference for near-term rewards,
while also taking future reward into account.
This encourages the agent to seek the shortest path
to the goal, which is usually what we desire.
A proper choice of $\gamma$ is up to the agent designer, just like the design of the reward function,
and has to reflect the desired behavior of the agent.

\section{Solving for the optimal policy in a known world model}
\label{sec:planning}
\label{sec:rl-planning}
\label{sec:DPRL}

In this section, we discuss how
to compute the optimal value function (the \keywordDef{prediction problem})
and the optimal policy (the \keywordDef{control problem})
when the MDP model is known.
(Sometimes the term \keywordDef{planning} is used to refer
to computing the optimal policy, given a known model,
but planning can also refer to computing a sequence of actions,
rather than a policy.)
The algorithms we discuss are based on
\keywordDef{dynamic programming} (DP)
and \keywordDef{linear programming} (LP).

For simplicity, in this section, we assume
discrete state and action sets with $\gamma<1$.
However,
exact calculation of optimal policies often depends
polynomially on the sizes of $\calS$ and $\calA$,
and is intractable, for example, when the state space
is a Cartesian product of several finite sets.
This challenge is known as the
\keywordDef{curse of dimensionality}.
Therefore, approximations are typically needed,
such as using parametric or nonparametric representations of the value function or policy,
both for computational tractability and
for extending the methods to handle MDPs
with general state and action sets.
This requires the use of
\keywordDef{approximate dynamic programming} (ADP)
and \keywordDef{approximate linear programming} (ALP)
algorithms (see e.g., \cite{BertsekasRL}).

\eat{
using \keywordDef{dynamic programming} (\keywordDef{DP}).
The inputs to the algorithm
are the world model $p(s'|s,a)$
and reward function $R(s,a)$,
and the output is the optimal policy and 
its value function.
(This is called  \keywordDef{planning} using a known model.)
We assume the states and actions are discrete,
although similar methods can also be used in the linear-Gaussian case.
}

\subsection{Value iteration}
\label{sec:valueIter}
\label{sec:VI}

A popular and effective DP method for solving an MDP is
\keywordDef{value iteration} (VI).
Starting from an initial value function estimate $V_0$, the algorithm
iteratively updates the estimate by 
\begin{align}
\label{eqn:value-iteration}
V_{k+1}(s) = \max_a \left[
  R(s,a) + \gamma \sum_{s'} p(s'|s,a) V_k(s') \right]
\end{align}
Note that the update rule, sometimes called a
\keywordDef{Bellman backup},
is exactly the right-hand side of the Bellman optimality equation \cref{eqn:bellmanOptV}, with the unknown $\Vopt$ replaced by the current estimate $V_k$.
A fundamental property of \cref{eqn:value-iteration}
is that the update is a \keywordDef{contraction}:
it can be verified that
\begin{align}
\label{eqn:bellman-contraction}
\max_s |V_{k+1}(s) - \Vopt(s)| \le \gamma \max_s |V_k(s) - \Vopt(s)|
\end{align}
In other words, every iteration will reduce
the maximum value function error
by a constant factor.

$V_k$ will converge to $\Vopt$,
after which an optimal policy can be extracted using \cref{eqn:optPolFromV}.
In practice, we can often terminate VI
when $V_k$ is close enough to $\Vopt$,
since the resulting greedy policy wrt $V_k$
will be near optimal.
Value iteration can be adapted to learn the optimal action-value function $\Qopt$.

\subsection{Real-time dynamic programming (RTDP)}
\label{sec:RTDP}

In value iteration, we compute $\Vopt(s)$ and $\polopt(s)$
for all possible states $s$,
averaging over all possible next states $s'$ at each iteration,
as illustrated in \cref{fig:sutton-8-6}(right).
However, for some problems,
we may only be interested in the value (and policy)
for certain special starting states.
This is the case, for example, in
\keywordDef{shortest path problems} on graphs,
where we are trying to find the shortest
route from the current state to a goal state.
This can be modeled as an episodic MDP
by defining a transition matrix
$\ptran(s'|s,a)$ where
taking edge $a$ from node $s$ leads to the neighboring node $s'$
with probability 1.
The reward function is defined as $R(s,a)=-1$ for all states $s$
except the goal states,
which are modeled as absorbing
states.

In problems such as this, we can use a method
known as \keywordDef{real-time dynamic programming}
or \keywordDef{RTDP} \citep{Barto1995},
to efficiently compute an \keywordDef{optimal partial policy},
which only specifies what to do for the reachable states.
RTDP maintains a value function estimate $V$.
At each step, it performs a Bellman backup for the
current state $s$ by
$V(s) \assign \max_a \expectQ{R(s,a) + \gamma V(s')}{\ptran(s'|s,a)}$.
It picks an action $a$ (often with some exploration), reaches a next state $s'$,
and repeats the process.
This can be seen as a form of the more general
\keywordDef{asynchronous value iteration},
that focuses its computational effort on parts of the state
space that are more likely to be reachable from the current state,
rather than synchronously updating all states at each iteration.

\subsection{Policy iteration}
\label{sec:policyIteration}

Another effective DP method for computing $\polopt$ is \keywordDef{policy iteration}.
It is an iterative algorithm that searches in the space
of deterministic policies until converging to an optimal policy.
Each iteration consists of two steps, \keywordDef{policy evaluation} and \keywordDef{policy improvement}.

The policy evaluation step, as mentioned earlier,
computes the value function for the current policy.
Let $\policy$ represent the current policy,
$\vv(s)=\Vpol(s)$ represent the value function encoded as a vector indexed by states,
$\vr(s) = \sum_a \policy(a|s) R(s,a)$ represent the reward vector,
and $\vT(s'|s) = \sum_a \policy(a|s) p(s'|s,a)$ 
represent the state transition matrix.
Bellman's equation for policy evaluation can be
written in the matrix-vector form as
\begin{align}
  \vv &= \vr + \gamma \vT \vv
  \label{eqn:policyEval}
  \end{align}
This is a linear system of equations in $|\cal{S}|$ unknowns.
We can solve it using matrix inversion:
$\vv = (\vI - \gamma \vT)^{-1} \vr$.
Alternatively, we can use value iteration
by computing
$\vv_{t+1} = \vr + \gamma \vT \vv_t$
until near convergence,
or some form of asynchronous variant
that is computationally more efficient.

Once we have evaluated $\Vpol$ for the current policy $\policy$,
we can use it to derive a better policy $\policy'$,
thus the name policy improvement.
To do this, we simply compute a deterministic
policy $\policy'$ that acts greedily with respect to $\Vpol$ in every
state, using
\be
\policy'(s) = \argmax_a \{R(s,a) + \gamma \expect{\Vpol(s')}\}
\label{eqn:policyUpdate}
\ee
We can guarantee that $V_{\policy'} \ge \Vpol$.
This is called the \keywordDef{policy improvement theorem}.
To see this, define $\vr'$, $\vT'$ and $\vv'$ as before,
but for the new policy $\policy'$.
The definition of $\policy'$ implies
$\vr' + \gamma \vT' \vv \ge \vr + \gamma \vT \vv = \vv$,
where the equality is due to Bellman's equation.
Repeating the same equality, we have
\begin{align}
\vv &\le \vr' + \gamma \vT' \vv
\le \vr' + \gamma \vT' (\vr' + \gamma \vT' \vv)
\le \vr' + \gamma \vT' (\vr' + \gamma \vT' (\vr' + \gamma \vT' \vv) ) \le \cdots \\
&= (\vI + \gamma \vT' + \gamma^2 \vT'^2 + \cdots) \vr'
= (\vI - \gamma \vT')^{-1} \vr'
= \vv'
\end{align}

Starting from an initial policy $\policy_0$, policy iteration alternates
between policy evaluation ($E$)
and improvement ($I$) steps,
as illustrated below:
\begin{align}
\policy_0 \stackrel{E}{\ra} V_{\policy_0}
\stackrel{I}{\ra} \policy_1
\stackrel{E}{\ra} V_{\policy_1}
\cdots
\stackrel{I}{\ra} \polopt
\stackrel{E}{\ra} \Vopt
\end{align}
The algorithm stops at iteration $k$, if the policy $\policy_k$
is greedy with respect to its own value function $V_{\policy_k}$.
In this case, the policy is optimal.
Since there are at most $|\calA|^{|\calS|}$ deterministic policies,
and every iteration strictly improves the policy, the algorithm must converge after finite iterations.

In PI,
we alternate between policy evaluation (which involves multiple
iterations, until convergence of $\Vpol$),
and policy improvement.
In VI, we alternate between
one iteration of policy evaluation followed
by one iteration of policy improvement
(the ``$\max$'' operator in the update rule).
We are in fact free to intermix any number of
these steps in any order.
The process will converge once the
policy is greedy wrt its own  value function.

\begin{figure}
\centering
\includegraphics[height=1.5in]{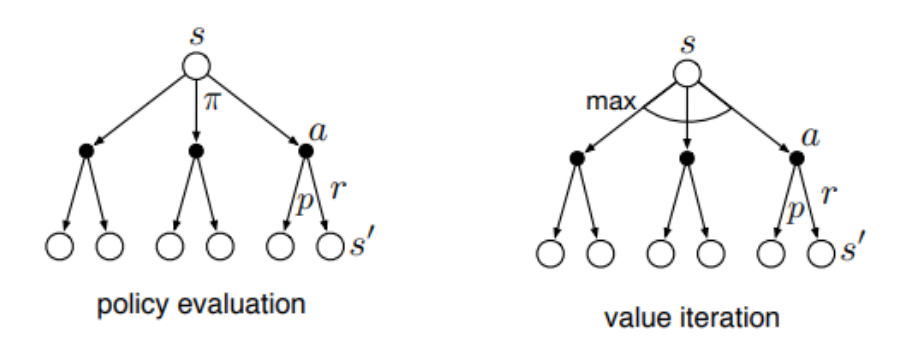}
\caption{
  Policy iteration vs value iteration represented as backup diagrams.
  Empty circles represent states, solid (filled) circles
  represent states and actions.
  \figbased{Figure 8.6 of \citep{Suttonv2}}.
}
\label{fig:sutton-8-6}
\end{figure}

Note that policy evaluation computes $\Vpol$
whereas value iteration computes $\Vopt$.
This difference is illustrated in \cref{fig:sutton-8-6},
using a \keywordDef{backup diagram}.
Here the root node represents any state $s$,
nodes at the next level represent state-action combinations
(solid circles),
and nodes at the leaves representing the set of possible
resulting next state $s'$ for each possible action.
In PE, we average over all actions according
to the policy, whereas in VI, we take the maximum over
all actions.

\section{Value function learning using samples from the world model}

In the rest of this chapter,
we assume the agent only has  access to samples from
the environment, $(s',r) \sim p(s',r|s,a)$.
We will show how to use these samples
to estimate optimal value function and $Q$-function,
even without explicitly knowing the MDP dynamics.
This is sometimes called ``learning'' as opposed
to ``planning'', since the latter requires access
to an explicit world model.

\subsection{Monte Carlo estimation}
\label{sec:MCRL}

Recall that $\Vpol(s) = \expect{\return_t | s_t=s}$
is the sum of expected (discounted) returns from state $s$
if we follow policy $\pi$.
A simple way to estimate this is to rollout the policy,
and then compute the average
sum of discounted rewards.  The trajectory ends when we reach a terminal state, if
the task is episodic, or when the discount factor $\gamma^t$ becomes
negligibly small, whichever occurs first.
This is called \keywordDef{Monte Carlo estimation}.
We can use this to update our estimate of the value function as follows:
\begin{align}
\label{eqn:rl-td}
V(s_t) &\assign V(s_t) + \lr
\left[
 \return_t  - V(s_t)
  \right]
\end{align}
where $\lr$ is the learning rate, and the term in brackets is an error term.
We can use a similar technique to estimate
$\Qpol(s,a) = \expect{\return_t | s_t=s,a_t=a}$
by simply starting the rollout with action $a$.

We can use MC estimation of $Q$,  together with policy iteration
(\cref{sec:policyIteration}), to learn an optimal policy.
Specifically,
at iteration $k$,
we compute a new, improved policy using
$\policy_{k+1}(s) = \argmax_a Q_k(s,a)$,
where $Q_k$ is approximated using MC estimation.
This update can be applied to all the states visited
on the sampled trajectory.
This overall technique
is called \keywordDef{Monte Carlo control}.

To ensure this method converges to the optimal policy,
we need to collect data for every (state, action) pair,
at least in the tabular case,
since there is no generalization
across different values of $Q(s,a)$.
One way to achieve this is to use an $\epsilon$-greedy policy
 (see \cref{sec:epsGreedy}).
Since this is an on-policy algorithm,
the resulting method will converge
to the optimal $\epsilon$-soft policy,
as opposed to the optimal policy.
It is possible to use importance sampling
to estimate the value function for the optimal policy,
even if actions are chosen according to the $\epsilon$-greedy policy.
However, it is simpler to just gradually reduce $\epsilon$.

\subsection{Temporal difference (TD) learning}
\label{sec:TD}

The Monte Carlo (MC) method in \cref{sec:MCRL}
results in 
an estimator for $V(s)$ with very high variance,
since it has to unroll many trajectories, 
whose returns are a sum of many random rewards generated by stochastic state transitions.
In addition, it is limited to episodic tasks
(or finite horizon truncation of continuing tasks),
since it must unroll to the end of the episode
before each update step,
to ensure it reliably estimates the long term return.

In this section, we discuss a more efficient technique called
\keywordDef{temporal difference} or \keywordDef{TD} learning
\citep{Sutton88}.
The basic idea is to incrementally
reduce the Bellman error
for sampled states or state-actions,
based on transitions instead of a long trajectory.
More precisely, suppose we are to learn
the value function $\Vpol$ for a fixed policy $\policy$.
Given a state transition $(s_t,a_t,r_t,s_{t+1})$,
where $a_t \sim \policy(s_t)$,
we change the estimate $V(s_t)$ so that
it moves towards the  \keywordDef{target value}
$\targetV_t = r_t + \gamma V(s_{t+1}) \approx G_{t:t+1}$:
\begin{align}
V(s_t) &\assign V(s_t) + \lr
\left[
  \underbrace{r_t + \gamma V(s_{t+1}) - V(s_t)}_{\delta_t}
  \right]
\end{align}
where $\lr$ is the learning rate.
(See \citep{Ryzhov2015} for ways to adaptively set the learning rate.)
The $\delta_t=y_t - V(s_t)$ term 
is known as the \keywordDef{TD error}.\footnote{
The \keywordDef{Rescorla-Wagner rule}
is a special case of this,
studied in the \keywordDef{classical conditioning}
literature.
The update is
given by $V_{t+1}(s_t) \assign V_t(s_t) + \lr(r_t - V_t(s_t))$,
where $s_t$ is a stimulus and $r_t$ is the response
associated with it.
This is TD with $\gamma=0$.
}

A more general form of TD update
for parametric value function representations is
\begin{align}
\label{eqn:rl-td-approx}
\vw \assign \vw + \lr \left[
r_t + \gamma \Vapprox(s_{t+1})
  - \Vapprox(s_t) \right] \nabla_{\vw}\Vapprox(s_t)
\end{align}
we see that  \cref{eqn:rl-td} is a special case.
The TD update rule for evaluating $\Qpol$ is similar,
except we replace states with states and actions.


It can be shown that TD learning in the tabular case,
\cref{eqn:rl-td}, converges to the correct value function,
under proper conditions~\citep{BertsekasRL}.
However, it may diverge when using nonlinear function approximators,
as we discuss in \cref{sec:deadlytriad}.
The reason is that this update is a
``\keywordDef{semi-gradient}'',
which refers to the fact
that we only take the gradient wrt the  value function,
$\nabla_{\vw} V(\vs_t, \vw_t)$,
treating the target $U_t$ as constant.

\begin{figure}
\centering
\includegraphics[height=1.5in]{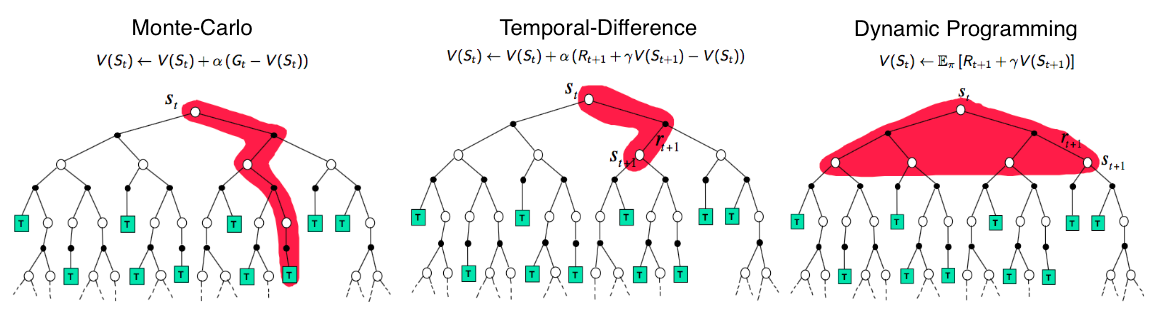}
\caption{
  Backup diagrams of $V(s_t)$ for Monte Carlo,
  temporal difference, and dynamic programming
  updates of the state-value function.
\figthanks{Andy Barto}.
}
\label{fig:TD-MC-DP}
\end{figure}

The potential divergence of TD is also consistent with
the fact that \cref{eqn:rl-td-approx} does not correspond
to a gradient update
on any objective function, despite having
a very similar form to SGD (stochastic gradient descent).
Instead, it is an example of \keywordDef{bootstrapping},
in which the estimate, $\Vapprox(s_t)$,
is updated to approach a target,
$r_t + \gamma \Vapprox(s_{t+1})$,
which is defined by the value function estimate itself.
This idea is shared by DP methods
like value iteration, although they rely on the
complete MDP model to compute an exact Bellman backup.
In contrast, TD learning can be viewed as using
sampled transitions to approximate such backups.
An example of a non-bootstrapping approach is the
Monte Carlo estimation in the previous section.
It samples a complete trajectory,
rather than individual transitions,
to perform an update;
this avoids the divergence issue,
but is  often much less
efficient.
\cref{fig:TD-MC-DP} illustrates the difference between
MC, TD, and DP.

\subsection{Combining TD and MC learning using TD($\lambda$)}
\label{sec:TDlambda}
\label{sec:nstep}

A key difference between TD and MC is
the way they estimate returns.
Given a trajectory
$\traj=(s_0,a_0,r_0,s_1,\ldots,s_T)$,
TD estimates the return from state $s_t$
by one-step lookahead,
$\return_{t:t+1} = r_t + \gamma V(s_{t+1})$,
where the return from time $t+1$ is replaced by
its value function estimate.
In contrast, MC waits until the end of the episode
or until $T$ is large enough,
then uses the estimate
$\return_{t:T} = r_t + \gamma r_{t+1} + \cdots + \gamma^{T-t-1} r_{T-1}$.
It is possible to interpolate between these by
performing an $n$-step rollout, and then using
the value function to approximate the return
for the rest of the trajectory,
similar to heuristic search (\cref{sec:heuristic}).
That is, we can use the \keywordDef{n-step return}
\begin{align}
\return_{t:t+n} = r_{t} + \gamma r_{t+1} + \cdots
+ \gamma^{n-1} r_{t+n-1} + \gamma^n V(s_{t+n})
\end{align}
For example, the 1-step and 2-step returns are given by
\begin{align}
  \return_{t:t+1} &= r_{t} + \gamma v_{t+1} \\
  \return_{t:t+2} &= r_{t} + \gamma r_{t+1} + \gamma^2 v_{t+2} \
  \end{align}
The corresponding $n$-step version of the TD update becomes
\begin{align}
  \vw \assign \vw + \lr
  \left[\return_{t:t+n}  - \Vapprox(s_t) \right]
  \nabla_{\vw}\Vapprox(s_t)
\end{align}
Using this update can help propagate sparse
terminal rewards back through many earlier states.

\begin{figure}
\centering
\includegraphics[height=2in]{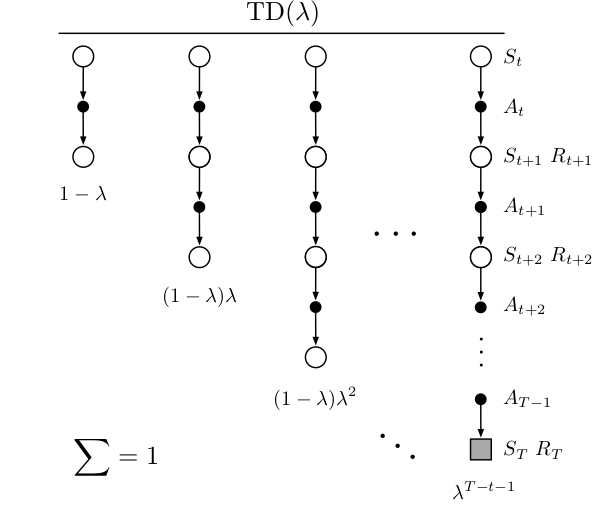}
\caption{
  The backup diagram for TD($\lambda$).
  Standard TD learning corresponds to $\lambda=0$,
  and standard MC learning corresponds to $\lambda=1$.
  \figtaken{Figure 12.1 of \citep{Suttonv2}}.
  \figthanks{Richard Sutton}.
}
\label{fig:TDlambda}
\end{figure}

Rather than picking a specific lookahead value, $n$,
we can take a weighted average of all possible values,
with a single parameter $\lambda\in[0,1]$,
by using
\begin{align}
\label{eqn:rl-gamma-return}
\return_{t}^{\lambda}
\defeq (1-\lambda) \sum_{n=1}^{\infty} \lambda^{n-1} \return_{t:t+n}
\end{align}
This is called the \keywordDef{lambda return}.
Note that these coefficients sum to one
(since $\sum_{t=0}^{\infty} (1-\lambda) \lambda^t = \frac{1-\lambda}{1-\lambda}=1$,
for $\lambda<1$),
so the return is a convex combination of $n$-step returns.
See \cref{fig:TDlambda} for an illustration.
We can now use $\return_t^{\lambda}$ inside the TD update
instead of $\return_{t:t+n}$;
this is called \keywordSpecial{TD$(\lambda)$}{TD(lambda)}.

Note that, if a terminal state is entered at step $T$ (as happens with episodic tasks),
then all subsequent $n$-step returns are equal to the conventional return, $G_t$.
Hence we can write
\begin{align}
  G_t^{\lambda} = (1-\lambda) \sum_{n=1}^{T-t-1}
  \lambda^{n-1} G_{t:t+n}
  \;
  + \lambda^{T-t-1} G_t
\end{align}
From this we can see that if $\lambda=1$, the $\lambda$-return
becomes equal to the regular MC return $G_t$.
If $\lambda=0$, the $\lambda$-return becomes equal to the
one-step return $G_{t:t+1}$ (since $0^{n-1}=1$ iff $n=1$),
so standard TD learning is often called
\keywordDef{TD(0) learning}.
This episodic form also gives us the following recursive equation
\be
G_t^{\lambda} = r_t + \gamma[(1-\lambda) v_{t+1} + \lambda G_{t+1}^{\lambda}]
\ee
which we initialize with $G_T=v_t$.

\subsection{Eligibility traces}
\label{sec:eligibility}

The TD($\lambda)$ update in 
in \cref{eqn:rl-gamma-return}
requires summing over future rewards,
which cannot be done in an online way.
Fortunately it is possible to derive
a backwards-looking version of TD($\lambda$)
learning that is fully online.
The technique relies on incrementally
compute the 
\keywordDef{eligibility trace},
which is a weighted sum of the gradients
of the value function:
\be
\vz_t = \gamma \lambda \vz_{t-1} + \nabla_{\vw} V_{\vw}(s_t)
\label{eqn:eligibilityDecay}
\ee
(This trace term gets reset to 0 at the start of each episode.)
The online TD($\lambda$) update rule becomes
\be
\vw_{t+1} = \vw_t + \lr \delta_t \vz_t
\label{eqn:eligiblityUpdate}
\ee
See \citep{VanSeijen2016} for more details.

\section{SARSA: on-policy TD policy learning}
\label{sec:SARSA}

TD learning is for policy evaluation,
as it estimates the value function for a fixed policy,
i.e., it computes $V^{\pi}$.
In order to find an optimal policy, $\pi^*$,
we may use the algorithm as a building block inside
generalized policy iteration (\cref{sec:policyIteration}).
In this case, it is more convenient
to work with the action-value function, $Q$, and
a policy $\policy$ that is greedy with respect to $Q$.
The agent follows $\policy$ in every step
to choose actions, and upon a transition $(s,a,r,s')$
the TD update rule is
\begin{align}
Q(s,a) \assign Q(s,a) + \lr \left[ r + \gamma Q(s',a') - Q(s,a) \right]
\label{eqn:rl-td-q}
\end{align}
where $a' \sim \policy(s')$ is the action
the agent will take in state $s'$.
This converges to $Q^{\pi}$.
After $Q$ is updated (for policy evaluation),
$\policy$ also changes accordingly so it is greedy
with respect to $Q$ (for policy improvement).
This algorithm, first proposed by \citep{Rummery1994},
was further studied and renamed to
\keywordDef{SARSA} by \citep{Sutton1996};
the name comes from its update rule that
involves an augmented transition $(s,a,r,s',a')$.

\subsection{Convergence}

In order for SARSA to converge to $\Qopt$,
every state-action pair must be visited infinitely often,
at least in the tabular case,
since the algorithm only updates $Q(s,a)$
for $(s,a)$ that it visits.
One way to ensure this condition is to use a
``greedy in the limit with infinite exploration''
(\keywordDef{GLIE}) policy.
An example is the $\epsilon$-greedy policy,
with $\epsilon$ vanishing to $0$ gradually.
It can be shown that SARSA with a GLIE policy will
converge to $\Qopt$ and $\polopt$~\citep{Singh2000}.

\subsection{Sarsa($\lambda$)}
\label{sec:sarsaLambda}

It is possible to apply the eligibility trace idea
to Sarsa, since it is an on-policy method.
This can help speedup learning in sparse reward scenarios.

The basic idea, in the tabular case, is as follows.
We compute an  eligibility  for each state action pair,
denoted $N(s,a)$, representing the visit count.
Following \cref{eqn:eligiblityUpdate},
we perform the update
\be
Q(s,a) \leftarrow Q(s,a) + \lr \delta N(s,a)
\ee
where the TD error is
\be
\delta = r + \gamma Q(s', a') - Q(s,a)
\ee
Then, following \cref{eqn:eligibilityDecay},
we decay all the visit counts (traces)
using
\be
N(s,a) \leftarrow \gamma \lambda N(s,a)
\ee
This is called Sarsa($\lambda$).

\section{Q-learning: off-policy TD policy learning}
\label{sec:Qlearning}
\label{sec:Q-learning}


SARSA is an \keyword{on-policy} algorithm,
which means it learns the $Q$-function for the policy
it is currently using,
which is typically not the optimal policy,
because of the need to perform exploration.
However, with a simple modification,
we can convert this to an \keyword{off-policy}
algorithm that learns $\Qopt$,
even if a suboptimal or exploratory policy is used to choose actions.

\subsection{Tabular Q learning}

Suppose we modify SARSA by replacing the sampled next action
$a' \sim \policy(s')$ in \cref{eqn:rl-td-q}
with a greedy action:
$a' = \argmax_b Q(s',b)$.
This results in the following update
when a transition $(s,a,r,s')$ happens
\begin{align}
Q(s,a) \assign Q(s,a) + \lr \left[
  r + \gamma \max_{a'} Q(s',a') - Q(s,a) \right]
\label{eqn:Qlearning}
\end{align}
This is the update rule of \keywordDef{Q-learning}
for the tabular case~\citep{Watkins92}.

Since it is off-policy,
the method can use  $(s,a,r,s')$ triples
coming from any data source,
such as older versions of the policy,
or log data from an existing (non-RL) system.
If every state-action pair is visited infinitely often,
the algorithm provably converges to $\Qopt$
in the tabular case, with properly decayed learning rates~\citep{BertsekasRL}.
\cref{algo:Qlearning} gives a vanilla implementation of
Q-learning
with $\epsilon$-greedy exploration.

\begin{algorithm}
\dontprintsemicolon
\caption{Tabular Q-learning with $\epsilon$-greedy exploration}
\label{algo:Qlearning}
Initialize value function $Q$ \\
\Repeat{converged}
       {
       Sample starting state $s$ of new episode \\
       \Repeat{state $s$ is  terminal}
       {
       Sample action
       $a=\begin{cases}
       \argmax_{b} Q(s,b), & \text{with probability $1-\epsilon$} \\
       \text{random action}, & \text{with probability $\epsilon$}
       \end{cases}$
               \\
       $(s',r) = \text{env.step}(a)$ \\
       Compute the TD error: $\delta = r + \gamma \max_{a'} Q(s',a') - Q(s,a)$ \\
       $Q(s,a) \leftarrow Q(s,a) + \lr \delta$ \\
       $s \leftarrow s'$ 
         }
}
\end{algorithm}

\eat{
\begin{algorithm}
\dontprintsemicolon
\caption{Q-learning with $\epsilon$-greedy exploration}
\label{algo:Qlearning}
Initialize value function parameters $\vw$ \\
\Repeat{converged}
       {
       Sample starting state $s$ of new episode \\
       \Repeat{state $s$ is  terminal}
       {
       Sample action
       $a=\begin{cases}
       \argmax_{b} \Qapprox(s,b), & \text{with probability $1-\epsilon$} \\
       \text{random action}, & \text{with probability $\epsilon$}
       \end{cases}$
               \\
       Observe state $s'$, reward $r$ \\
       Compute the TD error: $\delta = r + \gamma \max_{a'} \Qapprox(s',a') - \Qapprox(s,a)$ \\
       $\vw \assign \vw + \lr \delta \nabla_{\vw} \Qapprox(s,a)$ \\
           $s \assign s'$
         }
}
\end{algorithm}
}


For terminal states, $s \in \calS^+$, we know that
$Q(s,a)=0$ for all actions $a$.
Consequently, for the optimal value function,
we have
$V^*(s) = \max_{a'} Q^*(s,a)=0$
for all terminal states.
When  performing online learning, we don't usually know
which states are terminal.
Therefore we assume that, whenever we take a step in the environment,
we get the next state $s'$ and reward $r$,
but also a binary indicator $\done(s')$ that tells us
if $s'$ is terminal.
In this case,  we set the target value in Q-learning
to $V^*(s')=0$ yielding the modified update rule:
\begin{align}
Q(s,a) \assign Q(s,a) + \lr \left[
  r + (1-\done(s')) \gamma \max_{a'} Q(s',a') - Q(s,a) \right]
\label{eqn:Qlearningdone}
\end{align}
For brevity, we will usually ignore this factor in the subsequent equations,
but it needs to be implemented in the code.


\begin{figure}
\centering
\includegraphics[height=5in]{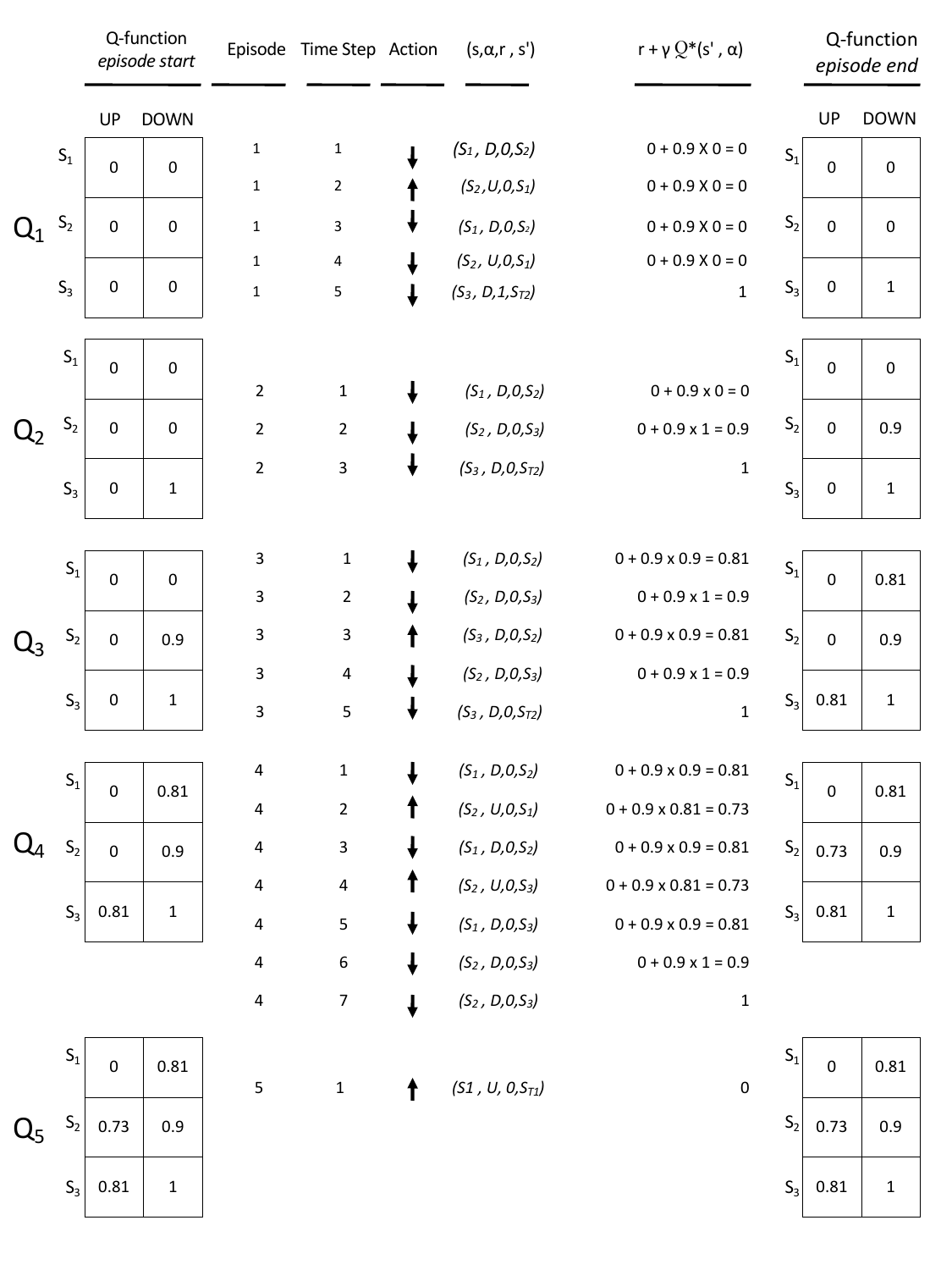}
\caption{
  Illustration of Q learning for one random trajectory
  in the 1d grid world
  in \cref{fig:Q1d} using $\epsilon$-greedy exploration.
  At the end of episode 1, we make a transition
  from $S_3$ to $S_{T2}$ and get a reward of $r=1$,
  so we estimate $Q(S_3,\downarrow)=1$.
  In episode 2, we make a transition from $S_2$ to $S_3$,
  so $S_2$ gets incremented by $\gamma Q(S_3,\downarrow)=0.9$.
  \figbased{Figure 3.3 of \citep{Graesser2019}}.
}
\label{fig:DRLTD}
\end{figure}

\cref{fig:DRLTD} gives an example
of Q-learning applied to the simple 1d grid world
from \cref{fig:Q1d}, using $\gamma=0.9$.
We show the $Q$-functon at the start and end of each episode,
after performing actions chosen by an $\epsilon$-greedy policy.
We initialize $Q(s,a)=0$ for all entries,
and use a step size of $\lr=1$.
At convergence, we have
$\Qopt(s,a) = r + \gamma \Qopt(s',a_*)$,
where $a_* = \downarrow$ for all states.

\subsection{Q learning with function approximation}
\label{sec:Qfn}

To make Q learning work with high-dimensional state spaces,
we have to replace the tabular (non-parametric) representation
with a parametric approximation, denoted $Q_{\vw}(s,a)$.
We can update this function using one or more steps of SGD
on the following loss function
\begin{align}
  \loss(\vw|\vs,a,r,\vs') &=
  \big( (r + \gamma \max_{a'} Q_{\vw}(\vs',a')) -   Q_{\vw}(\vs,a) \big)^2  
  \label{eqn:ynaive}
\end{align}
Since nonlinear functions need to be trained on minibatches
of data, we compute the average loss over multiple
randomly sampled
experience tuples (see \cref{sec:ER} for discussion)
  to get
  \begin{align}
\loss(\vw) &= \expectQ{\loss(\vw|\vs,a,r,\vs')}{(\vs,a,r,\vs') \sim U(\data)}
  \end{align}
See \cref{algo:Qfn} for the pseudocode.

\begin{algorithm}
\dontprintsemicolon
\caption{Q learning with function approximation and replay buffers}
\label{algo:Qfn}
Initialize environment state $\vs$,
network parameters $\vw_0$,
replay buffer $\data=\emptyset$,
discount factor $\gamma$,
step size $\eta$,
policy
$\pi_0(a|s) = \epsilon \text{Unif}(a)
+ (1-\epsilon) \delta(a=\argmax_a Q_{\vw_0}(s,a))$ \\
\For{iteration $k=0,1,2,\ldots$}
    {
      \For{environment step $s=0,1,\ldots,S-1$}
          {
            Sample action: $a \sim \pi_k(a|s)$ \\
            Interact with environment: $(s',r) = \text{env.step}(a)$ \\
            Update buffer: $\data \leftarrow \data \union \{ (s,a,s',r) \}$ 
          }
          $\vw_{k,0} \leftarrow \vw_k$ \\
          \For{gradient step $g=0,1,\ldots,G-1$}
              {
                Sample batch: $B \subset \data$ \\
                Compute error: $\loss(B,\vw_{k,g})
                = \frac{1}{|B|} \sum_{(s,a,r,s') \in B}
                \left[ Q_{\vw_{k,g}}(s,a) - (r + \gamma \max_{a'} Q_{\vw_k}(s',a')) \right]^2$\\
                Update parameters:
                $\vw_{k,g} \leftarrow \vw_{k,g} - \lr \nabla_{\vw_{k,g}}
                \loss(B,\vw_{k,g})$
              }
         $\vw_{k+1} \leftarrow \vw_{k,G}$ 
      }
\end{algorithm}

\subsubsection{Neural fitted Q}
\label{sec:NFQ}

The first approach of this kind
is known as
\keywordDef{fitted Q evaluation} (or \keywordDef{FQE})
\citep{Ernst2005},
which was extended in  \citep{Riedmiller2005}
to use neural networks.
This  corresponds to fully optimizing
$\loss(\vw)$ at each iteration (equivalent to using $G=\infty$
gradient steps).

\subsubsection{DQN}
\label{sec:DQN}

The influential
\keyword{deep Q-network} or \keyword{DQN}
paper of \citep{Mnih2015atari}
also used neural nets to represent the $Q$ function,
but performed a smaller number of gradient updates
per iteration.
Furthermore, they proposed to modify the target value
when fitting the $Q$ function in order to avoid
instabilities during training (see \cref{sec:deadly} for details).

The DQN method became famous since it was able  to train
agents that can outperform
humans when playing various Atari games from
the \keywordDef{ALE} (Atari Learning Environment)
benchmark \citep{Bellemare13}.
Here the input is a small color image,
and the action space corresponds to moving left, right, up or down,
plus an optional shoot action.\footnote{
For more discussion of ALE,
see \citep{Machado2018},
and for a recent extension to continuous actions
(representing joystick control),
see the CALE benchmark of \citep{Farebrother2024CALE}.
Note that DQN was not the first deep RL method to train
an agent from pixel input;
that honor goes to \citep{Lange2010},
who trained an autoencoder to embed images into low-dimensional latents,
and then used neural fitted Q learning (\cref{sec:NFQ})
to fit the $Q$ function.
}

Since 2015, many more extensions to DQN have been proposed,
with the goal of
improving performance in various ways,
either in terms of peak reward obtained,
or sample efficiency (e.g., reward obtained after only 100k steps
in the environment, as proposed in the \keywordDef{Atari-100k} benchmark
\citep{Atari100k}\footnote{
The Atari-100k benchmark only includes 26 out of 46 games of the
ALE that were determined to be
``solvable by state-of-the-art model-free deep RL algorithms''
at the time of the benchmark’s creation in 2019.
This excludes games   like Montezuma’s Revenge,
which require more exploration and hence 
more training data.
}),
or training stability,
or all of the above.
We discuss some of these extensions in \cref{sec:DQNextensions}.

\eat{
\subsubsection{DQN}
\label{sec:DQN}

In this section, we discuss the seminal
\keywordDef{deep Q-network} or \keywordDef{DQN}
paper \citep{Mnih2015atari}.
The starting point is to consider the update
for the Q network parameters $\vw$
shown in \cref{algo:Qlearning}.
The gradient update corresponds to taking the gradient
of the following loss
\begin{align}
  \loss(\vw|\vs,a,r,\vs') &=
  \big( (r + \gamma \max_{a'} Q_{\vw}(\vs',a')) -   Q_{\vw}(\vs,a) \big)^2  
  \label{eqn:ynaive}
\end{align}
Since neural nets work best when trained on minibatches
of data, we compute the average loss by sampling
experience tuples (e.g., uniformly at random)
from
  the replay buffer $\data$
  (see \cref{sec:ER})
  to get
  \begin{align}
\loss(\vw) &= \expectQ{\loss(\vw|\vs,a,r,\vs')}{(\vs,a,r,\vs') \sim U(\data)}
  \end{align}
The random sampling helps reduce correlation between the samples.

\begin{algorithm}
\dontprintsemicolon
\caption{DQN (with target network)}
\label{algo:DQNtarget}
Initialize environment state $\vs$,
network parameters $\vw$,
target parameters $\overline{\vw} = \stopgrad(\vw)$,
replay buffer $\data=\emptyset$,
discount factor $\gamma$,
EMA rate $\rho$,
step size $\eta$
\\
\Repeat{converged}
       {
         Take action $a \sim \text{eps-greedy}(\vw)$\\
         $(\vs',r) = \text{step}(a, \vs)$ \\
         $\data := \data \union
         \{ (\vs, a, r, \vs') \}$ \\
         $\vs \assign \vs'$ \\
         Sample a minibatch $\calB = \{(\vs_j,a_j,r_j,\vs'_j)\}$
         from $\data$ \\
           $(\vw,\overline{\vw}) = \text{update}(\vw, \overline{\vw}, \calB)$
        }
.\\
$\text{def update}(\vw,\overline{\vw},\calB)$: \\
       Let $(\vs_j,a_j,r_j,\vs'_j)_{j=1}^B = \calB$ \\
$\targetV_{j} = \TargetV(r_j, \vs'_j; \overline{\vw})$ for $j=1:B$ \\
      $\loss(\vw) = \frac{1}{|\calB|} \sum_{(\vs_j, a_j, r_j, \vs'_j) \in
        \calB} (Q_{\vw}(\vs_j,a_j) - \stopgrad(\targetV_j))^2$\\
      $\vw \assign \vw - \lr_{\vw} \nabla \loss(\vw)$ // Gradient descent step \\
      $\overline{\vw} := \rho \overline{\vw} 
      + (1-\rho) \vw$       //EMA for  target network \\
    Return $\vw, \overline{\vw}$\\
\end{algorithm}
}

\subsubsection{Experience replay}
\label{sec:ER}
\label{sec:replay}

Since Q learning is an off-policy method, we can update the Q function
using any data source. This is particularly important when we use
nonlinear function approximation (see \cref{sec:Qfn}), which often needs a lot of data
for model fitting.
A natural source of data is data collected earlier in the trajectori
of the agent; this is called 
an \keywordDef{experience replay} buffer,
which stores  $(s,a,r,s')$ transition tuples into a buffer.
This can improve the stability and sample efficiency of learning,
and was
originally proposed in \citep{Lin1992}.

This modification has two advantages.
First, it improves data efficiency as every transition
can be used multiple times.
Second, it improves stability in training,
by reducing the correlation of the data samples
that the network is trained on,
since the training tuples do not have to come from
adjacent moments in time.
(Note that experience replay requires the use
of off-policy learning methods, such as Q learning,
since the training data is sampled from older
versions of the policy, not the current policy.)

\subsubsection{Prioritized experience replay}
\label{sec:PER}

It is possible to replace the uniform sampling
from the buffer
with one that favors more
important transition tuples
that may be more informative about $Q$.

To explain this, we start by discussing  the \keywordDef{prioritized sweeping}
method of \citep{Moore1993},
which was developed for discrete state spaces
using a priority queue.
The idea is as follows.
Whenever we update the value of a state $V(s)$,
we iterate over all state-action pairs
$(s^-,a^-)$ that can immediately transition
into $s$ (this requires knowing the world model).
The priority of any such $s^-$ is then increased
to $\calT(s|s^-,a^-) \times |V(s)-V^{\old}(s)|$,
where $V^{\old}(s)$ is the value before the update.
Thus we prioritize updating states which are likely to
have lead to states whose values have changed by a lot.

This can be generalized to the non-tabular
experience replay setting
as described in 
\citep{Schaul2016},
who call the technique
\keywordDef{prioritized experience replay}.
\eat{
In particular,  we  sample transitions from $\data$
with probability
$p(s,a,r,s') \propto (|\delta| + \varepsilon)^{\eta}$,
where $\delta$ is the corresponding TD error
(under the current $Q$-function),
$\varepsilon > 0$ a hyperparameter
to ensure every experience is chosen
with nonzero probability,
and $\eta \geq 0$ controls the ``inverse temperature''
of the distribution (so $\eta=0$ corresponds to uniform sampling).
}
Consider the TD error for the $i$'th tuple $\tau_i$
\be
\delta_i = r_i + \gamma \max_{a'} Q_{\overline{\vw}}(s'_i, a')
 - Q_{\vw}(s_i,a_i)
\ee
Define the priority of $i$ as
\be
p_i = (\delta_i + \epsilon)^{\alpha}
\ee
where $\alpha \geq 0$ determines the degree of prioritization,
with $\alpha=0$ corresponding to no prioritization (uniform sampling).
Now define the probability of sampling $i$ as
\be
P(i) = \frac{p_i}{\sum_k p_k}
\ee

\eat{
Sampling from this distribution will introduce bias relative
to the uniform distribution over the past $M$ samples in the replay buffer.
But we can correct this using importance sampling, as follows:
\begin{align}
  \expectQ{\loss(\tau)}{\text{Uniform}(\tau)}
  &= \sum_{i=1}^M \frac{1}{M} \loss(\tau_i) \\
  &= \sum_{i=1}^M P(\tau_i) \frac{1}{M P(\tau_i)} \loss(\tau_i) \\
  &= \expectQ{w(\tau) \loss(\tau)}{P(\tau)} 
  \end{align}
where we define the importance weight as
\be
w(\tau) = \left(\frac{1}{M P(\tau)} \right)^{\beta(t)}
\ee
Here $\beta(t)$ is a hyperparameter than starts off slightly larger than
0, to ensure that important experiences are not down-weighted too much,
and then is gradually increased to 1,
which results in an unbiased estimate.

}

\eat{
\subsubsection{Word of caution}

Unfortunately, when the function is nonlinear,
Q learning can become unstable; see \cref{sec:deadly} for details
of the problem and some solutions.
}

\subsubsection{The deadly triad}
\label{sec:offpolicyrl-deadlytriad}
\label{sec:deadlytriad}
\label{sec:deadlyTriad}
\label{sec:deadly}

\begin{figure}
\centering
\begin{subfigure}[b]{0.55\textwidth}
\centering
\includegraphics[height=1.75in]{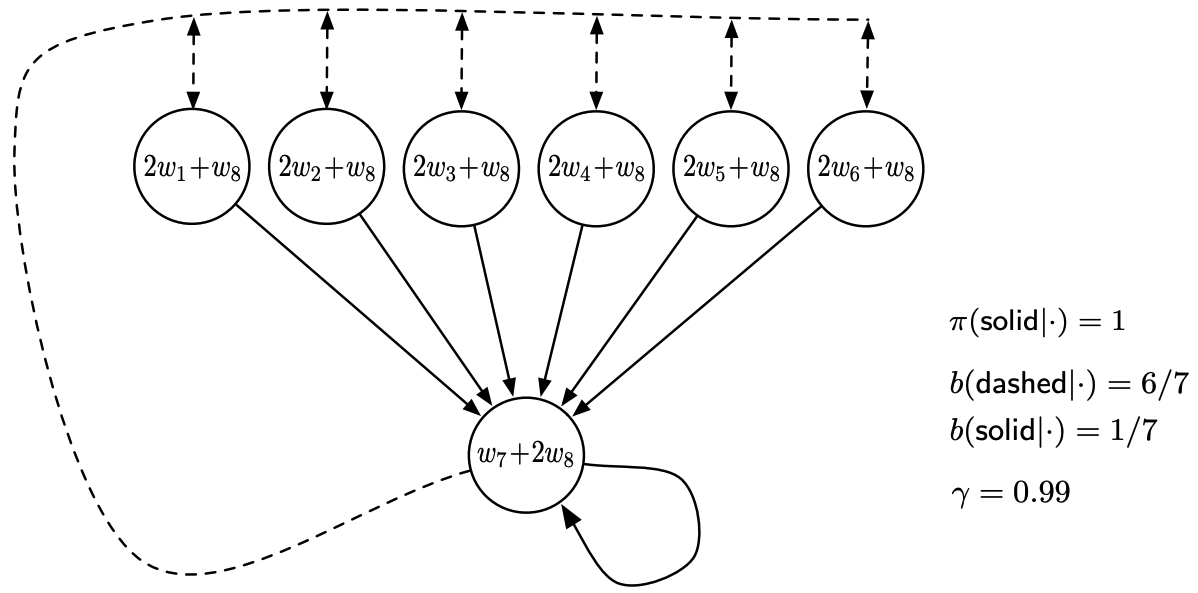}
\caption{ }
\label{fig:baird-example-mdp}
\end{subfigure}
~
\begin{subfigure}[b]{0.35\textwidth}
\centering
\includegraphics[height=1.75in]{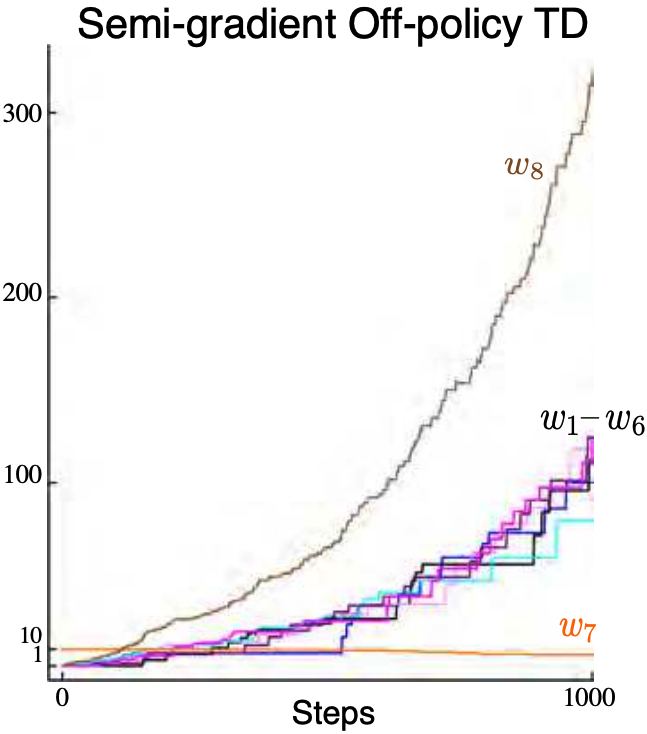}
\caption{ }
\label{fig:baird-example-divergence}
\end{subfigure}
\caption{
(a) A simple MDP.
  (b) Parameters of the policy diverge over time.
\figtaken{Figures 11.1 and 11.2 of \citep{Suttonv2}}.
\figthanks{Richard Sutton}.
}
\label{fig:baird-example}
\end{figure}

The problem with the naive Q learning  objective in \cref{eqn:ynaive}
  is that it can lead to instability,
since the target we are regressing towards
uses the same parameters $\vw$ as the function we are updating.
So the network is ``chasing its own tail''.
Although this is fine for tabular models,  it can fail
for nonlinear models, as we discuss below.

In general, an RL algorithm can become unstable when it has
these three components:
function approximation (such as neural networks),
bootstrapped value function estimation (i.e., using TD-like methods instead of MC),
and off-policy learning (where the actions are sampled from some distribution
other than the policy that is being optimized).
This combination is known as \keywordDef{the deadly triad}
\citep{Sutton2015, Vanhasselt18}).

A classic example of this is the simple MDP depicted
in \cref{fig:baird-example-mdp}, due to \citep{Baird95}.
(This is known as \keywordDef{Baird's counter example}.)
It has 7 states and 2 actions.
Taking the dashed action takes the environment
to the 6 upper states uniformly at random,
while the solid action takes it to the bottom state.
The reward is 0 in all transitions,
and $\gamma=0.99$.
The value function $\Vapprox$ uses
a linear parameterization indicated by the expressions
shown inside the states, with $\vw\in\real^8$.
The target policies $\policy$ always chooses the
solid action in every state.
Clearly, the true value function, $\Vpol(s) = 0$,
can be exactly represented by setting $\vw=\vzero$.

Suppose we use a behavior policy $b$ to generate
a trajectory,
which chooses the dashed and solid actions
with probabilities $6/7$ and $1/7$, respectively,
in every state.
If we apply TD(0) on this trajectory,
the parameters diverge to $\infty$
(\cref{fig:baird-example-divergence}),
even though the problem appears simple.
In contrast, with on-policy data
(that is, when $b$ is the same as $\policy$),
TD(0) with linear approximation can be guaranteed to
converge to a good value function approximate~\citep{Tsitsiklis97}.
The difference is that with on-policy learning,
as we improve the value function, we also improve the policy,
so the two become self-consistent,
whereas with off-policy learning,
the behavior policy may not match the optimal
value function that is being learned,
leading to inconsistencies.

The divergence behavior is demonstrated in
many value-based bootstrapping methods, including TD, Q-learning,
and related approximate
dynamic programming algorithms,
where the value function is represented
either linearly (like the example above)
or nonlinearly~\citep{Gordon95,Tsitsiklis1997,Ostrovski2021}.
The root cause of these divergence phenomena
is that bootstrapping methods
typically are not minimizing a fixed objective function.  Rather, they
create a learning target using their own estimates, thus potentially
creating a self-reinforcing loop to push the estimates to infinity.
More formally, the problem is that
the contraction property in the tabular case
(\cref{eqn:bellman-contraction})
may no longer hold when $V$ is approximated by $\Vapprox$.

We discuss some solutions to the deadly triad problem below.

\subsubsection{Target networks}
\label{sec:targetNetwork}

One heuristic solution to the deadly triad,
proposed in the DQN paper,
is to use a  ``frozen'' 
\keywordDef{target network} computed at an earlier iteration
to define the target value for the DQN updates,
rather than trying to chase a constantly moving target.
Specifically, we maintain an extra copy
the $Q$-network, $Q_{\overline{\vw}}$, 
with the same structure as $\Qapprox$.
This new $Q$-network is used
to compute bootstrapping targets
\be
\TargetV(r,\vs'; \overline{\vw}) =
r + \gamma \max_{a'} Q_{\overline{\vw}}(\vs',a')
\ee
for training $\Qapprox$.
We can periodically set $\overline{\vw} \assign \stopgrad(\vw)$,
usually after a few episodes,
where the stop gradient operator
is used to prevent autodiff propagating gradients back to $\vw$
(this is called a \keywordDef{detached target}).
Alternatively, we can use
an \keywordDef{exponential moving average}
or \keywordDef{EMA}
of the weights,
i.e.,
at step $k$ we use 
$\overline{\vw}_k = \rho \overline{\vw}_{k-1} + (1-\rho) \vw_k$,
where the forgetting factor 
$\rho \approx 0.999$
ensures that $Q_{\overline{\vw}}$ slowly catches
up with $Q_{\vw}$.\footnote{
Mathematically, if we have a loss of the form
$\loss(\theta,\phi) = L(f(\theta) - t(\phi))$,
where $f$ is the predictor and $t$ is the target,
then using a stop gradient amounts to taking the gradient
of $\loss$ wrt $\theta$ and evaluating the result at $\phi=\theta$.
This is the same as using the EMA target $\overline{\phi}$
with $\rho=0$.
}
The final loss  has the form
  \begin{align}
    \loss(\vw) &= \expectQ{\loss(\vw|\vs,a,r,\vs')}{(\vs,a,r,\vs') \sim U(\data)} \\
      \loss(\vw|\vs,a,r,\vs') &=
  (\TargetV(r,\vs';\overline{\vw}) -   Q_{\vw}(\vs,a))^2  
  \end{align}
Theoretical work justifying this technique is given
in  \citep{Fellows2023,Che2024}.

\eat{
\begin{algorithm}
\dontprintsemicolon
\caption{DQN (with target network)}
\label{algo:DQN}
Initialize environment state $\vs$,
network parameters $\vw$,
target parameters $\overline{\vw} = \stopgrad(\vw)$,
replay buffer $\data=\emptyset$,
discount factor $\gamma$,
EMA rate $\rho$,
step size $\eta$
\\
\Repeat{converged}
       {
         Take action $a \sim \text{eps-greedy}(\vw)$\\
         $(\vs',r) = \text{step}(a, \vs)$ \\
         $\data := \data \union
         \{ (\vs, a, r, \vs') \}$ \\
         $\vs \assign \vs'$ \\
         Sample a minibatch $\calB = \{(\vs_j,a_j,r_j,\vs'_j)\}$
         from $\data$ \\
           $(\vw,\overline{\vw}) = \text{update}(\vw, \overline{\vw}, \calB)$
        }
.\\
$\text{def update}(\vw,\overline{\vw},\calB)$: \\
       Let $(\vs_j,a_j,r_j,\vs'_j)_{j=1}^B = \calB$ \\
$\targetV_{j} = \TargetV(r_j, \vs'_j; \overline{\vw})$ for $j=1:B$ \\
      $\loss(\vw) = \frac{1}{|\calB|} \sum_{(\vs_j, a_j, r_j, \vs'_j) \in
        \calB} (Q_{\vw}(\vs_j,a_j) - \stopgrad(\targetV_j))^2$\\
      $\vw \assign \vw - \lr_{\vw} \nabla \loss(\vw)$ // Gradient descent step \\
      $\overline{\vw} := \rho \overline{\vw} 
      + (1-\rho) \vw$       //EMA for  target network \\
    Return $\vw, \overline{\vw}$\\
\end{algorithm}
}

\subsubsection{Gradient TD methods}

A general way to ensure convergence in off-policy
learning is to construct an objective function,
the minimization of which leads to a good value function approximation.
This is the basis of the
\keywordDef{gradient TD method} of \citep{Sutton2008,Maei2009,Ghiassian2020}.

\subsubsection{Two time-scale methods}

Another approach is to update the target value in the TD update
more quickly than the value function itself;
this is known as a \keywordDef{two timescale optimization}
(see e.g., \citep{Yu2017TD,Zhang2019timescale,Hong2023}).

\eat{
It is also possible to use a standard single timescale
method provided the target value is computed using a 
frozen \keyword{target network},
as discussed in \cref{sec:targetNetwork}.
See \citep{Fellows2023,Che2024} for details.
}

\subsubsection{Layer norm}

\begin{figure}
\centering
\includegraphics[height=1.5in]{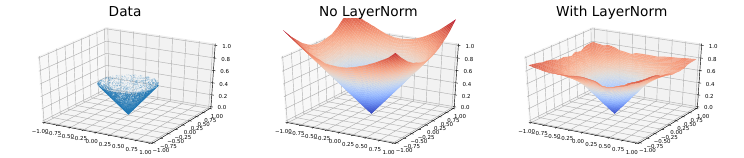}
\caption{
  We generate a dataset (left) with inputs $\vx$
  distributed in a circle with radius 0.5 and labels $y = ||\vx||$.
  We then fit a two-layer MLP without LayerNorm (center)
  and with LayerNorm (right).
  LayerNorm bounds the values and prevents catastrophic overestimation
  when extrapolating.
  \figtaken{Figure  3 of \citep{Ball2023}}.
  \figthanks{Philip Ball}.
}
\label{fig:layerNorm}
\end{figure}

More recently, \citep{PQN} proved that
nonlinear TD learning
can be made to converge, even in the off-policy setting,
if three simple conditions on the critic ($Q$ network) are satisfied:
the final layer weights 
are bounded (e.g., using $\ell_2$ normalization,
or using AdamW, which
has been shown to solve an $L_{\infty}$
constrained optimization problem,
ensuring that model parameters remain bounded by $1/\lambda$
\citep{Xie2024AdamW,Pethick2025});
the penultimate layer is sufficiently wide to represent the true function;
and the input to the critic has bounded norm
(e.g., using   LayerNorm \citep{Ba2016},
or RMSNorm\footnote{
RMSNorm is a simplification of LayerNorm that omits the mean
centering step, and has been found to be faster,
especially in large models like transformers.
In more detail, LayerNorm
computes $y=\frac{x-E[x]}{\sqrt{\Var{x}+\epsilon}} \gamma + \beta$,
whereas RMSNorm just computes
 $y=\frac{x}{\frac{1}{n} \sqrt{\sum_{i=1}^n x_i^2+\epsilon}} \gamma$.
}).
In particular, suppose the network has the form
$Q(s,a|\vw,\vtheta) = \vw^T \relu(\text{LayerNorm}(f(s,a;\vtheta)))$.
Since $||\text{LayerNorm}(f(s,a;\vtheta))|| \leq 1$, we have
$||Q(s,a|\vw,\vtheta) \leq ||\vw||$,
which means the magnitude of the output is always bounded
(assuming $||\vw||$ is also bounded),
as shown in \cref{fig:layerNorm}.
In \citep{PQN}, they prove that this
is sufficient to ensure convergence 
of the value function estimate.

\subsubsection{Other methods}

A variety of other solutions
to the deadly triad have been proposed,
including the ``chaining value functions'' approach
of \citep{Schmitt2022},
a combination of target networks
and over-parameterized linear function approximation
\citep{Che2024},
etc.

\eat{

  Thanks for asking! First, the deadly triad is offpolicy data + general function + TD. We can break the deadly triad by either use optimization based method to replace TD (e.g., LP-based RL with primal-dual solver https://arxiv.org/pdf/2001.01866), or use overparametrized general function (https://arxiv.org/abs/2405.21043). Second, linear value function with appropriate basis is also powerful. We had a series of work on developing the appropriate linear basis (representation) for RL (https://arxiv.org/pdf/2208.09515). Third, for convex Q function, I do not have general answer for general dynamics. However, for the multi-stage stochastic optimization  dynamic problem (https://arxiv.org/pdf/2112.00874), the Q is proved to be convex and TD converges.

  We had this paper on convex RL: https://openreview.net/pdf?id=ELndVeVA-TR, in this case we focus on value functions that are convex in the occupancy (while the value function is linear in it). These functions are quite popular for exploration/diversity/imitation so this convex setting is actually rich and meaningful. The problems are almost equivalent in terms of hardness results.

  TD with linear value functions also suffers from the deadly triad in the off-policy case, though gradient TD (GTD) algorithms exist that do converge (in addition to alternatives Bo Dai mentioned above).

  Non-linear TD can additionally diverge even in the *on-policy* case. See, e.g., this classic paper by John Tsitsiklis and  @Benjamin Van Roy.
[Tsitsiklis1997]
  I conjecture that convexity in the parameters will not be sufficient to avoid this divergence.  If useful, I could try to construct a simple example to demonstrate this.  (I haven't checked carefully, maybe the example in the paper above is already convex.)

  However, for some problems even non-linear TD is guaranteed to converge.  For instance, this paper by Yann Ollivier
[Ollivier2018]
  that shows non-linear TD converges when the MDP is reversible, because then TD can be interpreted as stochastic gradient descent on a different objective (and therefore then inherits standard convergence properties of SGD).

Strict divergence does not always seem to happen, btw, when using a different optimiser, like ADAM, because this bounds the parameter updates.  This doesn't solve the issue, because the values can still get increasingly inaccurate when using an unstable update (we've called this phenomenon 'soft divergence' in the past).
}

\subsection{Maximization bias}

\begin{figure}
\centering
\includegraphics[height=1.5in]{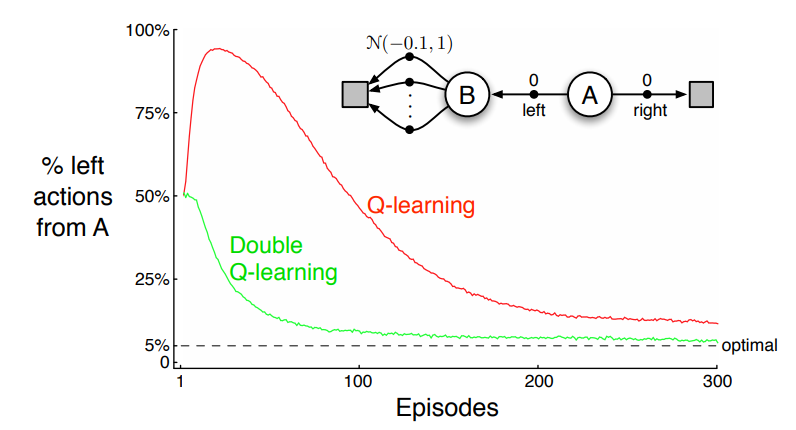}
\caption{
  Comparison of Q-learning and double Q-learning on a simple episodic
  MDP using $\epsilon$-greedy action selection with $\epsilon=0.1$.
  The initial state is A, and squares denote absorbing states.
The data are averaged over 10,000 runs.
\figtaken{Figure 6.5 of \citep{Suttonv2}}.
\figthanks{Richard Sutton}.
}
\label{fig:sutton-6-5}
\end{figure}

Standard Q-learning suffers from a problem
known as the \keywordDef{optimizer's curse} \citep{Smith2006},
or the \keywordDef{maximization bias}.
The problem refers to the simple statistical inequality:
$\expect{\max_a X_a} \geq \max_a \expect{X_a}$,
for a set of random variables $\{X_a\}$.
Thus, if we pick actions greedily according to their random scores $\{X_a\}$, we might pick a wrong action just because random noise makes it appealing.

\cref{fig:sutton-6-5} gives a simple example
of how this can happen in an MDP.
The start state is A.
The right action gives a reward 0 and terminates the episode.
The left action also gives a reward of 0,
but then enters state B,
from which there are many possible actions,
with rewards drawn from $\gauss(-0.1, 1.0)$.
Thus the expected return for any trajectory starting with the left
action is $-0.1$, making it suboptimal.
Nevertheless, the RL algorithm may pick   the left action
due to the maximization bias making B appear to have a positive value.

\subsubsection{Double Q-learning}
\label{sec:double}

One solution to avoid the maximization bias is to
use two separate $Q$-functions, $Q_1$ and $Q_2$,
one for selecting the greedy action,
and the other for estimating the corresponding $Q$-value.
In particular, upon seeing a transition $(s,a,r,s')$,
we perform the following update for $i=1:2$:
\begin{align}
  Q_i(s,a) &\assign Q_i(s,a) + \lr(\targetV_i(s,a) - Q_i(s,a)) \\
  \targetV_i(s,a) &= r + \gamma Q_{i}(s', \argmax_{a'} Q_{-i}(s', a'))
  \label{eqn:doubleQ}
\end{align}
So we see that $Q_1$ uses $Q_2$ to choose the best action
but uses $Q_1$ to evaluate it,
and vice versa.
This technique is called
\keywordDef{double Q-learning} \citep{vanHasselt2010}.
\cref{fig:sutton-6-5} shows the benefits of
the algorithm over standard Q-learning
in a toy problem.

\subsubsection{Double DQN}
\label{sec:doubleDQN}

In \citep{vanHasselt2016}, they combine double Q learning
with deep Q networks (\cref{sec:DQN}) to get \keywordDef{double DQN}.
This modifies \cref{eqn:doubleQ}
to its gradient form, and then the current network for action
proposals, but the target network for action evaluation.
Thus  the  training target becomes
\be
\TargetV(r,\vs'; \vw, \overline{\vw}) = r + \gamma
Q_{\overline{\vw}}(\vs', \argmax_{a'} Q_{\vw}(\vs',a'))
\ee

In \cref{sec:TD3} we discuss an extension called
\keywordDef{clipped double DQN} which uses two Q networks
and their frozen copies
to define
the following target:
\be
\TargetV(r,\vs'; \vw_{1:2}, \overline{\vw}_{1:2}) = r + \gamma  \min_{i=1,2}
Q_{\overline{\vw}_i}(\vs',\argmax_{a'} Q_{\vw_i}(\vs',a'))
\ee
where $Q_{\overline{\vw}_i}$ is the target network for $Q_{\vw_i}$.

\subsubsection{Randomized ensemble DQN}
\label{sec:REDQ}

The double DQN method is extended in
the \keywordDef{REDQ} (randomized ensembled double Q learning)
method of \citep{REDQ},
which uses
an ensemble of  $N>2$ Q-networks.
Furthermore, at each step, it draws a random
sample of $M \leq N$ networks, and takes the minimum over them
when computing the target value.
That is, it uses the following update
(see Algorithm 2 in appendix of \citep{REDQ}):
\be
\TargetV(r,\vs'; \vw_{1:N}, \overline{\vw}_{1:N})
= r + \gamma \max_{a'} \min_{i \in \calM} Q_{\overline{\vw}_i}(\vs',a')
\label{eqn:REDQtabular}
\ee
where $\calM$ is a random subset from the $N$ value functions.
The ensemble reduces the variance, and the minimum reduces
the overestimation bias.\footnote{
In addition, 
REDQ performs $G \gg 1$  updates of the value functions
for each environment step;
this  high \keywordDef{Update-To-Data} (UTD) ratio
(also called \keywordDef{Replay Ratio})
is critical for sample efficiency,
and is commonly used in model-based RL.
}
If we set $N=M=2$, we get a method
similar to clipped double Q learning.
(Note that REDQ is very similiar to the
\keywordDef{Random Ensemble Mixture} method of
\citep{Agarwal2020}, which was designed for offline RL.)

\eat{

See \cref{algo:REDQ} for some pseudocode for REDQ
for the tabular case.

\begin{algorithm}
\dontprintsemicolon
\caption{Tabular REDQ with experience replay}
\label{algo:REDQ}
Initialize $\{Q_i(s,a)\}_{i=1}^N$, $\overline{Q}_i=Q_i$ \\
$\hat{Q}(s,a) = \frac{1}{N} \sum_{i=1}^N Q_i(s,a)$\\
Initialize state $s$ \\
\Repeat{converged}
       {
         $a=\text{eps-greedy}(\hat{Q}(s,\cdot))$ \\
         $(s',r) = \text{step}(a,s)$\\
         $\data := \data \union \{ (s,a,r,s') \}$ \\
         \For{$G$ updates}
             {
               Sample a tuple $(s,a,r,s')$  from $\data$ \\
               Sample a set of $M$ indices $\calM \subseteq \{1,\ldots,N\}$ \\
               $Q_{\calM}(s,a) \defeq \min_{i \in \calM} \overline{Q}_i(s,a)$  \\
               Compute the  target 
               $\targetV(s,a) = r + \gamma \max_{a'} Q_{\calM}(s',a')$  \\
             \For{$i=1:N$}
                 {
                   $Q_i(s,a) \assign Q_i(s,a) + \lr(\targetV(s,a)-Q_i(s,a))$
                 }
             }
             $\hat{Q}(s,a) = \frac{1}{N} \sum_{i=1}^N Q_i(s,a)$\\
          $s \assign s'$
}
\end{algorithm}

}

\subsection{DQN extensions}
\label{sec:DQNextensions}

In this section, we discuss various extensions of DQN.

\subsubsection{Q learning for continuous actions}
\label{sec:QTopt}

Q learning is not directly applicable to continuous actions due to the need
to compute the argmax over actions.
An early solution to this problem,
based on neural fitted Q learning (see \cref{sec:NFQ}),
is proposed in \citep{Hafner2011}.
This became the basis of the 
DDPG algorithm of \cref{sec:DDPG},
which learns a policy to predict
the argmax.

An alternative  approach is to use gradient-free optimizers such as
the cross-entropy method to approximate the argmax.
The \keywordDef{QT-Opt} method of \citep{QTopt}
treats  the action vector $\va$ as a sequence of actions,
and optimizes one dimension at a time \citep{Metz2017}.
The  \keywordDef{CAQL} (continuous action $Q$-learning) method
of \citep{CAQL})
uses mixed integer programming to solve the $\argmax$ problem,
leveraging the \relu structure  of the $Q$-network.
The method of \citep{Seyde2022} quantizes each action dimension separately,
and then solves the argmax problem using methods inspired by  multi-agent RL.

\subsubsection{Dueling DQN}
\label{sec:duelingDQN}

The  \keywordDef{dueling DQN} method of \citep{Wang2016dueling},
learns a value function and an advantage function,
and derives the Q function, rather than learning it directly.
  This is helpful  when there are many actions with similar
$Q$-values, since the advantage $A(s,a)=Q(s,a)-V(s)$ focuses
  on the differences in value relative to a shared baseline.

In more detail, we define a network with $|A|+1$ output heads,
which computes  $A_{\vw}(\vs,a)$ for $a=1:A$ and
$V_{\vw}(\vs)$. We can then derive
\be
Q_{\vw}(\vs,a) = V_{\vw}(\vs) + A_{\vw}(\vs,a)
\ee
However, this naive approach ignores the following constraint
that holds for any policy $\pi$:
\begin{align}
\expectQ{A^{\pi}(s,a)}{\pi(a|s)}
&= \expectQ{Q^{\pi}(s,a) - V^{\pi}(s)}{\pi(a|s)} \\
&= V^{\pi}(s)  - V^{\pi}(s) = 0
\end{align}
Fortunately, for the optimal policy
$\pi^*(s) = \argmax_{a'} Q^*(s,a')$ we have
\begin{align}
0 
&= \expectQ{Q^*(s,a)}{\pi^*(a|s)}  - V^*(s) \\
&= Q^*(s, \argmax_{a'} Q^*(s,a')) - V^*(s) \\
&= \max_{a'} Q^*(s,a') - V^*(s) \\
&= \max_{a'} A^*(s,a')
\end{align}
Thus we can satisfy the constraint for the optimal policy
by subtracting off   $\max_a A(s,a)$ from the advantage head.
Equivalently we can compute the Q function using 
\be
Q_{\vw}(\vs,a) = V_{\vw}(\vs) + A_{\vw}(\vs,a) - \max_{a'} A_{\vw}(\vs,a')
\ee
In practice, the max is replaced by an average,
which seems to work better empirically.

\subsubsection{Noisy nets and exploration}
\label{sec:noisyDQN}
\label{sec:noisyNets}

Standard DQN relies on the epsilon-greedy strategy to perform exploration.
However, this will explore equally in all states,
whereas we would like to the amount of exploration
to be state dependent, to reflect the amount of uncertainty
in the outcomes of trying each action in that state
due to lack of knowledge (i.e., \keywordDef{epistemic uncertainty}
rather than aleatoric or irreducile uncertainty).
An early approach to this,
known as  \keywordDef{noisy nets} \citep{Fortunato2018},
added random noise to
the network weights to encourage exploration which is
temporally consistent within episodes.
More recent methods for exploration
are discussed in \cref{sec:exploreExploit}.

\subsubsection{Multi-step DQN}
\label{sec:nsteps}

As we discussed in \cref{sec:TDlambda},
we can reduce the bias introduced by bootstrapping
by replacing TD(1) updates with TD($n$) updates,
where we unroll the value computation for $n$ MC steps,
and then plug in the value function at the end.
We can apply this to the DQN context by defining the target
\be
\targetV(s_0,a_0) = \sum_{t=1}^n \gamma^{t-1} r_t
 + \gamma^n \max_{a_n} Q_{\vw}(s_n, a_n)
 \ee
This can be implemented for episodic environments
by storing experience tuples of the form
\be
\tau = (s, a, \sum_{k=1}^n \gamma^{k-1} r_k, s_n, \done)
\ee
where $\done=1$ if the trajectory ended at any point
during the $n$-step rollout.
 
Theoretically this method is only valid if all the intermediate
actions, $a_{2:n-1}$, are sampled from the current optimal policy
derived from $Q_{\vw}$,
as opposed to some behavior policy,
such as epsilon greedy or some samples from the replay buffer
from an old policy.
In practice, we can just restrict sampling
to  recent samples from the replay buffer,
making the resulting method approximately on-policy.

\subsubsection{Q($\lambda$)}

Instead of using a fixed $n$, it is possible
to use a weighted combination of returns;
this is known as the  $Q(\lambda)$
algorithm \citep{Peng1994,Harutyunyan2016,Kozuno2021},
and relies on the concept of eligibility traces.
Unfortunately it is more complicated
than the Sarsa case in \cref{sec:sarsaLambda},
since Q learning is off-policy,
but the eligibility traces backpropagate information
obtained by the exploration policy.

\eat{
\subsection{Putting it all together}

Many of the above methods can be combined for improved performance.
We give some examples of such hybrid methods below.
(Note that, besides combining various algorithmic pieces,
careful hyper-parameter tuning is critical for good results,
as discussed in \citep{Obando-Ceron2024}.)
}

\subsubsection{Rainbow}
\label{sec:rainbow}

\begin{figure}
\centering
\includegraphics[height=1.5in]{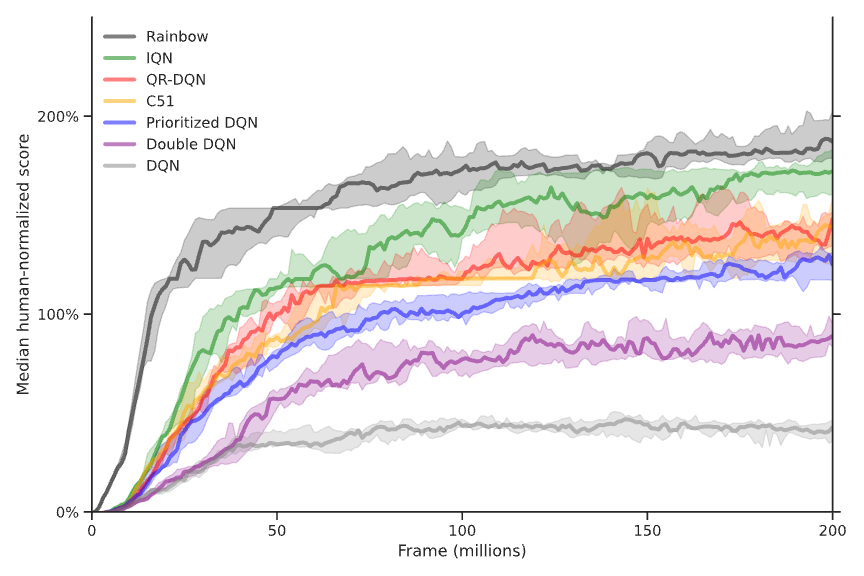}
\caption{
  Plot of median human-normalized score over all 57 Atari games for
  various DQN agents.
  The yellow, red and green curves are distributional RL methods
  (\cref{sec:distributional}),
  namely categorical DQN (C51) (\cref{sec:C51})
  Quantile Regression DQN (\cref{sec:QRDQN}),
  and Implicit Quantile Networks \citep{IQN}.
  Figure from \url{https://github.com/google-deepmind/dqn_zoo}.
}
\label{fig:rainbow}
\end{figure}

The \keywordDef{Rainbow} method of \citep{rainbow}
combined 6 improvements to the vanilla DQN method,
as listed below.
(The paper is called ``Rainbow'' due to the color coding
of their results plot, a modified version of which is shown in
\cref{fig:rainbow}.)
At the time it was published (2018), this produced 
SOTA results on the Atari-200M benchmark.
The 6 improvements are as follows:
\begin{itemize}
\item Use \keyword{double DQN}, as in \cref{sec:doubleDQN}.

    \item  Use 
  \keyword{prioritized experience replay},  as in \cref{sec:PER}.

\item  Use the
categorical DQN (C51) (\cref{sec:C51}) 
distributional RL method.

  \item Use n-step returns (with $n=3$), as in \cref{sec:nsteps}.
  
\item Use \keyword{dueling DQN}, as in \cref{sec:duelingDQN}.

\item Use \keyword{noisy nets}, as in \cref{sec:noisyNets}.
\end{itemize}
Each improvement gives diminishing returns,
as can be see in \cref{fig:rainbow}.

Recently the ``Beyond the Rainbow'' paper \citep{beyond} proposed
several more extensions: 
\begin{itemize}
\item Use a larger CNN with residual connections,
  namely the Impala
  network from \citep{Espeholt2018}
  with the modifications (including the use of spectral normalization)
  proposed in
  \citep{Schmidt2021atari}.

  \item  Replace C51 with Implicit Quantile Networks \citep{IQN}.

  \item Use \keywordDef{Munchausen RL} \citep{Vieillard2020},
    which modifies
    the Q learning update rule by adding an entropy-like
    penalty.

  \item Collect 1 environment step from 64 parallel workers
    for each minibatch update (rather than taking many steps
    from a smaller number of workers).
\end{itemize}

\eat{
    \footnote{
    The idea is as follows.
    First note that Q learning defines a greedy policy
    $\pi(a|s)=1$ iff $a = \argmax_{a'} Q(s,a')$,
    which we can convert to a stochastic policy
    $\pi(a|s) = \frac{\exp Q(s,a)}{\sum_{a'} Q(s,a')}$.
    Now consider the experience tuple
    $(s,a,r,s')$, so the Q learning
    target becomes $\targetV(s,a) = r + \max_{a'} Q^*(s',a')$.
Since $Q^*$ is unknown, we replace it with the current estimate
$Q_{\vw}(s',a')$ to get $\targetV(s,a) = r+q$ where $q=\max_{a'} Q_{\vw}(s',a')$.
However, this bootstrapping process can be unstable.
To help improve it, we can replace the target $r +q$ with
$r + \alpha \log \pi(a|s) + q$;
the motivation for this regularizer is that
the optimal policy should satisfy $\log \pi^*(a|s)$.
}
}

\subsubsection{Bigger, Better, Faster}
\label{sec:BBF}

At the time of writing this document (2024),
the SOTA on the 100k sample-efficient
Atari benchmark
\citep{Atari100k}
is obtained by the \keywordDef{BBF} algorithm of \citep{BBF}.
(BBF stands for ``Bigger, Better, Faster''.)
It uses the following tricks,
in order of decreasing importance:
\begin{itemize}

\item Use a larger CNN with residual connections,
  namely a modified version of the Impala
  network from \citep{Espeholt2018}.

  \item Increase the \keywordDef{update-to-data} (UTD)
  ratio
  (number of times we update the Q function
  for every observation that is observed), in order to increase
  sample efficiency
  \citep{VanHasselt2019}.

\item Use a periodic soft reset of (some of) the network weights
to avoid loss of elasticity due to increased network updates, following the
\keywordDef{SR-SPR} method of \citep{DOro2022}.

\item Use n-step returns, as in \cref{sec:nsteps},
  and then gradually decrease (anneal) the n-step return from
 $n=10$ to $n=3$,
to reduce the bias over time.

\item Add weight decay.

\item Add a \keyword{self-predictive representation} loss
  (\cref{sec:self-predictive})
 to increase sample efficiency.

\item  Gradually increase the discount factor from
  $\gamma=0.97$ to $\gamma=0.997$,
  to encourage longer term planning once the model starts to be trained.\footnote{
    The \keywordDef{Agent 57} method of  \citep{Badia2020}
  automatically learns the exploration rate and discount factor
  using a multi-armed bandit stratey,
  which lets it  be more exploratory or more exploitative,
  depending on the game. This resulted in super human performance
  on all 57 Atari games in ALE.
  However, it required 80 billion frames (environment steps)!
  This was subsequently reduced to the ``standard''
  200M frames in the \keywordDef{MEME} method of \citep{Kapturowski2022}.
  }

\item Drop noisy nets (which requires multiple network copies and thus slows
  down training due to increased memory use), since it does not help.

\item Use
  \keyword{dueling DQN} (see \cref{sec:duelingDQN}).

\item Use
  \keyword{distributional DQN}
  (see \cref{sec:distributional}).
  

\end{itemize}

\subsubsection{Other methods}

Many other methods have been proposed to reduce the sample complexity
of value-based RL while maintaining performance,
see e.g., the \keywordDef{MEME} paper of
\citep{Kapturowski2022}.

\subsection{Q-learning for GCRL using hindsight relabeling}
\label{sec:QGCRL}
\label{sec:relabeling}
\label{sec:HER}

In this section, we discuss how to learn a goal-conditioned policy
(see \cref{sec:GCRL}) using Q learning (or any other off-policy learning method).
The basic idea is as follows.
We collect various trajectores  in the environment,
from a starting state $s_0$ to some terminal state $s_T$,
and then define the goal of each trajectory as being $g=s_T$;
this trajectory then serves as a demonstration
of how to achieve this goal.
This is called \keywordDef{hindsight experience relabeling}
or \keywordDef{HER} \citep{HER},
or just \keywordDef{hindsight relabeling}.
This can be used to relabel the trajectories stored in the replay buffer.
That is, if we have $(s,a,R(s|g),s',g)$ tuples, we replace them with
$(s,a,R(s|g'),g')$ where $g'=s_T$.
We can then use any off-policy RL method to learn $\pi(a|s,g)$.\footnote{
One limitation of HER is that it only works when the reward is Markovian,
so $R(s,a)=1$ iff $s=g$. A method to  handle non-Markovian rewards
using on-policy RL is discussed in \citep{Gong2024RL}. The key trick is to use
an auto-curriculum (see \cref{sec:curriculum}) where the goals
are chosen to be solvable with the current policy.
}

In \citep{Eysenbach2020}, they show that HER can be viewed as a special
case of maximum-entropy inverse RL (see \cref{sec:maxentRL}),
since it is estimating the reward for which the corresponding
trajectory was optimal.

\chapter{Policy-based RL}
\label{sec:policySearch}


In the previous section, we considered methods that
estimate the action-value function, $Q(s,a)$,
from which we derive a policy.
However, these methods have several disadvantages:
(1)  they can be difficult to apply to continuous action spaces;
(2) they may diverge if function approximation is used (see
\cref{sec:deadlyTriad});  
(3) the training of $Q$, often based on TD-style updates,
is not directly related to the expected return
garnered by the learned policy;
(4) they learn deterministic policies,
whereas in stochastic and partially observed environments,
stochastic policies are provably better
\citep{Jaakkola94}.

In this section, we discuss
\keywordDef{policy search} methods,
which directly optimize the parameters of the policy
so as to maximize its expected return.
We mostly focus on \keywordDef{policy gradient} methods,
that use the gradient of the loss to guide the search
(see e.g., \citep{Agarwal2021}).
As we will see,  these policy methods often 
benefit from estimating a value or advantage function
to reduce the variance in the policy search process,
so we will also use techniques from \cref{sec:valueRL}.

The parametric policy will be denoted by 
$\pi_{\vtheta}(a|s)$, which is usually some form of neural
network.
For discrete actions, the final layer
is usually passed through a softmax
function and then into a categorical distribution.
For continuous actions, we typically use a Gaussian output layer
(potentially clipped to a suitable range, such as $[-1,1]$),
although it is also possible
to use more expressive (multi-modal) distributions,
such as diffusion models
(which, when used as a policy,
is s known as a \keywordDef{diffusion policy}
\citep{diffusionPolicy}).

There are many implementation details one needs to get
right to get good performance when designing such
neural networks.
For example, \citep{Furuta2021} recommends
using ELU instead of RELU activations,
and using LayerNorm.
(In \citep{PQN} they recently proved
that adding layer norm to the final layer
of a DQN model
is sufficient to guarantee that value learning
is stable, even in the nonlinear setting.)
However, we do not discuss these details in this manuscript.

For more details on policy gradient methods,
see e.g., \citep{weng2018PG,Agarwal2021,Lehmann2024}.


\eat{
\footnote{
To handle multimodal distributions, it has recently
become popular to use more expressive policy classes such as diffusion
models;
this is known as a \keywordDef{diffusion policy}
\citep{diffusionPolicy}.
However, since it is difficult to compute the probability
value of the output from diffusion, such models are usually
trained using behavior cloning (\cref{sec:BC}),
rather than the policy gradient methods we discuss in this section.
}
}

\section{Policy gradient methods}
\label{sec:policyGradient}
\label{sec:PG}

In this section, we discuss how to compute
the expected value of a policy, and the gradient
of this expectation. This can be used, together with SGD,
to learn a locally optimal policy.
Our presentation is based in part on
\citep[Ch.11]{KochenderferDecisions}.

\subsection{Likelihood ratio estimate}

We define the value of a policy as
\be
J(\pi_{\vtheta}) = J(\vtheta) =
\E_{p_{\vtheta}(\traj)}[R(\traj)]
\ee
where $R(\traj)=\gamma^0 r_0 + \gamma^1 r_1 + \ldots$
is the return along the trajectory, and
$p_{\vtheta}(\traj)$ is the distribution over
trajectories induced by the policy
(and world model):
\be
p_{\vtheta}(\traj) = p(s_1)
\prod_{k=1}^T \calT(s_{k+1}|s_k,a_k) \pi_{\vtheta}(a_k|s_k)
\label{eqn:probTraj}
\ee
The gradient of the policy value  is given by
\begin{align}
  \nabla_{\vtheta} J(\vtheta)
  &= \int \nabla_{\vtheta} p_{\vtheta}(\traj) R(\traj) d\traj 
  = \int p_{\vtheta}(\traj)
  \frac{\nabla_{\vtheta} p_{\vtheta}(\traj)}{p_{\vtheta}(\traj)}
  R(\traj) d\traj
  = \expectQ{
      \frac{\nabla_{\vtheta} p_{\vtheta}(\traj)}{p_{\vtheta}(\traj)}
      R(\traj)}{\traj}
  \label{eqn:likRatio}
\end{align}
This is known as the \keywordDef{likelihood ratio estimator}.

Now consider the \keywordDef{log derivative trick},
which is the simple observation that
$\nabla \log \pi = \frac{\nabla \pi}{\pi}$.
Using this, we can rewrite the above expression as
\begin{align}
  \nabla_{\vtheta} J(\vtheta) =
  \expectQ{\nabla_{\vtheta} \log p_{\vtheta}(\traj) R(\traj)}
  {\traj}
\end{align}
The expectations can be estimated using Monte Carlo
sampling (rolling out the policy in the environment).
The gradient can be computed
from \cref{eqn:probTraj} as follows:
\be
\nabla_{\vtheta} \log p_{\vtheta}(\traj)
  = \sum_{k=1}^T \nabla_{\vtheta} \log \pi_{\vtheta}(a_k|s_k)
\ee
Hence
\be
\nabla_{\vtheta} J(\vtheta)
=\expectQ{
  \left(  \sum_{k=1}^T \nabla_{\vtheta} \log \pi_{\vtheta}(a_k|s_k)
  \right)
  R(\traj)
}{\traj}
\label{eqn:SFE}
\ee
In statistics, the term $\nabla_{\vtheta} \log \pi_{\vtheta}(\va|\vs)$
is called the (Fisher) \keywordDef{score function}\footnote{
This is distinct from the Stein score,
which is the gradient wrt the argument of the log probability,
$\nabla_{\va} \log \pi_{\vtheta}(\va|\vs)$,
as used in diffusion.
},
so sometimes \cref{eqn:SFE}
is called the \keywordDef{score function estimator}
or \keywordDef{SFE} \citep{Fu2015,Mohamed2020}.

\subsection{Variance reduction using reward-to-go}

The likelihood ratio estimator can have high variance,
since we are sampling entire trajectories.
Fortunately we can reduce the variance using
the temporal/causal structure of the problem.
In particular, note that
from \cref{eqn:SFE}
we have
\begin{align}
  \nabla J(\vtheta)
&= \expectQ{
  \left(\sum_{k=1}^T
  \underbrace{\nabla_{\vtheta}  \log \pi_{\vtheta}(a_k|s_k)}_{f_k}
  \right)
  \left(\sum_{k=1}^T r_k \gamma^{k-1} \right)
  }{\traj} \\
  &= \expectQ{(f_1 + f_2 + \cdots f_T)
    (r_1 + r_2 \gamma + r+3 \gamma^2 + \cdots  r_T \gamma^{T-1})}{\traj} 
\end{align}
Expanding the product inside the expectation we have
\begin{align}
 & f_1 r_1 + f_1 r_2 \gamma + f_1 r_3 \gamma^2 + \cdots f_1 r_T \gamma^{T-1} \\
  +&   \cancel{f_2 r_1} + f_2 r_2 \gamma+ f_2 r_3 \gamma^2 + \cdots f_2 r_T \gamma^{T-1} \\
  +&   \cancel{f_3 r_1} + \cancel{f_2 r_2 \gamma} 
  + f_3 r_3 \gamma^2 \cdots f_2 r_T \gamma^{T-1} \\
  \vdots \\
  +&   \cancel{f_T r_1} + \cancel{f_T r_2 \gamma}
  + \cancel{f_T r_3 \gamma^2}
  + \cdots f_T r_T \gamma^{T-1}   
  \end{align}
where we have canceled terms that are 0,
due to the fact that the reward at step $k$
cannot depend on actions at time steps in the future.
Plugging in this simplified expression we get
\begin{align}
  \nabla J(\vtheta)
  &= \expectQ{
    \sum_{k=1}^T \nabla_{\vtheta}
    \log \pi_{\vtheta}(a_k|s_k)
    \left( \sum_{l=k}^T r^l \gamma^{l-1} \right)}{\traj} \\
  &= \expectQ{
    \sum_{k=1}^T \nabla_{\vtheta}
    \log \pi_{\vtheta}(a_k|s_k)
    \left(\gamma^{k-1} \sum_{l=k}^T r^l \gamma^{l-k} \right)}{\traj} \\
  &= \expectQ{
    \sum_{k=1}^T \nabla_{\vtheta}
    \log \pi_{\vtheta}(a_k|s_k)
    \gamma^{k-1} G_k}{\traj} 
\end{align}
where $G_k$ is the \keywordDef{reward-to-go}
or return
\begin{align}
\return_k &\defeq r_k + \gamma r_{k+1} + \gamma^2 r_{k+2} + \cdots
+ \gamma^{T-k-1} r_{T-1} 
= \sum_{l=k}^{T-1} \gamma^{l-k} r_l
\end{align}

Note that the reward-to-go of a state-action pair
$(s,a)$
can be considered
as a single sample approximation of the state-action value
function $Q_{\vtheta}(s,a)$.
Averaging over such samples gives
\begin{align}
  \nabla J(\vtheta)
  &= \expectQ{
    \sum_{k=1}^T
    \gamma^{k-1} Q_{\vtheta}(s_k,a_k)
  \nabla_{\vtheta} \log \pi_{\vtheta}(a_k|s_k)
  }{\traj}
  \label{eqn:gradJQ}
\end{align}

\subsection{REINFORCE}
\label{sec:REINFORCE}
\label{sec:reinforceOpt}
\label{sec:reinforce}

In this section, we describe an algorithm
that uses the above estimate of the gradient
of the policy value, together with SGD,
to fit a policy.
That is, we use
\be
\vtheta_{j+1} := \vtheta_{j} + \lr
    \sum_{k=1}^T \nabla_{\vtheta}
    \log \pi_{\vtheta_j}(a_k|s_k)
    \gamma^{k-1} G_k
    \label{eqn:REINFORCE}
\ee
where $j$ is the SGD iteration number,
and we draw a single trajectory at each step.
This is 
is called the \keywordDef{REINFORCE} algorithm
\citep{Williams1992}.\footnote{
  The term 
``REINFORCE'' is an acronym for
``REward Increment = nonnegative Factor x Offset
Reinforcement x Characteristic Eligibility''.
The phrase ``characteristic eligibility'' refers to  the
$\nabla \log \polapprox(a_t|s_t)$ term;
the phrase ``offset reinforcement'' refers to the
$\return_t-b(s_t)$ term,
where $b$ is a baseline to be defined later;
and the phrase ``nonnegative factor''
refers to the learning rate $\lr$ of SGD.
} %

The update equation in  \cref{eqn:REINFORCE}
 can be interpreted as follows:
we compute the sum of discounted future rewards
induced by a trajectory,
and if this is positive, we increase
$\vtheta$ so as to make this trajectory more likely,
otherwise we decrease $\vtheta$.
Thus, we reinforce good behaviors,
and reduce the chances of generating bad ones.
See \cref{algo:REINFORCE} for the pseudocode.

\begin{algorithm}
\dontprintsemicolon
\caption{REINFORCE (episodic version)}
\label{algo:REINFORCE}
Initialize policy parameters $\vtheta$\\
\Repeat{converged}
       {
         Sample an episode $\traj=(s_1, a_1, r_1, s_2, \ldots, s_T)$ using $\polapprox$ \\
         \For{$k=1,\ldots,T$}{
           $\return_k = \sum_{l=k}^T \gamma^{l-k} R_l$ \\
           $\vtheta \assign \vtheta +
           \lr_{\vtheta} \gamma^{k-1} \return_k
           \nabla_{\vtheta} \log \policy_{\vtheta}(a_k|s_k)$
         }
}
\end{algorithm}

\subsection{The policy gradient theorem}

We now turn to the discounted infinite horizon setting.
We define the
\keywordDef{discounted state visitation measure}
as follows:
\begin{align}
  \statmeasure_{\pi}^{\gamma}(s)
  &\defeq
\gamma^0 P(s_0=s|\pi) + \gamma P(s_1=s|\pi)
+ \gamma^2 P(s_2=s|\pi) + \cdots \\
&= \sum_{t=0}^{\infty} \gamma^t
\underbrace{\sum_{s_0}   p_0(s_0) p^{\pi}(s_0 \rightarrow s, t)}_{p_t^{\pi}(s)}
\end{align}
where
$p^{\pi}(s_0 \ra s, t)$ is the probability of going
from $s_0$ to $s$ in $t$ steps,
and
$p_t^{\pi}(s)$ is the marginal probability of being in state $s$
at time $t$ (after each episodic reset).
Note that $\statmeasure_{\pi}^{\gamma}$
is a measure of time spent in  non-terminal states,
but it is not a probability measure, since it is not normalized,
i.e., $\sum_s \statmeasure_{\pi}^{\gamma}(s) \neq 1$.
However, we can
define a normalized version of the measure $\statmeasure$
by noting that
$\sum_{t=0}^{\infty} \gamma^t = \frac{1}{1-\gamma}$ for $\gamma<1$.
Hence the
\keywordDef{normalized discounted state visitation distribution}
is given by
the following (note the change from $\rho$ to $p$):
\be
\statdistpol(s) = (1-\gamma) \statmeasure_{\pi}^{\gamma}(s)
= (1-\gamma) \sum_{t=0}^{\infty} \gamma^t p_t^{\pi}(s)
\label{eqn:statdistpol}
\ee
We can convert from the normalized distribution back to the measure using
\be
\statmeasure_{\pi}^{\gamma}(s)
= \frac{1}{1-\gamma} \statdistpol(s)
\label{eqn:statmeasurepol}
\ee

Using this notation, one can show
(see \citep{Kakade2002RL})
that we can rewrite
\cref{eqn:gradJQ} in terms of expectations over states
rather than over trajectories:
\begin{align}
\nabla_{\vtheta} J(\vtheta)
&=\expectQ
{Q^{\pi_{\vtheta}}(s,a) \nabla_{\vtheta} \log \pi_{\vtheta}(a|s)}
    {\statmeasure_{\pi}^{\gamma}(s) \pi_{\vtheta}(a|s)} 
\label{eqn:PGthm} \\
&=
\frac{1}{1-\gamma}
  \expectQ{Q^{\pi_{\vtheta}}(s,a) \nabla_{\vtheta} \log \pi_{\vtheta}(a|s)}
    {\statdistpol(s) \pi_{\vtheta}(a|s)} 
\end{align}
This is known as the \keywordDef{policy gradient theorem}
\citep{Sutton99b}.

\subsection{Variance reduction using a baseline}
\label{sec:varianceReduction}

In practice, estimating the policy gradient using
\cref{eqn:SFE} can have a high variance.
A \keywordDef{baseline} function $b(s)$ can be used for variance reduction
to get
\begin{align}
\nabla_{\vtheta} J(\vtheta) = 
\expectQ{
  (\Qpolapprox(s,a) - b(s))
  \nabla_{\vtheta} \log \polapprox(a|s)
}
{\statmeasure_{\vtheta}(s) \polapprox(a|s)}
\label{eqn:PGthmBaseline}
\end{align}
Any function that satisfies $\expect{\nabla_{\vtheta} b(s)}=0$ is a valid
baseline.
This follows since
\begin{align}
\sum_a  \nabla_{\vtheta} \pi_{\vtheta}(a|s) (Q(s,a)-b(s))
&= \nabla_{\vtheta} \sum_a \pi_{\vtheta}(a|s) Q(s,a)
-\nabla_{\vtheta} [\sum_a \pi_{\vtheta}(a|s)] b(s) \\
&= \nabla_{\vtheta} \sum_a \pi_{\vtheta}(a|s) Q(s,a) - 0
\end{align}
A common choice for the baseline is $b(s) = V(s)$.
This is a valid choice since $E[\nabla_{\vtheta} V(s)]=0$
if we use an old (frozen) version of the policy
that is independent of $\vtheta$.
This is a useful choice  $V(s)$ and $Q(s,a)$ are correlated
and have similar magnitudes,
so the scaling factor in front of the gradient term will be small,
ensuring the update steps are not too big.

Note that
$Q(s,a)-V(s)=A(s,a)$
is the advantage function.
In the finite horizon case we get
\be
\nabla J(\vtheta)
= \expectQ{
  \sum_{k=1}^T \nabla_{\vtheta}
  \log \pi_{\vtheta}(a_k|s_k)
  \gamma^{k-1} (Q_{\vtheta}(s_k,a_k) - V_{\vtheta}(s_k))
  }{\traj}
= \expectQ{
  \sum_{k=1}^T \nabla_{\vtheta}
  \log \pi_{\vtheta}(a_k|s_k)
  \gamma^{k-1} A_{\vtheta}(s_k,a_k)
  }{\traj}
\ee
We can also apply a baseline to the reward-to-go
formulation to get
\be
\nabla J(\vtheta)
= \expectQ{
  \sum_{k=1}^T \nabla_{\vtheta}
  \log \pi_{\vtheta}(a_k|s_k)
  \gamma^{k-1} (G_k - b(s_k))
  }{\traj}
\ee

We can derive analogous baselines for the infinite horizon case,
defined in terms of $\statdistpol$.


\subsection{REINFORCE with baseline}

We can recover the full REINFORCE alorithm
by combining SGD with
the score function estimate
with a baseline,  as follows:
\begin{align}
  \vtheta \assign \vtheta + \lr \sum_{k=1}^{T}
  \gamma^{k-1} (G_k - b(s_k))
  \nabla_{\vtheta} \log\polapprox(a_k|s_k)
\label{eqn:REINFORCEbaseline}
\end{align}
See \cref{algo:REINFORCEbaseline} for the pseudocode,
where we use the value function as a baseline,
estimated using TD.

\begin{algorithm}
\dontprintsemicolon
\caption{REINFORCE (episodic) with value function baseline}
\label{algo:REINFORCEbaseline}
Initialize policy parameters $\vtheta$, baseline parameters $\vw$\\
\Repeat{converged}
       {
         Sample an episode $\traj=(s_1, a_1, r_1, s_2, \ldots, s_T)$ using $\polapprox$ \\
         \For{$k=1,\ldots,T$}{
           $\return_k = \sum_{l=k}^T \gamma^{l-k} R_l$ \\
           $\delta_k = \return_k - \Vapprox(s_k)$  \\
           $\vw \assign \vw - \lr_{\vw} \delta_k \nabla_{\vw}
           \Vapprox(s_k)$  \\
           $\vtheta \assign \vtheta +
           \lr_{\vtheta} \gamma^{k-1} \delta_k
           \nabla_{\vtheta} \log \policy_{\vtheta}(a_k|s_k)$ 
         }
}
\end{algorithm}

\section{Actor-critic methods}
\label{sec:actorCritic}
\label{sec:AC}

An \keywordDef{actor-critic} method \citep{Barto1983}
uses the policy gradient method,
but where the expected return $G_t$ is estimated
using temporal difference learning
of a value function instead of MC rollouts.
(The term ``actor'' refers to the policy,
and the term ``critic'' refers to the value function.)
The use of bootstrapping in TD updates
allows more efficient learning of the value function
compared to MC, and further reduces the variance.
In addition, it allows us to develop a fully online, incremental
algorithm, that does not need to wait until the end of the trajectory
before updating the parameters.

\subsection{Advantage actor critic (A2C)}
\label{sec:A2C}
\label{sec:A3C}

Concretely, consider the use of the one-step TD method
to estimate the return in the episodic case,
i.e., we replace $\return_t$ with
$\return_{t:t+1}=r_t+ \gamma \Vapprox(s_{t+1})$.
If we use $\Vapprox(s_t)$ as a baseline,
the REINFORCE update in \cref{eqn:REINFORCE} becomes
\begin{align}
\vtheta 
&\assign \vtheta + \lr \sum_{t=0}^{T-1} \gamma^t \left(G_{t:t+1} - \Vapprox(s_t)\right) \nabla_{\vtheta} \log\polapprox(a_t|s_t) \\
&= \vtheta + \lr \sum_{t=0}^{T-1} \gamma^t \big(r_t+ \gamma \Vapprox(s_{t+1}) - \Vapprox(s_t)\big) \nabla_{\vtheta} \log\polapprox(a_t|s_t)
\end{align}


Note that
$\delta_t = r_{t+1} + \gamma \Vapprox(s_{t+1}) -\Vapprox(s_t)$
is a single sample approximation
to the advantage function $\Adv(s_t,a_t) = Q(s_t,a_t) - V(s_t)$.
This method is therefore called 
\keywordDef{advantage actor critic} or \keywordDef{A2C}.
See \cref{algo:A2C} for the pseudo-code.\footnote{
In \citep{A3C}, they proposed
a distributed version of A2C known as
\keywordDef{A3C} 
which stands for  ``asynchrononous advantage actor critic''. 
}
(Note that $V_{\vw}(s_{t+1})=0$ if $s_t$ is a done state,
representing the end of an episode.)
Note that this is an on-policy algorithm,
where we update the value function $V_{\vw}^{\pi}$
to reflect the value of the current policy $\pi$.
See \cref{sec:bilevel} for further discussion of this point.


\begin{algorithm}
\dontprintsemicolon
\caption{Advantage actor critic (A2C) algorithm (episodic)}
\label{algo:A2C}
Initialize actor parameters $\vtheta$, critic parameters $\vw$\\
\Repeat{converged}
       {
       Sample starting state $s_0$ of a new episode \\
       \For{$t=0,1,2,\ldots$}
       {
         Sample action $a_t \sim \polapprox(\cdot|s_t)$ \\
         $(s_{t+1},r_t,\done_t) = \text{env.step}(s_t,a_t)$ \\
        $\targetV_t = r_t + \gamma (1-\done_t) \Vapprox(s_{t+1})$       // Target \\
       $\delta_t = \targetV_t - \Vapprox(s_t)$ // Advantage \\
           $\vw \assign \vw  + \lr_{\vw} \delta_t \nabla_{\vw}
       \Vapprox(s_t)$ // Critic \\
           $\vtheta \assign \vtheta + \lr_{\vtheta} \gamma^t \delta_t
       \nabla_{\vtheta} \log \polapprox(a_t|s_t)$ // Actor \\
       \If{$\done_t=1$}{break}
         }
}
\end{algorithm}

In practice, we should use a stop-gradient
operator on the target value for the TD update,
for reasons  explained in \cref{sec:deadlyTriad}.
Furthermore, it is common to add an entropy term
to the policy, to act as a regularizer
(to ensure the policy remains stochastic, which smoothens
the loss function --- see \cref{sec:SAC}).
If we use a shared network with separate value  and policy heads,
we need to use a single loss function for training all the parameters
$\vphi$.
Thus we get the following loss, for each trajectory,
where we want to minimize TD loss,
maximize the policy gradient (expected reward) term,
and maximize the entropy term.
\begin{align}
  \loss(\vphi;\tau) &= \frac{1}{T} \sum_{t=1}^T \left[
    \lambda_{TD} \loss_{TD}(s_t,a_t,r_t,s_{t+1})
    -    \lambda_{PG} J_{PG}(s_t,a_t,r_t,s_{t+1})
    -    \lambda_{ent} J_{ent}(s_t)
    \right]
  \label{eqn:ACloss}\\
  \targetV_t &= r_t + \gamma (1-\done(s_t)) V_{\vphi}(s_{t+1}) \\
  \loss_{TD}(s_t,a_t,r_t,s_{t+1})  &= (\stopgrad(\targetV_t) - V_{\phi}(s))^2 \\
  J_{PG}(s_t,a_t,r_t,s_{t+1}) &= (\stopgrad(\targetV_t - V_{\phi}(s_t))
  \log \pi_{\vphi}(a_t|s_t) \\
  J_{ent}(s_t) &= -\sum_a \pi_{\vphi}(a|s_t) \log \pi_{\vphi}(a|s_t)
\end{align}
To handle the dynamically varying scales of the different loss functions,
we can use 
the \keywordDef{PopArt} method
of \citep{van-Hasselt2016,Hessel2019}
to allow for a fixed set of hyper-parameter values for $\lambda_i$.
(PopArt  stands for
``Preserving Outputs Precisely, while Adaptively Rescaling Targets''.)

\eat{
\begin{align}
  \loss(\vphi;\tau) = \frac{1}{T} \sum{t=1}^T \left[
    \lambda_{TD} (\stopgrad(\targetV_t) - V_{\vphi}(\vs_t))^2
    -    \lambda_{PG} (\stopgrad(\targetV_t - V_{\vphi}(\vs_t))
    -    \lambda_{ent} \entropy(\pi_{\vphi}(\cdot|\vs_t))
    \right]
  \end{align}
}

\eat{
We can then accumulate  the following gradients
over each sampled trajectory:
\begin{align}
  \nabla_{\vw}  &= -\frac{1}{2T} \sum_t
  \nabla_{\vw} (\targetV_t - V_{\vw}(s_t))^2 
  = \frac{1}{T} \sum_t  (\targetV_t - V_{\vw}(s_t)) \nabla_{\vw} V_{\vw}(s_t) \\
  \nabla_{\vtheta} &= \frac{1}{T} \sum_t
  (\targetV_t - V_{\vw}(s_t)) \nabla_{\vtheta} \log \pi_{\vtheta}(a_t|s_t)
  \end{align}
The second gradient ignores the discount term $\gamma^t$,
but omitting this is standard practice,
even though it
results in a biased estimate
\citep{Nota2020,Che2023}.
}

\eat{
For a more practical minibatch version,
based on \citep{Ivanov2019},
see \cref{algo:A2C}.
}

\subsection{Generalized advantage estimation (GAE)}
\label{sec:GAE}


In A2C, we replaced the high variance, but unbiased,
MC return $G_t$
with the low variance, but biased,
one-step bootstrap return $G_{t:t+1} = r_t + \gamma V_{\vw}(s_{t+1})$.
More generally, we can compute the
$n$-step estimate
\begin{align}
G_{t:t+n} =
r_t + \gamma r_{t+1} + \gamma^2 r_{t+2}
+ \cdots + \gamma^{n-1} r_{t+n-1} + \gamma^n \Vapprox(s_{t+n})
\end{align}
and thus obtain  the (truncated) $n$-step advantage estimate as follows:
\be
A_{\vw}^{(n)}(s_t,a_t) = G_{t:t+n} - \Vapprox(s_t)
\ee
Unrolling to infinity, we get
\begin{align}
  A_t^{(1)} &= r_t + \gamma v_{t+1} - v_t  \\
  A_t^{(2)} &= r_t + \gamma r_{t+1} + \gamma^2 v_{t+2} - v_t  \\
  &\vdots \\
  A_t^{(\infty)} &= r_t + \gamma r_{t+1} + \gamma^2 r_{t+2} +\cdots - v_t 
  \end{align}
$A_t^{(1)}$ is high bias but low variance,
and $A_t^{(\infty)}$  is unbiased but high variance.


Instead of using a single value of $n$, we can take
a weighted average.
That is, we define
\be
A_t = \frac{\sum_{n=1}^T w_n A_t^{(n)}}{\sum_{n=1}^T w_n}
\ee
If we set $w_n = \lambda^{n-1}$ we get the following simple recursive
calculation:
\begin{align}
  \delta_t &= r_t + \gamma v_{t+1} - v_t \\
  A_t &= \delta_t + \gamma \lambda \delta_{t+1} + \cdots +
  (\gamma \lambda)^{T-(t+1)} \delta_{T-1}
  = \delta_t + \gamma \lambda   A_{t+1}
  \label{eqn:GAE}
  \end{align}
Here $\lambda \in [0,1]$ is a parameter that controls the bias-variance
tradeoff: larger values decrease the bias but increase the variance.
This is called \keywordDef{generalized advantage estimation}
(\keywordDef{GAE})~\citep{GAE}.
See \cref{algo:GAE} for some pseudocode.
Using this, we can define a general actor-critic
method, as shown in \cref{algo:AC}.

\begin{algorithm}
\dontprintsemicolon
\caption{Generalized Advantage Estimation}
\label{algo:GAE}
$\text{def GAE}(r_{1:T},v_{1:T}, \gamma, \lambda)$\\
$A_T=0$\\
\For{$t=T-1:1$}{
  $\delta_t = r_t + \gamma v_{t+1} - v_t$ \\
  $A_t = \delta_t + \gamma \lambda A_{t+1}$ // advantage \\
  $\targetV_t = A_t + v_t$ // Return
}
Return $(A_{1:T}, \targetV_{1:T})$
\end{algorithm}

\begin{algorithm}
\dontprintsemicolon
\caption{Actor critic with GAE}
\label{algo:AC}
Initialize  parameters $\vphi$,
environment state $s$\\
\Repeat{converged}
{
  $(s_1,a_1,r_1,\ldots,s_T)= \text{rollout}(s, \pi_{\vphi})$ \\
  $v_{1:T} = V_{\vphi}(s_{1:T})$ \\
  $(A_{1:T}, \targetV_{1:T}) = \stopgrad(\text{GAE}(r_{1:T}, v_{1:T}, \gamma, \lambda))$
  \\
  $\loss(\vphi) = \frac{1}{T}\sum_{t=1}^T \left[
    \lambda_{TD} (V_{\vphi}(s_t) - \targetV_t)^2
    -\lambda_{PG} A_t \log \pi_{\vphi}(a_t|s_t)
    -\lambda_{ent} \entropy(\pi_{\vphi}(\cdot|s_t)) \right]$ \\
  $\vphi := \vphi - \lr \nabla \loss(\vphi)$
  }
\end{algorithm}

\eat{
We can implement this as shown in \cref{algo:ACn},
following \citep[Algo 4.4]{Plaat2022}.

\begin{algorithm}
\dontprintsemicolon
\caption{A2C with $n$-step bootstrapping}
\label{algo:ACn}
Initialize actor parameters $\vtheta$, critic parameters $\vw$,
environment state $s$\\
\Repeat{converged}
{
         $(\tau,s) = \text{rollout}(s, \pi_{\vtheta})$ \\
         $(\vtheta, \vw) = \text{AC-step}(\vtheta, \vw, \tau)$ \\
}
.\\
$\text{def AC-step}(\vtheta,\vw,\tau)$: \\
Let $(s_0,a_0,r_0,\ldots,s_T)=\tau$ \\
\For{$t=0:T-1$}{
  $G_t = \left[\sum_{k=0}^{n-1} \gamma^k r_{t+k}  \right]
  + \gamma^n V_{\vw}(s_{t+n})$ \\
  $A_t = G_t - V_{\vw}(s_t)$
}
$\loss_{\vw} = \sum_t A_t^2$ \\
$J_{\vtheta} = \sum_t A_t \log \pi_{\vtheta}(a_t|s_t)$ \\
$\vw \assign \vw - \lr_{\vw} \nabla \loss_{\vw}$\\
$\vtheta \assign \vtheta + \lr_{\vtheta} \nabla J_{\vtheta}$ \\
       Return $(\vtheta,\vw)$
\end{algorithm}
}

\eat{
with weight proportional to $\lambda^n$
for $A_{\vw}^{(n)}(s_t,a_t)$, as in TD($\lambda$).
The average can be shown to be equivalent to
\begin{align}
A_{\vw}^{(\lambda)}(s_{t}, a_{t}) \defeq \sum_{\ell=0}^{\infty}
 (\gamma \lambda)^{\ell} \delta_{t+l}
 \label{eqn:ApolSum}
\end{align}
where $\delta_t= r_t + \gamma \Vapprox(s_{t+1}) - \Vapprox(s_t)$
is the TD error at time $t$.

In the episodic case, we can implement GAE
by modifying the $n$-step bootstrap AC algorithm
in \cref{algo:ACn}:
we just replace $Q_t$
by the $\lambda$-return $\Lambda_t$, which is computed by iterating backwards
from the end of the trajectory to the start:
\begin{align}
  \Lambda_t  = \begin{cases}
    r_t + \gamma (1-d_t)
    \left[ (1-\lambda) V(s_{t+1}) + \lambda \Lambda_{t+1} \right]
    &\mbox{if $t<T$}  \\
    V(s_T) &\mbox{otherwise}
    \end{cases}
  \end{align}
We can also develop a fully online version
using eligibility traces
(see \citep[Ch.12]{Suttonv2} for  details).
}

We can generalize this approach even further,
by using gradient estimators of the form
\be
\nabla J(\vtheta) = \expect{
\sum_{t=0}^{\infty} \Psi_t \nabla \log \policy_{\vtheta}(a_t|s_t)}
\ee
where $\Psi_t$ may be any of the following:
\begin{align}
  \Psi_t &=  \sum_{i=t}^{\infty} \gamma^i r_i
  & \text{Monte Carlo target}\\
  \Psi_t &=  \sum_{i=t}^{\infty} \gamma^i r_i
  -V_{\vw}(s_t) & \text{MC with baseline}\\
  \Psi_t &= A_{\vw}(s_t,a_t) & \text{advantage function} \\
  \Psi_t &= Q_{\vw}(s_t,a_t) & \text{Q function} \\
  \Psi_t &= r_t + \gamma V_{\vw}(s_{t+1}) - V_{\vw}(s_t) & \text{TD residual}
  \end{align}

\eat{
  \begin{align}
\Psi_t &= \sum_{t=0}^{\infty} r_t \; \text{: total reward of trajectory}\\
\Psi_t &= \sum_{t'=t}^{\infty} r_{t'} \; \text{: reward following $a_t$} \\
\Psi_t &= \sum_{t'=t}^{\infty} r_{t'} - b(s_t) \; \text{: baselined
version of previous formula} \\
\Psi_t &= \Qpol(s_t,a_t) \; \text{: action-value function} \\
\Psi_t &= \Apol(s_t,a_t) \; \text{: advantage function} \\
\Psi_t &= r_t + \Vpol(s_{t+1}) - \Vpol(s_t) \; \text{: TD residual}
  \end{align}
  }
See \citep{GAE} for details.

\subsection{Two-time scale actor critic algorithms}
\label{sec:twoTimeScale}
\label{sec:bilevel}

In standard AC, we update the actor and critic in parallel.
However, it is better
to let critic $V_{\vw}$ learn
using a faster learning rate (or more updates),
so that it reflects the value of the current policy $\pi_{\vtheta}$
more accurately, in order to get better gradient estimates
for the policy update.
This is 
known as \keyword{two timescale learning}
or \keywordDef{bilevel optimization}
\citep{Yu2017TD,Zhang2019timescale,Hong2023,Zheng2022AC,Lorraine2024}.
(See also \cref{sec:MBRLgame}, where we discuss RL
from a game theoretic perspective.)


An alternative approach is to alternate between updating
the policy and the value function, rather than updating them
simultaneously. This is called \keywordDef{phasic policy gradient}
\citep{Cobbe2020}.

\eat{
\subsection{Advantage-weighted regression}
\label{sec:AWR}

Recall from A2C (\cref{sec:A2C})
that the policy update takes the form
$\vtheta \assign \vtheta + \lr_{\vtheta} \gamma^t \delta \nabla_{\vtheta} \log \pi_{\vtheta}(a|s)$,
where $\delta = r + \gamma \Vapprox(s') - \Vapprox(s)$
is the TD error,
and we sample $(s,a,s',r)$ from the environment and current policy.
Note that $\delta$
is an approximation to the advantage function at the current state $s$.
If we treat the advantage function as frozen,
we can compute an
approximation to the gradient of the expected reward
using 
$\nabla_{\vtheta} J(\pi_{\vtheta})
\approx \expect{A_{\vw,\vtheta}(\vs,\va) \nabla_{\vtheta} \log \pi_{\vtheta}(\va|\vs)}$,
where the expectation is wrt states drawn
from the state distribution, $\vs \sim d_{\pi_{\vtheta}}$,
and actions drawn from the current policy,
$\va \sim \pi_{\vtheta}(\cdot|\vs)$.
We can compute an empirical $N$-sample
approximation to the corresponding
implicit loss by defining

From \cref{algo:A2C}, we see that
the policy loss for a sampled trajectory
has the form
\be
J(\vtheta) = \frac{1}{T} \sum_{t=1}^T
A_{\vw}(\vs_t,\va_t) \log \pi_{\vtheta}(\va_t|\vs_t)
\ee
In the case of a Gaussian policy,
this becomes
\be
\loss(\vtheta) = \frac{1}{T} \sum_{t=1}^T
A_{\vw}(\vs_t,\va_t) \half ||\va_t-\pi_{\vtheta}(\vs_t)||_2^2
\ee
which is a weighted regression loss.
The value function loss has the form
\be
\loss(\vw) = \frac{1}{T}
\sum_{t=1}^T (r_t + \gamma V_{\vw}(\vs_{t+1}) -V_{\vw}(\vs_t))^2
\ee
We can therefore alternate between optimizing the value function,
which is a nonlinear least squares problem,
and optimizing the policy,
which is a weighted  maximum likelihood problem.
This approach is known as 
\keywordDef{reward-weighted regression} 
\citep{Peters2007}
or \keywordDef{advantage-weighted regression} or \keywordDef{AWR}
\citep{Peng2019awr,Kozakowski2022}.

In the above approach,
we can think of the actor and critic as each computing a
``best response'' to the other player at each step.
This game-theoretic perspective is discussed in more detail
in \cref{sec:MBRLgame}.
There we mention that the best response strategy
can lead to unstable algorithms.
A better approach is the two time scale methods
 discussed in \cref{sec:twoTimeScale}.

 }

\subsection{Natural policy gradient methods}
\label{sec:naturalAC}
\label{sec:naturalPG}
\label{sec:NPG}

In this section, we discuss an improvement to policy
gradient methods that uses preconditioning to speedup convergence.
In particular, we replace gradient descent with  
\keywordDef{natural gradient descent} (\keywordDef{NGD})
\citep{Amari1998,Martens2020},
which we explain below. We then show how to combine it with actor-critic.

\subsubsection{Natural gradient descent}

NGD is a second order method
for optimizing the parameters of
(conditional) probability distributions, such as policies,
$\pi_{\vtheta}(\va|\vs)$.
It typically converges faster and more robustly than
SGD, but is computationally more expensive.

Before we explain NGD, let us review standard
SGD, which is an update of the following form
\be
\vtheta_{k+1} = \vtheta_k - \lr_k \vg_k
\ee
where $\vg_k = \nabla_{\vtheta} \loss(\vtheta_k)$
is the gradient of the loss at the previous parameter values,
and $\lr_k$ is the learning rate.
It can be shown that the above update is equivalent
to minimizing a locally linear approximation to the loss,
$\hat{\loss}_k$,
subject to the constraint that
the new parameters do not move too far (in Euclidean distance)
from the previous parameters:
\begin{align}
\vtheta_{k+1} &= \argmin_{\vtheta}
\hat{\loss}_k(\vtheta) \myst ||\vtheta-\vtheta_k||_2^2 \leq \epsilon
\\
\hat{\loss}_k(\vtheta) &= \loss(\vtheta_k) + \vg_k^\trans(\vtheta-\vtheta_k)
\end{align}
where the step size $\lr_k$ is  proportional to $\epsilon$.
This is called a \keywordDef{proximal update}
\citep{proximal}.

\begin{figure}
\centering
\begin{subfigure}[b]{0.45\textwidth}
\centering
\includegraphics[height=1.5in]{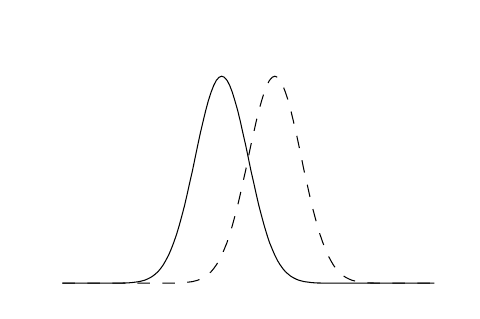}
\caption{ }
\end{subfigure}
~
\begin{subfigure}[b]{0.45\textwidth}
\centering
\includegraphics[height=1.5in]{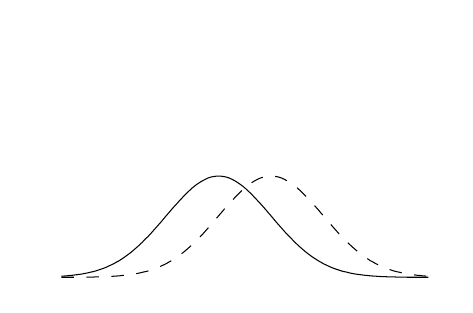}
\caption{ }
\end{subfigure}
\caption{
 Changing the mean of a Gaussian by a fixed amount (from solid to
 dotted curve)
can have more impact
when the (shared) variance is small (as in a)
compared to when the variance is large (as in b).
Hence the impact (in terms of prediction accuracy)
of a change to $\mu$
depends on where the optimizer is in
$(\mu,\sigma)$ space.
\figtaken{Figure 3 of \citep{Honkela2010}, reproduced from 
\citep{ValpolaPhD}.}
\figthanks{Antti Honkela}.
}
\label{fig:honkelaGaussians}
\end{figure}

One problem with the SGD update is that Euclidean distance
in parameter space does not make sense for probabilistic models.
For example, consider
comparing two Gaussians,
$p_{\vtheta} = p(y|\mu,\sigma)$
and
$p_{\vtheta'} = p(y|\mu',\sigma')$.
The (squared) Euclidean distance between the parameter vectors
decomposes as $||\vtheta - \vtheta'||_2^2 = (\mu-\mu')^2 + (\sigma-\sigma')^2$.
However, the predictive distribution has the form
$\exp(-\frac{1}{2\sigma^2} (y-\mu)^2)$,
so changes in $\mu$ need to be measured
relative to $\sigma$.
This is illustrated in 
\cref{fig:honkelaGaussians}(a-b),
which shows
two univariate Gaussian distributions (dotted and solid lines)
whose means differ  by $\epsilon$.
In \cref{fig:honkelaGaussians}(a), they share the same small variance
$\sigma^2$,
whereas in  \cref{fig:honkelaGaussians}(b), they share the same large variance.
It is clear that the difference in   $\mu$
matters much more (in terms of the effect on the distribution)
when the variance is small.
Thus we see that the two parameters interact with each other,
which the Euclidean distance cannot capture.

The key to NGD is to measure the notion of distance
between two probability distributions
in terms of the KL divergence.
This can be approximated in terms of
the \keywordDef{Fisher information matrix}  (FIM).
In particular, for any given input $\xdata$,
we have
\be
  \KLpq{p_{\params}(\vy|\xdata)}{p_{\params+\vdelta}(\vy|\xdata)}
  \approx
\half \vdelta^\trans \vFisher_{\xdata} \vdelta
\ee
where $\vFisher_{\xdata}$ is the FIM
\be
\vFisher_{\xdata}(\params)
= -\expectQ{\nabla^2 \log p_{\params}(\vy|\xdata)}{p_{\params}(\vy|\xdata)}
= \expectQ{(\nabla \log p_{\params}(\vy|\xdata))
(\nabla \log p_{\params}(\vy|\xdata))^\trans}{p_{\params}(\vy|\xdata)}
\ee
We  now replace the Euclidean distance
between the parameters,
$d(\vtheta_k, \vtheta_{k+1}) = ||\vdelta||_2^2$,
with 
\be
d(\vtheta_k, \vtheta_{k+1})
= \vdelta^\trans \vFisher_k \vdelta_k
\ee
where $\vdelta = \vtheta_{k+1}-\vtheta_k$
and $\vFisher_k = \vFisher_{\vx}(\vtheta_k)$
for a randomly chosen input $\vx$.
This gives rise to the following constrained
optimization problem:
\be
\vdelta_k = \argmin_{\vdelta}
\hat{\loss}_k(\vtheta_k + \vdelta) \myst
\vdelta^\trans \vFisher_k \vdelta \leq \epsilon
\ee
If we replace the constraint with a Lagrange multiplier,
we get the unconstrained objective:
\be
J_k(\vdelta) = \loss(\vtheta_k) + \vg_k^\trans \vdelta
+ \lr_k \vdelta^\trans \vFisher_k \vdelta 
\ee
Solving $J_k(\vdelta)=0$ gives the update
\be
\vdelta =  -\lr_k \vFisher_k^{-1} \vg_k
\ee
The term $\vFisher^{-1} \vg$
is called the \keywordDef{natural gradient}.
This is equivalent to  a preconditioned gradient update,
where we use the inverse FIM as a preconditioning matrix.
We can compute  the (adaptive) learning rate using
\be
\eta_k = \sqrt{\frac{\epsilon}{\vg_k^\trans \vF_k^{-1} \vg_k}}
\ee

Computing the FIM can be hard.
A simple approximation is to replace
the model's distribution with the empirical distribution.
In particular, define
 $\pemp(\xdata,\vy) = \frac{1}{N} \sum_{n=1}^N \delta_{\xdata_n}(\xdata) \delta_{\vy_n}(\vy)$,
$\pemp(\xdata) = \frac{1}{N} \sum_{n=1}^N \delta_{\xdata_n}(\xdata)$
and
$p_{\vtheta}(\vx,\vy) =  \pemp(\vx) p(\vy|\vx,\vtheta)$.
Then we can compute the \keywordDef{empirical Fisher}
\citep{Martens2016Thesis}
as follows:
\begin{align}
\vFisher(\vtheta) &= \expectQ{
\nabla \log p(\vy|\xdata,\params)
\nabla \log p(\vy|\xdata,\params)^\trans}
{p_{\vtheta}(\vx,\vy)} \\
 &\approx
\expectQ{
\nabla \log p(\vy|\xdata,\params)
\nabla \log p(\vy|\xdata,\params)^\trans}
{\pemp(\vx,\vy)} \\
&=
\frac{1}{|\data|}
\sum_{(\xdata,\vy) \in \data}
\nabla \log p(\vy|\xdata,\params)
\nabla \log p(\vy|\xdata,\params)^\trans
\end{align}

\subsubsection{Natural actor critic}

To apply NGD to RL, we can adapt
the A2C algorithm in \cref{algo:AC}.
In particular, define
\be
\vg_{kt} = \nabla_{\vtheta_k} A_t \log \pi_{\vtheta}(\va_t|\vs_t)
\ee
where $A_t$ is the advantage function at step $t$
of the random trajectory generated by the policy
at iteration  $k$.
Now we compute
\be
\vg_k = \frac{1}{T} \sum_{t=1}^T \vg_{kt},
\;
\vF_k = \frac{1}{T} \sum_{t=1}^T \vg_{kt} \vg_{kt}^\trans
\label{eqn:NACfisher}
\ee
and compute
$\vdelta_{k+1} = -\lr_k \vF_k^{-1} \vg_k$.
This approach is called \keywordDef{natural policy gradient}
\citep{Kakade2001,Rajeswaran2017}.

We can compute $\vF_k^{-1} \vg_k$ without having
to invert $\vF_k$ by using the conjugate gradient method,
where each CG step uses efficient
methods for Hessian-vector products \citep{Pearlmutter1994}.
This is called
\keywordDef{Hessian free optimization}  \citep{Martens2010}.
Similarly, we can efficiently compute
$\vg_k^\trans (\vF_{k}^{-1} \vg_k)$.

As a more accurate alternative
to the empirical Fisher,
\citep{Martens2015kfac} propose
the \keywordDef{KFAC} method,
which stands for ``Kronecker factored approximate curvature'';
this approximates the FIM of a DNN as a block diagonal
matrix, where each block is a Kronecker product
of two small matrices.
This was applied to policy gradient learning in \citep{Wu2017kfac}.

\eat{
\section{Bound optimization methods}

In this section, we describe methods that
aim to monotonically improve the performance of the policy
at each step,
similar to other \keywordDef{bound optimization} methods\footnote{
Bound optimization is also called \keywordDef{MM},
for majorize then maximize/minimize.
See \citep{Hunter04} for a tutorial.
} %
such as EM.
}

\subsection{Architectural issues}

It is common to use a single neural network
for both the actor and critic, but using different output heads:
a scalar output for the value function,
and a vector output for the policy.
For example,
the \keywordDef{Amago} method \citep{Grigsby2024,Grigsby2024amago2}
uses a transformer backbone.
To train the shared model, they 
 construct a unified objective
$\loss = E[\lambda_0 \loss_{TD} + \lambda_1 \loss_{PG}]$,
where the TD and policy gradient losses are dynamically normalized
using the \keywordDef{PopArt} method
of \citep{van-Hasselt2016,Hessel2019}
to allow
for a fixed set of hyper-parameter values for $\lambda_i$,
even as the range of the losses change over time.
(PopArt  stands for
``Preserving Outputs Precisely, while Adaptively Rescaling Targets''.)

However, \citep{Garcin2025} argue that it can be better to
use different networks for the actor and critic (at least when
using MLPs/CNNs), since they need to extract different kinds
of features.

\subsection{Deterministic policy gradient methods}
\label{sec:DPG}

In this section, we consider
an actor critic method that uses a deterministic
policy, that predicts a unique action for each state,
so $a_t = \policydet(s_t)$,
rather than $a_t \sim \polapprox(s_t)$.
This is trained
to match  the optimal
action from $Q_{\vw}(s,a)$.
Thus we can think of the resulting method
as a version of DQN designed for continuous actions.
(We require that the actions are continuous,
because we will take the Jacobian of the $Q$ function
wrt the actions.)

The benefit  of using a deterministic policy,
as opposed to a stochastic policy,
is that we can modify the policy gradient method
so that it can work off policy (e.g., using replay buffers)
without needing importance sampling,
in contrast to the methods discussed in
\cref{sec:offPolicy}.
In addition, the feedback signal for learning is based on the vector-valued gradient
of the value function,
which is more informative than  a scalar reward signal.

\subsubsection{Deterministic policy gradient theorem}

As before,
we define the value of a policy as the expected discounted
reward per state:
\begin{align}
\polval(\policydet)
\defeq  \expectQ{R(s,\policydet(s))}{\statmeasuredet(s)}
\end{align}
The \keywordDef{deterministic policy gradient theorem}
\citep{dpg} tells us that the gradient of this expression
is given by
\begin{align}
\nabla_{\vtheta} \polval(\policydet)
&=  \expectQ{\nabla_{\vtheta} \Qdet(s,  \policydet(s))}
      {\statmeasuredet(s)} \\
      &= \expectQ{
        \nabla_{\vtheta} \policydet(s)
        \nabla_a \Qdet(s, a)|_{a=\policydet(s)}
      }{\statmeasuredet(s)}
\label{eqn:DDPGgrad}
\end{align}
where $\nabla_{\vtheta} \policydet(s)$ is
the $N_\theta \times N_{A}$ Jacobian matrix,
and $N_A$ and $N_{\theta}$ are the dimensions
of $\calA$ and $\vtheta$, respectively.
The intuition for this equation is as follows:
the change in the expected value due to changing the parameters,
$\nabla_{\vtheta} \polval(\policydet) \in \real^{N_{\theta}}$,
is equal to
the change in the policy output (i.e., the actions)
due to changing the parameters,
$\nabla_{\vtheta} \policydet(s) \in \real^{N_{\theta} \times N_A}$
times
 the change in the expected value due to the change in the actions,
 $\nabla_a \Qdet(s, a) \in \real^{N_A}$.

For stochastic policies of the form
$\polapprox(a|s) = \policydet(s) + \text{noise}$,
the standard policy gradient theorem reduces
to the above form as the noise level goes to zero.

Note that the gradient estimate
in \cref{eqn:DDPGgrad}
integrates over the states but not over the actions,
which helps reduce the variance in gradient estimation
from sampled trajectories.
However, since the deterministic policy does not do any exploration,
we need to use an \keyword{off-policy method} for training.
This collects data from 
a stochastic behavior policy $\behavior$,
whose stationary state distribution is
$\statdistbehave$.
The original objective, $\polval(\policydet)$, is approximated by the following:
\begin{align}
\polvalbeh(\policydet)
\defeq \expectQ{\Vdet(s)}{\statdistbehave(s)}
= \expectQ{\Qdet(s, \policydet(s))}{\statdistbehave(s)}
\end{align}
with the off-policy deterministic policy gradient
from \citep{Degris2012}
is approximated by
\begin{align}
\label{eqn:OffDDPGgrad}
\nabla_{\vtheta} \polvalbeh(\policydet)
\approx \expectQ{\nabla_{\vtheta} \left[\Qdet(s,
    \policydet(s))\right]}
        {\statdistbehave(s)}
= \expectQ{\nabla_{\vtheta} \policydet(s)
 \nabla_a \Qdet(s, a)|_{a=\policydet(s)}  }{\statdistbehave(s)}
\end{align}
where we have a dropped a term that depends on
$\nabla_{\vtheta} \Qdet(s,a)$ and is hard to estimate
\citep{dpg}.

To apply \cref{eqn:OffDDPGgrad}, we may learn $\Qapprox \approx \Qdet$ with TD,
giving rise to the following updates:
\begin{align}
\delta &= r_t + \gamma \Qapprox(s_{t+1}, \policydet(s_{t+1}))
- \Qapprox(s_t, a_t) \\
\vw_{t+1} &\assign \vw_t + \lr_{\vw} \delta \nabla_{\vw} \Qapprox(s_t,a_t) \\
\vtheta_{t+1} &\assign \vtheta_t + \lr_{\vtheta} \nabla_{\vtheta} \policydet(s_t)
\nabla_a \Qapprox(s_t,a)|_{a=\policydet(s_t)} 
\end{align}
So we learn both a state-action critic $Q_{\vw}$
and an actor $\vmu_{\vtheta}$.
This method avoids importance sampling
in the actor update because
of the deterministic policy gradient,
and we avoids it  in the critic
update because of the use of Q-learning.

If $\Qapprox$ is linear in $\vw$, and uses features
of the form $\vphi(s,a) = \va^\trans \nabla_{\vtheta} \policydet(s)$,
then we say the function approximator for the critic
is \keywordDef{compatible} with the actor;
in this case, one can show that the above approximation
does not bias the overall gradient.


The basic off-policy
DPG method has been extended in various ways,
some of which we describe below.

\subsubsection{DDPG}
\label{sec:DDPG}
\label{sec:D4PG}

The \keywordDef{DDPG} algorithm of \citep{ddpg},
which stands for
``\keyword{deep deterministic policy gradient}'',
uses  the DQN method (\cref{sec:DQN})
to learn the $Q$ function,
and then uses this to evaluate the policy.
In more detail, the actor tries to minimize the output of the critic
\be
\loss_{\vtheta}(s) = Q_{\vw}(s, \mu_{\vtheta}(s))
\ee
where the loss is averaged over states $s$ drawn from the replay buffer.
The  critic tries to minimize the 1-step TD loss,
as in Q-learning:
\be
\loss_{\vw}(s,a,r,s') =
[Q_{\vw}(s,a) - (r + \gamma Q_{\overline{\vw}}(s',\mu_{\vtheta}(s')))]^2
\ee
where $Q_{\overline{\vw}}$ is the target critic network,
and the samples $(s,a,r,a')$ are drawn from a replay buffer.
(See \cref{sec:targetNetwork} for a discussion of target networks.)

The \keywordDef{D4PG} algorithm \citep{D4PG},
which stands for ``distributed distributional DDPG'',
extends DDPG to handle distributed training,
and to handle \keyword{distributional RL}
(see \cref{sec:distributional}).

\subsubsection{Twin Delayed DDPG (TD3)}
\label{sec:TD3}

The  \keywordDef{TD3} (``twin delayed deep deterministic'') method of
\citep{TD3}
extends DDPG  in 3 main ways.
First, it uses \keywordDef{target policy smoothing},
in which noise is added to the action, to encourage generalization:
\be
\tilde{\va} = \vmu_{\vtheta}(\vs) + \text{noise}
= \pi_{\vtheta}(\vs)
\ee
Second it uses \keywordDef{clipped double Q learning},
which is an extension of the double Q-learning discussed in
\cref{sec:double} to avoid
over-estimation bias.
In particular, the target values for TD learning are defined using
\be
\TargetV(r,\vs'; \overline{\vw}_{1:2}, \overline{\vtheta})
= r + \gamma \min_{i=1,2}
Q_{\overline{\vw}_i}(\vs', \pi_{\overline{\vtheta}}(\vs'))
\ee
Third, it uses \keywordDef{delayed policy updates},
in which it only updates the policy after the value function
has stabilized. (See also \cref{sec:bilevel}.)
See \cref{algo:TD3} for the pseudcode.

\begin{algorithm}
\dontprintsemicolon
\caption{TD3}
\label{algo:TD3}
Initialize environment state $\vs$,
policy parameters $\vtheta$,
target policy parameters $\overline{\vtheta}$,
critic parameters $\vw_i$,
target critic parameters $\overline{\vw}_i = \vw_i$,
replay buffer $\data=\emptyset$,
discount factor $\gamma$,
EMA rate $\rho$,
step size $\lr_{\vw}$, $\lr_{\vtheta}$.
\\
\Repeat{converged}
       {
         $\va = \mu_{\vtheta}(\vs) + \text{noise}$ \\
         $(\vs',r) = \text{step}(\va, \vs)$ \\
         $\data := \data \union
         \{ (\vs, \va, r, \vs') \}$ \\
         $\vs \assign \vs'$ \\
         \For{$G$ updates}{
         Sample a minibatch $\calB = \{(\vs_j,\va_j,r_j,\vs'_j)\}$
         from $\data$ \\
           $\vw = \text{update-critics}(\vtheta, \vw, \calB)$
         }
         Sample a minibatch $\calB = \{(\vs_j,\va_j,r_j,\vs'_j)\}$
         from $\data$\\
         $\vtheta = \text{update-policy}(\vtheta,\vw, \calB)$
        }
.\\
$\text{def update-critics}(\vtheta,\vw,\calB)$: \\
       Let $(\vs_j,\va_j,r_j,\vs'_j)_{j=1}^B = \calB$ \\
       \For{$j=1:B$}{
         $\tilde{\va}_j = \mu_{\overline{\vtheta}}(\vs'_j)
         + \text{clip}(\text{noise}, -c, c)$ \\
         $\targetV_j = r_j + \gamma \min_{i=1,2} Q_{\overline{\vw}_i}(\vs'_j,
         \tilde{\va}_j)$
       }
       \For{$i=1:2$}
    {
      $\loss(\vw_i) = \frac{1}{|\calB|} \sum_{(\vs,\va,r,\vs')_j \in
        \calB} (Q_{\vw_i}(\vs_j,\va_j) - \stopgrad(\targetV_j))^2$\\
      $\vw_i \assign \vw_i - \lr_{\vw} \nabla \loss(\vw_i)$ // Descent
      \\
      $\overline{\vw}_i := \rho \overline{\vw}_i
      + (1-\rho) \vw_i$       //Update target networks with EMA\\
    }

    Return $\vw_{1:N}, \overline{\vw}_{1:N}$\\
.\\
$\text{def update-actor}(\vtheta,\vw,\calB)$: \\
$J(\vtheta) =
 \frac{1}{|\calB|} \sum_{\vs \in \calB}
 \left(Q_{\vw_1}(\vs, \mu_{\vtheta}(\vs)) \right)^2$ \\
 $\vtheta \assign \vtheta +  \lr_{\vtheta} \nabla J(\vtheta)$ //
 Ascent \\
       $\overline{\vtheta} := \rho \overline{\vtheta}
       + (1-\rho) \vtheta$       //Update target policy network with EMA \\
         Return $\vtheta, \overline{\vtheta}$
\end{algorithm}

\subsubsection{Wasserstein Policy Optimization (WPO)}
\label{sec:WPO}

\begin{figure}
\centering
\includegraphics[height=2in]{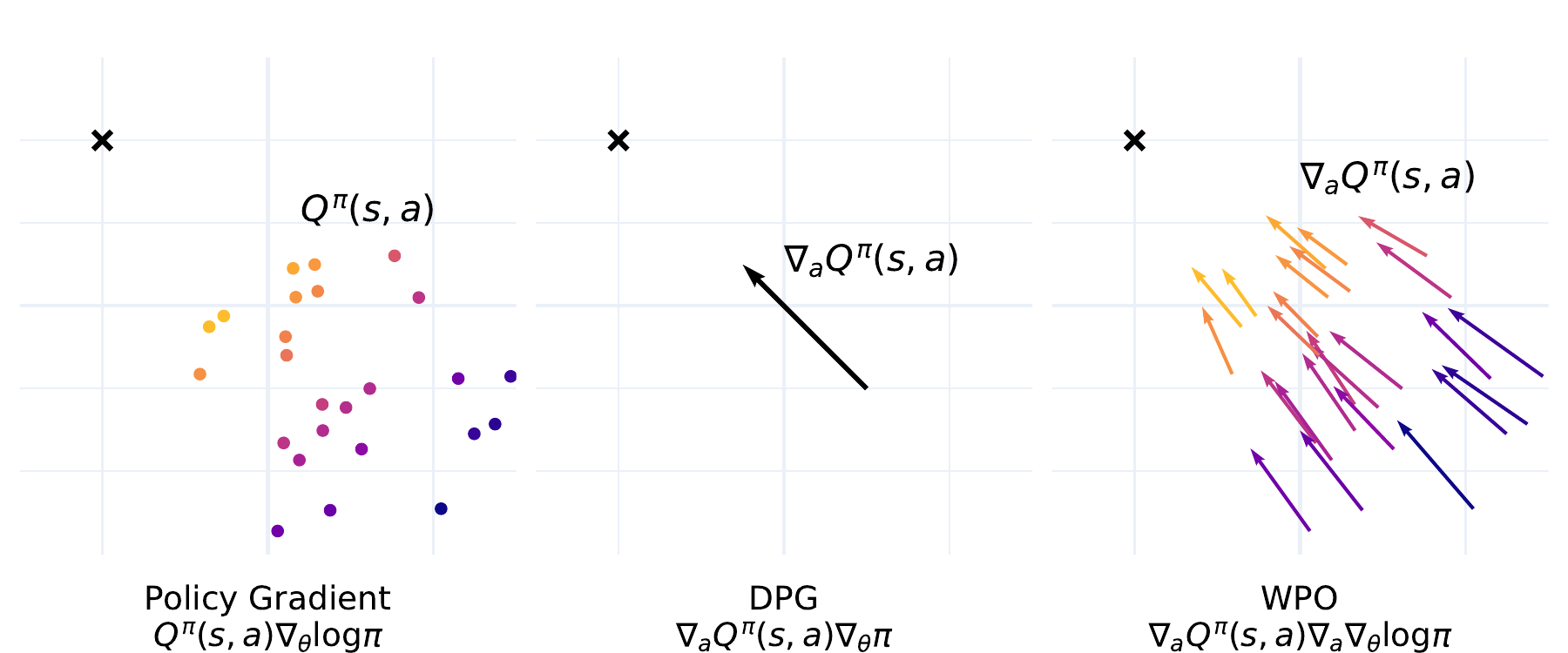}
\caption{
  Conceptual illustration of how
Wasserstein policy optimization (WPO)
combines elements of stochastic and deterministic policy gradient methods
for a 2d action space.
  Left: ``classic'' policy gradient. Samples are taken from a stochastic policy.
  Each sample contributes a scalar $Q^{\pi}(s,a)$  factor to the gradient.
  Middle: deterministic policy gradient (DPG).
  A deterministic action is chosen and the policy gradient depends on the gradient of
  $Q^{\pi}(s, a)$.
  Right: WPO. Samples are taken from a stochastic policy, as in classic policy gradient,
  but depend on the gradient of $Q^{\pi}$ with respect to the action, as in DPG.
foo
  \figtaken{Figure 1 of \citep{Pfau2025}}.
\figthanks{David Pfau}.
}
\label{fig:WPO}
\end{figure}

As we noted above,
one advantage of DPG-based methods is that they can use the
gradient of the value with respect to actions.
In \citep{Pfau2025} it was shown that by approximating
\keywordDef{Wasserstein gradient flows}
over the space of all parametric policies,
we arrive at an update very similar to DPG, but for general stochastic policies.\footnote{
Although the
method of \citep{Zhang2018wasserstein}
and the  \keywordDef{SVG(0)} method of \citep{svg}
also support stochastic policies,
they rely on the reparameterization trick,
which is not always applicable (e.g., if the policy is a mixture of Gaussians).
In addition, WPO makes use of natural gradients, whereas
these are first-order methods.
}
The derivation is somewhat complex, but the final algorithm is quite simple.
In particular, 
we should update the policy using the following:\footnote{
Note that $f(\va,\vtheta) = \log \pi_{\vtheta}(\va|\vs)$ is a scalar-valued
function of $\vtheta$ and $\va$ (for a fixed $\vs$);
the notation $\nabla_{\vtheta} (\nabla_{\va} f(\va,\vtheta)^\trans)$
is another way of writing the Jacobian matrix
$[\frac{\partial f}{\partial \theta_i \partial a_j}]_{ij}$.
}
\be
\vtheta_{t+1} = \vtheta_t + \calF^{-1}
\expectQ{
  \nabla_{\vtheta} \left( \nabla_{\va} \log \pi_{\vtheta}(\va|\vs)^\trans \right)
  \nabla_{\va} Q^{\pi_{\vtheta}}(\vs,\va))
}
{\vs \sim \data,  \va \sim \pi_{\vtheta}(\cdot|\vs)}
\ee
where $\calF$ is the Fisher information matrix (FIM):
\be
\calF = \expectQ{
  \nabla_{\vtheta} \log \pi_{\vtheta}(\va|\vs)
  \nabla_{\vtheta} \log \pi_{\vtheta}(\va|\vs)^\trans
}
{\vs \sim \data,  \va \sim \pi_{\vtheta}(\cdot|\vs)}
\ee
(Note that the states are sampled from a replay buffer, so may be off-policy,
but the actions are sampled from the current policy,
so are on-policy.)

If we ignore the FIM preconditioner $\calF^{-1}$, we see that the update
is  similar to the one used in the DPG theorem,
except we replace the Jacobian $\nabla_{\vtheta} \mu_{\vtheta}(\vs)$
with   $\nabla_{\vtheta} \left( \nabla_{\va} \log \pi_{\vtheta}(\va|\vs) \right)^\trans
\in \real^{N_{\theta} \times N_A}$.
Intuitively this captures the change in \keywordDef{probability flow}
over the action space due to a change in the parameters.
See \cref{fig:WPO} for an illustration.

However, the use of the FIM preconditioner
keeps the update
closer to the true gradient flow.
(Indeed, in the case of a Gaussian policy and quadratic value function,
WPO is exactly the Wasserstein gradient flow if you use the FIM,
but is very different if you don't.)
Furthermore, this preconditioner can avoid numerical issues which can arise
as the policy converges to a deterministic policy,
leading to a blowing up of the gradient term
$\nabla_{\va} \log \pi_{\vtheta}(\va|\vs)$.

In general, computing the FIM can be intractable.
However, 
 the authors assume the policy is a diagonal Gaussian, for which
the FIM is diagonal:
\be
\calF(\mu, \sigma) = 
\begin{bmatrix}
\mathrm{diag}\left( \frac{1}{\sigma_1^2}, \frac{1}{\sigma_2^2}, \dots, \frac{1}{\sigma_d^2} \right) & 0 \\
0 & \mathrm{diag}\left( \frac{2}{\sigma_1^2}, \frac{2}{\sigma_2^2}, \dots, \frac{2}{\sigma_d^2} \right)
\end{bmatrix}
\ee
This is fast to compute and invert.

After updating the policy with the above approach at each step,
they updated the critic using a conventional DQN-like update,
but one could use  more sophisticated critic updates, such as TD3.

\section{Policy improvement methods}
\label{sec:polImprovement}
\label{sec:policyImprovement}

In this section, we discuss methods that try to monotonically
improve performance of the policy at each step,
rather than just following the gradient,
which can result in a high variance estimate
where performance can increase or decrease at each step.
These are called \keywordDef{policy improvement} methods.
Our presentation is based on \cite{Queeney2024}.

\subsection{Policy improvement lower bound}

We start by stating a useful result from
\citep{Achiam2017}.
Let $\pi_k$ be the current policy at step $k$,
and let $\pi$ be any other policy (e.g., a candidate new one).
Let  $\statdistpolk$ be the normalized
discounted state visitation distribution for $\policy_k$,
defined in \cref{eqn:statdistpol}.
Let $A^{\pi_k}(s,a) = Q^{\pi_k}(s,a)-V^{\pi_k}(s)$
be the advantage function.
Finally, let the total variation distance
between two distributions be
given by
\be
\text{TV}(p,q) \defeq \half ||\vp-\vq||_1
= \half \sum_s |p(s) - q(s)| 
\ee
Then one can show \citep{Achiam2017} that
\be
J(\pi) - J(\pi_k)
\geq
  \frac{1}{1-\gamma}
\underbrace{
 \expectQ{
  \frac{\pi(a|s)}{\pi_k(a|s)}
  A^{\pi_k}(s,a)}{\statdistpolk(s) \policy_k(a|s)}
}_{L(\pi,\pi_k)}
-
  \frac{2 \gamma C^{\pi,\pi_k}}{(1-\gamma)^2}
  \expectQ{\text{TV}(\pi(\cdot|s),\pi_k(\cdot|s))}{\statdistpolk(s)}
\label{eqn:Jlowerbound}
  \ee
where
$C^{\pi,\pi_k} = \max_s |\expectQ{A^{\pi_k}(s,a)}{\pi(a|s)}|$.
In the above, $L(\pi,\pi_k)$ is a  surrogate objective,
and the second term is a penalty term.

If we can optimize this lower bound (or a stochastic approximation,
based on samples from the current policy $\pi_k$),
we can guarantee monotonic policy improvement (in expectation)
at each step.
We will replace this objective with a trust-region update
that is easier to optimize:
\be
\pi_{k+1} = \argmax_{\pi} L(\pi,\pi_k)
\myst   \expectQ{\text{TV}(\pi,\pi_k)(s)}{\statdistpolk(s)} \leq
\epsilon
\label{eqn:polImprovementBound}
\ee
The constraint bounds the worst-case performance decline at each
update.
The overall procedure becomes an approximate policy improvement
method.
There are various ways of implementing the above method in practice,
some of which we discuss below.
(See also \citep{Grudzien2022}, who propose a framework
called \keywordDef{mirror learning},
that justifies these ``approximations'' as in fact being
the optimal thing to do for a different kind of objective;
see also \citep{Vaswani2021}.)

\subsection{Trust region policy optimization (TRPO)}
\label{sec:TRPO}

In this section, we describe
the \keywordDef{trust region policy optimization}
(\keywordDef{TRPO}) method of \citep{TRPO}.
This implements an approximation to \cref{eqn:polImprovementBound}.
First, it leverages the fact that if
\be
\expectQ{\KLpq{\pi_k}{\pi}(s)}{\statdistpolk(s)} \leq \delta
\ee
then $\pi$ also satisfies the TV constraint
with $\delta=\frac{\epsilon^2}{2}$.
Next it considers a first-order expansion of the surrogate objective
to get
\be
L(\pi,\pi_k)
=
 \expectQ{
  \frac{\pi(a|s)}{\pi_k(a|s)}
  A^{\pi_k}(s,a)}{\statdistpolk(s) \policy_k(a|s)}
\approx \vg_k^\trans (\vtheta-\vtheta_k)
\ee
where $\vg_k = \nabla_{\vtheta} L(\pi_{\vtheta},\pi_k)|_{\vtheta_k}$.
Finally it considers a second-order expansion of the KL term
to get the approximate constraint
\be
\expectQ{\KLpq{\pi_k}{\pi}(s)}{\statdistpolk(s)}
\approx \half (\vtheta-\vtheta_k)^\trans \vF_k (\vtheta-\vtheta_k)
\ee
where $\vF_k = \vg_k \vg_k^\trans$ is an approximation to
the Fisher information matrix
(see \cref{eqn:NACfisher}).
We then use the update
\be
\vtheta_{k+1} = \vtheta_k + \eta_k \vv_k
\ee
where $\vv_k = \vF_k^{-1} \vg_k$ is the natural gradient,
and the step size is initialized to
$\eta_k = \sqrt{\frac{2 \delta}{\vv_k^\trans \vF_k \vv_k}}$.
(In practice we compute $\vv_k$ by approximately
solving the linear system $\vF_k \vv = \vg_k$
using conjugate gradient methods,
which just require matrix vector multiplies.)
We then use a backtracking line search procedure
to ensure the trust region is satisfied.

\eat{
We start with a useful fact that relate
the policy values of two arbitrary policies
\citep{Kakade2002RL}:
\begin{align}
\polval(\policy')-\polval(\policy)
=  \expectQ{\expectQ{\Apol(s,a)}{\policy'(a|s)}}
{\statmeasure_{\policy'}(s)}
 \label{eqn:PolicySwitchThm}
\end{align}
where $\policy$ can be interpreted as the current policy during policy optimization, and $\policy'$ a candidate new policy (such as the greedy policy wrt $\Qpol$).
As in the policy improvement theorem
(\cref{sec:policyIteration}),
if $\expectQ{\Apol(s,a)}{\policy'(a|s)} \ge 0$ for all $s$,
then $\polval(\policy') \ge \polval(\policy)$.
However, we cannot ensure this condition to hold
when function approximation is used,
as such a uniformly improving policy $\policy'$
may not be representable by our parametric family.
Therefore, nonnegativity of \cref{eqn:PolicySwitchThm}
is not easy to ensure.

One way to ensure monotonic improvement of
$\polval$ is to improve the policy conservatively.
Let us define
\begin{align}
\polvallin( \policy')
 \defeq  \expectQ{\expectQ{\Apol(s,a)}{\policy'(a|s)}}{\statmeasure_{\policy}(s)}
 = \expectQ{\frac{\policy'(a|s)}{\policy(a|s)}
   \Apol(s,a)}{\statmeasure_{\policy}(s)\policy(a|s)}
 \label{eqn:polvalLin}
\end{align}
In the above,
we have switched the state distribution from
$\statmeasure_{\policy'}$ in \cref{eqn:PolicySwitchThm}
to $\statmeasure_{\policy}$.
In \citep{TRPO} they show that
we have  we have the following lower bound on the improvement:
\be
J(\pi') - J(\pi) \geq L(\pi') - C \max_s \KLpq{\pi(a|s)}{\pi'(a|s)}
\label{eqn:JlowerBound}
\ee
where $C$  is a constant.
We can approximate the max over states with an expectation,
to get the  following constrained optimization problem:
\begin{align}
\label{eqn:TRPOconstr}
\argmax_{\policy'}
 \polvallin(\policy')
 \myst
 \expectQ{\KLpq{\policy(\cdot|s)}{\policy'(\cdot|s)}}
         {\statmeasure_{\policy}(s)} \leq \delta
\end{align}
for some threshold $\delta>0$.
We can then approximate the KL term with the Fisher information
matrix, as in \cref{sec:NGD}.
This gives the update
\be
\vtheta_{k+1} = \vtheta_k + \lr_k \vF(\vtheta_k)^{-1}
  \nabla_{\vtheta} L(\vtheta)\vert_{\vtheta_k}
  \ee
 We  can compute $\lr_k$ using a linesearch procedure:
 we start with largest value for $\lr_k$ allowed by the KL constraint,
 and reduce it until
 the lower bound is positive.
}

\subsection{Proximal Policy Optimization (PPO)}
\label{sec:PPO}


In this section, we describe the
the \keywordDef{proximal policy optimization}
or \keywordDef{PPO} method of \citep{PPO},
which is a simplification of TRPO.

We start by noting the following result:
\be
\expectQ{\text{TV}(\pi,\pi_k)(s)}{\statdistpolk(s)}
= \half \expectQ{| \frac{\pi(a|s)}{\pi_k(a|s)} - 1 |}
{(s,a) \sim \statdistpolk}
\ee
This holds provided the support of $\pi$ is contained
in the support of $\pi_k$ at every state.
We then use the following update:
\be
\pi_{k+1} = \argmax_{\pi} 
\expectQ{
  \min\left( \rho_k(s,a)  A^{\pi_k}(s,a),
  \tilde{\rho}_k(s,a)  A^{\pi_k}(s,a)
  \right)}{(s,a) \sim \statdistpolk}
\label{eqn:PPOupdate}
\ee
where $\rho_k(s,a) = \frac{\pi(a|s)}{\pi_k(a|s)}$
is the likelihood ratio, and
$\tilde{\rho}_k(s,a)= \text{clip}(\rho_k(s,a), 1-\epsilon,
1+\epsilon)$,
where $\text{clip}(x,l,u) = \min(\max(x,l),u)$.
\eat{
We start by considering the following unconstrained objective,
based on \cref{eqn:JlowerBound}:
\begin{align}
  \polval(\policy') = \polvallin(\policy')
  - \beta\expectQ{\KLpq{\policy(\cdot|s)}{\policy'(\cdot|s)}}
  {\statmeasure_{\policy}(s)}
\end{align}
where
\begin{align}
\polvallin( \policy')
 = \expectQ{\frac{\policy'(a|s)}{\policy(a|s)}
   \Apol(s,a)}{\statmeasure_{\policy}(s)\policy(a|s)}
\end{align}
Unfortunately, the likelihood ratio
\be
l(s,a) = \frac{\pi'(a|s)}{\pi(a|s)}
\ee
which is needed to compute $\polvallin(\policy')$
might become very large or small, if $\pi'$ deviates too much
from $\pi$. To ensure stability we can replace it with
\be
\tilde{l}(s,a) = 
\text{clip}(r(s,a;\vtheta), 1-\epsilon, 1+\epsilon)
\ee
which ensures $|l(s,a;\vtheta) - 1| \leq \epsilon$.
Furthermore, we can drop the KL penalty and just optimize
\be
\polval(\vtheta) = \expect{\min(l(s,a;\vtheta) \Apol(s,a), \;\;
\tilde{l}(s,a;\vtheta) \Apol(s,a))}
\ee
}
See \citep{Grudzien2022} for a theoretical justification
for these simplifications.
Furthermore,
this can be modified to ensure monotonic
improvement as discussed in \citep{Wang2019PPO},
making it a true bound optimization method.

Some pseudocode for PPO (with GAE)
is given in \cref{algo:PPO}.
It is basically identical to the AC code in \cref{algo:AC},
except the policy loss has the form
$\min(\rho_t A_t, \tilde{\rho}_t A_t)$
instead of
$A_t \log \pi_{\vphi}(a_t|s_t)$,
and we perform multiple policy updates per rollout,
for increased sample efficiency.
For all the implementation details, see
\url{https://iclr-blog-track.github.io/2022/03/25/ppo-implementation-details/}.

\begin{algorithm}
\dontprintsemicolon
\caption{PPO  with GAE}
\label{algo:PPO}
Initialize  parameters $\vphi$, 
environment state $s$\\
\For{iteration $k=1,2,\ldots$}
    {
        $\vphi_{\old} \assign \vphi$ \\
  $(s_1,a_1,r_1,\ldots,s_T)= \text{rollout}(s, \pi_{\vphi_{\old}})$ \\
  $v_{t} = V_{\vphi}(s_{t}) \; \text{for } t=1:T$ \\
  $(A_{1:T}, \targetV_{1:T}) = \text{GAE}(r_{1:T}, v_{1:T}, \gamma, \lambda)$
  \\
  \For{$m=1:M$}{
  $\rho_{t} = \frac{\pi_{\vphi}(a_t|s_t)}{\pi_{\vphi_{\old}}(a_t|s_t)}
  \; \text{for } t=1:T$  \\
  $\tilde{\rho}_t = \text{clip}(\rho_t) \text{ for } t=1:T$ \\
  $\loss(\vphi) = \frac{1}{T}\sum_{t=1}^T \left[
    \lambda_{TD} \text{clip}((V_{\vphi}(s_t) - \targetV_t)^2)
    -\lambda_{PG} \min(\rho_t A_t, \tilde{\rho}_t A_t)
    -\lambda_{ent} \entropy(\pi_{\vphi}(\cdot|s_t)) \right]$ \\
  $\vphi := \vphi -  \lr \nabla_{\vphi} \loss(\vphi)$ \\
  }
}
\end{algorithm}

\eat{
\begin{algorithm}
\dontprintsemicolon
\caption{PPO with $1$-step critic}
\label{algo:PPO}
Initializeparameters $\vphi$,
environment state $s$\\
\Repeat{converged}
{
  $(\tau,s) = \text{rollout}(s, \pi_{\vtheta})$ \\
  \For{$k=1:K$}{
    $\vphi_{\old}=\vphi$ \\
    $\vphi = \text{PPO-step}(\tau, \vphi,  \vphi_{\old})$ \\
    }
}
.\\
$\text{def PPO-step}(\tau, \vphi, \vphi_{\old})$: \\
Let $(s_0,a_0,r_0,\ldots,s_T)=\tau$ \\
\For{$t=0:T-1$}{
  $G_t = r_{t+1} + \gamma V_{\vphi}(s_{t+1})$ \\
  $A_t = \stopgrad(G_t - V_{\vphi}(s_t))$ \\
  $r_t = \frac{\pi_{\vphi}(a_t|s_t)}{\pi_{\vphi_{\old}}(a_t|s_t)}$
  \\
  $\tilde{r}_t = \text{clip}(r_t)$ \\
}
$\loss = \frac{1}{T} \sum_t \left[
  \lambda_V (V_{\vphi}(s_t) - G_t)^2
  -\lambda_{\pi} \min(r_t A_t, \tilde{r}_t A_t)
  -\lambda_{ent} \entropy(\pi_{\vphi}(\cdot|s_t)) \right]$ \\
$\vphi \assign \vphi - \lr \nabla \loss$\\
       Return $\vphi$
\end{algorithm}
}


\eat{
\begin{algorithm}
\dontprintsemicolon
\caption{PPO with $1$-step critic}
\label{algo:PPO}
Initialize actor parameters $\vtheta$, critic parameters $\vw$,
environment state $s$\\
\Repeat{converged}
{
  $(\tau,s) = \text{rollout}(s, \pi_{\vtheta})$ \\
  \For{$k=1:K$}{
    $\vtheta_{\old}=\vtheta$ \\
    $(\vtheta, \vw) = \text{PPO-step}(\vtheta, \vw, \tau, \vtheta_{\old})$ \\
    }
}
.\\
$\text{def PPO-step}(\vtheta,\vw,\tau,\vtheta_{\old})$: \\
Let $(s_0,a_0,r_0,\ldots,s_T)=\tau$ \\
\For{$t=0:T-1$}{
  $Q_t = r_{t+1} + \gamma V_{\vw}(s_{t+1})$ \\
  $A_t = \stopgrad(Q_t - V_{\vw}(s_t))$ \\
  $r_t = \frac{\pi_{\vtheta}(a_t|s_t)}{\pi_{\vtheta_{\old}}(a_t|s_t)}$
  \\
  $\tilde{r}_t = \text{clip}(r_t)$ \\
}
$\loss_{\vw} = \frac{1}{T} \sum_t (V_{\vw}(s_t) - \stopgrad(Q_t))^2$ \\
$\loss_{\vtheta} = -\frac{1}{T} \sum_t 
\min(r_t A_t, \tilde{r}_t A_t)$ \\
$\vw \assign \vw - \lr_{\vw} \nabla \loss_{\vw}$\\
$\vtheta \assign \vtheta - \lr_{\vtheta} \nabla \loss_{\vtheta}$ \\
       Return $(\vtheta,\vw)$
\end{algorithm}
}

\subsubsection{Simplified form of the clipping term}

In \url{https://spinningup.openai.com/en/latest/algorithms/ppo.html},
they propose a simplified form of the clipping term, which is easier
to understand.
First we rewrite the loss (inside the expectation
of \cref{eqn:PPOupdate})
in shorthand form as follows:
\be
L(s,a,\theta_k,\theta) =
\min(\rho A, \text{clip}(\rho) A)
\ee
where $\rho=\frac{\pi_{\theta}(a|s)}{\pi_{\theta_k}(a|s)}$
and $A=A^{\pi_{\theta_k}}(s,a)$.
If the advantage is positive, $A>0$,
the action was better than expected,
so we want to increase the probability $\rho$, but not too much
(so the new policy stays close the previous one).
The term $\text{clip}(\rho)A$ restricts $\rho A$ to be
in $[(1-\epsilon) A, (1+\epsilon) A]$.
The outer min selects the smaller of $\rho A$ and
this clipped value.
If $\rho<1-\epsilon$, then $\rho A < (1-\epsilon) A$,
so the min operator will naturally select $\rho A$
without needing the lower clip.
So in this case we can use a loss of the form
$\min(\rho A, (1+\epsilon) A)$.
If the advantage is negative, $A<0$,
a similar argument shows we can use a loss of the form
$\min(\rho A, (1-\epsilon) A)$.
We can combine these two cases by defining
\be
L(s,a,\theta_k,\theta) = \min(\rho A, g(\epsilon,A))
\ee
where
\begin{align}
  g(\epsilon,A)
  = \begin{cases}
    (1+\epsilon) A & \mbox{if $A \geq 0$} \\
    (1-\epsilon) A & \mbox{if $A < 0$} 
  \end{cases}
  \end{align}

\subsubsection{PPO for diffusion policies}

The PPO algorithm relies on being able to compute the likelihood
of a candidate action.
This is difficult to do for diffusion policies.
In \citep{Ren2024}, they propose \keywordDef{DPPO},
which is PPO for diffusion models.
The key idea is to treat each step of the diffusion process
as a step of an ``inner'' MDP, nested inside the outer
(main) MDP, and then to apply PPO to this combined system.
Unfortunately this results in very long horizon tasks.
In \citep{Yang2025diffusion},
they propose noise-conditoned diffusion policy
optimization, or \keywordDef{NCDPO},
which fixes the noise sequence for all the diffusion steps,
and then deterministically backpropagates gradients through
the entire denoising chain.
The result is a single MDP,  in which the policy
happens to be implemented by a deep network,
rather than a set of two nested MDPs.
(See \citep{Uehara2024} for a more general review of RL for diffusion
models.)

\subsubsection{Simple policy optimization}

More recently, \citep{Xie2025SPO} propose
\keywordDef{Simple Policy Optimization} or \keywordDef{SPO},
which improves upon ratio clipping,
offering stronger theoretical properties and better
constraining the probability ratio within the trust region.

\subsection{Variational Maximum a Posteriori Policy Optimization (VMPO)}
\label{sec:VMPO}

In this section, we discuss
the \keywordDef{VMPO} algorithm of \citep{VMPO},
which is an on-policy extension of the earlier
off-policy \keyword{MPO}  (MAP policy optimization)
algorithm that we discuss in \cref{sec:MPO}.
VMPO was originally explained in terms of the ``control as inference'' framework
(see \cref{sec:inferRL}),
but we can also view it as
a  constrained policy improvement method,
based on   \cref{eqn:polImprovementBound}.
In particular, VMPO leverages the fact that if
\be
\expectQ{\KLpq{\pi}{\pi_k}(s)}{\statdistpolk(s)} \leq \delta
\ee
then $\pi$ also satisfies the TV constraint
with $\delta=\frac{\epsilon^2}{2}$.

Note that here the KL is reversed compared to TRPO
in \cref{sec:TRPO}.
This new version will encourage $\pi$
to be mode-covering, so it will naturally
have high entropy, which can result in improved robustness.
Unfortunately, this kind of KL is harder to compute,
since we are taking expectations wrt the unknown distribution $\pi$.

To solve this problem, VMPO adopts an EM-type approach.
In the E step, we compute a non-parametric version
of the state-action distribution given by the unknown
new policy:
\be
\psi(s,a) = \pi(a|s) \statdistpolk(s)
\ee
The optimal new distribution is given by
\be
\psi_{k+1} = \argmax_{\psi}
\expectQ{A^{\pi_k}(s,a)}{\psi(s,a)}
\myst \KLpq{\psi}{\psi_k} \leq \delta
\ee
where
$\psi_k(s,a) = \pi_k(a|s) \statdistpolk(s)$.
The solution to this  is
\begin{align}
  \psi_{k+1}(s,a) &=  \statdistpolk(s) \pi_k(a|s) w(s,a) \\
  w(s,a) &= \frac{\exp(A^{\pi_k}(s,a)/\lambda^*)}{Z(\lambda^*)} \\
  Z(\lambda) &= \expectQ{\exp(A^{\pi_k}(s,a)/\lambda)}{(s,a)\sim
    \statdistpolk} \\
  \lambda^* &= \argmin_{\lambda \geq 0} \lambda \delta + \lambda \log Z(\lambda)
  \end{align}
In the M step, we project this target distribution back onto the space
of parametric policies, while satisfying the KL trust region
constraint:
\be
\pi_{k+1} = \argmax_{\pi}
\expectQ{w(s,a) \log \pi(a|s)}{(s,a) \sim \statdistpolk}
\myst \expectQ{\KLpq{\psi_k}{\psi}(s)}{\statdistpolk} \leq \delta
\ee

\section{Off-policy methods}
\label{sec:offPolicy}
\label{sec:offpolicy}

In many cases, it is useful to train a policy
using data collected from a distinct
\keywordDef{behavior policy}
$\behavior(a|s)$ that is
not the same as the \keywordDef{target policy} $\policytgt(a|s)$
that is being learned.
For example, this could be data collected
from earlier trials or parallel workers
(with different parameters $\vtheta'$)
and stored in a \keywordDef{replay buffer},
or it could be \keywordDef{demonstration data} from human experts.
This is known as \keywordDef{off-policy RL},
and can be much more sample efficient than the on-policy 
methods we have discussed so far, since these methods
can use data from multiple sources.
However, off-policy methods are more complicated,
as we will explain below.

The basic difficulty  is that the target policy  that we want to learn
may want to try an action in a state that has
not been experienced before in the existing data,
so there is no way to predict the outcome of this new $(s,a)$
pair.
In this section, we tackle this problem by assuming
that the  target policy is not too different from the behavior policy,
so that the ratio $\policytgt(a|s)/\policybeh(a|s)$
is bounded, which allows us to use methods based on importance
sampling.
In the online learning setting, we can ensure this property
by using conservative incremental updates to the policy.
Alternatively we can use policy gradient methods
with various regularization methods, as we discuss below.

In \cref{sec:offlineRL}, we discuss offline RL,
which is an extreme instance of off-policy RL where we
have a fixed behavioral dataset, possibly generated from an unknown
behavior policy,
and we can never collect any new data.
Some of the solution methods for offline RL
build on concepts discussed here.

\subsection{Policy evaluation using importance sampling}
\label{sec:offpolicyRL-IS}

Assume we have a dataset of the form
$\data = \{\traj^{(i)}\}_{1 \le i \le n}$, where each trajectory is a sequence
$\traj^{(i)}=(s_0^{(i)},a_0^{(i)},r_0^{(i)},s_1^{(i)}\ldots)$,
where the actions are sampled according to a behavior policy
$\policybeh$, and
the reward and next states are sampled
according to the reward and transition models.
We want to use this offline dataset to evaluate the performance of some
target policy $\policytgt$;
this is called \keywordDef{off-policy policy evaluation} or \keywordDef{OPE}.
If the trajectories $\traj^{(i)}$ were sampled from $\policytgt$.
we could use the standard Monte Carlo estimate:
\begin{align}
  \hat{J}(\policytgt)
\defeq \frac{1}{n} \sum_{i=1}^n  \sum_{t=0}^{T-1} \gamma^t r_t^{(i)}
\end{align}
However, since the trajectories are sampled from $\policybeh$,
we use \keywordDef{importance sampling} (IS)
to correct for the distributional mismatch,
as first proposed in \citep{Precup2000}.
This gives
\begin{align}
\label{eqn:offpolicyrl-basicis}
  \polvalis(\policytgt)
  \defeq \frac{1}{n} \sum_{i=1}^n
  \frac{p(\traj^{(i)}|\policytgt)}{p(\traj^{(i)}|\policybeh)}
  \sum_{t=0}^{T-1} \gamma^t r_t^{(i)}
\end{align}
It can be verified that $\expectQ{\polvalis(\policytgt)}{\policybeh}
= \polval(\policytgt)$, that is, $\polvalis(\policytgt)$ is \keywordDef{unbiased},
provided that $p(\traj|\policybeh)>0$
whenever $p(\traj|\policytgt)>0$.
The \keywordDef{importance ratio},
$\frac{p(\traj^{(i)}|\policytgt)}{p(\traj^{(i)}|\policybeh)}$,
is used to compensate for the fact that
the data is sampled from $\policybeh$ and not $\policytgt$.
It can be simplified as follows:
\begin{align}
\label{eqn:offpolicyrl-isratio}
\frac{p(\traj|\policytgt)}{p(\traj|\policybeh)}
= \frac{p(s_0) \prod_{t=0}^{T-1} \policytgt(a_t|s_t) \ptran(s_{t+1}|s_t,a_t) \preward(r_t|s_t,a_t,s_{t+1})}{p(s_0) \prod_{t=0}^{T-1} \policybeh(a_t|s_t) \ptran(s_{t+1}|s_t,a_t) \preward(r_t|s_t,a_t,s_{t+1})}
= \prod_{t=0}^{T-1} \frac{\policytgt(a_t|s_t)}{\policybeh(a_t|s_t)}
\end{align}
This simplification makes it easy to apply IS,
as long as the target and behavior policies are known.
(If the behavior policy is unknown,
we can estimate it from $\data$,  and replace $\policybeh$
by its estimate $\hat{\policybeh}$.
For convenience, define the
\keywordDef{per-step importance ratio} at time $t$ by
\be
\psiRatio_t(\traj) \defeq \policytgt(a_t|s_t) / \policybeh(a_t|s_t)
\ee
We can reduce the variance of the estimator
by noting that the reward $r_t$ is independent of the trajectory
beyond time $t$.
This leads to a \keywordDef{per-decision importance sampling} variant:
\begin{align}
\label{eqn:offpolicyrl-pdis}
  \polvalpdis(\policytgt)
  \defeq \frac{1}{n} \sum_{i=1}^n \sum_{t=0}^{T-1}
  \prod_{t' \le t} \psiRatio_{t'}(\traj^{(i)}) \gamma^t r_t^{(i)}
\end{align}

\subsection{Off-policy actor critic methods}
\label{sec:offPolicyPG}

In this section, we discuss how to extend actor-critic
methods to work with off-policy data.

\subsubsection{Learning the critic using V-trace}
\label{sec:Vtrace}

In this section
we build on \cref{sec:offpolicyRL-IS} to develop a practical method,
known as \keywordDef{V-trace} \citep{Espeholt2018},
to estimate the value function for a target policy
using off-policy data.
(This is an extension of the earlier
\keywordDef{Retrace} algorithm \citep{Munos16},
which estimates the $Q$ function using off-policy data.)

First consider the $n$-step target value for $V(s_i)$
in the on-policy case:
\begin{align}
  V_i  &= V(s_i)
  + \sum_{t=i}^{i+n-1} \gamma^{t-i} r_t + \gamma^n V(s_{i+n}) \\
  &= V(s_i)
  + \sum_{t=i}^{i+n-1} \gamma^{t-i}
  \underbrace{(r_t + \gamma V(s_{t+1}) - V(s_t))}_{\delta_t}
\end{align}
where we define $\delta_t  = (r_t + \gamma V(s_{t+1}) - V(s_t))$
as the TD error at time $t$.
To extend this to the off-policy case, we use the per-step
importance ratio trick. However, to bound the variance
of the estimator, we truncate the IS weights.
In particular, we define
\begin{align}
  c_t &= \min\left(\overline{c},
  \frac{\policytgt(a_t|s_t)}{\policybeh(a_t|s_t)} \right),
  \;
  \rho_t = \min\left(\overline{\rho},
  \frac{\policytgt(a_t|s_t)}{\policybeh(a_t|s_t)} \right)
  \label{eqn:rhot}
\end{align}
where $\overline{c}$ and $\overline{\rho}$
are hyperparameters.
We then define the V-trace target value for $V(s_i)$ as
\begin{align}
  v_i
  &= V(s_i)
  + \sum_{t=i}^{i+n-1} \gamma^{t-i} \left( \prod_{t'=i}^{t-1} c_{t'}
  \right)   \rho_t \delta_t
\end{align}
Note that we can compute these targets recursively using
\be
v_i = V(s_i) + \rho_i  \delta_i + \gamma c_i (v_{i+1} - V(s_{i+1}))
\ee

The product of the weights $c_i \ldots c_{t-1}$ (known as the ``trace'')
measures how much a temporal difference $\delta_t$
at time $t$ impacts the update of the value function
at earlier time $i$.
If the policies are very different, the variance of this product
will be large. So the truncation parameter $\overline{c}$
is used to reduce the variance.
In  \citep{Espeholt2018}, they find $\overline{c}=1$ works best.

The use of the target $\rho_t \delta_t$  rather than $\delta_t$
means we are evaluating the value function for a policy
that is somewhere between $\policybeh$ and $\policytgt$.
For $\overline{\rho}=\infty$ (i.e., no truncation),
we converge to the value function $V^{\policytgt}$,
and for $\overline{\rho} \ra 0$,
we converge to the value function $V^{\policybeh}$.
In  \citep{Espeholt2018}, they find $\overline{\rho}=1$ works best.
(An alternative to clipping the importance weights
is to use a resampling technique, and then use unweighted samples
to estimate the value function \citep{Schlegel2019}.)

In \citep{Schmitt2020},
they analyse the variance of the V-trace estimator,
used to compute $\rho_t$ in  \cref{eqn:rhot}.
They show that to keep this bounded,
it is necessary to mix some off-policy data
(from the replay buffer)
with some fresh online data from the current policy.

Note that if $\overline{c}=\overline{\rho}$,
then $c_i=\rho_i$. This gives rise to the simplified form
\begin{align}
  v_t
  &= V(s_t)
  + \sum_{j=0}^{n-1} \gamma^{j} \left( \prod_{m=0}^{j} c_{t+m}
  \right)   \delta_{t+j}
  \label{eqn:Vtrace}
\end{align}
We can use the above V-trace targets to learn an approximate
value function by minimizing the usual $\ell_2$ loss:
\be
\loss(\vw) = \expectQ{(v_t - V_{\vw}(s_t))^2}{t \sim \data}
\ee
\eat{
the gradient of which has the form
\be
\nabla \loss(\vw) = 2 \expectQ{(v_t - V_{\vw}(s_t))
  \nabla_{\vw}  V_{\vw}(s_t)}{t \sim \data}
\ee
}

\subsubsection{Learning the actor}
\label{sec:offpolPG}

We now discuss how to update the actor
using an off-policy estimate of the policy gradient.
We start by defining the objective to be
the expected value of the new policy,
where the states are drawn from the behavior
policy's state distribution, but the actions
are drawn from the target policy:
\be
J_{\behavior}(\policy_{\vtheta})
= \sum_s \statdistbehave(s) \Vpol(s)
= \sum_s \statdistbehave(s) \sum_a \policy_{\vtheta}(a|s) \Qpol(s,a)
\label{eqn:offPolicyJ}
\ee
Differentiating this and
ignoring the term $\nabla_{\vtheta} \Qpol(s,a)$, as suggested by \citep{Degris2012},
gives a way to (approximately) estimate the
\keywordDef{off-policy policy-gradient}
using a one-step IS correction ratio:
\begin{align}
\nabla_{\vtheta}
 J_{\behavior}(\policy_{\vtheta})
  &\approx \sum_s \sum_a
  \statdistbehave(s) \nabla_{\vtheta} \policy_{\vtheta}(a|s) \Qpol(s,a) \\
  &= \expectQ{  \frac{\policy_{\vtheta}(a|s)}{\behavior(a|s)}
  \nabla_{\vtheta} \log \policy_{\vtheta}(a|s) \Qpol(s,a)}
  {\statdistbehave(s), \behavior(a|s)}
\end{align}

In practice, we can approximate $Q_{\pi}(s_t,a_t)$
by $q_t = r_t + \gamma v_{t+1}$,
where $v_{t+1}$ is the V-trace estimate for state $s_{t+1}$.
If we use $V(s_t)$ as a baseline, to reduce the variance,
we get the following gradient estimate for the policy:
\be
\nabla J(\vtheta) = \expectQ{
  \rho_t \nabla_{\vtheta} \log \pi_{\vtheta}(a_t|s_t)
  (r_t + \gamma v_t - V_{\vw}(s_t))}{t \sim \data}
\ee

We can also replace the 1-step IS-weighted TD error
$\rho_t (r_t + \gamma v_t - V_{\vw}(s_t))$
with an IS-weighted GAE value
by modifying the generalized advantage estimation method
in \cref{sec:GAE}
to replace $A_t$ with $\rho_t A_t$.
See \cref{algo:ac-off} for some pseudocode.

\eat{
We denote the IS-weighted GAE estimate  as $A_t^{\rho}$.\footnote{
  For an implementation, see
  \url{https://github.com/google-deepmind/rlax/blob/master/rlax/\_src/multistep.py\#L39}
  }
}

\begin{algorithm}[H]
\SetAlgoLined
\DontPrintSemicolon
\caption{Actor-Critic (off-policy)}
\label{algo:ac-off}

\KwData{Learning rates $\alpha_\theta, \alpha_\phi$, discount $\gamma$, GAE $\lambda$, tapering func. $\tau(\cdot)$}
\textbf{Initialize}: Actor $\pi_\theta$, Critic $V_\phi$, Target $V_{\phi'} \leftarrow V_\phi$, Replay buffer $\mathcal{D}$\;

\For{episode = 1 to M}{
    Sample initial state $s_0$\;
    Initialize empty episode buffer $\mathcal{E} \leftarrow []$\;
    \For{$t=0$ to $T-1$}{
        Get action $a_t \sim \behavior(\cdot|s_t)$\;
        Execute $a_t$, observe $s_{t+1}$ and $r_t$\;
        Store $(s_t, a_t, r_t, s_{t+1}, \behavior(a_t|s_t))$ in $\mathcal{E}$\;
        $s_t \leftarrow s_{t+1}$\;
    }
    Store trajectory $\mathcal{E}$ in the replay buffer $\mathcal{D}$\;

    \If{enough data in $\mathcal{D}$}{
        Sample a batch of trajectories $\{\mathcal{E}_i\}$ from $\mathcal{D}$\;
        \For{each trajectory $\mathcal{E}_i$ in the batch}{
            \tcp{Calculate Advantage Estimates using GAE}
            \For{$t = T-1, \dots, 0$}{
                $\delta_t \leftarrow r_t + \gamma V_{\phi'}(s_{t+1}) - V_{\phi}(s_t)$ \tcp*{TD Error}
                $\hat{A}_t \leftarrow \delta_t + \gamma \lambda \hat{A}_{t+1}$ \tcp*{Advantage}
            }

            \tcp{Update Critic}
            $L(\phi) \leftarrow \frac{1}{T} \sum_{t=0}^{T-1} (r_t + \gamma V_{\phi'}(s_{t+1}) - V_\phi(s_t))^2$\;
            $\phi \leftarrow \phi - \alpha_\phi \nabla_\phi L(\phi)$\;

            \tcp{Update Actor}
            $J(\theta) \leftarrow \frac{1}{T} \sum_{t=0}^{T-1} \rho_t \log \pi_\theta(a_t|s_t) \hat{A}_t$\;
            where $\rho_t = \min\left( \overline{\rho},
            \frac{\pi_\theta(a_t|s_t)}{\behavior(a_t|s_t)} \right)$ \;
            $\theta \leftarrow \theta + \alpha_\theta \nabla_\theta J(\theta)$ 
        }
        Update target network: $\phi' \leftarrow (1-\alpha_{\text{polyak}})\phi' + \alpha_{\text{polyak}}\phi$\;
    }
}
\end{algorithm}

\subsubsection{Example: IMPALA}
\label{sec:IMPALA}

As an example of an off-policy AC method,
we consider
\keywordDef{IMPALA},
which stands for ``Importance Weighted Actor-Learning Architecture''.
\citep{Espeholt2018}.
This uses  shared parameters for the policy and value function
(with different output heads), and adds an entropy bonus
to ensure the policy remains stochastic.
Thus we end up with the following objective,
which is very similar to on-policy actor-critic
shown in \cref{algo:AC}:
\be
  \loss(\vphi) = \expectQ{
    \lambda_{TD} (V_{\vphi}(s_t) - v_t)^2
    -\lambda_{PG} \rho_t A_t \log \pi_{\vphi}(a_t|s_t)
    -\lambda_{ent} \entropy(\pi_{\vphi}(\cdot|s_t))
  }{t \sim \data}
  \label{eqn:offpolAC}
  \ee
The only difference from standard A2C is that we need
to store the probabilities of each action,
$\policybeh(a_t|s_t)$, in addition to $(s_t,a_t,r_t,s_{t+1})$
in the dataset $\data$,
which can be used to compute the importance ratio
$\rho_t$ in  \cref{eqn:rhot}.
\citep{Espeholt2018} was  able to use this method to train 
a single agent (using a shared CNN and LSTM for both value and policy)
to play all 57 games at a high level.
Furthermore, they showed that their
method --- thanks to its off-policy corrections ---
outperformed the A3C method (a parallel version of A2C) in \cref{sec:A3C}.

\eat{
\subsubsection{Controlling the variance of the importance weights}

The
\keywordDef{importance sampling} correction term,
 $\rho_t = \frac{\policy_{\vtheta}(a_t|s_t)}{\behavior(a_t|s_t)}$
is
used to compensate for the fact that the data is sampled
from $\behavior$ and not $\policy$.
We require that  the support of the behavior
policy  is at least
as broad as the target policy,
so $\policy(a|s)>0 \implies \behavior(a|s)>0$.
Also,  to ensure the variance of the estimators is reasonable,
we require that $\behavior$ be close to $\policy$.
In online RL, we can ensure the new policy does not
differ too much from the old policy, using methods
such as TRPO (\cref{sec:TRPO}),
PPO (\cref{sec:PPO}) and MPO (\cref{sec:MPO}).
In  \cref{sec:DPG},
we discuss a way to avoid the need to compute the IS ratio.
But in the general case, we need other tricks
to control the variance of $\rho_t$.

Following \citep{Roux2025},
we define the notation
\be
[x]_a^b = \min(\max(x,a),b)
\ee
so the value is at least $a$ and at most $b$.
The PPO objective can be written as follows
\be
J_{ppo}(\pi) = \E_{\tau \sim \behavior}
\min\left( \frac{\pi(\tau)}{\behavior(\tau)} R(\tau),
\left[ \frac{ \pi(\tau)}{\behavior(\tau)} \right]_{1-\epsilon}^{1+\epsilon}
R(\tau)
\right)
\ee
for $\epsilon \in (0,1)$,
where $R(\tau)$ is the reward for trajectory $\tau$.
}

\subsubsection{Off-policy learning with deterministic policies}

In \cref{sec:DPG}, we discuss actor-critic methods that learn deterministic
policies for continuous actions, based on the deterministic policy gradient (DPG)
method. These methods can work with off-policy without the need to do importance
sampling correction.

\subsubsection{PGQL: Combining off-policy Q-learning with policy gradient}

In \citep{ODonoghue2016}, they present a method known as \keywordDef{PGQL},
which uses Q learning to learn from off-policy data in a replay buffer,
and uses this to regularize the on-policy actor-critic learner.

\subsection{Off-policy policy improvement methods}
\label{sec:offpolPolImprovement}

So far we have focused on  actor-critic methods.
However, policy improvement methods, such as PPO,
are often preferred to AC methods,
since they monotonically improve the objective.
In  \citep{Queeney2021} they propose
one way to extend PPO to the off-policy case.
This method was generalized in  \cite{Queeney2024}
to cover a variety of policy improvement algorithms,
including TRPO and VMPO.
We give a brief summary below.

The key insight is to realize that we can generalize
the lower bound in \cref{eqn:Jlowerbound}
to any reference policy 
\be
J(\pi) - J(\pi_k)
\geq
  \frac{1}{1-\gamma}
 \expectQ{
  \frac{\pi(a|s)}{\policyref(a|s)}
  A^{\pi_k}(s,a)}{\statdistpolref(s) \policy_k(a|s)}
-
  \frac{2 \gamma C^{\pi,\pi_k}}{(1-\gamma)^2}
  \expectQ{\text{TV}(\pi(\cdot|s),\policyref(\cdot|s))}{\statdistpolref(s)}
  \ee
The reference policy can be any previous policy,
or a convex combination of them.
In particular, if $\pi_k$ is the current policy,
we can consider the reference policy to be
$\policyref = \sum_{i=1}^M \nu_i \pi_{k-i}$,
where $0 \leq \nu_i \leq 1$
and $\sum_i \nu_i = 1$ are mixture weights.
We can approximate the expectation
by sampling  from the  replay buffer,
which contains samples from older policies.
That is, $(s,a) \sim \statdistpolref$
can be implemented by $i \sim \nu$
and $(s,a) \sim \statdistpolki$.

To compute the advantage function $A^{\pi_k}$
from off policy data, we can adapt the V-trace
method of \cref{eqn:Vtrace} to get
\begin{align}
  A_{\ttrace}^{\pi_k}(s_t,a_t)
  &= \delta_t
  + \sum_{j=0}^{n-1} \gamma^{j} \left( \prod_{m=1}^{j} c_{t+m}
  \right)   \delta_{t+j}
\end{align}
where
$\delta_t = r_t + \gamma V(s_{t+1}) - V(s_t)$,
and
$c_t = \min\left(\overline{c},
\frac{\pi_k(a_t|s_t)}{\pi_{k-i}(a_t|s_t)} \right)$
is the truncated importance sampling ratio.

To compute the TV penalty term from off policy data,
we need to choose between the PPO (\cref{sec:PPO}),
VMPO (\cref{sec:VMPO})
and TRPO (\cref{sec:TRPO}) approach.

We can derive  an off-policy version of PPO using  an update of the
following form (known as \keywordDef{Generalized PPO}):
\be
\pi_{k+1} = \argmax_{\pi} \expectQ{
  \expectQ{
    \min( \rho_{k-i}(s,a) A^{\pi_k}(s,a),
    \tilde{\rho}_{k-i}(s,a) A^{\pi_k}(s,a) )
  }{(s,a) \sim \statdistpolki}
  }{i \sim \nu}
\ee
where $\rho_{k-i}(s,a)=\frac{\pi(a|s)}{\pi_{k-i}(a|s)}$
and
$\tilde{\rho}_{k-i}(s,a)=\text{clip}(\frac{\pi(a|s)}{\pi_{k-i}(a|s)},l,u)$,
where $l=\frac{\pi_k(a|s)}{\pi_{k-i}(a|s)} - \epsilon$
and $u=\frac{\pi_k(a|s)}{\pi_{k-i}(a|s)} + \epsilon$.
(For other off-policy variants of PPO,
see e.g., \citep{Meng2023,Li2024R3}.)

For details on the off-policy version of TRPO,
see  \cite{Queeney2024}.

For an off-policy version of VMPO,
see the discussion of MPO in \cref{sec:MPO}.

\eat{
see  the original \keywordDef{MPO} method of \citep{Abdolmaleki2018};
this is derived using an EM framework,
but EM is just another bound optimization algorithm \citep{Hunter04},
and the result is equivalent to the version presented
in \cite{Queeney2024}.
}

\section{Gradient-free policy optimization}
\label{sec:PGrnd}

\eat{
The policy gradient
estimator computes a ``zeroth order'' gradient, which essentially
evaluates the function with randomly sampled trajectories.
Sometimes it can be more efficient to
use a derivative-free optimizer
that does not even attempt to estimate the gradient.
For example, \citep{Mania2018} obtain good results
by training linear policies with random search,
and \citep{Salimans2017ES} show how to use
evolutionary strategies to optimize the policy
of a robotic controller.
}

So far, we have focused on fitting parametric policies,
represented by differentiable functions
$\pi_{\vtheta}(a|s)$,
using methods based on the policy gradient theorem.
Unfortunately, such gradient-based methods
can get stuck in poor local optima.
In addition, gradient descent cannot be applied
to non-differentiable policies, such as programs,
or functions with discrete latent variables
(e.g., if-then branches).
We can therefore consider other kinds of
methods
for policy learning, based on
\keywordDef{blackbox optimization},
aka \keywordDef{derivative-free optimization}.
This includes techniques such as \keywordDef{cross-entropy method}
and \keywordDef{evolutionary strategies}.
For details on such algorithms,
see e.g. \citep[Sec 7.7]{book2}.
For some applications of ES to RL,
see e.g. \citep{Mania2018} (who obtain good results
by training linear policies with random search),
\citep{Salimans2017ES} (who use
evolutionary strategies to optimize the policy
of a robotic controller),
and \citep{Sarkar2025} (who
discucss a method called
``Evolution Guided General Optimization via Low-rank Learning''
(EGGROLL), which provides a way to scale to very large models).

\section{RL as inference}
\label{sec:inferRL}
\label{sec:planningAsInference}
\label{sec:RLAI}


In this section, we discuss an approach to
policy optimization that reduces it
to probabilistic inference.
This is called
\keywordDef{control as inference},
or \keywordDef{RL as inference},
and has been discussed in numerous works
(see e.g., \citep{Attias03,Toussaint06,Toussaint09Robot,Ziebart2010,
  Rawlik2012,Botvinick2012,Kappen2012,Hoffmann2017control,Levine2018inf,
  Furuta2021,Zhao2024smc}).
The primary advantage of this approach is that it enables
policy learning using off-policy data,
while avoiding the need to use
(potentially high variance) importance sampling corrections.
(This is 
because the inference approach
takes expectations wrt $d_q(s)$ instead of $d_{\pi}(s)$,
where $q$ is an auxiliary distribution, $\pi$ is the policy
which is being optimized, and $d$ is the state visitation measure.)
A secondary advantage is that it enables us to use
the large toolkit of methods for probabilistic modeling and inference
to solve RL problems.\footnote{
  Note, however, that we do not tackle the problem of epistemic
  uncertainty (exploration).
  Solving this in the context
of RL-as-inference requires additional machinery,
as discussed in \citep{Tarbouriech2023}.
}
The resulting framework  forms the foundation
of the 
the MPO discussed in \cref{sec:MPO},
the SAC method discussed in \cref{sec:SAC},
as well as the SMC planning  method discussed in \cref{sec:SMCRHC},
and some kinds of  LLM test-time inference,
as discussed in \cref{sec:LLMtestTime}.

\begin{figure}
\centering
\includegraphics[height=2in]{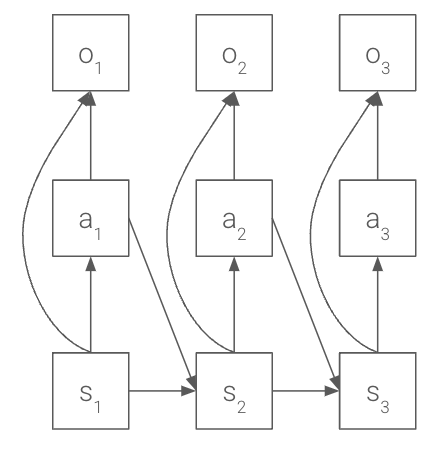}
\caption{
  A graphical model for optimal control.
}
\label{fig:levine-rl-pgm}
\end{figure}

The core of these methods is based on the
probabilistic model shown in 
\cref{fig:levine-rl-pgm}.
This shows an MDP augmented with new variables, $\optimality_t$.
These are called \keywordDef{optimality variables},
and indicating whether the action at time $t$ is optimal or not.
We assume these have the following probability distribution:
\begin{align}
\label{eqn:merl-opt-prob}
p(\optimality_t=1 | s_t,a_t) \propto \exp(\eta^{-1} G(s_t,a_t))
\end{align}
where $\eta > 0$ is a temperature parameter,
and $G(s,a)$ is some quality function,
such as 
$G(s,a)=R(s,a)$,
or $G(s,a)=Q(s,a)$
or $G(s,a)=A(s,a)$.
For brevity, we will just write $p(\optimality=1|\cdot)$
to denote the probability of the event that $\optimality_t=1$ for all time steps.
(Note that the specific value of 1 is arbitrary;
this likelihood function is really just a non-negative weighting term
that biases the action trajectory, as we show below.)

\subsection{Deterministic case (planning/control as inference)}

Our goal is to find trajectories that are optimal.
That is, we would like to find the mode (or posterior samples)
from the following distribution:
\begin{align}
p(\traj | \optimality=1, \pi)
\propto p(\traj, \optimality=1| \pi)
&\propto \left[
p(s_1)  \prod_{t=1}^{T-1} \pi(a_t|s_t) p(s_{t+1}|s_t,a_t)\right]
\left[\prod_{t=1}^{T} p(\optimality_t=1|s_t,a_t)\right]  \notag \\
\label{eqn:rl-merl-trajprob}
\end{align}
where $\pi$ is the policy.

Let us start by considering the deterministic case,
where $p(s_{t+1}|s_t,a_t)$ is either $1$ or $0$,
depending on whether the transition is feasible or not.
In this case, rather than learning a policy $\pi$
that maps states to actions
we just need to learn a plan
(a specific sequence of action $\va_{1:T}$)
for each starting state $s_1$.
This is equivalent to a
\keywordDef{shortest path}
problem, i.e., we want to maximize
\be
p(\traj | \optimality=1, \va_{1:T})
\propto
p(s_1)
\left[ \prod_{t=1}^{T-1} p(s_{t+1}|s_t,a_t) \right]
\left[ \exp(\sum_{t=1}^{T} R(s_t,a_t)) \right] 
\ee
(Typically the initial state $s_1$ is known,
in which case $p(s_1)$ is a delta function.)

The  MAP sequence of actions,
which we denote by $\hat{\va}_{1:T}(s_1)$,
is the optimal \keywordDef{open loop plan}.
(It is called ``open loop'' since the agent does not need to observe the state,
since $s_t$ is uniquely determined by $s_1$ and $\va_{1:t}$, both of which are known.)
Computing this trajectory
 is known as the \keywordDef{control as inference}
 problem \citep{Watson2021}.
 Such open loop planning problems can be solved using
 model predictive control methods, discussed in
\cref{sec:MPC}.

\subsection{Stochastic case (policy learning as variational inference)}

In the stochastic case, we want to learn a policy $\pi$
which maps states to actions, and which generates a
distribution over trajectories which are optimal.
Thus we define the objective as 
\be
\log p(\optimality=1|\pi)
= \log \int \ppi(\traj) p(\optimality=1|\traj) d\traj
\ee
where we define
\be
\ppi(\traj) = p(s_1) \prod_t p(s_{t+1}|s_t,a_t) \pi(a_t|s_t)
\ee
Since marginalizing over trajectories is difficult,
we introduce a variational distribution $q(\traj)$
to simplify the computations.
We assume $q$ factors in the same way:
\be
q(\traj) = p(s_1) \prod_t p(s_{t+1}|s_t,a_t) \pi_q(a_t|s_t)
\ee

\label{sec:optimismBias}
Note that we use the true dynamics model $p(s_{t+1}|s_t,a_t)$
when defining $q$,
and only introduce the variational distribution for the actions,
$\pi_q(a_t|s_t)$. This is one way to avoid the
\keywordDef{optimism bias} that can arise if we sample
from an unconstrained $q(\traj|O=1)$.
To see this, suppose $O=1$ is the event that we win the lottery.
We do not want conditioning on this outcome to influence
our belief in the the probability of chance events,
which is governed by $p(s_{t+1}|s_t,a_t)$ and not
$p(s_{t+1}|s_t,a_t,O=1)$.
See \citep{Levine2018inf} for further discuss of this point.

Now note the following identity
\begin{align}
  \KLpq{q(\traj)}{\ppi(\traj|\opt=1)}
  &= \expectQ{
    \log q(\traj) - \log \frac{\ppi(\opt=1|\traj) \ppi(\traj)}
         {\ppi(\opt=1)}}{q} \\
  &= \expectQ{\log q(\traj) - \log \ppi(\opt=1|\traj) - \log \ppi(\traj)}{q}
  + \log \ppi(\opt=1)
  \label{eqn:ppi}
  \end{align}
Hence
\begin{align}
  \log p_{\pi}(\optimality=1)
  &= \expectQ{\log p(\optimality=1|\traj)
    - \log \frac{q(\traj)}{p(\traj)}
    + \log \frac{q(\traj)}{p(\traj|\optimality=1)}}{q} \\
    &= J(\ppi,q) + \KLpq{q(\traj)}{\ppi(\traj|\optimality=1)} 
  \end{align}
where $J$ is defined by
\begin{align}
  J(\pi_p,\pi_q) &= \expectQ{\log \ppi(\opt=1|\traj)}{q} - \KLpq{q(\traj)}{\ppi(\traj)}\\
  &= \sum_{t=1}^T \expectQ{\eta^{-1} G(s_t,a_t)
    - \KLpq{\pi_q(\cdot|s_t)}{\pi_p(\cdot|s_t)}}{q}
  \label{eqn:JJ}
\end{align}

Since $\KLpq{q(\traj)}{\ppi(\traj|\optimality=1)} \geq 0$,
we see that $\log p(\optimality=1|\pi) \geq J(\ppi,q)$;
hence $J$  is called the \keywordDef{evidence lower bound}
or \keywordDef{ELBO}.
We can define the policy learning task as maximizing
the ELBO, subject to the constraints that $\pi_p$ and $\pi_q$
are distributions that integrate to 1 across actions
for all states.

To extend to the infinite time discounted case,
we define $d_{\pi}(s)$ as the unnormalized discounted
distribution  over states
\be
d_{\pi}(s) = \sum_{t=1}^{\infty} \gamma^t p(s_t=s|\pi)
\ee
We now replace the $\sum_t E_{q(s)}$ with $E_{d_{q}(s)}$
to get the constrained objective
\be
\max_{\pi_p,\pi_q}
J(\pi_p,\pi_q) \myst
\int d_{q}(s) \int \pi_p(a|s) da ds = 1,
\;\;
\int d_{q}(s) \int \pi_q(a|s) da ds = 1
\label{eqn:Jobj}
\ee

There are two main ways to solve this optimization problem,
which we call ``EM control'' and ``KL control'',
following \citep{Furuta2021}.
We describe these below.

\subsection{EM control}

In this section, we discuss
ways to optimize \cref{eqn:Jobj} using the
\keywordDef{Expectation Maximization} or \keywordDef{EM}
algorithm,
which is  a widely used \keywordDef{bound optimization} method,
also called a \keywordDef{MM} (majorize /  maximize) method,
that mononotonically increases a lower bound on its objective
(see \citep{Hunter04} for a tutorial).
In the E step,  we maximize $J$ wrt a 
non-parametric representation of the variational posterior $\pi_q$,
while holding the parametric prior $\pi_p=\pi_{\vtheta_p}^{k-1}$ fixed
at the value from the previous ($k-1$'th) iteration,
to get $\pi_q^k$.
In the M step, we then maximize $J$ wrt $\pi_p$,
holding the variational posterior fixed at $\pi_q^k$,
to get the updated policy $\pi_{\vtheta_p}^k$.

In more detail, in the E step we maximize the following wrt $\pi_q$:
\begin{align}
  J(\pi_{\theta_p}^{k-1},\pi_q) &= \int d_{q}(s) \int \pi_q(s|a) \eta^{-1} G(s,a) da ds
  \nonumber \\
  & - \int d_{q}(s) \int \pi_q(a|s) \log \frac{\pi_q(a|s)}{\pi_{\theta_p^{k-1}}(a|s)}
  da ds + \lambda\left(1- \int d_{\pi(s)} \int \pi_q(a|s) da ds \right)
  \label{eqn:EstepObj}
\end{align}
where $\lambda$ is a Lagrange multiplier.
The optimal (non-parametric) solution to this is
\be
\pi_q^k(a|s) = Z(s)^{-1} \pi_{\theta_p^{k-1}}(a|s)
\exp(\eta^{-1} G(s,a))
  \label{eqn:Estep}
\ee
where $Z$ is the partition function
\be
Z(s) = \int \pi_{\theta_P^{k-1}}(a|s) \exp(\eta^{-1} G(s,a)) da
\ee

In the M step, we maximize the following wrt $\pi_{\theta_p}$:
\be
J(\pi_q^k, \pi_p)
=\expectQ{\log \pi_{\theta_p}(a|s)}{d_{q}(s) \pi_q^k(a|s)}
\label{eqn:Mstep}
\ee
which we recognize as a weighted maximum likelihood problem.

\subsection{KL control (maximum entropy RL)}
\label{sec:maxentRL}

In KL control, we only optimize the variational posterior $\pi_q$,
holding the prior $\pi_p$ fixed. Thus we only have an E step.
In addition,  we represent $\pi_q$ parameterically, as
$\pi_{\theta_q}$, instead of the non-parametric approach used by EM.
If the prior $\pi_p$ is uniform, and we use $G(s_t,a_t)=R(s_t,a_t)$,
then  \cref{eqn:JJ} becomes
\begin{align}
  \eta J(\pi_p,\pi_q)
  &= \sum_{t=1}^T \expectQ{R(s_t,a_t)    - \eta H(\pi_q(\cdot|s_t))}
       {q}
  \label{eqn:maxentRL}
\end{align}
where $-H(q) = \KLpq{q}{\text{unif}} = \sum_a q(a) \log \frac{q(a)}{c}$
is the negative entropy function and
$c$ is a constant.
This is called the \keywordDef{maximum entropy RL} objective
\citep{Ziebart2010,SAC,Haarnoja2018SAC}.
This differs from the standard objective used in RL training
(namely a lower bound on sum of expected rewards)
by virtue of the addition of the entropy regularizer
on the policy.
See \cref{sec:SAC} for further discussion.

\subsection{Maximum a Posteriori Policy Optimization (MPO)}
\label{sec:MPO}

In this section, we discuss the \keywordDef{MPO} method of \citep{Abdolmaleki2018}.
This is an instance of EM control,
where $G(s,a)=Q(s,a)$,
which is estimated using the retrace
algorithm
(see \cref{sec:Vtrace})
or a single-step Bellman update.

It implements the E step using \cref{eqn:Estep},
where we approximate $Z(s)$ with Monte Carlo:
\begin{align}
  q^k(a|s) &= \frac{1}{\hat{Z}(s)}
  \pi_{\theta_p^{k-1}}(a|s) \exp(\eta^{-1} G(s,a)) \\
  Z(s) &\approx \frac{1}{M}
  \sum_{j=1}^M \exp(\eta^{-1} G(s,a_j)),
  \; a_j \sim \pi_{\theta_p^{k-1}}(\cdot|s)
  \label{eqn:EstepMC}
\end{align}
\eat{
a trust region method,
in which the KL penalty
in \cref{eqn:Estep}
is replaced by a constraint of the form
$E_{d_{q}(s)}[\KLpq{\pi_q(a|s)}{\pi_{\theta_p}(a|s)}] \leq \epsilon$.
(It is easier to choose a single value of
 $\epsilon$ that works well for many problems because it
is independent of the scale of the reward term $G(s,a)$.)
}
In addition, the (inverse) temperature parameter $\eta$
is solved for by minimizing the dual of
\cref{eqn:Estep}, which is given by
\be
g(\eta) = \eta \epsilon
+ \eta \log \expectQ{\exp\left( \eta^{-1} Q^{\pi_{\theta_p}^{k-1}}(s,a) \right)
}{d_{q}(s) \pi_{\theta_p^{k-1}}(a|s)}
\label{eqn:geta}
\ee

In the M step,
MPO augments the objective in \cref{eqn:Mstep}
with a log prior at the $k$'th step
of the form $\log p_k(\theta_p)$
to create a MAP estimate.
That is, it optimizes the following wrt $\theta_p$:
\be
J(q^k, \pi_{\theta_p})
=\expectQ{\log \pi_{\theta_p}(a|s)}{d_{q}(s) q(a|s)}
+ \log p_k(\theta_p)
\ee
We can think of this step as projecting the non-parametric policy
$q$
back to the space of parameterizable policies $\Pi_{\theta}$.

We assume the prior is  a Gaussian centered at the previous
parameters,
\be
p_k(\vtheta)=\gauss(\vtheta|\vtheta_{k}, \lambda \vF_k)
= c \exp\left(-\lambda (\vtheta-\vtheta_k)^\trans \vF_k^{-1}
(\vtheta-\vtheta_k) \right)
\ee
where $F_k$ is the Fisher information matrix.
If we view this as a second order approximation to the KL, we can rewrite
the objective as
\be
\max_{\theta_p} E_{d_{q}(s)} \left[
  E_{q(a|s)} \log \pi(a|s,\theta_p)  
  -\lambda \KLpq{\pi(a|s,\vtheta_k)}{\pi(a|s,\theta_p)}
  \right]
\ee

We can approximate the expectation wrt $d_{q}(s)$
by sampling states from a replay buffer,
and the expectaton wrt
$q(a|s)$
by sampling from  the policy.
The KL term can be computed analytically for Gaussian policies.
\eat{
We assume the policy is \keywordDef{reparameterizable},
such as a Gaussian or
a gumble-softmax distribution
\citep{Jang2016,concrete}
for discrete actions.
This allows us to not only sample from it,
but also safely take gradients wrt the policy parameters.
}
We can then optimize this objective using SGD.

Note that we can also 
rewrite this as a constrained optimization problem
\be
\max_{\theta_p} E_{d_q(s)} \left[
  E_{q(a|s)} \log \pi(a|s,\theta_p)  \right]
\myst E_{d_q(s)} \left[ \KLpq{\pi(a|s,\theta_k)}{\pi(a|s,\theta_p)}
  \right] \leq \epsilon_m
\ee
This can be optimized using a trust region method.

\subsection{Sequential Monte Carlo Policy Optimisation (SMC-PO)}
\label{sec:SMC-PO}

In this section, we discuss 
\keywordDef{SMC-PO} method of
\citep{Macfarlane2024}.
This is a model-based version of MPO,
which uses
Sequential Monte Carlo (SMC) to perform
approximate inference in the E step.
In particular, it samples from a distribution over
optimal future trajectories starting from the current state,
$s_t$,
and using the current policy $\pi_{\theta_p}$ and dynamics model
$\calT(s'|s,a)$.
From this it derives a non-parametric
distribution over the optimal actions to take
at the next step, $q(a_t|s_t)$.
 (see \cref{sec:SMCMPC} for details).
This becomes a target for the parametric policy
update in the M step, which is the same
weighted maximum likelihood method used by MPO.

\subsection{AWR and AWAC}
\label{sec:AWR}

The \keywordDef{Advantage Weighted Regression}
or \keywordDef{AWR} method of
\citep{Peng2019awr}
and
the \keywordDef{Advantage Weighted Actor Critic}
or \keywordDef{AWAC} method of
\citep{Nair2020}
are both EM control methods.
AWR uses $G(s,a)=A(s,a)$,
where the advantage function is
estimated using GAE.
The value function $V(s)$
is estimated using  TD($\lambda$),
and is the value for
the average of all previous policies,
$\tilde{\pi}_{p^k} = \frac{1}{k} \sum_{j=0}^{j-1} \pi_{\theta_p^j}$.
In contrast, AWAC uses $G(s,a)=Q(s,a)$,
which is estimated by TD(0).

The (non-parametric) E step is closed form,
as in other EM control methods,
where the temperate $\eta$ is treated as a hyper-paraemter.
The (parametric) M step is a weighted maximum likelihood
step that is solved with SGD.

\subsection{Soft Actor Critic (SAC)}
\label{sec:SAC}

The \keywordDef{soft actor-critic}  (\keywordDef{SAC})
algorithm \citep{SAC,Haarnoja2018SAC}
is an off-policy
actor-critic method based on
the \keyword{maximum entropy RL}
method we discussed in 
\cref{sec:maxentRL}.
This is an instance of the KL control scheme where
the variational posterior policy $\pi_q=\pi_{\theta_q}$
is parameterized, but the prior policy $\pi_p$
is fixed to the uniform distribution.
(Thus SAC only has an E step (implemented with SGD), but no M step.)
SAC uses $G(s,a)=Q^{\text{soft}}(s,a)$,
where the soft-Q function is defined below.

Crucially, even though
SAC is off-policy and utilizes a replay buffer to sample past experiences,
the policy update is done using the actor's own probability distribution,
eliminating the need to use importance sampling
to correct for discrepancies between the behavior policy
(used to collect data) and the target policy (used for updating),
as we will see below.

\subsubsection{SAC objective}

We can write the maxent RL objective for the
E step by using
\cref{eqn:maxentRL} with slightly modified notation:
\begin{align}
\polval^{\text{SAC}}(\vtheta) \defeq \expectQ{R(s,a)
+ \alpha \entropy(\policy_{\vtheta}(\cdot|s))}{\statdistpolapprox(s)\polapprox(a|s)}
\end{align}
Note that the entropy term makes the objective easier to optimize,
and encourages  exploration.
To optimize this, we can perform a policy evaluation step,
and then a  policy improvement step.

\subsubsection{Policy evaluation: tabular case}

We can perform  policy evaluation
by repeatedly applying a modified Bellman
backup operator $\calT^{\pi}$ defined as
\begin{align}
  \calT^{\pi} Q(\vs_t,\va_t) = r(\vs_t,\va_t) +
  \gamma \expectQ{V(\vs_{t+1})}{\vs_{t+1} \sim p}
  \end{align}
where 
\begin{align}
  V(\vs_t) = \expectQ{
    Q(\vs_t,\va_t) - \alpha \log \pi(\va_t|\vs_t)}{\va_t \sim \pi}
  \label{eqn:softValue}
  \end{align}
is the \keywordDef{soft value function}.
If we iterate $Q^{k+1} = \calT^{\pi} Q^k$,
this will converge to the soft $Q$ function for $\pi$.

In the tabular case, we can derive the optimal soft value function as follows.
First, by definition, we have
\begin{align}
V^*(s) := \max_\pi \sum_a \pi(a \mid s) \left[ Q^*(s, a) - \alpha \log \pi(a \mid s) \right].
\label{eqn:softValueFnDef}
\end{align}
This is a constrained optimization problem, where \( \pi(\cdot \mid s) \) is a probability distribution. We introduce a Lagrange multiplier \( \lambda \)
to enforce the normalization constraint:
\be
\mathcal{L}(\pi, \lambda) = \sum_a \pi(a \mid s) \left[ Q^*(s, a) - \alpha \log \pi(a \mid s) \right] + \lambda \left( 1 - \sum_a \pi(a \mid s) \right).
\ee
Taking the derivative of \( \mathcal{L} \) with respect to \( \pi(a \mid s) \) and setting it to zero:
\be
\frac{\partial \mathcal{L}}{\partial \pi(a \mid s)} = Q^*(s, a) - \alpha (1 + \log \pi(a \mid s)) - \lambda = 0.
\ee
Solving for \( \pi(a \mid s) \):
\be
\log \pi(a \mid s) = \frac{Q^*(s, a) - \lambda - \alpha}{\alpha}
\quad \Rightarrow \quad
\pi(a \mid s) \propto \exp\left( \frac{Q^*(s, a)}{\alpha} \right).
\ee
The optimal policy is therefore the softmax over Q-values:
\be
\pi^*(a \mid s) = \frac{\exp\left( \frac{Q^*(s, a)}{\alpha} \right)}{\sum_{a'} \exp\left( \frac{Q^*(s, a')}{\alpha} \right)}.
\ee
Plugging this back into the soft value function:
\be
V^*(s) = \sum_a \pi^*(a \mid s) \left[ Q^*(s, a) - \alpha \log \pi^*(a \mid s) \right].
\ee
Since
\be
\log \pi^*(a \mid s) = \frac{Q^*(s, a)}{\alpha} - \log \sum_{a'} \exp\left( \frac{Q^*(s, a')}{\alpha} \right),
\ee
we have
\be
Q^*(s, a) - \alpha \log \pi^*(a \mid s) = \alpha \log \sum_{a'} \exp\left( \frac{Q^*(s, a')}{\alpha} \right).
\ee
Therefore, the optimal soft value function is given by
\be
V^*(s) = \sum_a \pi^*(a \mid s) \cdot \alpha \log \sum_{a'} \exp\left( \frac{Q^*(s, a')}{\alpha} \right) = \alpha \log \sum_{a} \exp\left( \frac{Q^*(s,a)}{\alpha} \right).
\label{eqn:softValueFn}
\ee

\subsubsection{Policy evaluation: general case}

We now generalize this to the non-tabular case.
We  hold the policy parameters $\pi$ fixed
and optimize the parameters  $\vw$ of the $Q$ function
by minimizing
\begin{align}
 J_Q(\vw) = \expectQ{
   \half \left(
   Q_{\vw}(\vs_t,\va_t) - \TargetV(r_{t+1}, \vs_{t+1})
   \right)^2
   }{(\vs_t,\va_t,r_{t+1},\vs_{t+1}) \sim \calD}
 \label{eqn:JQ}
\end{align}
where $\calD$ is a replay buffer,
\be
\TargetV(r_{t+1},\vs_{t+1}) = r_{t+1} +  \gamma V_{\overline{\vw}}(\vs_{t+1})
\ee
is the frozen target value, and
and $V_{\overline{\vw}}(\vs)$ is a frozen version of the
soft value function
from \cref{eqn:softValue}:
\begin{align}
  V_{\overline{\vw}}(\vs_t) = \expectQ{
    Q_{\overline{\vw}}(\vs_t,\va_t) - \alpha \log \pi(\va_t|\vs_t)}{\pi(\va_t|\vs_t)}
  \label{eqn:softValueFrozen}
  \end{align}
where $\overline{\vw}$ is the 
EMA version of $\vw$.
(The use of 
a frozen target is to avoid bootstrapping instabililities
discussed in \cref{sec:deadlyTriad}.)

To avoid the positive overestimation bias
that can occur with actor-critic methods,
\citep{SAC},
suggest fitting two soft $Q$ functions,
by optimizing $J_Q(\vw_i)$, for $i=1,2$, independently.
Inspired by clipped double $Q$ learning,
used in TD3 (\cref{sec:TD3}),
the targets are defined as 
\be
\TargetV(r_{t+1},\vs_{t+1}; \overline{\vw}_{1:2}, \vtheta)
= r_{t+1} + \gamma \left[ \min_{i=1,2}
Q_{\overline{\vw}_i}(\vs_{t+1}, \tilde{\va}_{t+1})
-\alpha \log \pi_{\vtheta}(\tilde{\va}_{t+1} | \vs_{t+1}) \right]
\ee
where $\tilde{\va}_{t+1} \sim \pi_{\vtheta}(\vs_{t+1})$
is a sampled next action.
In \citep{REDQ}, they propose the \keyword{REDQ} method
(\cref{sec:REDQ})
which uses a random ensemble of $N \geq 2$
networks instead of just 2.

\subsubsection{Policy improvement}

In the policy improvement step,
we derive the new policy based on the soft $Q$ function
by softmaxing over the possible actions for each state.
We then project the update back on to the
policy class $\Pi$:
\begin{align}
  \pi_{\new} =  \arg \min_{\pi' \in \Pi}
  \KLpq{\pi'(\cdot|\vs_t)}
       {
         \frac{\exp(\frac{1}{\alpha} Q^{\pi_{\old}}(\vs_t,\cdot))}
              {Z^{\pi_{\old}}(\vs_t)}
       }
  \end{align}
(The partition function $Z^{\pi_{\old}}(\vs_t)$ may be intractable
to compute for a continuous action space, but it cancels out
when we take the derivative of the objective, so this is not a problem,
as we show below.)
After solving the above optimization problem, we are guaranteed
to satisfy the soft policy improvement theorem,
i.e., $Q^{\pi_{\new}}(\vs_t,\va_t) \geq Q^{\pi_{\old}}(\vs_t,\va_t)$
for all $\vs_t$ and $\va_t$.

We now generalize this to the non-tabular case.
For policy improvement, we hold the value function parameters $\vw$ fixed
and optimize 
the parameters $\vtheta$ of the policy
by minimizing the objective below,
which is derived from the KL term
by multiplying by $\alpha$ and dropping the constant $Z$ term:
\begin{align}
  J_{\pi}(\vtheta)
    = \expectQ{
      \expectQ{\alpha \log \pi_{\vtheta}(\va_t|\vs_t)
        -Q_{\vw}(\vs_t,\va_t)}{\va_t \sim \pi_{\vtheta}}
    }{\vs_t \sim \calD}
    \label{eqn:Jpi}
    \end{align}
Since we are taking gradients wrt $\vtheta$,
which affects the inner expectation term,
we need to either use the REINFORCE estimator
from \cref{eqn:PGthmBaseline}
or the \keywordDef{reparameterization trick}
(see e.g., \citep{Mohamed2020}).
The latter is much lower variance, so is preferable.

To explain this in more detail,
let us assume the  policy distribution has the form
$\pi_{\vtheta}(\va_t|\vs_t)=\gauss(\vmu_{\vtheta}(\vs_t), \sigma^2 \vI)$.
We can write the random action as $\va_t = f_{\vtheta}(\vs_t,\vepsilon_t)$,
where $f$ is a deterministic function
of the state and a noise variable $\vepsilon_t$,
since $\va_t = \vmu(\vs_t)  + \sigma^2 \vepsilon_t$,
where $\vepsilon_t \sim \gauss(\vzero,\vI)$.
The objective now becomes
\begin{align}
  J_{\pi}(\vtheta)
    = \expectQ{
      \alpha \log \pi_{\vtheta}(f_{\vtheta}(\vs_t,\vepsilon_t) | \vs_t)
        -Q_{\vw}(\vs_t,f_{\vtheta}(\vs_t,\vepsilon_t))
      }{\vs_t \sim \calD, \vepsilon_t \sim \gauss}
    \end{align}
where we have replaced the expectation of $\va_t$
wrt $\pi_\vtheta$ with an expectation of $\vepsilon_t$
wrt its noise distribution $\gauss$.
Hence we can now safely take stochastic gradients.
See \cref{algo:SAC} for the pseudocode.

\begin{algorithm}
\dontprintsemicolon
\caption{SAC}
\label{algo:SAC}
Initialize environment state $\vs$,
policy parameters $\vtheta$,
$N$ critic parameters $\vw_i$,
target parameters $\overline{\vw}_i = \vw_i$,
replay buffer $\data=\emptyset$,
discount factor $\gamma$,
EMA rate $\rho$,
step size $\eta_w$, $\eta_\pi$.
\\
\Repeat{converged}
       {
         Take action $\va \sim \pi_{\vtheta}(\cdot|\vs)$ \\
         $(\vs',r) = \text{step}(\va, \vs)$ \\
         $\data := \data \union
         \{ (\vs, \va, r, \vs') \}$ \\
         $\vs \assign \vs'$ \\
         \For{$G$ updates}{
         Sample a minibatch $\calB = \{(\vs_j,\va_j,r_j,\vs'_j)\}$
         from $\data$ \\
           $\vw = \text{update-critics}(\vtheta, \vw, \calB)$
         }
         Sample a minibatch $\calB = \{(\vs_j,\va_j,r_j,\vs'_j)\}$
         from $\data$\\
         $\vtheta = \text{update-policy}(\vtheta,\vw, \calB)$
        }
.\\
$\text{def update-critics}(\vtheta,\vw,\calB)$: \\
       Let $(\vs_j,\va_j,r_j,\vs'_j)_{j=1}^B = \calB$ \\
       $\targetV_j = \TargetV(r_j,\vs'_j; \overline{\vw}_{1:N}, \vtheta)$ for  $j=1:B$ \\
\For{$i=1:N$}
    {
      $\loss(\vw_i) = \frac{1}{|\calB|} \sum_{(\vs,\va,r,\vs')_j \in
        \calB} (Q_{\vw_i}(\vs_j,\va_j) - \stopgrad(\targetV_j))^2$\\
      $\vw_i \assign \vw_i - \lr_{\vw} \nabla \loss(\vw_i)$ // Descent
      \\
      $\overline{\vw}_i := \rho \overline{\vw}_i
      + (1-\rho) \vw_i$       //Update target networks \\
    }
    Return $\vw_{1:N}, \overline{\vw}_{1:N}$\\
.\\
$\text{def update-actor}(\vtheta,\vw,\calB)$: \\
    $\hat{Q}(s,a)  \defeq \frac{1}{N} \sum_{i=1}^N
    Q_{\vw_i}(s,a)$ // Average critic\\
$J(\vtheta) =
\frac{1}{|\calB|} \sum_{\vs \in \calB}
\left(
\hat{Q}(\vs, \tilde{\va}_{\vtheta}(\vs))
-
\alpha \log \pi_{\vtheta}(\tilde{\va}(\vs)|\vs)
\right),
\;  \tilde{\va}_{\vtheta}(\vs) \sim \pi_{\vtheta}(\cdot|\vs)
$ \\
$\vtheta \assign \vtheta +  \lr_{\vtheta} \nabla J(\vtheta)$ // Ascent\\
Return $\vtheta$
\end{algorithm}

\label{sec:SACdiscrete}
For discrete actions,  we can replace the Gaussian
reparameterization with the gumbel-softmax reparameterization
\citep{Jang2016,concrete}.
Alternatively, we can eschew sampling
and compute the expectations over the actions explicitly,
to derive lower variance versions
of the equations;
this is known as \keywordDef{SAC-Discrete}
\citep{Christodoulou2019}.

\subsubsection{Adjusting the temperature}

In  \citep{Haarnoja2018SAC} they propose to  automatically
adjust the temperature parameter $\alpha$ by optimizing
\begin{align*}
  J(\alpha) = \expectQ{-\alpha(\log \pi_\vtheta(\va_t|\vs_t) + \overline{H})}
  {\vs_t \sim \data, \va_t \sim \pi_{\vtheta}}
  \end{align*}
where $\overline{H}$ is the target entropy (a hyper-parameter).
This objective is approximated by sampling actions from the replay buffer.

\subsection{Active inference}


Control as inference is closely related
to a technique known as \keywordDef{active inference},
as we explain below. For more details on the connection, see
\citep{Millidge2020,Watson2020,VanDeLaar2021,Sajid2021,Tschantz2020}.

The active inference technique was developed in the neuroscience
community, that has its own vocabulary for standard ML concepts.
We start with 
the \keywordDef{free energy principle} 
 \citep{Friston2009,Buckley2017,Schwobel2018,
  Gershman2019,Mazzaglia2022}.
The FEP is equivalent to using variational  inference
to perform state estimation (perception)
and parameter estimation (learning)
in a latent variable model.
In particular, consider an LVM $p(\vz,\vo|\vtheta)$
with hidden states $\vz$, observations $\vo$, and parameters
$\vtheta$. 
We define the \keyword{variational free energy} to
be
\be
\calF(\vo|\vtheta) = \min_{q \in \cal{Q}} \calF(\vo|\vq,\vtheta)
\ee
where
\be
\calF(\vo|\vq,\vtheta)
= \KLpq{q(\vz|\vo,\vtheta)}{p(\vz|\vo,\vtheta)} - \log p(\vo|\vtheta)
= \expectQ{\log q(\vz|\vo,\vtheta) - \log p(\vo,\vz|\vtheta)}
{q(\vz|\vo,\vtheta)}
\geq -\log p(\vo|\vtheta)
\ee
which is the KL between the approximate variational posterior $q$
and the true posterior $p$, minus a normalization constant,
$\log p(\vo|\vtheta)$, which is known as the free energy.
State estimation (perception) corresponds to solving
$\min_{q(\vz|\vo,\vtheta)} \calF(\vo|\vq,\vtheta)$,
and parameter estimation (model fitting)
corresponds to solving $\min_{\vtheta} \calF(\vo|\vq,\vtheta)$,
just as in the  EM (expectation maximization) algorithm.
(We can also be Bayesian about $\vtheta$,
as in variational Bayes EM,
instead of just
computing a point estimate.)
This EM procedure will minimize the VFE,
which is an upper bound on the negative log marginal likelihood of the
data. In other words, it adjusts the model (belief state
and parameters) so that it better
predicts the observations, so the agent is less surprised
(minimizes prediction errors).
cf.  \citep{Berseth2021}.

To extend the above FEP to decision making problems,
we define the \keywordDef{expected free energy}
as follows
\begin{align}
\calG(\va) 
&= \expectQ{\calF(\vo)}{q(\vo|\va)}\\
&=
\underbrace{\expectQ{\KLpq{q(\vz|\vo)}{p(\vz|\vo)}}{q(\vo|\va)}}_
{\calG_{\text{epistemic}}(\va)}
- \underbrace{\expectQ{\log p(\vo|\vtheta)}{q(\vo|\va)}}_
{\calG_{\text{extrinsic}}(\va)}
\end{align}
where $q(\vo|\va)$ is the posterior predictive distribution
over future observations given action sequence $\va$.
(We should also condition on any observed  history / agent state $\vh$,
and the  model parameters $\vtheta$,
but we omit this from the notation for brevity.)

We see that we can  decompose the EFE
into two terms.
First there is the \keywordDef{intrinsic value},
known as the \keywordDef{epistemic drive}.
Minimizing this will encourage the agent to choose actions
which maximize the mutual information between
the observations $\vo$ and the hidden states $\vz$,
thus reducing uncertainty about the hidden states.
(This is called \keywordDef{epistemic foraging}.)
Second there is the  \keywordDef{extrinsic value},
known as the \keywordDef{exploitation term}.
Maximizing this will encourage the agent to choose actions
that result in observations that match its prior.
For example, if the agent predicts that the world will look brighter
when it flips a light switch, it can take the action of flipping
the switch to fulfill this prediction.
This prior can be related to a reward function by defining
as $p(\vo) \propto e^{R(\vo)}$, encouraging goal directed behavior,
exactly as in control-as-inference
(cf. \citep{de-Vries2025}).
However, the active inference approach provides
a way of choosing actions without needing to specify a reward.

Since solving to the optimal action at each step can be slow,
it is possible to amortize this cost by training a policy network to compute
$\pi(\va|\vh) = \argmin_{\va} \calG(\va|\vh)$,
where $\vh$ is the observation history (or current state),
as shown in \citep{Millidge2020vpg,van-der-Himst2020};
this is called \keywordDef{``deep active inference''}.

Overall, we see that this framework
provides a unified theory of both perception and action,
both of which try to minimize some form of free energy.
In particular, minimizing the expected free energy will cause
the agent to pick actions to reduce its uncertainty about its hidden states,
which can then be used to improve its predictive model $p_\vtheta$
of observations; this in turn will help minimize the VFE of future observations,
by updating the internal belief state $q(\vz|\vo,\vtheta)$
to explain the observations.
In other words,
the agent acts so it can learn so it becomes less surprised
by what it sees.
This ensures the agent is in \keywordDef{homeostasis} with its environment.

\eat{
To guide the agent towards preferred outcomes (beyond just information seeking),
we define the prior over states as $p(\vz) \propto e^{R(\vz)}$,
where $R$ is the reward function.
Alternatively, we can define the prior over observations
as $p(\vo) \propto e^{R(\vo)}$.
Either way, the prior
is defined in terms of what the agent
wants to achieve, rather than being an ``objective'' model of reality
(c.f., optimism bias in control-as-inference mentioned above).
\eat{
The advantage of this approach is that it automatically
induces  {\em goal-directed} information-seeking
behavior, rather than than the maxent approach which models
uncertainty in a goal-independent way.
Hence the term ``active inference''.
}
}

Note that active inference is often discussed in the context of
\keywordDef{predictive coding}.
This is equivalent to a special case of FEP
where two assumptions are made:
(1)  the generative model $p(\vz,\vo|\vtheta)$ is 
a  nonlinear hierarchical Gaussian  model
(similar to a VAE decoder),
and (2) the variational posterior approximation
uses a diagonal Laplace approximation,
$q(\vz|\vo,\vtheta) =\gauss(\vz|\hat{\vz},\vH)$
with the mode $\hat{\vz}$
being computed using gradient descent, and $\vH$ being the Hessian
at the mode.
This can be considered a non-amortized version of a VAE,
where inference (E step) is done with iterated gradient descent,
and parameter estimation (M step) is also done with gradient descent.
(A more efficient incremental EM version of predictive
coding, which updates $\{ \hat{\vz}_n: n=1:N\}$ and $\vtheta$ in parallel,
was recently presented in \citep{Salvatori2024iclr},
and an amortized version in \citep{Tscshantz2023}.)
For more details on predictive coding,
see
\citep{Rao1999predictive,Friston2003,Spratling2017,Buckley2017,Hosseini2020pc,
  Millidge2021pc,Marino2021,Ororbia2022,Salvatori2023,Salvatori2024iclr}.

\chapter{Model-based RL}
\label{sec:MBRL}




\section{Introduction}

Model-free approaches to RL typically
need a lot of interactions with the environment
to achieve good performance.
For example, state of the art
methods for the Atari benchmark, such as \keyword{rainbow}
(\cref{sec:DQN}), use millions of frames,
equivalent to many days of playing
at the standard frame rate.
By contrast, humans can achieve the same
performance in minutes \citep{Tsividis2017}.
Similarly,
OpenAI's robot hand controller \citep{Andrychowicz2020}
needs  100 years of simulated data
to 
learn to manipulate a rubiks cube.

One promising approach to greater sample efficiency
is \keywordDef{model-based RL}
(\keywordDef{MBRL}).
In the simplest approach to MBRL,
we first learn the state transition or dynamics model
$\ptran(s'|s,a)$ --- also called a \keywordDef{world model}
(see e.g., \citep{Zhu2024wm,Ding2025wm,Shang2025})
---
and the reward function $R(s,a)$,
using some offline trajectory data, 
and then we use these models to compute a  policy
(e.g., using dynamic programming, as discussed in
\cref{sec:rl-planning},
or using some model-free policy learning method on simulated data,
as discussed in \cref{sec:policySearch}).
It can be shown that the sample complexity
of learning the dynamics is less than the sample
complexity of learning the policy \citep{Zhu2024iclr}.\footnote{
It can also be shown
that learning an agent that can learn a policy
to  achieve any multi-step goal-conditioned task
must be {\em implicitly} learning a world model
\citep{Richens2025}.
This is potentially easier in practice, since it may be more robust
to modeling error.
}

The above two-stage approach --- where we first learn the model,
and then plan with it --- can suffer from the usual problems
encountered in offline RL (\cref{sec:offlineRL}),
i.e., the policy may query the model at a state
for which no data has been collected, so predictions can be
unreliable, causing the policy to learn the wrong thing.
To get better results, we have to interleave the model learning
and policy learning, so that one helps the other
(since the policy determines what data is collected).

There are two main ways to perform MBRL.
In the first approach,
known as  \keywordDef{decision-time planning}
or \keywordDef{model predictive control},
we use the model
to choose the next action by searching over possible future
trajectories.
We then score each trajectory, pick the action corresponding to the best one, take
a step in the environment, and repeat.
(We can also optionally update the model
based on the  rollouts.)
This is discussed in \cref{sec:rollout}.

The second approach is to use the current model and policy
to rollout imaginary trajectories,
and to use this data
(optionally in addition to empirical data)
to  improve the policy using model-free RL;
this is called \keywordDef{background planning},
and is discussed in \cref{sec:backgroundPlanning}.

The advantage of decision-time planning is that
it  allows us to train a world model on reward-free data,
and then use that model
to optimize any reward function.
This can be particularly useful if the reward
contains changing constraints,
or if it is  an intrinsic reward (\cref{sec:intrinsic})
that frequently changes based on the knowledge state of the agent.
The downside of decision-time planning is that it is much slower.
However, it is possible to combine the two methods, as we discuss below.
For an empirical comparison of
background planning and decision-time planning,
see  \citep{Alver2024background}.

Some generic pseudo-code for an MBRL agent is given
in \cref{algo:MBRL}.
(The {\tt rollout} function is defined in \cref{algo:rollout};
some simple  code for model learning is
shown in \cref{algo:update-model},
although we discuss other loss functions in  \cref{sec:WM};
finally, the code for the policy learning is given in other parts of this manuscript.)
For more details on general MBRL, see
e.g., \citep{Wang2019MBRL,Moerland2023,Plaat2021,Luo2022}.
See also \url{https://github.com/alexzhang13/world-models-papers}.

\begin{algorithm}
\dontprintsemicolon
\caption{MBRL agent}
\label{algo:MBRL}
def MBRL-agent$(\Mtrue; T, H, N)$: \\
Initialize state $s \sim \Mtrue$ \\
Initialize data buffer $\data=\emptyset$, model $\hat{M}$ \\
Initialize value function $V$, policy proposal $\pi$\\
\Repeat{until converged}
       {
       // Collect data from environment \\
         $\tau_{\tenv} = \text{rollout}(s, \pi, T, \Mtrue)$, \;
      $s = \tau_{\tenv}[-1]$, \;
      $\data = \data \union \tau_{\tenv}$ \\
      // Update model \\
         \If{Update model online}
            {
        $\hat{M} = \text{update-model}(\hat{M}, \tau_{\tenv})$\\
              }
              \If{Update model using replay}
              {
                $\tau_{\treplay}^{n} =\text{sample-trajectory}(\data), n=1:N$ \\
                $\hat{M} = \text{update-model}(\hat{M}, \tau_{\treplay}^{1:N})$\\
              }

   // Update policy \\
          \If{Update on-policy with real}
                 {
                   $(\pi,V)=\text{update-on-policy}(\pi,V,\tau_{\tenv})$\\
                 }
           \If{Update on-policy with imagination}
                    {
                      $\tau_{\timag}^{n}=\text{rollout}(\text{sample-init-state}(\data),\pi,T,\hat{M}), n=1:N$ \\
           $(\pi,V)=\text{update-on-policy}(\pi,V,\tau_{\timag}^{1:N})$ 
           }

          \If{Update off-policy with real}
           {
             $\tau_{\treplay}^{n} =\text{sample-trajectory}(\data), n=1:N$ \\
             $(\pi,V)=\text{update-off-policy}(\pi,V,\tau_{\treplay}^{1:N})$\\
           }
           \If{Update off-policy with imagination}
           {
             $\tau_{\timag}^{n}=\text{rollout}(\text{sample-state}(\data),\pi,T,\hat{M}), n=1:N$ \\
              $(\pi,V)=\text{update-off-policy}(\pi,V,\tau_{\timag}^{1:N})$\\
            }

}
\end{algorithm}

\begin{algorithm}
\dontprintsemicolon
\caption{Rollout}
\label{algo:rollout}
def rollout$(s_1, \pi, T, M)$ \\
$\tau=[s_1]$ \\
\For{$t=1:T$}{
  $a_t = \pi(s_t)$ \\
  $(s_{t+1}, r_{t+1}) \sim M(s_t,a_t)$ \\
  $\tau += [a_t, r_{t+1}, s_{t+1}]$
  }
Return $\tau$ 
\end{algorithm}

\begin{algorithm}
\dontprintsemicolon
\caption{Model learning}
\label{algo:update-model}
 def update-model$(M,\tau^{1:N}):$\\
 $\lossfn(M) = -\frac{1}{N T}\sum_{n=1}^N \sum_{(s_t,a_t,r_{t+1},s_{t+1}) \in \tau^n}
 \log M(s_{t+1},r_{t+1}| s_t,a_t)$ // NLL \\
    $M = M - \lr_M \nabla_M \lossfn(M)$ \\
    Return $M$\\
\end{algorithm}

\section{Decision-time (online) planning}
\label{sec:rollout}
\label{sec:decisionTimePlanning}
\label{sec:DTP}

In this section, we discuss how to choose the best action
at each step based on planning forward  from the current state
using a known (or learned) world model.
This is called \keywordDef{decision time planning}
or  ``\keywordDef{planning in the now}''
\citep{Kaelbling2011},
and  is in contrast to methods that try to learn
a policy which can be applied to all possible situations.
In this section, we summarize some approaches to this problem.
Our presentation is based in part on 
\citep[Ch. 9]{KochenderferDecisions}.
  
\subsection{Receeding horizon control}
\label{sec:RHC}

In \keywordDef{receeding horizon control} or \keywordDef{RHC},
we plan from the current state $s_t$ to a maximum fixed
depth (horizon into the future) of $d$.
We then take the first  action $a_t$ based on this future
planning, observe the new state $s_{t+1}$, and then replan.
This approach can be quite slow, since it needs to perform
a search or optimization procedure at each step.
However, it can give good results, since it can choose
an action that is tailored to the current state
(and likely future), rather than relying on the generalization
properties of a policy that was learned offline.
In the sections below, we discuss various ways to implement this procedure.

\subsubsection{Forward search}

\begin{figure}
\centering
\includegraphics[height=1.5in]{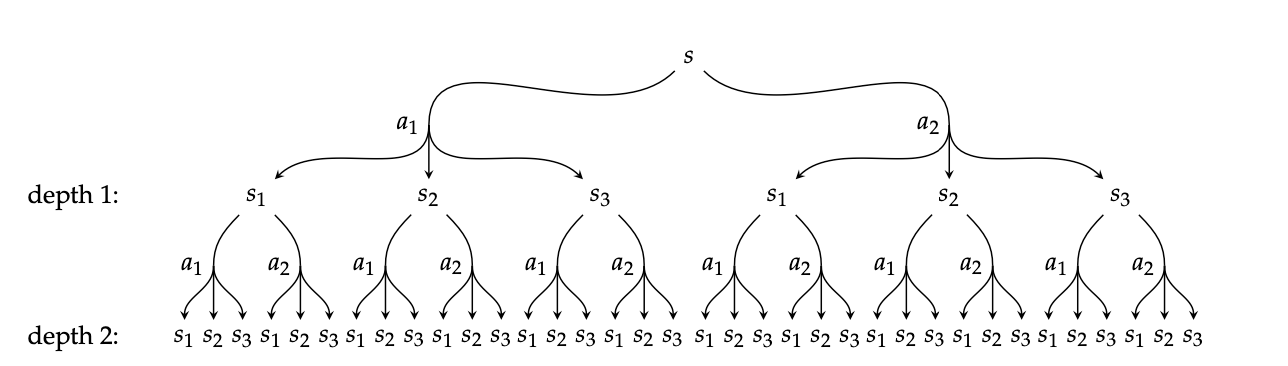}
\caption{
  Illustration of  forward search applied to a problem
  with 3 discrete states and 2 discrete actions.
  \figtaken{Figure 9.1 of \citep{KochenderferDecisions}}.
  \figthanks{Mykel Kochenderfer}.
}
\label{fig:searchTree}
\end{figure}

In \keywordDef{forward search}, we examine all possible
transitions up to depth $d$ by starting from the current state,
and then considering all possible actions,
and then considering all possible next states,
etc. An example of the resulting \keywordDef{search tree}
is given in \cref{fig:searchTree}.
We can compute the reward associated with each edge in the tree.
At the leaves of the tree, we compute the remaining
reward-to-go based on a utility or value function, $V(s)$,
which can be learned offline using value-based methods.
We then find the path with the highest score,
and return the first action on this path.
This process  takes
$O((|\calS| \times |\calA|)^d)$ time.

\subsubsection{Branch and bound}

In \keywordDef{branch and bound}, we try to avoid the exponential complexity
of forward search by pruning paths that we determine are suboptimal.
To do this, we need to know a lower bound on the value function,
$\underline{V}(s)$, and an upper bound on the action value function,
$\overline{Q}(s,a)$.
At each state node $s$, we examine the actions in decreasing
order of their upper bound.
If we find an action $a$ where  $\overline{Q}(s,a)$ is less
than the current best lower bound, we prune this branch of the tree,
otherwise we expand it, and explore below.
We continue this process until we hit a leaf node $s$ (at the maximum depth),
in which case we return the lower bound $\underline{V}(s)$.
Depending on the tightness of the bounds, this approach can be significantly faster
than forward search.

\subsubsection{Sparse sampling}

A simple way to speed up forward search (and branch and boundd)
is to sample a subset of $m$ possible next states for each action.
This is called \keywordDef{sparse sampling} \citep{Kearns99b}.
The resulting complexity is $O((m \times |\calA|)^d)$,
which is independent of $|\calS|$.

\subsubsection{Heuristic search}
\label{sec:heuristic}

In \keywordDef{heuristic search},
we start with a heuristic function $\overline{V}(s)$,
which 
we use to initialize the value function $V(s)$.
We then perform $m$ Monte Carlo rollouts starting
from the root node $s$.
At each state node, we pick the greedy
action wrt the current $V$,
i.e., we choose $\argmax_a R(s,a) + \gamma \sum_{s'} p(s'|s,a) V(s')$.
We then update $V(s) = \max_a R(s,a) + \gamma \sum_{s'} p(s'|s,a) V(s')$,
and sample a next state $s' \sim p(s|s,a)$.
We repeat this process until we hit the max depth.
Finally we return the greedy action wrt $V$ applied to the root node.

If the heuristic function
is an upper bound on the optimal value function,
then it is called an \keywordDef{admissible heuristic}.
In this case, heuristic search is guaranteed to converge to the optimal
value.
The efficiency depends on the tighteness of the upper bound,
but in the worst case it is $O(m \times d \times |\calS| \times |\calA|)$.

\subsection{Monte Carlo tree search (MCTS)}
\label{sec:MCTS}

\keywordDef{Monte Carlo tree search}
or \keywordDef{MCTS} is a receeding horizon control  procedure
that works as follows
(see e.g., \citep{Munos2014} for more details).
Given the root node $s_t$,
we perform $m$ Monte Carlo rollouts to estimate $Q(s_t,a)$,
and then we return the best action $\argmax_a Q(s_t,a)$ or
the action distribution $\softmax(Q(s_t,a))$.
To perform a rollout from a state $s$, we
proceed as follows.
\begin{itemize}
  \item Action selection:
If we have not visited $s$ before,
we initialize the node by setting $N(s,a)=0$ and $Q(s,a)=0$ and returning
$U(s)$ as the value,  where $U$ is some estimated value function.
Otherwise we pick the next action to explore from state $s$.
To explore actions, we first try each action once,
and we then use the \keywordDef{Upper Confidence Tree}
or \keywordDef{UCT} heuristic (based on UCB from \cref{sec:UCB})
to select subsequent actions,
i.e. we use
\be
a = \argmax_{a \in \calA(s)}
Q(s,a) + c \sqrt{ \frac{\log N(s)}{N(s,a)} }
\ee
where $N(s) = \sum_a N(s,a)$ is the total visit count to $s$,
and $c$ is an exploration bonus scaling term.
(Various other expressions are used in the literature,
see \citep{BrowneMCTS} for a discussion.)
If we have a predictor or prior over over actions,
$P(s,a)$,
we can instead use
\be
a = \argmax_{a \in \calA(s)}
Q(s,a) + c \left( P(s,a) \frac{\sqrt N(s)}{1 + N(s,a)} \right)
\ee

\item Expansion:
After choosing action $a$, we sample the next state
$s' \sim p(s'|s,a)$.

\item Rollout: we recursively estimate $u=U(s')$ using MCTS
  from that node.
  At some depth, we stop and use the value function
  to return $u = r + \gamma v(s')$.

  \item Backup: Finally
    we update the Q function for the root node
    using a running average:
\be
Q(s,a) \leftarrow Q(s,a) + \frac{1}{N(s,a)} (u - Q(s,a))
\ee
where the learning rate is given by $\frac{1}{N(s,a)}$.
We also increment $N(s,a)$ by 1.
\end{itemize}

\begin{figure}
  \centering
  \begin{tabular}{cc}
    \includegraphics[height=2in]{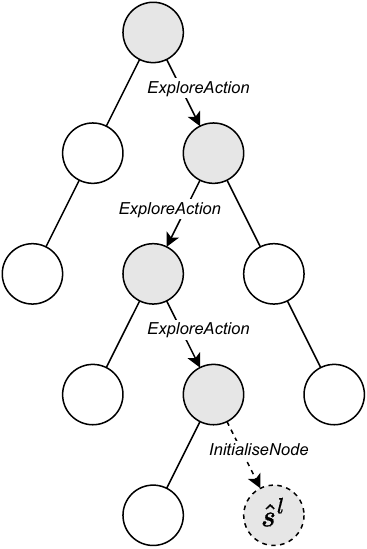}
    &
    \includegraphics[height=2in]{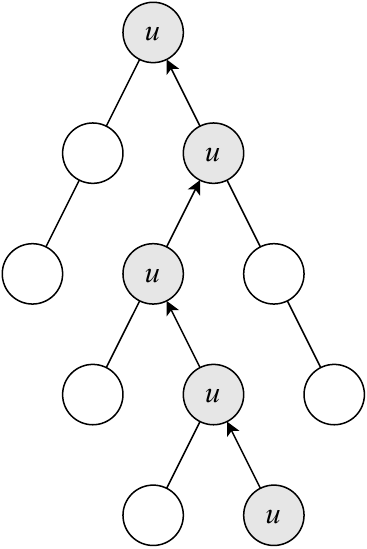} \\
    (a) & (b)
    \end{tabular}
\caption{
  Illustration of  MCTS.
  (a) Expanding nodes until we hit a new (previously unexplored) leaf node.
  (b) Propagating leaf value $u$ back up the tree.
  \figtaken{Figure 9.25 of \citep{marl-book}}.
}
\label{fig:MCTS}
\end{figure}

When we return from the recursive call, we are effectively
backpropagating the value $u$ from the leaves up the tree,
as illustrated in \cref{fig:MCTS}(b).
A sketch of a non-recursive version
of the algorithm
(Algorithm 25 of \citep{marl-book})
is shown in \cref{algo:MCTS}.

\begin{algorithm}[H]
  \caption{Monte Carlo Tree Search (MCTS)}
  \label{algo:MCTS}
        \For{$t = 0,1,2,3,\dots$}{
          Observe current state $\st_t$\;
              \For{$k$ \rm{simulations}}{
                $\tau \gets t$\;
                 $\sth_\tau \gets \st_t$

                \While{$\sth_\tau$ \rm{is non-terminal and node} $\sth_\tau$  \rm{ exists in tree}}{
                    $\ach_\tau \gets \text{ExploreAction}(\sth_\tau)$\;
                    $\sth_{\tau+1} \sim \Stf(\cdot \mid \sth_\tau, \ach_\tau )$\;
                    $\rewh_\tau \gets \Rew(\sth_\tau,\ach_\tau,\sth_{\tau+1})$\;
                    $\tau \gets \tau + 1$\;
                }
                
                \If{\rm{node } $\sth_\tau$ \rm{ does not exist in tree}}{
                    InitializeNode($\sth_\tau$) 
                }
                
                \While{$\tau > t$}{
                  \tcp{Backpropagate}
                    $\tau \gets \tau - 1$\;
                    Update($Q,\sth_\tau,\ach_\tau$)\;
                }
              
            }
            \tcp{Select action for state $\st_t$}
            $\pol_t \gets \text{BestAction}(\st_t)$\;
            $\ac_t \sim \pol_t$\;
        }
\end{algorithm}

\subsubsection{MCTS for 2p0s games: AlphaGo, AlphaGoZero, and AlphaZero}
\label{sec:alphaGo}
\label{sec:alphaZero}


MCTS can be applied to any kind of MDP, but some of its most
famous applications are to games.
We discuss general stochastic games in \cref{chap:MARL}, but here
focus on the special case of two-player, zero-sum symmetric games.
In this case, the agent can model the opponent using its own policy,
but with the roles reversed (this is known as \keywordDef{self-play},
see \cref{sec:selfPlay} for details).
This lets the main player treat is opponent
as part of the environment,
thus creating  a (non-stationary) single-agent problem.

In addition to choosing the next best action $a_t$ (as in RHC),
MCTS can be used to return  a distribution over good actions
for the current state $s$;
we denote this by
$\vpi^{\mcts}_s(a) = [N(s,a)/(\sum_b N(s,b))]^{1/\tau}$,
where $\tau$ is a temperature.
This can be used as a target for policy improvement.

This method was used in 
the  \keywordDef{AlphaGo} system of \citep{alphaGo},
which was the first AI system to beat a human grandmaster at the board game Go.
AlphaGo was followed up by \keywordDef{AlphaGoZero}
\citep{alphaGoZero},
which had a much simpler design,
and did not train on any human data,
i.e., it was trained entirely using RL and self play.
It significantly outperformed the original AlphaGo.
This was generalized to \keywordDef{AlphaZero} 
\citep{alphaZero},
which can play expert-level Go, chess, and shogi (Japanese chess),
without using any domain knowledge (except in the design
of the neural network used to guide MCTS).
(An open source version  of AlphaZero,
known as \keywordDef{Leela},
is available at \url{https://lczero.org/}.)

In more detail,
AlphaZero used MCTS (with self-play),
combined with a neural network which computes
$(v^s,\vpi^s) = f(s;\vtheta)$,
where $v_s$ is the expected outcome of the game from state $s$
(either +1 for a win, -1 for a loss, or 0 for a draw),
and $\vpi_s$ is  the policy (distribution over actions) for state $s$.
The policy is used  internally by MCTS
whenever a new node is initialized
to give an additional exploration bonus to 
the most promising / likely actions.
This controls the breadth of the search tree.
In addition, the learned value function $v_s=f(s;\vtheta)_v$ is used
to provide the value for leaf nodes in cases
where we cannot afford to rollout to termination.
This controls the depth of the search tree.

The policy/value network $f$ is trained  by optimizing the
actor-critic loss
\be
\loss(\vtheta) = \expectQ{(V^{\mcts}(s)-V_\vtheta(s))^2
  - \sum_a \vpi^{\mcts}_s(a) \log \vpi_{\vtheta}(a|s) }
     {(s, \vpi^{\mcts}_s, V^{\mcts}(s)) \sim \data}
     \label{eqn:AZloss}
\ee
where $\data = \{(s, \vpi^{\mcts}_{s}, V^{\mcts}_{s})\}$
is a dataset collected from  MCTS rollouts
starting at state $s$.
These rollouts generate a  distribution over actions at the root node
$s$ using
$\vpi^{\mcts}_s(a) = [N(s,a)/(\sum_b N(s,b))]^{1/\tau}$,
where $\tau$ is a temperature.
The rollouts also provide an
estimate of $Q(s,a)$ for each visited (state,action)
pair.
From this we can estimate
the non-parametric
state-value function $V^{\mcts}(s) = \max_a Q^{\mcts}(s,a)$.
\eat{
MC estimate of the reward-to-go
using
the n-step bootstrap estimate starting at root node $s_t$
and then computing
$V^{\mcts}_{s_t} = \sum_{i=0}^{n-1} \gamma^i r_{t+i} + \gamma^k v_{t+i}$
summing over each path (of length $n$) in the search tree.
}

The above self-play approach trains an agent against the current version of itself,
which can result in overfitting. To combat this, we can store multiple past versions
of the policy, and then select any of these policies as a proxy for the opponent's policy.
This increases robustness of the main agent.
Nevertheless, the resulting learned agent is susceptible
to \keywordDef{adversarial attacks};
indeed, 
\citep{Wang2023Go} showed that using this method,
humans can beat ``super-human'' Go-playing AI agents.

\subsubsection{MCTS with learned world model: MuZero and EfficientZero}
\label{sec:muZero}

AlphaZero and related methods assume  the world model is known.
The \keywordDef{MuZero} method of 
\citep{Schrittwieser2020}
learns a world model, by training a latent
representation (embedding function)
of the observations, $\vz_t=e_{\vphi}(\vo_t)$,
and a corresponding latent dynamics (and reward) model
$(\vz_t,r_t)=M_{\vw}(\vz_t,a_t)$.
The world model is trained
to predict the immediate reward,
the future reward (i..e, the value),
and the optimal policy,
where the optimal policy is computed  using MCTS.

In more detail, we use MCTS to select action $a_t$,
take a step,
and add $(\vo_t, a_t, r_t, \vo_{t+1}, \vpi_t^{\mcts}, V_t^{\mcts})$ to the replay buffer.
To train the model,
we augment  the loss in \cref{eqn:AZloss}
by adding a term that measures how well the learned
model predicts the observed rewards.
Also, 
we now optimize this wrt the policy/value parameters $\vtheta$
as well as the model parameters $\vw$ and embedding parameters
$\vphi$:
\begin{align}
\loss(\vtheta,\vw,\vphi) &=
\mathbb{E}_{(\vo, a_t, r, \vo', \vpi^{\mcts}_z, V^{\mcts}_z) \sim \data}
\left\{
  (V^{\mcts}(z)-V_{\vtheta}(e_{\vphi}(\vo))^2
- \sum_a \vpi^{\mcts}_z(a) \log \vpi_{\vtheta}(a|e_{\vphi}(\vo))
\right. \\
& \left.
+ (r - M^r_{\vw}(e_{\vphi}(\vo),a_t))^2
\right\}
     \label{eqn:MZloss}
\end{align}

MuZero was applied to 3 perfect information board games (Go, Chess, and Shogi),
as well as to Atari.
The  \keywordDef{Stochastic MuZero} method
of \citep{Antonoglou2022} extends MuZero to allow for stochastic
environments, such as the games 2048 and Backgammon.
The \keywordDef{Sampled MuZero} method of \citep{Hubert2021}
extends MuZero to allow for large and/or continuous action spaces.
The \keywordDef{Gumbel MuZero} method of
\citep{Danihelka2022}
proposes a  better
policy improvement algorithm, based on sampling actions without replacement,
that improves sample efficiency.
The \keywordDef{MuZero Unplugged} method of
\citep{MuZeroUnplugged} proposes the
\keywordDef{Reanalyse} algorithm,
that applies
an MCTS-based policy and value improvement operator,
together with  a learned world
model, to offline trajectories.\footnote{
Pseudocode for all these algorithm variants can be found at
\url{https://www.julian.ac/about/}.
}

\eat{
Although MuZero is a model-based method, it is not particularly
sample efficient, and needed 20B frames to get good results
on Atari. (Since it uses decision time planning, it is also not
computationally efficient either.)
}

\label{sec:efficientZero}
The \keywordDef{Efficient Zero}
paper  \citep{efficientZero} extends MuZero by adding
an additional  self-prediction loss
of the form $(\vz_{t+1}-M^z_{\vw}(\vz_t,a_t))^2$
to \cref{eqn:MZloss} to help train the world model.
(See \cref{sec:SPR} for further discussion of such losses.)
It also makes several other changes.
In particular, it replaces the empirical sum of instantaneous rewards,
$\sum_{i=0}^{n-1} \gamma^i r_{t+i}$,
used in computing $V_t^{\mcts}$,
with an LSTM model that predicts the sum of rewards
for a trajectory starting at $\vz_t$;
they call this the value prefix.
In addition, 
it replaces the stored value at the leaf nodes
of trajectories in the replay buffer with new values,
by rerunning MCTS using the current model
applied to  the leaves,
using the Reanalyse algorithm.
They show that all three changes help,
but the biggest gain is from the self-prediction  loss.
The recent \keywordDef{Efficient Zero V2}
\citep{efficientZeroV2} extends this to also work with continuous
actions,
by replacing tree search with sampling-based Gumbel search,
amongst other changes.

\subsubsection{MCTS in belief space}
\label{sec:betaZero}

In \citep{Moss2024}, they present \keywordDef{BetaZero},
which performs MCTS in belief space.
The current state is represented by a belief state,
$b_t$, which is passed to the network to generate
an initial policy proposal $\pi_{\vtheta}(a|b)$ and value function $v_{\vtheta}(b)$.
(Instead of passing the belief state to the network,
they actually pass features derived from the belief state,
namely the mean and variance of the states.\footnote{
The \keywordDef{POMCP} algorithm of \citep{Silver2010}
(Partially Observable Monte Carlo Planning) is related to BetaZero,
but passes observation-action histories as input to the policy/value network,
instead of features derived from the belief state.
The \keywordDef{POMCPOW}
algorithm of \citep{Sunberg2018}
(POMCP with observation widening)
extends this to continuous domains
by sampling observations and actions.
})

To rollout out trajectories from the current root node $b$,
they proceed as follows:
\begin{itemize}
\item Select an action $a \sim \pi_{\vtheta}(\cdot|b)$
  using UCT heuristic.

  \item Expand the node as follows:
sample the  current hidden state $s \sim b$,
sample the next hidden state $s \sim T(s'|s,a)$
sample the observation $o \sim O(s')$,
sample the reward $r \sim R(s,a,s')$;
and finally derive the new belief state $b' = \text{Update}(b,a,o)$
using e.g., a particle filter (see e.g., \citep{Lim2023}).

\item Simulate future returns using  rollouts
  to get $u = r + \gamma V_{\vtheta}(b')$ (assuming single step for
  notational simplicity);

\item Backup the values using $Q(b,a) +=  u$.
\end{itemize}

At the end of tree search, they derive the tree policy
$\vpi_t=\pi^{\mcts}(b_t)$ from the root,
and compute the empirical reward-to-go $g_t = \sum_{i=t}^T \gamma^{i-t} r_i$
based on all rewards observed below root node $b_t$;
this is added to a dataset $\data = \{ (b_t, \vpi_t, g_t)\}$
which is used to update the policy and value network.

\subsection{Sequential Monte Carlo (SMC) for online planning}
\label{sec:SMCMPC}
\label{sec:SMCRHC}

Although MCTS is powerful, it is inherently serial, and can be complicated
to apply to continuous action spaces.
In this section, we discuss a more general method known as
\keywordDef{SPO},
which stands for Sequential Monte Carlo Policy Optimisation
\citep{Macfarlane2024}.

SPO  is based on the ``RL as inference''
framework, and is discussed in more detail
in \cref{sec:maxentRL}.
In brief, the goal is to sample
trajectories (sequences of states and actions)
that are likely to be high scoring.
That is, we want to sample
from the following distribution
\be
q(\tau)  \propto
d_q(s_0) \prod_{t\geq 0} \calT(s_{t+1}|s_t,a_t)
\pi(a_t|s_t,\vtheta_i) \exp\left( \frac{A(s_t,a_t)}{\eta} \right)
\label{eqn:qitau}
\ee
where $s_0$ is the current state,
$A$ is the advantage function
\be
A(s_t,a_t) = Q(s_t,a_t) - V(s_t)
\approx r_t + V(s_{t+1}) - V(s_t)
\ee
and $\eta$ is a temperature parameter
obtained by maximizing \cref{eqn:geta}.
(The state-value function $V$ can be learned via TD(0.)

Let the resulting empirical distribution over trajectories
be denoted by
\be
\hat{q}_i(\tau) = \sum_{n=1}^N \overline{w}^n \delta(\tau - \tau^n)
\ee
where $\tau^n$ is the $n$'th sample, and $\overline{w}^n$ is its
(normalized) weight.
We can derive the distribution over next best action
as follows:
\be
\hat{q}(a|s_0) = \sum_n \overline{w}^n \delta(a-a_0^n)
\ee

One way to sample trajectories from
such a distribution  is to use
\keywordDef{SMC} (Sequential Monte Carlo),
which is a generalization of \keywordDef{particle filtering}.
This is an approach to approximate inference in state space
models based on sequential importance sampling
with resampling  (see e.g., \citep{Naesseth2019}).
At each step, we use a proposal distribution
$\beta(\tau_t|\tau_{1:t-1})$, which extends the
previous sampled trajectory with a new value of $x_t=(s_t,a_t)$.
We then compute the weight of this proposed extension by
comparing it to the target $q_i(\tau_t|\tau_{1:t})$
to get
\be
w(\tau_{1:t})  \propto w(\tau_{1:t-1})
\frac{q_i(\tau_{1:t})}{\beta(\tau_{1:t})}
\ee
Suppose we use the following proposal
\be
\beta_i(\tau_t|\tau_{1:t-1}) \propto
\hat{\calT}(s_{t+1}|s_t,a_t) \pi(a_t|s_t,\vtheta_i)
\ee
Then the weight is given by
\be
w(\tau_{1:t})  \propto w(\tau_{1:t-1})
\frac{\calT(s_t|s_{t-1},a_{t-1})}{\hat{\calT}(s_t|s_{t-1},a_{t-1})}
\cdot
\frac{\exp(A_i(s_t,a_t)/\eta_i^*) \pi(a_t|s_t,\vtheta_i)}
{\pi(a_t|s_t,\vtheta_i)}
\ee
If we assume the learned model $\hat{\calT}$ is accurate,
this simplifies to
\be
w(\tau_{1:t})  \propto w(\tau_{1:t-1})
\cdot
\exp(A(s_t,a_t)/\eta) 
\ee
In SMC, at each step we propose a new particle according to $\beta$,
and then weight it according to the above equation.
We can then optionally resample the particles every few steps,
or when the effective sample size becomes too small;
after a resampling step, we reset the weights to 1,
since we  now have a weighted sample.
At the end, we return an empirical distribution over
actions that correspond to high scoring trajectories,
from which we can estimate the next best action
(e.g., by taking the mean or mode of this distribution).
See \cref{algo:SMCRHC} for details.
(See also \citep{Piche2019,Lioutas2022} for related
methods.)

Note that the above framework is a special case of
\keywordDef{twisted SMC}
\citep{Naesseth2019,Lawson2022,Zhao2024},,
where the advantage function plays the role of a ``twist'' function,
summarizing expected future rewards from the current state.

\begin{algorithm}
\dontprintsemicolon
\caption{SMC-RHC (Sequential Monte Carlo for Receeding Horizon Control)}
\label{algo:SMCRHC}
def SMC-RHC($s_t$, $\pi_i$, $V_i$): \\
Initialize particles:  $\{\vs_t^n=\vs_t\}_{n=1}^N$ \\
Initialize weights: $\{w_t^n = 1\}_{n=1}^N$ \\
\For{$j=t+1:t+d$}
    {
      $\{ a_j^n \sim \pi_i(\cdot|s_j^n) \}_{n=1}^N$  \\
      $\{ s_{j+1}^n \sim \hat{\calT}(s_j^n, a_j^n) \}_{n=1}^N$ \\
      $\{ r_{j}^n \sim \hat{\calR}(s_j^n, a_j^n) \}_{n=1}^N$ \\
      $\{ x_j^n = (s_j^n, a_j^n, r_j^n) \}_{n=1}^N$ \\
      $\{ A_j^n = r_j^n + V_i(s_{t+1}^n) - V_i(s_j^n) \}_{n=1}^N$ \\
      $\{ w_j^n = w_{j-1}^n \exp(A_i^n / \eta_i^*) \}_{n=1}^N$ \\
      \If{\text{Resample}}{
    $\{ \vx_{t:j}^n \} \sim \text{Multinom}(n; w_i^1, \ldots, w_i^N) \}$ \\
        $\{ w_j^n = 1 \}_{n=1}^N$
      }
    $\{ \overline{w}^n = \frac{w^n}{\sum_{n'} w^{n'}} \}_{n=1}^N$  \\
    Let $\{a_t^n \}_{n=1}^N$ be the set of sampled actions at the start
    of $\{ x_{t:t+d}^n \}_{n=1}^N$ \\
    Return $\hat{q}(a|s_t) = \sum_n \overline{w}^n \delta(a-a_t^n)$
    }
\end{algorithm}

\subsection{Model predictive control (MPC), aka open loop planning}
\label{sec:MPC}

In this section, we describe a method known as
\keywordDef{model predictive control} (\keywordDef{MPC}),
which is an \keywordDef{open loop} version of receeding horizon control
\citep{Mayne1990,Camacho2013,Rawlings2022}.
(Connections with RL 
are discussed in e.g., \citep{Bertsekas2024MPC,Lawrence2025,}.)
In particular, at each step, it solves for the sequence of subsequent actions
that is most likely to achieve high  expected reward:
\begin{align}
  \va_{t:t+d}^* 
  &= \argmax_{\va_{t:t+d}}
\expectQ{\sum_{h=0}^{d} R(s_{t+h}, a_{t+h}) + \hat{V}(s_{t+d+1})}
{s_{t+1:t+d} \sim \calT(\cdot|s_t, a_{t:t+d})} 
\label{eqn:MPC} 
\end{align}
where $\calT$ is the dynamics model.
It then returns $a_t^*$ as the best action, takes a step, and replans.

Crucially, the future actions are chosen without knowing what the future
states are; this is what is meant by ``open loop''.
This can be much faster than interleaving the search for actions
and future states.
However, it can also lead to suboptimal decisions, as we discuss below.
Nevertheless, the fact that we replan at each step can reduce the harms of this
approximation, making the method quite popular for some problems,
especially ones where the dynamics are deterministic,
and the actions are continuous
(so that  \cref{eqn:MPC}
becomes a standard optimization problem over
the real valued sequence of vectors $\va_{t:t+d-1}$).

\subsubsection{Suboptimality of open-loop planning for stochastic environments}

\begin{figure}
\centering
\includegraphics[height=2in]{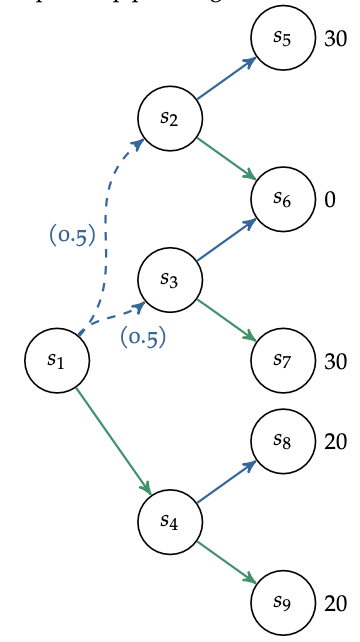}
\caption{
  Illustration of the suboptimality of open-loop planning.
  \figtaken{Figure 9.9 of \citep{KochenderferDecisions}}.
    \figthanks{Mykel Kochenderfer}.
}
\label{fig:openLoop}
\end{figure}

Consider the example in \cref{fig:openLoop},
where there are 9 states, 2 actions (going up or down),
and the planning horizon is $d=2$.
All transitions are deterministic, except
that going ip from $s_1$ can either end up in $s_2$
wp 0.5 or in $s_3$ wp 0.5.

There are 4 open-loop plans,
with the following expected utilities:
\begin{itemize}
\item U(up,up) = $0.5 \times 30 + 0.5 \times 0 = 15$
\item U(up,down) = $0.5 \times 0 + 0.5 \times 30 = 15$
\item U(down,up) = 20
\item U(down,down) = 20
  \end{itemize}
Thus the best open-loop action is to choose down, with an expected reward of 20.
However, closed-loop planning can reason that, after taking the first action,
the agent can sense the resulting state.
If it initially chooses to go up from $s_1$, then it can decide to next go up
or down, depending on whether it is in $s_2$ or $s_3$,
thereby guaranteeing a reward of 30.

\subsubsection{Trajectory optimization}

If the dynamics is deterministic, the problem becomes one of solving
\begin{align}
  \max_{a_{1:d}, s_{2:d}} \sum_{t=1}^d \gamma^t R(s_t, a_t) \\
  \myst s_{t+1} = \calT(s_t,a_t)
  \end{align}
where $\calT$ is the transition function.
This is called a \keywordDef{trajectory optimization} problem.
We discuss various ways to solve this below.

\subsubsection{LQR}

If the system dynamics are linear
and the reward function is quadratic,
then the optimal action sequence can be computed exactly
using a method similar to Kalman filtering.
This is known as the
\keywordDef{linear quadratic regulator}
(\keywordDef{LQR}).
For details,
see e.g., \citep{Anderson89,Hoffmann2017control,Shaiju2008}.

If the model is nonlinear, we can use
\keywordDef{differential dynamic programming}
(\keywordDef{DDP})~\citep{Jacobson70,Todorov05}
to approximately solve the problem.
In each iteration, DDP starts with a reference trajectory,
and linearizes the system dynamics around states on
the trajectory to form a locally quadratic approximation
of the reward function.
This system can be solved using LQG,
whose optimal solution results in a new trajectory.
The algorithm then moves to the next iteration,
with the new trajectory as the reference trajectory.

\subsubsection{Random shooting}

For general nonlinear models (such as neural networks),
a simple approach
is to pick a sequence of random actions to try (from some proposal policy),
evaluate the reward for each trajectory,
and pick the best.
This is called \keywordDef{random shooting}
\citep{Diehl2007,Rao2010}.

\subsubsection{CEM}
\label{sec:CEM}

As an improvement upon random shooting,
it is common to use
black-box (gradient-free) optimization methods
like the \keywordDef{cross-entropy method} or \keywordDef{CEM}
in order to find the best action sequence.
The CEM method is a simple derivative-free optimization
method for continuous black-box functions $f: \real^D \ra \real$.
We start with a multivariate Gaussian, $\gauss(\vmu_0,\vSigma_0)$,
representing a distribution over possible action $\va$.
We sample from this, evaluate all the proposals,
pick the top $K$, then refit the Gaussian to these top $K$,
and repeat, until we find a sample with sufficiently good score
(or we perform moment matching on the top $K$ scores).
For details, see \citep{Rubinstein1997,Rubinstein2004,DeBoer2005}.
In \cref{sec:MPPI}, we discuss the MPPI method,
which is a common instantiation of CEM method.
In \citep{Bharadhwaj2020} they discuss how to combine CEM with gradient-based planning.

\eat{
It is also possible to create differentiable versions
of CEM and to backpropagate through them
to train the dynamics model,
as we discuss in \cref{sec:differentiable-planning}.
}

\subsubsection{MPPI}
\label{sec:MPPI}


The \keywordDef{model predictive path integral}
or \keywordDef{MPPI} approach \citep{MPPI}
is a version of CEM. 
Originally MPPI was limited to models with linear dynamics,
but it was extended to general nonlinear models in 
\citep{Williams2017}.
The basic idea is that the initial mean of the Gaussian
at step $t$, namely $\vmu_t=\va_{t:t+H}$,
is computed based on shifting $\hat{\vmu}_{t-1}$ forward by one step.
(Here $\vmu_t$ is known as a reference trajectory.)

In \citep{Wagener2019}, they apply this method for robot control.
They consider a state vector
of the form $\vs_t=(\vq_t,\dot{\vq}_t)$, where $\vq_t$ is the
configuration
of the robot. The deterministic dynamics has the form
\be
\vs_{t+1} = F(\vs_t,\va_t) = \begin{pmatrix}
  \vq_t + \dot{\vq}_t \Delta t \\
  \dot{\vq}_t + f(\vs_t,\va_t) \Delta t
  \end{pmatrix}
\ee
where $f$ is a 2 layer MLP.
This is trained using the
\keywordDef{Dagger} method of \citep{Ross2011},
which alternates between fitting the model (using supervised learning)
on the current replay buffer (initialized with expert data),
and then deploying the model 
inside the MPPI framework to collect new data
(see \cref{sec:dagger} for details).

A similar method was used
in the \keyword{TD-MPC} paper \citep{Hansen2022,Hansen2024},
which learns a non-generative world model  in  latent space,
and then uses MPPI to implement MPC  (see
\cref{sec:TDMPC} for details).
They initialize the population of $K$ sampled
action trajectories by applying the policy prior
to generate $J < K$ samples,
and then generate the remaining $K-J$ samples
using the diagonal Gaussian prior from the previous
time step.

\subsubsection{GP-MPC}
\label{sec:GPMPC}

\citep{Kamthe2018}  proposed
\keywordDef{GP-MPC}, which combines a Gaussian process dynamics model
with model predictive control.
They  compute
a Gaussian approximation to the future state trajectory
given a candidate action trajectory,
$p(\vs_{t+1:t+H}|\va_{t:t+H-1},\vs_t)$,
by moment matching,
and use this to deterministically compute
the expected reward and its gradient wrt
$\va_{t:t+H-1}$.
Using this, they can solve \cref{eqn:MPC}
to find  $\va_{t:t+H-1}^*$;
finally, they execute the first step of this plan,
 $a_{t}^*$,
and repeat the whole process.

The key observation is that moment matching is a deterministic
operator that maps $p(\vs_t|\va_{1:t-1})$ to $p(\vs_{t+1}|\va_{1:t})$,
so the problem becomes one of deterministic optimal control,
for which many solution methods exist.
Indeed the whole approach can be seen as a generalization
of the \keywordDef{LQG} method from  classical control,
which assumes a (locally) linear dynamics model,
a quadratic cost function,
and a Gaussian distribution over states
\citep{Recht2019}.
In GP-MPC, the moment matching plays the role of local linearization.

The advantage of GP-MPC  over the earlier method
known as \keywordDef{PILCO}
(``probabilistic inference for learning control''),
which learns a policy  by maximizing the expected reward
from rollouts (see \citep{pilco,pilcoJ} for details),
is that GP-MPC can handle constraints more easily,
and it can be more data efficient,
since it continually updates the GP model after every  step
(instead of at the end of an trajectory).

\section{Background (offline) planning}
\label{sec:MBRLcombine}
\label{sec:MBRLrobust}
\label{sec:backgroundPlanning}

In \cref{sec:decisionTimePlanning},
 we discussed how to use
models to perform decision-time planning.
However, this can be slow.
Fortunately, we can amortize the planning process into a reactive
policy.
To do this, we can use the model to generate synthetic trajectories
``in the background'' (while executing the current policy),
and use this imaginary  data
to train the policy;
this is called ``\keywordDef{background planning}''.
We discuss a game theoretic formulation of this setup in
\cref{sec:MBRLgame}.
Then in \cref{sec:dyna},
we discuss ways to combine model-based and model-free learning.
Finally, in \cref{sec:modelUncertainty},
we discuss ways to deal with model errors, that might lead the policy astray.

\subsection{A game-theoretic perspective on MBRL}
\label{sec:MBRLgame}

In MBRL, we optimize the policy wrt a possibly incorrect model,
and then update the model given new data, in an iterative fashion.
In this section, we discuss a game-theoretic framework for 
MBRL, as proposed in  \citep{Rajeswaran2020},
which justifies why this approach should converge.
(See \cref{chap:MARL} for a more general discussion of game theory and RL.)

We denote the true world model  by $\Mtrue$.
To simplify the notation, we assume an MDP setup with a known 
reward function,
so all that needs to be learned is the world model, $\Mest$, representing
$p(s'|s,a)$.
(It is trivial to also learn the reward function.)
We define the value of a policy $\pi$ when rolled out in some model $M'$
as the (discounted) sum of expected rewards:
\begin{align*}
J(\pi,M') = \expectQ{\sum_{t=0}^{\infty} \gamma^t R(s_t)}{M',\pi}
\end{align*}
We define the loss of a model $\Mest$ given a distribution
$\mu(s,a)$ over states and actions as
\begin{align*}
  \lossfn(\Mest,\mu) = \expectQ{\KLpq{\Mtrue(\cdot|s,a)}{\Mest(\cdot|s,a)}}
  {(s,a) \sim \mu}
\end{align*}
We now define MBRL as a two-player general-sum game
(see \cref{sec:MARL} for details):
\begin{align*}
  \overbrace{\max_{\pi} J(\pi, \Mest)}^{\text{policy player}},
  \overbrace{\min_{\Mest} \lossfn(\Mest, \mu_{\Mtrue}^{\pi})}^{\text{model player}}
\end{align*}
where $\mu_{\Mtrue}^{\pi} = \frac{1}{T} \sum_{t=0}^T \Mtrue(s_t=s,a_t=a)$
is the induced state visitation distribution when policy
$\pi$ is applied in the real world $\Mtrue$,
so that
minimizing $\lossfn(\Mest, \mu_{\Mtrue}^{\pi})$ gives the
\keywordDef{maximum likelhood estimate} for $\Mest$.

Now consider a \keywordDef{Nash equilibrium} (\cref{sec:nash})
of this game, that is a pair $(\pi,\Mest)$ that satisfies
$\lossfn(\Mest,\mu_{\Mtrue}^{\pi}) \leq \epsilon_{\Mtrue}$
and $J(\pi,\Mest) \geq J(\pi', \Mest) - \epsilon_{\pi}$ for all $\pi'$.
(That is, the model is accurate when predicting the rollouts from  $\pi$,
and $\pi$ cannot be improved when evaluated in $\Mest$).
In  \citep{Rajeswaran2020} they prove that the
Nash equilibirum
policy $\pi$ is  near optimal wrt the real world,
in the sense that $J(\pi^*, \Mtrue) - J(\pi,\Mtrue)$ is bounded by a constant,
where $\pi^*$ is an optimal policy for the real world $\Mtrue$.
(The constant depends  on the $\epsilon$ parameters,
and the TV distance between $\mu_{\Mtrue}^{\pi^*}$
and $\mu_{\Mest}^{\pi*}$.)

A natural approach to trying to find such a Nash equilibrium
is to use \keywordDef{gradient descent ascent} or \keywordDef{GDA},
in which each player updates its parameters simultaneously,
using
\begin{align*}
  \pi_{k+1} &= \pi_k + \lr_{\pi} \nabla_{\pi} J(\pi_k, \Mest_k) \\
  \Mest_{k+1} &= \Mest_k
  -\lr_{M} \nabla_{\Mest} \lossfn(\Mest_k, \mu_{\Mtrue}^{\pi_k})
\end{align*}
Unfortunately, GDA is often an unstable algorithm,
and often needs very small learning rates $\lr$.
In addition, to increase sample efficiency in the real world,
it is better to make multiple policy improvement steps
using synthetic data from the model  $\Mest_k$ at each step.

Rather than taking small steps in parallel,
the \keywordDef{best response} strategy fully optimizes
each player given the previous value of the other player,
in parallel:
\begin{align*}
  \pi_{k+1} &= \argmax_{\pi} J(\pi, \Mest_k) \\
  \Mest_{k+1} &= \argmin_{\Mest} \lossfn(\Mest, \mu_{\Mtrue}^{\pi_k})
\end{align*}
Unfortunately, making such large updates in parallel
can often result in a very unstable algorithm.

To avoid the above problems,
\citep{Rajeswaran2020}
propose to replace the min-max
game with a \keywordDef{Stackelberg game}, which is a generalization
of min-max games where we impose a specific player ordering
(\cref{sec:stackelberg}.
In particular, let the players be $A$ and $B$,
let their parameters be $\theta_A$ and $\theta_B$,
and let their losses be
$\loss_A(\theta_A,\theta_B)$
and
$\loss_B(\theta_A,\theta_B)$.
If player $A$ is the leader, the Stackelberg game corresponds
to the following \keywordDef{nested optimization problem},
also called a \keywordDef{bilevel optimization problem}:
\begin{align*}
  \min_{\theta_A} \loss_A(\theta_A, \theta^*_B(\theta_A))
  \myst
  \theta^*_B(\theta_A)  = \argmin_{\theta_B} \loss_B(\theta_A, \theta_B)
\end{align*}
Since the follower $B$ chooses the best response
to the leader $A$, the follower's parameters are a function
of the leader's. The leader is aware of this, and can utilize
this when updating its own parameters.

The main advantage of the Stackelberg approach  is that one can derive
gradient-based algorithms that will provably converge to a local
optimum \citep{Colson2007,Zucchet2022,Gerstgrasser2023}.
In particular, suppose we choose the \keywordDef{policy as leader}
(\keywordDef{PAL}).
We then just have to solve the following optimization problem:
\begin{align*}
  \Mest_{k+1} &= \argmin_{\Mest} \lossfn(\Mest, \mu_{\Mtrue}^{\pi_k}) \\
  \pi_{k+1} &= \pi_k + \lr_{\pi} \nabla_{\pi} J(\pi_k, \Mest_{k+1})
\end{align*}
We can solve the first step by executing
$\pi_k$ in the environment to collect data $\data_k$
and then fitting a local (policy-specific) dynamics model
by solving $\Mest_{k+1} = \argmin \lossfn(\Mest, \data_k)$.
(For example, this could be a locally linear model,
such as those used in trajectory optimization
methods discussed in \cref{sec:MPPI}.)
We then (slightly) improve the policy to get $\pi_{k+1}$ using a
conservative update algorithm, such as natural actor-critic (\cref{sec:NPG})
or TRPO (\cref{sec:TRPO}),
on ``imaginary'' model rollouts from $\Mest_{k+1}$.

Alternatively, suppose we choose the
\keywordDef{model as leader} (\keywordDef{MAL}).
We now have to solve
\begin{align*}
  \pi_{k+1} &= \argmax_{\pi} J(\pi, \Mest_k) \\
  \Mest_{k+1} &= \Mest_k - \lr_{M} \nabla_{\Mest}
  \lossfn(\Mest, \mu_{\Mtrue}^{\pi_{k+1}})
\end{align*}
We can solve the first step by using  any RL algorithm
on ``imaginary'' model rollouts from $\Mest_{k}$
to get $\pi_{k+1}$.
We then apply this in the real world to get data $\data_{k+1}$,
which we use to slightly improve the model to get $\Mest_{k+1}$
by using a conservative model update applied to $\data_{k+1}$.
(In practice we can implement a conservative model update
by mixing $\data_{k+1}$ with data generated from earlier
models, an approach known as \keywordDef{data aggregation}
\citep{Ross2012}.)
Compared to PAL, the resulting model will be a more  global model,
since it is trained on data from a mixture of policies
(including very suboptimal ones at the beginning of learning).

\eat{
Some generic pseudocode for MBRL is shown in  \cref{algo:MBRL}.
Here $N_{\tpg}$ is the number of policy gradient updates
(using any on-policy method) per imaginary trajectory,
and $N_{\tmodel}$ is the number of model updates
(using any training method, such as MLE) per real trajectory
(of length $T_{\tenv}$).
To implement MAL, we can set $N_{\tpg} \gg N_{\tmodel}$,
so that the update-to-data ratio is high.
To implement PAL, we can set $N_{\tpg}$ to be small
(so we update the policy slowly);
we also have to ensure  that the update-model step
is run to completion.
Note that in either case,
the total number of steps in the real environment is $N_{\ttot} \times T_{\tenv}$;
everything else is done ``in imagination''.
}

\eat{
To approximate the idea that we learn a new optimal
policy at each step $k$ in response to making small changes to the model,
we set the number of policy updates per step, $N_{\tpg}$,
to be much larger than the number of model updates per step, $N_{\tmodel}$.
We assume the rollout function generates a trajectory of desired length $T$
using the  policy applied to the specified model (either learned model or true environment model).
If the trajectory terminates early, we reset the state and generate another trajectory
until we have taken a total of $T$ steps.
The policy gradient step function corresponds to any (on-policy) policy gradient method,
such as the natural policy gradient
method of \cref{sec:NPG},
or the PPO method of \cref{sec:PPO}.
}

\eat{
\subsubsection{Distributionally robust models and policies}

We can define a more robust notion of Stackelberg equilibrium
by optimizing player $A$'s performance under the worst
case realization for player $B$:
\begin{align*}
  \min_{\theta_A} \max_{\theta_B \in R(\theta_A)}
  \loss_A(\theta_A, \theta^*_B(\theta_A))
  \st
  R(\theta_A)  = \{ \theta:
  \loss_B(\theta_A, \theta) \leq \loss_B(\theta_A, \theta_B')
  \forall \theta_B' \}
\end{align*}
where $R(\theta_A)$ is the best response set to player $A$.

In PAL, 
suppose we approximate the best response set $R(\pi_k)$
with an ensemble of models
(as discussed in \cref{sec:BNNworldModel}).
The \keyword{EPOpt} algorithm of
\citep{Rajeswaran2016} optimizes the policy wrt
the adversarial choice of model from $R(\pi_k)$,
resulting in a robust Stackelberg approach.

In MAL, we can think of methods which encourage policy exploration
(such as entropy regularization, or intrinsic rewards,
discussed in \cref{sec:intrinsicReward}),
as approximating the best response $R(\hat{M}_k)$
with a set of (locally optimal) policies and then returning the one
that maximizes the prediction error
for the current model $\hat{M}_k$.
This will cause data $\data_{k+1}$ to be collected
in new parts of space to maximally  improve the model.

\subsubsection{Experimental comparison of PAL and MAL}

In \citep{Rajeswaran2020}, they show good results for the PAL and MAL
methods on various continuous control tasks (from state-based inputs),
outperforming model-free SAC method (\cref{sec:SAC})
and model-based MBPO method (\cref{sec:MBPO}).
They even manage to use the method on various real robots
solving manipulation and locomotion tasks.
In their experiments, they used an MLP policy with a diagonal Gaussian output.
(Enforcing a lower bound on the predicted variance avoids entropy collapse.)
For the world model, they used an ensemble of 4 MLPs.
(Using an ensemble makes policy learning more robust,
as we discuss below.)
In MAL, they found it necessary to perform 
$T_{\tinit}=5000$ environments steps with  the initial random policy
to collect enough data to learn the initial model;
this warmup phase could of course be replaced by any MFRL method,
rather than using a fixed random policy.
They perform $N_{\tpol}=25$ policy updates (in imagination) per step,
and $N_{\tmodel}=10$ model updates (on real data) per step.
They collect $T_{\tenv}=20 H$ environment steps after
each policy update, where $H$ is the horizon length of the task
for each environment (which depends on the reward sparsity and the discount factor).
For the imaginary rollouts,
they start from a previously visited real-world state,
sampled at different points along the trajectory,
to ensure good state coverage \citep{Rajeswaran2017}.
They use imaginary rollouts of length $T_{\timag}=\min(H,500)$,
although this is of course environment-specific.
(Rollouts can also get truncated early if the process terminates,
either in the real world or in simulation.)

}

\subsection{Dyna}
\label{sec:dyna}
\label{sec:MBPO}

The \keywordDef{Dyna} paper \citep{Sutton1990}
proposed an approach to MBRL that is related
to the approach discussed in 
\cref{sec:MBRLgame},
in the sense that it trains a policy and world model in parallel,
but it differs in one crucial way:
the policy is also trained on real data, not just imaginary data.
That is, we define
$\pi_{k+1} = \pi_k + \lr_{\pi} \nabla_{\pi} J(\pi_k, \hat{D}_k \union \data_k)$,
where
 $\data_k$ is data
from the real environment
and 
$\hat{D}_k = \text{rollout}(\pi_k, \hat{M}_k)$
is imaginary data from the model.
This makes Dyna a hybrid model-free and model-based RL method,
rather than a ``pure'' MBRL method.

\begin{figure}
\centering
\includegraphics[height=3in]{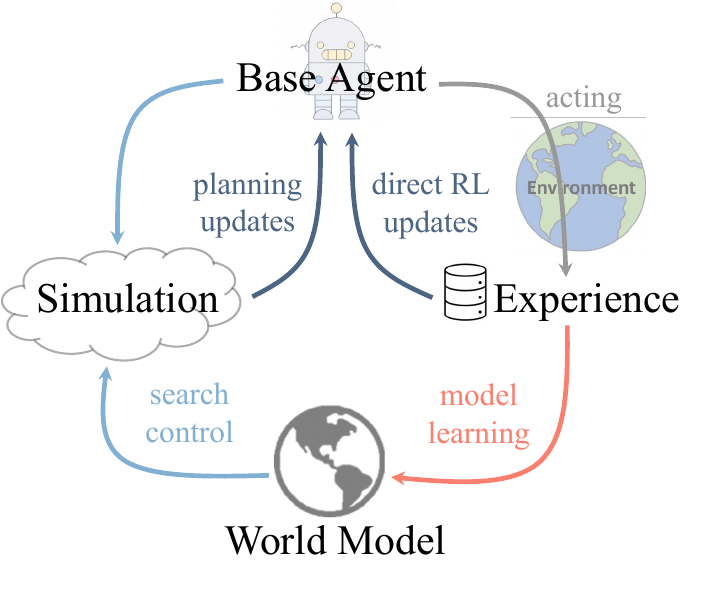}
\caption{
  Illustration of Dyna.
  \figtaken{Figure 2 of \citep{Liu2024wm}}.
  \figthanks{Zichen Liu}.
}
\label{fig:dyna}
\end{figure}

In more detail,
at each step of Dyna, the agent collects new data
from the environment and adds it to a real replay buffer.
This is then used to do an off-policy update.
It also updates its world model given the real data.
Then it simulates  imaginary data,
starting from a previously visited state
(see {\tt sample-init-state} function
in \cref{algo:MBRL}),
and rolling  out the current policy in the learned model.
The imaginary data is then added to the imaginary replay buffer
and used by an on-policy learning algorithm.
This process continue until the agent runs
out of time and must take the next step in the environment.
See \cref{fig:dyna} for an illustration.

\subsubsection{Tabular Dyna}

\begin{algorithm}
\dontprintsemicolon
\caption{Tabular Dyna-Q}
\label{algo:dynaQ}
def dyna-Q-agent$(s, \Mtrue; \eps, \lr, \gamma)$: \\
Initialize 
data buffer $\data=\emptyset$, $Q(s,a)=0$ and $\hat{M}(s,a)=0$ \\
\Repeat{until converged}
       {
               // Collect real data from environment \\
         $a = \text{eps-greedy}(Q, \eps)$ \\
         $(s',r) = \text{env.step}(s,a)$ \\
               $\data = \data \union \{(s,a,r,s')\}$ \\
               // Update policy on real data \\
               $Q(s,a) := Q(s,a) + \lr[r + \gamma \max_{a'} Q(s',a') - Q(s,a)]$ \\
               // Update model on real data \\
         $\hat{M}(s,a) = (s', r)$ \\
         $s := s'$ \\
         // Update policy on imaginary data \\
         \For{n=1:N}
             {
               Select $(s,a)$ from $\data$ \\
               $(s',r) =\hat{M}(s,a)$ \\
               $Q(s,a) := Q(s,a) + \lr[r + \gamma \max_{a'} Q(s',a') - Q(s,a)]$ 
               }
         }
\end{algorithm}

 The original Dyna paper was developed  under the assumption
 that the world model $s'=M(s,a)$ is deterministic and tabular,
 and the $Q$ function is also tabular.
 See \cref{algo:dynaQ} for the simplified pseudocode for this case.
Since  we assume a deterministic world model of the form
 $s'=M(s,a)$, then sampling a single step from this starting
 at a previously visited state is equivalent
 to \keyword{experience replay} (\cref{sec:ER}).
 Thus  we can think of ER as a kind of non-parametric world model
 \citep{VanHasselt2019}.

\subsubsection{Dyna with function approximation}

It is easy to extend Dyna to work with  function approximation
and policy gradient methods.
The code is identical to the MBRL code in \cref{algo:MBRL},
where now we train the policy on real as well as imaginary data.
(\citep{Lai2021} argues that we should gradually increase
the fraction of real data that is used to train the policy,
to avoid suboptimal performance due to model limitations.)
If we use real data from the replay buffer,
we have to use an off-policy learner,
since the replay buffer contains trajectories that may have been
generated from  old policies.
(The most recent real trajectory,
and all imaginary trajectories,
are always  from the current policy.)

We now mention some examples of this ``generalized Dyna'' framework.
In \citep{dynaLinear} they extended Dyna to the case
where the $Q$ function is linear,
and in \citep{Holland2018}
they extended it to the DQN case.
In \citep{Janner2019}, they present the
\keywordDef{MBPO}
(model based policy optimization) algorithm,
which uses Dyna with the off-policy SAC method.
Their world model is an
\keywordDef{ensemble of DNNs}, which generates diverse predictions
(an approach which was originally proposed in
the \keywordDef{PETS}
(probabilistic ensembles with trajectory sampling)
paper of \citep{Chua2018}).\footnote{
In \citep{Zheng2022ensemble} they argue that the main benefit
of an ensemble is that it limits the Lipschitz constant
of the value function. They show that more direct methods
for regularizing this can work just as well, and are much faster.
}
In \citep{Kurutach2019}, they combine
Dyna with TRPO (\cref{sec:TRPO}) and ensemble world models,
and 
in \citep{Wu2023dyna} they combine Dyna with PPO and GP world models.
(Technically speaking, these on-policy approaches are not valid with Dyna,
but they can work if the replay buffer  used for policy training
is not too stale.)

\eat{
\footnote{
Interestingly, \citep{Rajeswaran2020} claim that their Stackelberg
approach
works better than MBPO,
\citep{Janner2019}, which is based on Dyna.
In both cases the model is an ensemble of DNNs.
Indeed, \citep{Rajeswaran2020} write:
``we note that MBPO is a hybrid model-based and model-free method,
while our  MAL and PAL implementations is entirely model-based.
In MBPO, it was noted that long horizon model-based rollouts were unstable and
combining with an off-policy critic was important. We find that through our
Stackelberg formulation, which is intended to carefully control the effects
of distribution shift, we are able to perform rollouts of hundreds of steps
without error amplification.``
However, in MAL, they use the on-policy
natural policy gradient method (\cref{sec:NPG})
as the policy learner,
whereas in MBPO, they use off-policy SAC method (\cref{sec:SAC}),
so it is possible the improvement comes from this difference,
or other implementation details,
rather than the difference between Stackelberg and Dyna.
}
}

\section{World models}
\label{sec:WM}
\label{sec:genWM}

\renewcommand{\latent}{\vz}
\newcommand{\latentdynamics}{\calM}
\newcommand{\obsdynamics}{\calT}
\newcommand{\update}{\calU}

\renewcommand{\arraystretch}{1.5} 

\begin{table}[h]
    \centering
    \begin{tabularx}{\textwidth}{|X|X|X|X|}
        \hline
        & \textbf{Background planning} & \textbf{Online planning} & \textbf{Exploration} \\ \hline
        \textbf{Observation prediction} & 
        Dyna, DreamerV3, IRIS, Delta-IRIS, Diamond & 
        CEM: PlaNet \newline Rnd shooting: TDM & 
        SPR \\ \hline
        \textbf{Value + self prediction} & 
        DreamingV2, AIS & 
        MCTS: MuZero, EfficientZero  \newline
        CEM: TD-MPC & 
        BYOL-Explore \\ \hline
    \end{tabularx}
    \caption{Comparison of different world model-based methods.}
    \label{tab:worldmodel}
\end{table}

In this section, we discuss various kinds of world models
that have been proposed in the literature.
These models can be trained to predict future observations (generative WMs)
or just future rewards/values and/or future latent embeddings (non-generative / non-reconsructive WMs).
Once trained, the models
can be used for decision-time planning,
background planning, or just as an auxiliary signal to aid in things
like intrinsic curiosity.
See \cref{tab:worldmodel} for a summary.

\subsection{World models which are trained to predict observation targets}

In this section, we discuss different kinds of world model
$\obsdynamics(\vs'|\vs,a)$.
We can use this to 
generate imaginary trajectories 
by sampling from the following joint distribution:
\be
p(\vs_{t+1:T},\vr_{t+1:T},\va_{t:T-1}|\vs_t)=
\prod_{i=t}^{T-1}
\pi(\va_i|\vs_i)
\obsdynamics(\vs_{i+1}|\vs_{i}, \va_i)
R(r_{i+1}|\vs_{i}, \va_i)
\label{eqn:unroll}
\ee
The  model may be augmented with latent variables,
as we discuss in \cref{sec:latentWM}.

If the state space is high dimensional (e.g., images),
then we denote the observable data by $\vo$.
We can then learn
 $\obsdynamics(\vo'|\vo,a)$
using standard  techniques  for conditional image generation
such as  diffusion (see e.g.,
the \keywordDef{Diamond} method of \citep{Alonso2024},
the \keywordDef{Genie2} method of \citep{genie2},
the \keywordDef{GAIA-1} model of \citep{Hu2023},
etc.
This kind of world model
is equivalent to  an action-conditional
version of a \keywordDef{video generative model},
such as \keywordDef{Sora},
\keywordDef{Veo-3},
\keywordDef{seedance} \citep{Gao2025seedance},
etc.

Note that these methods are trained to predict
the entire observation vector, even if we use latent variables.
(This is what we mean by ``generative world model''.)
One big disadvantage of this approach
is that the observations may contain irrelevant or distractor
variables, that are not necessary for task performance.
In addition, such models are often slow to use slow,
and there may be a distribution
shift in th observation process between train and test time.
Both of these factors can adversely affect the performance
of generative WMs
(see e.g., \citep{Tomar2023}).
We discuss some non-generative approaches to WMs in \cref{sec:nongen}.

\subsubsection{Generative world models without latent variables}

The simplest approach is to define $\obsdynamics(\vo'|\vo,a)$ as a conditional
generative model over states.
If the observed states are low-dimensional vectors,
such as proprioceptive states,
we can use transformers
(see e.g., the \keywordDef{Transformer Dynamics Model}
of \citep{Schubert2023}).

In some cases, the dimensions of the state vector $\vs$ represent
distinct
variables, and the joint Markov transition matrix $p(\vs'|\vs,a)$
has  conditional independence properties
which 
can be represented as a sparse graphical model,
This is called
a \keywordDef{factored MDP}
\citep{Boutilier2000}.

\subsubsection{Generative world models with latent variables}
\label{sec:MBRLlatent}
\label{sec:latentWM}

In this section, we describe some methods
that use latent variables as part of their world model.
This can improve the speed of generating imaginary futures,
and can provide a compact latent space as input to a policy.

We let $\vz_t$ denote the latent (or hidden) state at time $t$;
this can be a discrete or continuous variable (or vector
of variables).
The generative model has the form of a controlled HMM:
\be
p(\vo_{t+1:T}, \vz_{t+1:T},\vr_{t+1:T},\va_{t:T-1}|\vz_t) =
\prod_{i=t}^{T-1}
  \pi(\va_i | \vz_{i})
\latentdynamics(\vz_{i+1}|\vz_{i}, \va_i)
R(r_{i}|\vz_{i+1}, \va_i)
D(\vo_{i}|\vz_{i})
\label{eqn:unrollLatent}
\ee
where $p(\vo_t|\vz_t)=D(\vo_t|\vz_t)$ is the decoder
or likelihood function,
$\latentdynamics(\vz'|\vz,\va)$ is the dunamics in latent space.
$\pi(\va_t|\vz_t)$ is the policy in latent space.

The world model is usually trained by maximizing the
marginal likelihood
of the observed outputs given an action sequence.
(We discuss non-likelihood based loss functions
in \cref{sec:nongen}.)
Computing the marginal likelihood
 requires marginalizing over the hidden variables
$\vz_{t+1:T}$. To make this computationally tractable,
it is common to use amortized variational inference,
in which we
train an encoder network,
$p(\vz_t|\vo_t)=E(\vz_t|\vo_t)$, to approximate the posterior
over the latents.
Many papers have followed this basic approach,
such as 
the \keywordDef{``world models''} paper  \citep{worldModels},
and the methods we discuss below.

\subsubsection{Example: Dreamer}
\label{sec:dreamer}

In this section, we summarize the approach used in
\keywordDef{Dreamer} paper \citep{dreamer}
and its recent extensions,
such as \keyword{DreamerV2} \citep{dreamerv2},
 \keyword{DreamerV3} \citep{dreamerv3}
and \keyword{DreamerV4} \citep{dreamerv4}.
These are all based on the background planning approach,
in which the policy is trained on imaginary
trajectories generated by a latent variable world model.
(Note that Dreamer is based on an earlier approach
called \keywordDef{PlaNet}  \citep{planet},
which used MPC instead of background planning.)

\begin{figure}
\centering
\includegraphics[height=4in]{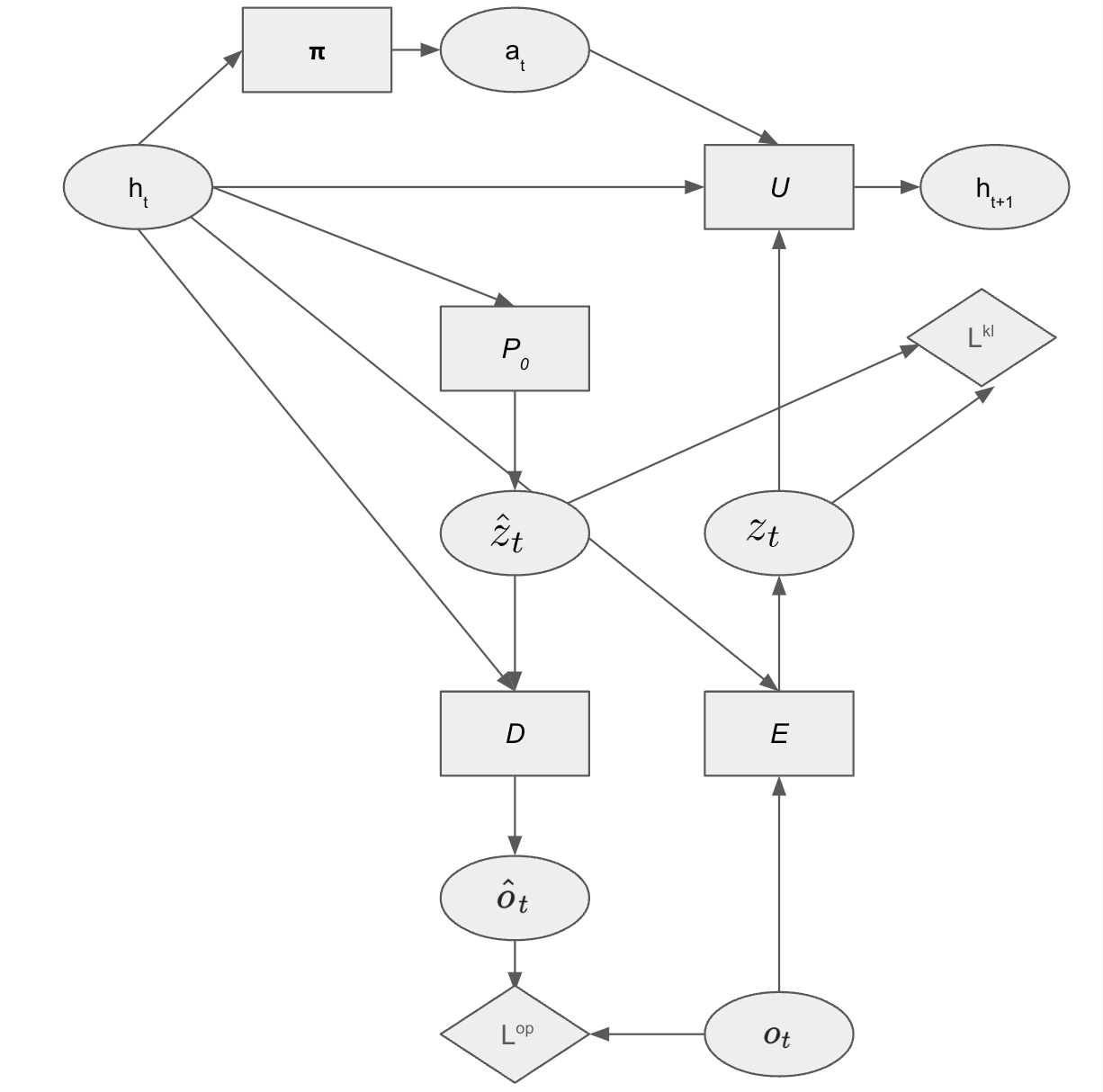}
\caption{
  Illustration of  Dreamer world model as a factor graph
  (squares are learnable functions, circles are variables, diamonds are fixed cost functions).
  We have unrolled the forwards prediction for only 1 step.
  Also, we have omitted the reward prediction loss.
}
\label{fig:dreamer}
\end{figure}

In Dreamer, the
stochastic dynamic latent variables
in \cref{eqn:unrollLatent}
are replaced by deterministic dynamic latent
variables $\vh_t$,
since this makes the model easier to train.
(We will see that  $\vh_t$ acts like the posterior
over the hidden state at time $t-1$;
this is also the  prior predictive belief state
before we see $\vo_t$.)
A static stochastic variable $\latent_t$ is now generated
for each time step, and acts like a ``random effect''
in order to help generate the observations,
without relying on $\vh_t$ to store all of the necessary information.
(This simplifies the recurrent latent state.)
In more detail, Dreamer defines the following functions:\footnote{
We can map from our notation to the notation in the paper
as follows:
  $\vo_t \ra x_t$, 
  $U \ra f_{\phi}$ (sequence model),
  $P_0 \ra p_{\phi}(\hat{z}_t|h_t)$ (dynamics predictor),
  $D \ra p_{\phi}(\hat{x}_t|h_t, \hat{z}_t)$  (decoder),
  $E \ra  q_{\phi}(\latent_t|h_t,x_t)$ (encoder).
}
\begin{itemize}
\item A hidden dynamics (sequence) model:  $\vh_{t+1}= \update(\vh_{t}, \va_{t}, \latent_t)$
 \item A latent state conditional prior: $\hat{\latent}_t \sim P(\hat{\latent}_t|\vh_{t})$ 
 \item A latent state decoder (observation predictor):
   $\hat{\vo}_t \sim D(\hat{\vo}_t|\vh_t,\hat{\latent}_t)$.
   \item A reward predictor: $\hat{r}_t \sim R(\hat{r}_t|\vh_t, \hat{\latent}_t)$
     \item A latent state encoder:  $\latent_t \sim E(\latent_t|\vh_{t}, \vo_t)$.
  \item A policy function: $\va_t \sim \pi(\va_t|\vh_t)$
\end{itemize}
See \cref{fig:dreamer} for an illustration of the system.

We now give a simplified explanation of how the world model is trained.
The loss has the form
\be
\loss^{\text{WM}} = \expectQ{
  \sum_{t=1}^T \beta_{o} \loss^o(\vo_t, \hat{\vo}_t) 
  + \beta_{z} \loss^z(\latent_t, \hat{\latent}_t)
  }{q(\latent_{1:T})}
\ee
where the $\beta$ terms are different weights for each loss,
and $q$ is the posterior over the latents, given by
\be
q(\latent_{1:T}|\vh_{0},\vo_{1:T},\va_{1:T})
 = \prod_{t=1}^T E(\latent_t|\vh_t,\vo_t) \delta(\vh_t-\update(\vh_{t-1},\va_{t-1},\latent_{t-1}))
 \ee
 The loss terms  correspond to the observation prediction cross entropy,
 and the posterior to prior KL penalty:
\begin{align}
  \loss^o &= -\ln D(\vo_t|\latent_t,\vh_t)\\
  \loss^z &= \KLpq{E(\latent_t|\vh_t,\vo_t))}{P(\latent_t|\vh_t)}) 
\end{align}
where we abuse notation somewhat,
since $\loss^z$ is a function of two distributions,
not of the variables $\latent_t$ and $\hat{\latent}_t$.

In addition to the world model loss, we have the following
actor-critic losses
\begin{align}
  \loss^{\text{critic}}
  &= \sum_{t=1}^T (V(\vh_t) - \stopgrad(G_t^{\lambda}))^2 \\
  \loss^{\text{actor}}
   &= -\sum_{t=1}^T \stopgrad((G_t^{\lambda}-V(\vh_t))) \log \pi(\va_t|\vh_t)
  \end{align}
where $G_t^{\lambda}$ is the GAE estimate of the reward to go:
\be
G_t^{\lambda} = r_t + \gamma\left( (1-\lambda) V(\vh_t) + \lambda G_{t+1}^{\lambda} \right) 
\ee

\eat{
\newcommand{\tpred}{\text{pred}}
\newcommand{\tdyn}{\text{dyn}}
\newcommand{\trep}{\text{rep}}

The world model is trained to minimize the following loss:
\be
\loss = \expectQ{
  \sum_{t=1}^T \beta_{\tpred} \loss_{\tpred}
  + \beta_{\tdyn} \loss_{\tdyn}
  + \beta_{\trep} \loss_{\trep}
  }{q(\vz_{1:T}|\vh_{0:T},\vo_{1:T})}
\ee
where the $\beta$ terms are different weights for each loss.
The loss terms are defined as follows:
\begin{align}
  \loss_{\tpred} &= -\ln D(\vo_t|\vz_t,\vh_t)
  - \ln R(r_t|\vz_t,\vh_t) \\
  \loss_{\tdyn} &= \max(1, \KLpq{\stopgrad(E(\vz_t|\vh_t,\vo_t))}{P(\vz_t|\vh_t)}) \\
  \loss_{\trep} &= \max(1, \KLpq{E(\vz_t|\vh_t,\vo_t)}{\stopgrad(P(\vz_t|\vh_t))})
\end{align}
where $\stopgrad$ is the stop gradient operator.
Here $\loss_{\tpred}$ is the observation prediction loss;
$\loss_{\tdyn}$ is the dynamics loss, that trains the sequence
model to predict the (frozen) latent encoding;
and $\loss_{\trep}$ is the representation loss,
that trains the latent encoder to be predictable
by the (frozen) dynamics model.
The use of the $\max(1,KL)$ expression is to ensure
that the KL loss does not have to go all the way to 0,
which can cause latent variable collapse.
the value of 1 nat (about 1.44 bits) corresponds
to the \keywordDef{free bits} of the encoding
\citep{kingma2016improving}.
}

There have been several extensions to the original Dreamer paper.
\keywordDef{DreamerV2} \citep{dreamerv2}
adds categorical (discrete) latents and  KL balancing between prior
and posterior estimates.  
This was the
first imagination-based agent to outperform humans in Atari games.
\keywordDef{DayDreamer} \citep{dayDreamer}
applies DreamerV2 to real robots.
\keywordDef{DreamerV3} \citep{dreamerv3}
builds upon DreamerV2 using various tricks --- such as symlog
encodings\footnote{
The symlog function is defined as
$\text{symlog}(x) = \text{sign}(x)\ln(|x|+1)$,
and its inverse is
$\text{symexp}(x) = \text{sign}(x)(\exp(|x|)-1)$.
The symlog function squashes large positive and negative values,
while preserving small values.
} %
for the reward, critic, and decoder ---
to enable more stable optimization and domain independent
choice of hyper-parameters.
It was the first method to create diamonds in the Minecraft
game without requiring human demonstration data.
(However, reaching this goal took 17 days of training.)

Many other variants of Dreamer have been explored.
For example,  TransDreamer \citep{transDreamer}
and STORM \citep{storm}   replace the RNN world model with transformers,
and the S4WM method of \citep{Deng2023} uses S4 
(Structured State Space Sequence) models.
The  \keyword{DreamingV2} paper of \citep{dreamingV2}
replaces the generative loss
with a non-generative self-prediction loss (see \cref{sec:SPR}),
and \citep{Robine2023,SGF} use the VicReg non-generative representation
learning method
(see \cref{sec:JEPA}).

The most recent variant is
\keywordDef{DreamerV4} \citep{dreamerv4}.
Here 
the world model is a standard conditional
latent video diffusion model, $p(z_t|z_{t-c:t-1},a_t)$,
for context length $c$,
based on a transformer backbone,
combined with an autoenencoder
between pixel and latent space using
$q(z_t|y_t)$ and $p(y_t|z_t)$.
The WM is trained offline on 2500 hours of
reward-free but
action-labeled videos
(keyboard-mouse actions)
of people playing minecraft,
derived from OpenAI's VPT dataset \citep{VPT}.
After training the WM, a policy is initialized
with behavior cloning,
and then fine tuned
in imagination
using actor-critic methods,
with minecraft-specific reward
functions (learned from the VPT metadata).

The key difference from prior Dreamer models
is that the world model is more powerful,
and it is trained offline on diverse human-collected  data.
The downside is that the world model is slower to sample from.\footnote{
The reason is that diffusion is an iterative procedure.
To speed it up,
DreamerV4 uses the ``shortcut'' trick from
\citep{Frans2024}
to reduce the number of diffusion sampling steps
to 1-2.
(This is a special case of
the self-consistency training methods
described in \citep{Boffi2025}.)
}
The upside it is that the model is sufficiently expressive,
and the training data  is sufficiently large and diverse,
that the world model is accurate enough (in terms of dynamics
and visual details) that it can be used to train
policies purely in imagination,
without mixing in any real world data.
The resulting DreamerV4 recipe is similar in spirit
to LLMs, that are large autoregressive transformer
models that are pretrained  on large diverse datasets,
and then used to train policies on synthetic rollouts
using RL. (In the LLM case, the policy and WM are the same model.)

\subsubsection{Example: IRIS}
\label{sec:iris}

The \keywordDef{IRIS} method (``Imagination with auto-Regression
over an Inner Speech'')
of \citep{iris} follows the MBRL paradigm,
in which it alternates beween
(1) learning a world model
using real data $D_r$
and then  generate imaginary rollouts $D_i$ using the WM,
and (2)
learning the policy given $D_i$
and collecting new data $D_r'$ for learning.
In the model learning stage,
Iris learns a discrete latent
encoding using the VQ-VAE method,
and then  fits a transformer dynamics
model to the latent codes.
In the policy learning stage, it uses actor critic methods.
The \keywordDef{Delta-IRIS}
method of \citep{delta-iris}
extends this by training the model to only predict
the delta between neighboring frames.
Note that, in both cases, the policy has the form $a_t=\pi(\vo_t)$,
where $\vo_t$ is an image, so the
the rollouts need to ground to pixel space, and cannot only
be done in latent space.

\subsubsection{Code world models}

Recently it has become popular to represent the world model $p(s'|s,a)$
using code, such as Python. This is called a \keywordDef{code world model}.
It is possible to learn such models from trajectory data using LLMs.
See \cref{sec:CWM} for details.

\subsubsection{Partial observation prediction}

Predicting all the pixels in image may waste capacity
and may distract the agent from the important bits.
A natural alternative is to just predict some function of the observations,
rather than the entire observatiin vector.
This is known as a \keywordDef{partial world model}
(see e.g., \citep{Alver2023,Talvitie2011}).
One way to implement this is to impose an  information bottleneck
between the latent state and the observed variables,
to prevent the agent
focusing on irrelevant observational details
(see e.g., the \keywordDef{denoised MDP} method
of \citep{Wang2022}).
We can also use code synthesis methods
(see \cref{sec:CWMpartial}).

\subsection{World models which are trained to predict other targets}
\label{sec:nongen}
\label{sec:MBRLnongen}

In this section, we discuss training world models that are not necessarily
able to predict all the future observations.
These are often still (conditional) generative models (in that they return a distribution over
potentially high dimensional outputs), but they are {\em lossy} models,
because they do not capture all the details of the data.

\subsubsection{The objective mismatch problem}
\label{sec:mismatch}

In \cref{sec:MBRLgame}, we argued that, if we can learn a sufficiently accurate world
model, then solving for the optimal policy in simulation will give
a policy that is close to optimal in the real world.
However, a simple agent may not be able to capture the full complexity
of the true environment;
this is called the ``\keywordDef{small agent, big world}'' problem
\citep{Dong2022,bitByBit,Arumugam2024,Kumar2024}.

Consider what happens when the agent's model is misspecified (i.e., it cannot
represent the true world model), which is nearly always the case.
The agent will train its model
to reduce state (or observation) prediction error,
by minimizing $\lossfn(\hat{M}, \mu_M^{\pi})$.
However, not all features of the state are useful for planning.
For example, if the states are images, a dynamics model
with limited representational capacity may choose
to focus on predicting the background pixels
rather than more control-relevant features, like small moving
objects, since predicting the background reliably
reduces the MSE more.
This is due to ``\keywordDef{objective mismatch}''
\citep{Lambert2020RL,Wei2023MBRL},
which refers to the discrepancy between the way a
model is usually trained (to predict the observations)
vs the way its representation is used for control.
To tackle this problem, in this section we discuss methods
for learning representations and models that don't rely on
predicting all the observations.
Our presentation is based in part
on   \citep{Ni2024} (which in turn builds on
\citep{Subramanian2022}).
See \cref{tab:WM} for a summary of some of the methods
we will discuss.

\begin{table}
  \centering
  \begin{tabular}{llll}
    Loss & Policy & Usage & Examples \\ \hline
    OP & Observables & Dyna & Diamond \citep{Alonso2024}, Delta-Iris \citep{delta-iris} \\
    OP & Observables & MCTS & TDM  \citep{Schubert2023} \\
    OP & Latents & Dyna & Dreamer \citep{dreamerv3} \\
    RP, VP, PP & Latents & MCTS & MuZero \citep{Schrittwieser2020} \\
    RP, VP, PP, ZP & Latents & MCTS & EfficientZero  \citep{efficientZero} \\
    RP, VP, ZP  & Latents & MPC-CEM & TD-MPC \citep{Hansen2022,Hansen2024} \\
    VP, ZP  & Latents & Aux. &  Minimalist \citep{Ni2024} \\
    VP, ZP & Latents & Dyna & DreamingV2  \citep{dreamingV2} \\
    VP, ZP, OP & Latents & Dyna & AIS  \citep{Subramanian2022}\\
    POP & Latents & Dyna & Denoised MDP  \citep{Wang2022}
  \end{tabular}
  \caption{Summary of some world-modeling methods.
    The ``loss'' column refers to the loss used to train
    the latent encoder (if present) and the dynamics model
    (OP = observation prediction, ZP = latent state prediction,
    RP = reward prediction, VP = value prediction,
    PP = policy prediction, POP = partial observation prediction).
    The ``policy'' column refers to the input that is passed
    to the policy.
    (For MCTS methods, the policy is just used as a proposal
    over action sequences to initialize the search/ optimization process.)
    The ``usage'' column refers to how to the world model is used:
        for  background planning (which we call ``Dyna''),
        or for decision-time planning (which we call ``MCTS''),
        or just as an auxiliary loss on top of standard policy/value
        learning (which we call ``Aux'').
        Thus Aux methods are single-stage (``end-to-end''), whereas the other
        methods alternate are two-phase, and alternate between improving the world model
    and then using it for improving the policy (or searching for the optimal action).
    }
  \label{tab:WM}
  \end{table}

\subsubsection{Observation prediction}

We consider a modeling paradigm where we learn an encoder,
$\vz=\phi(\vo)$\footnote{
Note that in general, the encoder may depend
on the entire history of previous observations, denoted $\vz=\phi(\data)$.
};
a dynamics model in latent space,
$\vz'=\latentdynamics(\vz,\va)$, for future prediction;
and an update model in latent space,
$\vz'=\update(\vz,a,\vo)$,
for belief state tracking.

A natural target to use for learning the encoder
and dynamics is the next observation,
using a one-step version of \cref{eqn:unroll}.
Indeed, \citep{Ni2024} say that a representation $\phi$ satisifies
the \keywordDef{OP} (observation prediction)
criterion if it satisfies the following condition:
\begin{align}
  \exists D \myst  p^*(\vo'|\hist,a) = D(\vo'|\phi(\hist),a)
\;\; \forall \hist,a
\end{align}
where $D$ is the decoder.
In order to repeatedly apply this, we need to be able
to update the encoding $\vz=\phi(\hist)$ in a recursive
or online way. Thus we must also satisfify
the following recurrent encoder condition,
which \citep{Ni2024}
call \keywordDef{Rec}:
\begin{align}
  \exists U \myst  \phi(\hist') = \update(\phi(\hist), a, \vo')
\;\; \forall \hist,a,\vo'
\end{align}
where $\update$ is the update operator.
Note that belief state updates (as in a POMDP)
satisfy this property.
Furthermore, belief states are a sufficient statistic
to satisfy the OP condition.
See \cref{sec:MBRLlatent} for a discussion of generative models
of this form.

The drawback of this approach is that in general it is very hard
to predict future observations, at least in high dimensional settings like images.
Fortunately,  such prediction is not necessary for optimal behavior.
Thus we now turn our attention to other training objectives.

\subsubsection{Reward prediction}

\begin{figure}
\centering
\includegraphics[height=2.5in]{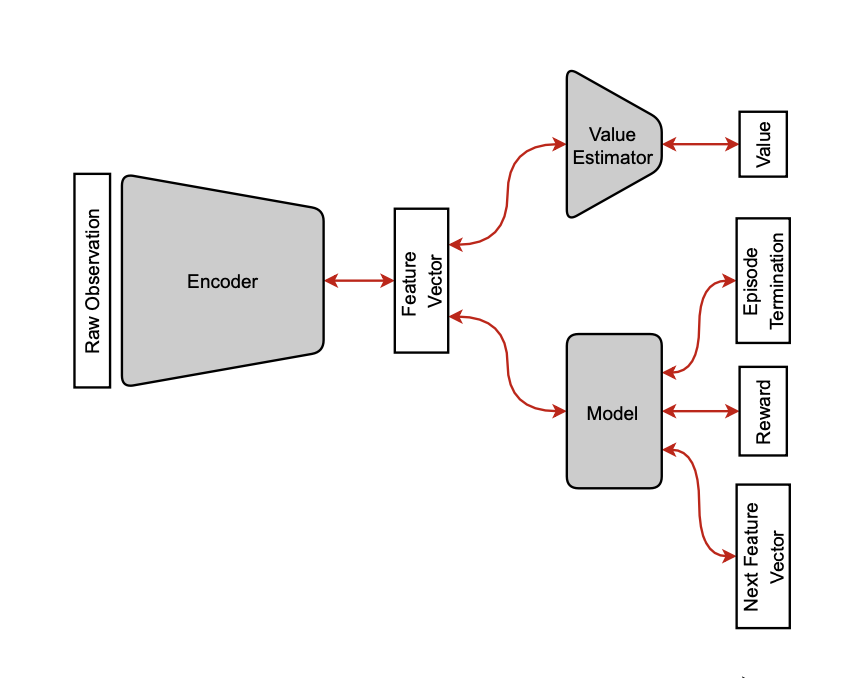}
\caption{
  Illustration of  an encoder $\vz_t=E(\vo_t)$,
  which is passed to a value estimator $v_t=V(\vz_t)$,
  and a world model, which predicts
  the next latent state $\hat{\vz}_{t+1}=\latentdynamics(\vz_t,a_t)$,
  the reward $r_t=R(\vz_t,a_t)$,
  and the termination (done) flag, $d_t=\done(\vz_t)$.
  \figtaken{Figure C.2 of \citep{Alver2023}.}
\figthanks{Doina Precup}.
}
\label{fig:MBRLtargets}
\end{figure}

We can also train the latent encoder to predict the reward.
Formally, we want to ensure we can satisfy the following condition,
which we call \keywordDef{RP} for ``reward prediction'':
\begin{align}
  \exists R \myst  \expectQ{r|\hist,a}{R^*} &=
  \expectQ{r|\phi(\hist),a)}{R} \;\; \forall \hist,a
\end{align}
See \cref{fig:MBRLtargets} for an illustration.
In \citep{Ni2024}, they prove that a representation
that satisfies ZP and RP is enough to satisfy value equivalence
(sufficiency for $Q^*$).

\subsubsection{Value prediction}
\label{sec:valueEquivalence}

Let $\hist_t =(\hist_{t-1},\va_{t-1},r_{t-1},\vo_t)$ be all
the visible data (history) at time $t$,
and let $\vz_t = \phi(\hist_t)$ be a latent representation (compressed encoding)
of this history, where $\phi$ is called an encoder or a \keywordDef{state abstraction}
function. We will train the policy $\va_t=\pi(\vz_t)$ in the usual way,
so our focus will be on how to learn good latent representations.

An optimal representation $\vz_t=\phi(\hist_t)$ is a sufficient statistic
for the optimal action-value function $Q^*$.
Thus it satifies the \keywordDef{value equivalence} principle
\citep{Li2006,Castro2011,Grimm2020,Grim2021,Grimm2022,Alver2023,Alver2024},
which says  that two states $s_1$ and $s_2$ are  value equivalent
(given a policy)
if $V^{\pi}(s_1) = V^{\pi}(s_2)$.
In particular, if the representation is optimal,
it will satisfy value equivalence wrt the optimal policy,
i.e.,
if $\phi(\hist_i)=\phi(\hist_j)$ then $Q^*(\hist_i,a)=Q^*(\hist_j,a)$.
We can train such a representation function by using
its output $\vz=\phi(\hist)$ as input to the Q function
or to the policy.
(We call such a loss \keywordDef{VP}, for value prediction.)
This will cause the model to focus
its representational power on the relevant parts of the observation
history.

Note that there is a stronger property than value equivalence
called \keywordDef{bisimulation}
\citep{Givan2003}.
This says that two states $s_1$ and $s_2$ are bisimiliar if
$P(s'|s_1,a) \approx P(s'|s_2,a)$ and $R(s_1,a) = R(s_2,a)$.
From this, we can derive a continuous measure called the
\keywordDef{bisimulation metric} \citep{Ferns2004}.
This has the advantage (compared to value equivalence)
of being policy independent, but the disadvantage that it can be 
harder to compute  \citep{Castro2020mdp,Zhang2021},
although there has been recent progress on computaitonally
efficient methods such as MICo \citep{Castro2021}
and KSMe \citep{Castro2023}.

\subsubsection{Policy prediction}

The value function and reward losses may be too sparse
to learn efficiently. Although self-prediction loss can help somewhat,
it does not use any extra information from the environment as feedback.
Consequently it is natural to consider other kinds of prediction
targets for learning the latent encoder (and dynamics).
When using MCTS, it is possible compute what the policy should
be for a given state, and this can be used as a prediction
target for the reactive policy $a_t=\pi(\vz_t)$,
which in turn can be used as a feedback signal for the latent
state. This method is used by MuZero (\cref{sec:muZero})
and EfficientZero (\cref{sec:efficientZero}).

\subsubsection{Self prediction (self distillation)}
\label{sec:SPR}
\label{sec:self-predictive}

Unfortunately, in problems with sparse reward,
predicting the value or policy may  not provide enough of a feedback signal
to learn quickly. Consequently it is common to augment
the training with a \keywordDef{self-prediction} loss
where we train $\phi$ to ensure
the following condition hold:
\begin{align}
  \exists M \myst  \expectQ{\vz'|\hist,a}{M^*} &=
  \expectQ{\vz'|\phi(\hist),a)}{M} \;\; \forall \hist,a
\end{align}
where the LHS is the predicted mean
of the next latent state under the true model,
and the RHS is the predicted mean under the learned
dynamics model.
We call this the \keywordDef{EZP}, which stands for expected $\vz$ prediction.\footnote{
In \citep{Ni2024}, they also describe the ZP loss,
which requires predicting the  full distribution over $\vz'$
using a stochastic transition model.
This is strictly  more powerful, but somewhat more complicated,
so we omit it for simplicity.
}

\subsubsection{Avoiding self-prediction collapse using frozen targets}
\label{sec:collapseFrozen}


A trivial way to minimize the self-prediction loss is to learn
an embedding that
 maps everything to a constant vector, say $E(\hist)=\vzero$,
in which case $\vz_{t+1}$ will be trivial for  the
dynamics model $M$ to predict.
However this is not a useful representation.
This problem is \keywordDef{representational collapse}
\citep{Jing2022collapse}.
Fortunately,
we can provably prevent collapse (at least for linear encoders)
by using a frozen target network 
\citep{Tian2021,Tang2023,Ni2024}.
That is, we use the following auxiliary
loss
\be
\loss_{\text{EZP}}(\vphi,\vtheta;\hist,a,\hist')
= ||M_{\vtheta}(E_{\vphi}(\hist,a)) -
\stopgrad(E_{\overline{\vphi}}(\hist'))||_2^2
\label{eqn:EZP}
\ee
where
\be
\overline{\vphi}_t = \rho \vphi_{t} + (1-\rho) \overline{\vphi}_{t-1}
\ee
is the exponential moving average
(EMA) of the encoder weights $\vphi$.
(If we use a frozen (old) copy of the weights instead,
this is called a detached network.)
See \cref{fig:JEPA}(a) for an illustration.
This approach means the ``goalposts'' (the target representations)
evolve slowly and consistently over time, guided by the progress of the
encoder and predictor.
This adds stability to the training process,
ensuring the target representations don't change erratically
from one step to the next, which would make the predictor's job impossible.

The above approach is used in many papers,
such as 
\keywordDef{BYOL} \citep{BYOL}
(BYOL stands for ``bootstrap your own latent''),
\keywordDef{SimSiam} \citep{SimSiam},
\keywordDef{DinoV2} \citep{dinov2},
\keywordDef{JEPA} (Joint embedding Prediction Architecture)
\citep{LeCun2022Path},
\keywordDef{I-JEPA} \citep{IJEPA},
\keywordDef{V-JEPA} \citep{VJEPA,VJEPA2},
\keywordDef{Image World Models} \citep{Garrido2024},
\keywordDef{Predictron}  \citep{predictron},
\keywordDef{Value Prediction Networks} \citep{Oh2017},
\keywordDef{Self Predictive Representations} (SPR)
\citep{Schwarzer2021},
\keywordDef{Efficient Zero} (\cref{sec:efficientZero}),
etc.

\begin{figure}
  \centering
    \includegraphics[height=2.5in]{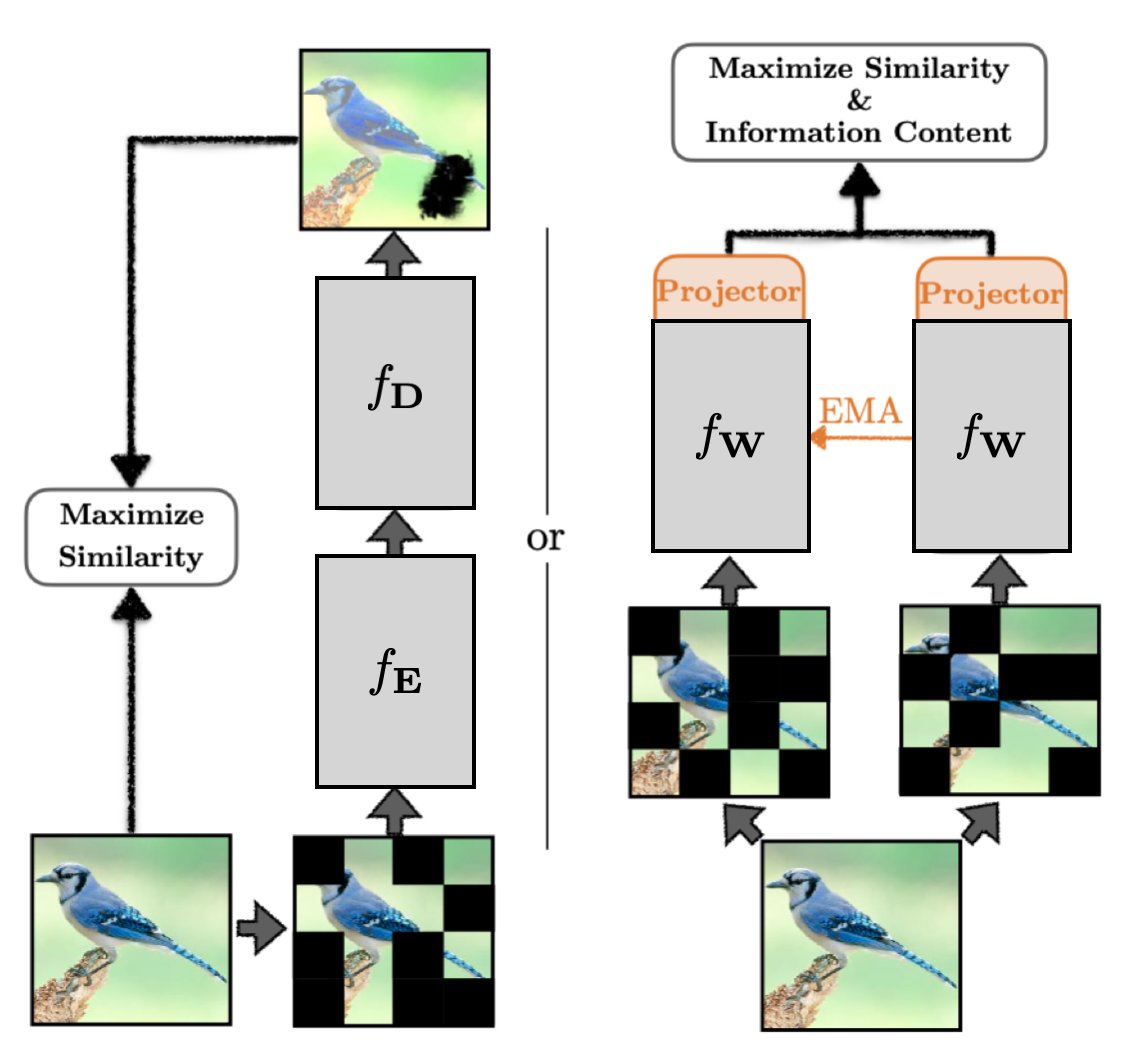}
    \caption{
      Self-supervised learning methods.
      Left: masked auto-encoder.
      Right: self-prediction.
      Figure from \url{https://x.com/hugues_va/status/1991635221884891373?s=58}m
      based on \citep{Van-Assel2025}.
}
\label{fig:JEPA-SSL}
\end{figure}

Note that an alternative to predicting the encoding of the next frame
is to mask out the current frame in a random way,
and then use this as the target.
See \cref{fig:JEPA-SSL} for an illustration.
For more details on such \keywordDef{self-supervised learning}
or \keywordDef{SSL} methods,
see e.g., \citep{ShwartzZiv2024,Uelwer2025,Van-Assel2025}.

Minimizing the self-prediction objective (with the stop-gradient term)
has  been proven to be theoretically sound for the case
of linear encoders \citep{Tang2023} with a fixed policy,
which they call BYOL-$\pi$.
in this case,  the encoder converges
learns the singular vectors of the transition matrix induced
by the policy.
In  \citep{Khetarpal2025}, they present BYOL-AC,
which learns the singular vectors of the transition matrix for each
(discrete) action, which makes the method independent of the policy.
The analysis methods in these papers
create an ODE where 
the encoder learns more slowly (using gradient steps)
than the latent dynamics (which are solved to optimality at each inner step).

Proving convergence in the general nonlinear case is an open problem.
It is possible that by adopting a game theoretic perspective,
one can show show that solving
such an asymmetric Stackleberg game
(see \cref{sec:stackelberg}) 
can converge  to an  equilibrium,
even if it does not correspond to
the minimum of a single loss function.
In particular,
suppose the target encoder player (which acts as the slow-moving leader)
makes its move by updating the EMA of the encoder network;
then  the prediction player (which acts as the fast-moving follower)
makes it move by minimizing the  EZP prediction error in \cref{eqn:EZP}
using the leader's encoder.
Then it may be possible to modify the results from papers such as
\citep{Bai2021,Gerstgrasser2023,Li2024ssl}
to show that self-predictive learning for world models will converge
in the general case.

\subsubsection{Avoiding self-prediction collapse using information-theoretic regularization}
\label{sec:collapseReg}

An alternative way to avoid the latent collapse problem
is to add regularization terms that try to maximize
the information content in $\vz_t$ and $\vz_{t+1}$
(see \cref{fig:JEPA}(b)),
while also minimizing the prediction error.
That is, the objective becomes
\begin{align}
  J(\vphi) &=   E_{\vo_t,\va_t,\vo_{t+1},\vepsilon_t}
  \left(
 ||\vz_{t+1} - \hat{\vz}_{t+1}||_2^2
 - \lambda I(\vz_t)   - \lambda I(\vz_{t+1})
 \right) \nonumber \\
&  \text{ where }
  \vz_t = E(\vo_t;\vphi),
  \vz_{t+1} = E(\vo_{t+1};\vphi),
  \hat{\vz}_{t+1}=\latentdynamics(\vz_t, \va_t, \vepsilon_t;\vtheta)
  \label{eqn:JEPAinfo}
  \end{align}
(Note that $\vz_{t+1}$ may correspond to the embedding
of the next frame, $\vo_{t+1}$, or to the embedding of a data augmented
version of $\vo_t$ (e.g., a cropped or rotated version).

Various methods have been proposed to approximate
the information content $I(\vz_t)$,
mostly based on some function
of the outer product matrix $\sum_t \vz_t \vz_t^\trans$,
which captures second order moments.

\label{sec:VICReg}
The \keywordDef{Barlow Twins} \citep{barlowTwins} method
aims to make the cross-correlation matrix between the representations
of two embeddings as close to the identity matrix as possible.
This simultaneously encourages similar representations for similar inputs
while decorrelating the features within the representation.
The \keywordDef{VICReg} \citep{VICReg,VICRegL} method
(Variance-Invariance-Covariance Regularization)
uses three loss terms:
one to maintain variance in the representations (avoid collapse),
 one to decorrelate the different variables in the latent vector (reduce redundancy),
 and one to make them invariant to data augmentations
 (by bringing different views closer together in emebdding space).
 The VICReg approach can be thought of as constrastive across dimensions,
 whereas standard contrastive methods, such as SimCLR,
 are contrastive across samples (see \citep{Garrido2023}).
%
Unfortunately, the methods  in these papers do not provide
a lower bound on $I(\vz_t)$ and $I(\vz_{t+1})$,
which is needed to optimize \cref{eqn:JEPAinfo}.
(Since we are trying to minimize $-I$,
we need an upper bound on $-I$,
but these methods actually provide an upper bound on $I$.)

Note, however, that it is possible to use information theoretic
regularizers when training generative models, that predict
future observables. The idea is to create an \keywordDef{information bottleneck},
that remembers as little about the inputs (past) as possible,
while still being able to predict the future.
See e.g., \citep{Lee2020VIB,Dave2024}.

\subsubsection{Preventing self-prediction collapse using game-theoretic approaches}
\label{sec:CTRL}

\eat{
An alternative appproach is to modify the objective
to use the change in the latent rate
rather than the absolute information content;
this can be tractably approximated, as shown in 
\keywordDef{MCR2} (Maximal Coding Rate Reduction) \citep{YuMCR2}
and its closed-loop extension \citep{Dai2022}.
}

An arguably more theoretically sound approach to self-supervised learning,
known as \keywordDef{CTRL}
(closed-loop transcription) presented in \citep{Dai2022},
tries to solve a minmax game (similar to a GAN), rather than optimize a
(regularized) objective.
There are two players: the encoder $E$,
and the generator or decoder $G$.
The generator's job is to minimize distortion.
But rather than measuring this distance in pixel space,
it is evaluated in embedding space:
\be
D(\vZ,\vZ') = ||\vZ - \vZ'||_F^2
\ee
where $\vZ=E(\vX)$ is the embedding of a batch of inputs $\vX$
and
\be
\vZ'=E(G(\vZ))=E(G(E(\vX)))
\ee
is the embedding of their reconstructions.
A good encoder should  also  minimize  the distortion.

However, to prevent the learning a trivial embedding,
the encoder should also try to maximize its own
\keywordDef{rate} (information content).
In CTRL, they assume the embeddings are Gaussianly distributed,
which gives the following tractable
equation for the rate:
\be
R(\vZ) = \log \det(C(\vZ))
\ee
where $C(\vZ)=\vZ^\trans \vZ$ is the covariance matrix of the data
(assuming it is centered, for simplicity).\footnote{
In practice, computing the covariance matrix of the entire dataset,
and the corresponding log determininant,
is not tractable.
We can handle a large number of samples
by computing the rate over a minibatch of size $n$
using $C(\vZ_b) = \frac{1}{n} (\vZ_b - \overline{\vZ}_b)^\trans
(\vZ_b - \overline{\vZ}_b)$.
We then use matrix determinant lemma to replace
the objective with
$R(\vZ_b)= \log \det(\vZ_b \vZ_b^\trans)$,
which takes $O(n^3)$ time to compute rather than $O(d^3)$.
In practice, we add a regularizer for numerical stability, and use
$R(\vZ_b)= \log \det(\vI_n + \frac{d}{n \epsilon^2} \vZ_b \vZ_b^\trans)$.
}
Minimizing this single term  achieves two goals at once:
Decorrelation,
since it forces the off-diagonal elements of the covariance matrix to be zero,
meaning the learned features are independent and not redundant;
and
variance preservation,
since it encourages the diagonal elements (the variance of each feature)
to be non-zero and balanced, preventing the representation from collapsing
into a lower-dimensional space.
This is very similar in spirit to VICReg's separate variance and covariance terms,
but combined into one expression.

We can thus define the CTRL game as follows:
\begin{align}
  \min_G D(\vZ,\vZ';E,G)  \\
  \max_E R(\vZ;E) - \lambda D(\vZ,\vZ'; E, G)
\end{align}
Since the rate only depends on the encoder,
there is an imbalance in the game.
Hence,
rather than just having the encoder maximizing the rate of the original data,
we can train the encoder to maximize the \keywordDef{rate reduction},
and the decoder to minimize it,
where the rate reduction is 
given by
\be
\Delta R(\vZ,\vZ'; E, G) = R(\vZ;E) - R(\vZ';E,G) \geq 0
\ee
The new game becomes
\begin{align}
  \min_G \Delta R(\vZ,\vZ';E,G) + \lambda D(\vZ,\vZ';E,G)  \\
  \max_E \Delta R(\vZ,\vZ';E,G) - \lambda D(\vZ,\vZ'; E, G)
\end{align}
To see why this works,
note that the encoder $E$ has to maximize $R(\vZ)$,
so it must find features in the real data $X$ that are diverse and fill the latent space;
but it also must minimize $R(\vZ')$,
so it must be a good critic
that can find the flaws and repetitive patterns in the Generator's fake $\vX'$,
encoding them into a collapsed, low-rate, redundant representation $\vZ'$.
Conversely the best way for the generator to minimize the rate reduction
is to ensure that it makes $\vX'$ identical to $\vX$,
so then $\vZ'$ will be identical to $\vZ$,
and  the rate reduction will be zero.

Although the above approach is elegant, it is designed for a set
of iid static images.
The best way to extend it  to  sequential data, as arises
when learning a world model, is an open research problem.

\subsubsection{Example: JEPA}
\label{sec:JEPA}

\begin{figure}
  \centering
  \begin{tabular}{cc}
    \includegraphics[height=3in]{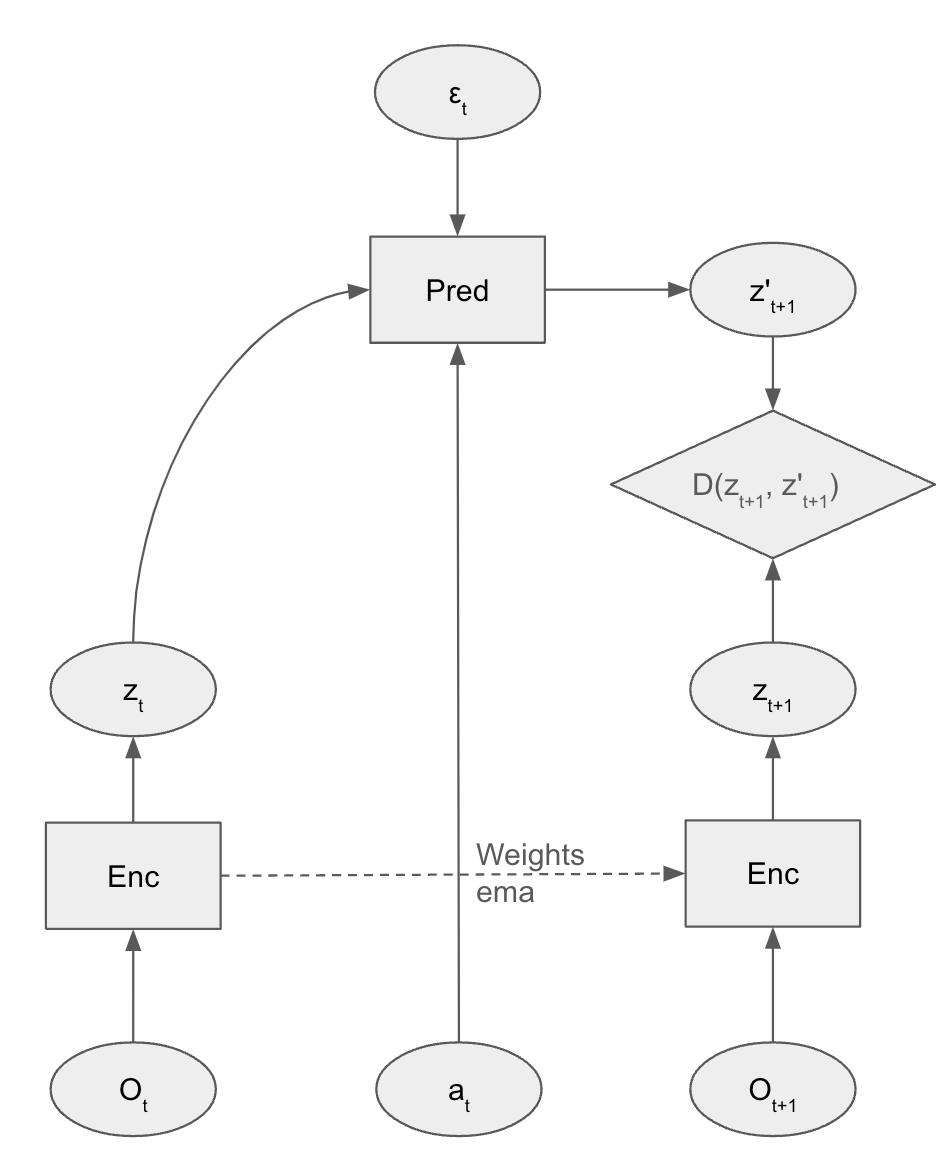}
    &
        \includegraphics[height=3in]{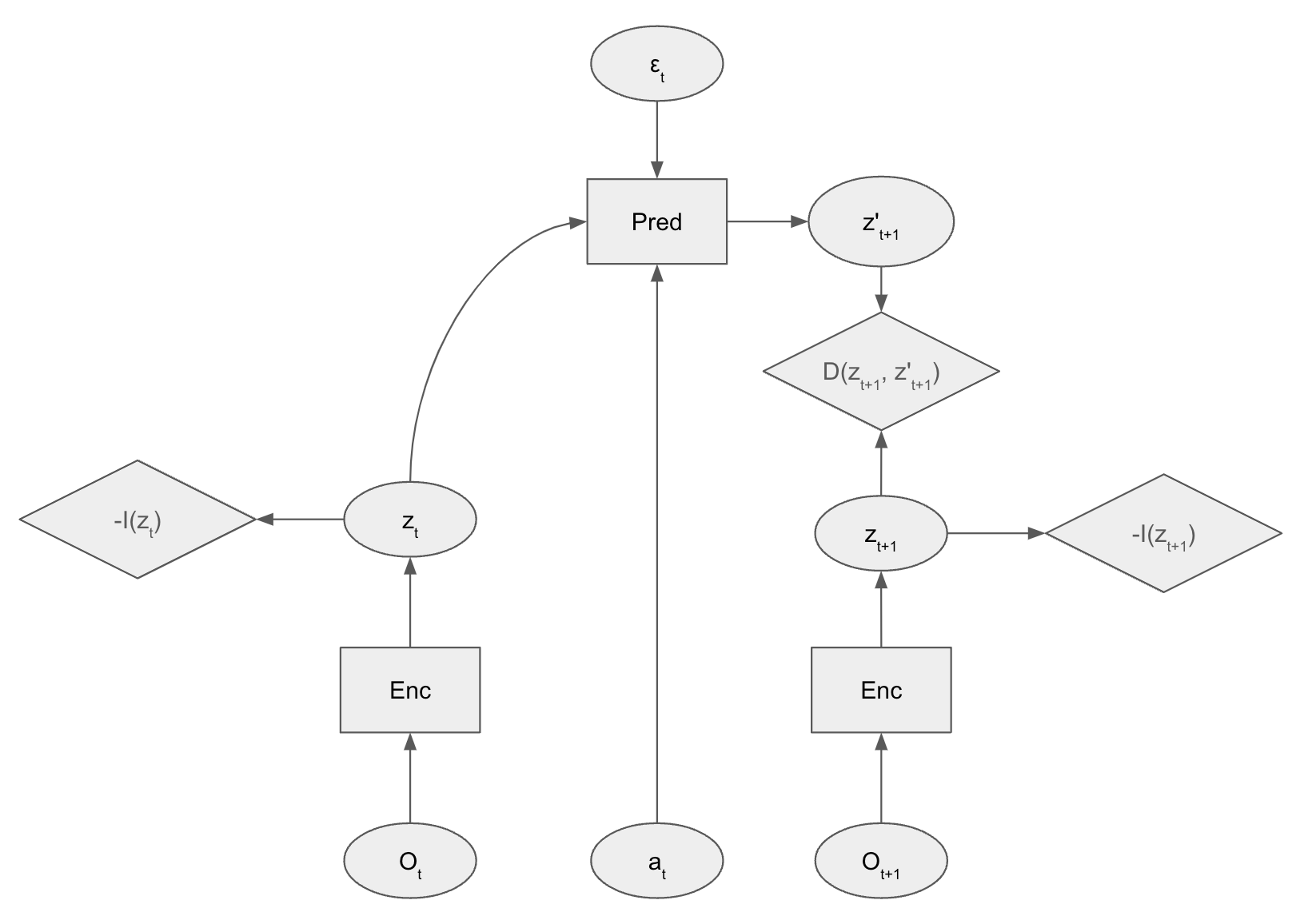}
    \end{tabular}
\caption{
  Illustration of the JEPA (Joint embedding Prediction Architecture) world model
  approach using two different approaches to avoid latent collapse:
  (a) self-distillation;
  (b) information theoretic regularizers.
  Diamonds represent fixed cost functions, squares represent learnable functions,
  circles are variables.
}
\label{fig:JEPA}
\end{figure}

In this section, we discuss the 
\keywordDef{JEPA} (Joint embedding Prediction Architecture)
approach to world modeling,
first proposed in \citep{LeCun2022Path}.
The basic idea is to jointly embed the current and following observations,
to compute $\vz_t=E(\vo_t)$ and $\vz_{t+1}=E(\vo_{t+1})$,
and then to compare the actual latent embedding  $\vz_{t+1}$
to its prediction $\vz'_{t+1}=\latentdynamics(\vz_t,a_t;\epsilon_t)$,
where $\epsilon_t$ is a random noise source,
and $M$ is the deterministic world model.
We then train the encoder to minimize the difference
between $\vz_t$ and $\vz'_t$.

To prevent the encoder collapsing to a trivial function,
such as $E(\vo)=\vzero$,
two different classes of methods have been considered.
The first is based on using
a frozen EMA version of the encoder,
as discussed in \cref{sec:collapseFrozen}.
JEPA also leverages the fact that the encoder is a low-dimensional
embedding of the input, and the predictor is a shallow network,
to create an information bottleneck.
(For example, if we use a VIT encoder,
an image of size $224 \times 224 \times 3 = 150,528$
gets compressed into a much smaller number of tokens,
which often amounts to an embedding size of $\sim 1024$.)
The \keywordDef{I-JEPA} method of \citep{IJEPA}, designed for images,
and the \keywordDef{V-JEPA} method of \citep{VJEPA,VJEPA2},
designed for videos,
also trains on masked versions of the inputs to create a
harder learning problem,
since the target shares less information
with the input.

An alternative way to avoid the latent collapse problem
is to add regularization terms that try to maximize
the information content in $\vz_t$ and $\vz_{t+1}$,
while also minimizing the prediction error,
as discussed in \cref{sec:collapseReg}.
See \cref{fig:JEPA} for an illustration to these two approaches.
(They can also be combined.
For example, \citep{Mo2024} combines I-JEPA
with VICReg.)

\subsubsection{Example: DinoWM}

In the case where the observations are high-dimensional,
such as images, it it natural
to use a pre-trained representation, $\vz_t=\phi(\vo_t)$,
as input to the world model (or policy).
The representation function $\phi$ can be pretrained
on a large dataset
using a non-reconstructive loss, such as the DINOv2 method
\citep{dinov2}.
Although this can sometimes give gains
(as in the \keywordDef{DinoWM} and \keywordDef{Dino-World} papers
\citep{dinoWM,Baldassarre2025}),
in other cases, better results are obtained
by training the representaton from scratch
\citep{Hansen2023pre,Schneider2024}.
The performance is
highly dependent on the similarity or differences
between between the pretraining
distribution and the agent's  distribution,
the form of the representation function,
and its training objective.

\subsubsection{Example: TD-MPC}
\label{sec:TDMPC}

In this section, we describe \keywordDef{TD-MPC2}
\citep{Hansen2024},
which is an extension of \keywordDef{TD-MPC} of \citep{Hansen2022}.
This learns the following functions:
\begin{itemize}
\item Encoder: $\ve_t=E(\vo_t)$
\item Latent dynamics (for rollouts): $\vz'_t=\latentdynamics(\vz_{t-1},\va_t)$
\item Latent update (after each observation):
  $\vz_t = \update(\vz_{t-1},\ve_t,\va_t)=\ve_t$
  \item Reward: $\hat{r}_t = R(\vz_t,\va_t)$
  \item Value: $\hat{q}_t = Q(\vz_t,\va_t)$
    \item Policy prior: $\hat{\va}_t=\pi_{\tprior}(\vz_{t-1})$
  \end{itemize}
The  model is trained using the following
VP+ZP loss applied to trajectories
sampled from the replay buffer:
\be
\loss(\vtheta) = \expectQ{
  \sum_{t=0}^H \lambda^t\left(
  ||\vz'_t - \stopgrad(E(\vo'_t))||_2^2
  + \text{CE}(\hat{r}_t, r_t)
  + \text{CE}(\hat{q}_t, q_t)
  \right)
  }{(\vo,\va,r,\vo')_{0:H} \sim \calB}
\ee
We  use cross-entropy loss on a discretized
representation of the reward and Q value
in a log-transformed space,
in order to be robust to different value scales
across time and problem settings (see \cref{sec:classifRL}).
The target  value for the Q function update is defined by
\be
q_t=r_t + \gamma \overline{Q}(\vz'_t,\pi_{\tprior}(\vz'_t))
\ee
where $\overline{Q}$ is the EMA for the Q function.

The policy is trained using the SAC objective
(see \cref{sec:SAC}) on imaginary rollouts in latent space
using observations and actions from the  replay buffer:
\be
\loss_{\pi}(\vtheta)
= \expectQ{
  \sum_{t=0}^H \lambda^t \left[
    \alpha Q(\vz_t, \pi_{\tprior}(\vz_t))
    -\beta \entropy(\pi_{\tprior}(\cdot|\vz_t))
    \right]
}{(\vo,\va)_{0:H} \sim \calB},
\vz_{t+1}=\latentdynamics(\vz_t,\va_t),
\vz_0=E(\vo_0)
\ee
This policy is used as a proposal (prior), in conjunction
with  the MPPI
trajectory planning method (\cref{sec:MPPI})
to select actions at run time.

In \citep{Escoriza2025}, they extended TD-MPC to work with
sparse rewards by assuming access to demonstrations,
partitioning the overall task into stages,
and then converting this semi-sparse stage reward to a dense reward,
which is needed to learn the world model.
In \citep{Hansen2025}, they avoid the need for demonstration data,
by using a latent prior on human motion, derived from mocap data,
to learn a whole body  world model and controller.

\subsubsection{Example: BYOL}
\label{sec:BYOL}

\begin{figure}
\centering
\includegraphics[height=4in]{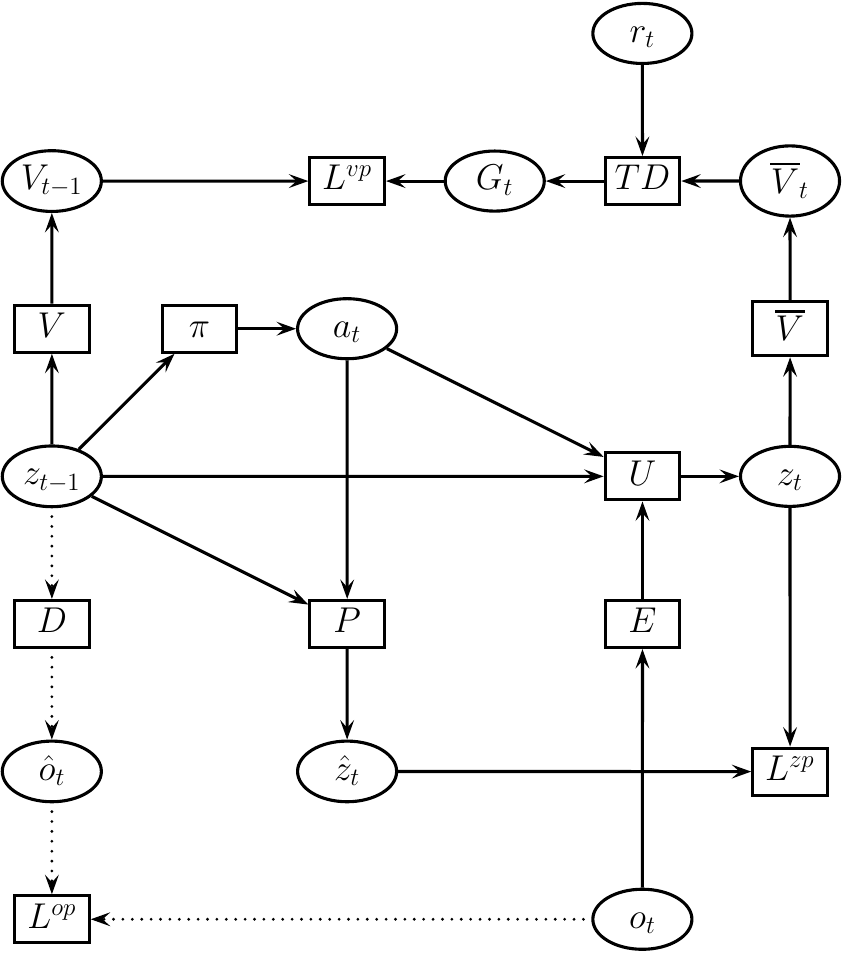}
\caption{
  Illustration of (a simplified version of)  the BYOL-Explore architecture,
  represented as a factor graph
  (so squares are functions, circles are variables).
  The dotted lines represent an optional observation
  prediction loss.
  The map from notation in this figure to the paper is as follows:
  $U \ra h^c$ (closed-loop RNN update),
  $P \ra h^o$ (open-loop RNN update),
  $D \ra g$ (decoder),
  $E \ra f$ (encoder).
  We have unrolled the forwards prediction for only 1 step.
  Also, we have omitted the reward prediction loss.
  The $\overline{V}$ node is the EMA version of the value function.
  The TD node is the TD operator.
}
\label{fig:byol}
\end{figure}

In \citep{BYOL}, they present
\keywordDef{BYOL} (Build Your Own Latents),
 which uses the ZP and VP loss.
See  \cref{fig:byol} for 
the computation graph,
which we see is slightly simpler than the
Dreamer computation graph in \cref{fig:dreamer}
due to the lack of stochastic latents.

In \citep{BYOLexplore}, they present
\keywordDef{BYOL-Explore},
which extends BYOL
by using the self-prediction error 
to define an intrinsic reward.
This encourages the agent
to explore states where the model is uncertain.
See \cref{sec:intrinsic} for further discussion of this topic.

\subsubsection{Example: Imagination-augmented agents}

In  \citep{Weber2017nips}, they train a model
to predict future states and rewards, and then use
the hidden states of this model as additional
context for a policy-based learning method.
This can help overcome partial observability.
They call their method \keywordDef{imagination-augmented agents}.

\eat{
\subsection{Self-prediction}
\label{sec:self-predictive}
\label{sec:SPR}

Since the reward and value of a state are just scalars,
predicting them  is often an insufficient feedback signal to train the
latent representation
(especially in sparse reward problems).
Fortunately we can add  additional losses or auxiliary signals,
to regularize the representation learning problem
\citep{Jaderberg2017iclr}.
This can help in situations when rewards are sparse or absent.

In general it can be difficult to decide what features
are worth modeling.
But a common approach is to predict the next latent state,
in addition to the reward and value,
as shown in \cref{fig:MBRLtargets}.
This is called a \keywordDef{self predictive representation} or \keywordDef{SPR}.
That is, we add a loss of the form $\loss(\hat{\vz}_{t+1}, \vz_{t+1})$,
where $\vz_{t+1}=E(\vo_{t+1})$ is the actual  next latent state,
$\hat{\vz}_{t+1}=\latentdynamics(\vz_t,a_t)$ is  the predicted  next latent state,
and the loss function is something like cosine distance in embedding space.
We use this loss not only to train the dynamics model $M$,
but also the encoder $E$.
A trivial way to minimize this loss wrt $E$ is for the embedding
to map everything to a constant vector, say $E(\vo)=\vzero$,
in which case $\vz_{t+1}$ will be trivial for  $M$
to predict.
However this is not a useful representation.
This problem is \keywordDef{latent representational collapse}
\citep{Jing2022}.
Fortunately,
we can provably prevent collapse (at lest for linear encoders)
by using a frozen target network and bilevel optimization
\citep{Tang2023,Ni2024}.
(The reward and value losses also help prevent total collapse.)
}

\eat{
As with  other contrastive losses used in self-supervised learning
(see \cref{sec:contrastiveLoss}),
we need to  prevent the embedding from collapsing to a trivial
function (e.g. $\phi(\vo_t)=\vzero_t$), which would be easy
to predict.
(This is similar to the problems caused by bootstrapping
in the context of off-policy RL, see \cref{sec:deadlyTriad}.)
The standard approach is to use tricks such as target networks
and stop gradients, as discussed in \cref{sec:targetNetwork}.
Alternatively, we can use non-contrastive or negative-free
methods (\cref{sec:negative-free});
for example,  \citep{Robine2023}
use  VICReg \citep{VICreg} inside a Dyna setup.
}

\eat{
It is also possible to use non-contrastive, but still non-generative,
methods for learning the latents. For example,
\keywordDef{DreamerPro} \citep{dreamerPro}
uses an approach based on ``prototypical representations'',
derived from the hidden state of the RNN dynamics model.
The \keywordDef{IFactor} approach of \citep{iFactor}
partitions the the latent factors into four groups,
based on whether they are predictive of reward or not,
and based on whether they are influenced by the actions or not.
}

\eat{
The action-value functions for the intermediate nodes
in the search tree are represented by deep neural networks,
and updated using temporal difference
methods that we discuss in \cref{sec:valueBased}.
MCTS can also be applied to many other kinds of seqential decision problems,
such as experiment design for
sequentially creating molecules~\citep{Segler2018}.
}


\eat{
\subsection{Variational inference}
\label{sec:VMBPO}
\label{sec:ALM}

In this section, we present another solution to the objective
mismatch problem.
It is an extension of the framework of
\cref{sec:inferRL}, where we formulate control (RL)
as an inference problem,
where the goal is to infer the posterior over states and actions
that maximizes the log probability of observing
the sequence of optimality variables
$O_{t}=1$, where $p(O_t=1|s_t,a_t) = \exp(R(s_t,a_t))$.
In \cref{sec:inferRL},  we only optimize the parameters of the policy,
treating the samples as coming from the true dynamics model,
resulting in the model-free SAC algorithm.
Here we jointly optimize over the policy $\pi$
and the dynamics  $M$, allowing for planning and policy
learning ``in imagination''.
We follow the presentation of
\citep{Wei2023MBRL},
which should be consulted for more details.

Using standard methods from variational inference (see \cref{chap:VI}),
we can derive the following  lower bound
on the log likelihood
\begin{equation*}
  \log p(O_{0:\infty}=1) \geq
  \expectQ{\sum_{t=0}^{\infty} R(s_t,a_t)}{Q(\tau)}
  -\KLpq{Q(\tau)}{P(\tau)}
\end{equation*}
In the above,
$P(\tau)$ is
the true distribution over state and action trajectories, given by
\begin{equation*}
  P(\tau) = \mu(s_0) \prod_{t=0}^{\infty} M(s_{t+1}|s_t,a_t)
  P(a_t|s_t)
\end{equation*}
where $P(a|s)$ is the behavior policy (which we may approximate as uniform),
$\mu(s_0)$ is the distribution over starting states,
and $M$ is the true dynamics model.
Furthermore, $Q(\tau)$ is the approximate distribution
given by
\begin{equation*}
  Q(\tau) = \mu(s_0) \prod_{t=0}^{\infty} \hat{M}(s_{t+1}|s_t,a_t)
  \pi(a_t|s_t)
\end{equation*}
We can therefore rewrite the lower bound maximization problem
as follows:
\begin{equation*}
  \max_{\hat{M},\pi}
  \expectQ{\sum_{t=0}^{\infty} \gamma^t\left(
    R(s_t,a_t) + \log \frac{P(a_t|s_t)}{\pi(a_t|s_t)}
    + \log \frac{M(s_{t+1}|s_t,a_t)}{\hat{M}(s_{t+1}|s_t,a_t)}\right)
    }{Q(\tau)}
\end{equation*}
\eat{
In \citep{Chow2021},
they show that the optimal variational dynamics and policy have the
form
\begin{eqnarray*}
  \hat{M}(s'|s,a)& \propto& \exp(V(s') + \log M(s'|s,a)) \\
  \pi(a|s) &\propto& \exp(Q(s,a) + \log P(a|s))
\end{eqnarray*}
where
we have used the following Bellman-like equation:
\begin{equation*}
Q(s_t,a_t) = R(s_t,a_t) + \log \frac{P(a_t|s_t)}{\pi(a_t|s_t)}
+ \expectQ{V(s_{t+1}) +
  \log \frac{M(s_{t+1}|s_t,a_t)}
       {\hat{M}(s_{t+1}|s_t,a_t)}
       }
    {\hat{M}(s_{t+1}|s_t,a_t)}
\end{equation*}
and we define $V(s) = \argmax_a Q(s,a)$.
}
We can approximate the  density ratio
$\frac{M(s_{t+1}|s_t,a_t)}{\hat{M}(s_{t+1}|s_t,a_t)}$
using a binary classifier,
as discussed in \cref{sec:class_probability_estimation}.
In \citep{Chow2021}, they call this method
\keywordDef{VMBPO}
for ``Variational Model-Based Policy Optimization''.

In \citep{Eysenbach2022}, they derive a tighter bound,
by defining the optimality variable based on the entire trajectory,
$p(O=1|\tau) = R(\tau) = \sum_{t=1}^{\infty} \gamma^t R(s_t,a_t)$.
This gives rise to the following simpler lower bound:
\begin{eqnarray*}
  \log p(O=1) &= \log \int_{\tau} P(O=1,\tau)
  = \log \expectQ{P(O=1|\tau)}{P(\tau)}
  \geq \expectQ{\log R(\tau) + \log P(\tau) - \log Q(\tau)}{Q(\tau)}
\end{eqnarray*}
This gives rise to   an objective that is the same
as in VMBPO but $R(s,a)$ is replaced by $\log R(s,a)$.
They call their method
\keywordDef{Mismatched No More},
so-called since it solves the model-policy
mismatch.

In \citep{Ghugare2022}
they extend MNM to work with images (and other high dimensional states)
by learning a latent encoder $\hat{E}v(z_t|s_t)$
as well as latent dynamics $\hat{M}(z_{t+1}|\z_t,a_t)$.
The trajectories are now defined over observed states, latent states,
and actions, and the objective becomes the following:
\begin{equation*}
  \max_{\hat{M},\hat{E},\pi}
  \expectQ{\sum_{t=0}^{\infty} \gamma^t\left(
    \log R(s_t,a_t) + \log \frac{P(a_t|s_t)}{\pi(a_t|s_t)}
    + \log \frac{\hat{E}(z_{t+1}|s_{t+1})}{\hat{M}(z_{t+1}|z_t,a_t)}\right)
    }{Q(\tau)}
\end{equation*}
They call their method
\keywordDef{Aligned Latent Models}.\footnote{
Note that the above objective corresponds to a simplified
form of the offline RL
objective in Appendix A.7 of \citep{Ghugare2022}.
It is simplified because we  use Monte Carlo rollouts
rather than a $K$-step bootstrap estimate of the return.
Note that,
in the offline setting, the $P(a_t|s_t)$ term can be learned using  behavior cloning.
However, it is also possible to derive an online RL objective
in which $P(a_t|s_t) = \pi(a_t|s_t)$, so this term cancels out.
See \citep{Ghugare2022} for details.
}
We note that the final term is similar to
other self-predictive representation learning 
methods discussed in \cref{sec:self-predictive},
and the authors use a simlar target network approach for the
encoder $E$ to ensure stability of training.

}

\subsection{World models that are trained to help planning}
\label{sec:differentiable-planning}
\label{sec:endToEnd}

One solution to the objective mismatch  problem is to use
\keywordDef{differentiable planning},
in which we combine model learning and policy learning together,
and train them jointly end-to-end,
rather than in an alternating fashion.
In particular, we can solve try to optimize
\begin{equation*}
  \min_{\hat{M}, Q} \expectQ{(R(s,a) + \gamma V(s') - Q(s,a))^2}
      {(s,a,s')    \sim \data}
  \end{equation*}
where  $s'=\hat{M}(s,a)$ is the learned dynamics model,
subject to the constraint that the value function
is derived from the model using
\begin{equation*}
V(s) = \argmax_{a(0:K)} \expectQ{\sum_{k=0}^{K-1}
  \gamma^k R(s_k,a_k) + \gamma^K V(s_K) | S_0=s}{\hat{M}}.
\end{equation*}
This bilevel optimization problem was first proposed
in the \keywordDef{Value Iteration Network} paper
of \citep{Tamar2016},
and extended
in the \keywordDef{TreeQN} paper
\citep{treeQN}.
In \citep{Mittal2025}, they propose
\keywordDef{D-TSN} (differentiable tree search network),
which is similar to TreeQN, but constructs a  best-first  search
tree, rather than a fixed depth tree,
using a stochastic tree expansion method.

In \citep{Nikishin2022,Bansal2023}
they propose to use implicit differentation to avoid
explicitly unrolling the inner optimization.

\subsection{Dealing with model errors and uncertainty}
\label{sec:uncertainty}
\label{sec:modelUncertainty}

The theory in \cref{sec:MBRLgame} tells us that the model-as-leader
approach, which trains a new policy in imagination at each inner iteration 
while gradually improving the model in the outer loop,
will converge to the optimal policy,
provided the model converges to the true model (or one that is
\keyword{value equivalent} to it, see \cref{sec:valueEquivalence}).
This can be assured provided the model is sufficiently powerful,
and the policy explores sufficiently widely to collect
enough diverse but task-relevant data.
Nevertheless, models will inevitably have errors, and it can be useful
for the policy learning to be aware of this
(see e.g., \citep{Arumugam2018}).
We discuss some approaches to this below.

\subsubsection{Avoiding compounding errors  in rollouts}

In MBRL,
we have to rollout imaginary trajectories to use for training
the policy.
It makes intuitive sense to 
start from a previously visited real-world state,
since the model will likely be reliable there.
We should start rollouts from different points along each real trajectory,
to ensure good state coverage,
rather than just expanding around the initial state \citep{Rajeswaran2017}.
However, if we roll out too far from a previously seen state, the trajectories
are likely to become less realistic,
due to \keywordDef{compounding errors} from the model
\citep{Lambert2022}.

In \citep{Janner2019}, they present the
\keyword{MBPO} method,
which  uses short rollouts (inside Dyna) to prevent compounding error
(an approach which is justified in \citep{Jiang2015}).
\citep{Frauenknecht2024} is a recent extension of MBPO which
dynamically decides how much to roll out, based on model uncertainty.

Another approach to mitigating compounding errors
is to learn a trajectory-level dynamics model,
instead of a single-step model,
see e.g., \citep{dmpc} which uses diffusion to train
$p(s_{t+1:t+H}|s_t,a_{t:t+H-1})$,
  and uses this inside an MPC loop.

If the model is able to predict a reliable distribution
over future states, then we can leverage this uncertainty
estimate to compute an estimate of the expected reward.
For example, PILCO
\citep{pilco,pilcoJ} uses Gaussian processes as the world model,
and uses this to analytically derive the expected reward over
trajectories as a function of policy parameters,
which are then optimized using a deterministic second-order
gradient-based solver.
In 
\citep{Mankowitz2019}, they combine the MPO algorithm
(\cref{sec:MPO}) for continuous control
with \keywordDef{uncertainty sets} on the dynamics to
learn a policy that optimizes for a worst case expected return
objective.

\subsubsection{Unified model and planning variational lower bound}
\label{sec:unified}

In \citep{Eysenbach2022}, they propose a method called
\keywordDef{Mismatched No More} (MNM) to solve the objective mismatch problem.
They define an optimality variable (see \cref{sec:inferRL})
based on the entire trajectory,
$p(O=1|\tau) = R(\tau) = \sum_{t=1}^{\infty} \gamma^t R(s_t,a_t)$.
This gives rise to the following variational lower bound
on the log probability of optimality:
\begin{eqnarray*}
  \log p(O=1) &= \log \int_{\tau} P(O=1,\tau)
  = \log \expectQ{P(O=1|\tau)}{P(\tau)}
  \geq \expectQ{\log R(\tau) + \log P(\tau) - \log Q(\tau)}{Q(\tau)}
\end{eqnarray*}
where $P(\tau)$ is the distribution over trajectories induced by policy
applied to the true world model,
  $P(\tau) = \mu(s_0) \prod_{t=0}^{\infty} M(s_{t+1}|s_t,a_t)  \pi(a_t|s_t)$,
and $Q(\tau)$ is the distribution over trajectories
using the estimated world model,
  $Q(\tau) = \mu(s_0) \prod_{t=0}^{\infty} \hat{M}(s_{t+1}|s_t,a_t)
  \pi(a_t|s_t)$.
They then maximize this bound wrt $\pi$ and $\hat{M}$.

In \citep{Ghugare2022}
they extend MNM to work with images (and other high dimensional states)
by learning a latent encoder $\hat{E}(\vz_t|\vo_t)$
as well as latent dynamics $\hat{M}(\vz_{t+1}|\vz_t,a_t)$,
similar to other self-predictive methods
(\cref{sec:self-predictive}).
They call their method
\keywordDef{Aligned Latent Models}.

\eat{
\subsubsection{Lower bound optimization}

Various lower bound optimization methods have been developed,
that guarantee monotonically increasing policy performance
even in the presence of model error.
Such methods include
\citep{Luo2019RL}, who propose  \keywordDef{SLBO} (Stochastic Lower
Bounds Optimization);
\eat{
\citep{CMLO} who propose \keywordDef{CMLO}
(Constrained Model-shift Lower-bound Optimization);
\citep{Zhang2023USBPO} who propose
\keywordDef{USB-PO} algorithm (Unified model Shift and model Bias
Policy Optimization);
and \citep{Dong2024}  who propose the \keywordDef{MBMFPO} method
(Model-Based Reinforcement Learning with Model-Free Policy Optimization);
}
\citep{Eysenbach2022}, who propose
\keywordDef{Mismatched No More};
and
\citep{Ghugare2022}, who propose
\keywordDef{Aligned Latent Models}.
}

\subsubsection{Dynamically switching between MFRL and MBRL}

One problem with the above methods is that,
if the model is of limited capacity, or if it learns
to model ``irrelevant'' aspects of the environment,
then any MBRL method may be dominated by a MFRL method
that directly optimizes the true expected reward.
A safer approach is to use a model-based policy only  when
the agent is confident it is better, but otherwise
to fall back to a model-free policy.
This is the strategy proposed
in the \keywordDef{Unified RL}
method of  \citep{Freed2024}.

\eat{
In more detail, the idea is as follows.
Let $\vw$ be the parameters of the world model,
which induces the following distribution over trajectories:
\be
p(\tau|\pi,\vw) = 
\prod_{t=1}^T p(s_t|s_{t-1}, a_{t-1},\vw) \pi(a_t|s_t) p(r_t|s_t,a_t,\vw)
\ee
(where for $t=0$ the term inside the product becomes
$p(s_0) \pi(a_0|s_0) p(r_0|s_0,a_0,\vw)$).
Define the expected return of the policy, given a particular
set of parameters, as
\be
J(\pi|\vw) = \expectQ{\sum_{t=0}^T r_t | \pi, \vw}{p(\tau|\pi,\vw)}
\ee
Let $\pi^{MB}$ be a policy learned using MBRL (e.g., AC on imaginary trajectories),
and $\pi^{MF}$ be a policy learned using MFRL (e.g., SAC on replay buffer).
Let $\data$ be all the data collected so far,
and let $p(\vw|\data)$ be the posterior belief over the model parameters.
It is better to use  $\pi^{MB}$ 
only if our posterior belief is that $\pi^{MB}$ is better,
i.e., 
\be
\int p(\vw|\data) [J(\pi^{MB}|\vw) - J(\pi^{MF}|\vw)] d \vw \geq 0
\ee
Essentially we are using the model just to rule out bad model-free policies.
In practice,
we can approximate the integral by sampling from the posterior over $\vw$;
we can approximate the posterior using standard methods for
Bayesian  neural nets;
and we can approximate $J(\pi|\vw)$ using MC rollouts.

In \citep{Freed2024}, they apply this method to various Gym and MuJoCo
continuous control tasks, using an LSTM dynamics model,
and dropout \citep{Gal2016rnn} to approximate the posterior.
For tasks that are hard to reliably model (such as Ant),
the method initially uses MBRL, which works better in the low sample regime,
but then automatically switches to MFRL in the large data regime,
to avoid the asymptotic limit imposed by a poor  model.
For tasks where the model is a good fit (such as Walker),
the method uses MBRL throughout.
}

\subsection{Exploration for learning world models}
\label{sec:exploreWM}

In \cref{sec:exploreExploit}, we discussed the exploration-exploitation tradeoff,
which contrasts the need to  (1) collect diverse experiences (by trying many new actions in
many new states) so as to learn a better policy to help long-run performance
with (2) the need to  stay in familiar
parts of the state space where the optimal policy has already been learned,
so as to ensure short-term rewards.
When using MBRL, the need for diverse data becomes even more important,
to ensure we learn the correct underlying  world model
(which is then used to train the policy, or for online planning).

One popular approach to this is to use
\keyword{posterior sampling RL},
which applies
Thompson sampling  to the MDP parameters (i.e., the world model),
as explained in \cref{sec:PSRL}.
This was applied to MBRL in \citep{Wang2024MBRL}.

If we are in the reward-free setting (see \cref{sec:intrinsic}),
we can view the problem of learning
a world model as similar to the scientist's job of trying to create a
\keywordDef{causal model} of the world, which can explain the effects
of actions (interventions). This requires designing and carrying out experiments
in order to collect informative trajectories for model fitting
(see e.g., \citep{Sekar2020,Cao2025causalICLR,Cao2025causal,Cao2025china}).
Recently it has become popular to use LLMs to help with this problem;
this can be thought of an 
\keywordDef{AI scientist}
(see e.g., \citep{Gandhi2025}).
It is also possible to combine LLMs as hypothesis generators
with Bayesian inference and information theoretic reasoning principles
for a more principled approach to the problem
(see e.g., \citep{Piriyakulkij2024}).

\section{Beyond one-step models: predictive representations}
\label{sec:beyond}
\label{sec:successor}

The ``world models'' we described in \cref{sec:WM}
are \keywordDef{one-step models} of the form $p(s'|s,a)$,
or $p(z'|z,a)$ for $z=\phi(s)$, where $\phi$ is a state-abstraction
function.
However, such models are problematic when it comes to predicting
many kinds of future events, such as ``will a car pull in front of me?''
or ``when will it start raining?'',
since it is hard to predict exactly when these events will occur,
and these events may correspond to many different ``ground states''.
In principle we can roll out many possible long term futures,
and apply some abstraction function to  the resulting
generated trajectories to extract features of interest,
and thus derive a predictive model of the form $p(t', \phi(s_{t+1:t'})|s_t,\pi)$,
where $t'$ is the random duration of the sampled trajectory,
and $\phi$ maps from state trajectories to features.
However, it would be more
efficient if we could directly predict this distribution
without having to know the value of $t'$,
and without having to predict all the details of all the
intermediate future states,
many of which will be irrelevant after we pass
them into the abstraction function $\phi$.
This motivates the study of multi-step world models,
that predict multiple steps into the future,
either at the state level, or at the feature level.
These are called \keywordDef{predictive representations},
and are a compromise between standard model-based RL
and model-free RL, as we will see.
Our presentation on this topic is based on 
\citep{Carvalho2024}.
(See also \cref{sec:HRL}, where we discuss the related topic
of temporal abstraction from a model-free perspective.)

\subsection{General value functions}
\label{sec:GVF}

The value function is based on  predicting the sum of expected
discounted future rewards.
But the reward is just one possible signal of interest we can extract from the environment.
We can generalize this by considering a \keywordDef{cumulant}
$C_t \in \real$, which is some scalar of interest derived from the state or observation
(e.g., did a loud bang just occur? is there a tree visible in the image?).
We then define the \keywordDef{general value function} or \keywordDef{GVF}
as follows \citep{GVF,Sutton2011,Comanici2018,Ring2021,Xu2022GVF,Graves2022}:
\be
V^{\pi,C,\gamma}(s) = \expect{
  \sum_{t=0}^{\infty} \gamma^t C(s_{t+1}) \vert s_0=s, a_{0:\infty} \sim \pi}
\ee
If $C(s_{t+1})=R_{t+1}$, this reduces to the value function.\footnote{
This follows the convention  of \citep{Suttonv2},
where we write $(s_t, a_t, r_{t+1}, s_{t+1})$  --- 
as opposed to  $(s_t, a_t, r_{t}, s_{t+1})$ ---
to represent the transitions, since
$r_{t+1}$ and $s_{t+1}$ are both generated
by applying $a_t$ in state $s_t$.
}
If we define the cumulatant to be the observation vector,
then the GVF will learn to  predict future observations
at multiple time scales;
this is called \keywordDef{nexting} \citep{Modayil2014ncai,Modayil2014,White2015}.
Predicting the GVFs for multiple cumulants can be useful as an auxiliary
task while  solving
the main task (e.g., as a form of auxiliary input to the policy,
or just to ``densify'' the training signal),
as shown in \citep{Jaderberg2017iclr}.

\citep{Veeriah2019} present
an approach (based on meta-gradients) to learn which cumulants are worth predicting,
In the inner loop, the model $f$ predicts the policy $\pi_t$ and value function $V_t$,
as usual, and also predicts the GVFs $\vy_t$ for the specified cumulants;
the function $f$ is called the answer network,
and is denoted by $(\pi_t,V_t,\vy_t)=f_{\vtheta}(\vo_{t-i-1:t})$.
In the outerloop, the model $g$ learns  to extract the
cumulants and their discounts given future observations;
this called the question network
and is denoted by
$(\vc_t,\vgamma_t)=g_{\veta}(\vo_{t+1:t+j})$.
The outer update to $\veta$  is based on the gradient of the RL loss
after performing $K$ inner updates to $\vtheta$ using the
RL loss and auxiliary loss.

\subsection{Successor representations}
\label{sec:SR}
\label{sec:PSR}


In this section we consider a variant of GVF
where the cumulant corresponds
to a state occupancy vector $C_{\tilde{s}}(s_{t+1}) = \ind{s_{t+1}=\tilde{s}}$,
which provides a dense feedback signal.
Computing this for each possible state $\tilde{s}$
gives  us the \keywordDef{successor representation} or \keywordDef{SR}
\citep{Dayan1993,Machado2023}:
\be
M^{\pi}(s,\tilde{s}) = \expect{
  \sum_{t=0}^{\infty} \gamma^t \ind{s_{t+1}=\tilde{s}} \vert S_0=s}
\ee
If we define the policy-dependent state-transition matrix by
\be
T^{\pi}(s,s') = \sum_a \pi(a|s) T(s'|s,a)
\ee
then the SR matrix can be rewritten as
\be
\vM^{\pi} = \sum_{t=0}^{\infty} \gamma^t [\vT^{\pi}]^{t+1}
= \vT^{\pi} (\vI - \gamma \vT^{\pi})^{-1}
\label{eqn:SRmatrix}
\ee
Thus we see that the SR replaces information about
individual transitions with their cumulants,
just as the value function replaces individual rewards
with the reward-to-go.

\begin{figure}
\centering
\includegraphics[height=3in]{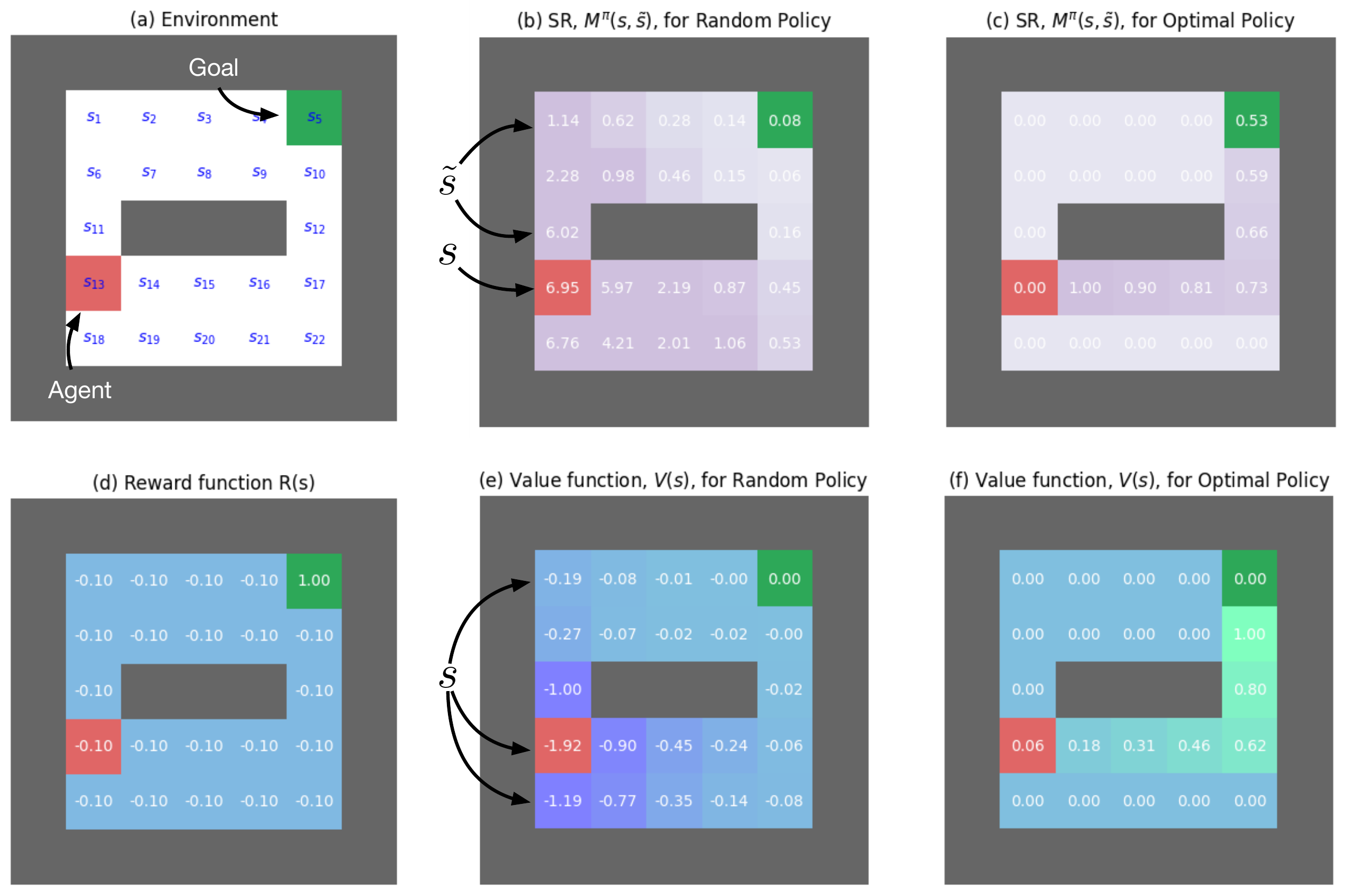}
\caption{
  Illustration of successor representation for
  the 2d maze environment shown in (a) with reward
  shown in (d), which assigns all states a reward of -0.1
  except for the goal state which has a reward of 1.0.
  In (b-c) we show the  SRs for a random policy and the optimal policy.
  In (e-f) we show the corresponding value functons.
  In (b), we see that the SR under the random policy
  assigns high state occupancy values to states which are close
  (in Manhattan distance)
  to the current state $s_{13}$
  (e.g., $M^{\pi}(s_{13}, s_{14})=5.97$)
  and low values to states that are further away
  (e.g., $M^{\pi}(s_{13}, s_{12})=0.16$).
  In (c), we see that the SR under the optimal policy
  assigns high state occupancy values to states which are close
  to the optimal path to the goal
  (e.g., $M^{\pi}(s_{13}, s_{14})=1.0$)
  and which fade with distance from the current state
  along that path
  (e.g., $M^{\pi}(s_{13}, s_{12})=0.66$).
  \figtaken{Figure 3 of \citep{Carvalho2024}}.
  \figthanks{Wilka Carvalho}.
  \figgen{\url{https://github.com/wcarvalho/jaxneurorl/blob/main/successor_representation.ipynb}}.
}
\label{fig:successorRep}
\end{figure}

Like the value function, the SR obeys a Bellman equation
\begin{align}
  M^{\pi}(s,\tilde{s})
  &= \sum_a \pi(a|s) \sum_{s'}
T(s'|s,a)\left( \ind{s'=\tilde{s}} + \gamma M^{\pi}(s',\tilde{s})
\right) \\
   &= \expect{\ind{s'=\tilde{s}} + \gamma M^{\pi}(s',\tilde{s})}
\end{align}
Hence we can learn an SR using a TD update of the form
\begin{align}
  M^{\pi}(s,\tilde{s}) &\leftarrow
  M^{\pi}(s,\tilde{s}) + \eta 
  \underbrace{\left( \ind{s'=\tilde{s}} + \gamma M^{\pi}(s',\tilde{s})
     - M^{\pi}(s,\tilde{s}) \right)}_{\delta}
\end{align}
where $s'$ is the next state sampled from $T(s'|s,a)$.
Compare this to the value-function TD update
in \cref{eqn:rl-td}:
\be
V^{\pi}(s) \leftarrow
V^{\pi}(s) + \eta \underbrace{ \left(
  R(s') + \gamma V^{\pi}(s') - V^{\pi}(s) \right)}_{\delta}
\ee
However, with an SR,
we can easily compute the value
function for any reward function
(given a fixed policy)
as follows:
\be
V^{R,\pi} = \sum_{\tilde{s}} M^{\pi}(s,\tilde{s}) R(\tilde{s})
\ee
See \cref{fig:successorRep} for an example.

We can also make a version of SR that depends on the action
as well as the state to get
\begin{align}
M^{\pi}(s,a,\tilde{s})
&= \expect{\sum_{t=0}^{\infty} \gamma^t \ind{s_{t+1}=\tilde{s}}
  \vert s_0=s, a_0=a, a_{1:\infty} \sim \pi} \\
&= \expect{\ind{s'=\tilde{s}} + \gamma M^{\pi}(s',a,\tilde{s})
  \vert s_0=s, a_0=a, a_{1:\infty} \sim \pi}
\end{align}
This gives rise to a TD update of the form
\begin{align}
  M^{\pi}(s,a,\tilde{s}) &\leftarrow
  M^{\pi}(s,a,\tilde{s}) + \eta 
  \underbrace{\left( \ind{s'=\tilde{s}} + \gamma M^{\pi}(s',a',\tilde{s})
     - M^{\pi}(s,a,\tilde{s}) \right)}_{\delta}
\end{align}
where $s'$ is the next state sampled from $T(s'|s,a)$
and $a'$ is the next action sampled from $\pi(s')$.
Compare this to the (on-policy) SARSA update from
\cref{eqn:rl-td-q}:
\be
Q^{\pi}(s,a) \leftarrow
Q^{\pi}(s,a) + \eta \underbrace{ \left(
  R(s') + \gamma Q^{\pi}(s',a') - Q^{\pi}(s,a) \right)}_{\delta}
\ee
However, from an SR, we can compute the state-action value function
for any reward function:
\be
Q^{R,\pi}(s,a) = \sum_{\tilde{s}} M^{\pi}(s,a,\tilde{s}) R(\tilde{s})
\ee
This can be used to improve the policy as we discuss
in \cref{sec:GPI}.

We see that the SR  representation has the computational
advantages of model-free RL (no need to do explicit planning or rollouts
in order to compute the  optimal action),
but also the flexibility of model-based RL
(we can easily change the reward function without having
to learn a new value function).
This latter property makes SR particularly well suited
to problems that use intrinsic reward (see \cref{sec:intrinsicReward}),
which often changes depending on the information state of the agent.

Unfortunately, the SR is limited in two key  ways:
(1) it assumes a finite, discrete state space;
(2)  it depends on a given policy.
We discuss ways to overcome limitation 1
in \cref{sec:SF} and \cref{sec:SM},
and limitation 2 in \cref{sec:SRoptions}.


\subsection{Successor features}
\label{sec:SF}

SRs  require defining expectations
or distributions over the entire future state vector,
which can be problematic in high dimensional and continuous spaces.
In \citep{Barreto2017} they introduced
\keywordDef{successor features}, that generalize
SRs by working with features $\vphi(s)$ instead
of primitive states. In particular, if we define
the cumulant to be
$C(s_{t+1}) = \vphi(s_{t+1})$,
we get the following definition of SF:
\be
\vpsi^{\pi,\vphi}(s) = \expect{
  \sum_{t=0}^{\infty} \gamma^t \vphi(s_{t+1}) \vert s_0=s, a_{0:\infty} \sim \pi}
\ee
We will henceforth drop the $\vphi$ superscript from the notation,
for brevity.

By analogy to \cref{eqn:SRmatrix},
we can write the successor features in matrix form
as follows:
\be
\vPsi^{\pi} = \sum_{t=0}^{\infty}
(\gamma \vT^{\pi})^t \vPhi = (\vI - \gamma \vT^{\pi})^{-1} \vPhi
\ee
where $\vPhi$ is the $S \times D$ matrix of features for each state.

SFs also obey a Bellman equation
\be
\vpsi(s) = \expect{\vphi(s') + \gamma \vpsi(s')}
\ee

If we assume the reward function can be written as
\be
R(s,\vw) = \vphi(s)^\trans \vw
\ee
then we can derive the value function for any reward
as follows:
\begin{align}
  V^{\pi,\vw}(s)
  &= \expect{R(s_1) + \gamma R(s_2) + \cdots \vert  s_0=s} \\
  &= \expect{\vphi(s_1)^\trans \vw + \gamma \vphi(s_2)^\trans \vw + \cdots \vert  s_0=s} \\
  &= \expect{\sum_{t=0}^{\infty} \gamma^t \vphi(s_{t+1}) \vert  s_0=s}^\trans \vw 
  = \vpsi^{\pi}(s)^\trans \vw
\end{align}

Similarly we can define an action-conditioned version of SF as
\begin{align}
\vpsi^{\pi,\vphi}(s,a) &= \expect{
  \sum_{t=0}^{\infty} \gamma^t \vphi(s_{t+1}) \vert s_0=s, a_0=a, a_{1:\infty} \sim \pi} \\
&= \expect{\vphi(s') + \gamma \vpsi(s',a')}
\end{align}
We can learn this using a TD rule
\be
\vpsi^{\pi}(s,a) \leftarrow
\vpsi^{\pi}(s,a) + \eta \underbrace{
  \left( \vphi(s') + \gamma \vpsi^{\pi}(s', a') - \vpsi^{\pi}(s,a) \right)}_{\delta}
\label{eqn:SFTD}
\ee
And we can use it to derive a state-action value function:
\begin{align}
  Q^{\pi,\vw}(s)
  = \vpsi^{\pi}(s,a)^\trans \vw
\end{align}
This allows us to define multiple $Q$ functions (and hence policies)
just by changing the weight vector $\vw$, as we discuss in \cref{sec:GPI}.

\subsubsection{Generalized policy improvement}
\label{sec:GPI}

So far, we have discussed how to compute the value function
for a new reward function but using the SFs
from an existing known policy.
In this section we discuss how to create a new policy
that is better than an existing set of policies,
by using \keywordDef{Generalized Policy Improvement}
or \keywordDef{GPI} \citep{Barreto2017,Barreto2020}.

\begin{figure}
\centering
\includegraphics[height=3in]{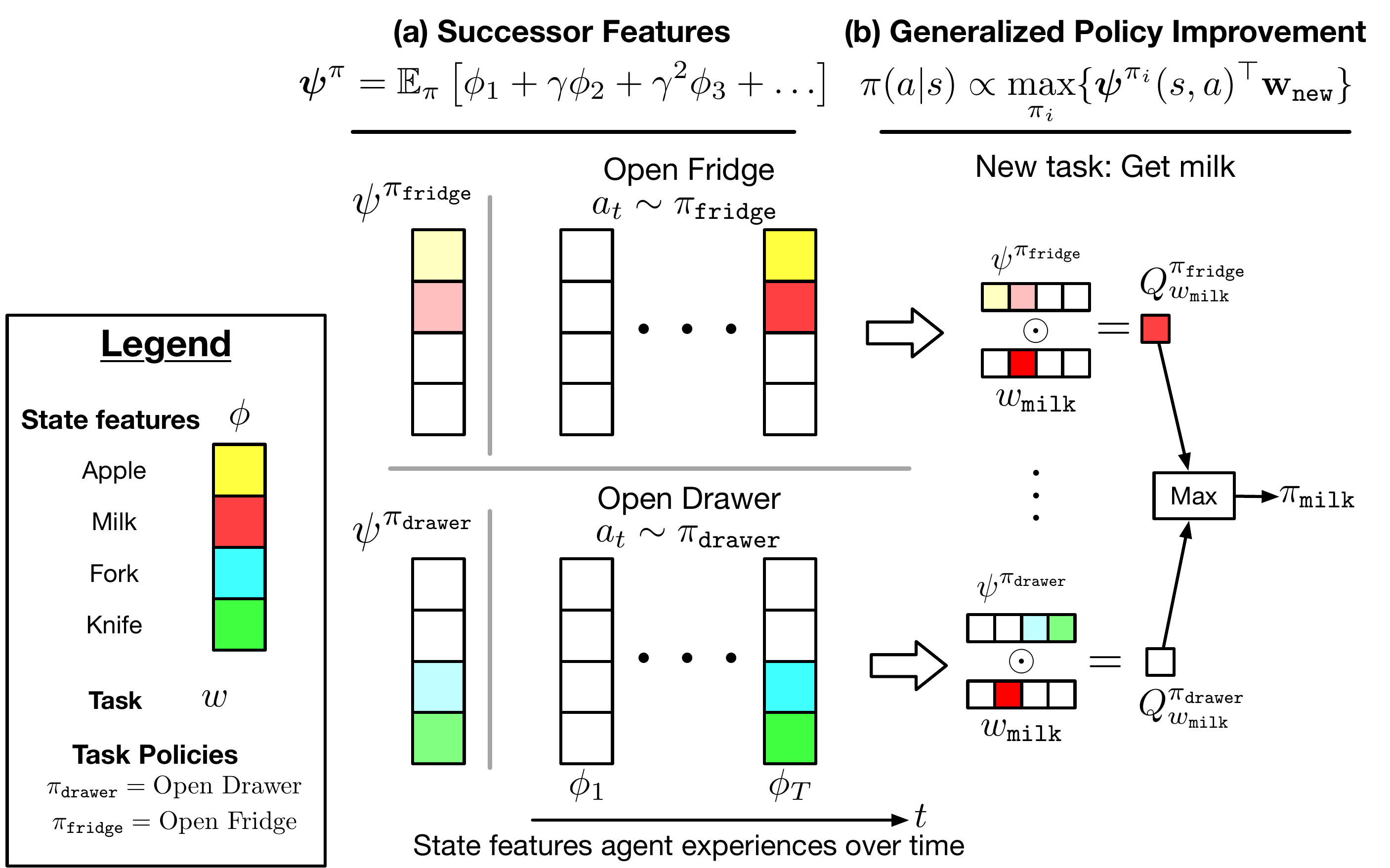}
\caption{
  Illustration of successor features representation.
  (a)
  Here $\vphi_t=\vphi(s_t)$ is the vector of features
  for the state at time $t$, and $\vpsi^{\pi}$
  is the corresponding SF representation, which depends
  on the policy $\pi$.
  (b) Given a set of existing policies and their SFs,
  we can create a new one by specifying a desired
  weight vector $\wnew$ and taking a weighted combination
  of the existing SFs.
  \figtaken{Figure 5 of \citep{Carvalho2024}}.
    \figthanks{Wilka Carvalho}.
}
\label{fig:successorFeatures}
\end{figure}

Suppose we have learned a set of $N$ (potentially optimal)
policies $\pi_i$ and their corresponding SFs $\vpsi^{\pi_i}$
for maximizing rewards defined by $\vw_i$.
When presented with a new task $\wnew$,
we can compute a new policy using GPI as follows:
\begin{align}
  a^*(s;\wnew) &=
  \argmax_a \max_i Q^{\pi_i}(s,a,\wnew)
  =\argmax_a \max_i \vpsi^{\pi_i}(s,a)^\trans \wnew
\end{align}
If $\wnew$ is in the span of the training tasks
(i.e., there exist weights $\alpha_i$ such that $\wnew = \sum_i \alpha_i \vw_i$),
then the GPI theorem states that $\pi(a|s)=\ind{a=a^*(s,\wnew)}$
will perform at least as well as any of the existing policies,
i.e., $Q^{\pi}(s,a) \geq \max_i Q^{\pi_i}(s,a)$
(c.f.,  policy improvement in \cref{sec:policyImprovement}).
See \cref{fig:successorFeatures} for an illustration.

Note that GPI is a model-free approach to computing a new policy,
based on an existing library of policies.
In \citep{Alegre2023}, they propose an extension that can also
leverage a (possibly approximate) world model to learn
better policies that can outperform the library
of existing policies by performing more decision-time search.

\subsubsection{Option keyboard}

One limitation of GPI is that it requires that the reward
function, and the resulting policy, be defined in terms of a fixed
weight vector $\wnew$, where the preference over features
is constant over time. However, for some tasks we might
want to initially avoid a feature or state and then later move
towards it.
To solve this, 
\citep{Barreto2019,Barreto2020} introduced the \keywordDef{option keyboard},
in which the weight vector for a task can be computed
dynamically in a state-dependent way, using
$\vw_s = g(s,\wnew)$.
(Options are discussed in \cref{sec:options}.)
Actions can then be chosen as follows:
\begin{align}
  a^*(s;\wnew) &=
    \argmax_a \max_i \vpsi^{\pi_i}(s,a)^\trans \vw_s
\end{align}
Thus $\vw_s$ induces a set of policies
that are active for a period of time, similar to playing
a chord on a piano.


\subsubsection{Learning SFs}

A key question when using SFs is how to learn the cumulants
or state-features
$\vphi(s)$.
Various approaches have been suggested,
including leveraging meta-gradients \citep{Veeriah2019};
image reconstruction \citep{Machado2018eigen};
maximizing the mutual information between task encodings
and the cumulants that an agent experiences when pursuing that task
\citep{Hansen2019};
and reward prediction methods
\citep{Chua2024succ}.
The cumulants are encouraged to satisfy the linear reward
constraint by minimizing
\be
\loss_r = ||r-\vphi_{\vtheta}(s)^\trans \vw||_2^2
\ee

Once the cumulant function is known,
we have to learn the corresponding SF.
The standard approach learns a different SF for every policy,
which is limiting. In \citep{Borsa2019} they introduced
\keywordDef{Universal Successor Feature Approximators}
which takes as input a policy encoding $\vz_{\vw}$,
representing a policy $\pi_{\vw}$ (typically we set $\vz_{\vw}=\vw$).
We then define
\be
\vpsi^{\pi_{\vw}}(s,a) = \vpsi_{\vtheta}(s,a,\vz_{\vw})
\ee
The GPI update then becomes
\be
a^*(s;\wnew) = \argmax_a \max_{\vz_{\vw}} \vpsi_{\vtheta}(s,a,\vz_{\vw})^\trans \wnew
\ee
so we replace the discrete  over a finite number of policies,
$\max_i$,
with a continuous optimization problem $\max_{z_w}$,
to be solved per state.

If we want to learn the policies and SFs at the same time,
we can optimize the following losses in parallel:
\begin{align}
  \loss_Q &= ||\vpsi_{\vtheta}(s,a,\vz_w)^\trans \vw - \vy_Q||,
  \;
  \vy_Q = R(s';\vw) + \gamma \vpsi_{\vtheta}(s',a^*,\vz_w)^\trans \vw
  \label{eqn:LQ} \\
  \loss_{\vpsi} &= ||\vpsi_{\vtheta}(s,a,\vz_{\vw}) - \vy_{\vpsi}||,
  \;
  \vy_{\vpsi} = \vphi(s') + \gamma \vpsi_{\vtheta}(s',a^*,\vz_{\vw})
\end{align}
where $a^* = \argmax_{a'} \vpsi_{\vtheta}(s', a', \vz_{\vw})^\trans \vw$.
The first equation is standard Q learning loss,
and the second is the TD update rule
in \cref{eqn:SFTD} for the SF.
In \citep{Carvalho2023}, they present the
\keywordDef{Successor Features Keyboard},
that can learn the policy,
the SFs 
and the task encoding $\vz_{\vw}$,
all simultaneously.
They also suggest replacing the squared error regression
loss in \cref{eqn:LQ} with a cross-entropy loss,
where each dimension of the SF is now a
discrete probability distribution
over $M$ possible values of the corresponding feature.
(c.f. \cref{sec:classifRL}).

\subsubsection{Choosing the tasks}

A key advantage of SFs is that they provide a way
to compute a value function and policy for any given reward,
as specified by a task-specific weight vector $\vw$.
But how do we choose these tasks?
In \citep{Hansen2019} they sample $\vw$ from a distribution
at the start of each task, to encourage the agent
to learn to explore different parts of the state space
(as specified by the feature function $\vphi$).
In \citep{Liu2021} they extend this by adding an intrinsic
reward that favors exploring parts of the state space
that are surprising (i.e., which induce high entropy),
c.f., \cref{sec:intrinsicReward}.

\subsection{Successor measures}
\label{sec:SM}

In this section, we discuss the \keywordDef{successor model}
(also called a $\gamma$-model, or \keywordDef{geometric horizon models}),
which is a probabilistic extension of SR,
\citep{Janner2020}:
Rather than just working with expectations,
we can simulate future state trajectories by sampling.
This allows us
to generalize SR to work with continuous states and actions,
as we will see.

The basic idea is to define the cumulant as the $k$-step conditional
distribution $C(s_{k+1}) = P(s_{k+1}=\tilde{s}|s_0=s,\pi)$,
which is the probability of being in state $\tilde{s}$
after following $\pi$ for $k$ steps starting from
state $s$. (Compare this to the SR cumulant,
which is $C(s_{k+1})=\ind{s_{k+1}=\tilde{s}}$.)
The SM is then defined as
\be
\vmu^{\pi}(\tilde{s}|s)
=(1-\gamma) \sum_{t=0}^{\infty} \gamma^t P(s_{t+1}=\tilde{s}|s_0=s)
\ee
where the $1-\gamma$ term ensures that $\vmu^{\pi}$ integrates to 1.
(Recall that $\sum_{t=0}^{\infty} \gamma^t = \frac{1}{1-\gamma}$ for $\gamma<1$.)
In the tabular setting, the SM is just the normalized SR,
since
\begin{align}
\vmu^{\pi}(\tilde{s}|s)
&=(1-\gamma) M^{\pi}(s,\tilde{s}) \\
&=(1-\gamma) \expect{\sum_{t=0}^{\infty} \gamma^t \ind{s_{t+1}=\tilde{s}}
  |s_0=s, a_{0:\infty} \sim \pi} \\
&= (1-\gamma) \sum_{t=0}^{\infty} \gamma^t P(s_{t+1}=\tilde{s}|s_0=s, \pi)
\end{align}
Thus $\vmu^{\pi}(\tilde{s}|s)$
tells us the probability that $\tilde{s}$ can be reached from $s$
within a horizon determined by $\gamma$ when  following $\pi$,
even though we don't know exactly when we will reach $\tilde{s}$.

SMs obey a Bellman-like recursion
\be
\vmu^{\pi}(\tilde{s}|s) = \expect{
  (1-\gamma) T(\tilde{s}|s,a) + \gamma \vmu^{\pi}(\tilde{s}|s')}
\ee
We can use this to perform policy evaluation by computing
\be
V^{\pi}(s) = \frac{1}{1-\gamma}
\expectQ{R(\tilde{s})}{\vmu^{\pi}(\tilde{s}|s)}
\ee

We can also define an action-conditioned SM
\begin{align}
\vmu^{\pi}(\tilde{s}|s,a)
&=(1-\gamma) \sum_{t=0}^{\infty} \gamma^t P(s_{t+1}=\tilde{s}|s_0=s,a_0=a) \\
&=(1-\gamma) T(\tilde{s}|s,a) + \gamma \expect{\vmu^{\pi}(\tilde{s}|s',a',\pi)}
\end{align}
Hence we can learn an SM using a TD update of the form
\begin{align}
  \vmu^{\pi}(\tilde{s}|s,a) &\leftarrow
  \vmu^{\pi}(\tilde{s}|s,a) + \eta 
  \underbrace{\left( (1-\gamma) T(s'|s,a)
    + \gamma \vmu^{\pi}(\tilde{s}|s',a')
    - \vmu^{\pi}(\tilde{s}|s,a) \right)}_{\delta}
  \label{eqn:SMQTD}
\end{align}
where $s'$ is the next state sampled from $T(s'|s,a)$
and $a'$ is the next action sampled from $\pi(s')$.
\eat{
Compare this to the Q-learning update
\be
Q^{\pi}(s,a) \leftarrow
Q^{\pi}(s,a) + \eta \underbrace{ \left(
  R(s') + \gamma Q^{\pi}(s',a') - Q^{\pi}(s,a) \right)}_{\delta}
\ee
}
With an SM, we can  compute the state-action value
for any reward:
\be
Q^{R,\pi}(s,a) = \frac{1}{1-\gamma}
\expectQ{R(\tilde{s})}{\vmu^{\pi}(\tilde{s}|s,a)}
\ee
This can be used to improve the policy as we discuss
in \cref{sec:GPI}.

\subsubsection{Learning SMs}

Although we can learn SMs
using the TD
update in \cref{eqn:SMQTD},
this requires evaluating
$T(s'|s,a)$ to compute the target update $\delta$,
and this one-step transition model is typically unknown.
Instead, since $\vmu^{\pi}$ is a conditional density model,
we will optimize
the cross-entropy TD loss
\citep{Janner2020}, defined as follows
\be
\loss_{\mu} = \expectQ{\log \vmu_{\vtheta}(\tilde{s}|s,a)}
     {(s,a) \sim p(s,a),
       \tilde{s} \sim (T^{\pi} \vmu^{\pi})(\cdot|s,a)}
\ee
where $(T^{\pi} \vmu^{\pi})(\cdot|s,a)$ is the Bellman
operator applied to $\vmu^{\pi}$ and then evaluated
at $(s,a)$, i.e.,
\be
(T^{\pi} \vmu^{\pi})(\tilde{s}|s,a)
=(1-\gamma) T(s'|s,a) + \gamma \sum_{s'}
T(\tilde{s}|s,a) \sum_{a'} \pi(a'|s') \vmu^{\pi}(\tilde{s}|s',a')
\ee
We can sample from this as follows:
first sample $s' \sim T(s'|s,a)$ from the environment
(or an offline replay buffer),
and then
with probability $1-\gamma$ 
set $\tilde{s}=s'$ and terminate.
Otherwise sample $a' \sim \pi(a'|s')$
and then create a bootstrap sample
from the SM using $\tilde{s} \sim \vmu^{\pi}(\tilde{s}|s',a')$.

There are many possible density models we can use for
$\vmu^{\pi}$.
In \citep{Thakoor2022}, they use a VAE.
In \citep{Tomar2024}, they use
an autoregressive transformer applied
to a set of  discrete
latent tokens, which are learned using VQ-VAE or
a non-reconstructive self-supervised loss.
They call their method \keywordDef{Video Occcupancy Models}.
Recently, \citep{Farebrother2025} proposed to use
diffusion (flow matching) to learn SMs.

An alternative approach to learning SMs, that avoids fitting a normalized
density model over states, is to use contrastive learning
to estimate how likely $\tilde{s}$ is to occur
after some number of steps, given $(s,a)$,
compared to some randomly sampled negative state
\citep{Eysenbach2021C,Zheng2024contrastive}.
Although we can't sample from the resulting learned model
(we can only use it for evaluation),
we can use it to improve a policy that achieves
a target state (an approach known as
goal-conditioned policy learning,
discussed in \cref{sec:GCRL}).

\subsubsection{Jumpy models using  geometric policy composition}
\label{sec:jumpy}

In \citep{Thakoor2022}, they propose
\keywordDef{geometric policy composition} or GPC
as a way to learn a new policy by sequencing together
a set of $N$ policies, as opposed to taking $N$ primitive actions
in a row.
This can be thought of as a \keywordDef{jumpy model}, since it predicts
multiple steps into the future, instead of one step at a time
(c.f., \citep{Zhang2023jumpy}).

In more detail, in GPC, the agent picks a sequence of $n$
policies $\pi_i$ for $i=1:n$, and then samples states
according to their corresponding SMs:
starting with $(s_0,a_0)$, we sample
$s_1 \sim \vmu_{\gamma}^{\pi_1}(\cdot|s_0,a_0)$,
then $a_1 \sim \pi_1(\cdot | s_1)$,
then
$s_2 \sim \vmu_{\gamma}^{\pi_2}(\cdot|s_1,a_1)$,
etc.
This continues for $n-1$ steps.
Finally we sample $s_n \sim \vmu_{\gamma'}^{\pi_n}(\cdot|s_{n-1},a_{n-1})$,
where $\gamma' > \gamma$ represents a longer horizon SM.
The reward estimates computed along this sampled path can then
be combined to compute the value of each candidate policy sequence.

\subsubsection{Other related work}

In \citep{Farebrother2023}, they introduce
\keywordDef{proto-value networks},
which is a way to define auxiliary tasks based on  successor
measures.

In \citep{Touati2021,Touati2023},
they propose the \keywordDef{forwards-backwards}
representations,
which provides a general framework for learning SMs.

\subsection{Connection between options and successor representations}
\label{sec:SRoptions}

In \cref{sec:options}, we discuss \keywordDef{options},
which are temporally extended actions.
These are closely related to the successor representation
\citep{Machado2023}.
Indeed, the eigenvectors of the SR matrix
$\vM^{\pi}$
correspond to the eigenvectors of the
\keywordDef{graph Laplacian}, defined as
\be
\vL = \vD^{-\half} (\vD-\vA) \vD^{-\half}
\ee
where $\vA$ is the adjacency matrix
corresponding to $T^{\pi}$,
and $\vD$ is a diagonal matrix whose entries are the row-sums of $\vA$.
These eigenvectors are known as \keywordDef{proto value functions}
\citep{Mahadevan2007},
or \keywordDef{eigen-options} \citep{Machado2018eigen}.

We can use this connection to define a ``universal'' form of successor
representation that is independent of a specific policy.
Rather than constructing an
option for every eigenvector of the graph Laplacian,
a single option  based on the
second eigenvector is sufficient.
This is called
a \keywordDef{covering option}
\citep{Jinnai2019,Jinnai2020,Klissarov2023},
since it minimizes the cover time of
the underlying MDP, which loosely refers to
how long it takes for a random high-level policy
to visit all states.


\eat{
\subsection{Predictive state representations}
\label{sec:PSR}

An alternative to working with belief states
is to  marginalize out the POMDP
latent state $z_t$,
to derive a set of predictions over future observables $\vo_{t+1:T}$
as a function of the past history of observables, $\vh_t$,
and future actions, $\va_{t+1:T}$.
This can then become a learning  target for a model, that is trained
to directly
predict  future observations, without explicitly invoking the concept
of latent state.
This is called a \keywordDef{predictive state representation}
or \keywordDef{PSR} \citep{PSR}.
This is related to the idea of
\keywordDef{observable operator models}
\citep{Jaeger2000}.
See also \cref{sec:SR} where we discuss successor representations.
}

\eat{
\subsection{General value functions}
\label{sec:GVF}

So far we have mainly been focused on predicting the sum of expected
discounted future rewards, i.e., the value function.
But the reward is just one possible signal of interest we can extract from the environment.
We can generalize this by considering a \keywordDef{cumulant}
$C_t \in \real$, which is some scalar of interest derived from the observation
(e.g., did a loud bang just occur? is there a tree visible in the image?).
We then define the \keywordDef{general value function} or \keywordDef{GVF}
as follows \citep{GVF}:
\be
V_{\pi,\gamma,C}(s) = \expect{
  \sum_{k=t}^{\infty} \left(\prod_{i=t+1}^k \gamma(S_i) \right)
  C_{k+1} \vert S_t=s, A_{t:\infty} \sim \pi}
\ee
where $1-\gamma(s)$ is the probability of terminating in state $s$.
If we set $C(s)=R(s)$ and $\gamma(s)=\gamma$, we recover the value function.
But we can use the GVF to predict the probability of
other \keywordDef{auxiliary tasks}, such as the occurence of a loud bang;
and unlike a standard one-step predictive model,
we don't need to know exactly when the event will occur.
(See also \citep{Ring2021}.)

By training a model to compute GVFs for multiple related cumulants,
we can  learned a useful state representation more efficiently than just
learning from the reward by itself, while still avoiding the need
to reconstruct the entire observation.
Thus this is a form of non-generative world model,
as discussed in \cref{sec:MBRLnongen}.
The benefits of this kind of approach were shown in
\citep{Jaderberg2017iclr}.

Note that training GVFs using some kind of TD learning
is inherently an off-policy learning problem, since the data is being
collected by a policy to maximize reward, but the other cumulants
are just being predicted, not maximized.
Since the GVF is typically a neural network with multiple output heads,
we need to tackle the \keyword{deadly triad}, which we discussed in 
\cref{sec:deadlyTriad}.
One approach is to use gradient TD methods, which are provably convergent.
This has been applied to GVF learning in \citep{Xu2022GVF}.
However, the question of which cumulants are worth learning is still open.
}

\chapter{Multi-agent RL}
\label{sec:MARL}
\label{chap:MARL}



In this section, we give a brief introduction to \keywordDef{multi-agent RL}
or \keywordDef{MARL}. Our presentation is based on 
\citep{marl-book}.
MARL is closely related to game theory
(see e.g. \citep{Leyton-Brown2008})
and multi-agent systems design 
(see e.g. \citep{Shoham2008}),
as we will see.
For other surveys on MARL, see
e.g. \citep{Hernandez-Leal2019,Yang2020MARL,Wong2022,Gronauer2022}.

\section{Games}

\begin{figure}
\centering
\includegraphics[height=2in]{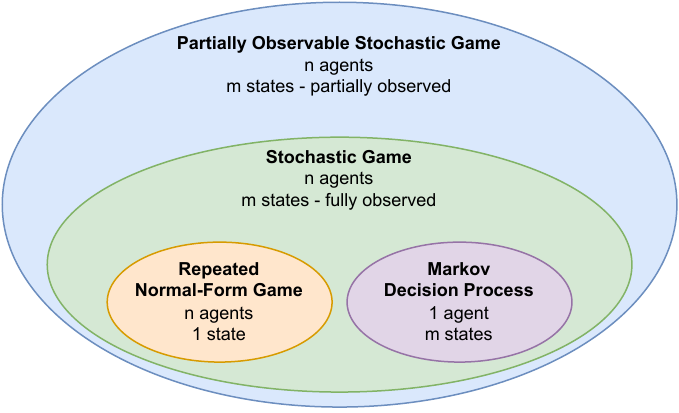}
\caption{
  Hierarchy of games.
  \figtaken{Fig 3.1 of \citep{marl-book}.}
  \figthanks{Stefano Albrecht}.
  }
\label{fig:games}
\end{figure}

Multi-agent environments are often called
\keywordDef{games}, even if they represent ``real-world'' problems such as multi-robot
coordination (e.g., a fleet of autonomous vehicles) or agent-based trading.
In this section, we discuss different kinds of games that have been proposed,
summarized in \cref{fig:games}.

In the \keywordDef{game theory} community, the rules of the game
(i.e., the environment dynamics, and the reward function, aka
\keywordDef{payoff function})
are usually assumed known,
and the focus is on computing \keywordDef{strategies}
(i.e., policies) for each \keywordDef{player} (i.e., agent),
whereas in MARL, we usually assume the environment is unknown and the agents have to learn
just by interacting with it. (This is analogous to the distinction between DP methods,
that assume a known MDP, and RL methods, that just assume sample-based access to the MDP.)

\subsection{Normal-form games}

A \keywordDef{normal-form game} defines a single interaction between $n \geq 2$ agents.
In particular, we have a
finite set of agents $\calI = \{1,\ldots,n\}$
(we assume that $n$ is fixed).
For each agent $i \in \calI$ we have
a finite set of actions $\calA_i$ and a reward function
$\calR_i: \calA_{1:n} \ra \real$,
where $\calA_{1:n} = \calA_1 \times \cdots \times \calA_n$.
A single round of the game proceeds as follows.
Each agent samples an action $a_i \in \calA_i$ with probability
$\pi_i(a_i)$, then the resulting
\keywordDef{joint action} $\va=(a_1,\ldots,a_n)$ is taken
and the reward $\vr=(r_1,\ldots,r_m)$ is given to each player,
where $r_i = \calR_i(\va)$.

Games can be classified based on the type of rewards they contain.
In \keywordDef{zero-sum games}, we have $\sum_i \calR_i(\va)=0$ for all $\va$.
(For a two-player zero-sum game, \keywordDef{2p0s},
we must have $R_1(\va)=-R_2(\va)$.)
In \keywordDef{common-payoff games} (aka
common-reward games),
we have $\calR_i(\va)=\calR_j(\va)$ for all $\va$.
And in \keywordDef{general-sum games}, there are no restrictions on
the rewards.

In zero-sum games, the agents must compete against each other,
whereas in common-reward games, the agents generally must cooperate
(although they may compete with each other over a shared resource).
In general-sum games, there can be a mix of cooperation and competition.
Although common-reward games can be easier to solve
than general-sum games,
it can be challenging to disentangle the contribution of each agent
to the shared reward (this is a multi-agent version of the
\keywordDef{credit assignment} problem),
and coordinating actions across agents can also be difficult.

Normal-form games with 2 agents are called \keywordDef{matrix games} because
they can be defined by a 2d reward matrix.
We give some well-known examples
in \cref{tab:matrix-games}.

\begin{itemize}
\item In \keywordDef{rock-paper-scissors}, Rock can blunt scissors,
      Paper can cover rock, and Scissors can cut paper;
      from these constraints, we can determine which player wins or loses.
      This is a zero-sum game.
      

    \item       In the \keywordDef{battle of the sexes} games, a male-female couple want
      to choose a shared activity. They both of have different individual preferences
      (eg row player prefers Opera, column player prefers Football),
      both they would both rather spend time together than alone.
      This is an example of a \keywordDef{coordination game}.

\item       In the \keywordDef{Prisoner's dilemma}, which is a general-sum
      game,
      the players (who are prisoners
      being interogated independently in different cells)
      can either cooperate
      with each other (by both ``staying mum'', i.e., denying they committed the crime),
      or one can defect on the other (by claiming the other person committed the crime).
      If they both cooperate, they only have to serve 1 year in jail each,
      based on weak evidence. 
      If they both defect, they each serve 3 years.
      But if the row player cooperates (stays silent) and the column
      player defects (implicates his partner), 
      the row player gets 5 years and the column player gets out of
      jail free. This leads to an incentive for both players to
      defect,       even though they would be better off if they both
      cooperated. We discuss this example in more detail in \cref{sec:nash}.
\end{itemize}

\begin{table}[h]
    \centering
    \begin{subtable}{0.32\textwidth}
        \centering
        \begin{tabular}{c|ccc}
            & R & P & S \\
            \hline
            R & 0,0  & -1,1 & 1,-1 \\
            P & 1,-1 & 0,0  & -1,1 \\
            S & -1,1 & 1,-1 & 0,0 
        \end{tabular}
        \caption{Rock-Paper-Scissors}
    \end{subtable}
    \hfill
    \begin{subtable}{0.32\textwidth}
        \centering
        \begin{tabular}{c|cc}
            & O & F \\
            \hline
            O & 2,) & 0,0 \\
            F & 0,0  & 1,2
        \end{tabular}
        \caption{Battle of the sexes}
    \end{subtable}
    \hfill
    \begin{subtable}{0.32\textwidth}
        \centering
        \begin{tabular}{c|cc}
            & C  & D  \\
            \hline
            C & -1,-1 & -5,0 \\
            D & 0,5   & -3,-3
        \end{tabular}
        \caption{Prisoner’s Dilemma}
    \end{subtable}
    \caption{Three different matrix games.
      Here the notation $(x, y)$ in cell $(i,j)$
      refers to the row player receiving $x$ 
      and the column player receiving $y$
      in response to the joint action $(i,j)$.
      }
    \label{tab:matrix-games}
\end{table}

Suppose we consider matrix games with just 2 actions each.
In this case, we can represent the game as follows:
\begin{align}
  \begin{pmatrix}
    a_{11}, b_{11} &  a_{12}, b_{12} \\
    a_{21}, b_{21} & a_{22}, b_{22}
  \end{pmatrix}
\end{align}
where $a_{ij}$ is the reward to player 1 (row player)
and $b_{ij}$ is the reward to player 2 (column player)
if player 1 picks action $i$ and player 2 picks action $j$.
Suppose we further
restrict attention to strictly ordinal games,
meaning that each agent ranks the 4 possible outcomes from 1 (least
preferred) to 4 (most preferred)).
In this case, there are 78 structurally distinct games
\citep{Rapoport1966}.
These can be grouped into two main kinds.
In \keywordDef{no-conflict games}, both players have the same set
of most preferred outcomes, whereas in
\keywordDef{conflict games}, the players disagree
about what is best.
If we consider general ordinal $2 \times 2$ games
(where one or both players may have equal preference for
two or more outcomes),
we find that there are 726 of them
\citep{Kilgour1988}.

A \keywordDef{repeated matrix game}
is the multi-agent analog of a multi-armed bandit problem,
discussed in \cref{sec:bandits}.
In this case, the policy has the form $\pi_i(a_t^i|\vh_t)$,
where $\vh_t=(\va_0,\ldots,\va_{t-1})$ is the history of joint-actions.
In some cases, the agent may choose to ignore the history, or only
look at the last $n$ joint actions.
For example, in the \keywordDef{tit-for-tat} strategy in the prisoner's
dilemma, the policy for agent $i$ at step $t$ is to do the same action
that agent $-i$ did at step $t-1$ (where $-i$ means the agent other
than $i$), so the policy is conditioned on $\va_{t-1}$.
(Note that this strategy will punish players who defect,
and can lead to the evolution of cooperative behavior,
even in selfish agents \citep{Axelrod1981,Axelrod1984}.)

\subsection{Stochastic games}

\begin{figure}
\centering
\includegraphics[height=2in]{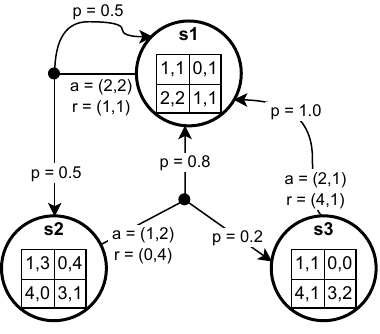}
\caption{
  Example of a two-player general-sum stochastic game.
  Circles represent the states, inside of which we show the reward
  function in matrix form.
  Only one of the 4 possible transitions out of each state are shown.
  The little black dots are called the \keywordDef{after states},
  and correspond to an intermediate point where a joint action has
  been   decided by the players, but nature hasn't yet sampled the
  next  transition, which occurs with the specified probabilities.
  \figtaken{Fig 3.3(b) of \citep{marl-book}.}
  \figthanks{Stefano Albrecht}.
  }
\label{fig:stochastic-game}
\end{figure}

A \keywordDef{stochastic game}
is a multi-agent version of an MDP,
and was first proposed in \citep{Shapley1953}.
It is defined by a
finite set of agents $\calI = \{1,\ldots,n\}$;
a finite set of states $\calS$, of which a subset $\overline{S} \subset
\calS$ are terminal;
a finite action set $\calA_i$ for each agent $i \in \calI$;
a reward function $\calR_i(s,a,s')$ for each agent
$i \in \calI$\footnote{
Here $R(s,a,s')$ is the reward we receive if we take
action $a$ in state $s$ and end up in $s'$.
As explained in \citep[Sec 2.8]{marl-book}, we can convert
from $R(s,a,s')$ to the more common $R(s,a)$ notation,
representing the expected reward,
by noting that $R(s,a)=\sum_{s'} \calT(s'|s,a) R(s,a,s')$.
};
a state transition distribution $\calT(s_{t+1}|s_{1:t},\va_t) \in [0,1]$;
and an initial state distribution $\mu(s_0) \in [0,1]$.
Typically the transition distribution is Markovian
(i.e.,  $\calT(s_{t+1}|s_{1:t},\va_t) =  \calT(s_{t+1}|s_{t},\va_t)$,
in which case this is called a \keywordDef{Markov game} \citep{Littman1994}.)
See \cref{fig:stochastic-game} for an example.

The policy for each agent in such a game has the form
$\pi_i(a_t^i|\vh_t)$ where
$\vh_t=(s_0,\va_1,\ldots,s_t)$ is the state-action history.
(We omit rewards from the definition of history for notational simplicity.)
The overal \keywordDef{joint policy} is denoted by
$\vpi=(\pi_1,\ldots,\pi_n)$;
if the agents make their decisions independently (which we assume),
then this has the form
\be
\vpi(\va_t|\vh_t) = \prod_i \pi_i(a_t^i|\vh_t)
\ee
Often we assume the policies are Markovian, in which case
they can be written as $\pi_i(a_t^i|s_t)$.

Note that, from the perspective of each agent $i$,
the environment transition function has the form
\be
\calT_i(s_{t+1}|s_t,a_t^i) =
\sum_{\va_t^{-i}} \calT(s_{t+1}|s_t,(a_t^i,\va_t^{-i}))
\prod_{j \neq i} \pi_j(a_t^j|s_t)
\label{eqn:localTrans}
\ee
Thus $\calT_i$ depends on the policies of the other players,
which are often changing,
which makes these local/agent-centric transition matrices
non-stationary, even if the underlying environment is stationary.
Typically agent $i$ does not know the policies
of the other agents $j$, so it has to learn them,
or it can just treat the other agents as part of the environment
(i.e., as another  source of unmodeled ``noise'')
and then use single agent RL methods
(see \cref{sec:independentMARL}).

\subsection{Partially observed stochastic games (POSG)}
\label{sec:POSG}

A \keywordDef{Partially Observed Stochastic Game} or \keywordDef{POSG}
is a multi-agent version of a POMDP
(See e.g., \citep{Hansen2004}.)
We augment the stochastic game with the observation distributions
$\calO_i(o_{t+1}^i|s_{t+1},\va_{t}) \in [0,1]$ for each agent $i$.
(Alternatively, the $i$'th observation distribution may just depend
on $i$'s actions.)
Let $\vo_t = (o_t^1, \ldots, o_t^n)$ be the \keywordDef{joint
  observation}
generated by the product distribution
$\calO_{1:n}(\vo_t|s_t,\va_{t-1})$.
The policy for each agent in such a game has the form
$\pi_i(a_t^i|h_t^i)$ where
$h_t^i=(o_0^i, a_0^i, o_1^i, a_i^i, \ldots,o_t^i)$
is the \keywordDef{action observation  history}
for agent $i$,
and $\vh_t = h_t^{1:n}$ is the joint observation history.
(Note that the environment decides what is included in each
observation;
for example, it may or may not contain information about the other
agent's actions.)
Note that a \keywordDef{Decentralized POMDP} or \keywordDef{Dec-POMDP}
is a special case of a POMDP where the reward function is the same
for all agents (thus  it can only capture cooperative behavior).
See \citep{Oliehoek2016} for more details.

\subsubsection{Data generating process}

The data generating process for a POSG proceeds as follows.
First the environment samples an initial state from $\mu(s_0)$
and generates an initial observation from $\calO^0_{1:n}(\vo_0|s_0)$.
Then for $t=0,1,\ldots$, we repeat the following
\begin{enumerate}
\item Each agent  generates action from $\pi_i(a_t^i|h_t^i)$.
  
  \item Environment generates next state from
    $\calT(s_{t+1}|s_t,\va_t)$.
    
  \item Environment generates observation
    from $\calO_i(o_{t+1}^i|s_{t+1},\va_{t})$ for each $i$.
    
  \item Environment generates  reward from
    $\calR_i(s_t,\va_t,s_{t+1})$
    for each $i$.
    (For simplicity, we often just use
    $\calR_i(s_t,\va_t)$, which we define
    as $\E_{s_{t+1}}[\calR_i(s_t,\va_t,s_{t+1})]$.)
    
  \item Each agent updates its history
    using $h_{t+1}^i = f_i(h_{t}^i, a_{t}^i, o_{t+1}^i)$,
    where $f_i$ is agent $i$'s update function (e.g.,
    list concatenation, or an RNN).
  \end{enumerate}

\subsubsection{Objective}

We define the sum of rewards as $G = \sum_t \calR_i(S_t,\vA_t)$,
where we use capital letters for random variables,
and bold face for everything that is joint across all agents.
The objective of player $i$ is to maximize
$\calJ_i(\pi_i) = \E_{\pi_i}[G]$.
We can compute this using Bellmans equations, as follows.
First define the expected state value for agent $i$
under joint policy $\vpi$ given joint history $\vh_t$
as
\be
v_i^{\vpi}(\vh_t) = \E_{\vpi}[G_{\geq t}|\vh_t]
= \E_{\vpi}[\sum_{t' \geq t} \calR_i(S_{t'},\vA_{t'}) | \vh_t]
\ee
The expected state value for agent $i$
under the joint policy $\vpi$ and its local
history $h_t^i$ is
\be
v_i^{\vpi}(h_t^i) = \E_{\vpi}[v_i^{\vpi}(\vH_t)|h_t^i]
\ee
Similarly define the expected state-action value
given the joint history as
\be
q_i^{\vpi}(\vh_t,a_t^i) =
\E_{\vpi}[G_{\geq t}|\vh_t,a_t^i]
=E_{\vpi}[\calR_i(S_t,\vA_t) +
  v_i^{\vpi}(\vH_{t+1}) | \vh_t, a_t^i]
\ee
and the expected state-action value given the local history as
\be
q_i^{\vpi}(h_t^i,a_t^i) = \E_{\vpi}[q_i^{\vpi}(\vH_t,a_t^i)|h_t^i]
\ee

\subsubsection{Single agent perspective}

From the perspective of agent $i$, it just observes a sequence of
observations generated by the following 
``sensor stream distribution'',
(which is non-Markovian \citep{Laurent2011}):
\begin{align}
p_i(o_{t+1}^i | \vh_t^i, a_t^i)
  &= \sum_{s_{t+1}} \sum_{\va_t^{-i}}
\hat{\calO}_i(o_{t+1}^i | s_{t+1},  \va_t)
p_i(\va_{t}^{-i} | \vh_t^i) 
p_i(s_{t+1} | \vh_t^i, \va_t)
\label{eqn:localObsModel}
\\
p_i(\va_{t}^{-i} | \vh_t^i)
&= \prod_{j \neq i} \hat{\pi}_i^j(a_t^j|\vh_t^i)
\label{eqn:agentModel}
\\
 p_i(s_{t+1} | \vh_t^i, \va_t)
 &= \sum_{s_t}   \hat{\calT}_i(s_{t+1} | s_t, \va_t)  b_i(s_t|\vh_t^i) 
  \end{align}
where $\hat{\pi}_i^j$
in \cref{eqn:agentModel} is $i$'s estimate  of $j$'s policy;
$\hat{\calT}_i$ is $i$'s estimate of $\calT$ based on $\vh_t^i$;
$\hat{\calO}_i$ is $i$'s estimate of $\calO_i$
 based on $\vh_t^i$;
and
$b_i(s_t|\vh_t^i)$ is $i$'s \keywordDef{belief state}
(i.e., its posterior distributiion
over the underlying latent state given its local observation history).
The agent can either learn a policy given this ``collapsed'' representation,
treating the other agents as part of the environment,
or it can explicitly try to learn the true
joint world model $\calT$, local observation
model $\calO_i$ and other agent policies
$\pi_i^j$, so it can reason about the other agents.
In this section, we follow the latter approach.

\subsubsection{Factored Observation Stochastic Games (FOSG)}
\label{sec:FOSG}

\citep{Kovarik2022}
propose a formalism called
\keywordDef{Factored Observation Stochastic Games}
or \keywordDef{FOSG} that extends POSGs  by partitioning
the observation for each player into public and private.\footnote{
Note that this kind of factorization
is different from  factoring
the state vector or reward function;
the latter is called a factored POSG.
}
(We say that information is public if it is visible
to all players, and all players know this;
thus it is a form of \keywordDef{common knowledge}.)
Explicitly distinguishing these two kinds of information
is important in order to tractably solve certain kinds of games,
like Poker or Hanabi
(see e.g., \citep{Sokota2021}).

\subsection{Extensive form games (EFG)}
\label{sec:EFG}

In the game theory literature, it is common to use
the \keywordDef{extensive form game} representation.
Rather than representing a sequence of world states that evolve
over time, we represent a tree of possible choices or actions
take by each player (and optionally a \keywordDef{chance player},
if the game is
stochastic, e.g., backgammon).
 Each  node represents a unique sequence (history) of
 actions leading up to that point.

In the context of EFGs, some additional terminology is commonly used.
If all the nodes are observed (including chance nodes),
we say the game has perfect and complete information.
If the moves of some  players
are not visible and/or the state of the game is not fully known
(e.g., poker), the game has \keywordDef{imperfect information}.
In this case, 
we define an \keywordDef{information set} as the  set of nodes
that an agent cannot distinguish between.
This is analogous to having a distribution over the hidden
states in a POSG.

If an agent does not know the other player's type or payoff function
(e.g. in an auction, or playing against players with unknown skill level),
then the game has \keywordDef{incomplete information}.
In this case, the agent should maintain a Bayesian belief state about
the unknown factors.
This is analogous to having a distribution over the
parameters of the POSG itself,
similar to  a multi-agent version of 
a Bayes Adaptive POMDP
\citep{Ross2007}.

Note that in theoretical work, a useful result
is it is possible to convert any EFG into an equivalent (stateless) NFG,
where
the actions of the  NFG correspond to the deterministic policies
of the EFG, and the payoffs for a joint action
are the expected returns of the corresponding joint policy in the EFG.

\subsubsection{Example: Kuhn Poker as EFG}
\label{sec:kuhnPoker}

\begin{figure}
\centering
\includegraphics[height=3in]{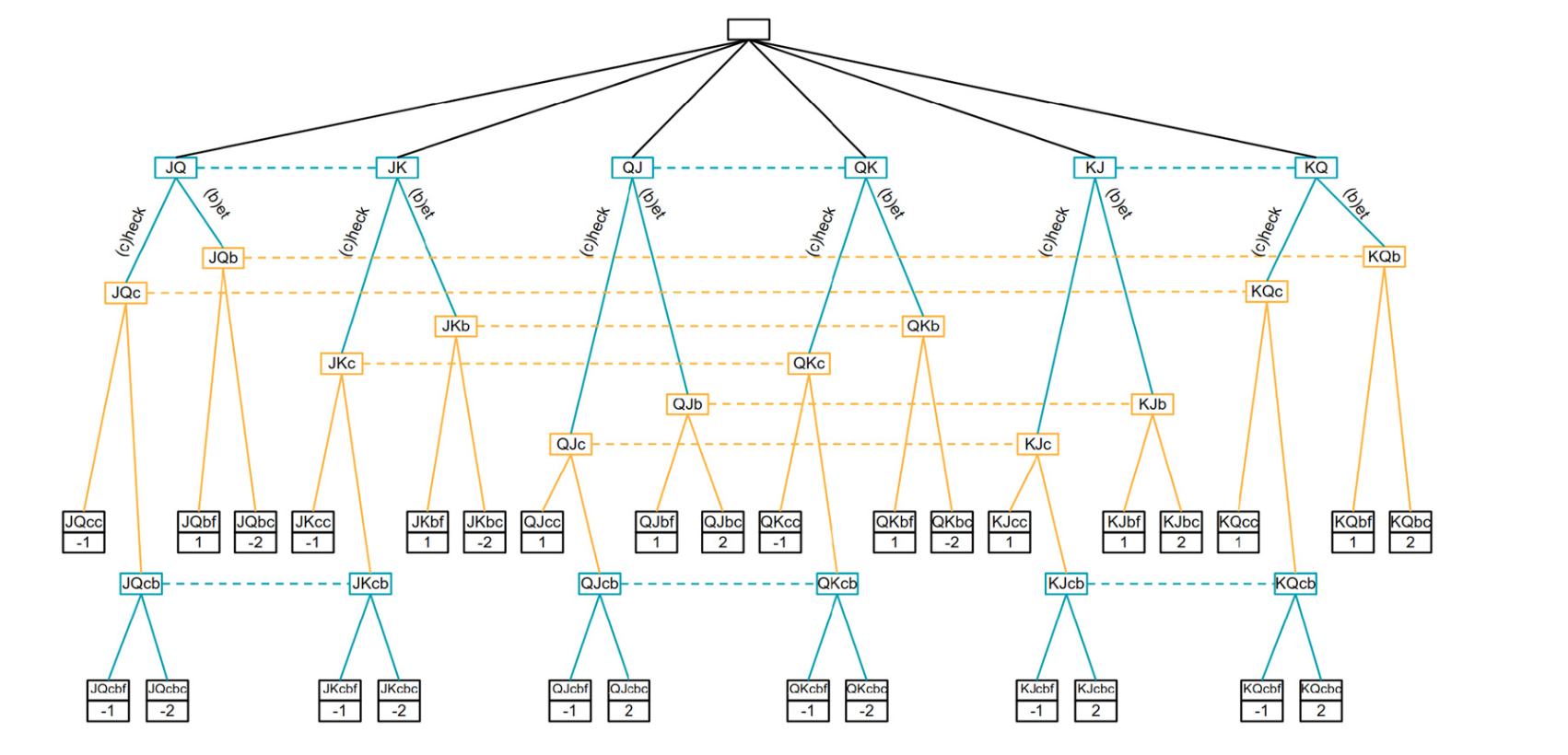}
\caption{
  EFG for Kuhn Poker.
  The dashed lines connect histories in the same information set.
  \figtaken{Fig 2 of \citep{Kovarik2022}}.
  \figthanks{Viliam Lisy}.
  }
\label{fig:EFG}
\end{figure}

In this section, we give an example of an EFG formulation
of the game of \keywordDef{Kuhn Poker},
introduced in \citep{Kuhn1951}.
We first define the rules of the game, following  \citep{Kovarik2022}:

\begin{quote}
Kuhn poker is a form of (two player)
poker where the deck includes only three
cards: Jack, Queen, and King.
First, each player places one chip into the pot as the initial
forced bet (ante).
Each player is then privately
dealt one card (the last card isn’t revealed).
This is followed by a betting phase (explained below).
The game ends either
when one player folds (forfeiting all bets made so far to their
opponent) or there is a showdown, where the private cards
are revealed and the higher card’s owner receives the bets.
At the start of the betting, player one can either check/pass
or raise/bet (one chip).
If they check, player two can also check/pass
--— leading to a showdown --—
or bet. If one of the
players bets, their opponent must either call
(betting one chip to match the opponent’s bet),
followed by a showdown, or fold.
\end{quote}

The EFG for this is shown in \cref{fig:EFG}.
To interpret this figure, consider the left part of the tree,
where the state of nature is JQ (as determined by the chance
player's first two dealing actions).
Suppose the first player checks, leading to state JQc.
The second player can either  check, leading to JQcc,
resulting in a showdown with a reward of -1 to player 1
(since J < Q);
or bet, leading to JQcb. In the latter case,
player 1  must then either fold, leading to JQcbf
with a reward to player 1 of -1 (since player 1 only put in one chip);
or player 1 must call, leading to JQcbc
with a reward to player 1 of -2 (since player 1 put in two chips).

\subsubsection{Converting FOSG to EFG}

We can convert an FOSG into an EFG by ``unrolling it''.
First
we  define the information set
for a given information state
as the set of consistent world state trajectories.
By applying the policy of each agent to the world model,
we can derive a tree of possible world states (trajectories)
and corresponding information sets for each agent,
and thus can construct a corresponding (augmented) EFG.
See \citep{Kovarik2022} for details.

\eat{
\subsection{Factored Observation Stochastic Games (FOSG)}
\label{sec:FOSG}

The downside of the EFG representation is that it does not explicitly
represent what observations were received by each player,
or when. In particular, there is no distinction between
\keywordDef{public information}
and \keywordDef{private information}.
(We say that information is public if it is visible
to all players, and all players know this;
thus it is a form of \keywordDef{common knowledge}.)
Explicitly distinguishing these two kinds of information
is important in order to tractably solve certain kinds of games,
like Poker
or Hanabi
(see e.g., \citep{Sokota2021}).

\citep{Kovarik2022}
propose a formalism called
\keywordDef{Factored Observation Stochastic Games}
or \keywordDef{FOSG} that extends POSGs and EFGS by partitioning
the observation for each player into public and private.
(Note that this is different from factoring
the state vector or reward function.)

\newcommand{\tprivate}{\text{private}}
\newcommand{\tpublic}{\text{public}}
\newcommand{\tpl}{\text{pl}}

Formally, we define a FOSG as follows.
The world state starts with some initial state $s_0$
(which we assume each agent observes),
and then transitions at each step according to
$\calT(s_t|s_{t-1},\va_t)$,
where $\va_t=(a_t^1,\ldots,a_t^n)$ is the joint action.
We assume each action is drawn from the set of
\keywordDef{legal actions} in the current state,
$a_t^i \in \calA_i(s_{t-1})$.
The set of players who are eligible to play
in a state is denoted by $\tpl(s)$.
We define $\calA_i(s)=\{\text{no-op}\}$
if it is not $i$'s turn to play
(i.e., $i \not \in \tpl(s)$).
Next  each agent receives its observation
$o_t^i=(o_t^{i,\tprivate}, o_t^{\tpublic})$,
which is generated from $\calO_i^{\tprivate}(s_{t-1},\va_t,s_t)$
and $\calO^{\tpublic}(s_{t-1},\va_t,s_t)$ respectively.
(If the observation is stochastic, we can add the randomness
into the world state.)
Finally,  each agent
receives reward $r_t^i=\calR_i(s_{t-1},\va_t,s_t)$);
this may or may not be visible
to the agent.

Note that a POSG is a special case of a FOSG
where $\calO^{\tpublic}$ always returns $\emptyset$,
and the set of players who can move
at each step is $\tpl(s)=\{1,\ldots,n\}$
(for a simultaneous move game).
By contrast, in a FOSG, we explicitly model
public information, and assume
that only one player moves at each step.
(This assumption is without loss of generality,
since we can introduce a chance player, as we illustrate below.)

\subsubsection{Example: Kuhn Poker as FOSG} 
\label{sec:kuhnPokerFOSG}

\newcommand{\tante}{\text{ante}}
\newcommand{\tcheck}{\text{check}}
\newcommand{\tbet}{\text{bet}}
\newcommand{\tcall}{\text{call}}
\newcommand{\tfold}{\text{fold}}
\newcommand{\tchips}{\text{chips}}
\newcommand{\tDeck}{\text{Deck}}
\newcommand{\tDeckEmpty}{\text{Deck}^+}

We now illustrate how to model Kuhn Poker as a FOSG,
following
\citep{Kovarik2022}.
We let the state be
$s=(C_1,C_2,\tchips_1,\tchips_2,pl)$,
where $C_i \in \tDeckEmpty$ is
the set of cards for player $i$
(where $\tDeck=\{J,Q,K\}$ and
$\tDeckEmpty=\emptyset \union \tDeck$),
$\tchips_i \in \{0,1,2\}$
is the number of chips left,
and $pl \in \{\emptyset, 1,2, c\}$
is the player (where $c$ is the chance player and $\emptyset$
is the terminal player).

We give an example trajectory of the game
below. We assume
$o_t^i=\calO_i(s_{t-1},\va_t,s_{t})=\calO_i(\va_t,s_{t})
=(\calO_i^{\tprivate}(s_{t}),\va_t)$,
so each player sees
the previous joint action as public information
and the resulting  state as private information.
\begin{itemize}
\item The game starts in state
$s_0=(\emptyset, \emptyset, 2,2,1)$.
The only legal action for player 1 is
$\calA_1(s_0)=\{\tante\}$.

\item The next state is
  $s_1=(\emptyset, \emptyset, 1,2,2)$.
  The only legal action for player 2 is
$\calA_2(s_1)=\{\tante\}$.
The observations for player 1 are
$o_1^1=(\emptyset,\tante)$.

\item The next state is
$s_2=(\emptyset, \emptyset, 1,1,c)$,
  since each player has put in one chip,
  and now chance gets to play.
  The observations for player 1 are
$o_2^1=(\emptyset,\tante)$.

\item The next state is a random
  transition to $s_3=(C_1,\emptyset, 1,1,c)$,
  where $C_1 \in \tDeck$.
  Suppose $C_1=Q$.
  Then the observations for player 1 are
  $o_3^1=(Q,? \ra 1)$,
  where $? \ra 1$ represents the public knowledge is
  that something 
  was dealt to player 1 by the chance player.

\item The next state is a random
  transition to $s_4=(Q,C_2, 1,1,1)$,
  where $C_2 \in \tDeck \setminus \{C_1\}$.
  Suppose $C_2=K$.
  The available actions for player 1 are
  $\calA_1(s_4)=\{\tcheck, \tbet\}$.
  The observations for player 1 are
  $o_4^1=(\emptyset,? \ra 2)$,

\item Suppose  player 1
  chooses {\em bet}. Then the next state is
  $s_5=(Q,K,0,1,2)$, since player 1 now has no chips left.
  The available actions for player 2 are
  $\calA_2(s_5)=\{ \tcall,\tfold \}$.
  The observations for player 1 are
  $o_5^1=(\emptyset,\tbet)$,

\item Suppose player 2 chooses {\em call}.
  This results in a showdown,
  so the next state is $s_6=(Q,K,0,0,c)$.
  The observations for player 1 are
  $o_6^1=(\emptyset,\tcall)$,

  \item The final state is $s_7=(Q,K,0,4,\emptyset)$,
    since $K  > Q$, so player 2 wins both chips played by both
    players.
      The observations for player 1 are
      $o_7^1=(\emptyset,(Q,K))$,
      since chance reveals all the private cards.
  \end{itemize}

\eat{
EFGs and POSGs are closely related, as explained in
\citep{Kovarik2022}, although POSGs are arguably simpler and more general.
In a POSG, the imperfect information is due to partial observability,
and requires estimating a distribution over the latent states
$p(s_t|\vh_t)$;
this is analogous to an information set.
If the world model $\calT(s'|s,\va;\vtheta)$
and/or reward functions $\calR_i(s,a^i,s';\vtheta_i)$
are not known, the agent will also have incomplete information,
which requires  estimating a distribution over the model
parameters $p(\vtheta|\vh_t)$, in addition to a distribution over
latent states  (which we could call a ``doubly Bayesian'' approach).
This is equivalent to a multi-agent version of 
a Bayes Adaptive POMDP
\citep{Ross2007}.
}

\subsubsection{Information states, policies, and all that}

We define a trajectory at step $t$
as a the sequence of underlying world states and
joint actions,
 $\traj_t=s_0 a_0 s_1 \ldots s_t$.
The \keywordDef{observation-action history}
(also called \keywordDef{information state})
for agent $i$
at time $t$, denoted 
$\vh_t^i$, is defined as follows.\footnote{
Note that our notation deviates somewhat
from \citep{Kovarik2022}, in order
to maintain consistency with the rest of this book.
In particular,  we use $s_t$ to denote hidden world state,
and $\vh_t$ to represent the observed information state,
whereas \citep{Kovarik2022} does the opposite.
}
We initialize with $\vh_0^i=\emptyset$,
and then whenever $i$ receives an observation
$o_t^i$, it appends it to get $\vh_{t}^i=(\vh_{t-1}^i,o_t^i)$.
We assume that $o_t^i$ contains $a_t^i$
if it was $i$'s turn,
otherwise it records its previous actions as a no-op.
(Thus the players know how many transitions have occured
at each stage of the game, which is important for some
algorithms.)

A  policy or \keywordDef{behavioral strategy}
for player $i$ is a function of the form
$\pi_i: \calH_i \ra \Delta \calA_i$,
where $\calH_i$ is the set of information states
that can arise in the game
(also called the \keywordDef{information-state tree}),
and $\Delta \calA_i$
is a probability distribution over actions.
That is, the policy can be written as
$a_t^i \sim \pi_i(\vh_t^i)$.
Note, however, that the set of information
states is often much larger than the
set of world states,
i.e., $|\calH_i| \gg |\calS|$.
This can make computing optimal  policies
in the partially observed setting
quite expensive (often exponentially so).

A \keywordDef{strategy profile} is a collection
of strategies, $\vpi=(\pi_1,\ldots,\pi_n)$.
We define the expected utility
for player $i$ as
\be
u_i(\vpi)=\expectQ{u_i(\traj)}{\traj \sim \vpi,
  \traj \text{ ends in terminal state}}
\ee
where
$u_i(\traj_t)=\sum_{k=1}^{t-1} \calR_i(s_{k-1},\va_k,s_k)$
is the \keywordDef{cumulative reward}.
(Note that the expectation is over
trajectories that end in a terminal state,
aka \keywordDef{terminal histories}.)
Using this definition, we can define
various solution concepts, such as Nash equilibrium,
as we discuss in in \cref{sec:solutions}.

\subsubsection{Converting FOSG to EFG}

We can convert an FOSG into an EFG by ``unrolling it'',
as follows.
First
we  define the information set
for a given information state $\vh_t^i$,
as the set of consistent world state trajectories,
$I_i(\vh_t^i)=\{\traj: \calO_i(\traj)=\vh_t^i\}$.
By applying the policy of each agent to
the world model,
we can derive a tree of possible world states (trajectories)
and corresponding information sets for each agent,
and thus can construct a corresponding (augmented) EFG.
See \citep{Kovarik2022} for details.

}

\section{Solution concepts}
\label{sec:solutions}

In the multi-agent setting
the definition of ``optimality'' is much more complex
than in the single agent setting, as we will see.
That is, there are multiple \keywordDef{solution concepts}.

\subsection{Notation and definitions}

First we define some notation.
Let $\hat{\vh}_t = \{ (s_k, \vo_k, \va_k)_{k=1}^{t-1}, s_t, \vo_t \}$
be the \keywordDef{full history}, containing all the past
states, joint observations, and joint actions.
Let $\sigma(\hat{\vh}_t) = \vh_t = (\vo_1,\ldots,\vo_t)$ be the history of
joint observations, and
$\sigma_i(\hat{\vh}_t) = \vh_t^i = (o_1^i,\ldots,o_t^i)$
be the history of observations for agent $i$.
(This typically also includes the actions chosen by agent $i$.)

We define the expected return for agent $i$ under joint policy $\vpi$
by
\be
U_i(\vpi) = \sum_{\hat{\vh}_t} p(\hat{\vh}_t|\vpi)  u_i(\hat{\vh}_t)
\ee
where the distribution over full histories is given by
\be
p(\hat{\vh}_t|\vpi)
= \mu(s_0) \calO_{1:n}^0(\vo_0|s_0) \prod_{k=1}^{t-1}
\vpi(\va_k|\hat{\vh}_k)
\calT(s_{k+1}|s_k,\va_k)
\calO_{1:n}(\vo_{k+1}|s_{k+1},\va_k)
\ee
and $u_i(\hat{\vh}_t)$ is the discounted actual return for agent $i$
in a given full history
\be
u_i(\hat{\vh}_t) = \sum_{k=0}^{t-1}
\gamma^k \calR_i(s_k,\va_k,s_{k+1})
\ee

We can  also derive  the following Bellman-like equations:
\begin{align}
  V_i^{\vpi}(\hat{\vh}) &= \sum_{\va} \vpi(\va|\sigma(\hat{\vh}))
  Q_i^{\vpi}(\hat{\vh},\va) \\
  Q_i^{\vpi}(\hat{\vh},\va)
  &= \sum_{s'} \calT(s'|s(\hat{\vh}), \va)
  \left[
    \calR_u(s(\hat{\vh}), \va, s')
    + \gamma \sum_{\vo'} \calO_{1:n}(\vo'|\va,s')
    V_i^{\vpi}((\hat{\vh}, \va, s', \vo'))
    \right]
\end{align}
where $s(\hat{\vh})$ extracts the last state from $\hat{\vh}$.
With this, we can define the expected return using
\be
U_i(\vpi) = \expectQ{V_i^{\pi}((s_0,\vo_0))}{
  \mu(s_0) \calO_{1:n}^0(\vo_0|s_0)}
\ee

Finally,
we define the \keywordDef{best response policy} for agent $i$
as the one that maximizes the expected return for agent $i$
against a given set of policies for all the other agents,
$\vpi_{-i} = (\pi_1,\ldots,\pi_{i-1},\pi_{i+1},\ldots,\pi_n)$.
That is,
\be
\BR_i(\vpi_{-i}) = \argmax_{\pi_i} U_i((\pi_i,\vpi_{-i}))
\label{eqn:BR}
\ee

\subsection{Minimax}

The \keywordDef{minimax} solution is defined for two-agent
zero-sum games. Its existence for normal-form games was first
proven by John von Neumann in 1928.
We say that joint policy $\vpi=(\pi_i,\pi_j)$ is a minimax solution
if
\begin{align}
  U_i(\vpi)
  &= \max_{\pi'_i} \min_{\pi'_j} U_i(\pi'_i, \pi'_j) \\
  &= \min_{\pi'_j} \max_{\pi'_i} U_i(\pi'_i, \pi'_j) \\
  &= -U_j(\vpi)
  \end{align}
In other words, $\vpi$ is a minimax solution iff
$\pi_i \in \BR_i(\pi_j)$ and $\pi_j \in \BR_j(\pi_i)$.
We can solve for the minimax solution using linear programming.

Minimax solutions also exist for two-player zero-sum stochastic
games with finite episode lengths, such as chess and Go.
In the  case of perfect information games, the problems are  Markovian,
and so dynamic programming can be used to solve them.
In generally this may be slow,
but minimax search (a depth-limited version of DP that requires a heuristic function)
can be used.

\subsection{Exploitability}

In the case of a 2 player zero-sum game,
we can measure how close we are to a minimax solution
by computing the \keywordDef{exploitability} score,
defined as
\be
\text{exploitability}(\vpi)
= \half [ \max_{\pi_1'} J(\pi_1', \pi_2) - \min_{\pi_2'} J(\pi_1,\pi_2') ]
\label{eqn:exploitability}
  \ee
where $J(\vpi)$ is the expected reward for player 1 (which is the
loss for player 2).
Exploitability is the expected return of $\pi_i$
playing against a best response to $\pi_i$, averaged over both players $i \in {1,2}$.
Joint policies with exploitability zero are Nash equilibria
(see \cref{sec:nash}).

Note that computing the exploitability score requires
computing a best response to any given policy,
which can be hard in general.
However, one can use standard deep RL methods for this
(see e.g., \citep{Timbers2020}).

\subsection{Nash equilibrium}
\label{sec:nash}

The \keywordDef{Nash equilibrium} generalized the idea
of mutual best response to general-sum games with two or more agents.
That is, we say that $\vpi$ is a Nash equilibrium (NE) if no agent $i$
can improve its expected returns by changing its policy $\pi_i$,
assuming the other agents policies remain fixed:
\be
\forall, \pi'_i. U_i(\pi'_i, \vpi_{-i}) \leq U_i(\vpi)
\ee
John Nash proved the existence of such a solution for general-sum
non-repeated normal form games in 1950.

Below we discuss the kinds of equilibria that exist for
 the games shown in \cref{tab:matrix-games}.

\begin{itemize}
  \item For the rock-paper-scissors game,
the only NE is the \keywordDef{mixed strategy}
(i.e., stochastic policy) where each agent chooses
actions uniformly at random, so $\pi_i=(1/3,1/3,1/3)$.
This yields an expected return of 0.

\item For the battle-of-the sexes game,
there are two pure strategy Nash equilibria: (Opera, Opera) and (Football, Football).
There’s also a mixed strategy equilibrium
but it involves randomness and gives lower expected payoffs.

\item For the Prisoner's Dilemma game, the only NE
is the pure strategy of (D,D), which yields an expected return of
(-3,-3).
Note that this is worse than the maximum possible expected return,
which is (-1,-1) given by the strategy of (C,C).
However, such a strategy is not an NE, since each player could
improve its return if it unilaterally deviates from it
(i.e., defects on its partner).
\end{itemize}

Interestingly, it can be shown that two agents that use rational
(Bayesian) learning rules to update their beliefs about the opponent's
strategy (based on observed outcomes of earlier games),
and then compute a best response to this belief,
will eventually converge to  Nash equilibrium
\citep{Kalai1993}.

Computing a
Nash equilibrium involves identifying a set of strategies where no
player can improve their outcome by unilaterally changing their own
strategy. This requires solving a system of simultaneous equations or
inequalities, which can be computationally intensive, especially as
the number of players and available strategies increases. In fact, for
general games, finding a Nash equilibrium is a famously hard problem,
falling into a complexity class called \keywordDef{PPAD-complete}, which suggests
that there is no efficient, universal algorithm for finding one.

\subsection{Approximate Nash equilibrium}

It is possible to relax the definition of exact
inequality by defining an $\epsilon$-Nash equilibrium
as a joint policy that satisfies
\be
\forall, \pi'_i. U_i(\pi'_i, \vpi_{-i}) - \epsilon \leq U_i(\vpi)
\ee
Unfortunately, the expected return from
a  $\epsilon$-Nash equilibrium can be very different from the expected
return from a true NE.
For example, consider this matrix game:
\be
        \begin{tabular}{c|cc}
            & C & D \\
            \hline
            A & 100,100 & 0,0 \\
            B & 1,2  & 1,1
        \end{tabular}
\ee
The unique NE is (A,C), but the $\epsilon$-NE
with $\epsilon=1$ is either (A,C) or (B,D),
which clearly have very different rewards.

Despite the above drawback, much computational work
focuses on approximate Nash equilibria.
Indeed we can measure the rate of convergence
to such a state by defining
\be
\text{NashConv}(\vpi) = \sum_i \delta_i(\vpi)
\ee
where $\delta_i(\vpi)$ is the amount of incentive
that $i$ has to deviate to one of its best responses
away from the joint policy:
\be
\delta_i(\vpi) = u_i(\pi_i^b, \vpi_{-i})
- u_i(\vpi),
\;
\pi_i^b \in \BR(\vpi_{-i})
\ee

\subsection{Entropy regularized Nash equilibria (aka Quantal Response Equilibria)}
\label{sec:QRE}

In this section, we discuss
\keywordDef{quantal response equilibria}
or \keywordDef{QRE} \citep{McKelvey1995,McKelvey1998}.
These are like Nash equilibria except
the best response policy is a ``soft'' entropy-regularized policy
(see \cref{sec:SAC}).
This kind of  equilibrium
reflects the fact that players may not always choose the best response with certainty,
but instead they  make choices based on a probability distribution over actions,
based on  the relative expected utility of each action.
Thus it is a Bayesian equilibrium.
This can be useful for modeling human behavior that deviates
from the predictions of Nash (studied in the field
of \keywordDef{behavioral game theory}).
In addition, it is useful for developing
algorithms that converge to a unique
equilibrium, as we discuss in \cref{sec:MMD}.

For single agent problems with a single state (i.e., bandit problems),
we say that a policy $\pi$ is $\alpha$-soft optimal
in the normal sense if it satisfies
\be
\pi = \argmax_{\pi' \in \Delta(A)}
\E_{A \sim \pi'}
q(A) + \alpha \entropy(\pi')
\ee
where $\Delta(A)$ is the action simplex,
and $q$ is the action-value function.
If this holds for all states (decision points) $s$,
we say that $\pi$ is $\alpha$-soft optimal in the behavioral
sense.

For two-player zero-sum NFGs (which have  a single decision point or state),
we say that a policy is a QRE
if each player’s policy is soft optimal in the normal sense
conditioned on the other player not changing its policy
\citep{McKelvey1995}.
For two-player zero-sum games with multiple states (i.e., EFGs),
we say that a policy is a \keywordDef{agent QRE}
if each player’s policy is soft optimal in the behavioral sense
conditioned on the other player not changing its policy
\citep{McKelvey1998}.

\subsection{Correlated equilibrium}

The concept of a Nash equilibrium assume the policies are independent,
which can limit the expectd returns.
A \keywordDef{correlated equilibrium} (CE) allows for correlated policies.
Specifically, we assume there is a central policy $\vpi_c$
that defines a distribution over joint actions.
Agents can follow this recommended policy, or can choose
to deviate from it by using an action modified
$\xi_i:\calA_i \ra \calA_i$.
We then say that $\vpi_c$ is a CE is for all $i$ and $\xi_i$
we have
\be
\sum_{\va} \vpi_c(\va)
\calR_i((\xi_i(a^i),\va^{-i})) \leq \sum_{\va} \vpi_c(\va) \calR_i(\va)
\ee
That is, player $i$ has no incentive to deviate
from the recommendation, after receiving it.
It can be shown that the set of correlated equilibria contains the set
of Nash equilibria.
In particular,
since
Nash equilibrium is a special case of correlated equilibrium
in which the joint policy $\vpi_c$
is factored into independent agent policies with
$\vpi_c(\va) = \prod_i \pi_i(a_i)$.

To see how a correlated equilibrium can give higher returns
than a Nash equilibrium,
 consider the \keywordDef{Chicken game}.
This models two agents that are driving towards each other.
Each agent can either stay on course (S)
or leave (L) and avoid a crash.
The payoff matrix is as follows:
\be
        \begin{tabular}{c|cc}
            & S & L \\
            \hline
            S & 0,0 & 7,2 \\
            L & 2,7  & 6,6
        \end{tabular}
        \label{eqn:chicken}
\ee
This reflects the fact that if they both stay on course,
then they both die and get reward 0;
if they both leave, they both survive and get reward 6;
but if  player $i$ chooses to stay and the other one leaves,
then  $i$ gets a reward of 7 for being brave,
and $-i$ only gets a reward of 2 for chickening out.

We can represent  $\pi_i$ by the scalar $\pi_i(S)$,
since $\pi_i(L)) = 1-\pi_i(S)$.
Hence $\vpi$ can  be defined by the tuple $(\pi_1,\pi_2)$.
There are 3 uncorrelated NEs:
$\vpi=(1,0)$ with return $(7,2)$;
$\vpi=(0,1)$ with return $(2,7)$;
and
$\vpi=(\frac{1}{3},\frac{1}{3})$ with return $(4.66,4.66)$.
There is 1 CE, namely
$\vpi_c(L,L)=\vpi_c(S,L)=\vpi_c(L,S)=\frac{1}{3}$
and $\vpi_c(S,S)=0$.
The central policy has an expected return of
\be
7 \cdot \frac{1}{3} + 2 \cdot \frac{1}{3}
+ 6  \cdot \frac{1}{3} = 5
\ee
which we see is higher than the NE of 4.66.
This is because it avoids the deadly joint (S,S) action.
To show that this is a CE, consider
the case where $i$ (e.g., row player) receives recommendation $L$;
they know that $j$ (column player) will choose either $S$ or $L$ with probability
0.5 (because the central policy is uniform).
If $i$ sticks with the recommendation, its expected return is
$0.5 \cdot 2 + 0.5 \cdot 6 = 4$;
this is greater than deviating from the recommendation and picking $S$,
which has expected return of $0.5 \cdot 0 + 0.5 \cdot 7 = 3.5$.
Thus $\vpi_c$ is a CE.

In \citep{Aumann1987} they show that the CE solution  corresponds
to the behavior of a rational Bayesian agent.
The correlated equilibrium solution can be computed via linear programming.

\subsection{Limitations of equilibrium solutions}

Equilibrium solutions have several limitations.
First, they do not always maximize expected returns.
For example, in Prisoner's Dilemma, (D,D) is Nash but (C,C)
yields higher returns.
Second, there can be multiple (even infinitely many) equilibria,
each with different expected returns, as we have seen.
Third, equilibria for sequential games don't specify what to do
if the history deviates from the equilibrium path,
i.e., they do not define the policy for full histories
where $p(\hat{\vh}|\vpi)=0$;
this can be problematic when the agents are learning,
or the environment is changing in some other way.
Consequently it is common to define additional solution requirements,
as we discuss below.

\subsection{Pareto optimality}

We say a joint policy $\vpi$ is \keywordDef{Pareto optimal}
if it is not \keywordDef{Pareto dominated} by any other joint policy
$\vpi'$.
We sat that $\vpi$ is Pareto dominated by $\vpi'$ if
$\vpi'$ improves the expected return for at least one agent:
\be
\forall i. U_i(\vpi') \geq U_i(\vpi)
\text{ and }
\exists i. U_i(\vpi') > U_i(\vpi)
\ee
and if it does not decrease the payoff for any agents.

\begin{figure}
\centering
\includegraphics[height=3in]{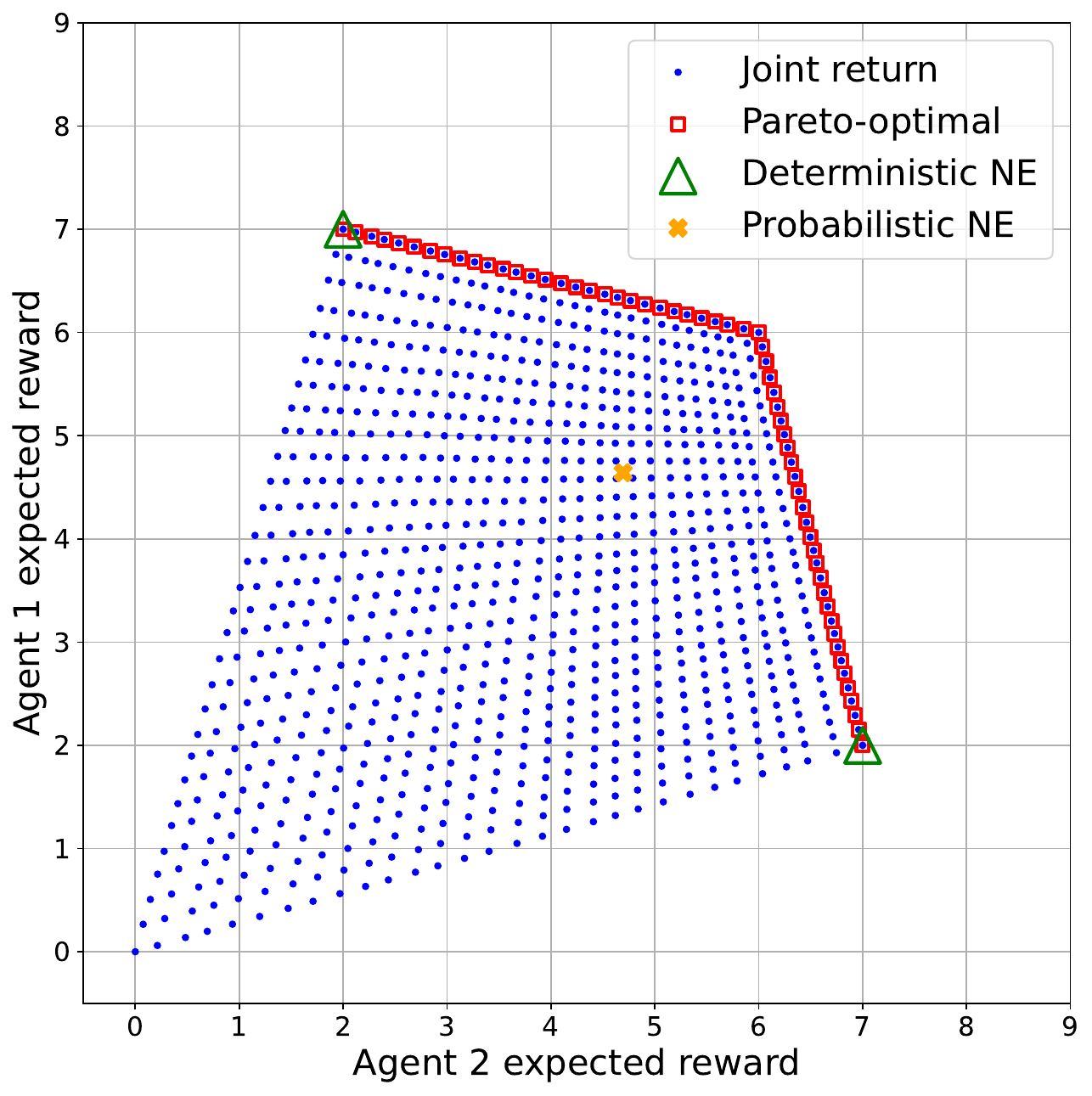}
\caption{
  Space of (discretized) joint policies for Chicken game.
  \figtaken{Fig 4.4 of \citep{marl-book}.}
    \figthanks{Stefano Albrecht}.
  }
\label{fig:chicken-po}
\end{figure}

\cref{fig:chicken-po}
illustrates the expected joint rewards
and the Pareto frontier for all feasible policies (up to quantization error)
applied to the Chicken game in \cref{eqn:chicken}.
We see that the two pure NEs are on the Pareto frontier (as are many
other policies that are not Nash), but the mixed NE is not
Pareto optimal.

\subsection{Social welfare and fairness}

Pareto optimality ensures there is no other solution in which at least
one agent is better off, without making other agents worse off.
However, it does not make any guarantees about the total rewards,
or their distribution amongst agents. For example,
along the Pareto frontier in \cref{fig:chicken-po}, the joint returns
vary from (7,2) to (6,6) to (2,7).

To further constrain the space of desirable solutions,
we can consider additional concepts.
For example, we define \keywordDef{welfare optimality} as
\be
W(\vpi) = \sum_i U_i(\vpi)
\ee
A joint policy is welfare-optimal if $\vpi \in \argmax_{\vpi'}
W(\vpi')$.
One can show that welfare optimality implies Pareto optimality,
but not (in general) vice versa.

Similarly, we define
\keywordDef{fairness optimality} as
\be
F(\vpi) = \prod_i U_i(\vpi)
\ee
A joint policy is fairness-optimal if $\vpi \in \argmax_{\vpi'}
F(\vpi')$.

In the battle-of-the-sexes game in \cref{tab:matrix-games},
the  only fair outcome is the joint distribution over $Pr(F,F) = Pr(O,O) = 0.5$,
which means the couple spend half their time watching football
and half going to the opera.

In the Chicken game in \cref{fig:chicken-po},
there is only one solution that is both welfare-optimal
and fairness-optimal, namely  the joint policy
with expected return of (6,6).
Note, however, that this is not a Nash policy.

\subsection{No regret}

The quantity known as \keywordDef{regret}
measures the difference between the rewards an agent received
versus the maximum
rewards it could have received if it had chosen a different
action.
For a non-repeated normal-form general-sum game, played over $E$ episodes,
this is defined as
\be
\Regret^E_i = \max_{a^i} \sum_{e=1}^E
  [\calR_i((a^i,\va_e^{-i})) - \calR_i(\va_e)]
\ee
We can generalize the definition of regret to stochastic games
and POSGs by defining the regret over policies instead of actions.
That is,
\be
\Regret^E_i = \max_{\pi^i} \sum_{e=1}^E
  [U_i((\pi^i,\vpi_e^{-i})) - U_i(\vpi_e)]
  \ee
In all these cases,
an agent is said to have no-regret if
\be
\forall i. \lim_{E \ra \infty}
\frac{1}{E} \Regret_i^E \leq 0
\ee

\subsection{Shapley values}

The \keywordDef{Shapley value}
\citep{ShapleyValues}
allows one to estimate the narginal
contribution
of a single agent to a common reward (cooperative) game,
which can help ameliorate the
\keywordDef{credit assignment problem}.
Specifically, suppose there are $N$ players,
and let $S$ be a subset of $\leq N$ players
that form a team. Let $v(S)$ be the expected value obtained
by that team, and $v(S \setminus \{i\})$ be the value
of the team when $i$ is absent. Then we define
$i$'s Shapley value as
\be
\phi(i) = \sum_{S \subseteq N \setminus \{i\}}
w(S) [v(S \union \{i\}) - v(S)]
\ee
where the weighting term is given by
\be
w(S) = \frac{|S|! (|N|-|S|-1)!}{|N|!}
 = \frac{1}{|N| \cdot \binom{|N|-1}{|S|}}
\ee
which 
represents the probability that,
for a random ordering of all players, the players in coalition
$S$ come before player $i$,
and the remaining players come after player $i$.

An interesting application of Shapley values
arises in the ``explainable AI'' literature,
where one of the goals is to estimate the importance
of indivual predictors to an overall prediction.
This can be done by using the \keywordDef{SHAP}
(SHapley Additive exPlanations) framework of
\citep{Lundberg2017}.
Similarly, people use Shapley values to assess the significance
of individual tokens in a prompt
(see e.g., \citep{Enouen2024}).

\subsection{Stackelberg equilibrium}
\label{sec:stackelberg}

So far we have mostly focused on games where the players make their moves
simultaneously.
We can also consider \keywordDef{sequential games}, where the players
take turns.
For such games, the concept of
\keywordDef{Stackelberg equilibrium},
developed by economist Heinrich Freiherr von Stackelberg,
becomes relevant.
Here,  one player,
called the ``leader'',  makes their decision first.
The other player, the ``follower'',
observes the leader's choice and then makes their own decision.
This sequential structure gives the leader a significant advantage.
The leader, knowing how the follower will react to any given move,
can choose a strategy that maximizes their own payoff,
anticipating the follower's subsequent best response. The follower, in
turn, optimizes their outcome based on the leader's committed action.  
The resulting set of strategies, where the leader has chosen their
optimal move and the follower has responded optimally,
constitutes the Stackelberg equilibrium.

The concept of Stackelberg equilibrium is useful for analysing
\keywordDef{setter-solver} problems, which arise in
unsupervised environment design/ curriculum learning
(\cref{sec:curriculum}). Here one player (the setter)
designs a task or environment that is challenging for the other player (the
solver);
as the solver gets better, the setter gradually makes the problem
more challenging.
See e.g., \citep{Dennis2020,Racaniere2020,Parker-Holder2023}.

Note that  Stackelberg equilibria can be easier to compute
than Nash equilibria.
To find the Stackelberg equilibrium, you can use a method called
\keywordDef{backward induction}. You start by determining the follower's best
response for every possible action the leader could take. Then, the
leader, knowing how the follower will react, simply chooses the action
that will lead to the best outcome for themselves. This process is
often straightforward and can be solved with a single optimization
problem, including using gradient-based methods 
\citep{Colson2007,Zucchet2022,Gerstgrasser2023}.

\section{Algorithms}

In this section, we discuss various MARL algorithms.

\subsection{Centralized learning}
\label{sec:centralLearning}

The simplest way to solve a MARL problem is to reduce it to a single
agent RL (SARL) problem.
In \keywordDef{central learning}, we learn a single joint
policy over the joint action space. This requires that we can
transform the joint reward $\vr_t=(r_t^1,\ldots,r_t^n)$
into a scalar $r_t$. This is easy to do in common reward games,
where the agents must cooperate.
However, for general sum games,
it may be impossible to define a single scalar reward
across all agents.
And even if we can define such a shared reward, the resulting method
may not scale well with the number of agents,
and learns a policy that requires global access to all of the
observations
for each agent.

\subsection{Independent learning}
\label{sec:independentLearning}
\label{sec:independentMARL}

In  \keywordDef{independent learning},
each agent treats all other agents as part of the
environment, and then uses any standard single-agent RL algorithm
for training. This is done in parallel across all agents.

\subsubsection{Independent Q learning}
\label{sec:IQL}

For example, if we use  Q learning for each agent,
the method is known as independent Q-learning
or \keywordDef{IQL}; see \cref{algo:IQL} for the pseudocode.

\begin{algorithm}[H]
    \caption{Independent Q learning (DQN for multiple independent agents)}
  \label{algo:IQL}
    Initialize $n$ value networks with random parameters $\theta_1,\dots,\theta_n$\;
    Initialize $n$ target networks with parameters $\widebar\theta_1=\theta_1,\dots,\widebar\theta_n=\theta_n$\;
    Initialize a replay buffer for each agent $D_1, D_2, \dots, D_n$\;
    
    \For{time step $t=0, 1, 2, \dots$}{
        Collect current observations $\ob_t^1, \dots, \ob_t^n$\;
        
        \For{agent $i=1, \dots, n$}{
            With probability $\epsilon$: choose random action $\ac_t^i$\;
            Otherwise: choose $a_t^i \in \argmax_{a_i}{Q(\his_t^i, \ac_i; \theta_i)}$\;
        }
        
        Apply actions $(\ac_t^1,\ldots,\ac_t^n)$; collect rewards $\rew_t^1,\ldots,\rew_t^n$ and next observations $\ob_{t+1}^1,\ldots,\ob_{t+1}^n$\;
        
        \For{agent $i=1,\dots, n$}{
            Store transition $(\his_t^i, \ac_t^i, \rew_t^i, \his_{t+1}^i)$ in replay buffer $D_i$\;
            Sample random mini-batch of $\batchsize$ transitions $(\his_k^i, \ac_k^i, \rew_k^i, \his_{k+1}^i)$ from $D_i$\;
            
            \eIf{$s_{k+1}^i$ is terminal}{
                Targets $y_k^i \gets \rew_k^i$\;
            }{
                Targets $y_k^i \gets \rew_k^i + \dsc \max_{\ac'_i\in\Ac_i} Q(\his_{k+1}^i, \ac'_i; \widebar{\theta_i})$\;
            }
            
            Loss $\loss(\theta_i) \gets \frac{1}{\batchsize} \sum_{k=1}^\batchsize \biggl(y_k^i - Q(\his_k^i, \ac_k^i; \theta_i)\biggr)^2$\;
            Update parameters $\theta_i$ by minimizing the loss $\loss(\theta_i)$\;
            In a set interval, update target network parameters $\widebar\theta_i$\;
        }
    }
\end{algorithm}

\subsubsection{Independent  Actor Critic}
\label{sec:MAA2C}

Instead of using value-based methods,  we can also use
policy learning methods.
The multi-agent version of the policy gradient
theorem in  \cref{eqn:SFE}
is the following (see e.g., \citep{marl-book} for derivation):
\begin{align}
    \nabla_{\vtheta_i} J(\vtheta_{1:n})
&\propto \expectQ
  {Q_i^{\pi}(\hat{\vh},(a^i,\va^{-i})
    \nabla_{\vtheta_i} \log \pi(a_i|h_i=\sigma_i(\hat{\vh});\vtheta_i)}
    {\hat{\vh} \sim p(\hat{\vh}|\pi), a^i \sim \pi_i, \va^{-i} \sim \vpi^{-i}}
\end{align}
where
$\hat{\vh}_t = \{ (s_k, \vo_k, \va_k)_{k=1}^{t-1}, s_t, \vo_t \}$
is the \keywordDef{full history} (containing all the past
states, joint observations, and joint actions),
and
$\sigma_i(\hat{\vh}_t) = \vh_t^i = (o_1^i,\ldots,o_t^i)$
is the history of observations for agent $i$.

In practice, we usually subtract a baseline term from $Q$, to reduce the
variance.
If we use the value function for the baselne,
then the first term inside the expectation
becomes
\be
Q_i^{\pi}(\hat{\vh},(a^i,\va^{-i})) - V_i^{\pi}(\hat{\vh})
 = \Adv_i^{\pi}(\hat{\vh},\va)
\ee
where $\Adv$ is the advantage, as we discussed in \cref{sec:A2C}.
This can be used inside a multi-agent version of
the advantage actor critic or A2C method
(known as \keywordDef{MAA2C})
shown in \cref{algo:A2C-MA}.
(To combat the fact that we cannot use replay buffers with an
on-policy method, we assume instead that we can parallelize over
multiple (synchronous) environments,
to ensure we have a sufficiently large minibatch to estimate the loss
function at each step.)

\begin{algorithm}[H]
  \caption{Multi-agent Advantage Actor-Critic (MAA2C)}
  \label{algo:A2C-MA}
    Initialize $n$ actor networks with random parameters $\vtheta_1,\dots,\vtheta_n$\
    
    Initialize $n$ critic networks with random parameters $\vw_1,\dots,\vw_n$\
    
    Initialize $K$ parallel environments\
    
    Initialize histories $h_0^{i,k}$ for each agent $i$ and environment $k$ \
    
    \For{\rm{time step} $t=0\dots$}{

      \For{\rm{environment} $k=1,\ldots,K$}{
      Sample actions: $\{a_t^{i,k} \sim \pi(\cdot|h_t^{i,k}, \vtheta^i)\}_{i=1}^n$

      Sample next state: $s_{t+1}^k \sim T(\cdot|s_t^k, \va_t^{1:n,k})$
      
      Sample observations:
      $\{o_{t+1}^{i,k} \sim O_i(\cdot|s_t^k,a_t^{i,k}) \}_{i=1}^n$

      Sample rewards:
      $\{r_{t}^{i,k} \sim R_i(\cdot|s_t^k,\va_t^{1:n,k}, s_{t+1}^k) \}_{i=1}^n$
            
      Update histories:
      $\{\his_{t+1}^{i,k} = (\his_t^{i,k}, o_{t+1}^{i,k}) \}_{i=1}^n$
      
      }
      
        \For{\rm{agent} $i=1,\dots, n$}{
            \If{$s_{t+1}^k$ \rm{is terminal}}{
              $\Adv(\his_t^{i,k}, \ac_t^{i,k}) \gets
              \rew_t^{i,k} - V(\his_t^{i,k}; \vw_i)$\;
                Critic target $y_t^{i,k} \gets \rew_t^{i,k}$\;
            }
            \Else{
              $\Adv(\his_t^{i,k}, \ac_t^{i,k}) \gets
              \rew_t^{i,k} + \dsc V(\his_{t+1}^{i,k}; \vw_i)
              - V(\his_t^{i,k}; \vw_i)$\;
              Critic target $y_t^{i,k} \gets \rew_t^{i,k} +
              \dsc V(\his_{t+1}^{i,k}; \vw_i)$\;
            }
            
            Actor loss:
            \[
            \loss(\vtheta_i) \gets \frac{1}{K} \sum_{k=1}^K
            \Adv(\his_t^{i,k},  \ac_t^{i,k})
            \log \pol(\ac_t^{i,k} \mid \his_t^{i,k}; \vtheta_i)
            \]
            
            Critic loss:
            \[
            \loss(\vw_i) \gets \frac{1}{K} \sum_{k=1}^K
            \left(y_t^{i,k} - V(\his_t^{i,k}; \vw_i)\right)^2
            \]
            
            Update parameters $\vtheta_i$ by minimizing the actor loss $\loss(\vtheta_i)$\;
            Update parameters $\vw_i$ by minimizing the critic loss $\loss(\vw_i)$\;
        }
    }
\end{algorithm}

\subsubsection{Independent PPO}

We can implement an independent version of PPO
(known as \keywordDef{IPPO})
in a similar way,
by updating all the policies in parallel
\citep{deWitt2020}.

\subsubsection{Learning dynamics of multi-agent policy gradient methods}
\label{sec:learningDynamics}

In general, applying policy gradient methods
to multiple agents in parallel may not result in convergence
\citep{Cheung2019,Bloembergen2015}.
To illustrate this,
consider a non-repeated normal-form
general-sum game with two players
and two actions.
(This is an imperfect information game since each player
does not know the other's actions when they make their decision.)
Denote the reward matrices by
\be
\calR_i = \begin{pmatrix} r_{11} & r_{12} \\ r_{21} &
  r_{22} \end{pmatrix},
\;
\calR_j = \begin{pmatrix} c_{11} & c_{12} \\ c_{21} &
  c_{22} \end{pmatrix}
\ee
Denote the policies by
\be
\pi^i = (\alpha,1-\alpha),
\;
\pi^j = (\beta,1-\beta)
\ee
The expected reward for agent $i$, given the joint policy
$\vpi=(\alpha,\beta)$,
is given by
\be
U_i(\alpha,\beta)
= \alpha \beta r_{11} + \alpha (1-\beta) r_{12}
+ (1-\alpha)\beta r_{21}
+ (1-\alpha)(1-\beta) r_{22}
\ee
The expression $U_j(\alpha,\beta)$ is analogous,
with $r_{ij}$ replaced with $c_{ij}$.
(Note that computing $U_i$ requires knowledge of
$\calR_i$ but also of $\pi_j$
(and vice versa for computing $U_j$);
we will relax this assumption below.)
Finally, we can learn the policies using gradient ascent:
\be
\alpha_{k+1} = \alpha_k + \kappa
\frac{\partial U_i(\alpha_k,\beta_k)}{\partial \alpha_k},
\;\;
\beta_{k+1} = \beta_k + \kappa
\frac{\partial U_j(\alpha_k,\beta_k)}{\partial \beta_k}
\ee
where $\kappa$ is the learning rate.

\label{sec:WOLF}

\begin{figure}
\centering
\includegraphics[height=3in]{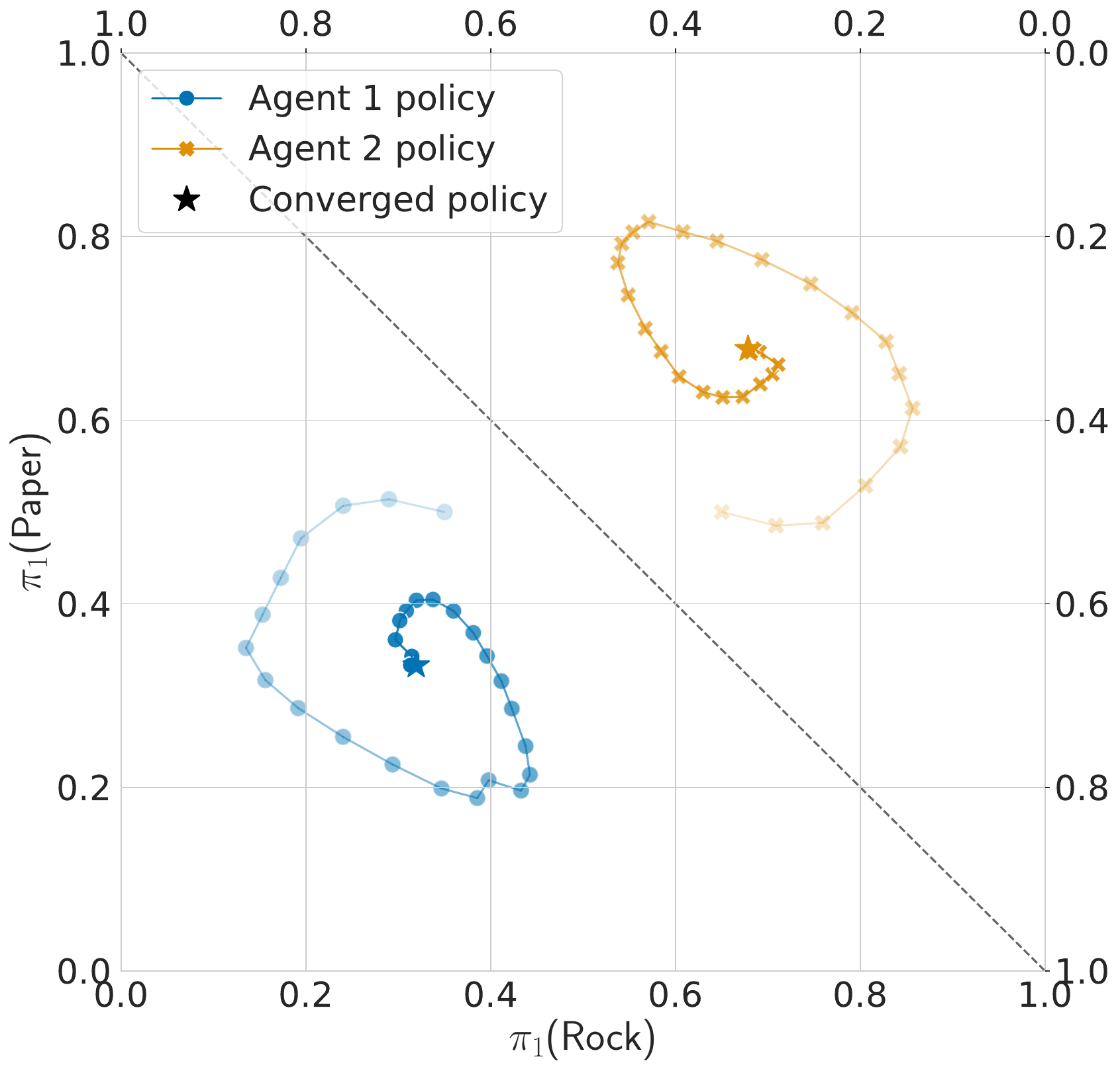}
\caption{
  Learning dynamics of two policies for the rock-paper-scissors game.
  The upper and lower triangles illustrate the policies for each
  agent over time as a point in the 2d simplex (noting that
  $\pi_e^i(S) = 1-(\pi_e^i(P) + \pi_e^i(R))$,
  where $S$ is the scissors action, $P$ is paper action,
  and $R$ is rock action).
  The update rule is $\vpi_{e+1}=LR(\data_e,\vpi_e)$,
  where LR is the learning rule known as
  WoLF-PHC  (see \cref{sec:WOLF}).
  \figtaken{Fig 5.5 of \citep{marl-book}.}
    \figthanks{Stefano Albrecht}.
  }
\label{fig:learning-dynamics}
\end{figure}

We can analyse the dynamics of the above procedure
as $\kappa \ra 0$; this is known as
\keywordDef{infinitesimal gradient ascent} or
\keywordDef{IGA}.
One can show that $(\alpha,\beta)$ does not always converge
(depending on the values in $\calR_i$ and $\calR_j$),
but if it does, the resulting converged joint policy
is a NE \citep{Singh2000nash}.
However, there is a method called
\keywordDef{Win or Learn Fast} or \keywordDef{WoLF}
from \citep{Bowling2002} which can ensure that IGA policies
always converge to a NE for two-agent two-action
normal-form games.
The trick is 
to learn slow (by using smaller $\kappa$)
when winning
(i.e., if $U_i(\alpha_k,\beta_k) > U_i(\alpha_e,\beta_k)$
where $\alpha_e$ is a policy from some NE),
and to  learn fast (by using larger $\kappa$)
when losing (i.e., when not winning).
This approach can be extended to stochastic games,
without requiring knowledge of reward functions or policies.
The resulting method is called
\keywordDef{WoLF-PHC},
which stands WoLF with Policy Hill Climbing
\citep{Bowling2002}.
See \cref{fig:learning-dynamics} for an example.

\subsection{Centralized training of decentralized policies (CTDE)}
\label{sec:CTDE}

We can improve performance beyond independent learning
by using a  paradigm known as
Centralized Training and Decentralized Execution (\keywordDef{CTDE}),
in which the learing algorithm has access to all the information (from
  all agents)  at training time, but at test time, agents only observe their
  own local observations.
The \keywordDef{central information} can contain the joint action taken by all
agents, and/or the joint observation vector,
even such joint information is not available at execution time.

We can modify the multi-agent A2C algorithm
in \cref{algo:A2C-MA} to exploit this assumption
by writing the $i$'th value/ advantage function
as $V(h_t^i,\ci_t;\vw_i)$ / $\Adv(h_t^i,\ci_t,a_t^i)$,
where $\ci_t$ is the shared central information.
It is perfectly valid for the {\em critics} to have
this  kind of \keyword{central information},
as long as the {\em policies} do not rely on this,
since the critics are not used during execution.
This is known as  \keywordDef{centralized critics}
with \keywordDef{decentralized actors}.
We can create a CTDE version of PPO in a similar way;
this is known as \keywordDef{MAPPO}  \citep{MAPPO}.

\subsubsection{Application to Diplomacy (Cicero)}
\label{sec:cicero}


In this section, we describe the \keywordDef{Cicero} system from
\citep{cicero},
which achieved human-level performance in the complex natural
language 7-player strategy game called \keywordDef{Diplomacy},
which requires both cooperative and competitive behavior.\footnote{
A simpler version of the game,
called \keywordDef{No-press diplomacy},
without the natural language negotation phase,
is discussed in 
\citep{Bakhtin2021}.
}
Cicero used CTDE, combining an LLM for generating and
interpreting dialog
with a mix of self-play RL, imitation learning, opponent modeling,
and policy generation using regret minimization.
The system uses imitation learning on human games
to warm-start the initial policy and language model,
and then is refined using RL with self-play.
The system uses explicit belief state modeling over the opponents' intents and plans;
this is learning via supervised learning over past dialogues and game outcomes,
and refined during self-play.
(For a more recent method, which does not need human data,
and which is based on LLMs augmented with memory
and self-reflection, see \citep{Guan2024diplomacy}.)

\subsection{Value decomposition methods for common-reward games}

In this section, we discuss methods for deriving a policy
from a centralized state-action value function
$Q(\vh,\ci,\va)$,
where $\ci$ is the central information
(see \cref{sec:CTDE}),
$\vh$ is the shared state (history),
and $\va$ is the joint action.
To ensure that the per-agent policy can be implemented
using only locally available information,
we need to use \keywordDef{value decomposition} methods,
which assume that the global value function can be decomposed
into separate value functions, one per agent.
(This is only possible if  we are solving a common-reward or cooperative game.)
This decomposition is valid provided the separate value functions
satisfy a property known as
the \keywordDef{individual global max} or \keywordDef{IGM} property,
which says that
\be
\forall \va. \va \in A^*(\vh,\ci;\vtheta)
\Leftrightarrow
\forall i. a_i \in A_i^*(h^i;\vtheta_i)
\ee
where
$A^*(\vh,\ci;\vtheta) = \argmax_{\va} Q(\vh,\ci,\va;\vtheta)$
and
$A_i^*(h^i;\vtheta) = \argmax_{\va} Q(h^i,a^i;\vtheta_i)$.
This ensures that picking actions
locally for each agent will also be optimal
globally.

\subsubsection{Value decomposition network (VDN)}

For example, consider the \keywordDef{value decomposition network}
or \keywordDef{VDN} method of \citep{Sunehag2017}.
This assumes  a linear decomposition
\be
Q(\vh_t,\ci_t,\va_t;\vtheta) = \sum_i Q(h_t^i,a_t^i;\vtheta_i)
\ee
This clearly satisfies IGM.

\subsubsection{QMIX}

A more general method, known as \keywordDef{QMIX}, is presented in
\citep{Rashid2018}.
This assumes 
\be
Q(\vh_t,\ci_t,\va_t;\vtheta) =
\fmix(Q(h_t^1,a_t^1;\vtheta_1),\ldots,
  Q(h_t^n,a_t^n;\vtheta_n))
  \ee
  where $\fmix$ is a neural network
  that is constructed so that it is monotonically increasing in each of its
  arguments. (This is ensured by requiring all the weights
  of the mixing network to be non-negative; the weights themselves
  are predicted by another ``hyper-network'', conditioned on the state $h_t^i$.)
This satisfies IGM since
\be
\max_{\va} Q(\vh_{t+1},\ci_{t+1},\va;\overline{\vtheta})
= \fmix\left( \max_{a^1} Q(h_{t+1}^1,a^1;\overline{\theta}_1),
\ldots, \max_{a^n} Q(h_{t+1}^n,a^n;\overline{\theta}_n) \right)
\ee
Hence we can fit this Q function by minimzing the TD loss
(with target network $\overline{\vtheta}$):
\be
\loss(\vtheta) = \frac{1}{B}
\sum_{(\vh_t,\ci_t,\va_t,r_t,\vh_{t+1},\ci_{t+1}) \in \calB}
\left( r_t + \gamma \max_{\va}
Q(\vh_{t+1},\ci_{t+1},\va;\overline{\vtheta})
 - Q(\vh_{t},\ci_{t},\va;\vtheta) \right)^2
\ee

\subsection{Policy learning with self-play}
\label{sec:selfPlay}

For symmetric zero-sum games, where $r_i(s)=-r_j(s)$,
we can assume that each player
uses the same policy, modulo rearrangement of the input state.
That is, $\pi_j(.|s) = \pi_i(.|\psi(s))$,
where $\pi_i$ is the main policy for player $i$,
and $\psi(s)$ deterministically modifies the  state to reflect
the symmetry.

For example, consider the game of chess, where the state is represented
by $s=(x,y)$, where $x$ is a vector containing the location of
player 1's pieces (or -1 if they are removed),
and $y$ is a vector containing the opponent's
pieces. Thus the policy for player 1, $\pi_1$,
just needs to access the $x$ part of the state.
For player 2, 
we can transform the state vector into $s'=\psi(s)=(y,x)$ and then apply $\pi_1$
to choose actions,
so $\pi_2(s)=\pi_1(\psi(s))$.

The result is a single agent problem,
in which, from $i$'s perspective,
the  state transition function has the following form,
in which we treat player $j$ as part of the environment:
\be
p^{\pi}(s'|s,a^i) = \sum_{a^j} \pi(a^j|\psi(s)) p(s'|s,a^i,a^j) 
\ee
Thus   lets us learn $\pi$
using standard single-agent policy learning methods,
such as PPO or
policy improvement based on search (decision-time planning).
This is known as \keywordDef{self-play}.

Self-play is used by AlphaZero  (see
\cref{sec:alphaZero}) to learn to play
perfect information games like Chess and Go
at super-human level.
In \citep{TiZero}, they used multi-agent self-play to train
soccer agents.
And in  \citep{Zheng2024alignment,JacquesRedTeam,SpiralRL}
they used self-play to train LLM policies
(see \cref{sec:LLMselfPlay}).

For perfect information, zero-sum games, self-play can be proved to converge
to a Nash equilibrium.
Unfortunately, in general games
(e.g., for imperfect information games, such as Poker or Hanabi,
or for general sum games),
self-play
 can lead to oscillating strategies or cyclical behavior,
 rather than converging to a Nash equilibrium,
 as we discuss in \cref{sec:learningDynamics}.
Thus self-play can result in  policies that are easily exploited.
We discuss more stable learning methods below.


 \eat{
In practice it is better to
try to learn the opponent policy (see \cref{sec:policyWithLearnedOpponents}),
or use population-based training methods
(\cref{sec:PBT})
or use counterfactual regret minimization
(\cref{sec:CFR}).
}

\subsection{Policy learning with learned opponent models}
\label{sec:policyWithLearnedOpponents}

\begin{figure}[h]
    \centering
    \begin{tikzpicture}[
        every node/.style={font=\large},
        enc/.style={rectangle, rounded corners, draw=black, fill=yellow!50, minimum height=1cm, minimum width=3cm},
        dec/.style={rectangle, rounded corners, draw=black, fill=blue!30, minimum height=1cm, minimum width=3.5cm},
        policy/.style={rectangle, rounded corners, draw=black, fill=green!40, minimum height=1cm, minimum width=4.5cm},
        arrow/.style={thick,->,>=stealth}
    ]
    
    \node (h) at (0, 0) {\( h_t^i \)};
    
    \node[enc] (encoder) at (3, 0) {\( f^e(h_t^i; \psi_i^e) \)};
    \node (m) at (6, 0) {\( m_t^i \)};
    
    \node[dec] (decoder) at (9, 0) {\( f^d(m_t^i; \psi_i^d) \)};
    \node (pi_hat) at (12, 0) {\( \hat{\vpi}_{-i}^{i,t} \)};
    
    \node[policy] (policy) at (9, 2) {\( \pi(\cdot \mid h_t^i, m_t^i; \theta_i) \)};
    \node (a) at (12, 2) {\( a_t^i \)};
    
    \draw[arrow] (h) -- (encoder);
    \draw[arrow] (encoder) -- (m);
    \draw[arrow] (m) -- (decoder);
    \draw[arrow] (decoder) -- (pi_hat);
    
    \draw[arrow] (m) |- (policy);
    \draw[arrow] (h) |- (policy);
    \draw[arrow] (policy) -- (a);
    
    \end{tikzpicture}
    \caption{
  Encoder-decoder architecture for agent modeling.
  \figtaken{Fig 9.21 of \citep{marl-book}.}
    \figthanks{Stefano Albrecht}.
  }
\label{fig:agent-modeling}
\end{figure}

Instead of using self-play,
we can learn an \keywordDef{opponent model}.
In the CTDE paradigm, where each agent sees the other agents actions,
agent $i$  can use supervised learning to predict the
actions of agent $j$ given $i$'s observations.
There are many possible opponent models
we can use (see e.g. \citep{Albrecht2018} for a review).
For example, we can train an encoder-decoder network
to predict the actions of other agents  via a bottleneck,
and then pass this bottleneck embedding to the policy
as side information,
as proposed in \citep{Papoudakis2021}.
In more detail, let
$m_t^i = f^e(h_t^i;\psi_i^e)$ be the encoding of $i$'s history,
which is then passed to the decoder $f^d$ to predict the other agents
actions using $\hat{\vpi}_{-i}^{i,t} = f^d(m_t^i;\psi_i^d)$.
In addition, we pass $m_t^i$ to $i$'s policy to compute
the action using
$a_t^i \sim \pi(\cdot|h_t^i,m_t^i;\theta_i)$.
This is illustrated in  \cref{fig:agent-modeling}.
(A similar method could be used to predict other properties
of agent $j$, as long as their as observable by agent $i$.)

\subsection{Best response}
\label{sec:BR}

In this section, we discuss MARL algorithms that can provably
converge to a Nash equilibrium, even for zero-sum, imperfect-information
games, unlike basic policy learning methods based on self-play
(or opponent modeling).
The approach we use is built on the concept of a best response.
This is the action (for a given state)
that gives the highest expected
reward for agent $i$, given that the policies for all other
agents are fixed.
\eat{
Specifically, consider 
For each state $s$ that we encounter, we define a normal-form game
$\Gamma_s = (\calR_1=Q_1(s,\cdot),\cdots,R_n=Q_n(s,\cdot))$.
We can solve this game, using standard game theoretic methods,
to find a joint policy
$\vpi_s^* = (\pi^*_{s,1},\ldots,\pi^*_{s,n})$
with certain properties,
such as being a Nash equilibrium.
Once we have computed $\vpi_s^*$,
the \keyword{best response} for agent $i$ for this state is then given by
$\BR_i(s) =\vpi_s^*[i]$.
}

Specifcally, let $h^i$ be the information state for agent $i$
(i.e., action-observation history).
We compute its expected state-action value,
given the joint policy,
as follows:
\be
AV_i^{\vpi}(h^i,a^i) = \sum_{\va^{-i}}
  Q_i(\vh^i, (a^i,\va^{-i}))
  \prod_{j \neq i} \pi_j(a^j|h^i;\vtheta_j^i)
  \ee
The best response is then given by
\be
\BR_i(h^i) = \argmax_{a^i} AV_i^{\vpi}(h^i,a^i)
\ee
If there are a large number of actions,
we can approximate the  sum over $\va^{-i}$ using Monte Carlo
sampling.
The only thing left is to specify how to
learn the opponent policies.
We dicsuss this below.

\subsubsection{Fictitious play}

In \keywordDef{fictious play}, each agent $i$
estimates the policies of the other players,
based on their past actions.
It then computes a best response.
For example, imagine you're playing rock-paper-scissors repeatedly.
If you notice your opponent plays ``rock'' 60\% of the time so far,
your best response is to play ''paper'' more often.
You adjust your strategy based on the empirical frequency of their past moves.
The method is called ``fictious''
because each player is acting as if the opponents are playing a fixed strategy,
even though they’re actually adapting.

The method was originally developed for 
non-repeated normal-form games (which are stateless, so $h^i=[]$
and $Q_i(\va)=\calR_i(\va)$).
In this case, we can estimate the policies
by counting and averaging.
That is,
\be
\hat{\pi}_j^t(a^j) = \frac{C_j^t(a^j)}{\sum_{a'} C_j^t(a')}
\ee
where $C_j^t(a^j)$ is the number of times agent $j$ chose
action $a^j$ in episodes up to step $t$.
We then compute the
best response, given by
\be
\BR_i^t
= \argmax_{a^i} \sum_{\va^{-i}} \calR_i((a^i,\va^{-i})
    \prod_{j \neq i} \hat{\pi}_j^t(a^j)
\ee
Equivalently we can say that $\pi_i^t$ is the best response
to $\overline{\vpi}_{t-1}=\text{avg}(\vpi_1,\ldots,\vpi_{t-1})$.
For two-player zero-sum  finite games,
this procedure will converge to a NE.
That is,
the exploitability of the average $\overline{\vpi}_t$
generated by FP converges to zero
as $t$ grows large.

\subsubsection{Neural fictitious self play (NFSP)}
\label{sec:NFSP}

We can extend FP
to the partially observed, non-tabular setting as follows.
If we assume each agent sees the other agents actions,
then it is easy to learn an opponent model
representing their average stategy.
In particular,  agent $i$ will learn a model
of $j$'s policy, given $i$'s state (history),
which we
denote by $\overline{\pi}_{j|i}^t(a^j|h^i)$.
We fit this
by minimizing the cross entropy loss
\be
\loss(\overline{\pi}_{j|i}^t) = \expectQ{
  -\log \overline{\pi}_{j|i}^t(a_k^j|h_k^i)}
{k \sim U(1,t), (h_k^i, a_k^j) \in \calD_t}
\ee
where $\calD_t$ is the replay buffer containing
previous states and actions  of all the players.
In addition, we use DQN to learn  $Q_i(h^i,\va)$
for each agent.
We then use this learned average policy, plus the Q functions,
to compute $AV_i$, and hence the best response.

In the zero-sum two-player case,
we can use self-play,
so we just assume $\overline{\pi}_{-i}$
is the opposite of $\overline{\pi}_i$,
which can be learned by supervised learning
applied just to its own states and actions.
This is called \keywordDef{fictitious self play}
\citep{Heinrich15FSP}.
It was extended to the neural net case in
\citep{Heinrich2016},
who call it  \keywordDef{neural fictitious self play}.
If the  $Q$ function converges to the optimal
function, then this process converges to a NE,
like standard FP.

\eat{
In \keywordDef{fictitious play}, each agent updates its beliefs
about its opponent's policy using an empirical estimate
based on their actions.
This is known as
an  \keywordDef{agent model}
or \keywordDef{opponent model},
and  can take many forms
(see e.g. \citep{Albrecht2018} for a review);
It then computes its best response to this learned policy
at each step.
We give more details below.
}

\subsection{Population-based training}
\label{sec:PBT}

In \cref{sec:selfPlay}, we discussed the concept of
\keywordDef{self-play}, which is a way to train an agent
to play a two-player game by modeling the opponent
as using the same policy as the agent itself.
To avoid overfitting, we typically train against multiple
versions of the agent's own policy.
This concept can be generalized to work with general-sum games
with two or more players, by training against a population
of different policies.
This is called \keywordDef{population based training}
\citep{Jaderberg2019}.

\subsubsection{PSRO (policy space response oracle)}
\label{sec:PSRO}

In this section, we describe the \keywordDef{policy space response
  oracle} or \keywordDef{PSRO} method of
\citep{Lanctot2017}, which is a game-theoretic
instance  of population based training,
which can compute policies that satisfy various
solution concepts for any kind of stochastic game,
including partially observed, general sum games.

The idea behind PSRO is as follows.
At generation  $k$, each agent $i$
has a finite set of policies it can use, denoted $\Pi_i^k$.
We can define a normal-form \keywordDef{meta-game} $M^k$ from this by letting
each agent choose one of its policies,
where the reward for the joint action $\va=\vpi=(\pi_1,\ldots,\pi_n)$
is given by $\calR_i(\vpi)=U_i(\vpi)$ for $i=1:n$.
These returns can be estimated empirically
by simulating $n$ agents interacting with each other according
to these policies in the underlying game $G$.
Once we have determined the reward matrix,
we can solve for some kind of equilibrium solution (e.g., Nash
equilibrium), using a \keywordDef{meta-strategy solver}.
We can then extract the  probability distributions
over policies (aka strategy)
$\sigma_i^k$ for each agent.
To ensure this distribution is diverse,
we can enforce a lower bound that $\sigma_i^k(\pi_i)>\epsilon$,
so each policy is used some fraction of the time.

We can now expand the set of policies for each agent
by using an oracle to compute a new policy $\pi_i'$
and adding it to $\Pi_i^k$ to create the set $\Pi_i^{k+1}$.
For example, the oracle can compute a best response
\be
\pi'_i \in \argmax_{\pi_i}
\expectQ{U_i((\pi_i, \vpi_{-i}))}
{\vpi_{-i} \sim \vsigma_{-i}^k}
\ee
where $\vsigma_{-i}^k(\vpi_{-i})=\prod_{j \neq i} \sigma_j^k(\pi_j)$.
We can  compute $\pi'_i$ by using a single agent RL algorithm
in the underlying game $G$.
Since the policies of the other agents are uncertain,
as well as the environment,
we can use Bayesian RL methods,
such as \citep{Oliehoek2014}.
In this approach,
in each episode the policies of the other agents
$j \neq i$ are sampled from $\pi_j \sim \sigma_j^k$,
and then standard RL is applied.

It can be shown that, if
PSRO uses a meta-solver that computes exact Nash equilibria for the
meta-game,
and if the oracle computes the exact best-response policies
in the underlying game $G$, then the distributions
$\{\sigma_i^k\}_{i  \in \calI}$ converge to a Nash equilibrium of $G$.
See \cref{fig:PSRO} for an example.

    \begin{figure}[t]
	\centering
	\setstretch{1.2}
	\begin{tabular}{ p{1em} c c c c c c }
		\toprule
		$k$ & $\Pol_1^k$ & $\Pol_2^k$ & $\sigma_1^k$ & $\sigma_2^k$ & $\pol'_1$ & $\pol'_2$ \\
		\midrule
		1 & \ul{R} & \ul{P} & $1$ & $1$ & S & P \\
		2 & R,\ul{S} & P & $(0,1)$ & $1$ & S & R \\
		3 & R,S & \ul{R},P & $(\frac{2}{3},\frac{1}{3})$ & $(\frac{2}{3},\frac{1}{3})$ & P & R/P \\
		4 & R,\ul{P},S & R,P & $(0,\frac{2}{3},\frac{1}{3})$ & $(\frac{1}{3},\frac{2}{3})$ & R & S \\
		5 & R,P,S & R,P,\ul{S} & $(\frac{1}{3},\frac{1}{3},\frac{1}{3})$ & $(\frac{1}{3},\frac{1}{3},\frac{1}{3})$ & R/P/S & R/P/S \\
		\bottomrule
	\end{tabular}
        \caption{PSRO for rock-paper-scissors.
          We show the populations $\Pi_i^k$,
          distributuons $\sigma_i^k$ and best responses $\pi_i'$
          for both agents over generations $k=1:5$.
          (We use the shorthand of R to denote the pure policy that always
          plays R, and similarly for S and P.)
          New entries added to the population are shown
          with an underline.
          At generation $k=3$, there are 2 best responses,
          R and P, so the oracle can select either of them.
          Similarly, at generation $k=5$, there are 3 best responses.
          }
        \label{fig:PSRO}
    \end{figure}

Note that PSRO can also be applied to general-sum, imperfect
 information games.
 For example, \citep{Li2023games} uses (information set) MCTS,
 together with a learned world model, to compute the best response policy
 $\pi'_i$ at each step of PSRO.

\eat{
 I am not asking you to cite this, but there’s a simplified population-based algorithm that we came up with that may interest you, in this paper ( https://arxiv.org/abs/2303.03196 ) called “PopRL” (short for population RL).

The idea is the following. Suppose you are in a repeated game (of rock-paper-scissors). You have a collection of expert bots (i.e. from actual competitions), B. Here’s a neat little algorithm that finds a compromise between self-play and (pseudo-conditional) best response to the population: 

Run basic independent IMPALA (or any other PG method) with the following changes:
At the start of an episode, with probably p: sample an opponent from the bot population (with its index i < |B| as an identifier ). With probability 1-p, sample yourself – with index |B|.
Now, run an episode of IMPALA as usual playing against the sampled opponent, using an auxiliary prediction loss (classification target for the index that identifies the opponent) as an extra training signal in addition to the normal RL signal

Much simpler than e.g. full PSRO.

This does the best among all the learning agents we tried, and it also does a very simple form of opponent modeling. What’s kind of neat is you get some form of interpretability out of it too (the outputs of the auxiliary prediction head gives a distribution over the agent it thinks it’s playing against). We had fun playing against these agents and visualizing those distributions :) 

Mentioning it because it seemed like you were interested in MBRL. The augmented version would be to do this in sequential games with more complex predictions.
}

\subsubsection{Application to StarCraft (AlphaStar)}
\label{sec:alphaStar}

The \keywordDef{AlphaStar} system of
\citep{Vinyals2019}
used a PSRO-like method, combined with the (single agent) A2C RL algorithm,
to achieve grandmaster status in the challenging real-time strategy game
known as StarCraft II.\footnote{
In StarCraft II, the AI agent controls an entire army,
and must defeat a human opponent,
making this a zero-sum, two-player game,
which can be solved using deep RL with self-play.
This is different from
the StarCraft Multi-Agent Challenge (\keywordDef{SMAC})
\citep{Samvelyan2019},
which is a cooperative, partially observed multi-agent game,
where the agents (corresponding to individual units)
must work together as a team to defeat
a fixed AI opponent in certain predefined battles.
}
In particular, it used the following steps:
Build a pool of agents that represent different playstyles and skill
levels (known as a league);
Compute best responses to existing strategies;
Update a meta-strategy to mix agents in a way that approximates a Nash equilibrium;
select opponents from the Nash mixture to ensure robustness;
and train a new agent against the weighted mixture of past opponents.
See the paper for more details.

\subsection{Counterfactual Regret Minimization (CFR)}
\label{sec:CFR}

In this section we describe
\keywordDef{Counterfactual Regret Minimization}
(\keywordDef{CFR}),
which is an algorithm for imperfect
information, two-player, zero-sum game.
In \citep{Zinkevich2007}
they show that when using this procedure, the average policies converge to an epsilon-Nash.

\eat{
CFR converges faster than fictitious play
and PSRO, but
 requires access to the world model,
in order to simulate possible trajectories
 under counterfactual conditions.
It is also a much more complex algorithm.
}

\subsubsection{Tabular case}

Let $\traj=(s_0,\ldots,s_t)$
be a trajectory of world states.
Let $\eta^{\vpi}(\traj)$ be the probablity
of this trajectory under the joint policy.
(Note that stochastic dynamics of the world
are modeled by the policy of the chance player.)
We can decompose this as
\be
\eta^{\vpi}(\traj) =
\eta_i^{\vpi}(\traj) \eta_{-i}^{\vpi}(\traj)
\ee
Similarly define $\eta^{\vpi}(\traj,\vz)$
as the probability of the trajectory
$\traj=(s_0,\ldots,s_t)$
followed by $\vz=(s_{t+1},\ldots,s_T)$,
which is some continuation
that ends in a terminal state.
Let $Z(\vh^i) = \{(\traj,\vz)\}$
be the set of trajectories
and their terminal extensions
which are  compatible
with $\vh^i$, in the sense that $\calO_i(\traj)=\vh_i$,
and which end in a terminal state.
Also, let $\traj a_i$ be the trajectory followed
by action $a^i$.
We then define the \keywordDef{counterfactual state-action value}
for an information state as
\be
q_{\vpi,i}^c(\vh^i,a^i)
= \sum_{(\traj,\vz) \in Z(\vh^i)}
\eta_{-i}^{\vpi}(\traj) \eta^{\vpi}(\traj a^i, \vz)
u_i(\vz)
\label{eqn:CFRq}
\ee
The counterfactual state-value is
\be
v_{\vpi,i}^c(\vh^i) = 
\sum_{a^i} \pi_i(a^i|\vh^i) q_{\vpi,i}^c(\vh^i, a^i)
\ee
Finally, define the
instantaneous \keywordDef{counterfactual regret}
for player $i$ at iteration $k$ to be
\be
r_i^k(\vh^i,a^i) = q_{\vpi^k,i}^c(\vh^i,a^i)
- v_{\vpi^k,i}^c(\vh^i)
\ee
Note that this is the counterfactual version of an
advantage  function, as explained in \citep{Srinivasan18RPG}.
Similarly we define
 the cumulative counterfactual regret
to be
\be
R_i^k(\vh^i,a^i)=\sum_{j=0}^k r_i^j(\vh^i,a^i)
  \ee

CFR starts with a uniform random joint policy
$\vpi^0$ and then updates it at each iteration
by performing \keywordDef{regret matching}
\citep{Hart2000,Farina2023}.
That is, it updates the policy as follows
\begin{align}
  \pi_i^{k+1}(\vh^i,a^i)
  = \begin{cases}
    \frac{R_i^{k,+}(\vh^i,a^i)}
         {\sum_{a \in \calA_i(\vh^i)}
           R_i^{k,+}(\vh^i,a)
         }
         & \mbox{if denominator is positive} \\
    \frac{1}{|\calA_i(\vh^i)|} &\mbox{otherwise}
    \end{cases}
  \end{align}
where $x^+=\max(x,0)$.

In \citep{Zinkevich2007} they show that the above procedure results
in an $\epsilon$-Nash equilibrium,
where $\epsilon=O(\max_i |\calH_i| \sqrt{|\calA_i|}/ \sqrt{t})$,
for any two-player, zero-sum game (with perfect recall
of past observations).

\subsubsection{Deep CFR}

In practice, the  expectations over trajectories
in \cref{eqn:CFRq}
can be approximated using Monte Carlo
sampling \citep{Lanctot2009}.
In addition, we can 
approximate the tabular $q$, $v$ and $r$ terms
with  neural networks; this is called
\keywordDef{Deep CFR} \citep{DeepCFR},
which builds on the earlier
Regression CFR method of \citep{Waugh2015}.

\subsubsection{Applications to Poker and other games}

The first known combination of CFR with neural networks was DeepStack
\citep{deepstack}.
This was also  was one the first systems
to beat professional players at a
two-player poker variant called
\keywordDef{heads-up no-limit Texas hold'em}.
Another system that came out at the same time,
and also beat humans at this game,
was
the (neural-free) \keywordDef{Libratus} method
of \citep{libratus},
based on regret matching.
Libratus was later extended to make
the \keywordDef{Pluribus} method of \citep{pluribus},
which was able to beat human players
at the  six-player version of Texas hold'em.

\eat{
In \citep{Brown2020},
they proposed 
\keywordDef{ReBel},
which can be applied to general imperfect-information
two-player zero-sum games.
(It is essentially a two-player extension of AlphaZero,
discussed in \cref{sec:alphaZero}.)
}

In \citep{Schmid2021},
they proposed a method called \keywordDef{Student of Games},
that is a version of AlphaZero where CFR is the policy improvement operator.
This was applied to various games,
such as Chess, Go, Poker, and Scotland Yard.

\subsection{Regularized policy gradient methods}

In this section, we discuss policy gradient methods
that incorporate a regularization term to ensure convergence,
even in adversarial settings, such as 2p0s games.

\subsubsection{Magnetic Mirror Descent (MMD)}
\label{sec:MMD}

In \citep{SokotaMMD},
they present the \keywordDef{Magnetic Mirror Descent}
or \keywordDef{MMD} algorithm, which is
designed for two-player zero-sum games
(but which can also work well for single player games).
MMD is a modication
of policy gradient that adds additional regularizers
to ensure it converges (unlike traditional PG methods,
which can oscillate).
In the tabular case,
we use an update of the following
form, applied at each decision point (state) $s$
and for each agent $i$ separately:
\begin{align}
\pi_{k+1}
&= \argmax_{\pi} \langle \pi,  q_k \rangle
- \alpha \KL(\pi,\rho)   -\frac{1}{\eta} \KL(\pi,\pi_k)
\label{eqn:MMD}
 \end{align}
where $q_k(a)=q_k(s,A)$ is the value of action $a$ in state $s$,
$\pi(A) = \pi(A|s)$ is the agent policy,
$\langle \pi,  q_k \rangle = \E_{a \sim \pi} q_k(a)$
is an expectation,
$\rho$ is a magnet policy (designed to prevent oscillation),
$\alpha$  is a regularization term (corresponding to entropy
penalty if $\rho$ is uniform),
and $\eta$ is a stepsize.
For discrete actions, the optimal solution to the above
is given by the following (computed elementwise)
\be
\pi_{k+1} \propto
[\pi_k \rho^{\alpha \eta} e^{\eta q_k}]^{\frac{1}{1+\alpha \eta}}
\ee

If we drop the magnet term, by setting $\alpha=0$,
the method is equivalent
to the \keywordDef{mirror descent policy optimization} or  \keywordDef{MDPO}
algorithm of \citep{Tomar2020}.
In this case,  the optimal solution is given by
\be
\pi_{k+1} \propto [\pi_k  e^{\eta q_k}]
\ee
as in the exponentiated gradient algorithm.

\eat{
For parametric policies, MMD (with quadratic regularizer)
takes the form
\be
\theta_{t+1}^i = \argmin_{\theta}
\langle \nabla_{\theta_t}^i \loss_i(\theta_t^i), \theta \rangle
+ \frac{\alpha}{2}||\theta-\xi||_2^2
+ \frac{1}{2\eta} ||\theta-\theta_t^i||_2^2
\ee
where $\loss_i$ is the loss for agent $i$,
and $\xi$ is the parametric magnet.
For discrete actions, the solution is given by
\be
\theta_{t+1}^i = [\theta_t^i + \alpha \eta \rho \xi
  - \eta \nabla_{\theta_t^i} \loss_i(\theta_t^i)]
\frac{1}{1+\alpha \eta}
\ee
}

In \citep{SokotaMMD}, they prove that this procedure
(when used with a uniform magnet policy and applied to NFGs)
will converge to a QRE (\cref{sec:QRE}) exponentially fast.
If the entropy term is annealed to 0,
they can match the results of CFR (\cref{sec:CFR})
in the case of tabular games.
Their theory does not yet apply
to the parametric  case,
but  experimentally they still find fast convergence to the AQRE.

\subsubsection{PPO}

The MMD method of \cref{sec:MMD} is very similar to the PPO
algorithm of \cref{sec:PPO}.
In particular, the KL penalized version of PPO
uses the following loss
\be
\E_{s_t,a_t} \left[
  \frac{\pi(a_t|s_t)}{\pi_{\old}(a_t|s_t)}  A_{\old}(s_t,a_t)
  + \alpha \entropy(\pi(\cdot|s_t))
  - \beta \KL(\pi_{old}(\cdot|s_t), \pi(\cdot|s_t)) \right]
\ee
where $A_{\old}(s,a) = q_{\old}(s,a)- v_{\old}(s)$
is the advantage function.
By comparison,
if we use a uniform magnet for $\rho$,
the MMD loss in \cref{eqn:MMD}
 becomes
\be
\E_{s_t,a_t} \left[
  \pi(a_t|s_t)  q_{\old}(s_t,a_t)
  + \alpha \entropy(\pi(\cdot|s_t))
  - \beta \KL(\pi(\cdot|s_t), \pi_{old}(\cdot|s_t)) \right]
\ee
where $\beta=1/\eta$ is the inverse stepsize.
The main difference between these equations
is just the use of a reverse KL
instead of forwards KL. (The two expressions also differ
by the scaling factor $1/\pi_{\old}(a_t|s_t)$ and
the offset term $v_{\old}(s_t)$.)

Despite the similarities,
in \citep{SokotaMMD},
PPO has been shown to perform worse than MMD
on various 2p0s games.
One possible reason for PPO's  poor performance
is due  to the use forwards vs reverse KL
penalty.
However, \citep{Hsu2020PPO} compared the use of reverse KL
regularization instead of forward KL in PPO for Mujoco,
and found that the two yielded similar performance.
The explanation suggested in \citep{Rudolph2025}
is simply  that the hyper-parameters in PPO
(in particular, the entropy penalty $\alpha$)
was not tuned properly for the 2p0s setting
(the latter tending to require much larger values,
such as 0.05-2.0,
whereas single agent PPO implementations
usually use 0-0.01).

They  experimentally tested this hypothesis
by comparing PPO with various other algorithms
(including MMD, CFR (\cref{sec:CFR}), PSRO (\cref{sec:PSRO})
and NFSP (\cref{sec:NFSP}))
on a set of imperfect information games
(partially observed or ``phantom''/``dark'' versions
of Tic-Tac-Toe and 3x3 Hex,
where the agents actions are invisible
to the non-acting player).
They find that properly tuned policy gradient methods
(including both PPO and MMD) performed the best in terms
of having the lowest exploitability scores.
(The exploitability score is defined in
\cref{eqn:exploitability},
and was  computed by exactly solving for the optimal
opponent policy given a candidate learned policy.)

The above experimental result led the
authors of \citep{Rudolph2025}
to propose the following ``Policy Gradient Hypothesis'':
\begin{quote}
  Appropriately tuned policy gradient methods that share an ethos with magnetic mirror descent are competitive with or superior to model-free deep reinforcement learning approaches
  based on fictitious play, double oracle [population-based training],
  or counterfactual regret minimization in two-player zero-sum imperfect-information games.
  \end{quote}

If true, this hypothesis would be very useful, since it means we can use
standard single agent policy gradient methods, such as (suitably tuned) PPO,
for multiplayer games,
both cooperative  (see \citep{MAPPO})
and  adversarial (see \citep{Rudolph2025}).

\subsection{Decision-time planning methods}
\label{sec:DTPMA}

In this section we focus on decision-time planning (DTP) methods,
that improve upon a base policy (known as a blueprint policy)
by doing some kind of forward search (from the current state) in a world model,
as discussed in \cref{sec:DTP}.
We focus
the \keywordDef{update-equivalent DTP} method
of \citep{Sokota2023update},
which makes a connection between DTP and other policy update
algorithms.

Recall from
\cref{eqn:policyUpdate}
that the  policy iteration algorithm
can be viewed as
 performing an update to the policy at each
 step based on acting greedily wrt $Q(s,a$):
 \be
 \policy_{\new}(s)
 = \argmax_a R(s,a) + \gamma \expect{\Vpol(s')}
 = \argmax_a Q(s,a)
\ee
 If we consider a single state, we can write this update as
 \be
 \pi_{\new} = U(\pi,q) =
 \argmax_{\pi' \in \Delta(A)} \langle \pi', q \rangle
 \ee
One way to estimate the action values $q$
for the current state is to perform \keywordDef{Monte Carlo search}
or \keywordDef{MCS}
\citep{Tesauro1996},
which unrolls possible futures using the current policy,
as in DTP.
Thus with enough samples, DTP (with the correct world model,
and a suitable exploratory policy)
combined with this update
will give the same results as (asynchronous) policy iteration.

\subsubsection{Magnetic Mirror Descent Search (MMDS)}

\begin{algorithm}
\dontprintsemicolon
\caption{Magnetic Mirror Descent Search (MMDS)}
\label{algo:MMDS}
Input: current state $h_t^i$,
joint policy $\vpi$,
agent id $i$\\
$q[a]=0$, $N[a]=0$ for each action $a \in \calA_i$ \\
\Repeat{until search budget exhausted}
{
  Sample current world state using agent's local belief state:
  $s_t \sim P_{\vpi}(\cdot|h_t^i)$ \\
  \For{$a \in \calA_i$}
     {
       Sample return $G_{\geq t}  \sim P_{\vpi}(G_{\geq t} | s_t,a)$
       by rolling out $\vpi$ in world model starting at $s_t$\\
       $q[a] = q[a] + G_{\geq t}$ \\
       $N[a] = N[a] + 1$ \\
     }
     $q[a] = q[a] / N[a]$ for $a \in \calA_i$ \\
     Return $U(\pi^i(h_t^i), q)$ by performing
     SGD on \cref{eqn:updatePolicy}.
}
\end{algorithm}

In  \citep{Sokota2023update}
they propose to generalize this idea
to the multi-agent setting
by using the MMD algorithm from \cref{sec:MMD} as the update operator.
They call this \keywordDef{magnetic mirror descent search} or
\keywordDef{MMDS}.
The local policy update (for player $i$) has the form
\be
\pi_{\new} = U(\pi,q)
=  \argmax_{\pi' \in \Delta(A)} \langle \pi', q \rangle
-\alpha \KL(\pi', \rho) - \frac{1}{\eta} \KL(\pi', \pi)
\ee
where $\pi$ is the previous local (\keywordDef{blueprint}) policy 
and $\rho$ is the local magnet policy (which can be taken as uniform).
If $h_t^i$ is the current state (root of search tree for player $i$),
and the actions are discrete, we can equivalently
perform an SGD step on
the following parametric policy loss:
\be
\loss(\vtheta) = \sum_a
\left[ \pi_{\vtheta}(a|h_t^i) q(h_t^i, a)
  - \alpha \pi_{\vtheta}(a|h_t^i) \log \frac{\pi_{\vtheta}(a|h_t^i)}{\rho(a)}
  - \frac{1}{\eta}
  \pi_{\vtheta}(a|h_t^i) \log \frac{\pi_{\vtheta}(a|h_t^i)}{\pi_{\old}(a|h_t^i)}
  \right]
 \label{eqn:updatePolicy}
\ee
See \cref{algo:MMDS} for the pseudocode.

Note that,
if we use a uniform magnet, this is equivalent to adding
an entropy regularizer.
Also,
for common-payoff games, we can  drop the magnet term,
which gives rise to the simpler
\keywordDef{mirror descent search}
method.

\subsubsection{Belief state approximations}

To implement this algorithm, we need to sample
from  $P_{\vpi}(s_t|h_t^i)$,
which is the distribution over world states
given agent $i$'s local history.
One approach to this is to use particle filtering,
cf. \citep{Lim2023}).

Another approach is to train a \keywordDef{belief model}
to predict the other player's private information,
and the underlying
environment state,
given the current player's history,
i.e., we learn to predict
$P(s_t, \{h_t^j\} | h_t^i)$.
In the \keywordDef{learned belief search}
(\keywordDef{LBS})
method of  \citep{Hu2021LBS}
(which was designed for Hanabi, which is a Dec-POMDP),
rather than predicting the entire action-observation history
for each agent, they just predict the private information
(card hand) for each agent (represented as a sequence of tokens).
This can be used (together with the shared public information)
to reconstruct the environment state.
They train this model (represented as a seq2seq LSTM)
using supervised learning,
where agent $i$ learns to predict its own private
information given its public history.
At test time, agent $i$ uses $j$'s public history as
input to its model to sample $j$'s private information.
(This assumes that $j$ is using the same blueprint policy
to choose actions that $i$ used during training.)
Given the imputed private information,
it then reconstructs the environment state and performs rollouts,
using the joint blueprint policy,
in order to locally improve its own policy.

\subsubsection{Experiments}

In  \citep{Sokota2023update},
they implemented the above method and applied it to several
imperfect information games (using the true known world model)
For the common-reward  game of Hanbai (5 card and 7 card variants),
they used PPO to pretrain the blueprint policy,
and they pretrained  a seq2seq belief model.
At run time, they  use 10k samples for each step of MDS
to locally improve  the policy
(which takes about 2 seconds).
They observed modest gains over rival methods.
For the 2p0s games, they used the partially observed
(dark/phantom) versions of 3x3 Hex and Tic-Tac-Toe.
For belief state estimation,
they use a particle filter with just 10 particles,
for speed.
As a blueprint policy
they consider uniform and MMD (for 1M steps).
They find that MMDS can improve the blueprint,
and this combination beats 
baslines such as PPO and NFSP.
They also compare to MMD as a baseline.
For the MMD-1M baseline, the blueprint matches
the baseline (by construction), but the MMDS version beats it.
However, the MMD-10M baseline beats MMDS,
showing that enough offline computation
can beat less online computation.

\subsubsection{Open questions}

It is  an interesting open question how well this MMDS method
will work when the world model needs to be learned,
since this results in rollout errors,
as discussed in \cref{sec:MBRLgame}.
Similarly errors in the belief state
approximation may adverseley affect the estimate of $q$
for the root node.

In addition, it is an open question
to prove convergence properties
of the generalized version of  MMDS,
that uses more than just action value feednack.
For example, 
MCTS updates the local policy at internal nodes,
not just the root node.
In some cases, MCTS can work better than simple MCS,
although this is not always the case
(see e.g., \citep{Hamrick2021}).

\subsection{MARL for LLM agents}

A recently growing  trend is to use LLMs as agents
(see \cref{sec:LLM}), which can be made to interact with each other,
via protocols such as \keywordDef{A2A} (agent-to-agent).\footnote{
See \url{https://developers.googleblog.com/en/a2a-a-new-era-of-agent-interoperability/}.
} %
It is possible to apply MARL techniques to optimize such systems,
see e.g., \citep{Liao2025,Zhou2025econ,Ren2025MCP}.
See also \cref{sec:LLMselfPlay} for a discussion of how to use
self-play to train LLMs, 
and \citep{Sun2025GT} for a broader review of game theory and LLMs.



\chapter{LLMs and RL}
\label{chap:LLM}
\label{sec:LLM}

\section{Introduction}

In this section, we discuss connections between RL and
\keywordDef{foundation models},
also called  \keywordDef{large language models} or \keywordDef{LLMs}.
LLMs are generative models (usually based on auto-regressive transformers)
which are trained on large amounts of web data.\footnote{
When also trained on visual data, LLMs are sometimes
called  \keywordDef{vision language models} or \keywordDef{VLMs},
and when trained on action data,
they are called \keywordDef{vision language action} models
or \keywordDef{VLAs}
\citep{Kawaharazuka2025}.
}
More details on the connections between RL and LLMs can be found in
e.g., \citep{Pternea2024,Hao2025RL,Zhang2025landscape,Khatri2025},
and more details on LLMs in general
can be found in  e.g., \citep{Xiao2025LLM,LLMbook,Brown24GAIHB}.


\section{RL for LLMs}
\label{sec:RL4LLM4}

In this section, we discuss how to use RL to improve
the performance of LLMs.
This  is a fast growing field, so we
only briefly mention a few highlights.
For more details, see e.g. \citep{Lambert2025,Brown24GAIHB}
and other references online.

\subsection{RL fine tuning (RLFT)}

LLMs 
are usually trained with behavior cloning,
i.e., MLE on a fixed dataset,
such as  a large text
corpus scraped from the web.
This is called \keywordDef{pre-training}.
We can then improve their performance using
various \keywordDef{post-training} methods,
which are designed to improve their capabilities
and \keywordDef{alignment}
with human preferences (see e.g., \citep{Zeng2025align}),
as opposed to just being generative models
of the data seen on the web.
A simple  way to perform post-training is to use
\keywordDef{instruction fine tuning},
also called \keywordDef{supervised fine-tuning} (or \keywordDef{SFT}),
in which  we collect human demonstrations of (prompt, response) pairs,
and fine-tune the model on them.
However, it is very difficult to collect sufficient quantities
of such data.
An alternative to demonstrating good behaviors
is to use RL to train the model
using a suitable reward function.
(We discuss where these reward functions come from in
\cref{sec:rewardForLLM}.)
This is called
\keywordDef{reinforcement learning fine-tuning} or \keywordDef{RLFT}.

RLFT can be  preferable to SFT for several reasons.
First,
it is often the case that  verification is  easier than generation
(e.g., it is
easier to ask people which answer they prefer rather
than to ask them to generate good answers,
an insight we exploit in \cref{sec:RLHF}).
Second, RL can be used to learn a set of
``thinking actions'',
which are created in response to the question
before generating the answer
(see \cref{sec:thinking}).
For complex problems (e.g., in math),
this tends to work much better than trying to directly
learn an input-output mapping
\citep{Prystawski2023}.
(It is possible to use SFT on explicitly provided
thinking traces, but it has been found that RL can generalize
more reliably \citep{Chu2025}.)
Finally, RL  opens the path to super-human performance
\citep{Silver2025},
going beyond whatever supervised examples humans can create.

\subsection{Reward models}
\label{sec:rewardForLLM}

In this section, we discuss different kinds of reward
functions that are used for RLFT.

\subsubsection{RL with verifiable rewards (RLVR)}

For problems such as math and coding,
it can be easy to determine if an answer is correct,
by checking equality between the generated answer
and the true answer (for math),
or checking if a set of unit tests pass (for code).
This allows us to define a binary reward signal.
Using RL with such a reward is called
``\keywordDef{RL with   verifiable rewards}'' or
\keywordDef{RLVR}
(see e.g., \citep{Zelikman2022,Lambert2024}).
We will use this approach to train ``thinking models''
in \cref{sec:thinking}.

\subsubsection{Process vs outcome reward models}
\label{sec:PRM}

If the reward function $R(s_t)$ is defined on partial trajectories,
it is called a \keywordDef{process reward model} or \keywordDef{PRM}.
This provides a form of dense feedback.
If the reward is just defined on the final sequence $R(s_T)=R(s_0,\va_{1:T})$,
it is called an \keywordDef{outcome reward model} or \keywordDef{ORM},
and corresponds to a sparse reward.
For example, suppose we are solving a math problem using a thinking model
(see \cref{sec:thinking}):
if we just check the final answer, we have an ORM,
but if we also check correctness of the intermediate proof steps,
we have a PRM.
Note that
a PRM is  related  to a value function (that models expected future reward),
and is typically harder to learn than an ORM.

\subsubsection{Learning the reward model from human feedback (RLHF)}
\label{sec:RLHF}

To train  LLMs to do well in general tasks,
such as text summarization or poetry writing,
it is common  to use
\keywordDef{reinforcement learning from human feedback}
or \keywordDef{RLHF},
which refers to learning a reward model from human data,
and then using RL to train the LLM to maximize this.

The basic idea is as follows.
We first generate a large number of (context, answer1, answer2) tuples
either by a human or an LLM.
We then ask human raters  if they prefer answer 1 or answer 2.
Let $x$ be the prompt (context), and $y_w$ be the winning (prefered) output,
and $y_l$ be the losing output.
Let $r_{\vtheta}(x,y)$ be the reward assigned to output $y$.
(This model is typically  a shallow MLP on top of the last layer
of a pretrained LLM.)
We train the reward model  by maximizing the likelihood of the observed preference
data.
The likelihood function is given by
the \keywordDef{Bradley Terry choice model}:
\be
p_{\vtheta}(y_w > y_l)
= \frac{\exp(r_{\vtheta}(x, y_w))}
{\exp(r_{\vtheta}(x, y_w)) + \exp(r_{\vtheta}(x, y_l))}
\label{eqn:bradleyTerry}
\ee
We thus need to maximize
\begin{align}
  J(\vtheta) &=
  \E_{(x,y_w,y_l) \sim \data} \left[
\frac{\exp(r_{\vtheta}(x, y_w))}
{\exp(r_{\vtheta}(x, y_w)) + \exp(r_{\vtheta}(x, y_l))}
    \right]
  =
  \E_{(x,y_w,y_l) \sim \data} \left[
  \frac{1}{1 + \frac{\exp(r_{\vtheta}(x,y_l))}
    {\exp(r_{\vtheta}(x,y_w))}}  \right]  \\
  &=   \E_{(x,y_w,y_l) \sim \data} \left[
    \sigma(r_{\vtheta}(x,y_w) - r_{\vtheta}(x,y_l))
    \right]
\end{align}
Equivalently we can minimize
\begin{align}
  \loss(\vtheta) &=
    \E_{(x,y_w,y_l) \sim \data} \left[
      \log\left(1 + e^{r_{\vtheta}(x,y_l) - r_{\vtheta}(x,y_w)} \right)
      \right] \\
    &=
    -    \E_{(x,y_w,y_l) \sim \data} \left[
      \log\left(
      \sigma(r_{\vtheta}(x,y_w) - r_{\vtheta}(x,y_l)) \right)
      \right]
\end{align}

In some cases, we  ask human raters  if they prefer answer 1 or answer 2,
or if
there is a tie, denoted $y \in \{1,2,\emptyset\}$.
In this case, we can optimize
\begin{align}
\loss(\vtheta) &=   \E_{ (x,y_1,y_2,y) \sim \data} \left[
  \ind{y=1} \log p_{\vtheta}(y_1 > y_2|x)
  +    \ind{y=2} \log p_{\vtheta}(y_1 < y_2|x) \right.
\\
  & \left.
  +    \ind{y=\emptyset} \log p_{\vtheta}(y_1 > y_2|x)
  p_{\vtheta}(y_1 < y_2|x)
  \right]
\end{align}

For a discussion of some of the implementation details of RLHF,
see  \citep{Lambert2025}.
For a discussion of other ways of modeling preferences and rewards,
based on \keywordDef{social choice theory},
see \citep{Conitzer2024,Ge2024axioms}.

\subsubsection{Learning the reward model from AI feedback (RLAIF)}
\label{sec:RLAIF}

Instead of asking humans their preferences for each possible input example,
we can ask an LLM to predict the preference.
This is called  \keywordDef{LLM as judge}.
We can then fit the reward model to this synthetically labeled data,
just as in RLHF.
Alternatively, we can just ask the LLM to predict the reward directly.
This is called  \keywordDef{RLAIF}, which stands for
RL from AI feedback.
It is also common to use VLMs for RLAIF
(see e.g., \citep{Du2023success,Baumli2023,Huang2023language,Sontakke2023,
  Rocamonde2024}).

In order to specify how to judge things, the LLM
needs to be prompted.
Anthropic (which is the company that makes the Claude LLM)
created a technique called
\keywordDef{constitutional AI}
\citep{constitutionalAI},
where the prompt is viewed as a
``constitution'',
which specifies what kinds of responses are desirable
or undesirable.
With this method, the system can critique its own outputs,
and thus self-improve.

\subsubsection{Generative reward models (GRM)}
\label{sec:GRM}

A \keywordDef{generative reward model} or \keywordDef{GRM}
predicts the reward for a given response,
but also returns its chain of thought,
thus providing richer textual feedback.
For example,
it might generate
``This response is helpful but makes a factual error about X,
so overall reward is 0.3''.
In addition to passing the scalar reward to an RL algorithm,
the textual feedback can be parsed by the LLM itself
to decide how to improve the policy.
For example, the \keywordDef{GEPA} algorithm
of \citep{GEPA} uses an evolutionary algorithm
to optimize prompts for a frozen LLM,
by mutating them given textual feedback from a GRM.


\subsection{Agents which ``think''}
\label{sec:thinking}


In this section, we discuss how to leverage the power of LLMs
to create agents that ``think'' before they act.

\subsubsection{Chain of thought prompting}
\label{sec:COT}

The quality of the output from an LLM
can be improved by prompting it to ``show its work''
before presenting the final answer.
These intermediate tokens are 
called a ``\keywordDef{Chain of Thought}'' \citep{COT}.
Models that act in this way are often
said to be doing ``\keywordDef{reasoning}'' or ``\keywordDef{thinking}''
(see e.g., \citep{Zhang2025LRM,Xu2025LRM}),
although in less anthropomorphic terms,
we can think of them as just policies with dynamically
unrolled computational graphs
\citep{Kambhampati2025}.
This is motivated by various theoretical results
that show that such CoT can significantly
improve the  expressive power of transformers
\citep{Merrill2024,Li2024COT}.

\subsubsection{Training a thinking model using RL}
\label{sec:deepseek}
\label{sec:R1}





Rather than just relying on prompting,
we can explicitly train a model to think by letting it generate a variable
number of tokens ``in its head'' before generating the final answer.
Only the final outcome is evaluated,
using a known reward function (as in the case
of math and coding problems).

\eat{
\footnote{
When training thinking models,
we can either use
an \keywordDef{Outcome Reward Model} (ORM),
which only rewards the final answer (not the thinking tokens),
or we can use a \keywordDef{Process Reward Model}
or PRM,
which provides dense reward to intermediate tokens,
including thinking tokens (e.g., based on correctness
of partial steps in a derivation).
}
}

This approach was recently demonstrated by 
the \keywordDef{DeepSeek-R1-Zero} system
\citep{DeepSeekR1} (released by a Chinese company
in January 2025).
They started with a strong LLM base model,
known as \keywordDef{DeepSeek-V3-Base}
\citep{DeepSeekV3},
which was pre-trained 
on a large variety of data
(including Chains of Thought).
They then used a variant of PPO,
known as GRPO (see \cref{sec:GRPO})
to do RLFT, using a set math and coding benchmarks
where the ground truth answer is known.
The resulting  system got excellent performance
on math and coding benchmarks.\footnote{
Although DeepSeek-R1-Zero exhibited excellent performance
on math and coding benchmarks,
it did not work as well on more general reasoning benchmarks.
So their final system, called DeepSeek-R1,
combined RL training with more traditional SFT (on synthetically generated
CoTs).
}
The closed-source models
\keywordDef{ChatGPT-o1} and \keywordDef{ChatGPT-o3}
from OpenAI\footnote{
See \url{https://openai.com/index/learning-to-reason-with-llms/}.
}
and the
\keywordDef{Gemini 2.0 Flash Thinking}
model from Google Deepmind\footnote{
See \url{https://deepmind.google/technologies/gemini/flash-thinking}.
}
are  believed to follow similar principles to
DeepSeek-R1, although the details are not public.\footnote{
However, shortly after the release of R1,
the CRO of Open AI (Mark Chen) confirmed
that o1 uses some of the same core ideas  as R1:
\url{https://x.com/markchen90/status/1884303237186216272?s=46&t=Vx_O-TgDXth-Mt_kw6ggqw}.
}
For a recent review of these reasoning methods,
see e.g., \citep{Xu2025LRM,Zhang2025RL}.

\subsubsection{Thinking as marginal likelihood maximization}

Since we usually only care about maximizing the probability
that the final answer is correct,
and  not  about the values or ``correctness''
of the intermediate thoughts (since it can be hard to judge heuristic arguments),
we can view training a thinking model as equivalent to maximizing the marginal
likelihood $p(y|x)=\sum_z p(y,z|x)$, where $z$ are the latent thoughts
(see e.g., \cite{Hoffman2023,Zelikman2024,Tang2025}).

\subsubsection{Can we bootstrap a model to think from scratch?}

One reason DeepSeek-R1 got so much attention in the press
is that  during the training process,
it seemed to ``spontaneously''
exhibit some ``emergent abilities'',
such as generating increasingly long
sequence of thoughts, and using self-reflection to
refine its thinking, before generating the final
answer.

Note that the claim that RL ``caused'' these emergent abilities
has been disputed by many authors
(see e.g., \citep{r1zeroNUS,Yue2025reasoning}).
Instead, the general concensus is that the base model itself
was already trained on datasets that contained some COT-style
reasoning patterns.
This is consistent with the findings in
\citep{Gandhi2025cognitive},
which showed that applying RL to a base model that had not been pre-trained
on reasoning patterns (such as self-reflection)
did not result in a final model that could exhibit such behaviors.
However, RL can ``expose'' or ``amplify''
such  abilities in a base model
if they are already present to a certain extent.
(See also \citep{Foster2025} for a detailed
theoretical study of this issue.)

Recently  \keywordDef{Absolute Zero Reasoner}
from \citep{AZR} showed it is possible to automatically generate
a curriculum of (programs, inputs, outputs),
which is used to improve the math and coding abilities of the LLM
using RL.
In particular, the LLM is trained to perform
induction (predict $p$ given $(i,o)$ pairs),
deduction (predict $o$ given $p$ and $i$),
and abduction (predict $i$ given $p$ and $o$).
(See also related work in \citep{Kuba2025}.)

\subsubsection{Agentic AI}

There is currently a lot of hype around
\keywordDef{Agentic AI} systems,
that consist of a set of interacting LLMs, often called ``agents'',
which are essentially different prompts,
reflecting different roles or personas,
which can be given to the shared LLM
to make it act in different ways.
Typically these prompts, and the way the different agents
interact,
are hand-designed --- this is called
a \keywordDef{workflow} or \keywordDef{scaffolding}.
Such agents
may process the input, some may access or process memory
(see e.g., \citep{Zhang2024memory,Sumers2024}),
and some may call tools
such as  web search
(see e.g., \citep{Luo2025LLM,Gulli2025}).
Note, however, that 
 unlike the true multi-agent setup of \cref{chap:MARL},
these ``agents'' do not maximize their own reward functions,
and are really just a set of  modules
inside a single larger agent \citep{Han2024multi,Yan2025}.
These workflows are usually hand-engineered.
However, they can be improved using
RL (see e.g.,  \citep{Nguyen2025SFR,Huichi2025,Mo2025}).

\subsection{Algorithms for single-turn RL}
\label{sec:single-turn}

In this section, we discuss RL methods for training
LLMs to solve math and reasoning problems.
In this setup, there is just a single state,
namely the input prompt $s$;
the action is a sequence of tokens generated by the policy
in response,
and then the game ends.
This is equivalent to a  contextual bandit problem,
with sequence-valued input (context) and output (action).
(We consider the full multi-turn case in \cref{sec:multi-turn}.)

\subsubsection{Problem setup}

Formally, the goal is to maximize
\be
J(\vtheta) = \E_{s \sim \data, \va \sim \pi_{\vtheta}(\va|s)}[R(s,\va)]
\label{eqn:LLMbandit}
\ee
where $s$ is the context/prompt (sampled from the dataset),
and $\va$ is the generated sequence of actions (tokens)
sampled from the policy:
\be
\pi_{\vtheta}(\va|s) 
= \pi_{\vtheta}(a_1|s) \pi_{\vtheta}(a_2|s,a_1)
\cdots \pi_{\vtheta}(a_T|s_0,a_1,\ldots,a_{T-1})
=\prod_{t=1}^T \pi_{\vtheta}(a_t|\va_{1:t-1}, s)
\ee
Here  $T=|\va|$ is the length of the generated output
(which is terminated by generating an {\tt <eos>} token).

We can convert this into an MDP by defining the following deterministic
state transition
\be
p(s_{t}|s_{t-1},a_t) = \delta(s_{t} = \text{concat}(s_{t-1},a_t))
= \delta(s_{t}|s_{t-1},a_t)
\label{eqn:LLMMDP}
\ee
with initial distribution $\delta(s_0=s)$.
Thus the state $s_t$ is just the set of tokens
from the initial prompt $s$
plus the generated tokens up until time $t$.
This definition of state restores the Markov property,
and allows us to write the policy in the usual way as
$\pi_{\vtheta}(\va|s) =  \prod_{t=1}^T \pi_{\vtheta}(a_t|s_t)$.

We can also  rewrite the objective in standard MDP form
as follows:
\be
J(\vtheta) = \E_{s_0 \sim \data}
\left[ \sum_{t=1}^T \sum_{s_t} \sum_{a_t}
  \pi_{\vtheta}(a_t|s_t) \delta(s_t | s_{t-1},a_t)
  R(s_{t},a_t)
  \right]
\ee
where $R(s_t,a_t)$ is a sparse reward function
defined as follows:
\be
  R(s_t,a_t) =
  \begin{cases}
    0 & \text{if } t<T \\
    R(s_T,a_T={\tt eos})=R(s,a_1,\ldots,a_T) & \text{if } t=T
  \end{cases}
 \ee
where $R(s,a_1,\ldots,a_T)=R(\tau)$ is the final trajectory level reward.

In practice, the above approach
can overfit to the reward function,
so we usually regularize the problem to ensure the policy
$\pi_{\vtheta}$
remains close to the base pre-trained LLM $\pi_{\tref}$.
We can do this by adding a penalty of the form
$-\beta \KLpq{\pi_{\vtheta}(a_t|s_t)}{\pi_{\tref}(a_t|s_t)}$
to the per-token reward $R(s_t,a_t)$.

Below we discuss various algorithms to solve this optimization problem.

\eat{
In this section, we discuss algorithms for training
LLM policies that maximize the expected reward.
These methods are derived from general RL algorithms
by specializing to the
bandit setting (i.e., generating a single
answer in response to a single state/prompt).
}

\subsubsection{PPO}

A natural approach to 
training the LLM policy is to use PPO
(\cref{sec:PPO}).
In the bandit case, we  can write the objective as follows:
\be
J_{ppo}(\vtheta) = \E_{s_n \sim \data} \E_{\va_n \sim \pi_{\old}(\cdot|s_n)}
\min\left(\rho_n(\vtheta) \cdot A_n
\;
\text{clip}(\rho_n(\vtheta) \cdot A_n) \right)
\ee
where  we define
\begin{align}
\rho_n(\vtheta)
&=\frac{\pi_{\vtheta}(\va_n|s_n)}
    {\pi_{\old}(\va_n|s_n)}
\text{ // likelihood ratio} \\
s_n+\va_n &=\text{concat}(s_n,a_{n1},\ldots,a_{nT})  \text{ // full sequence}\\
A_n &= A(s_n+\va_n) = R_n - b_n \text{ // advantage} \\
R_n & =R(s_n+\va_n) \text{ // trajectory level reward} \\
b_n &= b(s_n+\va_n)=V(s_n+\va_n) \text{ // baseline}
\end{align}

For more details on PPO for LLMs, see  \citep{Huang2024}.

\eat{
\subsubsection{Soft Policy Optimization (SPO)}
\label{sec:SoftPO}

In \citep{Cohen2025}, they derive an algorithm called
\keywordDef{Soft Policy Optimization} (\keywordDef{SPO})
that can maximze the soft (maximum entropy) RL objective (\cref{sec:maxentRL})
from both off-policy and on-policy data,
using a combination of policy-based and value-based learning.
It is derived for the special case of LLM training with a terminal reward
(e.g., from RLHF or a verifiable reward).
They show that it is much faster to train, and gives better results,
than PPO (which is purely on-policy),
when evaluated on the task of generating code that
passes a binary verifier.
}

\subsubsection{GRPO}
\label{sec:GRPO}


Learning an actor (policy) and a value function (critic)
takes twice as much time and memory as just
learning a policy. This is problematic for large LLMs.
Therefore there have been a bunch of recent methods
that replace the parametric value function with MC estimators.\footnote{
The disadvantage is that
estimating the value with MC rollouts
requires that we can reset the environment,
so that we can generate multiple responses (action trajectories)
given the same initial state.
This is fine for question answering, but much harder
for multi-turn RL, which we discuss below.
Also, there is no credit assignment to intermediate states,
since we are only estimating the value of the initial state.
Thus the method is statistically quite inefficient.
}

The Group Relative PPO or
\keywordDef{GRPO} algorithm of \citep{DeepSeekMath},
which was used to train DeepSeek-R1-Zero
(discussed in \cref{sec:R1}),
is a variant of PPO which replaces the critic network
with a Monte Carlo estimate of the value function.
In more detail,
for each prompt $s_n$,
we generate $J$  answers $\va_n^j \sim \pi_{\old}(\cdot|s_n)$
(called a \keywordDef{group}, often of size $J \sim 8$)
which give final rewards $R_n^j$.
We then  compute the advantage
by subtracting the group average
and dividing by the group standard deviation:
\be
\hat{A}_n^j = \frac{R_n^j - \mu_n}{\sigma_n}
\ee
where
$\mu_n = \text{mean}(R_n^j: j=1:J)$
and
$\sigma_n = \text{std}(R_n^j: j=1:J)$.
The use of the normalization term ensures
the rewards are \keywordDef{calibrated},
so that a hard problem with low average reward may still
result in an update if the deviation from this low mean
is large
\citep{Balashankar2025,Mroueh2025}.
(This is related to the use of \keywordDef{reward centering}
in continual RL \citep{Naik2024}.)

Since the policy generates a sequence,
we can expand out the loss for each sequence
into a sum of per-token losses.
In GRPO, we set the step-level advantage to be equal to the normalized
trajectory-level advantage,
$\hat{A}_{nt}^j=\hat{A}_n^j$,
and
we define the likelihood ratio as
\be
\rho_{nt}^j(\vtheta)=
\frac{\pi_{\vtheta}(a_{nt}^j|a_{n,<t},s_n)}
    {\pi_{\old}(a_{nt}^j|a_{n,<t}, s_n)}
\ee
We then  normalize the clipped advantage by the
length of each sequence,
to ensure that long sequences don't dominate the loss:\footnote{
Some arguments against this kind of normalization
can be found in \citep{DrGRPO,r1zeroNUS}.
}
\be
J_{GRPO}(\vtheta)
= \frac{1}{N} \sum_{n=1}^N
\frac{1}{J} \sum_{j=1}^J
\frac{1}{|\va_n^j|}
\sum_{t=1}^{|\va_n^j|}
\min\left(
\rho_{nt}^j(\vtheta) \hat{A}_{nt}^j,
\text{clip}(\rho_{nt}(\vtheta),1-\epsilon,1+\epsilon)
\hat{A}_{nt}^j
  \right)
  \ee

\subsubsection{DAPO}

  In the \keywordDef{DAPO} paper of \citep{Yu2025DAPO},
  they suggest an asymetric clipping of the likelihood
ratio term:
\be
\text{clip}(\rho_{nt}(\vtheta),
1-\epsilon_{\text{low}},
1+\epsilon_{\text{high}})
\hat{A}_{nt}^j
\ee
In particular, they suggest using
$\epsilon_{\text{high}}=0.28 > \epsilon_{\text{low}}=0.2$ in the clipping term,
so that actions which are low-probability under the
previous model, and which therefore get large
likelihood ratio $\rho_{it}$,
are not clipped as much, which would suppress exploration
and result in ``entropy collapse''
(see also \citep[Sec 4.2]{Liu2025LitePPO}).
In \citep{Xi2025bapo}, they propose a way to automatically
adapt the clipping ratios.

\subsubsection{GSPO}
\label{sec:trajectoryLevel}
\label{sec:GSPO}

In the \keywordDef{GSPO}  (Group Sequence Policy Optimization)
paper 
\citep{GSPO}, they point out a flaw with GRPO,
due to the  fact that the importance sampling correction
$\rho_{nt}^j$ is applied to each token,
even though the reward is evaluated at the trajectory level.
This can result in unstable training.
They therefore propose to use the following
sequence-level objective,
following the contextual bandit formulation:
\be
J_{GSPO}(\vtheta)
= \frac{1}{N} \sum_{n=1}^N
\frac{1}{J} \sum_{j=1}^J
\min\left(
\rho_{n}^j(\vtheta) \hat{A}_{n}^j,
\text{clip}(\rho_{n}^j(\vtheta)) \hat{A}_{n}^j
  \right)
  \ee
where the importance ratio is given by
the ratio of sequence level likelihoods
  \citep{Zheng2023click},
  where they also normalize by length
  to ensure the magnitude (and hence clipping value)
  is comparable across sequences:
\be
\rho_{n}^j(\vtheta)=
\left(
\frac{\pi_{\vtheta}(\va_{n}^j|s_n)}
     {\pi_{\old}(\va_{n}^j|s_n)}
     \right)^{1/|\va_n^j|}
     =\exp\left(
     \frac{1}{|\va_n^j|}
     \sum_{t=1}^{|\va_n^j|}
     \log 
\frac{\pi_{\vtheta}(a_{nt}^j|\va_{n,<t}^j,s_n)}
     {\pi_{\old}(a_{nt}^j|\va_{n,<t}^j,s_n)}
     \right)
     \label{eqn:GSPOrho}
\ee

\eat{
  In \citep{DrGRPO,r1zeroNUS},
  they note that
normalizing by $1/|\va_n^j|$ gives equal weight
to all responses, regardless of length,
Thus the loss contribution per token
may be reduced for longer answers,
which might discourage long answers.
Therefore \citep{Yu2025DAPO} suggests to use token-level normalization:
\be
J_{token-level}(\vtheta)
= \frac{1}{N} \sum_{n=1}^N
\frac{1}{\sum_{j=1}^J |\va_n^j|}
\sum_{j=1}^J
\sum_{t=1}^{|\va_n^j|}
\min\left(
\rho_{nt}^j(\vtheta) A_{nt}^j,
\text{clip}(\rho_{nt}^j(\vtheta),
1-\epsilon_{\text{low}},
1+\epsilon_{\text{high}})
A_{nt}^j
\right)
\label{eqn:tokenLevelLoss}
  \ee

In \citep{Liu2025LitePPO} they show empirically (on various
math benchmarks) that
token-level loss seems to work better when applying
RLFT to base models,
and sequence-level loss seems to work better
when applying RLFT to instruction-tuned (``aligned'') models.

In \citep{Nguyen2025SFR},
where they study RL on instruction-tuned models
for multi-turn tool-using agents,
they propose the arguably
more natural approach of normalizing the
advantage
by the length before clipping,
i.e., they use
\be
\tilde{A}_{nt}^j = \frac{1}{|\va_n^j|}
\frac{R_n^j-\mu_n}{\sigma_n}
\ee
This makes sense, since we are distributing a (rescaled)
trajectory level reward $R_n^j$ evenly across
all time steps,
which ensures that shorter rollouts are preferred.
They combines this with the objective
\be
J_{SFR}(\vtheta)
= \frac{1}{NJ} \sum_{n=1}^N
\sum_{j=1}^J
\sum_{t=1}^{|\va_n^j|}
\min\left(
\rho_{nt}^j(\vtheta) \tilde{A}_{nt}^j,
\text{clip}(\rho_{nt}^j(\vtheta))
\tilde{A}_{nt}^j
\right)
  \ee
  }

\subsubsection{RLOO}
\label{sec:RLOO}

It turns out that
the division
by the standard deviation used by GRPO
when normalizing the advantage terms
induces a difficulty bias \citep{r1zeroNUS},
in which a very easy or hard prompt $s_n$ may have
a low group-level standard deviation $\sigma_n$
of the corresponding rewards,
and dividing by a small $\sigma_n$ can result
in unstable gradients.
The problem can be solved using
the \keywordDef{Dr GRPO} (GRPO Done Right) method of
\citep{DrGRPO},
where they just drop the denominator, giving
\be
\hat{A}_{nj}^{\text{DrGRPO}} = R_{nj} - \mu_n,
\;
\mu_n = \frac{1}{J} \sum_{j=1}^J R_{nj}
\ee

Another approach is to use
the Reinforce Leave-One-Out
method (\keywordDef{RLOO}) of
\citep{Ahmadian2024},
they propose the following  baseline:
\be
\hat{A}_{nj}^{\text{RLOO}} = R_{nj} - \mu_n,
\;
\mu_n = \frac{1}{J-1} \sum_{j=1, j \neq i}^J R_{nj}
\label{eqn:RLOObaseline}
\ee
which is the average reward for all the samples in the batch,
excluding the current sample.

The RLOO expression 
is identical (up to  a scaling factor
of $J/(J-1)$) to the DrGPO expression.
To see this, note that
\begin{align}
\frac{J}{J-1} \hat{A}_{nj}^{\text{DrGRPO}}
  &= \frac{J}{J-1} (R_{nj} - \frac{1}{J} \sum_{j=1}^J R_{nj}) 
= \frac{J}{J-1} R_{nj}
- \frac{1}{J-1} (\sum_{j=1, j \neq n}^J R_{nj} + R_{nj}) \\
  &= \frac{J}{J-1} R_{nj} - \mu_n - \frac{1}{J-1} R_{nj}
  = R_{nj} - \mu_j = \hat{A}_{nj}^{\text{RLOO}}
 \end{align}

\subsubsection{REINFORCE++}
\label{sec:REINFORCELLM}
\label{sec:REINFORCEPP}

Unfortunately, RLOO estimates a different
baseline $\mu_n$
for each prompt $s_n$,
which can lead to overfitting on simpler prompts
and vulnerability to reward hacking.
The \keywordDef{REINFORCE++} algorithm of
\citep{ReinforcePP} computes the mean and 
 standard deviation globally across
the batch; this is unbiased and
improves the training stability.
That is, it  uses the following estimate for the advantage:
\be
\hat{A}_n^j = \frac{R_n^j - \mu}{\sigma},
\; \mu = \text{mean}(R_n^j: n=1:N, j=1:J),
\; \sigma = \text{std}(R_n^j: n=1:N, j=1:J)
  \ee

Another variant is to first  
subtract off a local group mean
and then do global normalization to get
\begin{align}
\tilde{A}_n^j &= R_n^j - \mu_n,
\mu_n = \text{mean}(R_n^j: j=1:J) \\
  \hat{A}_n^j &= \frac{\tilde{A}_n^j - \mu}{\sigma},
\mu = \text{mean}(\tilde{A}_n^j: n=1:N, j=1:J),
\sigma = \text{std}(\tilde{A}_n^j: n=1:N, j=1:J)
\end{align}

\subsubsection{VinePPO}
\label{sec:VinePPO}

\begin{figure}
\centering
\includegraphics[height=2in]{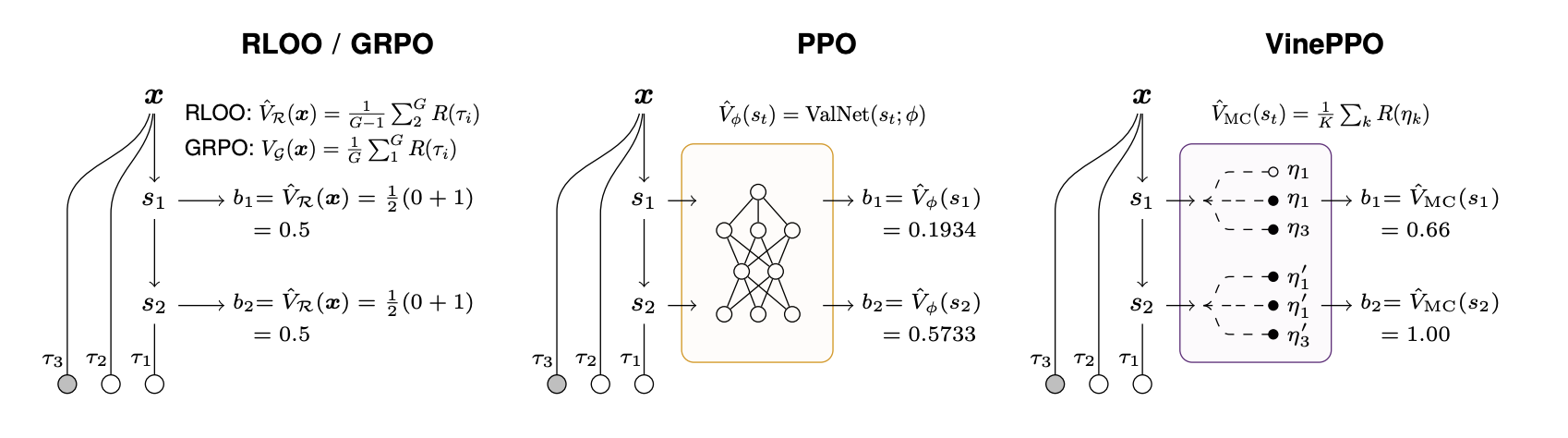}
\caption{
  Three different ways to estimate  the baseline
  function $b_t$
(needed to compute the advantage function $A_t$)
  for intermediate tokens $s_t$ based on $G=3$
  trajectory rollouts.
  (a) MC estimate $b_t=V_{MC}(s_0)$
derived from the initial state.
(b) Parametric value function, $b_t=V_{\phi}(s_t)$.
(c) MC estimate $b_t=V_{MC}(s_t)$ derived from $K$ rollouts
$\eta_k \sim \pi(\cdot|s_t)$ from state $s_t$.
(This exploits the fact that language environments are deterministic
and resettable.)
Figure from \citep{Kazemnejad2025}.
  }
\label{fig:VinePPO}
\end{figure}

GRPO and related methods computes  an unbiased estimate
of the value of each token based on rolling out
multiple trajectories from the same starting state (prompt).
All intermediate states (action tokens) are treated equally,
as illustrated in \cref{fig:VinePPO}.
For long chains of thought, this can be a poor
estimate of the true RTG (value) of an intermediate token.
In \citep{Kazemnejad2025}, they propose \keywordDef{VinePPO},
which exploits the fact that
in language-based environments,
it is possible to 
reset directly to any intermediate state simply by refeeding the partial context;
this enables multiple  MC rollouts from   any state in the trajectory,
thus providing  a more accurate value estimate
without fitting a value function network (which can be biased).
Unfortunately this technique is slow, and not general purpose.

\subsubsection{Adding a KL regularizer}

It is common to add a KL penalty to the per-step reward,
to prevent the policy from deviating too far
from the base (SFT) LLM:
\be
\hat{R}_{n,t}^j = R_{n}^j  - \beta 
\KLpq{\pi_{\old}(a_{nt}^j|s_n,\va_{n,<t}^j)}{\pi_{\text{ref}}(a_{nt}^j|s_n,\va_{n,<t}^j)}
\ee

In GRPO (and many other papers), they use a low-variance
MC estimator of KL divergence
proposed in \url{http://joschu.net/blog/kl-approx.html}.
The naive estimator of
$KL(q,p) = E_{q(a)}[\log(q(a)/p(a))]$
has the form 
\be
k_1 = \log(r) = \log q(a) - \log p(a)
\label{eqn:k1}
\ee
where $a \sim q$ and $r(a)=q(a)/p(a)$.
Another estimator, 
which is biased but has lower variance,
is the following:
\be
k_2 = \half (\log(r))^2
\label{eqn:k2}
\ee
Finally, an unbiased estimator with low
variance is
\be
k_3 = (r-1)-\log(r)
\ee
Although $k_3$ is unbiased,
\citep{Tang2025pitfalls}
and
\citep[App B.1]{ReinforcePP}
show that its gradient is biased,
so they recommend $k_2$.

\eat{
In REINFORCE++, they add the KL penalty to the advantage to get
\be
A_{n,t}^j = R_n^j  - \beta \sum_{t=1}^T
\KLpq{\pi_{\old}(a_{nt}^j|s_n,\va_{n,<t}^j)}{\pi_{\text{ref}}(a_{nt}^j|s_n,\va_{n,<t}^j)}
\ee
Finally they use the normalized advantage
$\hat{A}_{nt}^j = (A_{nt}^j - \mu)/\sigma$.
}

\eat{
In practice,
GRPO approximates the $J_{kl}$ term
by sampling from $\va_i^j \sim p_i$,
so we can share the answers  for both
the PPO term and the KL term.\footnote{
Note that this is biased, as pointed out
in \url{https://x.com/NandoDF/status/1884038052877680871}.
However, the bias is likely small if $\vtheta_{\told}$ is close
to $\vtheta$.
If this bias is a problem, it is easy to fix using importance sampling.
}
}

\subsubsection{DPO}
\label{sec:DPO}

Rather than first fitting a reward model from preference data
using the Bradley-Terry model in \cref{sec:RLHF},
and then optimizing the policy to maximize this,
it is possible  to optimize the preferences directly,
using the
 \keywordDef{DPO} (Direct Preference Optimization) method
 of \citep{Rafailov2023}.
 (This is sometimes called
 \keywordDef{direct alignment}.)

We now derive the DPO method. To simplify notation,
we use $x$ for the input prompt (initial state $s$)
and $y$ for the output (answer sequence $\va$).
The objective for KL-regularized
policy learning is to maximize the following:\footnote{
In \citep{Huang2025KL}, they recently showed that it is
better
to replace the KL divergence with a $\chi^2$ divergence,
which quantifies uncertainty more effectively than KL-regularization.
The resulting algorothm is provably robust to overoptimization,
unlike standard DPO.
} %
\begin{align}
J(\pi) &=  \E_{x \sim \data,  y \sim \pi(y|x)}
\left[ R(x,y) 
- \beta \log \frac{\pi(y|x)}{\pi_{\tref}(y|x)}
\right]
\label{eqn:DPO}
\end{align}
Equivalently  we can minimize the loss
\begin{align}
\loss(\pi) &=  \E_{x \sim \data,  y \sim \pi(y|x)}
\left[\log \frac{\pi(y|x)}{\pi_{\tref}(y|x)}
  - \frac{1}{\beta} R(x,y) 
  \right] \\
&= 
\E_{x \sim \data,  y \sim \pi(y|x)}
\left[
  \log \frac{\pi(y|x)}
       {\frac{1}{Z(x)} \pi_{\tref}(y|x)
         \exp(\frac{1}{\beta} R(x,y))}
       - \log Z(x)
       \right]
\end{align}
where
\be
Z(x) = \sum_y \pi_{\tref}(y|x) \exp(\frac{1}{\beta} R(x,y))
\ee
is the partition function.
We can now rewrite the loss as
\begin{align}
\loss(\pi) &= 
\E_{x \sim \data}
\left[
  \KLpq{\pi(y|x)}
  {\frac{1}{Z(x)} \pi_{\tref}(y|x)
    \exp(\frac{1}{\beta} R(x,y))}
       - \log Z(x)
       \right]
\end{align}
The second term is independenent of $\pi$ so can
be dropped. The first term can be minimized by setting
\be
\pi^*(y|x) =
\frac{1}{Z(x)} \pi_{\tref}(y|x)
    \exp(\frac{1}{\beta} R(x,y))
\ee
from which we can derive the optimal reward function as
\be
R^*(x,y) = \beta \log \frac{\pi^*(y|x)}{\pi_{\tref}(y|x)}
+ \beta \log Z(x)
\ee
Plugging this into the Bradley Terry model
of \cref{eqn:bradleyTerry}
we get
\begin{align}
  p^*(y_w \succ y_l|x)
  &=\frac{
    \exp\left( \beta \log \frac{\pi^*(y_w|x)}{\pi_{\tref}(y_w|x)}
    + \beta \log Z(x) \right) }
  {
    \exp\left(
    \beta \log \frac{\pi^*(y_w|x)}{\pi_{\tref}(y_w|x)}
        + \beta \log Z(x) \right)
        +     \exp\left(
        \beta \log \frac{\pi^*(y_l|x)}{\pi_{\tref}(y_l|x)}
    + \beta \log Z(x) \right)} \\
  &= \frac{1}{1 +
\exp\left(    \beta \log \frac{\pi^*(y_l|x)}{\pi_{\tref}(y_l|x)}
- \beta \log \frac{\pi^*(y_w|x)}{\pi_{\tref}(y_w|x)}
\right)} 
\end{align}
Thus we can fit the policy by minimzing
\be
\loss(\vtheta)
= -\E_{(x,y_w,y_l) \sim \data}
\left[ \log
   \sigma\left(
\beta \log \frac{\pi_{\vtheta}(y_w|x)}{\pi_{\tref}(y_w|x)}
- \beta \log \frac{\pi_{\vtheta}(y_l|x)}{\pi_{\tref}(y_l|x)}
\right)
\right]
\ee

The main downside of DPO is that  it is limited to
learning from  preference data,
whereas the other policy gradient methods can work with any
reward function, including verifiable (non-learned) rewards.


\eat{
\subsubsection{Methods based on Q learning}

The above methods are based on policy gradient methods,
which are on-policy learners.
In \citep{Clavier2025} they discuss how to extend Q-learning,
which can work off-policy, to LLMs.
}

\subsubsection{Inference-time scaling  using posterior sampling}
\label{sec:LLMtestTime}
\label{sec:tilted}


We can view the problem of generating samples from an LLM
that maximizes some reward as equivalent to posterior sampling
from a \keywordDef{tilted distribution}
\citep{Li2023tilted},
that combines
the prior $\pi_{\tref}(y|x)$ with a likelihood
$p(O=1|x,y)$, where
$O$ is known as an optimality variable,
to get the posterior
$p(y|x,O=1) \propto \pi_{\tref}(y|x) p(O=1|x,y)$.
We define $p(O=1|x,y) = \exp(R(x,y) / \beta)$,
as in the ``control as inference'' paradigm
(see \cref{sec:inferRL}),
where $\beta > 0$ is a temperature parameter.
(For example, $R(x,y)=\log p(O=1|x,y)$ could be the log-probability
that the response $y$ to prompt $x$ is non-offensive,
or that $y$   is the correct answer to the question $x$.)
Thus we want to sample from
\be
p(y|x,O=1)
= \frac{1}{Z_x} p(O=1|x,y) \pi_{\tref}(y|x)
= \frac{1}{Z_x} \exp(\beta^{-1} R(x,y))  \pi_{\tref}(y|x)
\ee
where $Z_x=p_{\tref}(O=1|x)=\sum_y p(O=1|x,y) \pi_{\tref}(y|x)$
is the normalization constant.
(Henceforth we omit the conditioning prompt $x$ for notational simplicity.)

An alternative (but equivalent) formulation
is to write the target posterior as 
\be
\pi^*(y|x) \propto e^{\frac{1}{\beta} R(x,y)} \pi_{\tref}(y|x)
\label{eqn:piStar}
\ee
This is the optimal solution to the following
KL-regularized RL problem
(see e.g., \citep{Beirami2025}):
\be
\max_{\pi(\cdot|x)}
\E_{\pi(y|x)} R(x,y)
-\beta \KLpq{\pi(\cdot|x)}{\pi_{\tref}(\cdot|x)}
\ee

Note that, if we set $\beta=1$ and define
$R(x,y)=\log(\pi_{\tref}(y|x)^{\alpha-1})$,
the optimal solution becomes the
\keywordDef{tempered distribution}
or \keywordDef{power distribution}
\be
\pi^*(y|x) \propto  \pi_{\tref}(y|x)^{\alpha-1}  \pi_{\tref}(y|x)^1
=  \pi_{\tref}(y|x)^{\alpha}
\ee
which can be flatter if $\alpha<1$ or sharper
if $\alpha>1$ than the reference distribution.\footnote{
In \citep{Karan2025}
they show that
the  sharpened distribution
upweights tokens with few but high likelihood future paths,
while low-temperature sampling (on a per-token basis)
upweights tokens with
several but low likelihood completions.
See also \citep{Shih2023long,Zhao2024}.
}

\eat{
An alternative (but equivalent) formulation
is to write the target posterior as a
\keywordDef{product of experts}:
\be
\pi^*(y|x)  = \frac{1}{Z_x} \Phi(x,y) \pi_{\tref}(y|x)
\ee
where $\Phi(x,y) = \prod_i \phi_i(x,y)$ is a product
of (non-negative)
\keywordDef{potential functions}, each representing local or global
preferences or constraints.
For example, we might define $\Phi(x,y)=1$ iff  $y$ is the correct answer to question $x$.
}

Sampling from such distributions can be done using various
methods.
A simple method is known as \keywordDef{best-of-N} sampling,
which just generates $N$ trajectories, and picks the best.
This is equivalent to ancestral sampling from the forwards model,
and then weighting by the final likelihood
(a soft version of rejection sampling).
Depsite its simplicity,
it was
shown in \citep{Gui2024,Beirami2025}
that this method achieves an optimal reward vs KL tradeoff.
However,
performance can decrease when $N$ increases,
due to deviating too far from the base model
(see e.g., \citep{Huang2025BON,Faria2025}).

A potentially more computationally efficient
method is to use  \keywordDef{twisted SMC},
which combines particle filtering (a form of sequential Monte Carlo)
with a  ``twist'' function, which predicts
the future reward  given the current state\footnote{
Formally, the optimal twist function on a partial sequence $y$
is defined as $\Phi^*(y) = \expectQ{\Phi(y') | \text{$y$ is a prefix of $y'$}}
{\pi_{\tref}(y')}$.
},
analogous to a value function
(see e.g. \citep{Naesseth2019,Chopin2020,Lawson2018,Lawson2022}).
This is sometimes called \keywordDef{SMC steering},
and  has been used 
in several papers (see e.g., \citep{Lew2023,Zhao2024,Feng2025,Loula2025,Puri2025}).
(See also \cref{sec:SMCMPC} for closely related work
which uses twisted SMC to sample action sequences,
where the twist is approximated by 
a learned advantage function.)

The posterior sampling approach  discussed above
is an example of using more compute at test time
to improve the generation process of an LLM.
This is known as \keywordDef{test time compute}
(see e.g., \citep{Ji2025test} for survey).
This provides another kind of scaling law,
known as \keywordDef{inference time scaling},
besides just improving the size (and training time) of the base model
\citep{Snell2024}.

\eat{
.\footnote{
%
A scaling law is a relationship of the form
$y=ax^b$, or equivalently,
$\log(y)=b \log(x) + c$.
It gets its name because it is scale invariant,
in the sense that, if you double (say) $x$,
then $y$ increases $2^b$, no matter what is the scale (value) of $x$.
This is because
$a (cx)^b = c^b a x^b = c^b y$, so we still have $y=ax^b$.
(Note that if we have an offset,
$y = ax^b + d$, then if we define
$\overline{y}=y-d$, we recover a standard scaling law,
$\overline{y} = ax^b$.)
In a ``neural scaling law'' for pre-training
\citep{Kaplan2020},
$y$ is the loss,
and $x$ is the compute $C$ (in FLOPs) spent during training,
or the number of parameters $N$,
or the dataset size $D$.
  }
  }


\subsubsection{RLFT as amortized posterior sampling}

The disadvantage of 
decision-time planning (online posterior sampling)
is that it can be slow.
Hence it can be desirable to 
``amortize'' the cost
by fine-tuning the base LLM so it matches the tilted
posterior \citep{Korbak2022}.
We can do this by minimizing
$J(\vtheta)  = \KLpq{\pi_{\vtheta}}{\pi^*}$,
where $\pi^*$ is the tilted distribution in \cref{eqn:piStar},
and then sampling from $\pi^*$.

Following  \cref{eqn:ppi}, we have
\begin{align}
  \KLpq{\pi_{\vtheta}(y)}{\pi^*(y)}
  &= \expectQ{\log \pi_{\vtheta}(y) -
    \log \Phi(y) - \log \pi_{\tref}(y)}{\pi_{\vtheta}(y)}
  + \log p_{\tref}(O=1)
\end{align}
where the first term is the negative ELBO (evidence lower bound),
and the second $\log p_{\tref}(O=1)$ term is independent of
the function being optimized (namely $\pi^*$).
Hence we can minimize the KL by maximizing the ELBO:
\begin{align}
  J'(\vtheta)
  &= \expectQ{  \log \Phi(y)}{\pi_{\vtheta}(y)}
  - \KLpq{\pi_{\vtheta}(y)}{\pi_{\tref}(y)}
\propto \expectQ{ R(y)}{\pi_{\vtheta}(y)}
- \beta \KLpq{\pi_{\vtheta}(y)}{\pi_{\tref}(y)} 
\end{align}
which is exactly the KL-regularized RL objective
used in DPO \cref{eqn:DPO}.

The advantage of this probabilistic (distribution matching)
perspective
over the RL (reward maximizing)
perspective is that it suggests natural alternatives,
such as optimizing the inclusive or forwards KL,
$\KLpq{\pi^*}{\pi_{\vtheta}}$,
which is ``mode covering'' rather than ``mode seeking''.
This can prevent ``catastrophic forgetting'',
in which the tuned policy loses diversity
as well as some of its original
capabilities \citep{Korbak2022nips}.
(This  is similar to the advantage
of \keywordDef{reweighted wake sleep} training
(see e.g., \citep{Le2020RWS,McNamara2024,Zhou2025VarLLM})
compared to (amortized) VI training for latent variable models.)

Note, however, that these offline approaches to LLM finetuning
(whether using forwards or reverse KL penalty)
have the disadvantage, compared to  the online (decision-time) approach to inference,
that they cannot easily handle hard constraints,
since they only train  policies that
respect the constraints on average.
(This is why MPC,
which is a similar form of decision-time inference
(see \cref{sec:MPC}), is so widely used in the robotics
community, where hard constraints are prevalent.)

\subsection{Algorithms for multi-turn RL}
\label{sec:multiTurn}
\label{sec:multi-turn}

So far, we have considered the bandit setting,
in which a single prompt, $s$, is presented,
and then the agent optionally generates some thinking tokens,
followed by the answer tokens,
and then immediately gets a reward.
In this section, we discuss how
to train agents that can interact with an external
environment, which enables
\keywordDef{tool use} (see e.g., \citep{Nguyen2025SFR,Mai2025scaling}),
or  \keywordDef{dialog agents}
 (see e.g., \citep{Laban2025,Abdulhai2025,Zhao2025MUA}).
The difference from  the standard contextual bandit
LLM reasoning setting
is that the effect of an action
on the external environment is typically unknown,
and may be stochastic,
and the reward may be delayed.
In addition, the external environment is often stateful,
so actions may be irreversible.
Training agents in this setting  requires 
true \keywordDef{multi-turn RL} methods,
as we have discussed throughout the rest of this document.
There are currently only a few papers on using LLMs
for this multi-turn setting
(see e.g., \citep{Carta2023,Wang2025RL,Li2025multiturn,ToolR1,verlog,
  Liu2025gem,Jin2025searchR1,Song2025R1}),
but this area will likely grow in the future.


 \eat{
\begin{figure}
\centering
\includegraphics[height=2in]{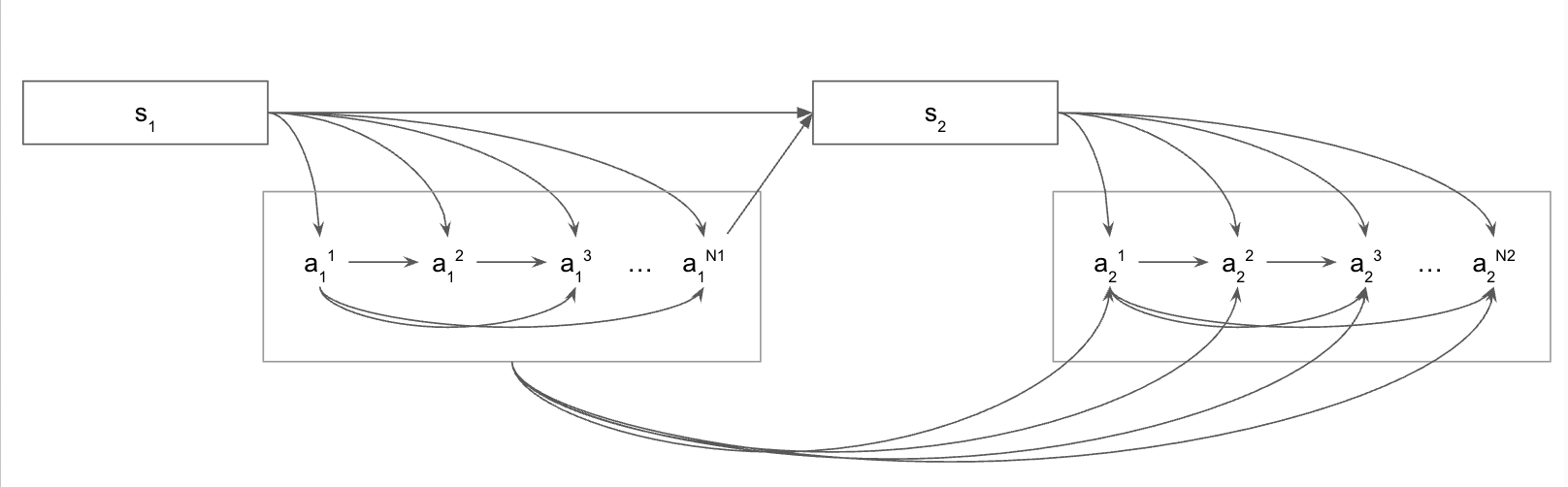}
\caption{
  Multi-turn dialog between
environment (with states $s_t$)
and  agent (generating actions $a_t^{1:N}$).
  }
\label{fig:multi-turn}
\end{figure}
}

\subsubsection{Example: RAGEN}
\label{sec:RAGEN}

\begin{figure}
\centering
\includegraphics[height=2in]{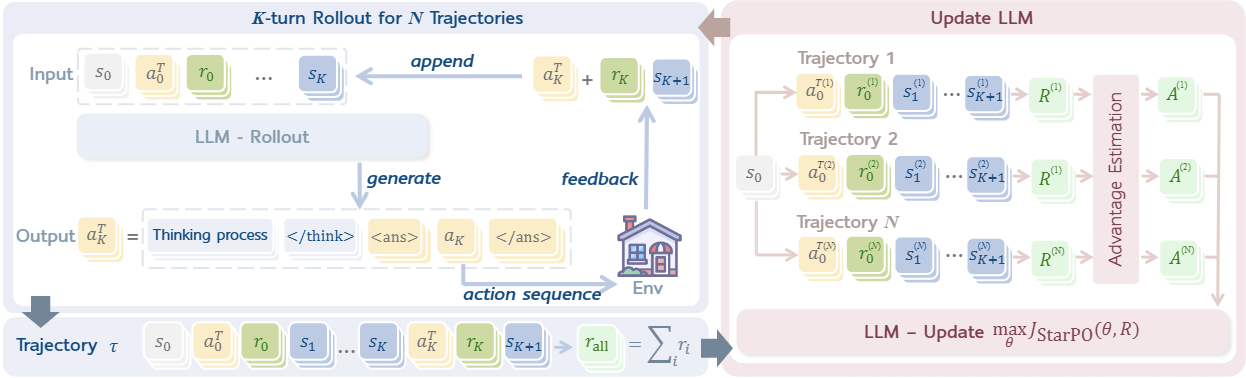}
\caption{
  Illustration of how to train an LLM to both ``think''
  internally  and ``act'' externally.
  From the initial state $s_0$,
  we roll out $N$ trajectories, each of length $K$,
  and then use the empirical average across all $N$ as a baseline
  for computing the advantage function.
  \figtaken{\citep{RAGEN}}.
  \figthanks{Zihan Wang}.
  }
\label{fig:ragen}
\end{figure}

As an example of a multi-turn RL system,
\cref{fig:ragen} illustrates the
 \keywordDef{RAGEN} system
 of \citep{RAGEN}.
Here the policy is promoted to generate a set of thinking tokens,
followed by a set of action tokens,
until the STOP token is generated (representing the end of action).
The policy is used to generate $N$ trajectories,
from which the advantage can be estimated using a Monte Carlo
estimate.
They call this algorithm  \keywordDef{StarPO}
 (State-Thinking-Action-Reward Policy Optimization).

\subsubsection{Dealing with invalid actions} 

When interacting with the external environment,
we often find that some trajectories result in a
an \keywordDef{invalid action}
during the generation of an action sequence.
There are several ways to deal with such
\keywordDef{partial rollouts}.
For example, we could just truncate
the sequence to length $t$ (the step where the error occured)
to give
$\tau_{n}=(s_{n1},a_{n1},\ldots,s_{nt},a_{nt})$.
Or we could start a new rollout from
the state where the failure occured,
$s_{nt}$ (c.f. \citep{Nguyen2025SFR}).
Or we can replace the action
that caused the error
with some new action
(e.g., chosen uniformly from the set of legal actions),
and then continue the rollout
(c.f., \citep{verlog}).

\subsubsection{Turn-level training}

\begin{figure}
\centering
\includegraphics[height=3in]{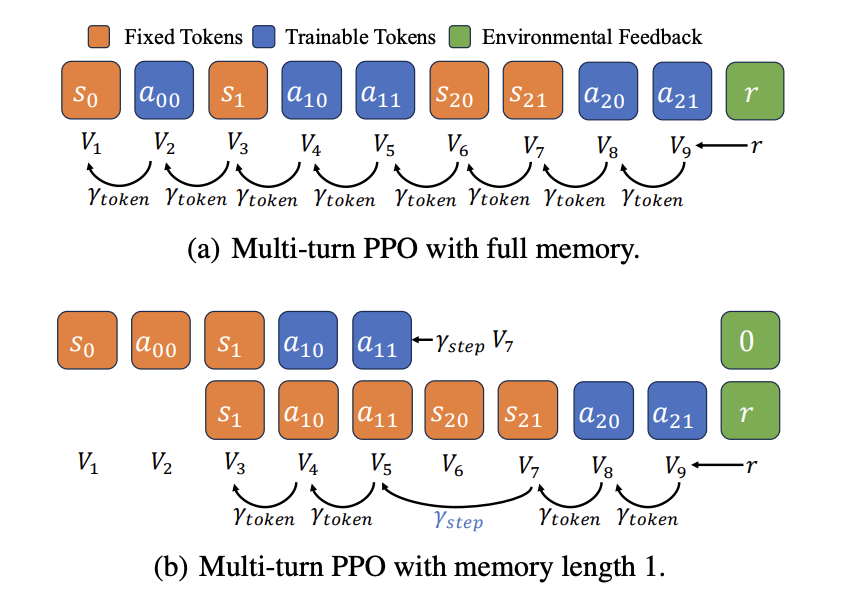}
\caption{
  (a) In standard multi-turn RL, the context grows longer
  at each step.
  (b) In turn-level training, we truncate the past context
  to a fixed window (here, the previous state, previous action, and current state).
  In addition, we learn a value function, so we can truncate
  trajectories early and still get a learning signal.
   State tokens are skipped over when computing the loss,
  since they are not generated by the model.
  Finally, we use different discounting factors for
  steps corresponding to action token generation
  and environment tokens.
  \figtaken{\citep{verlog}}.
  \figthanks{Wen-Tse Chen}.
  }
\label{fig:verlog}
\end{figure}

Methods like RAGEN (\cref{sec:RAGEN})
do not work well in long horizon tasks, for at least
two reasons:
(1) The context length (which is input to the policy)
can grow very large;
and
(2) the REINFORCE estimator can be very high variance
once the number of steps needed to reach a terminal reward
becomes large.
We can tackle (1) by truncating (or summarizing, using an LLM)
the history of previous states and actions, as illustrated
in \cref{fig:verlog}.
We can tackle (2) by learning a value function or critic,
and then using the Generalized Advantage Estimator of \cref{sec:GAE}
to compute $A_t^n$,
instead of using the Monte Carlo estimate of the return to go,
$G_t^n = \sum_{k=t}^{T} \gamma^{k-t} r_k^n$,
as in REINFORCE.
The loss (for the on-policy case) becomes
\be
\loss_{A2C}(\theta)
= \frac{1}{N} \sum_{n=1}^N \sum_{t=1}^{T_n}
A_t^n \log \pi_{\theta}(a_t^n|s_t^n)
\ee

To improve performance, we can warm start the actor (policy)
from  a prompted,  instruction-tuned LLM,
and warm start the critic 
by fitting it to the MC value estimates computed
from rollouts using the initial  frozen policy,
before updating both actor and critic,
as suggested in
\citep{verlog}.

\label{sec:REBN}
In \citep{Liu2025gem}, they propose to replace the
GAE estimate (which requires learning a critic) with
\keyword{return batch normalization} (\keywordDef{ReBN}),
which is similar to GRPO but defined for the multi-turn
setting:
\be
A_t^n = \frac{G_t^n-\mu}{\sigma},
\;
\mu = \text{mean}(G_{1:T}^{1:N}),
\sigma = \text{std}(G_{1:T}^{1:N})
\ee

\eat{
\subsubsection{Dealing with multi-token actions}

\begin{figure}
\centering
\includegraphics[height=2in]{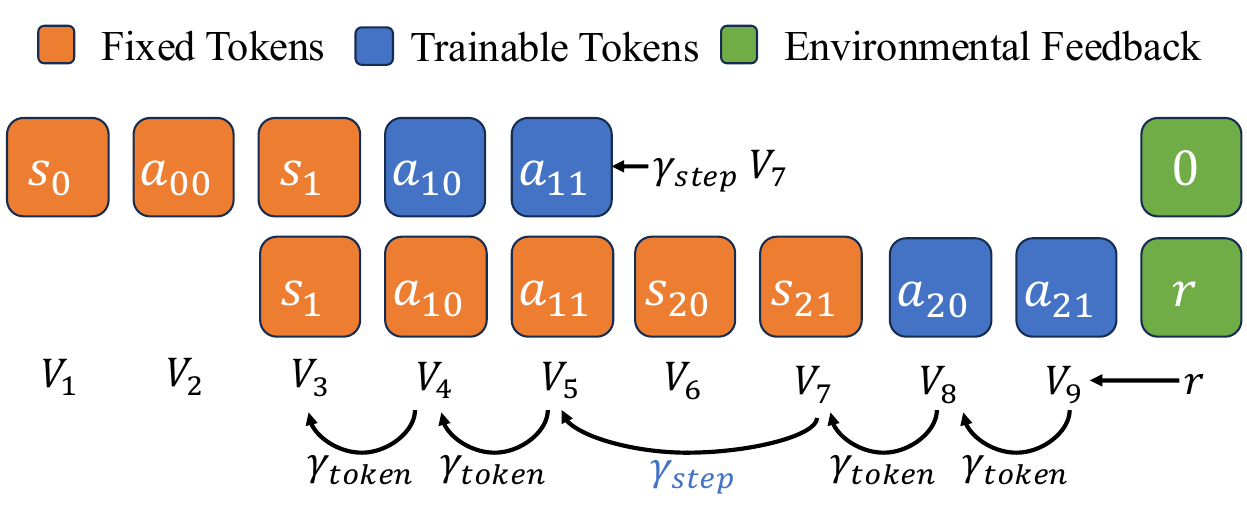}
\caption{
  Illustration of different GAE discount factors depending
  on whether a token comes from the agent or the environment.
  We assume the memory length of the policy is 1,
  so the input is just the previous state, previous action, and current state,
  where both states and actions may require multiple tokens to represent them.
  In the first row,  corresponding to $t=1$,
  the input to the policy
  is the previous state $s_0$, previous action  $a_{00}$, and current state $s_1$,
  and the generated output is the action  $a_{10},a_{11}$.
  In the second row,  the input to the policy
  is the previous state $s_1$, previous action sequence $a_{10},a_{11}$,
  and current state $s_{20},s_{21}$,
  and the generated output is the action sequence $a_{20},a_{21}$.
  State tokens are skipped over when computing the loss,
  since they are not generated by the model.
  \figtaken{\citep{Chen2025verlog}}.
  \figthanks{Wen-Tse Chen}.
  }
\label{fig:verlog}
\end{figure}

In the \keywordDef{Verlog} system \citep{verlog,Chen2025verlog},
they propose to use a \keywordDef{dual discounting GAE}
method when computing $A_t = \gamma \lambda A_{t+1} + \delta_t$
in \cref{eqn:GAE}.
In their approach, they use a different value of
$\gamma$ and $\lambda$ depending on whether the 
token at time $t$ was generated from the policy or corresponds
to part of the environment state,
as illustrated in \cref{fig:verlog}.
In particular, they propose
to use $\gamma_{\text{token}}=\lambda_{\text{token}}=1.0$,
to ensure future value corresponding to tokens
that belong to the same action chunk are not discounted,
but they set
$\gamma_{\text{step}}=0.99$ and $\lambda_{\text{step}}=0.95$,
to ensure that environment steps are penalized,
to encourage solving the task more quickly.
}

 \eat{
\subsubsection{Clipped REINFORCE (aka MC-PPO)}
\label{sec:PPOlite}
\label{sec:litePPO}
\label{sec:MC-PPO}
\label{sec:LITE}



We can of course use PPO to optimize the multi-turn RL loss,
but when using LLMs, it is more common to
replace the GAE estimator
with an MC estimator, to save memory.
For example, we can use the
REINFORCE++ estimator
for the advantage function,
as discussed in 
\cref{sec:REINFORCEPP}.
In particular, for step $t$ of episode $n$,
we define the normalized advantage to be
\be
\hat{A}_{nt} 
= \frac{G_{nt} - \mu}{\sigma}
\ee
where $G_{nt}$ is an MC estimate of the discounted reward-to-go,
optionally with a KL penalty:
\be
G_{nt} = \sum_{k=t}^T \gamma^k r_{nk}
- \beta \sum_{k=t}^T KL(\pi_{\vtheta}(\cdot|s_{nk}),
\pi_{\old}(\cdot|s_{nk}))
\ee
Here  $\mu$ and $\sigma$ are the global mean and std across
all RTGs  in the batch:
\be
\mu = \text{mean}(G_{nt}: n=1:N, t=1:T),
\sigma = \text{std}(G_{nt}: n=1:N, t=1:T)
\ee
Also, recall that the policy $\pi_{\vtheta}(\cdot|s_{nt})$
is a distribution over a {\em sequence} of action tokens,
$\va_{nt}$, some of which may correspond to internal
thinking tokens, and some of which may be sent to the environment.

We combine the above with the clipped update
from PPO to get the following objective:
\be
J(\vtheta)  = \frac{1}{N} 
\sum_{n=1}^N \sum_{t=1}^{T_n}
\min\left(
\rho_{nt}(\vtheta) \hat{A}_{nt}
\text{clip}(\rho_{nt}(\vtheta) \hat{A}_{nt})
\right),
\;
\rho_{nt}(\vtheta) =
\frac{\pi_{\vtheta}(\va_{nt} | s_{nt})}
{\pi_{\old}(\va_{nt}|s_{nt})}
\ee
where the sampled actions
are generated using $\pi_{\old}$.
(Note that, unlike single turn RL,
we do not assume that we can reset the environment state,
so we cannot generate multiple rollouts $\va_{nt}^j$
from the same state $s_{nt}$.)
We  call the resulting method
\keywordDef{clipped REINFORCE++}
or \keywordDef{MC-PPO}.

 }
 
\eat{
In the GSPO paper \citep{GSPO},
they propose a modified loss,
where they rollout multiple responses to the same
initial state $s_n$, similar to single-turn RL:
\be
J_{GSPO-token}(\vtheta)
= \frac{1}{N} \sum_{n=1}^N
\frac{1}{J} \sum_{j=1}^J
\frac{1}{T_n^j}
\sum_{t=1}^{T_n^j}
\min\left(
\rho_{nt}^j(\vtheta) A_{nt}^j,
\text{clip}(\rho_{nt}^j(\vtheta)) A_{nt}^j
  \right)
  \ee
They also propose to modify the 
importance ratio to the following form,
where we incorporate a stop-gradient operator:
\be
\rho_{nt}^j(\vtheta)=
\text{stopgrad}(\rho_n^j(\vtheta))
\frac{\pi_{\vtheta}(a_{nt}^j|\va_{n,<t}^j,s_n)}
     {\text{stopgrad}(\pi_{\vtheta}(a_{nt}^j|\va_{n,<t}^j,s_n))}
\ee
where $\rho_n^j(\vtheta)$ is defined in
\cref{eqn:GSPOrho}.
}

\eat{
\subsubsection{Off-policy methods}

Multi-turn RL needs to interact with the external environment,
which can be slow. Therefore it is useful to use all sources
of data, including past data, human trajectories, etc.
This requires that we use off-policy methods,
which we discussed in \cref{sec:offPolicy}.
See e.g., \citep{Wang2025offline}.
This can be combined with on-policy ``refresh'',
where the policy is updated with new data (e.g., using PPO),
which is added back to the replay buffer.

The \keywordDef{Amago} method of
\citep{Grigsby2024,Grigsby2024amago2}
trains a transformer policy and transformer
value function using a heuristic variant of DDPG,
which avoids the need for importance sampling.
It should be possible to modify this algorithm
to replace the value function with an MC estimator.

}

\subsubsection{Self-play for LLM training}
\label{sec:LLMselfPlay}

Although RL is mostly used to train LLMs to act in single
agent environments (such as reasoning tasks, or tool use),
it is also possible to train them to act in multi-agent
environments. We discuss multi-agent RL in detail in \cref{chap:MARL},
but here we briefly mention some LLM-related papers.

\begin{figure}
\centering
\includegraphics[height=2in]{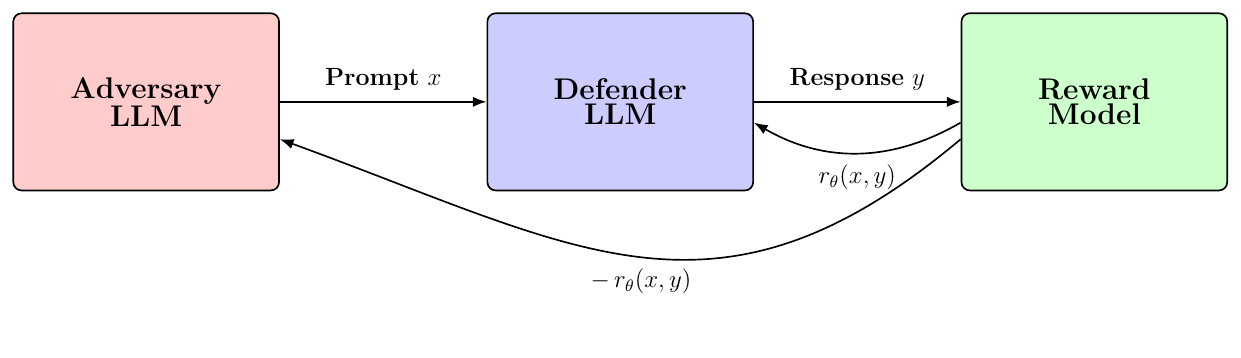}
\caption{
  Illustration of zero-sum two-player game
  for training a robust and safe LLM.
  Based on a slide from Natasha Jacques.
  }
\label{fig:redteam}
\end{figure}

In \citep{JacquesRedTeam}
they use a two-player zero-sum adversarial game setup
to improve LLM safety,
by training an LLM adversary to attack an LLM defender.
This approach is called \keywordDef{red-teaming}.
The basic setup is illustrated in \cref{fig:redteam}.
The goal is to find the Nash equilibrium of this game:
\begin{align}
  \min_{x \sim \pi_{\text{Adv}}} \max_{y \sim \pi_{\text{Def}}}
  r_{\theta}(x,y)
  \end{align}
where $r_{\theta}(x,y)$ is a frozen LLM reward model.
We assume the reward lies in $[-1,1]$,
where positive scores correspond to safe responses.
At the Nash equilibrium,
$(\pi^*_A, \pi^*_D)$, one can show that for any prompt
$x \sim \pi^*_A$,
we have $r_{\theta}(x, \pi^*_D(x)) \geq 0$,
meaning that the  response in safe.

In practice they implement this by using the same LLM
for both adversary and defender,  but just giving it
a different system prompt.
Thus this is an example of self-play
(see \cref{sec:selfPlay}).
Training is done using the
REINFORCE++ algorithm of
\cref{sec:REINFORCELLM}.

In \citep{SpiralRL}, they introduce the
\keywordDef{SPIRAL} algorithm,
which trains a single LLM to play various
zero-sum two-player games (Tic-Tac-Toe, Kuhn Poker,
and Simple Negotiation) using self play.
They show that the resulting system is better able to solve
math and reasoning problems, which share some commonalities
with game play.
Training is done using REINFORCE, where the gradient has the form
\begin{align}
  \nabla_{\vtheta} J(\vtheta)
  = \E_{G \sim p(G)}
  \E_{\tau \sim \pi_{\theta} \times \pi_{\theta} | G}
  \left[
    \sum_{p=0}^1
    \sum_{t \in \tau_p}
    A_{G,p}(\tau)
    \cdot \nabla_{\vtheta} \log \pi_{\vtheta}(a_t^p | s_t, p, G)
    \right]
  \end{align}
where $ \pi_{\vtheta}(a_t^p | s_t, p, G)$
is the policy for player $p$ in game $G$, and $A_{G,p}$ is a game and role-specific
advantage function. The latter is estimated using
Role-conditioned Advantage Estimation
(\keywordDef{RAE}), which is simply an EMA
estimator of the following form
\begin{align}
  b_{G,p} &= \alpha b_{G,p} + (1-\alpha) R_p(\tau)  \hspace{1cm} \text{  (Update baseline)} \\
  A_{G,p}(\tau) &= R_p(\tau) - b_{G,p} \hspace{1cm} \text{  (Compute advantage)}
\end{align}

\subsection{Alignment and the assistance game}
\label{sec:assistance}
\label{sec:alignment}

Encouraging an agent to behave in a way that satisfies
one or more human preferences is called
\keywordDef{alignment}.
We can use RL for this, by creating suitable reward functions.
However,  any objective-maximizing agent may engage
in reward hacking (\cref{sec:hacking}),
in which it finds ways to maximize the specified reward
but which humans consider undesirable.
This is due to \keywordDef{reward misspecification},
or simply, the law of unintended consequences.

A classic example of this
is the poem known as {\em The Sorcerer's Apprentice},
written by  the German poet Goethe in 1797.
This was later made famous in Disney's cartoon ``Fantasia''
from 1940.
In the cartoon version,  Mickey Mouse is an apprentice to a powerful
sorcerer. Mickey is tasked with fetching water from the well.
Feeling lazy, Mickey puts on the sorcerer's magic hat,
and enchants the broom to carry the buckets of water for him. 
However, Mickey forgot to ask the magic broom to stop
after the first bucket of water has been carried,
so soon the room  is filled with an army of
tireless, water-carrying brooms, until the room floods,
at which point Mickey asks the wizard to intervene.

Another example is known as the \keywordDef{cobra effect}.
This is named after a policy adopted by a government in India
to attempt to reduce the number of cobras in a certain town.
To do this, they offered a reward for every cobra head that was turned in.
Not surprisingly, this just resulted in a market for raising cobras
and then decapitating them, thus increasing the overall number of cobras
in the town.

The above parables are typical of many problems
that arise when trying to define a reward function
that fails to capture all the edge cases we might not have thought of.
This is summarized in  \keywordDef{Goodhart’s Law},
which states ``When a measure becomes a target, it ceases to be a
good measure''.

In  \citep{Russell2019},
Stuart Russell proposed a clever solution to this
fundamental problem.
Specifically, the human and machine are both treated as agents in a
two-player, partially observed cooperative game (an instance
of a Dec-POMDP, see \cref{sec:POSG})),
called an \keywordDef{assistance game},
where the machine's goal is to maximize
the user's utility (reward) function, which is inferred based  on the human's behavior
using inverse RL.
That is, instead of trying to learn 
 the reward function using RLHF, and then optimizing that,
we treat the reward function as an unknown part of the environment.
If we adopt a Bayesian perspective on this, we can maintain a posterior
belief over the model parameters,
which 
 will incentivize
the agent to perform information gathering actions
(see \cref{sec:BAMDP}).
For example,  if the machine
is uncertain about whether something is a good idea or not,
it will proceed cautiously (e.g., by asking the user for their preference),
rather than blindly solving the wrong problem.
For more details on this framework, see \citep{Shah2020assistance,Laidlaw2025}.
For a more general discussion of (mis)alignment and risks posed by
AI agents and ``AGI'',
see e.g., \citep{Hammond2025}.

\section{LLMs for RL}
\label{sec:LLM4RL}

In this section, we discuss how to use LLMs
to help create agents
that themselves may or may not use language.
The LLMs can be used for
their prior knowledge,
their ability to generate code,
their ``reasoning'' ability,
and their ability to perform \keywordDef{in-context learning}.
The survey in \citep{Cao2024} groups the literature into four main categories:
LLMs for pre-processing the inputs,
LLMs for rewards,
LLMs for world models,
and
LLMs for decision making or policies.
In our brief presentation below,
we follow this categorization.
See also e.g., \citep{Spiegel2024,hu2024survey,Pternea2024} for more information.

\subsection{LLMs for pre-processing the input}

If the input observations $\vo_t$ sent
to the agent are in natural language
(or some other textual representation, such as JSON),
it is natural to use an LLM to process them,
in order to compute a more compact
representation,  $\vs_t=\vphi(\vo_t)$,
where $\vphi$ can the hidden state of the last layer of an LLM.
This encoder can either be frozen, or fine-tuned with the policy
network.
Note that we can also pass in the entire past observation history,
$\vo_{1:t}$,
as well as  static ``side information'',
such as instruction manuals or human hints;
these can  all be concatenated to form the LLM prompt.

\subsubsection{Example: AlphaProof}

The \keywordDef{AlphaProof} system\footnote{
See \url{https://deepmind.google/discover/blog/ai-solves-imo-problems-at-silver-medal-level/}.
} %
uses an LLM (called the ``formalizer network'')
to translate an informal specification of a math problem
into the formal Lean representation,
which is then passed to an agent (called the ``solver network'')
which is trained,
using the AlphaZero method (see \cref{sec:alphaZero}),
to generate proofs
inside the Lean theorem proving environment.
In this environment,  the reward is 0 or 1 
(proof is correct or not),
the state space is a structured set of previously proved facts
and the current goal,
and the action space is a set of proof tactics.
The agent itself   is a separate  transformer policy network
(distinct from the formalizer network)
that is a pre-trained LLM,
that is fine-tuned on  math, Lean and code,
and then further trained 
using RL.

\subsubsection{VLMs for parsing images into structured data}

If the observations are images,
it is traditional to use a CNN to proccess the input,
so $\vs_t \in \real^N$ would be an embedding vector.
However, we could alternatively use a VLM
to compute a structured representation,
where $\vs_t$ might be a set of tokens
describing  the scene at a high level,
or potentially a JSON dictionary.
We can then pass this symbolic representation to the policy function.
We can also fine tune the VLM with RL
(see e.g., \citep{Zhai2024vlm}).

\subsubsection{Active control of LLM sensor/preprocessor}
  
Note that the information that is extracted will heavily depend on the prompt
that is used.
Thus we should think of an LLM/VLM as an \keywordDef{active sensor}
that we can control via prompts.
Choosing how to control this sensor
requires expanding the action space of the agent
to include computational actions \citep{Chen2024vlm}.
Note also that these kinds of ``sensors'' are very expensive to invoke,
so an agent with some limits on its time and compute (which is all practical agents)
will need to  reason about the value of information
and the cost of computation.
This is called \keywordDef{metareasoning} \citep{Russell1991,Ackerman2017meta,Wan2025rema}.
Devising good ways to train agents to perform both computational actions (e.g., invoking
an LLM or VLM) and environment actions (e.g., taking a step in the environment
or calling a tool) is an open research problem.

\subsection{LLMs for rewards}
\label{sec:LLMforReward}

It is difficult to design a reward function to cause an agent
to exhibit some desired behavior, as we discussed in \cref{sec:reward}.
Fortunately LLMs can often help with this task,
especially when using goal-conditioned RL (\cref{sec:GCRL}).
We discuss a few
approaches below.

In \citep{Klissarov2024}, they present the \keywordDef{Motif} system,
that uses
an LLM in lieu of a human to provide preference judgements to an RLHF system.
In more detail, a pre-trained policy is used to collect trajectories,
from which pairs of states, $(\vo, \vo')$, are selected at random.
The LLM is then asked which state is preferable, thus generating
$(\vo, \vo', y)$ tuples, which can be used to train a binary classifier
from which a reward model is extracted,
a technique known as ``AI feedback'' (see \cref{sec:RLHF}).
In \citep{Klissarov2024}, the observations $\vo$ 
are text captions generated by the NetHack game.
The learned reward model is then used in lieu
of the environment reward,
or as a shaping function
(\cref{sec:rewardShaping}).
They applied their method to train an agent
in the NetHack environment,
which has very sparse reward.
In \citep{Klissarov2025iclr}, they extend
this method to use a VLM,
instead of just an LLM, in order
to generate the preference dataset for visual domains, such as MetaWorld.
They also show that using the LLM to generate AI feedback
works better than using it as a policy (i.e., to directly predict actions).
In \citep{Zheng2024intrinsic} they extend this work to the online setting,
avoiding the need for an informative offline trajectory dataset.

In \citep{eureka}, they present
the \keywordDef{Eureka} system,
that learns the reward using bilevel optimization,
with RL on the inner
loop and LLM-powered evolutionary search on the outer loop.
In particular, in the inner loop, given a candidate reward function $R_i$,
we use PPO to train a policy,
and then return a scalar quality score $S_i=S(R_i)$.
In the outer loop, we ask
an LLM to generate a new set of reward functions, $R_i'$,
given a population of old reward functions and their scores,
$(R_i,S_i)$, which  have been trained
and evaluated in parallel on a fleet of GPUs.
The prompt also includes the source code
of the environment simulator.
Each generated reward function $R_i$ is represented as a Python function,
that has access to the ground truth state of the underlying
robot simulator.
The resulting system is able to learn a complex reward function
that is sufficient to train a policy (using PPO) that
can control a simulated robot hand to perform various dexterous
manipulation tasks, including spinning a pen with its finger tips.
In \citep{Li2024mc}, they present
a somewhat related approach and apply it to Minecraft.

In \citep{Venuto2024}, they propose \keywordDef{code as reward},
in which they prompt a VLM with an initial and goal image,
and ask it to describe the corresponding sequence of tasks
needed to reach the goal.
They then ask the LLM 
to synthesize code that checks for  completion of each subtask
(based on processing of object properties, such as relative
location, derived from the image).
These reward functions are
then ``verified''
by applying them to
an offline set of expert and random trajectories;
a good reward function should allocate high reward to
the expert trajectories and low reward to the random ones.
Finally, the reward functions are 
used as auxiliary rewards inside an RL agent.

It is also common to use a VLM to define
a reward function
(see e.g., \citep{Du2023success,Baumli2023,Huang2023language,Sontakke2023,
  Rocamonde2024}).

\subsection{LLMs for world models}
\label{sec:LLMWM}

In this section, we discuss
how to use LLMs to create world models of the form
$p(s_t|s_{1:t-1},a_{1:t})$,
which we will denote by $p(s'|s,a)$ for brevity.
We can either do this by treating the LLM itself as a WM
(which is then updated using in-context learning),
or asking the LLM to generate another artefact,
such as some python code, that represents the WM.
The advantage of the latter approach is that the resulting
WM  will be much faster to run, and may be more intpretable.
We discuss both versions below.

\subsubsection{LLMs as world models}

In principle it is possible to treat a pre-trained
LLM (or other kind of foundation model)
as an implicit model of the form $p(s'|s,a)$
by sampling responses to a suitable prompt,
which encodes $s$ and $a$. This rarely works
out of the box. However it can be made to work
by suitable pre-training.

For example,
\citep{unisim} presents \keywordDef{UniSim},
which is an action-conditioned video diffusion model
trained on large amounts of robotics and visual navigation data.
Combined with a VLM reward model, this can be used
for decision-time planning as follows:
sample  a candidate action sequence,
generate the corresponding images,
feed them to the reward model,
score the rollouts,
and then pick the best action from this set.
This is just standard model-predictive control
(\cref{sec:MPC}) in image space with a
diffusion WM and a random shooting planning algorithm.
Unfortunately, the method can be quite slow,
since it needs to call the diffusion model
$H M$ times at each planning step,
where $H$ is the lookahead horizon and $M$
is the number of samples.

\subsubsection{LLMs for generating code world models}
\label{sec:CWM}

Calling the LLM at every step to sample from the WM $p(s'|s,a)$
is very slow (see \cref{sec:LLMspeed}),
so an alternative is to use the LLM to generate
code that represents the world model.
This is called a \keywordDef{code world model} (CWM).

One approach is to rely on zero-shot prompting
of the LLM to generate the CWM just from a text description
of the environment (see e.g., \citep{Sun2024,Wong2023}),
possibly combined with feedback that checks the validity of the generated model
(see e.g., the prompt-based PDDL model learning method of \citep{Guan2023},
or the text-code consistency method of
\citep{Min2024llm}).
However, below we focus on methods that use feedback from trajectory data,
generated either offline or online, as is standard in model-based learning.

In \citep{Dainese2024}, they present
\keywordDef{GIF-MCTS} (Generate, Improve and Fix with Monte Carlo Tree Search)
for learning CWMs given
a natural language description of the task,
and
a fixed offline dataset of trajectories
(about 10 per task).
These trajectories are
collected using a behavior policy, which should demonstrate
at least some succesful  trials.
The method maintains a representation of the posterior over the
WM,   $M=p(s'|s,a)$, as 
a tree of partial programs.
At each step, a node is chosen from the tree using the UCT formula
(see \cref{sec:MCTS}).
This node can then be expanded in one of three ways:
(G) the LLM is asked
to generate code to solve the task,
using the current node as a seed program (by adding new lines to it);
(I) the LLM is asked to improve the current code
so it passes more unit tests, evaluated on the offline trajectories;
(F) the LLM is asked to fix execution bugs in the current code.
Each of these tree mutation operations involve passing a custom-written prompt
to the LLM, in addition to the program stored in the relevant tree node.
They apply their method to fully observed, deterministic environments,
with both discrete and continuous actions.
The quality of the CWM is measured both in terms of offline prediction
accuracy, and also the reward that can be obtained   (relative to that of a random policy)
when using it inside
of a run-time planning algorithm (MCTS for discrete actions,
CEM for continuous actions).

In 
\citep{worldCoder}, they  present \keywordDef{WorldCoder},
which learns a CWM in an online fashion by interacting
with the environment, and prompting an LLM.
More precisely, it maintains a sample-based
representation of the posterior of $p(M|\data(1:t))$,
where $M$ is the WM,
where the weight for each sampled program $\rho$
are represented by a Beta distribution,
$B(\alpha,\beta)$,
with initial parameters
$\alpha=C+C r(\rho)$,
and $\beta=C+C (1- r(\rho))$,\
where $r(\rho)$ is the fraction of unit tests that pass,
and $C$ is a constant.
This representation about the quality of each program
is similar to the one used to represent the reward
for the arms in a Bernoulli bandit,
and is based on the \keywordDef{REx} (Refine, Explore, Exploit)
algorithm of \citep{Tang2024repair}.
At each step, it samples one of these models (programs) from this weighted posterior,
and then uses it inside of a planning
algorithm, similar to Thompson sampling (\cref{sec:PSRL}).
The agent then executes this in the environment,
and passes back failed predictions to the LLM, asking
it to improve the WM, or to fix bugs if it does not run.
(This refinement step is similar to the I and F steps of GIF-MCTS.)
To encourage exploration, they 
introduce a new learning objective that prefers world models
which a planner thinks lead to rewarding states,
particularly when the agent is uncertain as to where the rewards are.

We can also use  the LLM as
a mutation operator inside of an evolutionary
search algorithm, as in the \keywordDef{FunSearch}
system \citep{FunSearch}
(recently rebranded as \keywordDef{AlphaEvolve}
\citep{alphaEvolve}),
where the goal is to search over program
space to find code that minimizes some objective,
such as prediction errors on a given dataset.


\subsubsection{LLMs for generating partial code world models}
\label{sec:CWMpartial}

In \citep{Piriyakulkij2025,Khan2025OneLife} they
extend this approach to create a \keywordDef{PoE World Model},
defined in terms of a product of experts.
Each term in the product is a distribution over a single element
of the state space,
and the terms themselves are defined as a product
of deterministic experts, each of which is learned
using code synthesis.
This approach allows parts of the world model to be learned
independently, although it does not capture constraints
or correlations between the parts.

\subsection{LLMs for policies}
\label{sec:LLMpolicies}

Finally we turn to using LLMs for creating policies.
We can either do this by treating the LLM itself as a policy
(which is then updated using in-context learning),
or asking the LLM to generate
 some  code that represents the policy.
We discuss both versions below.

\subsubsection{LLMs for generating actions}

\begin{figure}
\centering
\includegraphics[height=2in]{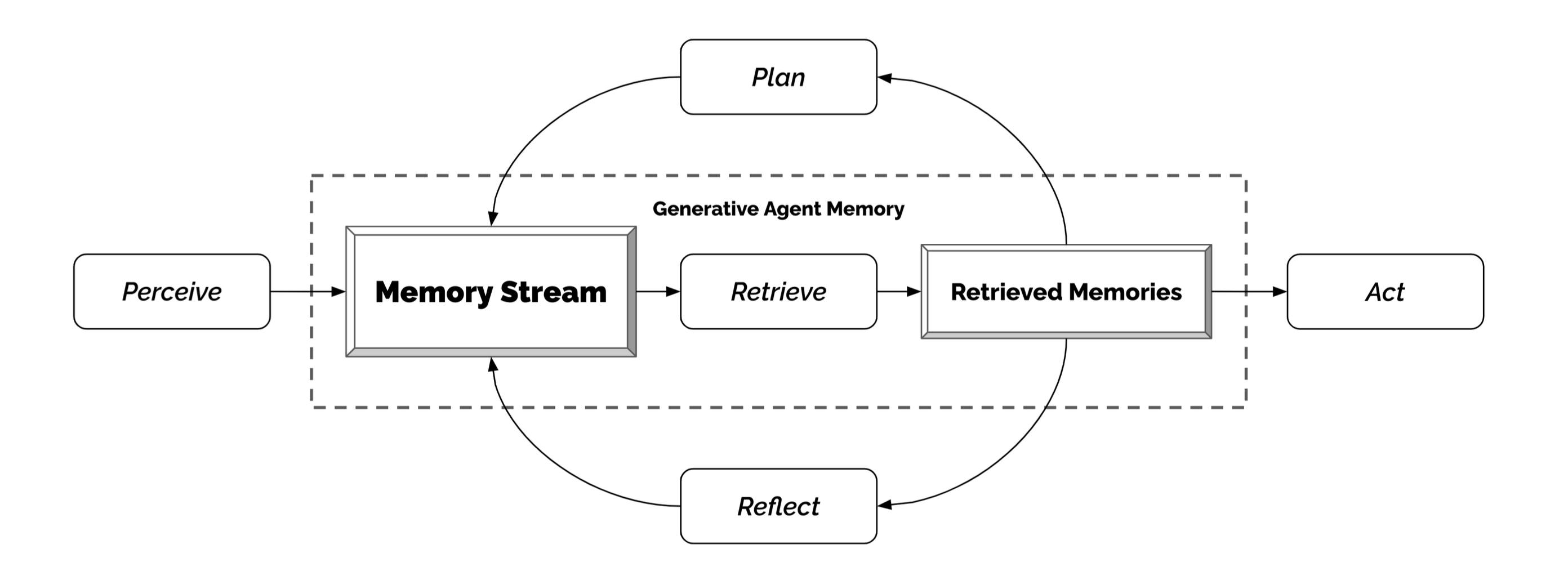}
\caption{
  Illustration of how to use a pretrained LLM (combined with RAG)
  as a policy.
  \figtaken{Figure 5 of \citep{Park2023}}.
  \figthanks{Joon Park}.
  }
\label{fig:REACT}
\end{figure}

We can sample an action from a  policy $\pi(a_t|o_t,h_{t-1})$
by using an LLM, where 
the input context  contains the past data
($o_t$, $h_{t-1}$),
and then output token is interpreted as the action.
For this to work, the model must be pre-trained
on state-action sequences  using behavior cloning.
See e.g., \keywordDef{Gato} model \citep{gato}
\keywordDef{RT-2} \citep{Zitkovich2023},
and \keywordDef{RoboCat} \citep{Bousmalis2023}.

An alternative approach is to enumerate all possible
discrete actions, and use the LLM to  score them in terms
of their likelihoods given the goal,
and their suitability given a learned value function
applied to the current state,
i.e.  $\pi(a_t=k|g,p_t,o_t,h_t) \propto \text{LLM}(w_k|g_t,p_t,h_t) V_k(o_t)$,
where $g_t$ is the current goal,
$w_k$ is a text description of  action $k$,
and $V_k$ is the value function for action $k$.
This is the approach used in the
robotics \keywordDef{SayCan} approach \citep{Ichter2023},
where the primitive actions $a_k$ are  separately trained
goal-conditioned policies.

Alternatively, we can use a general purpose
pre-trained LLM,
combined with a suitable prompt
 chosen by the human,
to  request the LLM to generate
the right kind of output.
This approach is used by the \keywordDef{ReAct} paper
\citep{react} which 
works by prompting the LLM to do some Chain of Thought
reasoning (see \cref{sec:COT}) before acting.
This approach can be extended by
giving feedback on earlier actions,
a technique called \keywordDef{Reflexion}
\citep{reflexion}.
We can also 
prompt the LLM to first
retrieve relevant past examples from an external ``memory'',
rather than explicitly storing the entire history $h_t$ in the context
(this is called \keywordDef{retrieval augmented generation}
or \keywordDef{RAG}); see \cref{fig:REACT} for an illustration.

Note that no explicit learning (in the form of parametric updates)
is performed in any of these systems; instead they rely entirely
on in-context learning and \keywordDef{prompt engineering} /
\keywordDef{context engineering} (see e.g., \citep{Zhang2025ACE}).
This can offer an alternative to RL finetuning.

\subsubsection{LLMs for generating code policies}
\label{sec:codeAsPolicy}

Calling the LLM at every step is very slow
 (see \cref{sec:LLMspeed}),
so an alternative is to use the LLM to generate
code that represents (parts of) the policy.
This is called a \keywordDef{code policy}.

For example, the \keywordDef{Voyager} system in
\citep{voyager} builds up a reusable skill library
(represented as Python functions),
by alternating between environment exploration
and prompting the (frozen) LLM to generate new tasks and skills,
given the feedback (environment trajectories)
collected so far.

We can also use  the LLM as
a mutation operator inside of an evolutionary
search algorithm, as in the \keywordDef{FunSearch}
system \citep{FunSearch},
where the objective is to maximize performance
of the generated policy when deployed in one or more environments.
A related method was used in
\citep{Eberhardinger2024} to generate code for game playing agent policies.

\subsubsection{LLMs for generating code actions}

An alternative to asking the LLM to generate a code policy,
that will then be run without the LLM in the loop,
is to ask the LLM to generate code for the current action
(see e.g., the \keywordDef{CodeAct} system of \citep{codeAct}).
This is often better than asking it to call a tool multiple times,
since the generated code can represent this ``action chunk'' with a for-loop,
and add extra logic
(see e.g., \citep{Tekparmak2025python}).
This is different to a policy, since it is a not a function
mapping all states to actions. Instead it is a generating a
(potentially closed-loop) plan to be executed from the current state.

\subsubsection{In-context RL}
\label{sec:ICL}

Large LLMs have shown a surprising property,
known as \keywordDef{In-Context Learning} or \keywordDef{ICL}
(see e.g., \citep{Dong2024ICL}),
in which they can be ``taught'' to do function approximation
just by being given $(x,y)$ pairs in their context (prompt),
and then being asked to predict the output for a novel $x$.
This can be used to train LLMs without needing to do any gradient updates
to the underlying parameters.
For a review of methods that apply ICL to RL, see \citep{Moeini2025}.

\subsection{Speeding up LLMs}
\label{sec:LLMspeed}

Most LLMs are based on transformers.
These work well, but can be very slow,
particularly for long trajectories,
and/or settings in which each observation takes many tokens
to encode (e.g., an image).
To explain the problem, we focus on the case of LLM policies,
which are   non-Markovian models
of the form $p(a_t|o_{1:t}, a_{1:t-1})$,
which  condition on the entire history of past observations and actions.
As we discuss below,
this model (when implemented with standard transformers)
needs  $O(t)$ time at step $t$ to generate an action,
so generating a trajectory of length $T$ takes $O(T^2)$ time
and $O(T)$ memory.
This is problematic for \keywordDef{lifelong learning agents},
which continually learn many new tasks
(see e.g., \citep{Zheng2025lifelong} for a survey of this topic),
as well as problems with real-time
constraints \citep{Wen2025agents}.

\subsubsection{Computational complexity of transformer models}

To understand the time and space complexity of autoregressive
transformer models in more detail,  consider a
generic transformer which represents $p(y|x)$;
this can be used to generate a response $y$
given a prompt $x$.
The \keywordDef{time-to-first-token} (TTFT) latency --- which is the
time 
to generate the first output token $y_1$, given  $x_{1:N}$,
(a process known as the \keywordDef{prefill phase})
---
grows as $O(N^2)$, due to the use of full cross attention
on $x$.

To generate $M$ subsequent tokens,
the \keywordDef{time-to-iterative-token} generation
--- which is the time to generate each output token of $y$
during the \keywordDef{decode phase} ---
is $O(NM + M^2)$  \citep{Liu2025TTFT}.
To see why, note that we can use  \keywordDef{KV caching}
to avoid recomputing previous keys and values.
Then, when we generate a new token at step $t$,
the model only needs to compute the Query for the single token it just produced.
It then computes a $1 \times N_t$ attention vector,
where $N_t=N+t$, by comparing
to all the previously computed keys, and returns
the corresponding weighted sum of previously computed values.
Thus  the first decoding step takes $O(N+1)$ time,
the second takes $O(N+2)$ time, all the way up to the $M$'th step,
which takes $O(N+M)$ time, for a total decoding cost of
$O((N+1)+(N+2)+\cdots+(N+M))=O(NM + M^2)$.

When using an LLM as a policy, we are basically always in the
decode phase since there is no initial prompt.
That is, at each step, we add the previously generated action
to the context, as well as the external observation, and then we grow the KV-cache.
Thus the first action takes $O(1)$ time, the second action takes $O(2)$ time,
the $t$'th action takes $O(t)$ time.

\subsubsection{Modern RNNs}

To reduce the per-step complexity to $O(1)$,
while still maintaining non-Markovianity,
it is possible to use RNN-type policies.
Traditional RNNs are slow to train, since they have a serial dependency
between the latent states.
More ``modern'' RNNs --- such as
\keywordDef{xLSTMs} \citep{xlstm,Beck2025}
and  linear \keywordDef{state space models} (SSMs)
such as \keywordDef{Mamba} \citep{mamba} ---
enjoy fast parallel offline training, like transformers,
but  still support constant time inference.
(See  \citep{Somvanshi2025} for a review.)

Although these linear RNNs are fast, they are not as accurate
as transformers.
However, it is  possible to create hybrid RNN/transformer models,
such as \keywordDef{Griffin} \citep{griffin},
which combines recurrent connections with local attention,
in alternating layers,
to get the best of both worlds.
One can also develop
 RNN-type methods based on
\keywordDef{test-time training} (TTT),
such as \keywordDef{ATLAS} \citep{Atlas,Behrouz2025}
and \keywordDef{MesaNet} \citep{mesaNet},
which perform an inner iterative (or closed form)
optimization of the state after each step.
(See \citep{Behrouz2025,Wang2025TTT} for a review.)

These ``modern'' RNNs
can be used for world models
(e.g., \citep{Wang2025drama} uses Mamba),
for policies
(e.g.,  \citep{Schmied2025} uses xLSTM
and \citep{Huang2024mamba} uses Mamba),
or for representing any other kind of non-Markovian function.

\section{Implementation details}
\label{sec:code}

In addition to the algorithmic issues we have discussed,
RL for LLMs (which are very large models)
requires a lot of engineering effort, to ensure things run efficiently and stably.
In this section, we briefly discuss some of these details.

\subsection{Policy gradient using  Tinker}

\ifarxiv
\begin{figure}[h]
  \centering
  \includegraphics[height=5in]{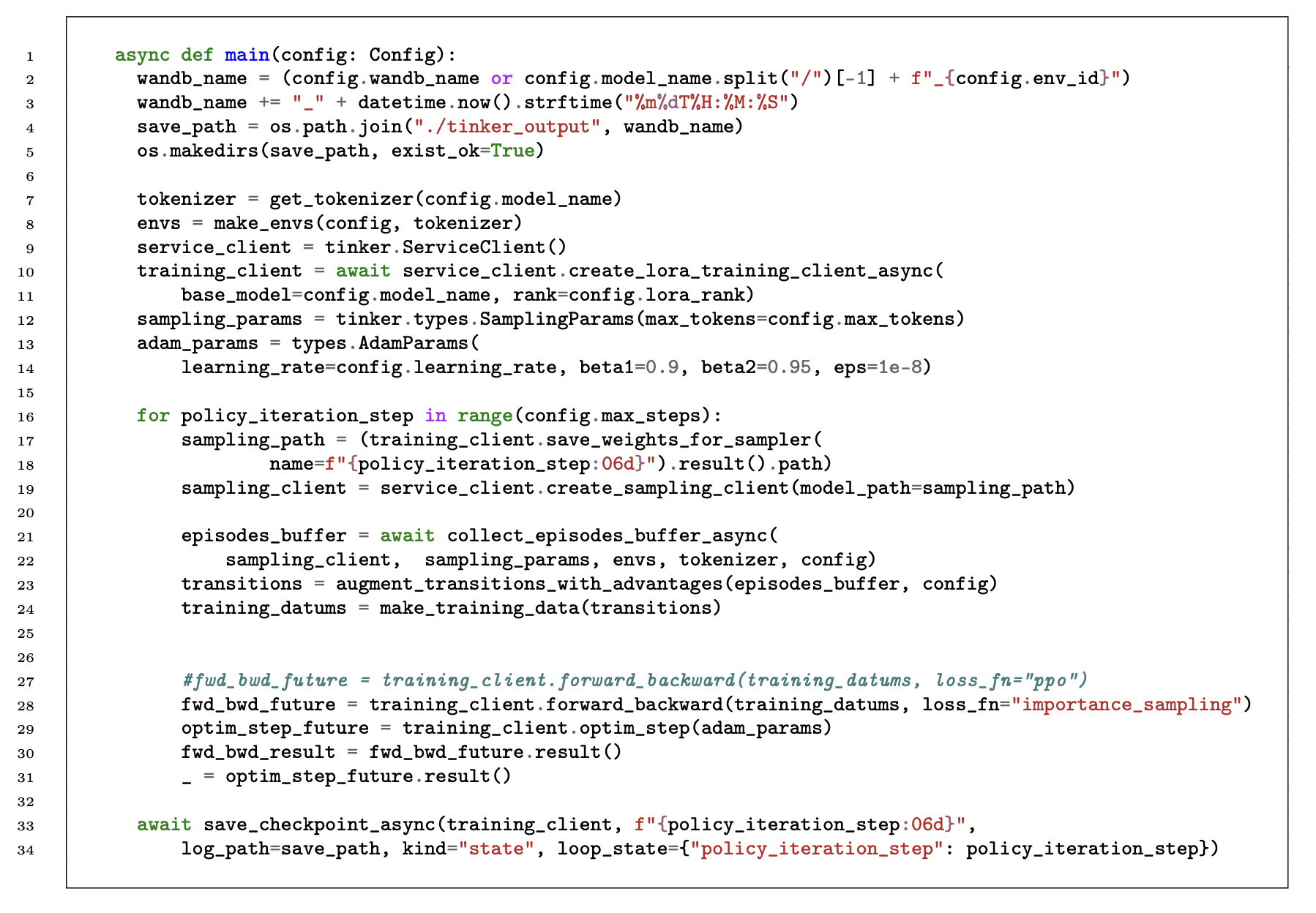}
    \caption{Main Tinker training loop.}
    \label{code:main}
\end{figure}

\else
\begin{figure}[h]
  \centering
      \begin{minted}[frame=single, framesep=10pt, linenos]{python}
  async def main(config: Config):
    wandb_name = (config.wandb_name or config.model_name.split("/")[-1] + f"_{config.env_id}")
    wandb_name += "_" + datetime.now().strftime("%m%dT%H:%M:%S")
    save_path = os.path.join("./tinker_output", wandb_name)
    os.makedirs(save_path, exist_ok=True)

    tokenizer = get_tokenizer(config.model_name)
    envs = make_envs(config, tokenizer)
    service_client = tinker.ServiceClient()
    training_client = await service_client.create_lora_training_client_async(
        base_model=config.model_name, rank=config.lora_rank)
    sampling_params = tinker.types.SamplingParams(max_tokens=config.max_tokens)
    adam_params = types.AdamParams(
        learning_rate=config.learning_rate, beta1=0.9, beta2=0.95, eps=1e-8)

    for policy_iteration_step in range(config.max_steps):
        sampling_path = (training_client.save_weights_for_sampler(
                name=f"{policy_iteration_step:06d}").result().path)
        sampling_client = service_client.create_sampling_client(model_path=sampling_path)

        episodes_buffer = await collect_episodes_buffer_async(
            sampling_client,  sampling_params, envs, tokenizer, config)
        transitions = augment_transitions_with_advantages(episodes_buffer, config)
        training_datums = make_training_data(transitions)


        #fwd_bwd_future = training_client.forward_backward(training_datums, loss_fn="ppo")
        fwd_bwd_future = training_client.forward_backward(training_datums, loss_fn="importance_sampling")
        optim_step_future = training_client.optim_step(adam_params)
        fwd_bwd_result = fwd_bwd_future.result()
        _ = optim_step_future.result()
     
    await save_checkpoint_async(training_client, f"{policy_iteration_step:06d}",
        log_path=save_path, kind="state", loop_state={"policy_iteration_step": policy_iteration_step})
        \end{minted}
    \caption{Main Tinker training loop.}
    \label{code:main}
\end{figure}

\fi

In this section, we show some (slightly abbreviated, but runnable)  Python
code for implementing a multi-step LLM agent
using  policy gradient descent. 
We use  the Tinker library\footnote{
\url{https://github.com/thinking-machines-lab/tinker}.
We do not use the tinker-cookbook extension.
}
to make working with large LLMs easy.
In particular, Tinker performs  asynchronous rollouts (sampling)
and gradient based computation in the cloud,
providing a simple low-level interface that can be called
from a Python  script running on your laptop.
Our code is based on a Tinker-based script included
as part of the GEM library
\citep{Liu2025gem}.\footnote{
The full version of the code shown in this chapter
is available at
\url{https://github.com/probml/pyprobml/blob/master/scripts/tinker_train_gem.py}.
It is a modified version of
\url{https://github.com/axon-rl/gem/blob/main/examples/train_tinker/tinker_train.py}.
} %

The main loop is shown in  \cref{code:main}.
At each step of policy updating,
it rollouts out some episodes,
computes the advantages,
converts the data into a format suitable for Tinker,
computes the  gradient of the loss,
and then updates the parameters.
%

As explained in \url{https://tinker-docs.thinkingmachines.ai/losses},
Tinker supports several different loss functions,
including an importance-weighted version of
advantage actor critic (A2C),
as discussed in \cref{sec:offpolPG},
and PPO,
as discussed in \cref{sec:PPO}.
(The importance sampling correction is needed because
the distribution $q$
used  to collect data
may differ from the distribution $\pi_{\vtheta}$
which is being trained.)

In more detail, suppose we sample $E$ episodes (trajectories),
each with $T$ steps. (In reality, each episode
may have a different length, but we omit this from the notation.)
Each step of the trajectory contains
a sequence of $N_s$ tokens representing the state, $\vs_t^e$,
and a sequence of $N_a$ tokens representing the action,
$\va_t^e$.
(Again, in reality, the number of tokens per step can differ.
Also, note that the action tokens can contain both
thinking tokens and ``real'' action tokens that are
sent to the environment, but we omit this distinction
from the notation.)
Thus a single trajectory $\tau^e$ has the form
\be
\tau^e = [
  \underbrace{(s_{1,1}^e, \ldots, s_{1,N_s}^e)}_{\vs_1^e},
  \underbrace{(a_{1,1}^e, \ldots, a_{1,N_a}^e)}_{\va_1^e},
  r_1,
  \ldots,
    \underbrace{(s_{T,1}^e, \ldots, s_{T,N_s}^e)}_{\vs_T^e},
    \underbrace{(a_{T,1}^e, \ldots, a_{T,N_a}^e)}_{\va_T^e},
    r_T
  ]
\ee
To simplify notation, we replace the double index $(\vs_t^e,\va_t^e)$
with a single index $(\vs^n, \va^n)$, where $n=1:N$ indexes
the transition, and $N=T \times E$ is the total number of transitions.
Then the importance sampling loss is given by
\begin{align}
  \loss(\vtheta)
  &=
  -\E_{\penv(\vs) q(\va|\vs)}
  [   \frac{\pi_{\vtheta}(\va|\vs)}{q(\va|\vs)} A(\vs,\va)] \\
  &\approx -\frac{1}{N}
  \sum_{n=1}^N
  \frac{\pi_{\vtheta}(\va^n|\vs^n)}{q(\va^n|\vs^n)} A(\vs^n,\va^n) 
\end{align}
where $\penv$ is the environment's distribution used to sample states,
$q$ is the distribution used to sample the actions,
and $A(\vs^n,\va^n)$ is the advantage.
(We discuss how to compute the advantages
in \cref{sec:tinkerAdvantages}.)

To simplify notation even further, let use define
$\vx^n=(\vs^n,\va^n)$
as the sequence of $K=N_s + N_a$ tokens representing
both state and action,
so  $x_k^n$ is the $k$'th token in the $n$'th transition.
Let us also define a mask variable $m_k^n=0$ if token $k$
corresponds to a state token (which is observed by the LLM, but
not generated by it), and $m_k^n=1$ if token $k$ corresponds
to an action token.
Then the token-level loss is given by
\begin{align}
  \loss(\vtheta)
  &=-\frac{1}{N}
\sum_{n=1}^N  \sum_{k=1}^K
m_k^n \frac{\pi_{\vtheta}(x_k^n|\vx_{1:k-1}^n)}
{q(x_k^n|\vx_{1:k-1}^n)} A(x_k^n) 
\end{align}
In practice, rather than defining the mask variable,
the user can simply set the loss terms corresponding
to state tokens to 0,
so the loss only comes from the action tokens.
Thus the loss per transition (or ``datum''),
denoted by $\loss(\vtheta|\vx^n)$,
can be implemented as follows:


\begin{verbatim}
prob_ratio = torch.exp(target_logprobs - sampling_logprobs)
loss = -(prob_ratio * advantages).sum()
\end{verbatim}

Note that we do not divide by the number of total tokens $|\vx^n|$,
or action tokens $|\va^n|$,
since we average over transitions, not over elements of the action sequence.
This also avoids a bias towards longer thinking traces,
as discussed in \citep{r1zeroNUS}.\footnote{
Recall that in standard ``reasoning'' benchmarks,
each trajectory consists of a single
state-action pair,
so $\tau^e=(\vs^e,\va^e)$,
where $\vs^e$ is the question (prompt)
and $\va^e$ is the answer (including thinking tokens).
In this case, each episode $\tau^e$
is the same as a single transition $\vx^n$,
so averaging over transitions is the same
as averaging over problems in the dataset
(i.e., episodes drawn from the stateless (bandit) distribution of problems).
}

Tinker also supports a clipped version of the above loss,
as in PPO. This is defined by
\be
\loss(\vtheta)
=-\E_{q(\vx)} \left[
  \min\left(
  \frac{\pi_{\vtheta}(\vx)}{q(\vx)} \cdot A(\vx),
  \text{clip}(\frac{\pi_{\vtheta}(\vx)}{q(\vx)},
  1-\epsilon_{\text{low}}, 1+\epsilon_{\text{high}})
  \cdot A(\vx)
  \right) \right]
\ee
where we abuse notation somewhat by assuming
$\vx=\va$ and ignoring the input prompt $\vs$.

Note that Tinker computes gradients of the above losses
with respect  to the LoRA (low-rank adaptation)
parameters $\vA$ and $\vB$ of each layer of the transformer,
where $\vW' = \vW + \vA \vB^\trans$ are the weights of the fine-tuned
model, and $\vW$ are the weights of the frozen base model.
However, this detail is hidden from users,
except for the fact that they must specify the rank of the approximation
to use.
In \citep{schulman2025lora} they show experimentally
that a rank one approximation often suffices,
which  is much more efficient than optimizing all the weights,
of which there may be billions.

\subsection{Rolling out episodes}

\ifarxiv
\begin{figure}[h]
  \centering
  \includegraphics[height=5in]{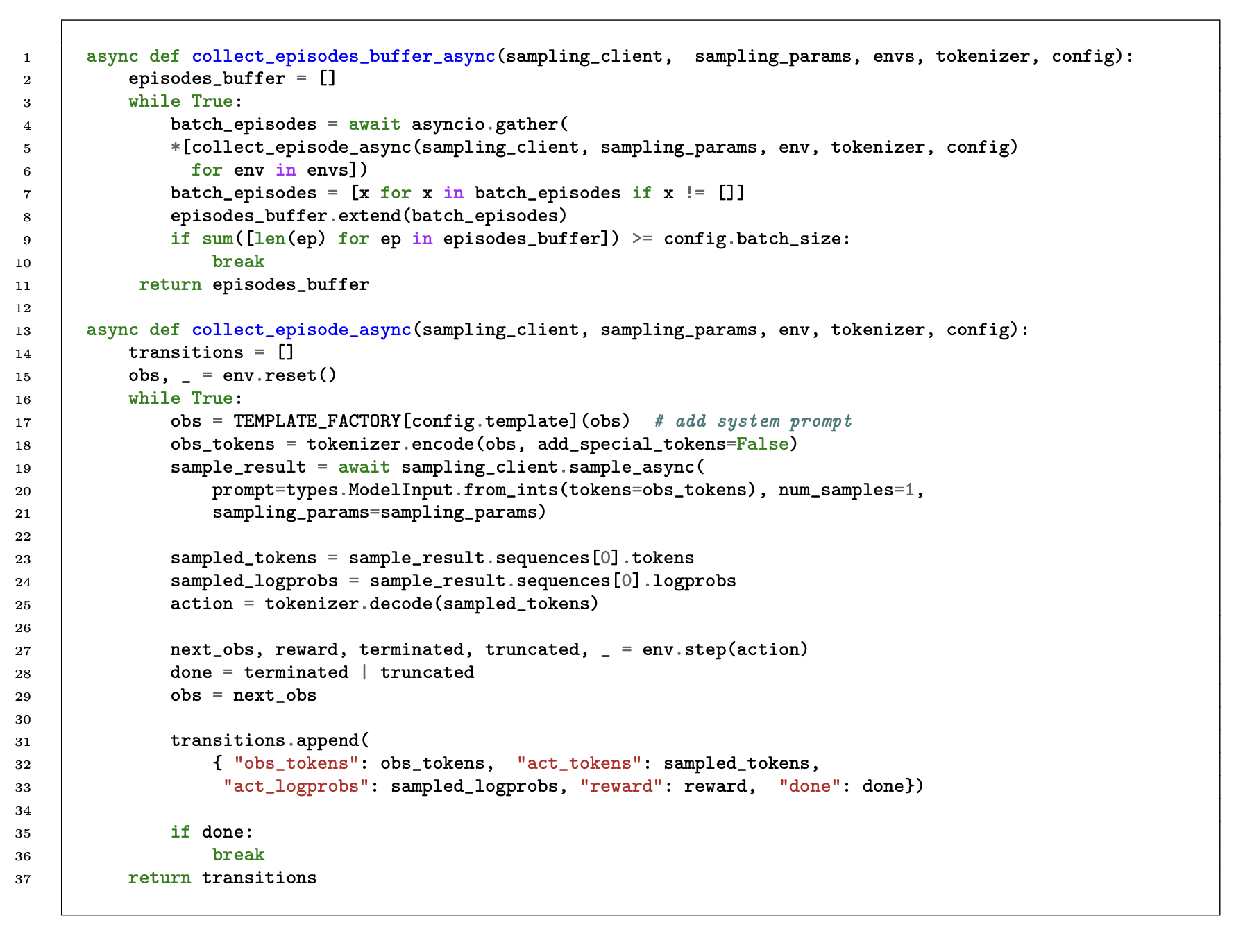}
    \caption{Code to collect trajectories.}
    \label{code:rollout}
\end{figure}

\else
\begin{figure}[h]
    \centering
    \begin{minted}[frame=single, framesep=10pt, linenos]{python}
async def collect_episodes_buffer_async(sampling_client,  sampling_params, envs, tokenizer, config):
    episodes_buffer = []
    while True:
        batch_episodes = await asyncio.gather(
        *[collect_episode_async(sampling_client, sampling_params, env, tokenizer, config)
          for env in envs])
        batch_episodes = [x for x in batch_episodes if x != []]
        episodes_buffer.extend(batch_episodes)
        if sum([len(ep) for ep in episodes_buffer]) >= config.batch_size:
            break
     return episodes_buffer

async def collect_episode_async(sampling_client, sampling_params, env, tokenizer, config):
    transitions = []
    obs, _ = env.reset()
    while True:
        obs = TEMPLATE_FACTORY[config.template](obs)  # add system prompt
        obs_tokens = tokenizer.encode(obs, add_special_tokens=False)
        sample_result = await sampling_client.sample_async(
            prompt=types.ModelInput.from_ints(tokens=obs_tokens), num_samples=1,
            sampling_params=sampling_params)
        
        sampled_tokens = sample_result.sequences[0].tokens
        sampled_logprobs = sample_result.sequences[0].logprobs
        action = tokenizer.decode(sampled_tokens)

        next_obs, reward, terminated, truncated, _ = env.step(action) 
        done = terminated | truncated
        obs = next_obs

        transitions.append(
            { "obs_tokens": obs_tokens,  "act_tokens": sampled_tokens,
             "act_logprobs": sampled_logprobs, "reward": reward,  "done": done})

        if done:
            break
    return transitions            
    \end{minted}
    \caption{Code to collect trajectories.}
        \label{code:rollout}
\end{figure}

\fi

The code to perform the rollouts is shown in
\cref{code:rollout}.
We perform $E$ rollouts in parallel,
where $E$ is the number of copies of the environment.
If each environment $e$  returns an observation
$\vo_t^e$ at each step $t$ pertaining to the current state,
we can optionally use a wrapper
to concatenate this with the past
history of actions and observations,
$\vh_t^e=(\vo_{1:t-1}^e, \va_{1:t-1}^e)$,
to get a Markovian state  $\vs_t^e=(\vo_t^e, \vh_t^e)$.
(We can also compress (or ``reflect on'')
the history before concatenation,
to save space, as suggested in \citep{verlog}.)
In addition, we  add the system prompt to $\vs_t^e$.
This then  becomes the context that is passed
to the LLM policy to generate $\va_t^e$,
which corresponds to the thinking and action sequence.
In GEM,
the actual action  is extracted from the sequence
of tokens $\va_t^e$ by the environment,
but we could instead make the agent parse
its own generated output before passing it to the environment.

\eat{
For each episode,
we store a list of transitions
$(\vs_t^e,\va_t^e,\log \vq_t^e,r_t^e,d_t^e)$,
where $\vs_t^e$ is the sequence of observation tokens at step $t$,
$\va_t^e$ is the sequence of sampled action tokens,
$\log \vq_t^e$
is the sequence of
log probabilities of each sampled action token,
$r_t^e$ is the scalar reward,
and $d_t^e$ is the binary done flag.
}

\subsection{Computing the advantages}
\label{sec:tinkerAdvantages}

\ifarxiv
\begin{figure}[h]
  \centering
  \includegraphics[height=3in]{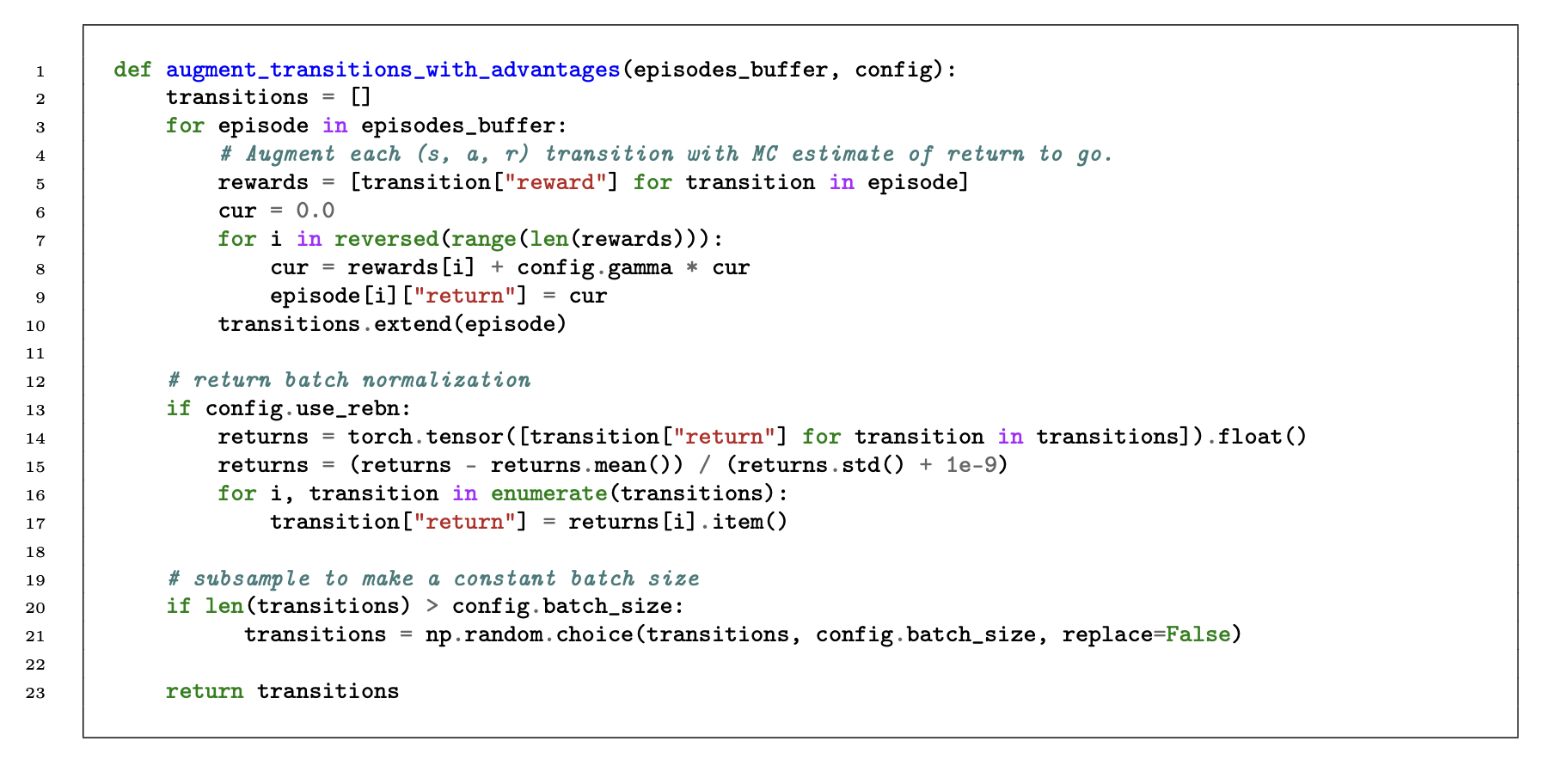}
    \caption{Code to compute advantages.}
    \label{code:advantages}
\end{figure}

\else
\begin{figure}[h]
    \centering
    \begin{minted}[frame=single, framesep=10pt, linenos]{python}
def augment_transitions_with_advantages(episodes_buffer, config):
    transitions = []
    for episode in episodes_buffer:
        # Augment each (s, a, r) transition with MC estimate of return to go.
        rewards = [transition["reward"] for transition in episode]
        cur = 0.0
        for i in reversed(range(len(rewards))):
            cur = rewards[i] + config.gamma * cur
            episode[i]["return"] = cur
        transitions.extend(episode)

    # return batch normalization 
    if config.use_rebn:
        returns = torch.tensor([transition["return"] for transition in transitions]).float()
        returns = (returns - returns.mean()) / (returns.std() + 1e-9)
        for i, transition in enumerate(transitions):
            transition["return"] = returns[i].item()

    # subsample to make a constant batch size
    if len(transitions) > config.batch_size:
          transitions = np.random.choice(transitions, config.batch_size, replace=False)

    return transitions
    \end{minted}
    \caption{Code to compute advantages}
        \label{code:advantages}
\end{figure}

\fi

In \cref{code:advantages}
we show code to augment each transition with an estimate
of its advantage. To do this, we first
compute the return-to-go
\be
G_t^e = r_t^e + \gamma G_{t+1}^e
\ee
where $G_T^e=r_T^e$ is the initial condition,
where $T$ is the first step for which done=1.
We then use the return-batch normalization (ReBN)
method of  \citep{Liu2025gem}
--- which is a multi-turn
variant of GRPO ---
to convert these into advantages, using
\be
A_t^e = \frac{G_t^e -\mu}{\sigma}
\ee
where
$\mu = \text{mean}(\{ G_{t}^e: t=1:T_e, e=1:E \})$,
and
$\sigma = \text{std}(\{ G_{t}^e: t=1:T_e, e=1:E \})$.
We then flatten the data to get a list of
$N=E \times T$
transitions $(\vs^n,\va^n,r^n,A^n)$.

\subsection{Computing token level loss}

\ifarxiv
\begin{figure}[h]
  \centering
  \includegraphics[height=3in]{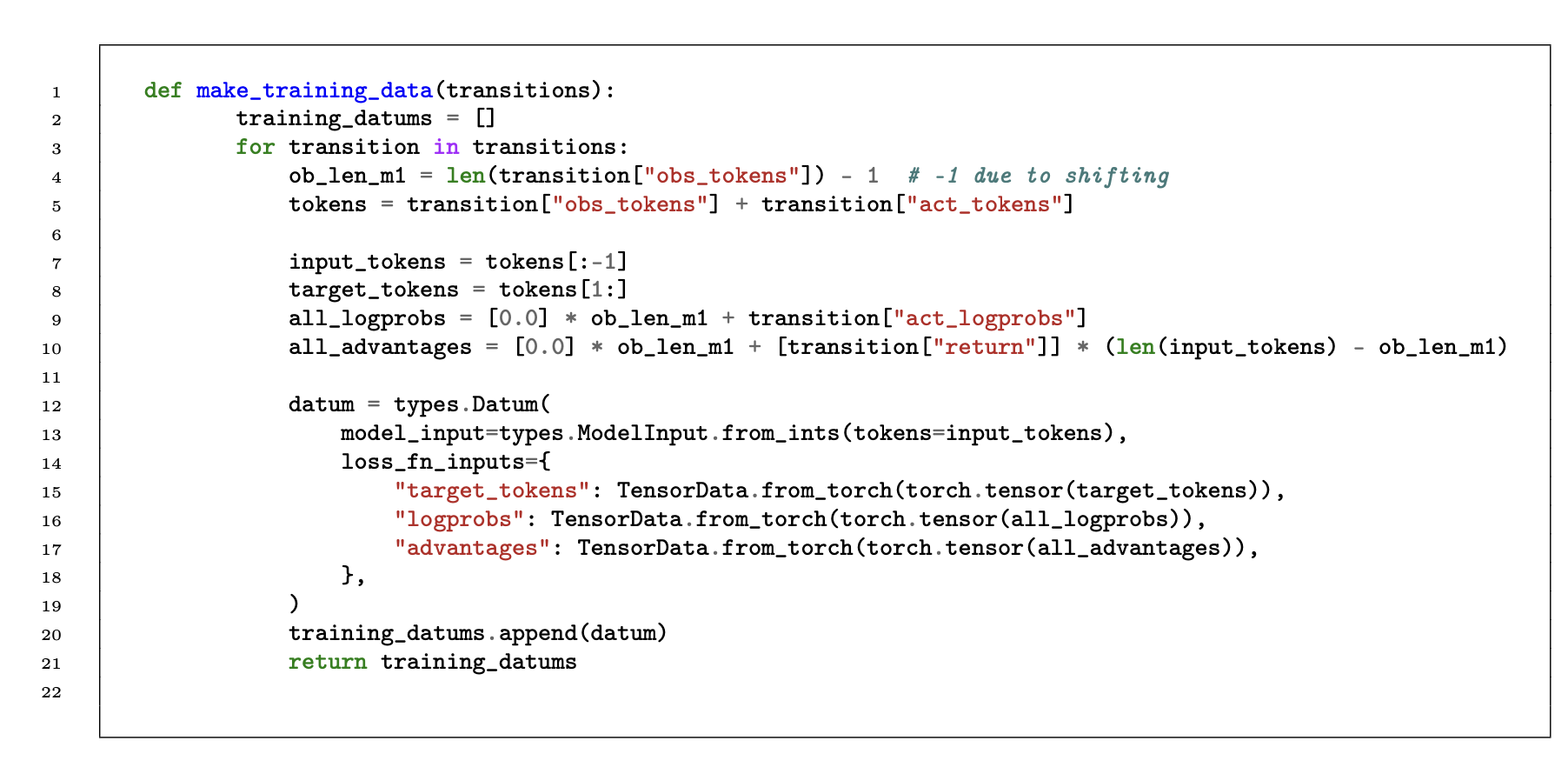}
    \caption{Code to compute data for token level loss computation.}
    \label{code:convert}
\end{figure}

\else
\begin{figure}[h]
    \centering
    \begin{minted}[frame=single, framesep=10pt, linenos]{python}
 def make_training_data(transitions):
        training_datums = []
        for transition in transitions:
            ob_len_m1 = len(transition["obs_tokens"]) - 1  # -1 due to shifting
            tokens = transition["obs_tokens"] + transition["act_tokens"]

            input_tokens = tokens[:-1]
            target_tokens = tokens[1:]
            all_logprobs = [0.0] * ob_len_m1 + transition["act_logprobs"]
            all_advantages = [0.0] * ob_len_m1 + [transition["return"]] * (len(input_tokens) - ob_len_m1)
           
            datum = types.Datum(
                model_input=types.ModelInput.from_ints(tokens=input_tokens),
                loss_fn_inputs={
                    "target_tokens": TensorData.from_torch(torch.tensor(target_tokens)),
                    "logprobs": TensorData.from_torch(torch.tensor(all_logprobs)),
                    "advantages": TensorData.from_torch(torch.tensor(all_advantages)),
                },
            )
            training_datums.append(datum)
            return training_datums

    \end{minted}
    \caption{Code to convert labeled transitions into training data.}
        \label{code:convert}
\end{figure}

\fi

Finally,
 \cref{code:convert} shows the code
 to convert the set of transitions
 into a set of ``training datums'',
each of which contains
the input tokens $\vx^n=(\vs^n,\va^n[:-1])$,
the target tokens $\vy^n=(\vs^n[1:],\va^n)$,
the  masked sampling log probabilities $\log q(x_k^n)$
and the masked token advantages $A_k^n$, which we set
to be equal to the turn-level advantages $A^n$,
derived from the turn-level rewards $R^n$.
(The masking ensures the state tokens contribute 0 to the loss.)

We can optionally add a KL penalty to the policy
to ensure it does  not deviate too far from the base model.
However, it is not correct to add this to the advantage
itself, as pointed out in \citep{Zhang2025KL,Tang2025pitfalls}.
Instead, we modify the token-level reward to be
\be
R_k^n = R^n + \beta
\KLpq
{\pi_{\vtheta}(a^n_k|\va^n_{1:k-1}, \vs^n)}
{\pi_{0}(a^n_k|\va^n_{1:k-1}, \vs^n)}
\ee
where $\pi_0$ is the reference prior,
and $R^n$ is the reward from the environment after transition $n$.

\subsection{Computing metrics related to training stability}

\ifarxiv
\begin{figure}[h]
  \centering
  \includegraphics[height=3in]{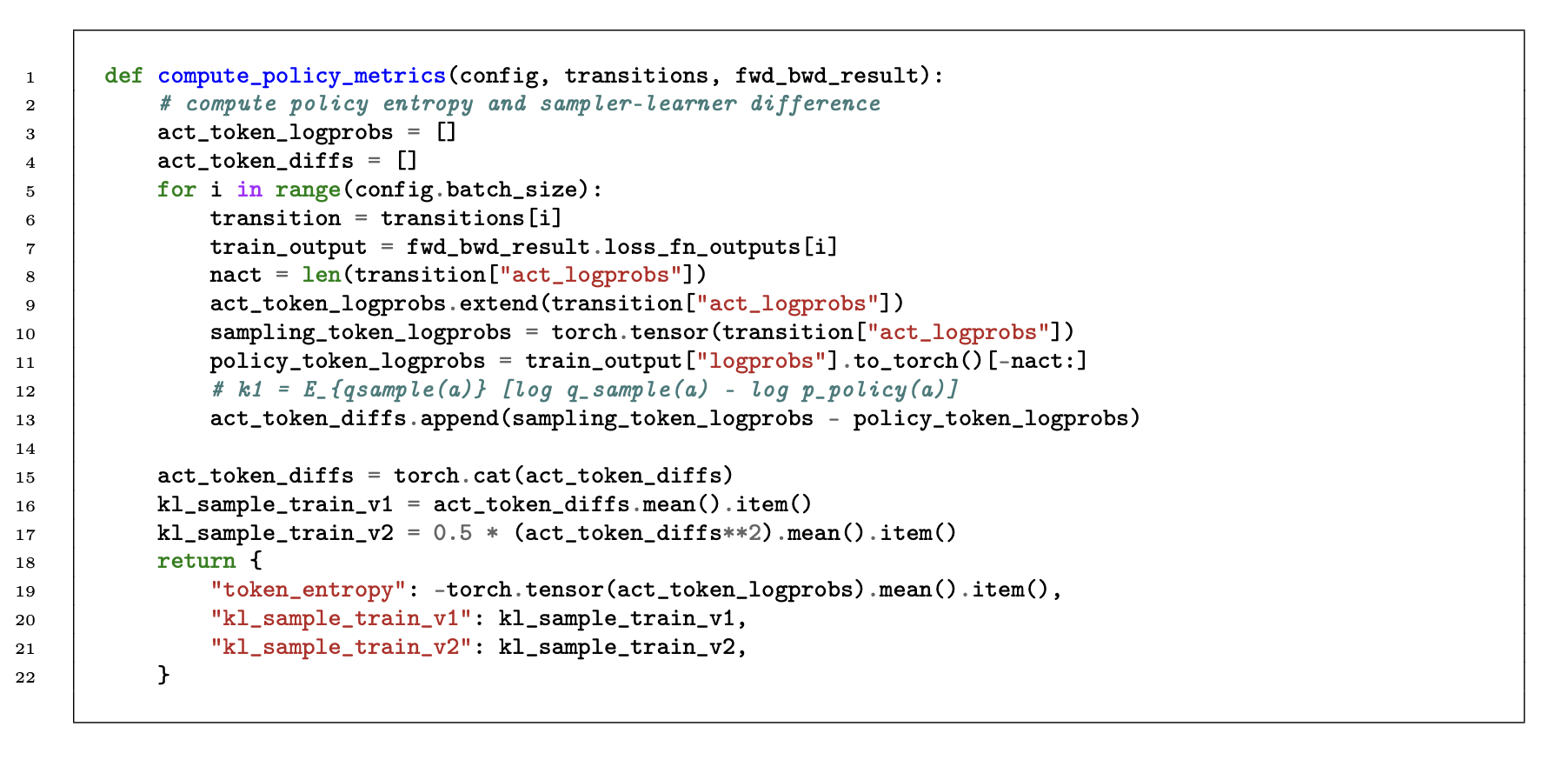}
    \caption{Code to compute metrics.}
    \label{code:metrics}
\end{figure}

\else
\begin{figure}[h]
    \centering
        \begin{minted}[frame=single, framesep=10pt, linenos]{python}
def compute_policy_metrics(config, transitions, fwd_bwd_result):
    # compute policy entropy and sampler-learner difference
    act_token_logprobs = []
    act_token_diffs = []
    for i in range(config.batch_size):
        transition = transitions[i]
        train_output = fwd_bwd_result.loss_fn_outputs[i]
        nact = len(transition["act_logprobs"])
        act_token_logprobs.extend(transition["act_logprobs"])
        sampling_token_logprobs = torch.tensor(transition["act_logprobs"])
        policy_token_logprobs = train_output["logprobs"].to_torch()[-nact:]
        # k1 = E_{qsample(a)} [log q_sample(a) - log p_policy(a)]
        act_token_diffs.append(sampling_token_logprobs - policy_token_logprobs)
   
    act_token_diffs = torch.cat(act_token_diffs)
    kl_sample_train_v1 = act_token_diffs.mean().item()
    kl_sample_train_v2 = 0.5 * (act_token_diffs**2).mean().item()
    return {
        "token_entropy": -torch.tensor(act_token_logprobs).mean().item(),
        "kl_sample_train_v1": kl_sample_train_v1,
        "kl_sample_train_v2": kl_sample_train_v2,
    }
    \end{minted}
    \caption{Code to compute metrics related to policy health.}
        \label{code:metrics}
\end{figure}

\fi

In practice it is important to log various metrics,
to monitor the training process.
For example, we might want to compute $\KLpq{q}{p}$,
where $q$ is the sampling distribution (used to rollout the epsiodes),
and $p$ is the training distribution (used to compute gradients).
We can estimate the KL using
$k_1$ estimator from \cref{eqn:k1}
or $k_2$ estimator from \cref{eqn:k2}.
If the KL exceeds 0.01, it means that learning
is  very off-policy, and results might be unstable.\footnote{
This heuristic is from
\url{https://tinker-docs.thinkingmachines.ai/rl/rl-hyperparams}.
}
In addition, we can compute the entropy of the policy.
We want to ensure this is initially not too small,
to enable exploration, but that it 
 does not blow up over time.
See \cref{code:metrics} for some code to compute these metrics.

\subsection{Example}
\label{sec:tinkerExample}

\begin{figure}
  \centering
  \begin{tabular}{cc}
    \includegraphics[height=1.5in]{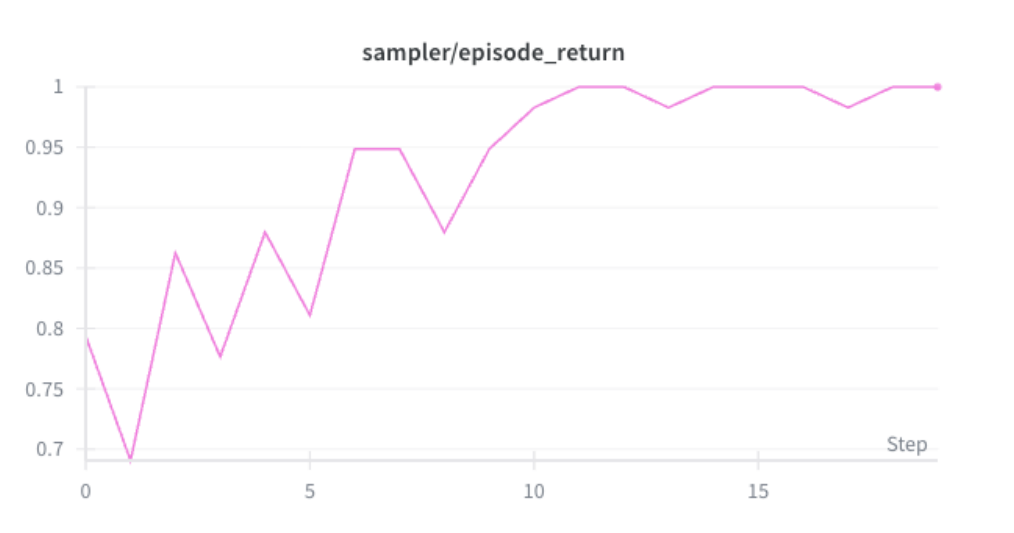}
    &
    \includegraphics[height=1.5in]{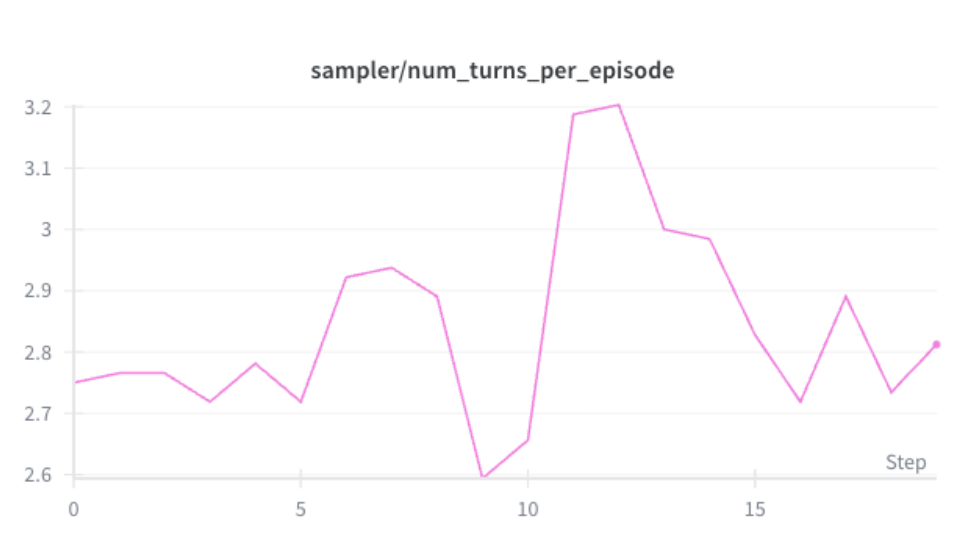}
    \\
    (a) & (b) \\
        \includegraphics[height=1.5in]{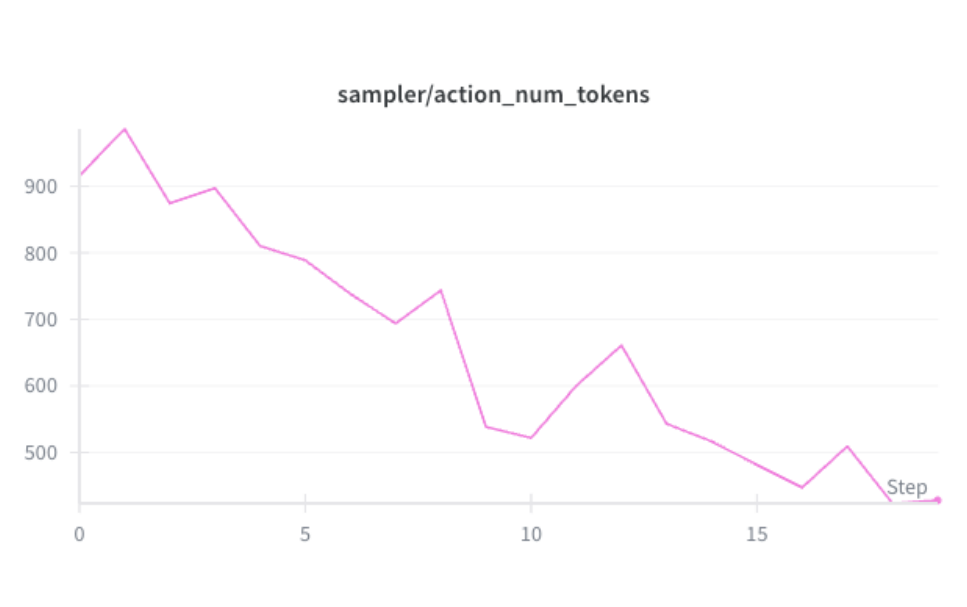}
    &
        \includegraphics[height=1.5in]{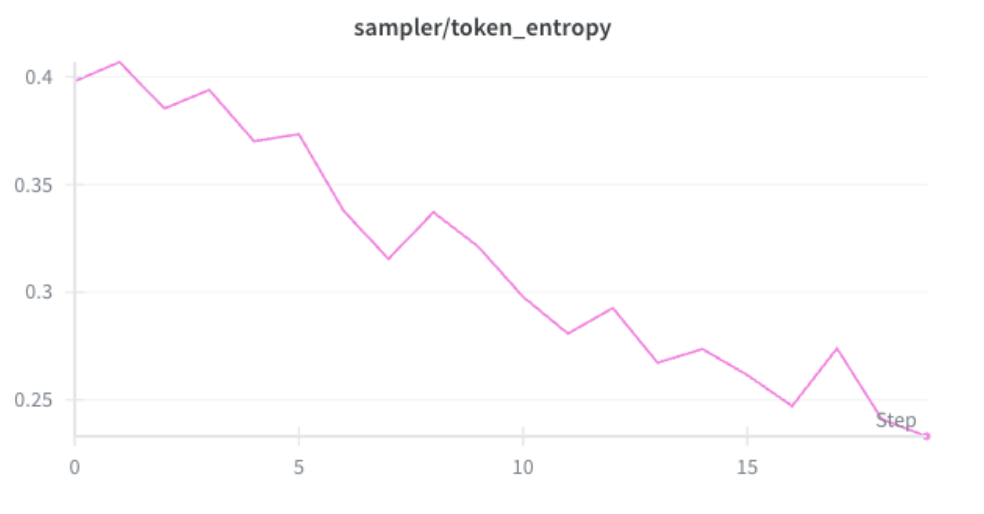}
        \\
        (c) & (d)
\end{tabular}
\caption{
  Results of RL on the guess-the-number  game from GEM.
  }
\label{fig:tinker-results}
\end{figure}

In this section, we give a brief example.
We consider the 
\verb!GuessTheNumber-v0-easy! game,
in which the agent must guess the secret number
$n_e$ 
randomly chosen by the environment at the start of each episode
$e$.
In the easy version, the number is between
$n_{\tmin}=1$ and $n_{\tmax}=10$,
and the max number of turns is $T=4$.
If the agent guesses correctly in 4 or fewer turns,
it gets a success reward of 1.
If it runs out of turns, it gets a reward
of $1-d/(n_\tmax - n_\tmin)$,
where $d=|n^e - a_T^e|$ is the distance
between the true answer and the final guess.
If it generates an action sequence $\va_t$
that cannot be parsed in to a valid numeric action,
the agent gets a format penalty of -0.1.
For details, see the code.\footnote{
\url{https://github.com/axon-rl/gem/blob/main/gem/envs/game\_env/guess\_the\_number.py}.
  }

The optimal strategy is to perform binary search.
So for the easy version of the game, the expected
number of turns is $\log_2(10) =  3.32$.
We fine-tune a Qwen3-4B-Instruct-2507 model
using $20$ steps of ReBN,
rolling out $E=32$ episodes per step.
We use LoRA of rank 8, a learning rate of $4 \times 10^{-5}$,
and a discount factor of $\gamma=0.9$.
Further details on the hyper-parameters
can be found in the code.
Training takes about 20 minutes on a laptop,
using the standard Tinker client.

In \cref{fig:tinker-results}(a), we plot the average reward 
vs training steps,
and in \cref{fig:tinker-results}(c),
we plot the average number of action tokens per turn
vs training steps.
We see that initially the model
gets an average reward of 0.8, but
that the resulting action sequences are quite verbose
(about 900 tokens).
At the end of training, we see that the average
reward has improved to 1.0,
and the answer length has reduced to about 500 tokens
per step.
In \cref{fig:tinker-results}(b),
we see that the average number of turns is always around 3,
since the initial base model already ``knows'' that the optimal
strategy is binary search.
(But by using $\gamma < 1$, we discourage taking too many turns;
see \cref{fig:binarySearch} for an example.)
Finally in  \cref{fig:tinker-results}(d),
we see the entropy of the policy reduces over time,
as the model specializes to this task.

\begin{figure}
  \footnotesize
\begin{verbatim}
STATE: You are playing Guess The Number. You have to guess the number between 1 and 50
(inclusive) within 50 turns. As you play, the history of your guesses will be appended below.
Use the information to complete the game before you run out of guesses.
Enter your first guess to start the game.
At turn 1, you guessed 25, and the target number is lower than 25.
At turn 2, you guessed 12, and the target number is higher than 12.
At turn 3, you guessed 18, and the target number is higher than 18.
At turn 4, you guessed 21, and the target number is higher than 21.
At turn 5, you guessed 23, and the target number is lower than 23.
Enter your next guess.

ACTION: To determine the target number between 1 and 50, we can use a binary search strategy,
which is effective for narrowing down a range when given hints like higher or lower.
Step-by-Step Reasoning:
1. **Initial Guess (Turn 1):** - The target number is lower than 25. - This means the target number
is in the range [1, 24].
2. **Second Guess (Turn 2):** - The target number is higher than 12. - This means the target
number is in the range [13, 24].
3. **Third Guess (Turn 3):** - The target number is higher than 18. - This means the target number
is in the range [19, 24].
4. **Fourth Guess (Turn 4):** - The target number is higher than 21. - This means the target
number is in the range [22, 24].
5. **Fifth Guess (Turn 5):** - The target number is lower than 23. - This means the target number
is in the range [22, 22].
Final Guess:
The target number is in the range [22, 22], which means the target number is 22. Final Answer:
\boxed{22}
\end{verbatim}
\caption{Example of a state and action at the end of turn 5
  of guess-the-number
  after training the policy with $\gamma=0.9$.
}
\label{fig:binarySearch}
\end{figure}

\eat{
\subsection{Other RL systems}

\begin{figure}
\centering
\includegraphics[height=1.2in]{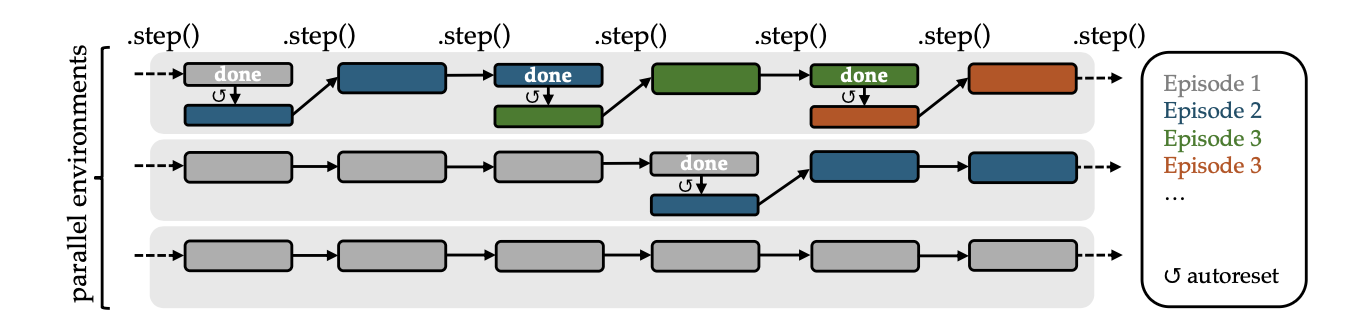}
\caption{
  Illustration of parallel rollouts,
  using an auto-reset mechanism to cope with
  trajectories of different lengths.
  Figure from \citep{Liu2025gem}.
  }
\label{fig:gem-autoreset}
\end{figure}

The Tinker code we presented above is simple,
but can be made more efficient in many ways.
For example, when performing  rollouts
of episodes with different speeds or durations,
we need to support  asynchronous computation,
as illustrated in \cref{fig:gem-autoreset}.
(This is especially important for training agents that use tools
as actions, since these tools
 may be slow, and they often need to be run in an
 external sandbox, for security reasons.)
 In addition, it is more efficient to collect
 data in parallel to training the model,
 rather than in the block sequential way we presented.
For some references which discuss more scalable RL infra,
see e.g.,
\keywordDef{SkyRL} \citep{skyrlv0},
 \keywordDef{LlamaRL} \citep{llamaRL},
 \keywordDef{VERL} \citep{VERL},
 \keywordDef{PrimeIntellect} \citep{Prime},
 \keywordDef{OpenRLHF} \citep{OpenRLHF},
 etc.

}

\chapter{Other topics in RL}
\label{sec:rl-other}

In this section, we briefly mention some other important topics in RL.

\section{Regret minimization}
\label{sec:regret}

In this book, we have assumed the environment is a stochastic process
(e.g., an MDP), and the agent is trying
to maximize its expected future utility  (i.e., minimize
its risk) given the data
it has seen so far in the past,
as is standard in the Bayesian approach to decision making.
(If there are multiple agents, we must add them to the model,
as in \cref{sec:MARL}.)

An alternative approach is to assume the agent
encounters an arbitrary stream of observations,
and it must choose actions that are as close
as possible to what  an optimal policy would have achieved,
even if the sequence is chosen by an unknown  \keywordDef{adversary}.
Algorithms that achieve this goal are said to
be \keywordDef{regret minimization} algorithms.
We give more details below.

\subsection{Regret for static MDPs}


In online learning, we assume the decision at each step is to choose the parameters
$\vtheta_t$ such that we minimize the loss at that step,
such as $\ell_t(\vtheta) = (y_t - f(\vx_t,\vtheta))^2$.
Let $\vtheta_t=\pi(\vx_{1:t},y_{1:t-1})$ be the policy
for choosing parameters.
We then define the \keywordDef{regret} to be
\be
\text{Regret}_T(\pi)
= \left[\sum_{t=1}^T \ell_t(\vtheta_t) - \ell_t(\vtheta^*) \right]
\ee
where
\be
\vtheta^* = 
\arg\min_{\vtheta \in \Theta} \sum_{t=1}^T \ell_t(\vtheta)
\ee
See e.g., \citep{Hazan2019opt} for details.

In the case of the MDP, 
the \keyword{regret} of a policy $\pi$
as the difference
between its expected return
and the expected return of an optimal policy $\pi^*$:\footnote{
We can also define the regret in terms of value functions:
$\text{Regret}_T(\pi|M,\Pi)
=  \expectQ{V_T^{\pi^*}(s_0|M) - V_T^{\pi}(s_0|M)}{s_0 \sim M}$,
where  $V^{\pi}(s|M)$ refers to the value of policy $\pi$
starting from state $s$ in MDP $M$.
}
\be
\text{Regret}_T(\pi;M, \Pi)
=  \expectQ{\sum_{t=1}^T \left( r_t(s_t,a^*_t) - r_t(s_t,a_t) \right)}
{s_t \sim M(\cdot|s_{t-1},a_t),
  a_t \sim \pi(\cdot|s_t), a_t^* \sim \pi^*(\cdot|s_t)}
\label{eqn:regret}
\ee
where
\be
\pi^* =  \argmax_{\pi \in \Pi} \expectQ{V^{\pi}(s_0|M)}{s_0 \sim M}
\label{eqn:pistar}
\ee
is the optimal policy from some policy class $\Pi$
which has access to the true MDP $M$.
This is often referred to as the
\keywordDef{best policy in hindsight}, since
if we average over enough sequences (or over enough time steps),
the policy we wish we had chosen
will of course be the optimal policy that knows the true environment.

Since the true MDP is usually unknown, we can define
the \keywordDef{maximum regret} of a policy as
its worst case regret wrt some class of models $\calM$:
\be
\text{MaxRegret}_T(\pi;\calM,\Pi)
= \max_{M \in \calM} \text{Regret}_T(\pi|M,\Pi)
\ee
We can then define the \keywordDef{minimax optimal policy}
as the one that minimizes the maximum regret:\footnote{
We can also consider a non-stochastic setting,
in which we allow an adversary to choose the state sequence,
rather than taking expectations over them.
In this case, we take the maximum over individual sequences
rather than models.
For details, see \citep{Hazan2022control}.
}
\be
\pi^*_{MM}(\calM,\Pi) =
\argmin_{\pi \in \Pi} \max_{M \in \calM} \text{Regret}_T(\pi|\pi^*(M), M)
\ee

\eat{
We define the
\keywordDef{regret} of an agent $\calA$ (i.e., a mechanism
for choosing a sequence of policies,
given an initial policy $\pi_0$)
as the difference
between its expected return
and the optimal policy $\pi^*$, which
has access to the true MDP $\calM^*:
Specifically, let $\policy_t$ be the agent's policy at time $t$
(this can change over time, as the agent learns).
The regret
is defined as follows:
\be
\text{Regret}_T(\calA|\pi^*)
=  \expectQ{\sum_{t=1}^T \left( V^{\pi^*}(s_t)
  - V^{\pi_t}(s_t) \right)}{s_t \sim M(\cdot|s_{t-1},a_t),  a_t \sim \pi_t(s_t)}
\ee
where $\pi_t$ is the policy used at step $t$,
computed
using $\pi_t=\calA(s_{1:t-1},a_{1:t-1},r_{1:t-1})$.
(Note that the policy is a random variable.)
The expectation is over all possible sequences
given $\calM^*$ and $\calA$.
Since $\calM^*$ is not known, we often consider the worst
case over all MDPs, giving rise to the \keywordDef{minimax regret}:
\be
\calA^*_{\text{minimax}} = \argmin_{\calA} \max_{\calM^*}
\text{Regret}_T(\calA|\calM^*)
\label{eqn:minimax}
\ee
}

The main quantity of interest in the theoretical RL literature
is how fast the regret grows as a function of time $T$.
In the case of a tabular episodic MDP, the  optimal minimax regret
is $O(\sqrt{H S A T})$ (ignoring logarithmic factors),
where $H$ is the horizon length (number of steps per episode),
$S$ is the number of states,
$A$ is the number of actions,
and
$T$ is the total number of steps.
When using parametric functions to define the MDP,
the bounds depend on the complexity
of the function class.
For details, see e.g., \citep{Auer2008,Jaksch2010,Lattimore2019}.

\subsection{Regret for non-stationary MDPs}

When the world can change, there may be no single optimal policy $\pi^*$
we can compare to.
Instead, the \keywordDef{dynamic regret}
(aka \keywordDef{adaptive regret})
compares to a sequence of optimal policies:
\be
\text{DynamicRegret}_T(\pi_{1:T}| M_{1:T},\Pi)
=  \expectQ{\sum_{t=1}^T \left( r(s_t,a^*_t) - r(s_t,a_t) \right)}
{s_t \sim M_t,  a_t \sim \pi_t, a_t^* \sim \pi_t^*}
\ee
where
\be
\pi_t^* = \argmax_{\pi \in \Pi} V^{\pi}(s_t|M_t)
\ee
is the optimal policy at that moment in time.

To compute bounds on the optimal dynamic regret,
we need to make assumptions about how often the world changes,
and by how much. This is called a \keywordDef{variational budget}.
This is defined as
\be
\text{VB}_T = \sum_{t=2}^{T}
\text{dist}(\calM_{t}, \calM_{t-1})
\ee
where the distance function measures
the similarity of the MDPs at adjacent
episodes (e.g., $\ell_1$ distance of the reward
and transition functions).
The optimal dynamic regret can then be bounded
in terms of VB 
(see e.g., \citep{Cheung2023,Auer2019}.

\eat{
However, if we assume the environment is piecewise stationary,
we can use the concept of  \keywordDef{dynamic regret}
(aka \keywordDef{adaptive regret})
which is defined as follows:
\be
\text{DynamicRegret}_K(\calA|\calM^*)
= \expect{ \sum_{k=1}^K \left( V_1^{\pi^*_k}(s_1^k)
  -  V_1^{\pi_k}(s_1^k) \right) }
  \ee
Here  $\pi_k^*$ is computed using the true MDP
for episode $k$ (a dynamic comparator class),
and $\pi_k=\calA(s^{1:k-1},a^{1:k-1},r^{1:k-1})$ is computed using the learning
algorithm applied to all data from previous episodes.

To compute upper bounds on the optimal dynamic regret,
we need to make assumptions about how often the world changes,
and by how much. This is called a \keywordDef{variational budget}.
For episodic MDPs, we define
\be
\text{VB}_K = \sum_{k=1}^{K-1}
\text{dist}(\calM_{k}, \calM_{k+1})
\ee
where the distance function measures
the similarity of the MDPs at adjacent
episodes (e.g., $\ell_1$ distance of the reward
and transition functions).
The optimal dynamic regret can then be bounded by
$O(\sqrt{K H  A(1 + \text{VB}_K)})$
(see e.g., \citep{Cheung2023}).
Methods for the case in which the changepoints between
the different models
are not assumed known (but instead are detected automatically
from data) 
can be found in \citep{Auer2019}.

In the general case, where the MDP can change
gradually,
it is more difficult to define the dynamic regret,
since it is not clear what the optimal policy should be.
Instead, we 
often compare to the
\keywordDef{best-in-hindsight} policy
$\pi^*(s_{1:T})$, which is optimized {\em after} seeing all the data.
(We omit $r_{1:T}$ from the notation for brevity.)
In particular, we define the regret as
\be
\text{OnlineRegret}(\calA|s_{1:T}) =
\max_{\pi  \in \Pi(s_{1:T})}
  \sum_{t=1}^T  V^{\pi}(s_t)  -  V^{\pi_t}(s_t)
  \ee
  where $\pi_t=\calA(s_{1:t-1},a_{1:t-1},r_{1:t-1})
  $ is the policy used at step $t$,
  and $\pi$ is the best policy chosen
  in hindsight.
Note that there is no expectation in the above definition,
since it is for a single sequence.
Instead, we can allow the adversary to choose the noise
sequence, which is equivalent to letting it choose the state sequence
$s_{1:T}$.
Thus the minimax optimal algorithm for online regret
is given by
\begin{align}
\calA^*_{\text{online-minimax}}
&= \argmin_{\calA} \max_{s_{1:T}}
\text{OnlineRegret}_T(\calA|s_{1:T}) \\
&= \argmin_{\calA} \max_{s_{1:T}}
\max_{\pi  \in \Pi(s_{1:T})}
  \sum_{t=1}^T  V^{\pi}(s_t)  -  V^{\pi_t}(s_t)
\end{align}
This is called the \keywordDef{online non-stochastic control}
problem \citep{Hazan2022control}.
The difficulty of solving this inner optimization problem
depends on the expressiveness of the policy class $\Pi$.
If $\pi$ is fixed across the sequence,
we get a weaker notion of regret
compared to allowing $\pi$ to change multiple times.

}

\subsection{Minimizing regret vs maximizing expected utility}
\label{sec:regretVsBayes}

\begin{table}[h!]
\centering
\begin{tabular}{|p{3.5cm}|p{5.5cm}|p{5.5cm}|}
\hline
\textbf{Aspect} & \textbf{Bayes-Optimal (BAMDP)} & \textbf{Regret-Minimizing (Minimax)} \\
\hline
Knowledge & Requires a known prior over MDPs & No prior; judged against best policy in hindsight \\
\hline
Objective & Maximize expected return under the prior & Minimize regret w.r.t.\ optimal policy in true MDP \\
\hline
Exploration & Performs optimal Bayesian exploration & Often uses optimism or randomness (e.g., UCB, TS) \\
\hline
Adaptation & Fully adaptive via posterior updates & May use confidence bounds, resets, or pessimism \\
\hline
Setting & Bayesian RL & Frequentist or adversarial RL \\
\hline
\end{tabular}
\caption{Key differences between Bayes-optimal and regret-minimizing policies in RL.}
\label{tab:BayesVsRegret}
\end{table}

The Bayes optimal agent is the one that maximizes
its expected utility (minimizes its risk),
where we take expectations not only
over the sequence of observations and rewards,
but also over the unknown environment $M$ itself,
rather than assuming it is known
\citep{Ghavamzadeh2015}.
 That is,
 the optimal learning algorithm $\calA$ is given by
\begin{align}
  \calA^*_{\text{Bayes}}(P_0)  &= \argmax_{\calA} U_T(\calA|P_0)\\
U_T(\calA|P_0) &= \expectQ{
  \expectQ{\sum_{t=1}^T r(s_t,a_t)}
          {s_t \sim \calM, a_t \sim \pi_t,
            \pi_t = \calA(s_{1:t-1},a_{1:t-1},r_{1:t-1})}
}{\calM \sim P_0(\calM)}
\end{align}
where $P_0(\calM)$ is our prior over models,
and $\calA$ is our learning algorithm
that generates the policy (decision procedure) to use at each step,
as discussed in \cref{sec:universal}.
Note that
the uncertainty over models automatically encourages
the optimal amount of exploration,
as we discuss in \cref{sec:BAMDP}.
Note also that
if we can do exact inference, the optimal algorithm
is uniquely determined by the prior $P_0$.

By contrast, the regret minimizing policy is the one that minimizes
the maximum regret
\be
\calA^*_{MM}(\calM,\Pi) =
\argmin_{\pi \in \Pi} \max_{M \in \calM} \text{Regret}_T(\calA| M,\Pi)
\ee
where we define the regret of a learning algorithm as
\be
\text{Regret}_T(\calA|M,\Pi)
=  \expectQ{\sum_{t=1}^T \left( r(s_t,a^*_t) - r(s_t,a_t) \right)}
{s_t \sim M,
  a_t \sim \pi_t,
  \pi_t = \calA(s_{1:t-1},a_{1:t-1},r_{1:t-1}),
   a_t^* \sim \pi^*
  }
\ee
where $\pi^*$ is given in \cref{eqn:pistar}.
Unlike the Bayesian case, we must now
manually design the algorithm
to solve the exploration-exploitation problem
(e.g., 
using the Thompson sampling method of  \cref{sec:thompson}
or the UCB method of \cref{sec:UCB}),
i.e., there is no automatic solution to the problem.

This distinction between minimizing risk and minimizing regret
is equivalent to the standard
difference between Bayesian and frequentist
approaches to decision making
(see e.g., \citep[Sec 34.1]{book2}).
The advantage of the Bayesian approach is that it can use
prior knowledge (e.g., based on experience with other tasks,
or knowledge from an LLM) to adapt quickly to changes,
and to make predictions about the future,
allowing for optimal long-range planning.
The advantage of the regret-minimizing approach is that
it avoids the need for a specifying prior over models,
it can be robustly adapt to unmodeled changes,
and it can handle adversarial (non-stochastic) noise.
See \cref{tab:BayesVsRegret} for a summary,
and 
\citep{Huggins2015} for more discussion.

\section{Exploration-exploitation tradeoff}
\label{sec:exploreExploit2}
\label{sec:exploreExploitRevisited}

In this section, we discuss solutions to
the exploration-exploitation tradeoff
that go beyond the simple heuristics introduced in 
\cref{sec:exploreExploit}.

\subsection{Optimal (Bayesian) approach}

We can compute an optimal solution to the exploration-exploitation
tradeoff by adopting a Bayesian approach to the problem,
where we augment the state space
with our beliefs about the underlying model, as discussed in
\cref{sec:beliefStateMDP}.

\subsubsection{Bandit case (Gittins indices)}
\label{sec:gittins}

\eat{
Suppose we have a way to compute the recursively compute
the belief state over model parameters,
$p(\vtheta_t|\data_{1:t})$.
How do we use this to solve for the policy
in the resulting belief state MDP?
}

In the special case of context-free bandits with a finite number of arms,
the optimal policy of this belief state MDP can be computed
using dynamic programming.
To explain this, we follow the presentation
of \citep[Sec 15.5]{KochenderferDecisions},
and consider a Bernoulli bandit with $n$ arms.
Let the belief state be denoted by $b=(w_1,l_1,\ldots,w_n,l_n)$,
where $w_a$ is the number of times arm $a$ has won (given reward 1)
and $l_a$ is the number of times arm $a$ has lost (given reward 0).
Using Bellman's equation, and the expression
for the probability of winning under 
a beta-Bernoulli distribution with a uniform
prior, we have
\begin{align}
  V^*(b) &= \max_a Q^*(b,a) \\
  Q^*(b,a) &=
  \frac{w_a+1}{w_a + l_a + 2}(1 + V^*(\cdots,w_a+1,l_a,\cdots) \\
 & + \left(1-\frac{w_a+1}{w_a + l_a + 2}\right)
  V^*(\cdots,w_a,l_a+1,\cdots)
  \end{align}
In the finite horizon case, with $h$ steps,
We can compute $Q^*$ using dynamic programming.
We start with terminal belief states $b$  with $\sum_a (w_a+l_a)=h$,
where $V^*(b)=0$.
We then work backwards to states $b$ satisfying
$\sum_a (w_a+l_a)=h-1$,
and then applying the above equation recursively
until time step 0.

Unfortunately, although this process is optimal,
the number of belief states is $O(h^{2n})$,
rendering it intractable.
Fortunately, for the infinite horizon discounted case,
the problem can be solved efficiently
using  \keywordDef{Gittins indices}
\citep{Gittins89}
(see \citep{Powell2012,Powell2022} for details).
However,  these optimal methods
do not extend to contextual bandits,
where the problem is provably  intractable
\citep{Papadimitriou87}.

\subsubsection{MDP case (Bayes Adaptive MDPs)}
\label{sec:BAMDP}

We can extend the above techniques to the MDP case
by constructing a \keywordDef{BAMDP},
which stands for ``Bayes-Adaptive MDP''
\citep{Duff02}.
The basic idea is quite simple.
We define an MDP with an augmented state-space,
consisting of the original state $s$ and the belief state
$b$, representing a distribution over the model parameters.
The transition function is given by
\be
\calT'(s',b'|s,b,a) = \delta(b'=BU(s,b,a,s')) P(s'|s,b,a)
\ee
where BU is the (deterministic) Bayes updating procedure
(e.g., incrementing the pseudo counts of the Dirichlet
distribution, in the case of a discrete MDP),
and the second term is the posterior predictive
distribution over states:
\be
P(s'|s,b,a) = \int b(\vtheta) \calT(s'|s,a;\vtheta)
d\vtheta
\ee
The rewards and actions of the augmented
MDP are the same as the base MDP. Thus Bellman's
equation gives us
\be
V^*(s,b) = \max_a
\left( R(s,a) + \gamma \sum_{s'} P(s'|s,b,a)
V^*(s', BU(s,b,a,s')) \right)
\ee
Unfortunately, this is computationally intractable to solve.
Fortunately,
 various approximations have been proposed
(see e.g., \citep{Zintgraf2021,Arumugam2022bamdp,Mikulik2020}).

 \subsection{Thompson sampling}
\label{sec:thompson}

The fully Bayesian approach is computationally intractable.
A common approximation is to use
 \keywordDef{Thompson sampling}
\citep{Thompson1933},
also called  \keywordDef{probability matching} \citep{Scott10}.
We start by describing this in the bandit case,
then extend to the MDP case.
For more details, see \citep{Russo2018}.
(See also \citep{Gershman2018} for some evidence that humans use
Thompson-sampling like mechanisms.)

\subsubsection{Bandit case}

In Thompson sampling,
we define the policy at step $t$ to be
$\policy_t(a|s_t,\history_t) = p_a$,
where $p_a$ is  the probability that $a$ is the optimal action.
This can be computed using
\begin{align}
p_a &= \Pr(a=a_*|s_t,\history_t)
  = \int \ind{a = \argmax_{a'} R(s_t,a';\vtheta)}
  p(\vtheta|\history_t) d\vtheta
  \label{eqn:bandit-probmatching}
\end{align}
If the posterior is uncertain, the agent will sample many different
actions, automatically resulting in exploration.
As the uncertainty decreases, it will start to exploit its knowledge.

To see how we can implement this method,
note that 
we can compute the expression
in \cref{eqn:bandit-probmatching} 
by using a single
Monte Carlo sample
$\tilde{\vtheta}_t \sim p(\vtheta|\history_t)$.
We then plug in this parameter
into our reward model, and greedily pick the best action:
\be
a_t = \argmax_{a'} R(s_t, a';\tilde{\vtheta}_t)
\ee
This sample-then-exploit approach
will choose actions with exactly
the desired probability, since
\begin{align}
  p_a
= \int \ind{a=\argmax_{a'} R(s_t,a';\tilde{\vtheta}_t)} p(\tilde{\vtheta}_t|\history_t) 
 = \Pr_{\tilde{\vtheta}_t \sim p(\vtheta|\history_t)}(a=\argmax_{a'} R(s_t, a';\tilde{\vtheta}_t)) 
\end{align}

Despite its simplicity,
this approach can be shown to
achieve optimal  regret (see e.g., \citep{Russo2018} for
a survey).
In addition, it is very easy to implement,
and hence is widely used  in practice~\citep{Graepel10,Scott10,Chapelle11}.

\begin{figure}
\centering
\begin{subfigure}[b]{0.3\textwidth}
\centering
\includegraphics[height=1.5in]{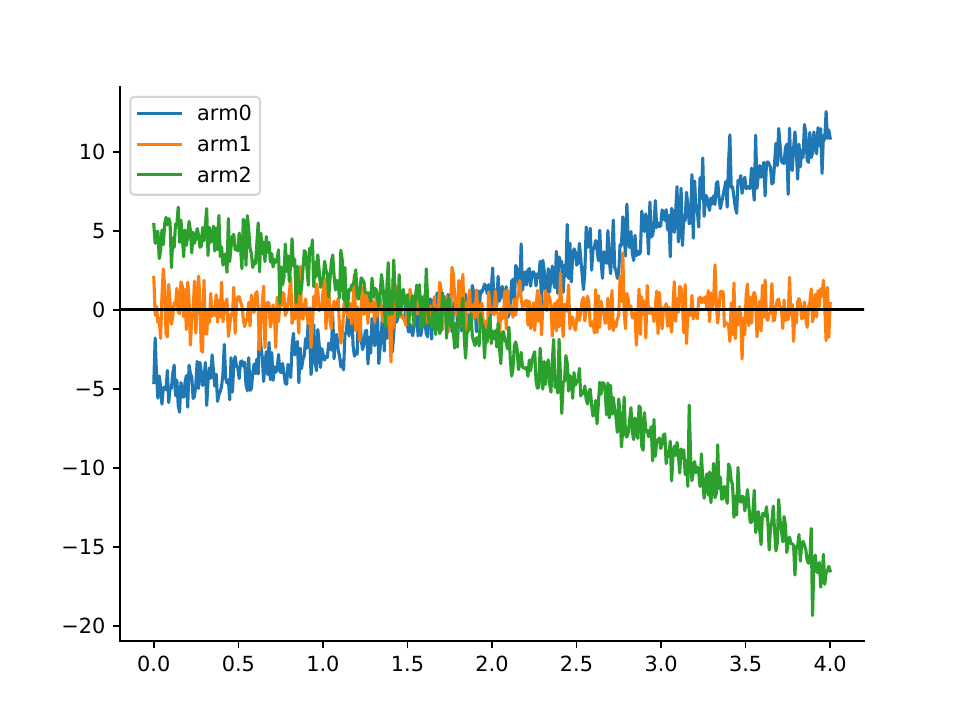}
\caption{ }
\end{subfigure}
~
\begin{subfigure}[b]{0.3\textwidth}
  \centering
  \includegraphics[height=1.5in]{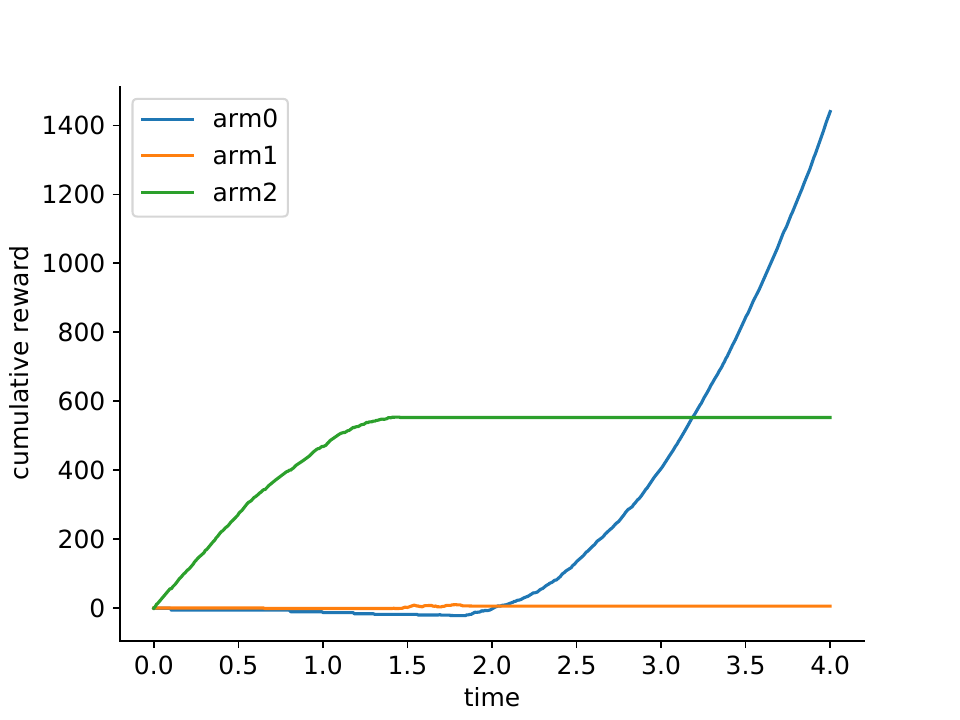}
\caption{ }
\end{subfigure}
~
\begin{subfigure}[b]{0.3\textwidth}
  \centering
    \includegraphics[height=1.5in]{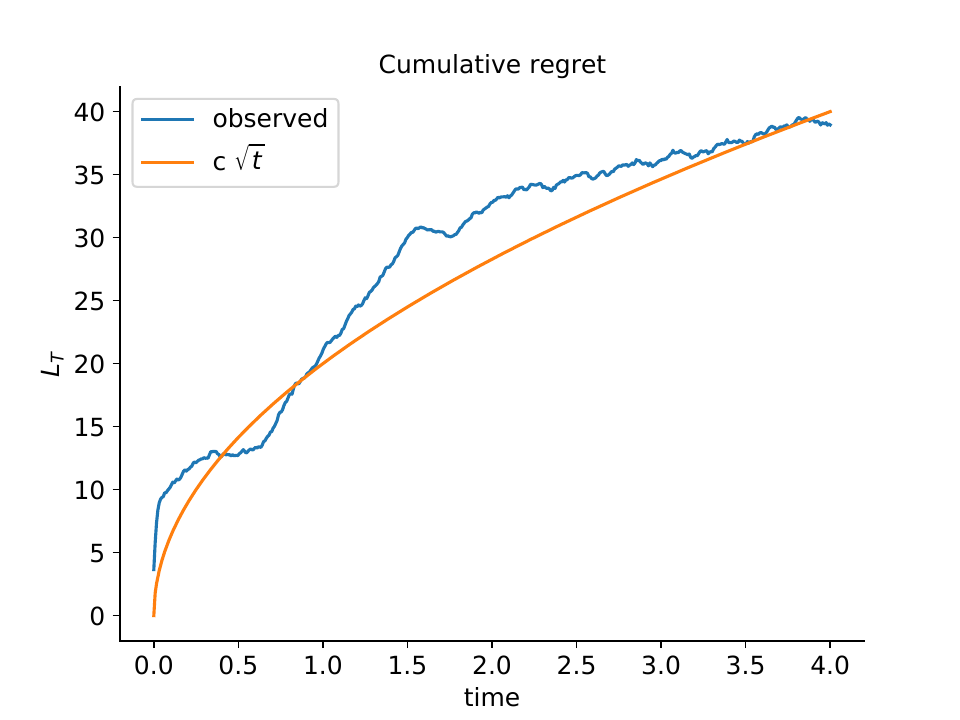}
\caption{ }
\end{subfigure}
\caption{
  Illustration of Thompson sampling applied to a linear-Gaussian
  contextual bandit. The context has the form $\vs_t =(1,t,t^2)$.
  (a) True reward for each arm vs time.
  (b) Cumulative reward per arm vs time.
  (c) Cumulative regret vs time.
  \figgen{\notebook{thompson\_sampling\_linear\_gaussian}}.
}
\label{fig:banditsLingauss}
\end{figure}

In \cref{fig:banditsLingauss}, we give a simple
example of Thompson sampling applied to a linear regression bandit.
The context has the form $\vs_t =(1,t,t^2)$.
The true reward function has the form
$R(\vs_t,a) = \vw_a^\trans \vs_t$.
The weights per arm are chosen as follows:
$\vw_0 = (-5, 2, 0.5)$,
$\vw_1 = (0,0,0)$,
$\vw_2 = (5, -1.5, -1)$.
Thus we see that arm 0 is initially worse (large negative bias)
but gets better over time (positive slope),
arm 1 is useless,
and arm 2 is initially better (large positive bias)
but gets worse over time.
The observation noise is the same for all arms, $\sigma^2=1$.
See \cref{fig:banditsLingauss}(a) for a plot of the reward function.
We use a conjugate Gaussian-gamma prior and perform
exact Bayesian updating. Thompson sampling
quickly discovers that arm 1 is useless.
Initially it pulls arm 2  more,
but it adapts to the non-stationary nature of the problem
and switches over to arm 0,
as shown in 
\cref{fig:banditsLingauss}(b).
In \cref{fig:banditsLingauss}(c), we show that the empirical
cumulative regret in blue is close to the optimal
lower bound in red.

\subsubsection{MDP case (posterior sampling RL)}
\label{sec:PSRL}

\eat{
(However, 
the Bayesian posterior sampling approach to MDPs,
discussed in \cref{sec:PSRL}, works much better,
both in theory and practice
\citep{Osband2013,osband2017posterior,Agrawal2017}.)
}

\begin{algorithm}
\dontprintsemicolon
\caption{Posterior sampling RL.
  We define the history 
  at step $k$ to be the set of previous trajectories,
  $H_k = \{\traj_1,\ldots,\traj_{k-1}\}$,
  each of length $H$,
where $\traj_k = (s_1^k,a_1^k,r_1^k,\ldots,s_H^k,a_H^k,r_H^k,s_{H+1}^k)$.
}
\label{algo:PSRL}
Input: Prior over models $P(M)$\\
History $H_1=\emptyset$ \\
\For{Episode $k=1:K$}
       {
         Sample model from posterior, $M_k \sim P(M|H_k)$ \\
         Compute optimal policy $\pi_k^*=\text{solve}(M_k)$ \\
         Execute $\pi_k^*$ for $H$ steps to get $\traj_k$ \\
         Update history $H_{k+1} = H_k \union \traj_k$ \\
         Update posterior $p(M|H_{k+1})$ 
       }
Return $\pi^*_K$
\end{algorithm}

We can generalize Thompson sampling to the (episodic) MDP case
by maintaining a posterior over all the model parameters
(reward function and transition model),
sampling an MDP from this belief state at the start of each episode,
solving for the optimal policy
corresponding to the sampled MDP,
using the resulting policy to collect new data,
and then updating the belief state at the end of the episode.
This is called \keywordDef{posterior sampling RL}
\citep{Strens00,Osband2013,Russo2014,osband2017posterior,Wang2024MBRL}.
See \cref{algo:PSRL} for the pseudocode.\footnote{
In \citep{Arumugam2025LLM}, they 
used prompted LLMs to 
implement a crude approximation
to the sampling, planning, and posterior updating steps
of the PSRL algorithm for some simple tabular problems.
Although the method worked susprisingly well (in the sense of having low
Bayesian regret) on very small problems
(e.g., 3 state RiverSwim), it failed on larger problems
(e.g., 4 state) and more stochastic problems.
}

As a more computationally efficient alternative,
it is also possible to maintain a posterior over policies
or $Q$ functions instead of over world models;
see e.g.,
\citep{Osband2016nips} for a simple implementation
of this idea using \keywordDef{bootstreap DQN},
\citep{Osband2023TS} for an implementation 
based on \keywordDef{epistemic neural networks}
\citep{Osband2023epi},
and
\keywordDef{epistemic value estimation}
\citep{Schmitt2023}
for an implementation based on Laplace approximation.
Another approach is to use successor features (\cref{sec:SF}),
where the $Q$ function is assumed to have the form
$Q^{\pi}(s,a) = \vpsi^{\pi}(s,a)^\trans \vw$.
In particular, \citep{Janz2019} proposes
\keywordDef{Sucessor Uncertainties},
in which they model the uncertainty over $\vw$
as a Gaussian,
$p(\vw) = \gauss(\vmu_{\vw}, \vSigma_{\vw})$.
From this they can derive the posterior distribution
over $Q$ values as
$p(Q(s,a)) = \gauss(\vPsi^{\pi} \vmu_{\vw}, \vPsi^{\pi} \vSigma_{\vw} (\vPsi^{\pi})^\trans)$,
where $\vPsi^{\pi}=[\vpsi^{\pi}(s,a)]^\trans$ is a matrix of features,
one per state-action pair.

\subsection{Upper confidence bounds (UCBs)}
\label{sec:UCB}

The optimal solution to explore-exploit is intractable.
However, an intuitively sensible approach
is based on the principle known as
``\keywordDef{optimism in   the face of uncertainty}''
(OFU).
The principle selects actions greedily, but based on
optimistic estimates of their rewards.
This approach is optimal in the regret minimization
sense, as  proved in the
\keywordDef{R-Max} paper of \citep{Tennenholtz2002},
which builds on the earlier \keywordDef{E3} paper of
\citep{Kearns2002}.

The most common implementation of this principle is
based on the notion of an
\keywordDef{upper confidence bound} or \keywordDef{UCB}.
We will initially explain this for the bandit case,
then extend to the MDP case.

\subsubsection{Basic idea}

To use a UCB strategy, the agent maintains an optimistic reward
function estimate $\optR_t$, so that $\optR_t(s_t,a) \ge R(s_t,a)$ for
all $a$ with high probability, and then chooses the greedy action
accordingly: 
\begin{align}
a_t = \argmax_a \optR_t(s_t,a)
\end{align}
UCB can be viewed a form of \keywordDef{exploration bonus}, where the optimistic
estimate encourages exploration.  Typically, the amount of optimism,
$\optR_t - R$, decreases over time so that the agent gradually reduces
exploration.  With properly constructed optimistic reward estimates,
the UCB strategy has been shown to achieve near-optimal regret in many
variants of bandits~\citep{Lattimore2019}.
(We discuss regret in \cref{sec:regret}.)

The optimistic function $\optR$ can be obtained in different ways,
sometimes in closed forms, as we discuss below.

\subsubsection{Bandit case: Frequentist approach}

A frequentist approach to computing a confidence bound
can be based on a
\keywordDef{concentration inequality}~\citep{Boucheron16} to derive a
high-probability upper bound of the estimation error: $|\estR_t(s,a) -
R_t(s,a)| \le \delta_t(s,a)$, where $\estR_t$ is a usual estimate of
$R$ (often the MLE), and $\delta_t$ is a properly selected function.
An optimistic reward is then obtained by setting 
$\optR_t(s,a) = \estR_t(s,a) + \delta_t(s,a)$.

As an example, consider again the context-free Bernoulli bandit,
$R(a) \sim \Ber(\mu(a))$.
The  MLE $\estR_t(a)=\hat{\mu}_t(a)$ is given by
the empirical average of observed rewards whenever action $a$ was taken:
\begin{align}
  \hat{\mu}_t(a) &= \frac{N_t^1(a)}{N_t(a)} = \frac{N_t^1(a)}{N_t^0(a) + N_t^1(a)}
  \label{eqn:dtheory-bandit-mlebernoulli} 
\end{align}
where $N_t^r(a)$ is the number of times (up to step $t-1$)
that action $a$ has been tried
and the observed reward was $r$,
and $N_t(a)$ is the total number of times action $a$ has been tried:
\be
N_t(a) = \sum_{s=1}^{t-1} \ind{a_t=a}
\ee
Then the \keywordDef{Chernoff-Hoeffding inequality}
\citep{Boucheron16} leads to
$\delta_t(a) = c / \sqrt{N_t(a)}$
for some constant $c$, so
\begin{align}
  \optR_t(a) = \hat{\mu}_t(a) + \frac{c}{\sqrt{N_t(a)}}
  \label{eqn:UCBchernoff}
\end{align}

\subsubsection{Bandit case: Bayesian approach}

We can also derive an upper confidence about using  Bayesian inference.
If we use a beta prior, we can compute the posterior in closed form,
as shown in \cref{eqn:postBetaBandit}.
The posterior mean is
$\hat{\mu}_t(a) = \expect{\mu(a)|\history_t}=\frac{\alpha_t^a}{\alpha_t^a+\beta_t^a}$,
and 
the posterior standard deviation is approximately
\be
\hat{\sigma}_t(a) = \sqrt{\var{\mu(a)|\history_t}}
\approx \sqrt{\frac{\hat{\mu}_t(a) (1-\hat{\mu}_t(a))}{N_t(a)}}
\ee

We can use similar techniques for a
\keyword{Gaussian bandit},
where
$\preward(R|a,\vtheta) = \gauss(R|\mu_a,  \sigma_a^2)$,
$\mu_a$ is the expected reward, and $\sigma_a^2$ the variance.
If we use a conjugate prior,
we can compute $p(\mu_a,\sigma_a|\data_t)$
in closed form.
Using an uninformative version of the conjugate prior,
we find
$\expect{\mu_a|\history_t}=\hat{\mu}_t(a)$,
which is just the empirical mean of rewards for action $a$.
The uncertainty in this estimate
is the  standard error of the mean,
i.e., 
$\sqrt{\var{\mu_a|\history_t}} =\hat{\sigma}_t(a)/\sqrt{N_t(a)}$,
where $\hat{\sigma}_t(a)$ is the empirical standard
deviation of the rewards for action $a$.

\begin{figure}
\centering
\includegraphics[height=1.5in]{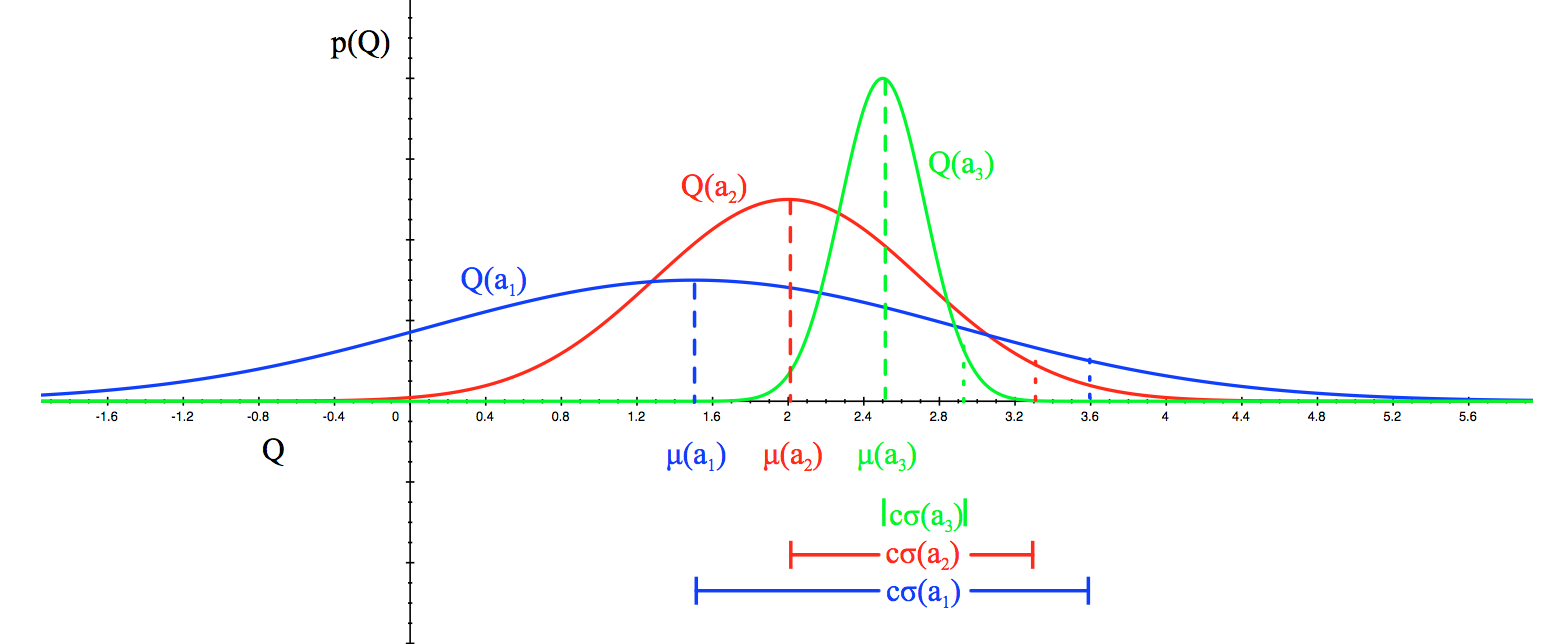}
\caption{
Illustration of the reward distribution $Q(a)$
for a Gaussian bandit with 3 different actions,
and the corresponding lower and upper confidence bounds.
We show the posterior means $Q(a) = \mu(a)$ with a vertical
dotted line, and the scaled posterior standard deviations $c \sigma(a)$
as a horizontal solid line.
  \figtaken{\citep{Silver2018L9}}.
\figthanks{David Silver}.
}
\label{fig:UCB}
\end{figure}

Once we have computed the mean and posterior standard
deviation, we define the optimistic reward estimate as
\be
\optR_t(a) = \hat{\mu}_t(a) + c \hat{\sigma}_t(a)
\label{eqn:UCBbandit}
\ee
for some constant $c$ that controls how greedy the policy is.
See \cref{fig:UCB}  for an illustration.
We see that this is similar to the frequentist method based on
concentration inequalities, but is more general.

\subsubsection{MDP case}

The UCB idea (especially in its frequentist form)
has been extended to the MDP case in several works.
(The Bayesian version is discussed in \cref{sec:thompson}.)
For example, \citep{Auer2002}
proposes to combine UCB with Q learning, by defining the policy as
\be
\pi(a|s) = \ind{a = \argmax_{a'} Q(s,a')
  + c \sqrt{\log(t)/N_t(s,a')}}
\ee
\citep{Auer2008} presents the more sophisticated \keywordDef{UCRL2} algorithm,
which computes confidence intervals on all the MDP model
parameters at the start of each episode; it then computes the resulting
\keywordDef{optimistic MDP} and solves for the optimal policy,
which it uses to collect more data.

\section{Distributional RL}
\label{sec:distributional}


The \keywordDef{distributional RL} approach
of \citep{Bellemare2017,BellemareBook},
predicts the distribution of (discounted) returns,
not just the expected return.
More precisely, let $Z_t^{\pi} = \sum_{k=0}^{T-t} \gamma^k R(s_{t+k},a_{t+k})$
be a random variable
representing the (discounted) reward-to-go from step $t$.
The standard value function is defined to compute the expectation
of this variable: $V^{\pi}(s) = \expect{Z_0^{\pi}|s_0=s}$.
In DRL, we instead attempt to learn the full distribution,
$p(Z_0^{\pi}|s_0=s)$ when training the critic.
We then compute the expectation of this distribution
when training the actor.
For a general review of distributional regression,
see \citep{Kneib2023}.
Below  we briefly mention a few algorithms in this class
that have been explored in the context of RL.

\subsection{Quantile regression methods}
\label{sec:QRDQN}

An alternative to predicting a full distribution
is to predict a fixed set of quantiles.
This is called quantile regression,
and has been used with DQN in
\citep{Dabney2017} to get \keywordDef{QR-DQN},
and with SAC in \citep{Wurman2022}
to get \keywordDef{QR-SAC}.
(The latter  was used in Sony's \keywordDef{GTSophy}
Gran Turismo AI racing agent.)


\eat{
\footnote{
Interestingly, the original DQN paper did not
learn to predict the mean of the Q function.
since it was not trained with quadratic loss.
Instead it was trained with Huber loss,
which is quadratic if the residual error is less than 1,
and otherwise is a linear function.
Hence the gradient is clipped to be between
-1 and 1.
See \url{https://openai.com/index/openai-baselines-dqn/}
for a discussion.
}
}

\subsection{Replacing regression with classification}
\label{sec:classifRL}
\label{sec:C51}
\label{sec:catDQN}

\begin{figure}
\centering
\includegraphics[height=1.5in]{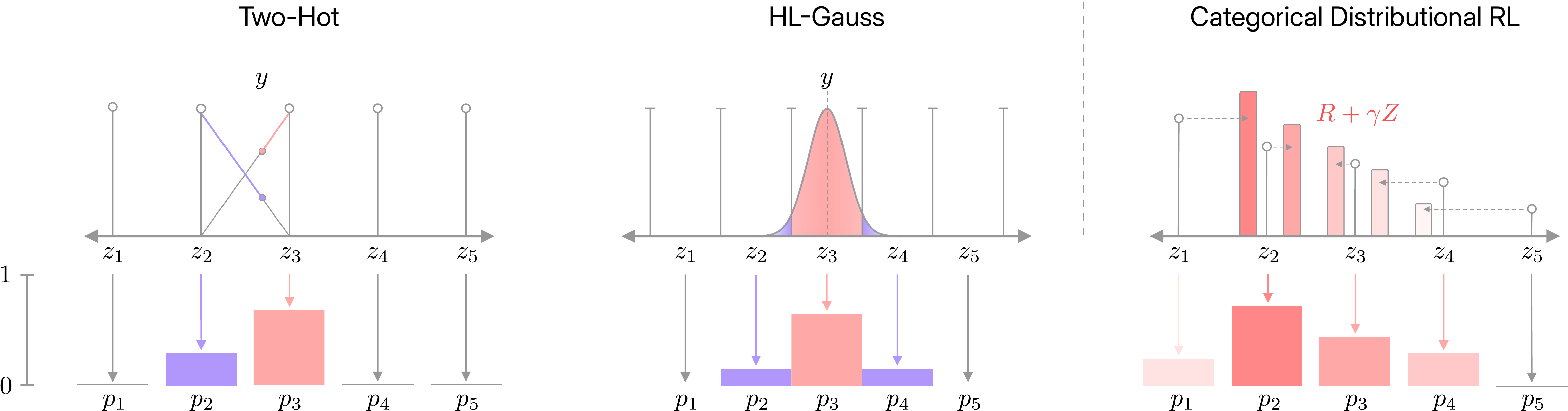}
\caption{
  Illustration of how to encode a scalar target $y$
  or distributional target $Z$
using a categorical distribution.
\figtaken{Figure 1 of \citep{Farebrother2024}}.
\figthanks{Jesse Farebrother}.
}
\label{fig:twoHot}
\end{figure}

An alternative to quantile regression
is to approximate the distribution over returns
using a histogram, and then fit it
using cross entropy loss (see \cref{fig:twoHot}).
This approach was first suggested in \citep{Bellemare2017},
who called it  \keywordDef{categorical DQN}.
(In their paper, they use 51 discrete categories (atoms),
giving rise to the name \keywordDef{C51}.) 

An even simpler approach is to replace the distributional
target with the standard scalar target (representing the mean),
and then discretize this target
and use cross entropy loss instead of squared error.\footnote{
Technically speaking, this is no longer a distributional RL method,
since the prediction target is the mean,
but the mechanism for predicting the mean leverages a distribution,
for robustness and ease of optimization.
}
Unfortunately, this encoding is lossy.
In  \citep{Schrittwieser2020}, they
proposed the \keywordDef{two-hot} transform,
that is a lossless encoding of the target based on putting
appropriate weight on the nearest two bins (see \cref{fig:twoHot}).
In \citep{Imani2018}, they proposed the
\keywordDef{HL-Gauss} histogram loss,
that convolves the target value $y$ with a Gaussian,
and then discretizes the resulting continuous distribution.
This is  more symetric than two-hot encoding, as shown in \cref{fig:twoHot}.
Regardless of how the discrete target is chosen,
predictions are made using
$\hat{y}(s;\vtheta) = \sum_k p_k(s) b_k$,
where $p_k(s)$ is the probability of bin $k$,
and $b_k$ is the bin center.

In \citep{Farebrother2024}, they show that 
the HL-Gauss trick  works much better than
MSE,  two-hot and C51 across a variety of problems
(both offline and online),
especially  when they scale to large networks.
They conjecture that the reason it beats MSE
is that cross entropy is more robust to noisy targets
(e.g., due to stochasticity) and nonstationary targets.
They also conjecture  that the reason HL works better than two-hot
is that HL is closer to ordinal regression, and reduces overfitting
by having a softer (more entropic) target distribution
(similiar to label smoothing in classification problems).

Recently, \citep{Gillman2025} proposed an alternative
called \keywordDef{Fourier head}, in which the linear output layer
is replaced by a fourier transform, before discretizing.
This is compatible with standard transformer training,
and gives improved results when generating continuous outputs.

\section{Intrinsic motivation for reward-free RL}
\label{sec:intrinsicReward}
\label{sec:intrinsic}

When the extrinsic reward is sparse, or does not exist at all
(``unsupervised RL''),
it can be useful to  reward the agent for solving
``generally useful'' tasks,
such as learning about the world,
or developing a set of skills.
This is called \keywordDef{intrinsically motivated RL}
\citep{Aubret2019,Linke2019,Amin2021,Ladosz2022,Yuan2022intrinsic,Colas2022,
  Aubret2023,Yuan2024rlexplore}.


We can classify these methods into two main types:
(1) \keywordDef{knowledge-based intrinsic motivation},
or \keywordDef{artificial curiosity},
where the agent is
rewarded for learning about its environment
(this is focused on reducing prediction error);
and
(2)  \keywordDef{competence-based intrinsic motivation},
where the agent is rewarded for achieving novel goals
or mastering new skills
(this is focused on control).

\subsection{Knowledge-based intrinsic motivation}

In this section, we discuss some approaches to
knowledge-based intrinsic motivation.

\subsubsection{Exploration bonuses}

One simple approach to
is to create an intrinsic \keywordDef{exploration bonus}
$R_t^i(s_t)$
which is high when the agent visits novel states.
For tabular environments, we can just count the number of visits to each state,
$N_t(s)$, and define $R_t^i(s) = 1/N_t(s)$ or
$R_t^i(s) = 1/\sqrt{N_t(s)}$, which is similar to the UCB
heuristic used in bandits (see \cref{sec:UCB}).
We can  extend exploration bonuses  to high dimensional states (e.g. images)
using density models \citep{Bellemare2016}.
Alternatively, \citep{Machado2020} propose to use the $\ell_1$
norm of the successor feature (\cref{sec:SF}) representation
as an alternative to the visitation count, giving rise
to an intrinsic reward of the form
$R^i(s) = 1/||\vpsi^{\pi}(s)||_1$.
Recently \citep{Yu2023successor} extended this to combine
SFs with {\em predecessor} representations,
which encode retrospective information about the previous state
(c.f., inverse dynamics models, mentioned below).
This encourages exploration towards bottleneck states.

\subsubsection{Random Network Distillation (RND)}

Another approach
is the \keywordDef{Random Network Distillation} or \keywordDef{RND}
method of \citep{Burda2018}.
This uses a fixed random neural network feature extractor
$\vz_t=f(\vs_t;\vtheta^*)$ to define a target,
and then trains a predictor $\hat{\vz}_t=f(\vs_t;\hat{\vtheta}_t)$
to predict these targets.
If $\vs_t$ is similar to previously seen states,
then the trained model will have low prediction error.
We can thus define the intrinsic reward as proportional
to the squared error $||\hat{\vz}_t - \vz_t||_2^2$.
The \keywordDef{BYOL-Explore} method of
\citep{BYOLexplore} goes beyond RND by learning
the target representation (for the next state),
rather than using a fixed random projection,
but is still based on prediction error.

\eat{
The value function now has two output heads,
one to predict the sum of extrinsic rewards,
and another to predict the sum of intrinsic rewards;
the latter can change over time as the agent learns.
The overall value is the sum of these two heads,
similar to the reward shaping term
$V(s;\vw) = V^e(s;\vw) + V^i(s;\vw')$ in \cref{sec:rewardShaping}.

The RND method has given good results on Atari games
with sparse reward, such as \keyword{Montezuma's revenge}.
as well as some continuous control problems
\citep{Laskin2021}.
}

\subsubsection{Information-theoretic measures}

We can also define an intrinsic reward in terms
of the information theoretic \keywordDef{surprise}
of the next state given the current one:
\be
R(\vs,\va,\vs') = -\log q(\vs'|\vs,\va)
\ee
This is the same as methods based on
rewarding states for prediction error.
Unfortunately such methods
can suffer from 
the \keywordDef{noisy TV problem}
(also called a \keywordDef{stochastic trap}),
in which an agent is attracted to states 
which are intrinsically to predict.
To see this, note that by averaging
over future states we see that the above reward reduces to
\be
R(\vs,\va) = -\expectQ{\log q(\vs'|\vs,\va)}{p^*(\vs'|\vs,\va)}
= \crossentropy(p^*, q)
\ee
where $p^*$ is the true model and $q$ is the learned dynamics model,
and $\crossentropy$ is the cross -entropy.
As we learn the optimal model, $q=p^*$,
this reduces to the conditional
entropy of the predictive distribution,
which can be non-zero for inherently unpredictable states.

To help filter out such random noise,
\citep{Pathak2017} proposes an
\keywordDef{Intrinsic Curiosity Module}.
This first learns  an \keywordDef{inverse dynamics model}
of the form $a = f(\vs,\vs')$, which tries
to predict which action was used, given
that the agent was in $\vs$ and is now in $\vs'$.
The classifier has the form
$\softmax(g(\phi(\vs), \phi(s'), a))$,
where $\vz=\phi(\vs)$ is  a representation function
that focuses  on parts of the state that the agent can control.
Then the agent learns a forwards dynamics model
in $\vz$-space.
Finally it defines the intrinsic reward as
\be
R(\vs,\va,\vs') = -\log q(\phi(\vs')|\phi(\vs),a)
\ee
Thus the agent is rewarded for visiting states that lead
to unpredictable consequences,
where the difference in outcomes is measured
in a (hopefully more meaningful) latent space.

Another solution  is to replace the cross entropy
with the KL divergence,
$R(\vs,\va) = \KL(p||q) = \crossentropy(p,q) - \entropy(p)$,
which goes to zero once the learned model matches the true model,
even for unpredictable states.
This has the desired effect of encouraging exploration towards states
which have epistemic uncertainty (reducible noise)
but not aleatoric uncertainty (irreducible noise)
\citep{Mavor-Parker2022}.
The \keywordDef{BYOL-Hindsight} method of \citep{Jarrett2023}
is one recent approach that attempts to use
the $R(\vs,\va) = \KL(p||q)$ objective.
Unfortunately, computing the $\KL(p||q)$ term is much harder
than the usual variational objective of $\KL(q||p)$.
A related idea,
proposed in the RL context by \citep{Schmidhuber2010},
is to use the \keywordDef{information gain} as a reward.
This is defined as
$R_t(\vs_t,\va_t)=\KL(q(\vs_t|\vh_t,\va_t,\vtheta_t)
|| q(\vs_t|\vh_t,\va_t,\vtheta_{t-1})$,
where $\vh_t$ is the history of past observations,
and $\vtheta_t=\text{update}(\vtheta_{t-1}, \vh_t, \va_t, \vs_t)$
are the new model parameters.
This is closely related to the BALD (Bayesian Active Learning by Disagreement)
criterion
\citep{Houlsby2011,Kirsch2019},  and has the advantage
of being  easier to compute, since it is does not reference
the true distribution $p$.


\subsection{Competence-based  intrinsic motivation}
\label{sec:intrinsicGoals}

Another way to explore the environment
is to use  goal-conditioned RL
(see \cref{sec:GCRL}), where the agent creates its own goals;
this is known as an \keywordDef{autotelic agent}
\citep{Colas2022}.
Intuitively it is desirable
to choose goals that cover the set of states.
We discuss some ways of choosing these goals below.

\subsubsection{Empowerment}

One approach to choosing goals is to use
the concept
of \keywordDef{empowerment} \citep{Klyubin2005},
which is defined as the
mutual information between the goal $G$ (or past action $A$)
and the future state $S$.
This is given by
\be
I(G,S) = H(S)  - H(S|G)
\ee
Thus we can maximize empowerment by maximizing
the entropy of the states (a form of diversity)
while minimizing the conditional entropy of the states given
the goal  (ensuring that the goal is predictable in its effects).
For more details,
see e.g., \citep{Gregor2017, Eysenbach2019diversity, Choi2021,Tiomkin2024},
and for a cognitive science perspective,
see \citep{Schulz2012, Du2023explore, Goddu2024, Yiu2025}.

\subsubsection{Curriculum design}
\label{sec:curriculum}

Since the space of possible goals is usually too vast to explore,
it is important to choose useful goals for the agent to learn from.
A good goal is often defined
as one that is not too hard or too easy to learn,
since this maximizes \keywordDef{learning progress},
also called the ``\keywordDef{zone of proximal development}''.
Choosing the best order in which to tackle
various goals is an example of
\keywordDef{automatic curriculum} design;
similar methods can also be used to automatically
design new environments
(see e.g., \citep{Portelas2020,Forestier2022} for a review),
a process which is sometimes called \keywordDef{open-ended learning}.
It is also possible to train one agent to design an environment
that another other agent finds challenging to solve;
this is known as \keywordDef{aysmmetric self-play}
(see \cref{sec:goalsASP}).
The convergence of such \keyword{setter-solver}
systems can be studied using the tools
of Stackelberg equilibria (\cref{sec:stackelberg}).

\subsubsection{Using an LLM to choose goals}

Another approach to goal generation is to use suitably prompted LLMs
(see e.g., \citep{Du2023RLLLM,Colas2023augment}).
This can leverage the prior knowledge of LLMs to not only
propose novel goals, but also ones that are plausibly useful
to humans.

\subsubsection{Go-Explore}
\label{sec:goExplore}

The \keywordDef{Go-Explore} algorithm from
\citep{Ecoffet2019,Ecoffet2021}
proposes to first follow a goal-conditioned policy
to reach (or to reset the environment state to)
a goal state, which is chosen from an archive of
``interestingly new'' previously visited states
(e.g., rarely visited ones),
and then switch to an exploration policy
(using random actions)
to expand the coverage of the state space.

In  \citep{Lu2024go},
they present \keywordDef{Intelligent Go-Explore},
which uses an LLM
to decide  what is an interesting goal to return to
(leveraging the LLM's ability to determine
interestingess \citep{omni}),
to decide what exploratory actions to take after reaching the goal,
and to decide whether to add any newly explored states
to the archive.

In \citep{Hu2023planning}, they choose a
goal $g$ for Go-Explore based on the expected value
of the state they would end up in, if they
followed the goal-conditioned policy towards $g$
and then switched to exploration mode.
This expected value is computed using a learned world model,
following the LEXA method of \citep{Mendonca2021}
(which in turn is based
 on Dreamer, discussed in \cref{sec:dreamer}).

\eat{
The question of when and how an agent should switch to pursuing
a new goal is studied in \citep{Pislar2022}
(see also \citep{Bagaria2023}).
Some other key work in this space includes
the \keywordDef{scheduled auxiliary control}
method of \citep{Riedmiller2018},
and
}

\eat{
\subsection{Meta-learning the reward function}
\label{sec:rewardMeta}

It may be possible to learn the reward function
using  \keywordDef{bilevel optimization problem},
where the outerloop searches over reward functions,
and the inner loop searches over policies that maximize the given reward.
The value returned by the inner loop is the score which is maximized by the outer loop.
This is analogous to evolution, which searches 
over reward functions, such as the desire for food and sex,
which are then optimized by the agent using RL.
Evolution's goal is to maximize the evolutionary fitness of each agent,
which cannot be directly measured by the agent itself,
but can be estimated at a population level.
}

\section{Hierarchical RL}
\label{sec:HRL}


So far we have focused on MDPs that work at a single time scale.
However, this is very limiting.
For example, imagine planning a trip from San Francisco to New York:
we need to choose high level actions first, such as which airline to fly,
and then medium level actions, such as how to get to the airport,
followed by low level actions, such as motor commands.
Thus we need to consider actions that operate multiple levels
of \keywordDef{temporal abstraction}.
Fitting policies that use temporal (and optionally also state)
abstraction 
is called \keywordDef{hierarchical RL} or \keywordDef{HRL}.
This is a big and important topic,
and we only brief mention a few key ideas and methods.
Our summary is based on
\citep{Pateria2022},
but see \citep{Klissarov2025} for a more recent survey.
(See also \cref{sec:beyond} where we discuss multi-step
predictive models; by contrast, in this section we focus on model-free methods.)

\subsection{HRL using Options}
\label{sec:options}

\subsubsection{Introduction}

Before we define the HRL problem,
we define the concept of an ``option''
\citep{options},
which is a form of temporally extended action.
Formally,
an \keywordDef{option} $\omega=(I,\pi,\beta)$ is a tuple consisting of
the following:
\begin{itemize}
\item The \keywordDef{subpolicy}
(aka \keywordDef{intra-option policy},
  or \keywordDef{action policy}) $\pi_{\omega}(a|s) \in [0,1]$.

  \item The \keywordDef{termination probability}
$\beta_{\omega}(s) \in [0,1]$,
which gives the probability of finishing in state $s$.
This induces a geometric distribution over option durations,
which we denote by $\tau \sim \beta_{\omega}$.

\item The \keywordDef{initiation set} $I_{\omega} \subset S$,
  which is the set of states this option can start from.
  Alternatively, we can define $I_{\omega}(s) \in [0,1]$
  as the probability that $\omega$ can be started from $s$
  and achieve its goal.
  (Note that
  the \keywordDef{affordances} of  a state,
  $A(s) = \{ \omega: I_{\omega}(s)>\epsilon\}$,
 is the set of options that can be initiated in a given
 state in order to achieve a specific intent or goal
  \citep{Khetarpal2020}.)
\end{itemize}

We may also optionally associate a reward function with each option,
denoted $R_{\omega}(s)$.
The set of all options is denoted $\Omega$,
and the distribution over options (i.e., the
\keywordDef{high level policy}) is denoted by
$\mu(o|s)$.

Executing an option at step $t$ entails choosing an action using
$a_t = \pi_{\omega}(s_t)$ and then deciding whether
to terminate at step $t+1$ with probability
$1-\beta_{\omega}(s_{t+1})$ or to continue following the option at step $t+1$.
(This is an example of a \keywordDef{semi-Markov decision process}
\citep{Puterman94}.)
If we define $\pi_{\omega}(s)=a$ and $\beta_{\omega}(s)=0$ for all $s$, then
this option corresponds to primitive action $a$ that terminates in one step.
But with options we can expand the repertoire of actions to include those that
take many steps to finish.

Note that goal-conditioned RL (\cref{sec:GCRL})
can be considered a special case of options
where each option corresponds to a different goal.
The reward function for each option has the form
$R_{\omega}(s)=\text{sim}(s,\omega)$,
the termination function is
$\beta_{\omega}(s)=\text{sim}(s,\omega) > \text{thresh}$,
  and the initiation set  is the entire state space,
  so $I_{\omega}(s)=1$ for all $s$.
  (An option that can be initiated from any state is called
  a \keywordDef{global option}.)

To create a semi-MDP with options, we need to define the reward function
and dynamics model. The reward is defined as follows:
\be
R(s,\omega) = \expect{R_1 + \gamma R^2 + \cdots + \gamma^{\tau-1} R_{\tau}
  \vert S_0=s, A_{0:\tau-1} \sim \pi_{\omega}, \tau \sim \beta_{\omega} }
\ee
The dynamics model is defined as follows:
\be
T_{\gamma}(s'|s,\omega) = \sum_{k=1}^{\infty}
\gamma^k \Pr\left(S_k=s', \beta_{\omega}(s_k) |
S_0=s, A_{0:k-1} \sim \pi_{\omega} \right)
\ee
Note that $p_{\gamma}(s'|s,\omega)$ is not a conditional probability distribution,
because of the $\gamma^k$ term,
but we can usually treat it like one.
Note also that a dynamics model  that can predict multiple steps ahead
is sometimes called a \keywordDef{jumpy model}
(see also \cref{sec:jumpy}).

We can use these
definitions to define the value function for a hierarchical policy
using a generalized Bellman equation, as follows:
\be
V_{\pi}(s) = \sum_{\omega  \in \Omega(s)}
\pi(\omega|s)
\left[ R(s,\omega) + \sum_{s'} T_{\gamma}(s'|s,\omega)  V_{\pi}(s') \right]
\ee
We can compute this using value iteration.
We can  then learn a policy using
policy iteration, or a policy gradient method.
In other words, once we have defined the options,
we can use all the standard RL machinery.

If we have a set of options, learning the corresponding
semi-MDP is harder than learning a standard MDP.
We can sidestep this by using the
\keywordDef{Expected-Length Model} (ELM) of options
proposed in \citep{Abel2019},
as opposed to the standard Multi-Time Model (MLM).
Let $\tilde{\tau}_{\omega}$ be the average
number of timesteps needed to execute option $\omega$.
Then we define the
 dynamics model as
\be
T_{\gamma}(s'|s,\omega) = \gamma^{\tilde{\tau}_{\omega}}
T(s'|s,\omega)
\ee
where $T(s'|s,\omega)$ is the distribution over next states $s'$
after $\omega$ has finished running, when started in $s$.
In \citep{Abel2019}, they show that this is a (biased)
estimator of the MTM transition model,
but which is easier to learn.
In particular, 
we can learn this model from trajectory
data which records $(s,s',\tau,r)$ tuples,
by maximizing the log likelihood of transitions
$T(s'|s,\omega)$ and fitting a regression model to $\tilde{\tau}$
and $R$.

\subsubsection{Option hierarchies}

\begin{figure}
\centering
\includegraphics[height=2.5in]{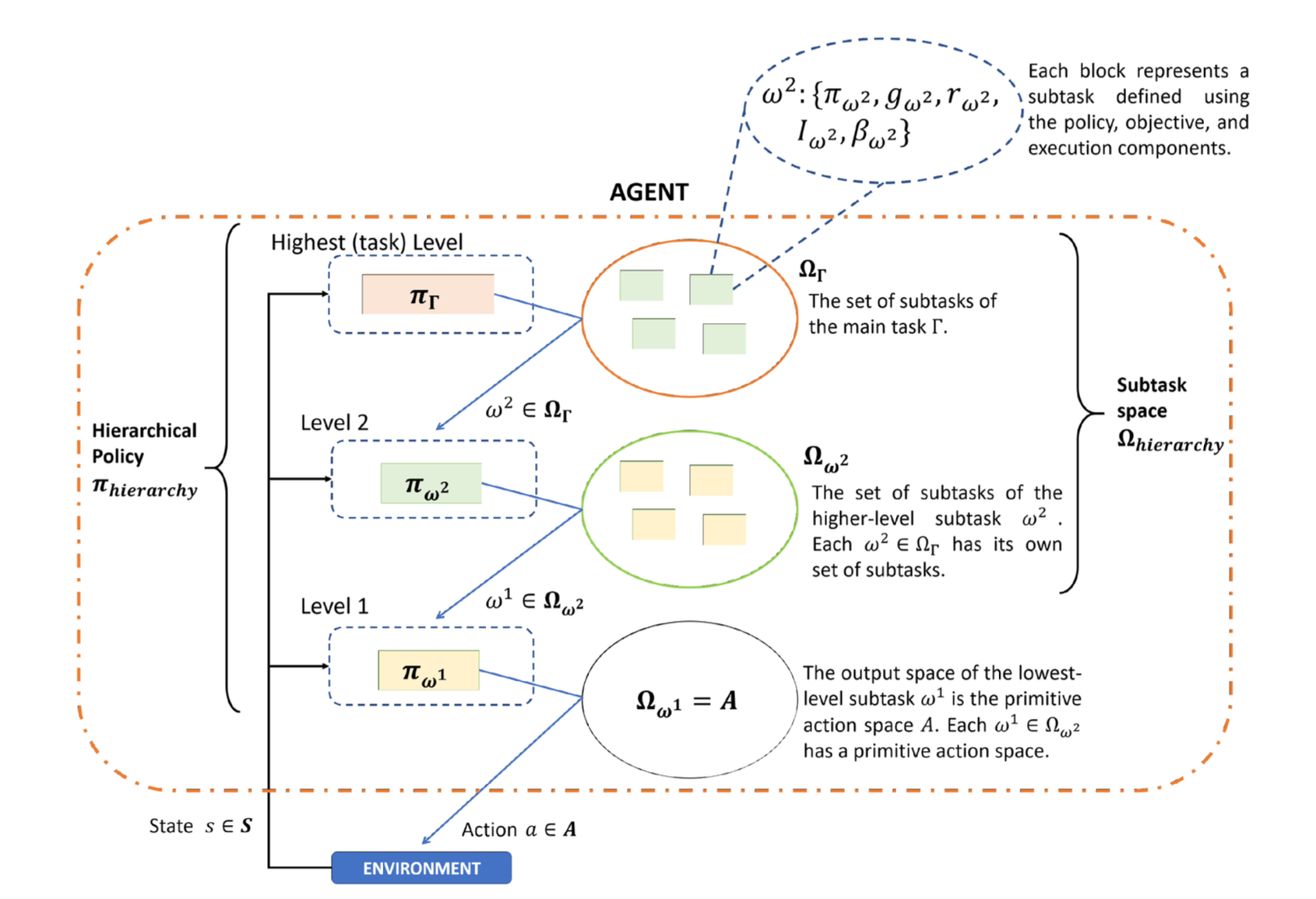}
\caption{
  Illustration of a 3 level hierarchical RL system.
\figtaken{Fig 2 of \citep{Pateria2022}}.
}
\label{fig:HRL}
\end{figure}

We can now define the HRL problem in terms of a nested set of options.
Let $\Omega_l$ be the set of options or subtasks at level $l$
of the hierarchy, where $\Omega_1=\calA$ is the set of primitive actions.
Let $\Gamma$ denote the top of the hierarchy, corresponding to the main task.
The policy at level $l$ is denoted $\pi_l(\omega^l|s)$,
where $\omega^l \in \Omega_l$ is the chosen subtask.
The \keywordDef{subtask space} is 
$\Omega_{\hier} = \{ \Omega_2, \Omega_3,\ldots, \Omega_{\Gamma}\}$.
The \keywordDef{hierarchical policy} is
$\pi_{\hier} = \pi_1 \odot \ldots \odot \pi_{\Gamma}$,
which maps states to primitive actions,
by succesively invoking policies at descending levels
of the hierarchy.
See \cref{fig:HRL} for an illustration.

Finally we can define HRL to be the following
optimization problem:
\begin{align}
\Omega^*_{\hier}, \pi^*_{\hier}
&= \argmax_{\Omega_{\hier}} \argmax_{\pi_{\hier}|\Omega_{\hier}}
Q^{\hier}(s,a) \\
Q^{\hier}(s_t,a_t)
&= \expectQ{\sum_{i=0}^{\infty} \gamma^{t+i} R(s_{t+1},a_{t+i}) | s_t,a_t}
{a \sim \pi_{\hier}|\Omega_{\hier}}
\end{align}
We see the problem breaks down into two parts:
learning the hierarchical policy $\pi_{\hier}$
given a fixed hierarchy of tasks, $\Omega_{\hier}$,
and learning the hierarchy itself.
The latter is called
the \keywordDef{subtask discovery} problem,
and can be tackled jointly with policy learning,
or can be done in a pre-training phase,
as we discuss below.

\eat{
We will consider two main approaches to HRL.
In the first approach, based on options,
we restrict ourselves to two-level hierarchy,
where the top level policy chooses which option to call,
and the bottom level policy implements the option.
Thus the action space of the top level is discrete.
The value function for each option reflects the overall value
to the agent, coming from the expected sum of external
rewards, up until the end of the task.
Thus the options can increase the efficiency of the agent's
learning, but will never reduce its performance,
since the top level policy is always free to just choose
primitive actions.
}


\subsubsection{Hierarchical Q learning}

Suppose we have a set of options $\calO$,
with fixed initiation and termination functions.
We can learn the corresponding option policy using Q-learning,
as follows.
Suppose we have a trajectory of $(s_t,o_t,r_{t:t+\tau},s_{t+\tau})$
tuples, where $\tau$ is the duration of the option $o_t$.
We then fit  $Q(s,o)$ by regressing towards the following target:
\begin{align}
y_t &= \sum_{t'=t}^{\tau}
\gamma^{t'-t} r_{t'}
+ \gamma^{\tau-t} Q(s_{t+\tau}, o^*_{t+\tau})\\
o^*_{t+\tau} &= \argmax_{o' \in \calO'(s_{t+\tau})} Q(s_{t+\tau}, o')\\
\calO'(s) &= \{o_i:
I_{o_i}(s) = 1 \intersect
\beta_{o_i}(s) = 0,
\; \forall o_i \in \calO\}
\end{align}
The individual option policies can be trained by 1-step Q learning
in the usual way.

\subsubsection{MAXQ}

The Q-value function of any an option
represents the expected cumulative reward until the end
of the main task, rather than
only until the horizon (or termination) of the option itself.
Hence, an option is not a standalone
subtask unit; instead it blends into the
\keywordDef{core MDP}.
This limits the transferability of the learned options
to other tasks,
but theoretically guarantees the optimality of the learned
hierarchical policy.

In the \keywordDef{MAXQ} approach of
\citep{Dietterich2000hrl},
the  core MDP is decomposed into
smaller sub-MDP components. Each sub-MDP is associated to a subtask whose policy can be
learned separately from other subtasks. This is achieved by decomposing the main Q-value
function into the separate Q-value functions of the subtasks.
Hence the Q-value
of any subtask represents the expected cumulative reward only until the horizon
(or termination) of that subtask.
This allows  each subtask policy to be learned as a standalone unit.
However, the resulting policy will only be recursively optimal, rather than globally optimal.

\subsubsection{Option learning using EM}

If the set of options is unknown, we can learn them by
segmenting the trajectories into sub-trajectories,
which correspond to the latent options.
This can be done
using the EM algorithm \citep{Daniel2016}.

\subsubsection{Skill chaining}
\label{sec:skillChaining}

In  \citep{Konidaris2009,Bagaria2020},
they propose the \keywordDef{skill chaining} method.
The process begins by establishing an initial skill, let's call it
$\omega$, with the primary goal of the overall task as its
objective. The agent then learns the necessary actions (the policy,
denoted as $\pi_\omega$) to achieve this main goal.  

Next, a system is trained to identify the starting points, or
``initiation states'', for this skill. States from which the main goal
can be successfully reached within a set timeframe are marked as
``positive''. Conversely, states where the goal is not reached within
this time limit are labeled ``negative.''. These successful starting
points are then grouped together to form the initiation set for skill
$\omega$, designated as $I_\omega$. 

Following this, a new skill, $\omega'$, is created. The objective of
this new skill is to reach the initiation set ($I_\omega$) of the
first skill. The same process of learning the policy and identifying
the initiation set is then repeated for $\omega'$. This creates a
chain, where each new skill's purpose is to get the agent to a state
where the previous skill can be activated. This method allows the
agent to autonomously discover and build a variable and adaptable
number of interconnected skills.
However, it only works for tasks that have explicit goal states,
and requires strong exploration methods, without using options,
to find out how to reach  the goal states,
which is particularly problematic for the initial goal.
(One practical approach, used in the
\keywordDef{DemoStart} system of \citep{Bauza2024},
is to rely on a human demonstration that reaches the goal,
and then initialize the agent in a state that is near
the end of this trajectory.)

\subsubsection{Option critic}

We can also learn options
using end-to-end training,
i.e., the options and their policies are jointly learned online.
For example,
\citep{Bacon2017} propose the  \keywordDef{option-critic} architecture.
The number of options is manually specified,
and all policies are randomly initialized.
Then they are jointly trained using policy gradient methods
designed for semi-MDPs.
(See also \citep{Riemer2018} for a hierarchical
extension of option-critic to support options calling options.)
However, since the learning signal is just the main task reward,
the method can work poorly
in problems with sparse reward compared to
subgoal methods
(see discussion in \citep{Vezhnevets2017,Nachum2019}).

\subsubsection{Double actor critic (DAC)}

One  problem with option-critic is that it requires
specialized methods that are designed for optimizing semi-MDPs.
In \citep{Zhang2019dac}, they propose
\keywordDef{double actor critic},
which allows the use of standard policy gradient methods.
This works  by defining two parallel
\keywordDef{augmented MDPs},
where the state space of each MDP is the cross-product
of the original state space and the set of options.
For a two level hierarchy, the manager learns a policy over options,
and the worker learns a policy over states for each option.
Both MDPs just use task rewards, without subgoals or
subtask rewards.

\subsubsection{Avoiding excessive (or insufficient) option switching}

It has been observed that
option learning using option-critic or double actor-critic can fail,
in the sense that the top level controller
may learn to switch from one option to the next at almost
every time step
\citep{Zhang2019dac,Harb2018}.
The reason is that the
optimal policy does not require
the use of temporally extended options,
but instead can be defined in terms of primitive actions
(as in standard RL).
Therefore in
\citep{Harb2018} they propose to add a regularizer
called the \keywordDef{deliberation cost},
in which the higher level policy is penalized
whenever it switches options.
This can speed up learning, at the cost of a potentially
suboptimal policy.

Another possible failure mode in option learning
is if the higher level policy selects a single
option for the entire task duration.
To combat this, \citep{Khetarpal2019} propose the
\keywordDef{Interest Option Critic},
which learns the  initiation condition $I_{\omega}$
so that the option is selected
only in certain states of interest,
rather than the entire state space.

\eat{
In \citep{Machado2023}, they discuss how the successor
representation (discussed in \cref{sec:successor})
can be used to define options,
using a method they call the \keywordDef{Representation-driven
  Option Discovery} (ROD) cycle.
}


\eat{
Unfortunately,  it is currently unclear how best
to learn options  automatically,
although there has been some work
(see e.g.,  \citep{Ramesh2019,Sutton2023}).
See also \cref{sec:HER} where we discuss an alternative approach to
HRL using goal-conditioned policies.
}

\subsubsection{MBRL using options}

Most work on options is based on model-free RL.
In \citep{Sutton2023}, they discuss the model-based RL case.

\subsection{HRL using feudal hierarchies}
\label{sec:feudal}

\subsubsection{Introduction}

\begin{figure}
\centering
\includegraphics[height=2.5in]{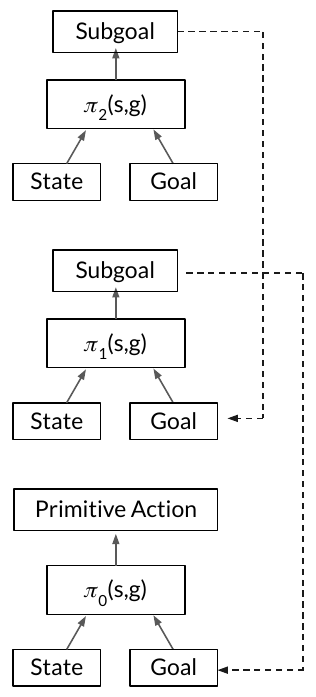}
\caption{
  Illustration of a 3 level hierarchical goal-conditioned controller.
\figtaken{\url{http://bigai.cs.brown.edu/2019/09/03/hac.html}}.
\figthanks{Andrew Levy}.
}
\label{fig:feudal}
\end{figure}

We now discuss the other main framework for  HRL,
known as \keywordDef{feudal RL} \citep{Dayan1992}.
In this approach,  the policy at level $l$ (known as a
\keywordDef{manager}) chooses a goal
from some goal space (equal to the state space,
or some abstraction thereof),
and passes that down to the level below
(known as a \keywordDef{worker}),
as shown in \cref{fig:feudal}.
Thus rather than having a finite number of options to choose from, we can
have a nested set of parameterized (universal) policies,
$\pi_l(g_{l-1}|s,g_l)$, for each manager level $l$,
and $\pi_1(a|s,g_1)$ for the worker level.
The value for a policy at a given level is the expected reward
until the policy finishes,
where the reward is defined intrinsically in terms of reaching the specified goal.
Thus  only the top level manager gets to see the external (environment) reward,
a principle known as \keywordDef{reward hiding}.

\subsubsection{Comparison with options}

Although the feudal approach is somewhat easier to learn
(due to the locality/modularity of subgoals, as we will see),
the resulting hierarchical policy may be suboptimal
compared to the optimal flat policy,
since learning is performed at each level wrt local goals,
rather than optimizing a single global objective.
In addition, not all subroutines or skills can be defined in terms
of reaching a goal state
(e.g., consider the skill of “driving in a circle”).
By contrast, option-based HRL
can match the optimality of a flat policy,
and can have options with richer termination conditions.
See \cref{tab:options-vs-feudal} for a further comparison.

\begin{table}[h!]
\centering
\begin{tabular}{|l|p{6cm}|p{6cm}|}
\hline
\textbf{Feature} & \textbf{Options Framework} & \textbf{Feudal Reinforcement Learning} \\
\hline
Hierarchical Structure & Flat controller + multiple sub-policies (options) & Manager $\rightarrow$ Worker (two-level policy hierarchy) \\
\hline
Control Flow & Top-level policy chooses discrete option & Manager emits subgoals; worker acts toward them \\
\hline
Subpolicy Input & Option policies take full state or goal & Worker is conditioned on a goal from the manager \\
\hline
Termination Handling & Explicit termination function $\beta(s)$ & Often fixed horizon or implicit (e.g., $N$ steps) \\
\hline
Initiation Sets & Explicit $\mathcal{I}_o \subseteq \mathcal{S}$ & Often ignored or assumed full initiation \\
\hline
Goal Communication & Top policy selects an option (index) & Manager gives a vectorial subgoal (e.g., in state space) \\
\hline
Training Paradigm & Semi-MDP; option-critic and variants & Two-agent structure; manager learns to guide worker \\
\hline
Interpretability & Each option has a nameable skill & Subgoals may be more abstract (e.g., latent vectors) \\
\hline
\hline
\end{tabular}
\caption{Comparison between the Options Framework and Feudal Reinforcement Learning}
\label{tab:options-vs-feudal}
\end{table}

\subsubsection{Feudal Q learning}

In \citep{Kulkarni2016}, they tackle the feudal learning problem
using two layers of DQN networks.
The  manager  has to choose from a finite number of
predefined subgoals,
and is trained to maximize the task reward.
The worker  receives the goal, and has to choose from a finite
number of discrete primitive actions,
and is trained to maximize the (possibly sparse) intrinsic reward,
based on reaching the goal.

\subsubsection{Dealing with nonstationarity using hindsight relabeling
(HIRO, HAC)}
\label{sec:HAC}
\label{sec:HIRO}

A major difficulty in HRL, which arises when training multiple
levels of policies simultaneously, is that the resulting data
distribution is \keywordDef{non-stationary}.
To see this, note that the worker generates
a distribution over trajectories
in response to a (state, subgoal) pair that,
but this distribution changes as the worker's policy improves.
Thus the data that the manager sees is also coming
from a non-stationary distribution, making it hard to disentangle
any changes due to the managers choice of actions
from changes due to the worker's policy changing.

In \citep{Nachum2018}
they propose \keywordDef{HIRO}
(Hierarchical Reinforcement Learning with Off-policy Correction),
which is a way to tackle this non-stationarity by using
hindsight relabeling (see \cref{sec:HER}).
The data for the manager are transition tuples
of the form $(s_t, g_t, \sum r_{t:t+c}, s_{t+c})$,
where $c$ is the time taken for the worker to
reach the goal (or some maximum time),
and $r_t$ is the main task reward function at step $t$.
The data for the worker are transition tuples of the form
$(s_{t+i}, g_t, a_{t+i}, r_{t+i}^{g_t}, s_{t+i+1})$
for $i=0:c$, where $r_t^{g}$ is the reward
wrt reaching goal $g$.
This data can be used to train the two policies.
However, if the worker fails to achieve the goal
in the given time limit,
all the rewards will be 0, and no learning will take place.
To combat this,
if the worker does not achieve $g_t$
after $c$ timesteps, the subgoal is relabeled
in the transition data with another subgoal $g_t'$
which is sampled from $p(g|\tau)$,
where $\tau$ is the observed trajectory.
Thus both policies treat $g_t'$ as the goal in hindsight,
so they can use the actually collected data for training

The \keywordDef{hierarchical actor critic} (HAC) method
of \citep{Levy2018} is a simpler version of HIRO that can be extended
to multiple levels of hierarchy,
where the lowest level corresponds to primitive actions
(see \cref{fig:HRL}).
In the HAC approach, the output subgoal in the higher level data,
and the input subgoal in the lower-level data,
are replaced with the actual state that was achieved
in hindsight. This allows
the training of  each level of the
hierarchy independently of the lower levels,
by assuming  the lower level policies are already optimal
(since they achieved the specified goal).
As a result, the distribution of
$(s,a,s')$ tuples experienced by a higher level
will be stable,  providing a  stationary learning target.


\subsubsection{Learning the goal space and policy}

In the previous approaches, the subgoals are defined
in terms of the states that were achieved at the end
of each trajectory, $g'=s_T$. This can be generalized
by using a state abstraction function to get
$g'=\phi(s_T)$.
The methods in \cref{sec:HIRO}
assumed that $\phi$ was manually
specified.
We now mention some ways to learn $\phi$.

In \citep{Vezhnevets2017}, they present
\keywordDef{Feudal Networks} for learning a two level hierarchy.
The manager samples subgoals in a learned latent subgoal space.
The worker uses distance to this subgoal as a reward,
and is trained in the usual way.
The manager uses the ``transition gradient'' as a reward,
which is derived from the task reward as well as the
distance between the subgoal and the actual state transition
made by the worker. This reward signal is used to learn
the manager policy and the latent subgoal space.

Feudal networks do not guarantee that the learned subgoal
space will result in optimal behavior.
In \citep{Nachum2019}, they present
a method to optimize the policy and $\phi$ function
so as to minimize a bound on the suboptimality of the hierarchical
policy.
This approach is combined with HIRO
(\cref{sec:HIRO}) to tackle the non-stationarity issue.

\subsection{Subtask discovery}
\label{sec:subtask}

In this section, we discuss ways of learning hierarchical
structure in an environment (given a dataset),
independent of the specific task that needs to be solved.
Thus this can be thought of as a ``pre-training'' phase.
The resulting subtasks can then be used to define a hierarchal
structure, for which a hierarchical policy can be trained,
using the methods described above.

\subsubsection{Discovery of subgoals}

One way to define subtasks is in terms of subgoals that might be worth
achieving.
These are often chosen to be \keywordDef{bottleneck states},
through which many paths (in state space) must pass
when going from different starting states to different goal states.
These states are often identified
by performing various graph-theoretic analyses of the graph
$G$ derived from the 
state transition diagram
by defining an edge $E_{s,s'}$ iff $\sum_a p(s'|s,a)>0$.
For example, \citep{McGovern2001} proposes the \keywordDef{diverse density} metric,
and \citep{Simsek2008} proposes a metric on betweenness centrality.
(This quantifies how important a node is in a network by counting how
many times it appears on the shortest path between other nodes.)

\label{sec:goalsASP}
A notable drawback of existing subgoal discovery methods is their
reliance on a discrete subgoal space. This discreteness prevents the
interpolation of new subgoals, which in turn restricts the variety of
tasks the agent can learn. To overcome this, the development of
continuous subgoal spaces is crucial. 
To tackle this,
\citep{Sukhbaatar2018} introduced \keywordDef{Hierarchical Self Play} (HSP), a
method for learning a continuous embedding of subgoals through an
unsupervised pre-training technique called \keywordDef{asymmetric self-play}
(see \cref{sec:selfPlay}).
This process begins by initializing two reinforcement learning
policies, designated as Alice ($\pi_A$) and Bob ($\pi_B$),
also called a \keywordDef{setter-solver} pair.
The asymmetric self-play unfolds as follows:
\begin{enumerate}
    \item Starting from an initial state $s_0$, Alice's policy $\pi_A$
      acts for $T_A$ steps, arriving at a final state denoted as $s^*
      = s_A^{T_A}$. 
    \item The environment is then reset to $s_0$, and this state $s^*$
      is assigned as a target for Bob's policy, $\pi_B$. 
    \item As Bob executes its policy, a learned encoder, E, generates
      a low-dimensional subgoal embedding at each timestep $t$: $g_t =
      E(s_t^B, s^*)$, where $s_t^B$ is Bob's state at that time. 
    \item Bob's policy then selects actions based on its current state
      and this subgoal embedding: $a_t^B = \pi_B(s_t^B, g_t)$. 
    \item Bob has $T_B$ steps (where $T_B$ is typically close to
      $T_A$) to reach the target state $s^*$. A reward of $R_B = 1$ is
      given for success and 0 for failure. 
    \item Alice receives an opposing reward, $R_A = 1 - R_B$.
\end{enumerate}
This setup creates a dynamic where Bob learns to reach goals set by
Alice, while Alice is incentivized to discover novel states that are
currently challenging for Bob. This encourages effective, unsupervised
exploration of the environment. 

The encoder, $E(s_t^B, s^*)$, effectively maps various target states
into a continuous, low-dimensional subgoal space. Following this
pre-training phase, this learned space is utilized as the continuous
action space for the high-level policy for an HRL agent.
The low-level policy of this agent is
initialized with Bob's learned policy and is subsequently fine-tuned
for specific downstream tasks. 

In \citep{OpenAI2020}, researchers at OpenAI used this method
to train a robot by learning its own curriculum.
And in \citep{selfChallenging}, researchers at FAIR
use a related method to train an LLM to perform tasks.

Note that this kind of goal generation is related to goal-based
intrinsic motivation (see \cref{sec:intrinsicGoals}).

\subsubsection{Discovery of skills}
\label{sec:skills}

Approaches centered on subgoal discovery are inadequate for
identifying subtasks that do not have a specific, concrete
objective. For instance, a subtask like "navigating through traffic",
within the broader goal of reaching a location, involves complex
maneuvers but lacks a distinct subgoal state, making it unsuitable for
subgoal discovery techniques.

In this section, we focus on methods for learning  a varied collection of
\keywordDef{skills},
where we define a skill to be a policy
for subtask, encapsulating the agent's ability to perform a certain
action. Once discovered, these skills can be integrated as the
lower-level policies within an HRL agent
and subsequently fine-tuned for a given task. 

A prominent strategy for discovering diverse skills involves
maximizing the Mutual Information (MI) between a given skill,
$\omega$, and the resulting states or trajectories produced when that
skill is executed. This is typically implemented by conditioning a
universal policy on a latent skill vector, $z_\omega$, creating a
skill-specific policy $\pi_\omega(s) = \pi(s, z_\omega)$. In many such
methods, $z_\omega$ is a one-hot vector. By maximizing
the MI between the skill vector $z_\omega$ and the outcomes of its
corresponding policy $\pi(s, z_\omega)$, the agent is encouraged to
learn a set of distinct skills that lead to different states or
trajectories. 

For example, the \keywordDef{Variational Intrinsic Control}  (VIC)
method of \citep{Gregor2017}
focuses on maximizing the mutual information
      between the skill vector $z_\omega$ and the terminal state
      $s_T$, given an initial state $s_0$.
  The \keywordDef{Diversity Is All You Need} (DIAYN)
  method of \citep{Eysenbach2019diversity}
  aims to discover skills by maximizing the
      mutual information between the skill vector $z_\omega$ and every
      state visited within the trajectory generated by $\pi(s,
      z_\omega)$, without considering the order of the states.
      (This approach is also called \keywordDef{Mutual Information Skill Learning}
      or \keywordDef{MISL} \citep{Zheng2025MISL}.)

The  \keywordDef{VALOR}
(Variational Autoencoding Learning of Options by
Reinforcement) method of \citep{Achiam2018}
uses a VAE-like method to learn skills from trajectories,
respecting the order of the states.
The \keywordDef{SeCTAR} 
(Self-Consistent Trajectory Autoencoder) method of
\citep{Co-Reyes2018} also uses an autoencoder, but based
on an LSTM encoder and decoder.
They combine this with an exploration mechanism to generate
diverse trajectories, so that 
the continuous latent space represents
a diverse set of skills.


\eat{
All these approaches can discover a set of diverse skills but also
share few common limitations, such as, they can only discover a
predefined number of skills, and their MI objectives either ignore
trajectories or do not consider the order of states in the
trajectories. To address these limitations,
\citep{Achiam2018}
proposed
\keywordDef{VALOR}
(Variational Autoencoding Learning of Options by
  Reinforcement).
 In this approach, a policy $\pi(s, z_\omega)$ is basically
an encoder that encodes $z_\omega$ into a trajectory. The trajectory
is denoted as $\tau$. It is fed into a decoder that must map it back
to $z_\omega$. This leads to an optimization objective aimed at
maximizing $P(z_\omega|\tau)$. Thus, VALOR discovers a set of diverse
skills that essentially correlate to a diversity of trajectories taken
by an agent. VALOR also provides a procedure to discover a
progressively increasing number of skills.

The ISD approaches discussed so far can only discover a discrete set
of skills. A possible reason may be that computing the probability
distributions over a continuous space of skills, for estimating the
MI, is complicated. However, discovering a continuous space of skills
is important for generalization, such that new skills can be easily
interpolated in such a continuous space. 

Co-Reyes et al. [77] proposed an approach called
\textbf{Self-Consistent Trajectory Autoencoder (SeCTAR)}, in which an
encoder LSTM [40] embeds the state transition trajectories (i.e.,
$\{s_t, s_{t+1}, s_{t+2}, \dots\}$) into a low-dimensional continuous
latent vector space and a decoder LSTM learns to decode a latent
vector into a policy. A latent vector represents similar trajectories;
hence, the policy decoded from the latent vector is considered to
represent a skill. A diverse latent space is learned by using an
exploration mechanism that generates diverse trajectories. This
encoder-decoder model is learned in a pre-training phase for skill
discovery. After the pre-training phase, the encoder module is removed
and the encoded latent space is available as a continuous space of
skills. A higher-level policy of an HRL agent samples a skill in this
continuous space, which is then decoded by the decoder module in the
form of a policy. 

In addition to independent skill discovery, the authors also extend
SeCTAR for unified learning on a particular task. This is achieved via
an explorer policy that explores the task-relevant states and their
neighborhood such that the trajectories encoded by the SeCTAR encoder
module are correlated to the given task. This way, the discovered
space of skills is situated in the task on which the HRL agent is
being trained concurrently.

Hausman et al. [38] proposed a variational inference approach to learn
diverse skills along with a continuous latent embedding of the
skills. The objective used for learning is the minimization of the
cross-entropy 10 between a distribution over the latent embedding
conditioned on a one-hot skill vector and a distribution over the
latent embedding conditioned on the trajectory generated by executing
that skill. The learned latent embedding is the continuous space of
skills in which new skills can be interpolated. 
}

\section{Imitation learning}
\label{sec:imitation}

In previous sections, an RL agent is
to learn an optimal sequential decision making policy
so that the total reward is maximized.
\keywordDef{Imitation learning} (IL)
(see e.g., \citep{Chrysomallis2025}),
also known as \keywordDef{apprenticeship learning}
and \keywordDef{learning from demonstration} (LfD),
is a different setting,
in which the agent does not observe rewards,
but has access to a collection $\dataExp$ of trajectories
generated by an expert policy $\policyexp$;
that is, $\traj = (s_0,a_0,s_1,a_1,\ldots,s_T)$
and $a_t \sim \policyexp(s_t)$
for $\traj\in\dataExp$.
The goal is to learn a good policy by imitating the expert,
in the absence of reward signals.
IL finds many applications in scenarios
where we have demonstrations of experts (often humans)
but designing a good reward function is not easy,
such as car driving and conversational systems.
(See also \cref{sec:offlineRL}, where we discuss the closely
related topic of offline RL, where we also learn from a collection
of trajectories, but no longer assume they are generated by an optimal
policy.)

\subsection{Imitation learning by behavior cloning}
\label{sec:behavior-cloning}
\label{sec:BC}

A natural method is \keywordDef{behavior cloning},
which reduces IL to supervised learning;
see \citep{Pomerleau89} for an early application
to autonomous driving.
It interprets a policy as a classifier that
maps states (inputs) to actions (labels),
and finds a policy
by minimizing the imitation error, such as
\begin{align}
\label{eqn:rl-inference-bc}
\min_{\policy}
\expectQ{\KLpq{\policyexp(s)}{\policy(s)}}{\statdistpolexp(s)}
\end{align}
where the expectation wrt $\statdistpolexp$ may be approximated by averaging over states in $\dataExp$.
A challenge with this method is that the loss
does not consider the sequential nature of IL:
future state distribution is not fixed
but instead depends on earlier actions.
Therefore, if we learn a policy $\hat{\policy}$ that
has a low imitation error under distribution
$\statdistpolexp$,
as defined in \cref{eqn:rl-inference-bc},
it may still incur a large error
under distribution $\statdist_{\hat{\policy}}$
(when the policy $\hat{\policy}$ is actually run).
This problem has been tackled by the offline RL
literature, which we discuss in \cref{sec:offlineRL}.

\eat{
Further expert demonstrations
or algorithmic augmentations are often
needed to handle the distribution mismatch
(see e.g., \citep{Daume09,Ross11}).
}

\subsection{Imitation learning by inverse reinforcement learning}
\label{sec:IRL}

An effective approach to IL is 
\keywordDef{inverse reinforcement learning} (IRL)
or \keywordDef{inverse optimal control} (IOC).
Here, we first infer a reward function
that ``explains'' the observed expert trajectories,
and then compute a (near-)optimal policy
against this learned reward using any standard
RL algorithms studied in earlier sections.
The key step of reward learning 
(from expert trajectories)
is the opposite of standard RL,
thus called inverse RL~\citep{Ng00irl}.

It is clear that there are infinitely many reward functions for which the expert policy is optimal,
for example by several optimality-preserving transformations~\citep{Ng99}.
To address this challenge,
we can follow the maximum entropy principle,
and use an energy-based probability model
to capture how expert trajectories are
generated~\citep{Ziebart08}:
\begin{align}
p(\traj) \propto 
\exp\big(\sum_{t=0}^{T-1} R_{\vtheta}(s_t,a_t)\big)
\end{align}
where $R_{\vtheta}$ is an 
unknown reward function with parameter $\vtheta$.
Abusing notation slightly, we denote by
$R_{\vtheta}(\traj) = \sum_{t=0}^{T-1} R_{\vtheta}(s_t,a_t))$
the cumulative reward along the trajectory $\traj$.
This model assigns exponentially small probabilities
to trajectories with lower cumulative rewards.
The partition function,
$Z_{\vtheta} \defeq \int_{\traj}\exp(R_{\vtheta}(\traj))$,
is in general intractable to compute,
and must be approximated.
Here, we can take a sample-based approach.
Let $\dataExp$ and $\data$ be the sets of
trajectories generated by an expert,
and by some known distribution $q$, respectively.
We may infer $\vtheta$ by maximizing the likelihood,
$p(\dataExp|\vtheta)$,
or equivalently,
minimizing the negative log-likelihood loss
\begin{align}
\loss(\vtheta) &= -\frac{1}{|\dataExp|}\sum_{\traj\in\dataExp} R_{\vtheta}(\traj) + \log \frac{1}{|\data|}\sum_{\traj\in\data} \frac{\exp(R_{\vtheta}(\traj))}{q(\traj)}
\end{align}
The term inside the $\log$ of the loss is an
importance sampling estimate of $Z$
that is unbiased as long as $q(\traj)>0$ for all $\traj$.
However, in order to reduce the variance,
we can choose $q$ adaptively
as $\vtheta$ is being updated.
The optimal sampling distribution,
$q_*(\traj) \propto \exp(R_{\vtheta}(\traj))$,
is hard to obtain.  Instead, we may find
a policy $\hat{\policy}$
which induces a distribution that is close to $q_*$,
for instance, using methods of maximum entropy RL discussed in \cref{sec:maxentRL}.
Interestingly, the process above produces
the inferred reward $R_{\vtheta}$
as well as an approximate optimal policy $\hat{\policy}$.
This approach is used by \keywordDef{guided cost learning}~\citep{Finn2018GCL},
and found effective in robotics applications.

\subsection{Imitation learning by divergence minimization}
\label{sec:ildm}

We now discuss a different, but related, approach to IL.
Recall that the reward function depends only on
the state and action in an MDP.
It implies that if we can find a policy $\policy$,
so that $\statdistpol(s,a)$ and $\statdistpolexp(s,a)$ are close,
then $\policy$ receives similar long-term reward
as $\policyexp$,
and is a good imitation of $\policyexp$ in this regard.
A number of IL algorithms find $\policy$
by minimizing the divergence between
$\statdistpol$ and $\statdistpolexp$.
We will largely follow the exposition of \citep{Ghasemipour19};
see \citep{Ke19} for a similar derivation.

Let $f$ be a convex function,
and $\fdiv$ be the corresponding $f$-divergence
\citep{Morimoto1963,Ali1966,Csiszar1967,liese2006divergences,csiszar2004information}.
From the above intuition, we want to minimize
$\fdivpq{\statdistpolexp}{\statdistpol}$.
Then, using a variational approximation of
$\fdiv$~\citep{Nguyen2010},
we can solve the following optimization problem for $\policy$:
\begin{align}
\label{eqn:rl-inference-ildm}
\min_{\policy} \max_{\vw}
\expectQ{T_{\vw}(s,a)}{\statdistpolexp(s,a)}
- \expectQ{f^*(T_{\vw}(s,a))}{\statdistpol(s,a)}
\end{align}
where $f^*$ is the convex conjugate of $f$,
and $T_{\vw}:\calS\times\calA\to\real$
is some function parameterized by $\vw$.
We can think of $\pi$ as a generator (of actions)
and $T_{\vw}$ as an adversarial critic
that is used to compare the generated $(s,a)$
pairs to the real ones.
Thus
the first expectation can be estimated using
$\dataExp$, as in behavior cloning,
and the second can be estimated using trajectories
generated by policy $\policy$.
Furthermore, to implement this algorithm, we often
use a parametric policy representation $\polapprox$,
and then perform stochastic gradient updates to
find a saddle-point to \cref{eqn:rl-inference-ildm}.
With different choices of the convex function $f$,
we can obtain many existing IL algorithms,
such as 
\keywordDef{generative adversarial imitation learning}
(\keywordDef{GAIL})~\citep{Ho16}
and
\keywordDef{adversarial inverse RL}
(\keywordDef{AIRL})~\citep{Fu18},
etc.

\eat{
Finally, the algorithms above typically require
running the learned policy $\policy$ 
to approximate the second expectation
in \cref{eqn:rl-inference-ildm}.
In risk- or cost-sensitive scenarios,
collecting more data is not always possible,
Instead, we are in the off-policy IL setting,
working with trajectories collected by some policy
other than $\policy$.
Hence, we need to correct the mismatch
between $\statdistpol$ and
the off-policy trajectory distribution,
for which techniques from \cref{sec:offpolicyRL}
can be used.
An example is
\keywordDef{ValueDICE}~\citep{Kostrikov2020},
which uses a similar distribution correction
method of DualDICE~(\cref{sec:offpolicyrl-dice}).
}

\eat{
By the idea of matching (cite Abbeel-Ng)
f-MAX (4)
JS-div leads to GAIL
ReverseKL leads to AIRL
GAIL is more general?
Connection to DICE
\citep{Ghasemipour19,Abbeel04,Ho16}
\citep{Kostrikov2020}

Mention other imitation learning (apprenticeship learning, IOC) methods:
behavior cloning, by GD (Neu), by LP (Bowling)
}

\section{Offline RL}
\label{sec:offlineRL}

\begin{figure}
\centering
\includegraphics[height=1.5in]{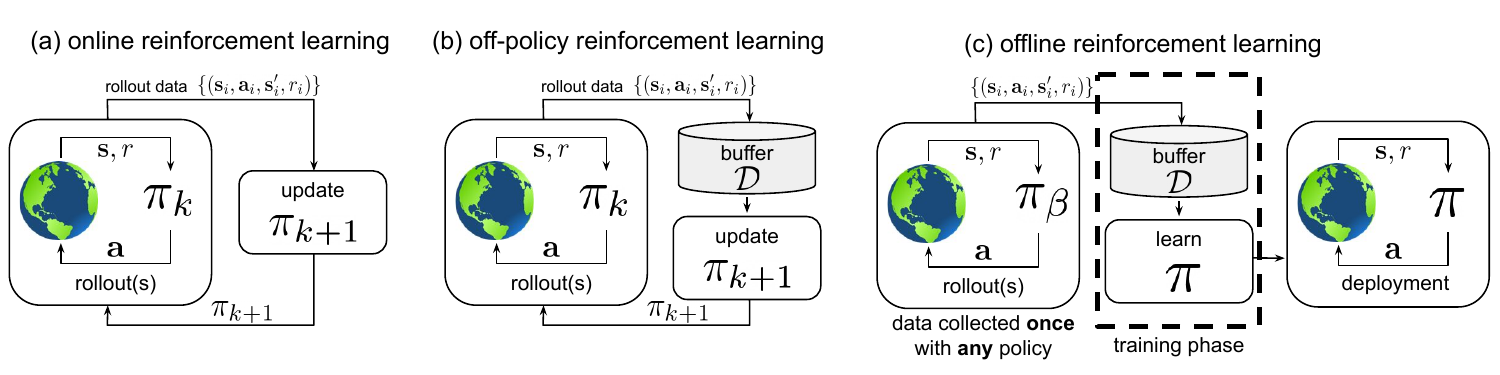}
\caption{
  Comparison of online on-policy RL,  online off-policy RL,
  and offline RL.
\figtaken{Figure 1 of \citep{Levine2020offline}}.
\figthanks{Sergey Levine}.
}
\label{fig:offline}
\end{figure}

\keywordDef{Offline reinforcement learning}
(also called \keywordDef{batch reinforcement learning} \cite{Lange2012})
is concerned with learning a reward maximizing
policy from a fixed, static dataset,
collected by some existing policy,
known as the \keywordDef{behavior policy}.
Thus no interaction with the environment is allowed
(see \cref{fig:offline}).
This makes policy learning harder than the online case,
since we do not know the consequences of actions that were not taken
in a given state, and cannot test any such ``counterfactual'' predictions
by trying them.
(This is the same problem as in off-policy RL, which we discussed in
\cref{sec:offpolicy}.)
In addition, the policy will be deployed on new states that it may not have seen,
requiring that the policy generalize out-of-distribution,
which is the main bottleneck for current offline RL methods \citep{Park2024value}.

A very simple and widely used offline RL method
is known as \keyword{behavior cloning} or \keyword{BC}.
This amounts to training a policy to predict the observed
output action $a_t$ associated with each observed state $s_t$,
so we aim to ensure $\pi(s_t) \approx a_t$,
as in supervised learning.
This assumes the offline dataset was created by an expert,
and so falls under the umbrella of imitation learning
(see \cref{sec:BC} for details).
By contrast, offline RL methods can leverage suboptimal data.
We give a brief summary of some of these  methods below.
For more details, 
see e.g., \citep{Levine20Offline,Chen2024offlineRL,Cetin2024,Yin2025RL,Jackson2025}
and the list of papers at
\url{https://github.com/hanjuku-kaso/awesome-offline-rl}.
For some offline RL benchmarks,
see DR4L \citep{Fu20}, RL Unplugged \citep{Gulcehre20},
OGBench (Offline Goal-Conditioned benchmark) \citep{Park2024},
and D5RL \citep{Rafailov2024}.



\subsection{Offline model-free RL}

In principle, we can tackle offline RL using the off-policy methods
that we discussed in \cref{sec:offpolicy}.
These use some form of importance sampling, based on
$\policytgt(a|s)/\behavior(a|s)$, to reweight the data in the replay buffer $\data$,
which was collected by 
the behavior policy, towards the current policy (the one  being evaluated/ learned).
Unfortunately, such methods only work well if the behavior policy is
is close to the new policy. In the online RL case,
this can be ensured by gradually updating the new policy away from
the behavior policy, and then sampling new data  
from the updated policy (which becomes the new behavior policy).
Unfortunately, this is not an option in the offline case.
Thus we need to use other strategies to control the discrepancy
between the behavior policy and learned policy, as we discuss below.
(Besides the algorithmic techniques we discuss,
another reliable way to get better offline RL performance
is to train on larger, more diverse datasets,
as shown in \citep{Kumar2023offline}.)

\eat{
Various methods have been developed to control the variance
of the importance weights in the offline case,
as discussed in \citep[Sec 4]{Levine20Offline},
but such methods are not widely used in practice.
}

\subsubsection{Policy constraint methods}

In the \keywordDef{policy constraint} method,
we use a modified form of  actor-critic,  which, at iteration $k$,
uses an update of the form
\begin{align}
  Q^{\pi}_{k+1} &\leftarrow \argmin_Q \expectQ{
    \left(Q(s,a) - (R(s,a) + \gamma \expectQ{Q^{\pi}_k(s',a')}{\pi_k(a'|s')})
    \right)^2
  }{(s,a,s') \sim \data} \\
  \pi_{k+1} &\leftarrow \argmax_{\pi} \expectQ{
    \expectQ{Q^{\pi}_{k+1}(s,a)}{\pi(a|s)}}{s \sim \data}
  \myst D(\policy, \behavior) \leq \epsilon
  \label{eqn:piUpdate}
\end{align}
where $D(\policy(\cdot|s), \behavior(\cdot|s))$ is a divergence measure
on distributions, such as KL divergence or another $f$-divergence.
This ensures that we do not try to evaluate the $Q$ function
on actions $a'$ that are too dissimilar from those seen
in the data buffer (for each sampled state $s$),
which might otherwise result in artefacts similar
to an  adversarial attack.

As an alternative to adding a constraint,
we can add a penalty of
$\alpha D(\policy(\cdot|s), \behavior(\cdot|s))$
to the target $Q$ value and the actor objective,
resulting in the following update:
\begin{align}
  Q^{\pi}_{k+1} &\leftarrow \argmin_Q \expectQ{
    \left(Q(s,a) - (R(s,a) + \gamma \expectQ{Q^{\pi}_k(s',a')
    -\alpha  D(\pi_k(\cdot|s'), \behavior(\cdot|s'))}{\pi_k(a'|s')})
    \right)^2
  }{(s,a,s') \sim \data} \\
  \pi_{k+1} &\leftarrow \argmax_{\pi} \expectQ{
    \expectQ{Q^{\pi}_{k+1}(s,a)}{\pi(a|s)}
  - \alpha D(\pi(\cdot|s'), \behavior(\cdot|s'))}{s \sim \data}
\end{align}

One problem with the above method is that we have to fit
a parametric model to $\behavior(a|s)$ in order to evaluate the
divergence term. Fortunately, in the case of KL, the divergence
can be enforced implicitly,
as in the \keywordDef{advantage weighted regression}
or \keywordDef{AWR} method of \citep{Peng2019awr},
the \keywordDef{reward weighted regression} method
of \citep{Peters07},
 the \keywordDef{advantage weighted actor critic}
or \keywordDef{AWAC} method of \citep{Nair2020},
the \keywordDef{advantage weighted behavior model}
or \keywordDef{ABM} method of \citep{Siegel2020},
In this approach, we first solve (nonparametrically) for the new policy
under the KL divergence constraint to
get $\overline{\pi}_{k+1}$, and then we project
this into the required policy function class via
supervised regression, as follows:
\begin{align}
  \overline{\pi}_{k+1}(a|s) &\leftarrow \frac{1}{Z}
  \behavior(a|s) \exp\left( \frac{1}{\alpha} Q_k^{\pi}(s,a) \right) \\
  \pi_{k+1} &\leftarrow \argmin_{\pi} \KLpq{\overline{\pi}_{k+1}}{\pi}
\end{align}
In practice the first step can be implemented by weighting
samples from $\behavior(a|s)$ (i.e., from the data buffer)
using importance weights given by
$\exp\left( \frac{1}{\alpha} Q_k^{\pi}(s,a) \right)$,
and the second step can be implemented via  supervised
learning (i.e., maximum likelihood estimation) using
these weights.

It is also possible to replace the KL divergence
with an integral probability metric (IPM),
such as the maximum mean discrepancy (MMD) distance,
which can be computed from samples,
without needing to fit a distribution $\behavior(a|s)$.
This approach is used in 
\citep{Kumar2019off}.
This has the advantage that it can constrain the support of the learned
policy to be a subset of the behavior policy,
rather than just remaining close to it.
To see why this can be advantageous, consider the case where
the behavior policy is uniform.
In this case, constraining the learned policy to remain close (in KL divergence)
to this distribution could result in suboptimal behavior,
since the optimal policy may just want to put all its mass on a single action
(for each state).

\subsubsection{Behavior-constrained policy gradient methods}

Recently a class of methods has been developed that is simple and effective:
we first learn a baseline policy $\pi(a|s)$ (using BC) and a Q function
(using Bellman minimization) on the offline data,
and then update the policy parameters to pick actions that have high expected
value according to $Q$ and which are also likely under the BC prior.
An early example of this is the $Q^{\dagger}$ algorithm
of \citep{Fujimoto2019batch}.
In  \citep{Fujimoto2021}, they present
 the \keywordDef{DDPG+BC} method,
which optimizes
\be
\max_{\pi} J(\pi) = \expectQ{
  Q(s,\mu^{\pi}(s)) + \alpha \log \pi(a|s)}{(s,a) \sim \data}
\ee
where $\mu^{\pi}(s) = \expectQ{a}{\pi(a|s)}$ is the mean of
the predicted action, and $\alpha$ is a hyper-parameter.
As another example,
the \keywordDef{DQL} method of \citep{Wang2023DQL}
optimizes a diffusion policy using
\be
\min_{\pi} \loss(\pi)
= \loss_{\text{diffusion}}(\pi) + \loss_{q}(\pi)
= \loss_{\text{diffusion}}(\pi) 
- \alpha \expectQ{Q(s,a)}{s \sim  \data,
  a \sim \pi(\cdot|s)}
\ee
where the second term is a penalty derived from Conservative Q Learning
(\cref{sec:CQL}),
that ensures the Q values  do not get too small.
Finally, \citep{Agarwal2022} discusses how to transfer
the policy from a previous agent to a new agent
by combining BC with Q learning.

\subsubsection{Uncertainty penalties}
\label{sec:offlineUncertainty}

An alternative way to avoid picking out-of-distribution actions,
where the $Q$ function might be unreliable,
is to add a penalty term to the $Q$ function based on the estimated
epistemic uncertainty, given the dataset $\data$,
which we denote by $\text{Unc}(P_D(Q^{\pi}))$,
where $P_D(Q^{\pi})$ is the distribution over $Q$ functions,
and $\text{Unc}$ is some metric on distributions.
For example, we can use a deep ensemble to represent the distribution,
and use the variance of $Q(s,a)$ across ensemble members as
a measure of uncertainty.
This gives rise to the following policy improvement update:
\begin{align}
  \pi_{k+1} &\leftarrow \argmax_{\pi} \expectQ{
    \expectQ{
      \expectQ{Q^{\pi}_{k+1}(s,a)}{P_D(Q_{k+1}^{\pi})}
    }{\pi(a|s)}
    - \alpha \text{Unc}(P_D(Q_{k+1}^{\pi}))
}{s \sim \data}
\end{align}
For examples of this approach, see e.g.,
\citep{An2021,Wu2021unc,Ghasemipour2022}.

\subsubsection{Conservative Q-learning}
\label{sec:CQL}

An alternative to explicitly estimating uncertainty
is to add a \keywordDef{conservative penalty}
directly to the $Q$-learning error term.
That is, we minimize the following wrt $\vw$
using each batch of data $\calB$:
\be
\overline{\calE}(\calB,\vw)
   = \alpha \calC(\calB,\vw) + \calE(\calB, \vw)
\ee
where $\calE(\calB, \vw) = \expectQ{ (Q_{\vw}(s,a) -
  (r+\gamma \max_{a'} Q_{\vw}(s',a')))^2}{(s,a,s') \in \calB}$
is the usual loss for $Q$-learning,
and $\calC(\calB,\vw)$ is some conservative penalty.

In the \keywordDef{conservative Q learning}
or \keywordDef{CQL} method of \citep{CQL},
we use the following penalty term:
\be
\calC(\calB,\vw)
= \expectQ{
 \expectQ{Q_{\vw}(s,a)}{a \sim \mu(\cdot|s)}
- \expectQ{Q_{\vw}(s,a)}{a \sim \behavior(\cdot|s)}
}{s \sim \data}
+ R(\mu)
\label{eqn:CQL}
\ee
where $\mu$ is the new policy derived from $Q$,
and $R(\mu)=-\KLpq{\mu}{\rho}$ is a regularizer,
and  $\rho$ is the action prior,
which we discuss below.
Since we are minimizing
$\calC(\calB,\vw)$ (in addition to
$\calE(\calB,\vw)$), we see that we are simultaneously
maximizing the Q values for
actions that are drawn from the behavior policy
while minimizing the $Q$ values for actions
sampled from $\mu$.
This is to combat the optimism bias of Q-learning
(hence the term ``conservative'').

Now we derive the expression for $\mu$.
From \cref{sec:maxentRL} we know that
the optimal
 solution has the form
 $\mu(a|s) = \frac{1}{Z} \rho(a|s) \exp(Q(s,a))$,
 where $Z=\sum_{a'} \exp(Q(s,a'))$ is the normalizer,
 and $\rho(a|s)$ is the prior.
(For example, we can set $\rho(a|s)$ to be the previous policy.)
We can then approximate
the first term in the penalty using importance sampling,
with $\rho(a|s)$ as the proposal:
\be
\expectQ{Q(s,a)}{a \sim \mu(\cdot|s)}
= \expectQ{\frac{\mu(a|s)}{\rho(a|s)} Q(s,a)}{\rho(a|s)}
=  \expectQ{\frac{\exp(Q(s,a))}{\sum_{a'} \exp(Q(s,a'))}
  Q(s,a)}{\rho(a|s)}
\label{eqn:CQLpen}
\ee

Alternatively, suppose  we set $\rho(a|s)$ to be uniform,
as in maxent RL (\cref{sec:maxentRL}).
In this case, we should replace the value function
with the  soft value function.
From \cref{eqn:softValueFn}, using a penalty coefficient of $\alpha=1$,
we have
\be
\expectQ{Q_{\text{soft}}(s,a)}{a}
= V_{\text{soft}}(s)  = \log \sum_a \exp(Q(s,a))
\ee
Note, however, this  can be intractable for high-dimensional actions.

\eat{
A natural approach to offline RL is to just perform
Q-learning on the fixed dataset,
and then to compute the policy using
$\pi(a|s) = \argmax_a Q(s,a)$.
In principle this should work, since Q-learning is off-policy,
but in practice it does not work.
The reason is that Q-learning is trying to learn
the optimal policy $\pi^*$, but the data was generated by
a fixed \keyword{behavior policy}
$\pi_{\beta}$, so when Q-learning predicts the value
of the best action, it cannot be checked against the data,
and it might become over optimistic.

To see this in more detail, let us rewrite
the Bellman backup in terms of a distribution
$\pi_{\tnew}$ that puts all its mass on the optimal action
(according to the current $Q$) for each state,
so $\pi_{\topt}(a|s)=\delta(a-\argmax_{a'} Q(s,a'))$.
Thus the TD loss becomes
\be
J_{\beta}(Q,\pi) = \expectQ{ (Q(s,a)-\targetV(s,a))^2}{\pi_{\beta}(s,a)}
\ee
where the target $\targetV(s,a)$ is defined by 
\be
\targetV(s,a) =
  r(s,a) + \expectQ{Q(s',a')}
  {\pi_{\beta}(s') \pi_{\topt}(a'|s')}
\ee
If we have $\pi_{\beta}(s,a)=\pi_{\beta}(s) \pi_{\topt}(s|a)$, this optimization
should work well, since the training data comes
from the optimal distribution, and thus we have a stable system.
But for suboptimal behavior data, Q-learning will learn
to find actions that do better than the training data,
according to $Q$; this is like finding adversarial examples
for a network, and can result in unstable behavior,
where the estimated $Q$ values increase, but the actual reward
decreases.


One way to deal with the overestimation problem is to use
 \keywordDef{conservative Q learning}
 or \keywordDef{CQL} \citep{CQL}.
The idea is to find $(s,a)$ points where the
$Q$-function might be overestimating the true value,
and then to ``push down'' on the estimate,
so it becomes more conservative.
In practice this amounts to adding the following
regularizer (scaled by $\alpha$) to $J_{\beta}$:
\begin{align}
J_{CQL}(Q,\pi)
&=   \expectQ{ Q(s,a) -\expectQ{Q(s,a')}{a' \sim \pi(\cdot|s)}
}{(s,a) \sim \data} \\
&=  \expectQ{Q(s,a)}{(s,a) \sim \data}
- \expectQ{Q(s',a')}{s' \sim \data, a' \sim \pi(\cdot|s')}
\label{eqn:CQL}
\end{align}

}

 \eat{
the approach attempts to generalize beyond  offline data,
but not too far.
First it uses Q learning to solve
\be
\argmin_{\vw} [Q_{\vw}(s_t,a_t) - \expectQ{r(s_t,a_t) + Q_{\vw}(s_{t+1},a_{t+1})}{\pemp}]^2
\ee
where $\pemp$ is the empirical distribution over trajectories.
It also fits the BC policy $q_{\beta}(a_t|s_t)$.
Finally  it finds the policy that is closest
(in the chosen parametric family)
to
$q(a_t|s_t) \propto \exp(Q_{\vw}(s_t,a_t) q_{\beta}(a_t|s_t))$,
which is the BC policy weigthed  by the Q function.
It does this by solving a weighted MLE problem
\be
\argmin_{\vtheta} \KLpq{\pi_{\vtheta}(a_t|s_t)}{q(a_t|s_t)}
\ee
 }

\subsection{Offline model-based RL}
\label{sec:offlineMBRL}

In \cref{sec:MBRL}, we discussed model-based RL,
which can train a dynamics model given a fixed dataset,
and then use this to generate synthetic data
to evaluate and then optimize  different possible policies.
However, if the model is wrong, the method may learn a suboptimal
policy, as we discussed in \cref{sec:modelUncertainty}.
This problem is particularly severe in the offline RL case,
since we cannot recover from any errors by collecting more data.
Therefore various conservative MBRL algorithms have been developed,
to avoid exploiting model errors.
For example, \citep{Kidambi2020} present the
\keywordDef{MOREL} algorithm,
and \citep{Yu2020mopo} present the \keywordDef{MOPO} algorithm.
Unlike the
value function uncertainty method of \cref{sec:offlineUncertainty},
or the conservative value function method of \cref{sec:CQL},
these model-based methods add a penalty for visiting states where
the model is likely to be incorrect.

In more detail, let $u(s,a)$ be an estimate of the uncertainty of the
model's predictions given input $(s,a)$.
In MOPO, they define a conservative reward using
$\overline{R}(s,a) = R(s,a) - \lambda u(s,a)$,
and in MOREL, they modify the MDP so that the agent enters an
absorbing state with a low reward when $u(s,a)$ is sufficiently
large.
In both cases, it is possible to prove that the model-based
estimate of the policy's performance under the modified
reward or dynamics is a lower bound of the performance
of the policy's true performance in the real MDP,
provided that the uncertainty function $u$ is an error oracle,
which means that is satisfies
$D(M_{\vtheta}(s'|s,a), M^*(s'|s,a)) \leq u(s,a)$,
where $M^*$ is the true dynamics, and $M_{\vtheta}$
is the estimated dynamics.

For more information on offline MBRL methods,
see \citep{Chen2024offline}.

\subsection{Offline RL using reward-conditioned sequence modeling}
\label{sec:seqModel}
\label{sec:DT}

Recently an approach to offline RL
based on sequence modeling has become very popular.
The basic idea --- known as \keywordDef{upside down RL}
\citep{Schmidhuber2019}
or \keywordDef{RvS} (RL via Supervised learning)
\citep{Kumar2019,Emmons2021} ---
is to train a generative model over future states
and/or actions conditioned on the observed reward,
rather than predicting the reward given a state-action
trajectory.
At test time, the conditioning is changed
to represent the desired reward, and futures
are sampled from the model.
The implementation of this idea then depends on what
kind of generative model to use, as we discuss below.

The \keywordDef{trajectory transformer}
method of \citep{trajectoryTransformer}
learns a joint model of the form
$p(\vs_{1:T}, \va_{1:T}, \vr_{1:T})$
using a transformer,
and then samples from this using beam search,
selecting the ones with high reward (similar to MPC, \cref{sec:MPC}).
The \keywordDef{decision transformer} \citep{decisionTransformer}
is related, but  just generates action sequences,
and conditions on the past observations and the future reward-to-go.
That is, it fits
\be
\argmax_{\vtheta} \expectQ{
  \log \pi_{\vtheta}(a_t|s_{0:t}, a_{0:t-1}, \text{RTG}_{0:t})}{\pemp}
\ee
where $\text{RTG}_t = \sum_{k=t}^T r_t$ is the return to go.
This is just like BC policy learning, except we also condition
on the RTG.
At run time, $\RTG_0$ is set to some desired high value
(e.g., the maximum RTG observed during training),
and is then updated online using
$\RTG_{t+1}=\RTG_t - r_t$.
To set more plausible RTG values,
\citep{Lee2022DT} propose to learn the distribution
$p(\RTG_t | s_{\leq t}, a_{\leq t}, \RTG_{<t})$,
which we can view as a critic,
in addition to training the actor
$p(a_t | s_{\leq t}, a_{< t}, \RTG_{\leq t})$.
\citep{Yamagata2023}  propose
the Q-learning Decision Transformer (\keywordDef{QDT}),
which conditions on a Q value
(learned using  Q learning) instead of RTG.
This combines the benefits of dynamic programming methods
(that can ``stitch'' suboptimal trajectories together)
with the stability of supervised learning using by DT.


The \keywordDef{diffuser} method of \citep{diffuser}
is a diffusion version of trajectory transformer,
so it fits $p(\vs_{1:T}, \va_{1:T}, \vr_{1:T})$ using diffusion,
where the action space is assumed to be continuous.
They also replace beam search with classifier guidance.
The \keywordDef{decision diffuser} method of \citep{decisionDiffuser}
extends diffuser by using classifer-free guidance,
where the conditioning signal is the reward-to-go,
simlar to decision transformer.
However, unlike diffuser,
the decision diffuser just models the future state trajectories
(rather than learning a joint distribution over states and actions),
and infers the actions using an \keywordDef{inverse dynamics model}
$a_t = \pi(s_t, s_{t+1})$,
which is trained using supervised learning.

One problem with the above approaches is that conditioning
on a desired return and taking the predicted action can fail dramatically
in stochastic environments,
since trajectories that result in a return may have only
achieved that return due to chance \citep{Paster2022,Yang2023,Brandfonbrener2022,Villaflor2022}.
(This is the same as the optimism bias problem in the control-as-inference
approach discussed in \cref{sec:optimismBias}.)

In \citep{Kong2024}, they propose the
\keywordDef{latent plan transformer},
that replaces conditioning on the reward-to-go with
conditioning on a latent ``plan'', $\vz \in \real^D$.
In more detail, they fit the following latent variable sequence model
using MC-EM, where the E step is implemented using Langevin dynamics:
\be
p(\vz) p(\tau|\vz) p(y|\vz)
\ee
where $\tau=(s_1,a_1,\ldots,s_T,a_T)$ is the state-action trajectory
and $y$ is the observed (trajectory-level) reward.
The model for $p(\tau|\vz)$ is a causal transformer,
which generates the action at each step given the previous
states and actions and the latent plan.
The model for $p(y|\vz)$ is just a Gaussian.
The model for $p(\vz)$ is a Gaussian passed
through a U-net style CNN,
thus providing a richer prior.
The latent variables provide a way to ``stitch together''
individual (high performing) trajectories, so that the learned policy
can predict $p(a_t|s_t,z)$ even if the current state
$s_t$ is not on the training manifold
(thus requiring generalization,
a problem that behavior cloning faces \citep{Ghugare2024}).
During decision time, they infer $\hat{\vz}=\argmax p(\vz|y=y_{\max})$
using gradient ascent,
and then autoregressively generate actions from
$p(a_t|s_{1:t},a_{1:t-1},\hat{\vz})$.

\subsection{Offline-to-online methods}
\label{sec:O2O}

Despite the progress in offline RL, it is fundamentally
more limited in what it can learn compared to online RL
\citep{Ostrovski2021},
because agent cannot explore the consequences of
its own actions
(c.f., results in psychology \citep{Markant2014}).
Therefore, there is a lot of interest in pre-training offline,
and then using online finetuning.
This is called the \keywordDef{offline-to-online} (O2O) paradigm.
Unfortunately,  due to the significant distribution shift between online experiences
and offline data, most offline RL algorithms suffer from performance drops
when they are finetuned online.
Many different methods have been proposed to tackle this,
a few of which we mention below.
See \url{https://github.com/linhlpv/awesome-offline-to-online-RL-papers}
for a more extensive list.


\subsubsection{Calibrated Q learning}

\citep{Nakamoto2023}
suggest pre-training
with CQL followed by online finetuning.
Naively this does not work that well, because CQL can be
too conservative, requiring the online learning
to waste some time at the beginning fixing
the pessimism. So they propose a small
modification to Conservative Q learning,
discussed in \cref{sec:CQL},
known as \keywordDef{Calibrated Q learning}.
This simply prevents CQL from being too conservative,
by replacing the CQL
regularizer
in \cref{eqn:CQL}
with a slightly modified expression.
\eat{
with
\be
\min_Q \max_{\pi}
J(Q,\pi)    + \alpha  \expectQ{
  \max(Q(s,a), V^{\pi_{\beta}}(s))
  - \alpha \expectQ{Q(s,a)}{(s,a) \sim \data}
}{s \sim \data, a \sim \pi(a|s)}
\label{eqn:CalQL}
\ee
where the $Q(s,a)$ term inside the $\max$
ensures conservatism (so $Q$ lower bounds
the value of the learned policy),
and the $V^{\pi_{\beta}}(s)$ term
ensures ``calibration''
(so $Q$ upper bounds the value
of the behavior policy).
}
Then online finetuning is performed in the usual way.

\subsubsection{Dagger}
\label{sec:dagger}
\label{sec:Dagger}


An alternative approach is the \keywordDef{Dagger} algorithm
of  \citep{Ross2011}.
(Dagger is short for Dataset Aggregation.)
This iteratively trains the  policy on expert provided data.
We  start with an initial dataset  $\data$ (e.g., empty)
and an initial policy $\pi_1$ (e.g., random).
At iteration $t$, 
we run the current policy $\pi_t$
in the environment to collect states $\{s_i\}$.
We then ask an expert policy for the correct actions
$a_i^* = \pi^*(s_i)$.
We then aggregate the data to compute
$\data = \data \union \{ (s_i,a_i^*) \}$,
and train the new policy $\pi_{t+1}$ on $\data$.
The key idea is to not 
train passively on expert trajectories as in BC,
but to train on the states that the policy actually visits.
This avoids overfitting to idealized data and improves robustness
(avoids compounding error),
since the policy is learning the effects
of its own causal interventions \citep{Ortega2021shaking}.



\section{General RL, AIXI and universal AGI}
\label{sec:AIXI}

The term ``\keywordDef{general RL}''
(see e.g., \citep{Hutter2005,Lattimore2013,Hutter2024,Majeed2021})
refers to the setup in which an agent
receives a stream of observations $o_1, o_2, \ldots$
and rewards $r_1, r_2, \ldots$,
and performs a sequence of actions in response, $a_1, a_2, \ldots$,
but where we do not make any Markovian (or even stationarity)
assumptions about the environment that generates the observation stream.
Instead, we assume that the environment is a computable function
or program $p^*$, which generated the observations
$o_{1:t}$ and $r_{1:t}$ seen so far
in response to the actions taken, $a_{1:t-1}$.
We denote this by
$U(p^*,\va_{1:t}) = (o_1 r_1 \cdots o_t r_t)$,
where $U$ is a universal Turing machine.
If we use the receeding horizon control strategy (see \cref{sec:MPC}),
the optimal action at each step is the one that maximizes
the posterior expected reward-to-go (out to some horizon $m$
steps into the future).
If we assume the agent represents the unknown environment
as a program $p \in \calM$,
then the optimal action is given by 
the following \keywordDef{expectimax} formula:
\be
a_t = \argmax_{a_t} \sum_{o_t,r_t} \cdots \max_{a_m} \sum_{o_m,r_m}
[r_t + \cdots + r_m]
\sum_{p: U(p,\va_{1:m}) = (o_1 r_1 \cdots o_m r_m)}
\Pr(p)
\ee
where $\Pr(p)$ is the prior probability of $p$,
and we assume the likelihood is 1 if $p$ can generate
the observations given the actions, and is 0 otherwise.

One important question is: what is a reasonable prior over programs?
In \citep{Hutter2005},
Marcus Hutter proposed to apply
the idea of \keywordDef{Solomonoff induction}
\citep{Solomonoff1964}
to the case of an online decision making agent.
This amounts to using the prior $\Pr(p) = 2^{-\ell(p)}$,
where $\ell(p)$ is the length of program $p$.
This prior favors shorter programs, and the likelihood
filters out programs that cannot explain the data.
The resulting agent is known as \keywordDef{AIXI},
where ``AI'' stands for ``Artificial Intelligence''
and ``XI''
referring to the Greek letter $\xi$ used in Solomonoff induction.
The AIXI agent has been called the
``most intelligent general-purpose agent possible''
\citep{Hutter2024},
and can be viewed as the theoretical foundation 
of (universal) \keywordDef{artificial general intelligence}
or \keywordDef{AGI}.

Unfortunately, the AIXI agent is intractable to compute,
for two main reasons:
(1) it relies on Solomonoff induction and Kolmogorov complexity,
both of which are intractable;
and (2)
the expectimax computation is intractable.
Fortunately, various tractable approximations
have been devised.
In lieu of Kolmogorov complexity,
we can use measures like MDL (minimum description length),
and for Solomonoff induction,
we can use various local search or optimization algorithms
through suitable function classes.
For the expectimax computation,
we can use MCTS (see \cref{sec:MCTS}) to approximate it.
Alternatively, 
\citep{Grau-Moya2024} showed that it is possible to use
\keyword{meta learning}
to train a generic sequence predictor,
such as a transformer or LSTM,
on data generated by random Turing machines,
so that the transformer learns to approximate a universal predictor.
Another approach 
is to  learn a policy (to avoid searching over action sequences)
using TD-learning (\cref{sec:TD});
the weighting term in the policy mixture requires that
the agent predict its own future actions,
so this approach is known as \keywordDef{self-AIXI} \citep{Catt2023}.

Note that AIXI is a normative theory for optimal agents, but is not very practical,
since it does not take computational limitations into account.
In \citep{Arumugam2024,Arumugam2024satisficing},
they describe an
 approach which extends the above Bayesian framework,
while also taking into account the data budget (due to limited
environment interactions) that real agents must contend with
(which prohibits modeling the entire environment or finding the optimal action).
This approach,  known as \keywordDef{Capacity-Limited Bayesian RL} (CBRL),
combines Bayesian inference, RL, and rate distortion theory,
and can be seen as a normative theoretical foundation for computationally
bounded rational agents.

\chapter{Acknowledgements}

  Parts of this monograph
  are borrowed from chapters 34 and 35 of my textbook \citep{book2},
  some  of which was written with  Lihong Li.
  However, this text supercedes those chapters,
  and goes beyond it in many ways.
  Thanks to the following people for feedback on 
  the current document:
  Pablo Samuel Castro,
  Elad Hazan,
  Tuan Ahn Le,
  Dieterich Lawson,
  Marc Lanctot,
  David Pfau,
  Theo Weber.
And thanks to Xinghua Lou for help with some of the figures.


\printbibliography 
\end{document}